# Stress-testing large language model agents in a robotic chemistry laboratory

Lulu Guo [†,1], Yingkai Sun[†,1], Xiaobo Li[†,1,2,*], Luyao Ge[1], Ziming Wang[1], Haitao Zheng[1], Jingyu Li[1], Huijuan Zhang[1], Bingxu Chen[1], Daobin Liu[1], Yuebo Liu[2], Jie Li[2], Xiaohui Li[2], Linjiang Chen[1,2,3,*], Yi Luo[1,2,*], Jun Jiang[1,2,*]

[1] State Key Laboratory of Precision and Intelligent Chemistry, Hefei National Research Center for Physical Sciences at the Microscale, School of Chemistry and Materials Science, University of Science and Technology of China, Hefei, China

[2] Center for Scientific Intelligence Innovation, Hefei, China

[3] School of Chemistry, School of Computer Science, University of Birmingham, Birmingham, UK

[†] These authors contributed equally

*Correspondence: xiaoboli@ustc.edu.cn; linjiangchen@ustc.edu.cn; yiluo@ustc.edu.cn; jiangj1@ustc.edu.cn

## Abstract

AI is evaluated through knowledge, reasoning and plan generation, yet scientific agency requires reliable physical action and adaptation to evidence. Here, we use a robotic chemistry laboratory as a physical-world testbed to make scientific agency measurable. Its 45 modular workstations exposed as machine-readable skills enabled 4,608 trials. Only 3.3% of trials produced expert-assessed executable workflows under laboratory constraints; even the best system achieved 28.1%. Long-horizon planning remained a challenge: only three executable workflows exceeded 30 operations, although the longest contained 44. Across five rounds, experimental feedback prompted local adjustments but no workflow-level replanning or analytical-method redesign. By making physical executability and evidence-driven replanning measurable, our study provides an evidence-based assessment of deployment readiness and a diagnostic framework to guide closed-loop improvements towards physically grounded autonomous research.

## Main

End-to-end autonomous research is a central ambition of artificial intelligence for science, but achieving it requires agents that connect scientific reasoning to reliable action in robotic laboratories. This capability extends beyond generating plausible hypotheses and experimental protocols. Agents must produce executable actions that respect constraints imposed by instruments, operating parameters, samples and vessels, while maintaining coherent plans and adapting across successive experimental rounds. AI agents have recently demonstrated increasing autonomy in scientific reasoning, computational analysis and experimental planning (*1–5*). In parallel, robotic laboratories have enabled automated and closed-loop physical experimentation (*6–13*), while recent studies have begun to connect AI agents directly to robotic instruments (*14–18*). Yet, to our knowledge, no systematic, quantitative assessment has established how reliably AI agents can convert scientific intent into executable action in a robotic laboratory (*19, 20*). This challenge is particularly pronounced in modular robotic laboratories that integrate heterogeneous capabilities for open-ended scientific discovery.  Here, we conduct an end-to-end stress test of AI agents powered by large language models in a modular robotic chemistry laboratory, evaluating their ability to plan, execute and adapt experiments. By revealing capability boundaries and failure modes, our study provides an evidence-based assessment of deployment readiness and a diagnostic basis for closed-loop improvements towards physically grounded autonomous research.

**A robotic laboratory testbed**

We built an integrated, general-purpose robotic catalysis laboratory. The laboratory capacities span the full research cycle from sample synthesis and characterization to performance test. It contains 45 modular automated workstations, including liquid handling, solid weighing, centrifugation and drying; electrochemical, photocatalysis and high-temperature/high-pressure reaction stations; and X-ray diffractometer (XRD), ultraviolet-visible (UV-Vis), Fourier-transform infrared (FT-IR) and fluorescence spectroscopy, microplate reading, and liquid and gas chromatography (Fig. 1 and fig. S1A). Standardized container formats, transfer units and robotic-gripper interfaces enable diverse vessels and sample carriers to be handled across workstations. Reactor-specific fixtures accommodate local requirements for positioning, fixation, reaction, heating and characterization within individual workstation modules (fig. S1B). Source-aligned counting summaries are provided in tables S1 to S5, and the complete workstation inventory is provided in table S6 and data S2. The catalogue exposes 45 modular automated workstations, 62 operations, 433 constraints and 456 parameters; a conservative stagewise enumeration yielded approximately $6.71 \times 10^6$ device-combination workflow families (see calculations in note 1 of the supplementary text).

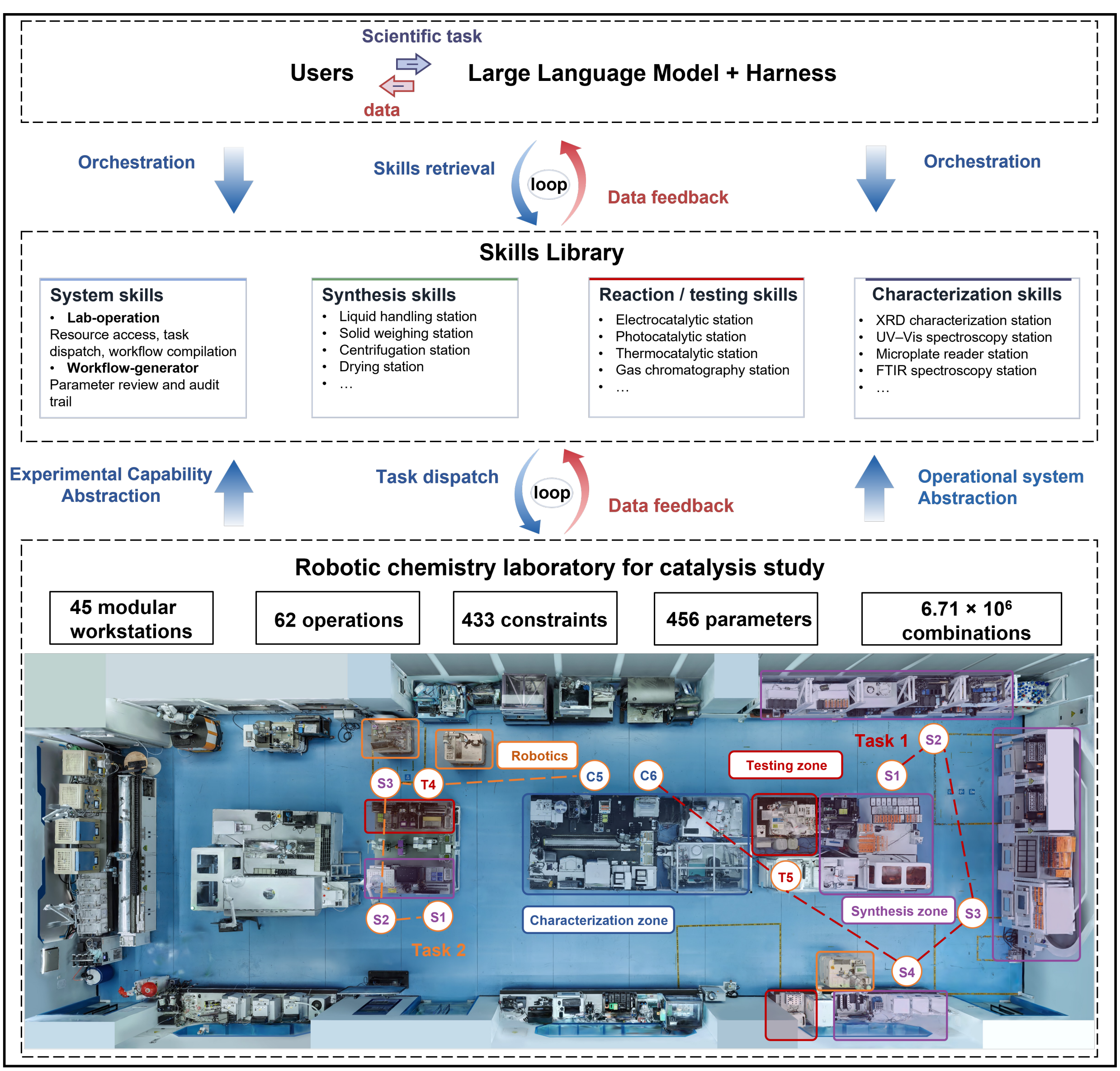

**Fig. 1. Architecture of the agent-readable robotic chemistry laboratory for catalysis study.** Scientific intent and tasks enter through the user interface, whereas results, data and execution state return through controlled pathways. The Skills Layer exposes system, synthesis, reaction, characterization and testing capabilities as machine-readable operations, parameters, input-output states and constraints. Agents retrieve these skills to propose workflows, but the system retains authority over validation, compilation and dispatch. The laboratory integrates synthesis, testing and characterization modular workstations through robotics.

The operation system and capability of robotic laboratory was exposed to agents through agent-readable skills. System skills expose resource access, workflow compilation, dispatch and result retrieval, whereas workstation skills describe synthesis, reaction and testing, and characterization, as detailed in note 2 of the supplementary text. Each workstation skill records supported operations, required parameters, vessel and sample input-output states, and local constraints, with the detailed skill structure described in note 3 of the supplementary text. For example, Liquid_Handling_Station_1ml_V2 separates uncapping, liquid addition and recapping. It accepts at most 16 source solutions, limits a transfer to 1 mL and total addition of one solution to 3 mL, enforces one solution per source bottle and fixed bottle order, and tracks whether the destination must be capped or uncapped (fig. S2 and table S7). Chemically plausible plans that violate these limits or vessel states are rejected. Laboratory-wide rules for authorization, routing, operation order and compiler interfaces are encoded alongside device definitions, allowing the same representation to guide agent planning and pre-compilation validation. The full skill libraries were available (see Materials and Methods).

Together, the architecture layers form three coupled loops: capability abstraction maps equipment into agent-readable skills; skill retrieval supplies planning rules and schemas; and dispatch with data feedback links accepted plans to task identifiers, execution state and results. The laboratory retains execution authority: only workflows matching available resources and passing schema, state and compilation checks enter dispatch.

## From scientific intent to robotic execution

Each evaluated agent comprised a large language model and an agent harness. The harness denotes the software infrastructure that provides the model with its working environment and tools, such as OpenClaw, Codex or Claude Code. We abstracted the pathway from scientific intent to robotic action into ten evidence endpoints organized across three stages (Fig. 2A): experimental design (N1-N7), workflow validation and laboratory dispatch (N8-N9), and the physical-execution boundary (N10). Agents reasoned and acted across N1-N9. N10 was represented by a separate expert assessment of workflow executability rather than by trial-level evidence of started or completed experiments.

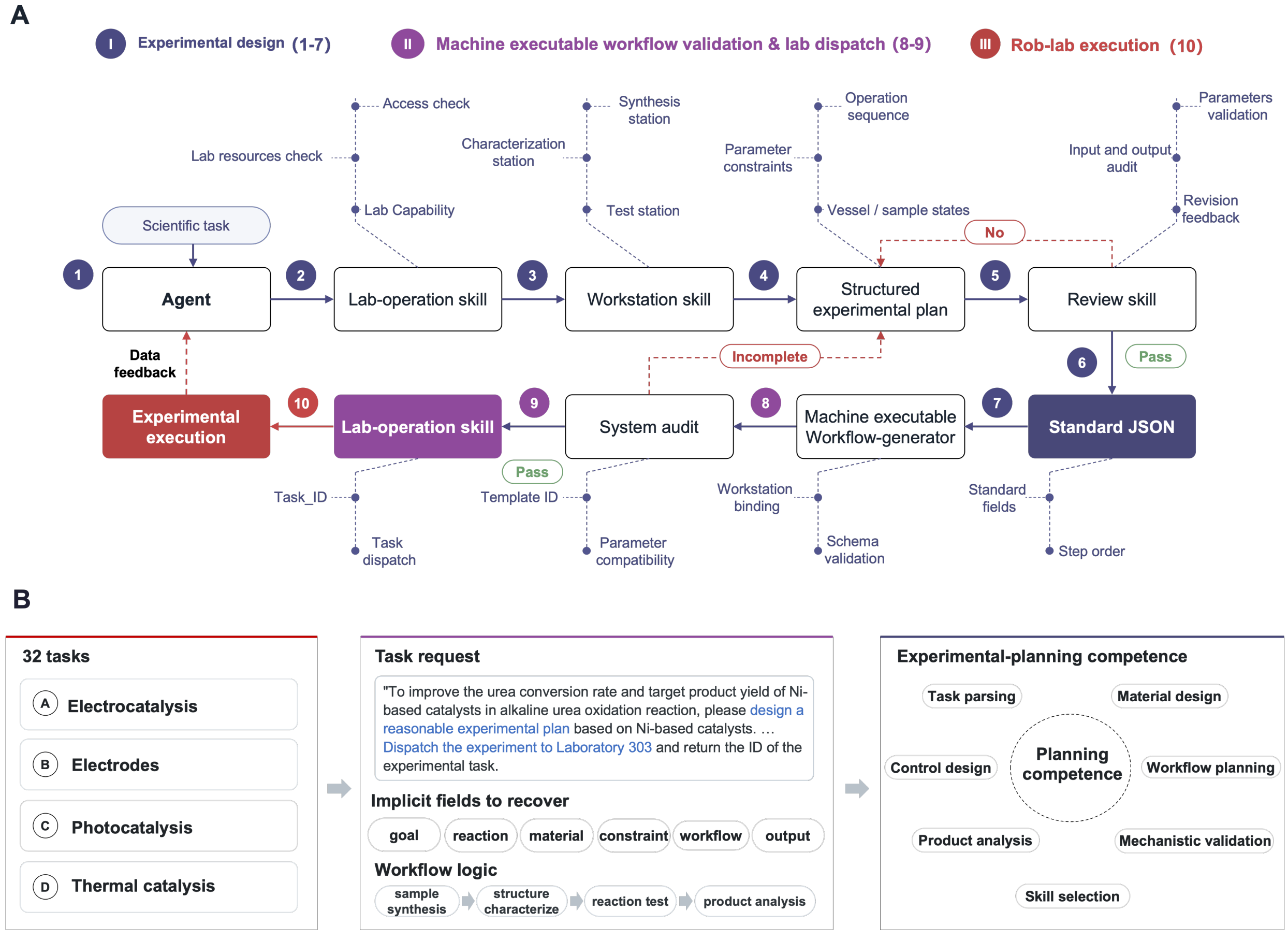

**Fig. 2. From a scientific task to a physically constrained robotic laboratory pathway.** A, Ten-endpoint route from experimental design (N1-N7) through workflow validation and laboratory dispatch (N8-N9) to the physical-execution boundary (N10). Agents operated across N1-N9; N10 was supplied by a separate expert assessment of workflow executability, not by trial-level started or completed experiments. B, The 32 tasks are organized into electrocatalysis, electrode-material and electrochemical testing, photocatalysis and thermal-catalysis domains. Each request requires recovery of an implicit goal, reaction, material, constraint, workflow and output. The urea-oxidation example links sample synthesis, structural characterisation, reaction testing and product analysis, thereby probing task parsing, material and control design, skill selection, workflow planning, product analysis and mechanistic validation.

At node 1, the agent receives a task specifying the scientific objective. At node 2, the lab-operation skill routes the request to an authorized laboratory, verifies access and identifies the required workstation capabilities. At node 3, the agent retrieves and reads the relevant workstation skills. These may include liquid handling for sample preparation, electrochemical testing, and X-ray diffraction or infrared spectroscopy for characterization. At node 4, the agent uses this machine-readable vocabulary to construct a structured experimental plan. The plan specifies the operation sequence, parameter values and constraints, together with the evolving states of vessels and samples. At node 5, review skill checks parameter validity and input-output consistency. Failed reviews return the workflow to node 4 for revision. After approval, node 6 serializes the plan as standard JSON with the required fields and step order. This representation marks the interface between planning and validation, but does not establish that the workflow can be dispatched or executed.

At node 7, the workflow-generator parses the JSON, validates its schema and binds each step to a workstation. Incomplete or unsupported workflows return to node 4 for revision. At node 8, system review verifies parameter compatibility and the sequence of workstation operations. Approved workflows are compiled into workflow templates and assigned template identifiers. At node 9, the lab-operation skill creates a task, obtains a task identifier and dispatches the compiled workflow. Only after these gates can the laboratory cross the physical-execution boundary at node 10. Execution states and result files then return through telemetry as data feedback for subsequent tasks. Verified dispatch did not by itself establish that a task had started or completed.

We collected 32 research tasks from domain experts, with eight tasks in each of four catalysis domains (Fig. 2B, note 4 of the supplementary text, and table S8). Together, these tasks formed a test suite for general catalysis research spanning electrocatalysis, electrochemical materials and testing, photocatalysis and thermal catalysis. Electrocatalysis tasks covered the oxygen evolution, hydrogen evolution, urea oxidation and ethanol oxidation reactions. Electrodes tasks covered cyclic voltammetry, impedance spectroscopy and ion-storage measurements. Photocatalysis tasks addressed $H_2$ and $H_2O_2$ production and selective oxidation, whereas thermal-catalysis tasks focused on selective hydrogenation. Rather than repeating a canonical protocol, the suite required different combinations of sample synthesis, structural characterization, reaction testing and product analysis.

A representative task asks the agent to improve urea conversion and target-product yield for Ni-based catalysts in alkaline urea oxidation. The agent must design and dispatch an experiment to robotic Laboratory 303 and return the resulting task identifier. To fulfil this request, it must recover six implicit fields: the goal, reaction, material, constraints, workflow and expected output. These fields must then be translated into a physically executable workflow linking sample synthesis, structural characterization, reaction testing and product analysis.

Across the full suite, the task prompts embedded diverse planning demands rather than testing success on a single reaction or canonical protocol. These demands implicitly involved task parsing, material design, control design, skill selection, workflow planning, product analysis and mechanistic validation. Agents were required to recover the relevant goal, constraints and expected output from partially specified requests and translate them into executable laboratory workflows.

**AI agent performances**

We paired six agent harnesses with nine frontier language models. Six family-locked pairings were excluded by design, yielding 48 observed configurations (table S9). Each configuration attempted 32 tasks in three independent trials, giving 96 trials per configuration and 4,608 trials overall. All observed configurations completed their assigned trial set.

Independent expert assessment targeted the primary physical question: whether a proposed workflow could be translated into robotic action without human intervention (criteria in table S10). The manual-review cohort comprised all 607 trials reaching verified dispatch at N9, of which 151 were labelled executable at N10. Fig. 3A therefore reports fixed-denominator N10 executable yield: for each configuration, the number of N10 executable labels was divided by the 96 benchmark trials assigned to that configuration. Claude Code with Claude Opus 4.7 contributed 27 executable labels (27/96, 28.1%); Codex with GPT 5.5 contributed 19 (19.8%); Hermes with GPT 5.5 contributed 16 (16.7%); and Hermes with Qwen 3.7 Max contributed 9 (9.4%) (tables S11 and S12). A representative executable Claude Code-Claude Opus 4.7 trajectory is provided in note 5 of the supplementary text and table S13.

Fig. 3B evaluates long-horizon planning among workflows labelled executable at N10. Planning depth was quantified by the number of workstation operations because longer workflows require agents to maintain operation order, parameter compatibility, and vessel and sample states across increasingly extended sequences. This metric therefore captures the ability to preserve physical and logical consistency throughout long-horizon experimental planning. Codex with GPT 5.5 achieved the greatest median planning depth among the displayed configurations, at 18 operations (range, 6-36). Although Claude Code with Claude Opus 4.7 contributed the largest number of executable labels, it did not produce the longest executable workflows. Hermes with MiMo V2.5 Pro generated a 44-operation workflow, whereas Codex with GPT-5.5 produced workflows comprising 36 and 34 operations, demonstrating that some current agents can maintain unusually long experimental horizons. Among the displayed configurations, GPT-5.5-based pairings combined relatively high fixed-denominator N10 yields with long plans; these descriptive comparisons do not estimate independent model or harness effects (tables S14 and S15).

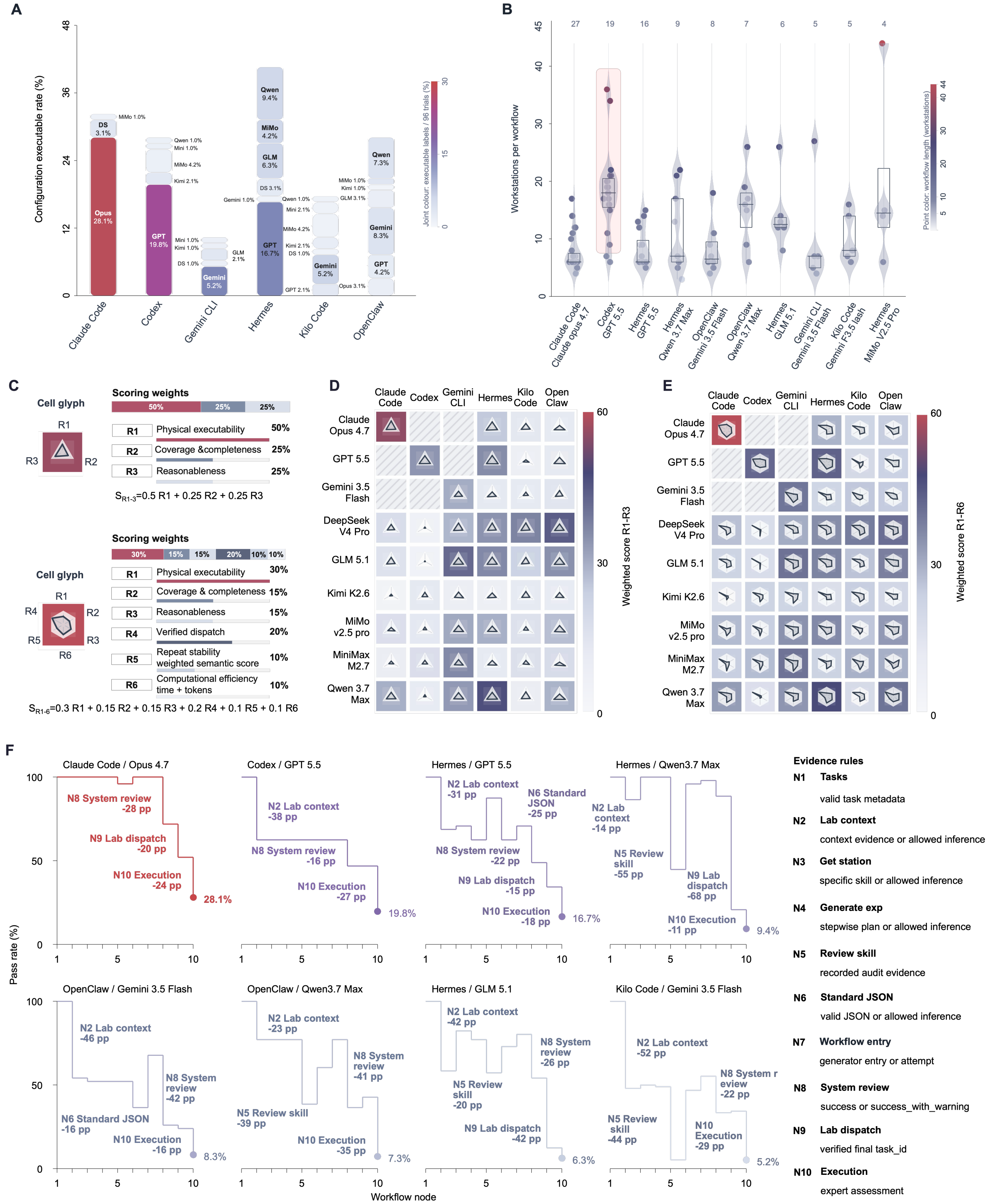

A
B
C
D
E
F
Configuration executable rate (%)
Claude Code
Codex
Gemini CLI
Hermes
Kilo Code
OpenClaw
Workstations per workflow
Cell glyph
Scoring weights
R1 Physical executability 50%
R2 Coverage &completeness 25%
R3 Reasonableness 25%
R4 Verified dispatch 20%
R5 Repeat stability weighted semantic score 10%
R6 Computational efficiency time + tokens 10%
Weighted score R1-R3
Weighted score R1-R6
Pass rate (%)
Workflow node
Evidence rules
N1 Tasks valid task metadata
N2 Lab context context evidence or allowed inference
N3 Get station specific skill or allowed inference
N4 Generate exp stepwise plan or allowed inference
N5 Review skill recorded audit evidence
N6 Standard JSON valid JSON or allowed inference
N7 Workflow entry generator entry or attempt
N8 System review success or success_with_warning
N9 Lab dispatch verified final task_id
N10 Execution expert assessment

**Fig. 3. AI-agent performance across executable planning, validation and laboratory dispatch.** A, Fixed-denominator N10 executable yield from expert adjudication of trials reaching verified dispatch at N9. Stacked segments show N10 executable-label counts divided by the 96 structured-benchmark trials assigned to the corresponding configuration. B, Operation counts for the ten displayed executable-workflow groups in the N10 cohort. Violins show distributions, boxes show medians and interquartile ranges, and points denote individual workflows. C, Glyph encodings and scoring weights for JSON-stage and dispatch-oriented evaluation. R1-R6 denote physical-system executability, problem coverage and workflow completeness, visible scientific-design reasonableness, verified dispatch, repeat stability of the weighted semantic score and computational efficiency, respectively. D, JSON-stage semantic quality across 48 configurations. Heat shows the weighted R1-R3 composite, and glyphs show conditional metric means. E, Dispatch-oriented quality using the weighted R1-R6 composite. Hatched cells in D and E indicate absent family-locked pairings. F, Node-specific workflow success rates across 96 trials for the eight highest-ranked agent–model pairings. Labels indicate adjacent decreases exceeding 10 percentage points, and the right-hand panel summarizes the evidence rules used to score nodes N1-N10.

We next evaluated plan quality and system-level performance using six prespecified dimensions: R1, physical-system executability; R2, problem coverage and workflow completeness; R3, visible scientific-design reasonableness; R4, verified dispatch; R5, repeat stability of the weighted primary semantic score; and R6, computational efficiency. Fig. 3C defines the two composite metrics reported in Fig. 3D and Fig. 3E. The semantic composite weighted R1, R2 and R3 by 0.50, 0.25 and 0.25, respectively. The six-axis composite extended this evaluation to R4, R5 and R6 with weights of 0.30, 0.15, 0.15, 0.20, 0.10 and 0.10 across the six dimensions. R5 quantified within-task repeat variation in the weighted primary semantic score (fig. S3), whereas R6 combined cohort-relative wall-time and token-efficiency scores. To prevent configurations that produced few scorable plans from being overestimated, the R1-R3 components used all 96 assigned trials as the denominator, with unscorable trials contributing zero (note 11 of the supplementary text and table S16). R1-R3 were independently scored by two model judges, whose paired scores were strongly correlated across all three dimensions (Pearson r = 0.892–0.930; fig. S4), whereas R4-R6 were derived from dispatch, runtime and repeat records (see Materials and Methods; note 6 of the supplementary text, table S17, and fig. S5).

On the 96-trial fixed-denominator R1-R3 composite, Claude Code with Claude Opus 4.7 ranked highest at 56.0 (Fig. 3D). Hermes with Qwen 3.7 Max and OpenClaw with DeepSeek V4 Pro followed at 47.0 and 44.4, respectively. Conditional scores calculated only among completed plans showed a different ordering. Hermes with GPT 5.5 achieved the highest conditional mean of 57.3, followed by Claude Code with Claude Opus 4.7 at 56.0 and Codex with GPT 5.5 at 54.8. However, each GPT 5.5 pairing yielded only 60 scorable plans, compared with 96 for Claude Code with Claude Opus 4.7. Their fixed-denominator composites consequently fell to 35.8 and 34.2, respectively. Plan-production reliability therefore explained much of the difference between conditional plan quality and end-to-end semantic performance (fig. S6; tables S18 to S23). Task-composition adjustment produced conditional means closely aligned with their unadjusted values (fig. S7), and equal weighting of R1-R3 preserved the four highest-ranked configurations (fig. S8; see the equal-weight sensitivity analysis in Materials and Methods).

Adding R4, R5 and R6 preserved the leading pairing but altered the remaining ranking (Fig. 3E and note 7 of the supplementary text). Claude Code with Claude Opus 4.7 achieved a six-axis score of 60.4. It was followed by Hermes with Qwen 3.7 Max (47.7), Hermes with GPT 5.5 (43.4) and Codex with GPT 5.5 (42.6). These scores did not directly track the number of verified dispatches. Claude Code with Claude Opus 4.7 produced 50 verified task records, Codex with GPT 5.5 produced 45, OpenClaw with Qwen 3.7 Max produced 41, Hermes with GPT 5.5 produced 33 and Hermes with Qwen 3.7 Max produced 20 despite ranking second overall (fig. S9 and tables S24 to S29).

The clearest divergence occurred between the two OpenClaw pairings. DeepSeek V4 Pro scored 44.4 at the structured-plan stage but yielded no verified dispatch. By contrast, Qwen 3.7 Max scored only 28.9 at that stage but produced 41 verified dispatches. Semantic plan quality and successful entry into the laboratory task infrastructure therefore represented related but distinct capabilities.

Independent node-level success profiles localized stage-specific variation across agent–model pairings (Fig. 3F). Each profile used a fixed denominator of 96 assigned trials per configuration. N1-N8 reported success at each process node independently, N9 represented verified laboratory dispatch, and N10 represented expert-adjudicated physical executability. Because N1-N8 were evaluated independently, rates were not constrained to decrease across nodes. Adjacent changes therefore describe relative node performance rather than cumulative trial attrition; only decreases exceeding 10 percentage points were annotated. Endpoint definitions and statistical procedures are provided in note 8 of the supplementary text, table S30, and data S4. Claude Code with Opus 4.7 achieved 100% success at N1-N4 and N6-N7, whereas system-review success was 71.9%, 28.1% points below workflow-entry success. Codex with GPT 5.5 and Kilo Code with Gemini 3.5 Flash showed their largest decreases at laboratory-context acquisition, from 100% to 62.5% and 47.9%, respectively. Hermes with Qwen3.7 Max illustrated the distinction between independent and cumulative scoring: review-skill success was 44.8%, whereas standard-JSON generation and workflow entry reached 95.8% and 97.9%, respectively. Its largest decrease occurred at verified dispatch, from 88.5% at system review to 20.8% at N9, a 67.7-percentage-point difference. By contrast, Claude Code with Opus 4.7 and Codex with GPT 5.5 showed comparatively robust and coherent profiles across stages: marked declines were confined to the functionally demanding stages of context acquisition, system review, dispatch and expert-adjudicated executability, and neither pairing showed the pronounced node-to-node reversals observed in several of the other displayed profiles. Across the Top 8, verified dispatch ranged from 12.5% to 52.1%, whereas executable endpoints ranged from 5.2% to 28.1%. N10 was assessed only among trials reaching N9 and remained nested within verified dispatch, although its displayed rate used all 96 assigned trials as the denominator. Together, these profiles localized pairing-specific constraints in context grounding, workstation retrieval, workflow review, system validation, laboratory dispatch and expert-adjudicated executability. Profiles for the remaining 40 pairings are provided in figs. S10 to S49.

These results show that laboratory competence is a property of the complete model-agent-laboratory stack, rather than the language model alone. High semantic plan quality did not ensure reliable context acquisition, system review or dispatch. Evaluations restricted to completed plans would therefore overestimate practical autonomy. The central capability boundary remains the reproducible conversion of scientific intent into validated and executable laboratory action. These pairing-specific failure profiles also identify targets for improvement. Feedback generated through agent interactions with the physical laboratory could strengthen context grounding, workflow review or system validation according to each pairing's dominant failure mode.

**Iterative adaptation test on open discovery task**

Next, we evaluated Codex/GPT-5.5, which combined relatively high executable-label contribution with long-horizon planning in the preceding analyses. Rather than another prespecified catalysis problem, we placed the agent in *goal* mode on an open-ended task (Fig. 4A). The prompt defined a closed loop of design, robotic execution, evidence retrieval and replanning, but prescribed neither the catalyst system, experimental protocol nor performance target, specifying only the available laboratory capabilities. The task tested context retention, long-horizon planning and adaptation to physical feedback, rather than catalyst discovery. The detailed closed-loop workflow is provided in note 9 of the supplementary text and fig. S50.

A

**Open-ended discovery task in "goal" mode:**

You are an independent PI with authority to design and dispatch scientific tasks to a real self-driving laboratory. The lab covers synthesis, post-processing, performance testing and spectroscopic characterization across photo-, electro-, and thermal-catalysis. Choose, on your own, a catalysis problem that is high-impact, high-difficulty, scientifically valuable, and feasibly executable within this lab's capability boundary (single-domain or photo-electro / electro-thermal / photo-electro-thermal coupling), and push it forward through multiple rounds of real experiments.
Each round follows the loop: design → dispatch to Lab 303 (record the returned task ID) → wait for the real returned data → let the data decide the next round.

B

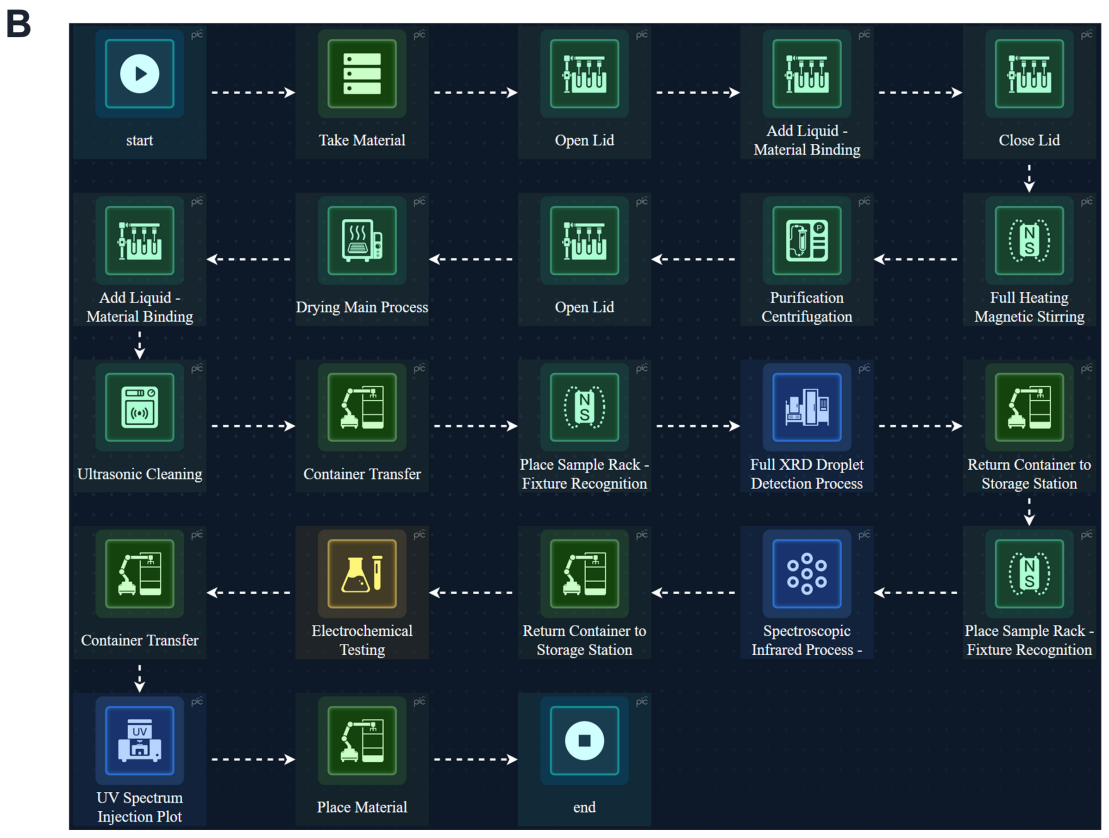

C

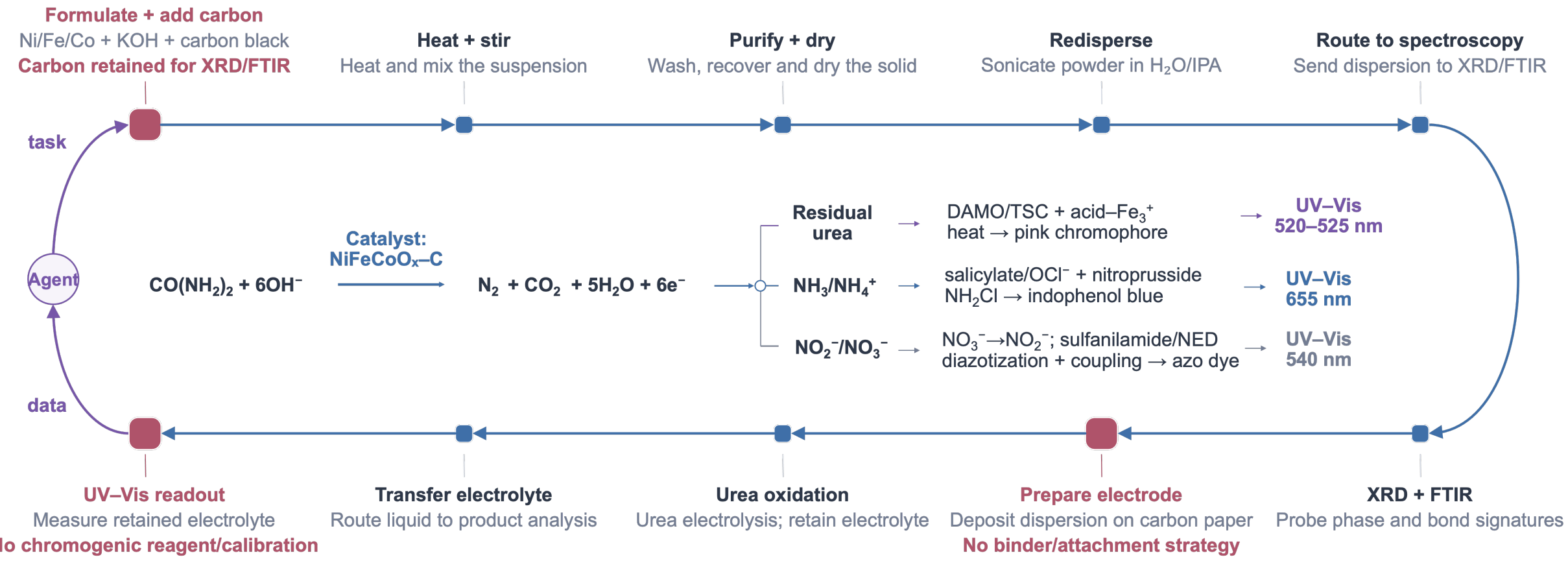

D

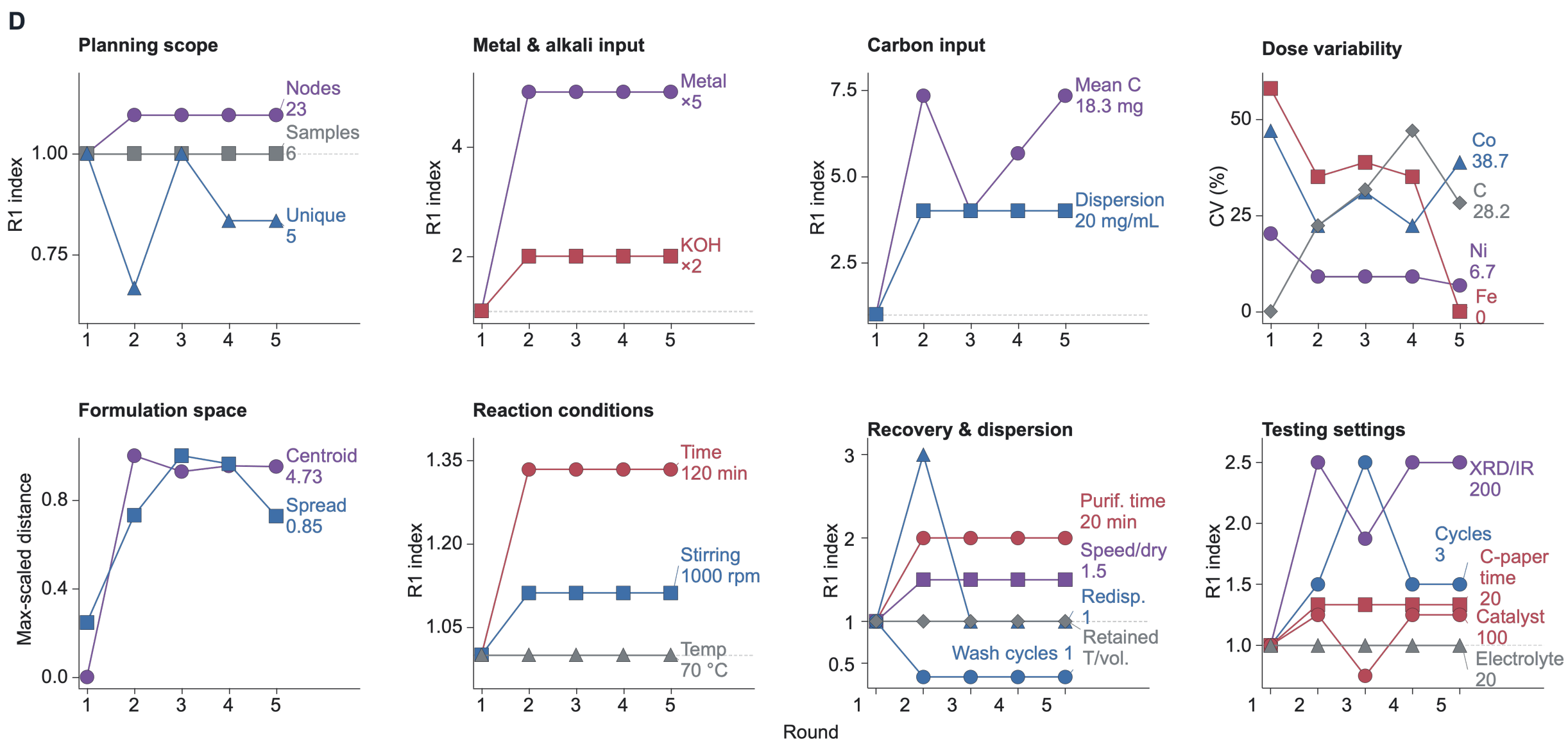

**Fig. 4. Agent-directed open-ended research across five planning rounds in an operational robotic laboratory.** A, Open-ended task assigned to Codex/GPT-5.5. B, Robot-executable workflow designed by the agent in round 1 for the synthesis, purification, dispersion, structural characterization and electrochemical evaluation of NiFeCoO$_x$-C catalysts for alkaline urea oxidation. C, Agent-driven closed-loop workflow for NiFeCoO$_x$-C synthesis, characterization and urea electrooxidation analysis. The central pathway presents the alkaline urea-oxidation reaction catalyzed by NiFeCoO$_x$-C. The three subsequent branches (red line) illustrate representative analytical methods and associated colour-development mechanisms that domain experts would use for UV-Vis detection: DAMO/TSC-based colour development for residual urea, salicylate-hypochlorite-induced indophenol formation for $NH_3$/$NH_4^+$, and nitrate reduction followed by sulfanilamide/NED-mediated diazotization and azo coupling for $NO_2^-$/$NO_3^-$. IPA, isopropyl alcohol; DAMO, diacetyl monoxime; TSC, thiosemicarbazide; NED, N-(1-naphthyl)ethylenediamine. D, Evolution of eight groups of planned variables and experimental parameters from round 1 to round 5: planning scope, metal and alkali input, carbon input, dose variability, formulation space, reaction conditions, recovery and dispersion, and testing settings. Parameters are reported directly or relative to round 1, whereas derived metrics quantify within-round dosing variability and changes in the explored formulation space. Definitions, calculation procedures and source data for the derived variables are provided in Materials and Methods.

From the initial prompt, the agent selected electrochemical alkaline urea oxidation and designed an initial study of NiFeCoO$_x$-C catalysts. The round 1 workflow integrated precursor addition, reaction, purification, drying, redispersion, substrate deposition and multimodal characterization across 21 operations (Fig. 4B). Six NiFeCo compositions were screened under a common synthesis and testing protocol. Fig. 4C expands this operational overview by resolving the individual chemical manipulations and the associated reaction and analytical logic. The resulting workflow was sufficiently specified for direct execution in the robotic laboratory, although several choices differed from conventional domain-expert practice. The large amount of carbon powder introduced during synthesis could mask catalyst signatures during subsequent XRD and infrared characterization (*21, 22*). Electrode preparation also omitted a suitable binder, which could reduce electrode integrity and stability during electrochemical testing (*23*). The central scheme presents the six-electron alkaline oxidation of urea over NiFeCoO$_x$-C: $CO(NH_2)_2 + 6OH^- \rightarrow N_2 + CO_2 + 5H_2O + 6e^-$ (*24*). The proposed UV-Vis measurements likewise omitted analyte-specific chromogenic reagents. Domain experts would typically use diacetyl monoxime (DAMO)/thiosemicarbazide (TSC) for residual urea (*25*) and the salicylate–hypochlorite–nitroprusside method for $NH_3$/$NH_4^+$ (*26*). $NO_2^-$/$NO_3^-$ analysis would instead involve nitrate reduction followed by diazotization with sulfanilamide and azo coupling with N-(1-naphthyl)ethylenediamine (NED) (*27*). We intentionally did not inject these science-based corrections into the initial plan. Instead, the workflow was retained as generated, allowing the agent to receive the experimental data and revise its strategy through subsequent analysis. This design therefore evaluated whether data-driven feedback could enable the agent to recognize and correct limitations in its own experimental planning. We then quantified the five-round planning trajectory across eight metric groups covering planning scope, material inputs, formulation diversity, processing conditions and testing settings (Fig. 4D). Definitions and calculation procedures are provided in Materials and Methods and note 12 of the supplementary text, with round-level parameters, planned formulations and returned modalities reported in tables S31 to S34. The complete five-round Codex/GPT-5.5 planning-and-dispatch trajectory is documented in note 13 of the supplementary text and table S35. Round-resolved characterization results for rounds 1 to 4 are shown in figs. S51 to S54.

Round 1 yielded little recovered material and no valid electrochemical result. Before round 2, a human researcher provided one documented scientific intervention, judging that product recovery was insufficient for reliable characterization and testing. No comparable scientific judgement was recorded in later transitions. Following this feedback, the agent expanded the workflow from 21 to 23 operations while retaining six samples per round (figs. S55 to S58). Metal-stock concentration increased from 0.05 to 0.25 M, KOH input doubled and carbon-dispersion concentration increased from 5 to 20 mg ml$^{-1}$. Reaction time increased from 90 to 120 min, purification speed from 8,000 to 12,000 r.p.m., and the remaining recovery and drying conditions were also intensified.

Rounds 3-5 retained the same planned process backbone while reallocating composition and carbon-dose comparisons, including revisiting previously examined conditions. Mean carbon input decreased to 10.0 mg in round 3 before increasing to 14.2 and 18.3 mg in rounds 4 and 5, respectively. The number of redispersion conditions increased from one to three in round 2 but returned to one thereafter. Electrode-deposition cycles were two, three, five, three and three across rounds 1-5, respectively. The number of unique core formulations followed 6, 4, 6, 5 and 5 across the five rounds.
Across successive rounds, Ni- and Fe-dose variability narrowed, whereas Co- and carbon-dose diversity increased during later exploration. Within-round formulation spread increased from 0.29 in round 1 to 1.17 in round 3 before decreasing to 0.85 in round 5. The formulation centroid shifted sharply from the round 1 region in round 2 and remained displaced thereafter.
Together, these results show that the agent maintained experimental context, preserved the selected research goal and sustained planning across five rounds. It used returned evidence from rounds to coordinate changes across multiple parameter dimension. However, its adaptation remained local. Characterization and performance feedback prompted further adjustments to formulations and operating conditions, but did not trigger workflow-level replanning or analytical-method redesign. Across rounds, the agent retained the carbon-powder loading and introduced neither an electrode binder nor analyte-specific chromogenic reagents for UV-Vis analysis.

## Discussion

Here we systematically stress-tested large language model agents in a robotic chemistry laboratory, tracing 32 catalysis tasks from scientific intent to machine execution. The agents proposed plausible experiments and generated long executable plans, including one spanning 44 workstation operations. Yet the best large language model–harness pairing achieved only 28.1% expert-assessed executability. This gap defines a central capability boundary: proficiency in scientific planning does not translate directly into physical autonomy. Performance therefore emerged from the coupled model–laboratory system, revealing operational limits that plan-only benchmarks cannot capture.
Measured executability was conditioned by how laboratory operations, constraints and state transitions were represented to the agent. More precise constraint encoding and machine-readable skill descriptions could improve the grounding of agent decisions. Direct interaction with the robotic laboratory provides diagnostic signals through rejected workflows, execution states and returned experimental data. These signals can locate failures in reasoning, harness design or laboratory abstraction and guide targeted improvement. End-to-end autonomy should therefore be treated as a co-design problem, requiring agents and robotic laboratory to be jointly optimized aroundphysical action and feedback.

**References and Notes**

1. C. Lu, C. Lu, R. T. Lange, Y. Yamada, S. Hu, J. Foerster, D. Ha, J. Clune, Towards end-to-end automation of AI research. *Nature* **651**, 914–919 (2026).
2. J. Gottweis, W.-H. Weng, A. Daryin, T. Tu, P. Sirkovic, A. Myaskovsky, G. Glowaty, F. Weissenberger, A. Orlandi, D. Popovici, A. Palepu, K. Rong, R. Tanno, K. Saab, F. Zhang, J. Blum, A. Carroll, K. Kulkarni, N. Tomašev, D. Zverinski, I. Rendulic, E. Vedadi, F. Hasler, L. Rimanic, M. Boia, I. Budiselic, B. Feinstein, M. Bellaiche, T. Sheffer, J. Freyberg, J. Ratcliff, O. Bertolli, K. Chou, A. Hassidim, B. Gokturk, A. Vahdat, Y. Guan, V. Dhillon, E. D. Vaishnav, B. Lee, T. R. D. Costa, J. R. Penadés, G. Peltz, Y. Matias, J. Manyika, D. Hassabis, Y. Xu, P. Kohli, A. Pawlosky, A. Karthikesalingam, V. Natarajan, Accelerating scientific discovery with Co-Scientist. *Nature* **655**, 487–496 (2026).
3. A. E. Ghareeb, B. Chang, L. Mitchener, A. Yiu, C. J. Szostkiewicz, D. Shved, G. J. Gyimesi, J. M. Laurent, S. M. Wright, M. T. Razzak, A. D. White, S. C. Finnemann, M. M. Hinks, S. G. Rodriques, A multi-agent system for automating scientific discovery. *Nature* **655**, 497–505 (2026).
4. F. Trost, B. Zhang, I. Aring, M. Bauer, L. Glamann, M. Wessolly, K. Johnson, H. Göbel, T. Lerbs, T. Sangenne, P. Herrmann, F. Mairinger, C. Kopp, S. Michels, A. Rasokat, M. Heldwein, S. Wagner, B. Schömig-Markiefka, J. Wolf, S. Hartmann, C. Wickenhauser, A. Bychkov, J. P. Klussmann, A. Quaas, R. Buettner, Y. Tolkach, An agentic framework for autonomous scientific discovery in cancer pathology. *Nat. Med.* **32**, 2254–2266 (2026).
5. J. R. Penadés, J. Gottweis, L. He, J. B. Patkowski, A. Daryin, W.-H. Weng, T. Tu, A. Palepu, A. Myaskovsky, A. Pawlosky, V. Natarajan, A. Karthikesalingam, T. R. D. Costa, AI mirrors experimental science to uncover a mechanism of gene transfer crucial to bacterial evolution. *Cell* **188**, 6654–6665.e2 (2025).

6. B. Burger, P. M. Maffettone, V. V. Gusev, C. M. Aitchison, Y. Bai, X. Wang, X. Li, B. M. Alston, B. Li, R. Clowes, N. Rankin, B. Harris, R. S. Sprick, A. I. Cooper, A mobile robotic chemist. *Nature* **583**, 237–241 (2020).
7. B. A. Koscher, R. B. Canty, M. A. McDonald, K. P. Greenman, C. J. McGill, C. L. Bilodeau, W. Jin, H. Wu, F. H. Vermeire, B. Jin, T. Hart, T. Kulesza, S.-C. Li, T. S. Jaakkola, R. Barzilay, R. Gómez-Bombarelli, W. H. Green, K. F. Jensen, Autonomous, multiproperty-driven molecular discovery: From predictions to measurements and back. *Science* **382**, eadi1407 (2023).
8. F. Strieth-Kalthoff, H. Hao, V. Rathore, J. Derasp, T. Gaudin, N. H. Angello, M. Seifrid, E. Trushina, M. Guy, J. Liu, X. Tang, M. Mamada, W. Wang, T. Tsagaantsooj, C. Lavigne, R. Pollice, T. C. Wu, K. Hotta, L. Bodo, S. Li, M. Haddadnia, A. Wołos, R. Roszak, C. T. Ser, C. Bozal-Ginesta, R. J. Hickman, J. Vestfrid, A. Aguilar-Granda, E. L. Klimareva, R. C. Sigerson, W. Hou, D. Gahler, S. Lach, A. Warzybok, O. Borodin, S. Rohrbach, B. Sanchez-Lengeling, C. Adachi, B. A. Grzybowski, L. Cronin, J. E. Hein, M. D. Burke, A. Aspuru-Guzik, Delocalized, asynchronous, closed-loop discovery of organic laser emitters. *Science* **384**, eadk9227 (2024).
9. N. J. Szymanski, B. Rendy, Y. Fei, R. E. Kumar, T. He, D. Milsted, M. J. McDermott, M. Gallant, E. D. Cubuk, A. Merchant, H. Kim, A. Jain, C. J. Bartel, K. Persson, Y. Zeng, G. Ceder, An autonomous laboratory for the accelerated synthesis of inorganic materials. *Nature* **624**, 86–91 (2023).
10. J. Bai, S. Mosbach, C. J. Taylor, D. Karan, K. F. Lee, S. D. Rihm, J. Akroyd, A. A. Lapkin, M. Kraft, A dynamic knowledge graph approach to distributed self-driving laboratories. *Nat. Commun.* **15**, 462 (2024).
11. R. D. King, J. Rowland, S. G. Oliver, M. Young, W. Aubrey, E. Byrne, M. Liakata, M. Markham, P. Pir, L. N. Soldatova, A. Sparkes, K. E. Whelan, A. Clare, The automation of science. *Science* **324**, 85–89 (2009).
12. A. Slattery, Z. Wen, P. Tenblad, J. Sanjosé-Orduna, D. Pintossi, T. den Hartog, T. Noël, Automated self-optimization, intensification, and scale-up of photocatalysis in flow. *Science* **383**, eadj1817 (2024).
13. S. Steiner, J. Wolf, S. Glatzel, A. Andreou, J. M. Granda, G. Keenan, T. Hinkley, G. Aragon-Camarasa, P. J. Kitson, D. Angelone, L. Cronin, Organic synthesis in a modular robotic system driven by a chemical programming language. *Science* **363**, eaav2211 (2019).
14. K. Huang, S. Zhang, H. Wang, Y. Qu, Y. Lu, R. Li, Y. Roohani, L. Qiu, S. Cao, G. Li, J. Zhang, D. Yin, R. Wierenga, D. Kavi, S. Liu, T. She, S. Marwaha, J. N. Carter, X. Zhou, M. T. Wheeler, J. A. Bernstein, M. Wang, P. He, J. Zhou, M. P. Snyder, L. Cong, A. Regev, J. Leskovec, Autonomous biomedical research with an artificial intelligence agent. *Science* **393**, eadz4351 (2026).
15. D. A. Boiko, R. MacKnight, B. Kline, G. Gomes, Autonomous chemical research with large language models. *Nature* **624**, 570–578 (2023).
16. T. Song, M. Luo, X. Zhang, L. Chen, Y. Huang, J. Cao, Q. Zhu, D. Liu, B. Zhang, G. Zou, G. Zhang, F. Zhang, W. Shang, Y. Fu, J. Jiang, Y. Luo, A multiagent-driven robotic AI chemist enabling autonomous chemical research on demand. *J. Am. Chem. Soc.* **147**, 12534–12545 (2025).
17. L. Huang, C. Zhang, Y. Fu, Y. Jiang, E. He, M.-Q. Qi, M.-H. Du, X.-J. Kong, J. Cheng, L. Cronin, C. Wang, Natural-language-interfaced robotic synthesis for AI-copilot-assisted exploration of inorganic materials. *J. Am. Chem. Soc.* **147**, 23014–23025 (2025).
18. Z. Chen, A. N. Petsch, A. J. Israelski, R. Plumley, L. Shen, C. Wang, C. Peng, Y. Ni, A. Bansil, S. Chowdhury, M. Li, J. B. Thayer, V. Thampy, J. J. Turner, An agentic artificially intelligent X-ray scientist. *Nat. Mach. Intell.* **8**, 1075–1086 (2026).
19. Z. Chen, S. Chen, Y. Ning, Q. Zhang, B. Wang, B. Yu, Y. Li, Z. Liao, C. Wei, Z. Lu, V. Dey, M. Xue, F. N. Baker, B. Burns, D. Adu-Ampratwum, X. Huang, X. Ning, S. Gao, Y. Su, H. Sun, "ScienceAgentBench: Toward Rigorous Assessment of Language Agents for Data-Driven Scientific Discovery" in *The Thirteenth International Conference on Learning Representations* (ICLR, Singapore, 2025).
20. B. P. Majumder, H. Surana, D. Agarwal, B. Dalvi Mishra, A. Meena, A. Prakhar, T. Vora, T. Khot, A. Sabharwal, P. Clark, "DiscoveryBench: Towards Data-Driven Discovery with Large Language Models" in *The Thirteenth International Conference on Learning Representations* (ICLR, Singapore, 2025).
21. P. Scardi, P. L. Antonucci, XRD characterization of highly dispersed metal catalysts on carbon support. *J. Mater. Res.* **8**, 1829–1835 (1993).
22. J. M. O'Reilly, R. A. Mosher, Functional groups in carbon black by FTIR spectroscopy. *Carbon* **21**, 47–51 (1983).
23. S. M. Alia, K. S. Reeves, J. S. Baxter, D. A. Cullen, The impact of ink and spray variables on catalyst layer properties, electrolyzer performance, and electrolyzer durability. *J. Electrochem. Soc.* **167**, 144512 (2020).
24. B. K. Boggs, R. L. King, G. G. Botte, Urea electrolysis: direct hydrogen production from urine. *Chem. Commun.* **2009**, 4859–4861 (2009).
25. A. Mather, D. Roland, The automated thiosemicarbazide-diacetyl monoxime method for plasma urea. *Clin. Chem.* **15**, 393–396 (1969).
26. C. E. Bower, T. Holm-Hansen, A salicylate–hypochlorite method for determining ammonia in seawater. *Can. J. Fish. Aquat. Sci.* **37**, 794–798 (1980).
27. K. M. Miranda, M. G. Espey, D. A. Wink, A rapid, simple spectrophotometric method for simultaneous detection of nitrate and nitrite. *Nitric Oxide* **5**, 62–71 (2001).

28. J. Pineau, P. Vincent-Lamarre, K. Sinha, V. Larivière, A. Beygelzimer, F. d'Alché-Buc, E. Fox, H. Larochelle, Improving reproducibility in machine learning research (a report from the NeurIPS 2019 reproducibility program). *J. Mach. Learn. Res.* **22**, 1–20 (2021).
29. J. Dodge, S. Gururangan, D. Card, R. Schwartz, N. A. Smith, "Show Your Work: Improved Reporting of Experimental Results" in *Proceedings of the 2019 Conference on Empirical Methods in Natural Language Processing and the 9th International Joint Conference on Natural Language Processing (EMNLP-IJCNLP)* (Association for Computational Linguistics, Hong Kong, China, 2019), pp. 2185–2194.
30. Y. Liu, D. Iter, Y. Xu, S. Wang, R. Xu, C. Zhu, "G-Eval: NLG Evaluation using GPT-4 with Better Human Alignment" in *Proceedings of the 2023 Conference on Empirical Methods in Natural Language Processing* (Association for Computational Linguistics, Singapore, 2023), pp. 2511–2522.
31. L. Zheng, W.-L. Chiang, Y. Sheng, S. Zhuang, Z. Wu, Y. Zhuang, Z. Lin, Z. Li, D. Li, E. P. Xing, H. Zhang, J. E. Gonzalez, I. Stoica, "Judging LLM-as-a-Judge with MT-Bench and Chatbot Arena" in *Advances in Neural Information Processing Systems 36* (Neural Information Processing Systems Foundation, 2023), pp. 46595–46623.
32. A. Kumar, N. Poungpeth, D. Yang, E. Farrell, B. L. Lambert, M. Groh, When large language models are reliable for judging empathic communication. *Nat. Mach. Intell.* **8**, 173–185 (2026).
33. P. Wang, L. Li, L. Chen, Z. Cai, D. Zhu, B. Lin, Y. Cao, L. Kong, Q. Liu, T. Liu, Z. Sui, "Large Language Models are not Fair Evaluators" in *Proceedings of the 62nd Annual Meeting of the Association for Computational Linguistics (Volume 1: Long Papers)* (Association for Computational Linguistics, Bangkok, Thailand, 2024), pp. 9440–9450.
34. J. M. Bland, D. G. Altman, Statistical methods for assessing agreement between two methods of clinical measurement. *Lancet* **327**, 307–310 (1986).
35. B. Efron, Bootstrap methods: another look at the jackknife. *Ann. Stat.* **7**, 1–26 (1979).
36. C. A. Field, A. H. Welsh, Bootstrapping clustered data. *J. R. Stat. Soc. B* **69**, 369–390 (2007).

# Acknowledgments

The AI-driven experiments, simulations and model training were performed on the robotic AI-Scientist platform of Chinese Academy of Sciences (CAS).

**Funding:**

J.J. acknowledges the CAS Strategic Priority Research Program (Grant XDB0450302), the National Natural Science Foundation of China (Grants 22025304, 22033007), and the CAS Project for Young Scientists in Basic Research (Grant YSBR-005) for funding. L.C. acknowledges the National Natural Science Foundation of China (Grant 22573101) and the University of Science and Technology of China (USTC) Startup Program for funding.

**Author contributions:**

Lulu Guo, Yingkai Sun and Xiaobo Li designed and established the benchmark. Lulu Guo and Yingkai Sun performed benchmark testing and scoring, and conducted the open-ended experimental tasks and subsequent analyses. Xiaobo Li, Daobin Liu, Lulu Guo and Luyao Ge contributed to the development of robotic laboratory. Lulu Guo, Luyao Ge, Yuebo Liu, Xiaohui Li and Jie Li developed and maintained the skill repository for the robotic laboratory. Yingkai Sun and Ziming Wang curated, organized and analysed the agent-testing results. Lulu Guo, Luyao Ge, Jingyu Li and Xiaobo Li independently assessed the robotic executability of the dispatched experimental workflows. Lulu Guo, Haitao Zheng, Huijuan Zhang and Bingxu Chen performed the experiments using the robotic laboratory. All authors discussed the results and approved the final manuscript. All authors contributed to the preparation of the manuscript. Xiaobo Li, Linjiang Chen, Yi Luo and Jun Jiang conceived and supervised the project.

**Competing interests:**

The authors declare no competing interests.

**Data, code and materials availability:**

The data supporting the findings of this study are available within the paper and its supplementary materials. The full library of custom skills was provided in Data S5.All code and scripts developed and used in this study, including scoring materials, data-processing and statistical-analysis scripts, figure-generation code and machine-readable summary tables, are available at https://github.com/pic-ai-robotic-chemistry/LabBench.

**Supplementary Materials**

Materials and Methods
Supplementary Text
Figs. S1 to S58
Tables S1 to S35
References
Data S1 to S5
Movies S1 to S6

# Supplementary Materials for

# Stress-testing large language model agents in a robotic chemistry laboratory

Lulu Guo[†,1], Yingkai Sun[†,1], Xiaobo Li[†,1,2,*], Luyao Ge[1], Ziming Wang[1], Haitao Zheng[1], Jingyu Li[1], Huijuan Zhang[1], Bingxu Chen[1], Daobin Liu[1], Yuebo Liu[2], Jie Li[2], Xiaohui Li[2], Linjiang Chen[1,2,3,*], Yi Luo[1,2,*], Jun Jiang[1,2,*]

[1] State Key Laboratory of Precision and Intelligent Chemistry, Hefei National Research Center for Physical Sciences at the Microscale, School of Chemistry and Materials Science, University of Science and Technology of China, Hefei, China

[2] Center for Scientific Intelligence Innovation, Hefei, China

[3] School of Chemistry, School of Computer Science, University of Birmingham, Birmingham, UK

[†] These authors contributed equally

*Correspondence: xiaoboli@ustc.edu.cn; linjiangchen@ustc.edu.cn; yiluo@ustc.edu.cn; jiangj1@ustc.edu.cn

The PDF file includes:

Materials and Methods

Supplementary Text

Figs. S1 to S58

Tables S1 to S35

References

Other Supplementary Material for this manuscript includes the following:

Data S1 to S5

Movies S1 to S6

## Contents

# Materials and Methods

### Agent-to-laboratory control boundary

Agents interacted with the laboratory through three interface classes. Workstation skills exposed admissible operations, parameters, units, ranges, vessel requirements, input and output states, and cross-step constraints. The workflow-generation interface accepted ordered experiment steps and returned either a compiled template identifier or a structured validation error. The laboratory-operation interface resolved the active laboratory, checked dispatch authority, generated tasks from accepted templates, queried task records and returned available result metadata.

Agents could inspect these interfaces, propose and revise steps, and write trial-local request files. They could not bypass laboratory identity, authorization, schema, state or template checks to command an instrument directly. The system-engineering layer resolved workstation type and instance, validated operation names and parameter structures, compiled ordered steps and mediated task creation. The machine-readable skill library used in this study is archived in the LabBench repository (https://github.com/pic-ai-robotic-chemistry/LabBench).

### Skill contracts and workflow representation

Each workstation contract used stable workstation and operation identifiers. It specified physical preconditions, including supported container types, container and sample states, required predecessor operations, and admissible parameter values. Output constraints specified the resulting container and sample states. File-based operations additionally identified the required template class and the fields that an agent could populate. Cross-workstation rules required continuity of container identity and count, explicit material introduction, compatible state transitions, and a terminal storage or analysis state.

Agents represented proposed experiments as ordered JSON steps. Each step identified a workstation, an operation and the parameters required by the corresponding contract. The workflow generator checked JSON structure, mandatory fields, operation naming, workstation binding, parameter type and admissibility, and the ordering constraints available to the service. A rejected request could be revised within the same trial. The final plan analysed in this study was the materialized JSON document retained at either prescribed trial output location; prose not materialized as a final plan was not semantically scored.

### Task construction

The 32 prompts comprised eight tasks in each of four domains: electrocatalysis; electrode materials and electrochemical testing; photocatalysis; and thermal catalysis. The electrocatalysis set covered oxygen evolution, hydrogen evolution, urea oxidation and ethanol oxidation; the electrode set covered cyclic voltammetry, impedance spectroscopy and ion-storage measurements; the photocatalysis set covered hydrogen production, hydrogen peroxide production and selective oxidation; and the thermal-catalysis set covered selective hydrogenation. Each prompt supplied a scientific objective, requested an experimental protocol and instructed the agent to dispatch a newly generated workflow rather than reuse an existing executable template. The complete workstation sequence was not enumerated. Agents therefore had to infer material preparation, controls, characterization or reaction measurements, product analysis and expected data return from the scientific objective and visible skills.

The 32 prompt texts and task codes were locked before data collection. Every observed configuration received the same 32 task codes, with three repeat trials per code. This task-centred design follows scientific-agent benchmarks that assess multiple systems against fixed domain tasks (*19, 20*). The complete prompts and task codes are reproduced in data S1.

### Harness-model configurations and runtime

A configuration combined one agent harness with one named model. The harnesses were Claude Code, Codex, Gemini CLI, Hermes Agent, Kilo Code and OpenClaw. The models were Claude Opus 4.7, GPT 5.5, Gemini 3.5 Flash, DeepSeek V4 Pro, GLM 5.1, Kimi K2.6, MiMo v2.5 Pro, MiniMax M2.7 and Qwen 3.7 Max. These named model identifiers were held fixed throughout data collection. Claude Code-GPT, Claude Code-Gemini, Codex-Claude, Codex-Gemini, Gemini CLI-Claude and Gemini CLI-GPT were unavailable by design and treated as not applicable.

The matrix was not interpreted as a complete factorial estimate of independent harness and model effects; results apply to the observed harness-model pairings and stated runtime conditions. We recorded implementation versions and execution conditions in accordance with reporting recommendations for reproducible machine-learning experiments (*28, 29*). Evaluated releases were Claude Code 2.1.169, Codex 0.138.0, Gemini CLI 0.43.0, Hermes Agent v0.16.0 (5 June 2026), Kilo Code 7.3.1 and OpenClaw 2026.4.15. For each model, we used the highest reasoning level that it supported and that was available through the selected harness; context windows followed official model-card limits. Each trial was limited to 30 min and ran in automatic tool-permission mode with unrestricted tool access. Trials disrupted by an orchestration-framework, network or API failure were discarded and rerun in full from the initial prompt.

### Trial retention and analytical cohort

The analytical cohort retained three accepted records for every observed configuration-task pair. A rerun occupied the predefined repeat slot of the record that it replaced and did not increase the configuration denominator. All retained trials remained in the cohort whether or not they produced a final plan, score or dispatch record, thereby preserving failures before plan materialization. Completeness required 96 retained trial directories and 96 index records for each observed configuration; all 48 configurations satisfied both checks.

The retained cohort comprised 4,608 trials. Of these, 2,596 contained a final JSON plan and 2,012 did not. Every final plan had six finite score observations, corresponding to two judges and three dimensions, for 15,576 score observations in total.

### Outcome assertions

A final JSON plan was a parseable plan materialized at either prescribed final-plan output location. A complete semantic triple was a final plan with finite primary-judge values for R1, R2 and R3. A verified dispatch required at least one numerical task identifier reported by the trial to resolve through the laboratory task-query interface to a task record or associated task evidence. Identifiers present only in model text were insufficient.

A manually executable plan was a verified-dispatch trial row independently adjudicated as executable under the laboratory constraints available to the reviewer. A started task would require affirmative task-start evidence, and a completed experiment would require terminal execution status and returned experimental data. The structured benchmark measured final plans, complete semantic triples and verified dispatches. The independent manual cohort measured manual executability; it did not measure started or completed experiments.

**Manual executability assessment**

The N10 cohort comprised all 607 row-level assessments of trials that reached verified dispatch at N9, spanning 43 harness-model configurations. The assessments were independently performed by domain experts: researchers specializing, respectively, in the integration of self-driving laboratories with electrochemistry and photocatalysis, and researcher specializing in the integration of self-driving laboratories with thermal catalysis. The number of N9 trials varied across configurations from 1 to 50. Of the 607 records, 151 were labelled executable and 456 were labelled non-executable. The cohort contained 601 unique task identifiers. Six identifiers occurred in two trial rows, and every duplicated identifier had concordant executable labels. The primary Fig. 3a analysis retained all row-level annotations without deduplication because the duplicated identifiers represented distinct benchmark trial rows.

For each observed harness–model configuration, Fig. 3a divided the number of records labelled manually executable at N10 by the 96 structured-benchmark trials assigned to that configuration. Segment widths therefore represent fixed-denominator N10 executable yield. Trials that did not reach verified dispatch at N9 were not eligible for N10 adjudication and therefore contributed zero at N10. Because N10 was evaluated only among trials reaching N9, N10/96 summarizes end-to-end retention from assignment to expert-adjudicated executability. These descriptive values were not used to estimate independent model or harness effects. The normalized N1-N10 decisions, declared count-level adjustments, reconstructed profiles and reproducibility scripts are provided in data S4.

**Workflow length**

Workflow length was calculated for the 151 trial records labelled manually executable. For each task identifier, the associated workflow representation was retrieved and its nodes classified by step type. The metric counted nodes whose step type was workstation and excluded process-control start and end nodes. The observed range was 3-44 workstation operations.

**Semantic scoring**

Two model-based judges independently scored each final plan. Model-based evaluation enables scalable, rubric-conditioned assessment (*30, 31*), but reliability is task-dependent (*32*) and systematic evaluator biases can arise (*30, 33*). We therefore prespecified the scoring prompts and anchors, used distinct model–harness pairs, and treated cross-judge agreement as a validation analysis rather than as evidence of agreement with expert human assessors. The primary judge was GPT 5.5 operated through Codex at the highest supported and available reasoning level; the validation judge was DeepSeek V4 Pro operated through Hermes Agent. Each judge assigned a score from 0 to 100 to three dimensions: R1, physical-system executability; R2, problem coverage and workflow completeness; and R3, visible scientific-design reasonableness. Scores were based on the materialized plan and visible task context, not on unrecorded intentions.

For trial i, the primary semantic score was

$S_i = 0.50\ R1_i + 0.25\ R2_i + 0.25\ R3_i.$

The weighting prioritised physical-system executability and used author-defined descriptive coefficients. The coefficients were held fixed across configurations and were not fitted separately by task or model. An equal-weight sensitivity analysis recomputed the denominator-96 semantic score using one-third weight for each dimension (fig. S8). The complete R1-R3 judge prompts, numerical anchors and R1 physical-system constraint reference are reproduced verbatim in data S3. Anchor targets were 92-96, 50-60, 20-30 and 0-20 for R1; 90-95, 55-65 and 20-30 for R2; and 85-92, 55-70 and 25-45 for R3. Four archived runtime copies of each prompt and the R1 constraint reference were SHA-256-identical to the corresponding source rule.

**JSON-stage composite**

The JSON-stage composite combined semantic quality with the probability of materializing a scorable plan. For configuration c and dimension k,

$R_{k,96,c} = \mathrm{mean}(R_{k,i} \mid \text{complete semantic triple in } c) \times n_{\mathrm{complete,c}} / 96.$

The displayed heat value was

$H_{\mathrm{JSON,c}} = 0.50\ R1_{96,c}/100 + 0.25\ R2_{96,c}/100 + 0.25\ R3_{96,c}/100.$

This is algebraically equivalent to assigning zero semantic contribution to trials without a complete triple and averaging over the fixed 96-trial denominator, without manufacturing synthetic trial-level score rows. The glyph in each cell displays the unpenalized conditional R1-R3 means to distinguish low plan yield from low quality among produced plans. Conditional means were diagnostic and were not the primary configuration outcome.

**Dispatch-oriented composite**

For configuration c, the dispatch-oriented composite was

$H_{\mathrm{dispatch,c}} = 0.30\ R1_{96,c}/100 + 0.15\ R2_{96,c}/100 + 0.15\ R3_{96,c}/100 + 0.20\ D_c/96 + 0.10\ E_c/100 + 0.10\ T_c/100,$

where D is the number of verified-dispatch trials, E is the computational-efficiency score and T is the repeat-stability score. Semantic and dispatch components used the fixed 96-trial denominator. Efficiency and stability were configuration-level summaries and were not multiplied by plan yield a second time. Both heatmaps used the same colour scale, with its upper limit set to the maximum observed composite across the two panels. Design-absent cells were displayed as not applicable.

**Computational efficiency**

Efficiency was defined relative to the finite primary-scored trial cohort. Token use and elapsed wall time were ranked separately using empirical mid-rank percentiles. Lower use received a higher score:

token efficiency$_i$ = 100 − percentile(total tokens$_i$),

time efficiency$_i$ = 100 − percentile(duration$_i$).

The trial efficiency score was the arithmetic mean of available components. If one component was missing, the available component was used; trials missing both were omitted from the configuration efficiency mean. Configuration efficiency E was the mean of trial scores. This metric is cohort-relative and does not estimate monetary cost, energy consumption, instrument time or environmental impact.

**Repeat stability**

Repeat stability quantified within-task variation in the primary semantic score. For every configuration-task group with at least two finite scores, the population standard deviation of S across retained repeats was calculated. Task-level standard deviations were then averaged within configuration, and stability was defined as

$T_c = \max\{0, 100 - \text{mean}_{t\in\mathcal{T}_c}[\text{SD}_{\text{pop}}(\{S_{c,t,r}\}_r)]\}$.

Here, $\mathcal{T}_c$ is the set of tasks with at least two finite retained repeat scores for configuration c, and r indexes retained repeats. Population rather than sample standard deviation described the observed repeats. The metric is a descriptive dispersion transformation; it is not an intraclass correlation, measurement-error model or claim of deterministic behaviour.

**Judge agreement**

Agreement used finite pairs in which GPT and DeepSeek scored the same trial and dimension. Because correlation alone does not establish agreement (*34*), Pearson and Spearman correlations were reported alongside mean absolute error and the mean DeepSeek-minus-GPT difference, overall and by dimension. Judge-dimension observations were not treated as independent benchmark trials. Overall Pearson correlation was 0.916, mean absolute error was 7.46 and the mean DeepSeek-minus-GPT difference was +1.59. Dimension-specific Pearson correlations were 0.930 for R1, 0.921 for R2 and 0.892 for R3. These statistics quantify concordance between the two model judges and do not establish validity against expert human ratings.

**Uncertainty and sensitivity analyses**

Configuration-level semantic uncertainty used a task-cluster bootstrap based on nonparametric resampling (*35*). The task, rather than the individual trial, was the resampling unit so that retained repeats remained clustered (*36*). Trial semantic scores were first averaged within each of the 32 tasks for a configuration, with absent scores contributing zero under the denominator-96 definition. The 32 task means were sampled with replacement 2,000 times. The 2.5th and 97.5th percentiles of bootstrap means formed the displayed 95% interval. The random-number generator used a fixed base seed of 20260711 with a deterministic configuration-specific offset.

To assess sensitivity to the composition of finite-scored tasks, each finite trial score was adjusted by subtracting the global mean for its task and adding the global mean across finite trial scores. Adjusted scores were averaged by configuration and compared with unadjusted conditional means. This descriptive adjustment does not replace a fully balanced factorial estimate.

**Open-ended autonomous-research study**

Codex 0.142.3 with GPT 5.5 was placed in goal mode for five successive planning rounds. The initial prompt, reproduced in note 10 of the supplementary text, supplied an open scientific objective and the agent-readable capabilities of Laboratory 303 but specified no catalyst system, experimental protocol or performance target. The agent selected alkaline urea oxidation and designed a study of $NiFeCoO_x$-C catalysts. Here it was used to assess context retention, long-horizon planning and adaptation to physical feedback, not to establish catalyst discovery. No result from the structured benchmark was used as evidence of autonomous discovery.

At each planning round, the agent generated a structured plan that was assessed for executability in Laboratory 303. A non-executable plan received structured constraint feedback and could be redesigned; the run terminated if no feasible plan was produced within ten redesign rounds. An accepted plan was submitted for dispatch. After a human operator confirmed experiment completion, result data were retrieved through the laboratory skill, analysed by the agent and used to update the evidence and objective for the next round. The operational cycle therefore comprised design, executability assessment, dispatch, physical execution, result retrieval, analysis and replanning.

Human scientific input was recorded separately from procedural confirmation of task completion. After round 1, a researcher made one documented scientific intervention: the recovered product was judged insufficient for reliable characterization and testing. No comparable scientific judgement was recorded in subsequent round transitions.

Round-level plans were summarized in eight display groups. Planning scope comprised the number of executable workstation nodes and planned samples. Metal and alkali input comprised metal-stock concentrations, Ni, Fe and Co dosing, and KOH input. Carbon input comprised carbon-dispersion concentration and per-formulation carbon dose. Dose variability summarized within-round variation in metal and carbon dosing. Formulation space comprised the number of unique core formulations, within-round formulation spread and displacement of the formulation centroid from its round 1 position. Reaction conditions comprised the agent-specified synthesis settings. Recovery and dispersion comprised centrifugation, drying and redispersion settings. Testing settings comprised electrode-deposition cycles and electrochemical test parameters. Direct parameters were reported in their stated units or relative to the corresponding round 1 value; derived quantities were calculated within round.

To quantify dose variability, the coefficient of variation was calculated separately for Ni, Fe, Co and carbon within each round as $CV = 100 \times s/\bar{x}$, where s is the sample standard deviation across the six planned samples (n − 1 denominator) and $\bar{x}$ is their arithmetic mean; computed values with absolute magnitude below $1 \times 10^{-12}$ were reported as zero. For the formulation-space metrics, Ni, Fe, Co, KOH and carbon were each globally z-standardized across all 30 planned samples as $z = (x - \mu)/\sigma$, where μ and σ are the global mean and population standard deviation. Each round centroid was the arithmetic mean of its six five-dimensional standardized vectors, and the five coordinates entered the Euclidean distance with equal unit weight. Within-round formulation spread was the median of the six Euclidean distances from the sample vectors to their round centroid. Formulation-centroid displacement was the Euclidean distance between each round centroid and the round 1 centroid in the same standardized space; Round 1 was zero by construction. For display in Fig. 4d, each formulation-space distance series was divided by its own maximum across rounds 1–5; this scaling compares temporal patterns within a metric and not absolute magnitudes between metrics. The underlying sample-level values and round-level summaries are provided in tables S31 and S32.

## Supplementary Text

### Note 1. Combinatorial complexity of the physical workflow space

To obtain a conservative, device-level estimate of the workflow space associated with materials synthesis, we abstracted the laboratory process into four ordered stages: synthesis, post-processing, testing and characterization. Workflows containing two or more stages were constrained to include both synthesis and post-processing. At the device-combination level, synthesis was represented by 117 route variants derived from liquid-only, solid-only and liquid–solid input modes combined with three mixing or heating options. Post-processing contributed seven non-empty combinations of drying, centrifugation and purification. When included, testing contributed seven non-empty combinations of electrochemical testing and two photocatalytic platforms, whereas characterization contributed 1,023 non-empty combinations across ten analytical modalities. Summing the four valid workflow classes—synthesis–post-processing, synthesis–post-processing–testing, synthesis–post-processing–characterization and synthesis–post-processing–testing–characterization—yielded 6,709,248 device-combination workflow families, approximated as $6.71 \times 10^6$. This value is a restricted combinatorial estimate based only on the material-synthesis stages and associated device combinations represented in the platform. It does not include the substantially larger space of operation-level parameters, such as reagent identities, quantities, temperatures, durations or device settings, and therefore should not be interpreted as the complete experimental parameter space. Moreover, these combinations were not individually validated for physical compatibility or end-to-end executability; accordingly, the estimate represents the size of the abstract device-combination design space rather than the number of experimentally executable workflows.

This document consolidates the counting rules applied to the current formal workstation skills, the audit files used for workflow review, and the two system-level skills. Counts are reported as traceable units rather than globally deduplicated concepts.

**Table S1. Source-aligned summary.**

| Metric | Count | Operational definition |
|---|---|---|
| Formal workstations | 45 | Workstation directories containing a formal SKILL.md in the three specified modules. |
| Executable operation interfaces | 62 | Formal operation sections, identified by workstation and operation together. |
| Documented constraint units | 433 | Interface-state, local workflow, authorization, and workflow-generation rules under the stated scope. |
| Configurable parameter fields | 456 | Actual data rows in parameter-setting tables, including required, optional, parent, and nested fields. |

### 1. Scope and Counting Principles

Operation and parameter counts use only the Characterization, Reaction and Testing, and Synthesis module directories. Constraint counts use operation input-output fields, entries under dedicated audit workflow-constraint sections, and unique declared rules from lab-operation and workflow-generator. Repeated interface fields remain counted at each operation because they constrain different executable transitions.

**Table S2. Source boundaries used for the recommended count.**

| Source | Included content | Treatment |
|---|---|---|
| Three formal workstation modules | 45 workstation skills, 62 operations, and parameter-setting tables | Primary source for workstation, operation, and parameter counts. |
| references_audit | Input-output state fields and entries under Overall Workflow Constraints | Matched to the 45 formal workstations; one missing audit operation is recovered from its formal skill. |
| lab-operation | Authorization, naming, identity, and conditional validation rules | Repeated statements across documentation and implementation are counted once. |
| workflow-generator | Request-schema and workflow-validation rules | Repeated schema descriptions are counted once; examples and response messages are excluded. |

| Source | Included content | Treatment |
|---|---|---|
| Workstation identifiers | One ID mapping per formal workstation | Reported separately because an identifier is metadata rather than a behavioral workflow constraint. |

## 2. Workstation, Operation, and Parameter Counts

**Table S3. Counts by formal workstation module.**

| Module | Workstations | Operations | Parameter tables | Parameter fields |
|---|---|---|---|---|
| Synthesis | 28 | 42 | 42 | 269 |
| Reaction and Testing | 5 | 6 | 6 | 88 |
| Characterization | 12 | 14 | 14 | 99 |
| Total | 45 | 62 | 62 | 456 |

## 3. Constraint Count

$$\textbf{Constraint units} = (6 \times 62) + 35 + 11 + 15 = 433$$

**Table S4. Reproducible constraint calculation.**

| Constraint layer | Count | Calculation basis | Interpretation |
|---|---|---|---|
| Operation input-output state fields | 372 | 62 operations x 6 fields | Input and output each contribute container type, container state, and sample state. |
| Workstation-local workflow rules | 35 | 29 bulleted entries + 6 plain-text entries | All non-empty entries under dedicated Overall Workflow Constraints sections. |
| Lab-operation system rules | 11 | 6 authorization + 2 naming + 2 identity + 1 conditional validation | Unique declared rules governing laboratory access and task handling. |
| Workflow-generator rules | 15 | 9 schema + 1 JSON + 1 token format + 4 workflow validation | Unique request and sequence requirements; examples are excluded. |
| Total | 433 | 372 + 35 + 11 + 15 | Behavioral and interface constraint units aligned to the formal workstation scope. |

The six interface fields are counted for every operation even when values repeat across workstations or contain an unrestricted value. Each occurrence governs a separate transition and is therefore retained. Only repeated documentation of the same system rule is consolidated.

## 4. Audit Directory Reconciliation

**Table S5. Raw audit directory versus the formal 45-workstation scope.**

| Measure | Raw audit directory | Matched formal scope | Reason for difference |
|---|---|---|---|
| Audit files | 46 | 45 | One audit file has no corresponding formal workstation skill. |

The audit directory contains 46 records because it includes separate entries for Liquid_Handling_Station_1ml_V2 and Liquid_Handling_Station_1ml_V3; because these workstations provide identical functionality, Liquid_Handling_Station_1ml_V3 was treated as a redundant implementation and was not counted as an independent formal workstation, resulting in 45 workstations.

## Note 2. Agent-visible skill architecture

Agents did not receive unrestricted instrument-control code. They interacted with the laboratory through three machine-readable skills, each with a distinct control responsibility. The chemistry-experiment-workstation skill exposed operations, parameters, container states, sample states and cross-step constraints. The workflow-generator skill compiled ordered steps or returned structured validation errors. The lab-operation skill selected the laboratory context, created tasks, queried task records and returned result metadata. Execution authority remained within laboratory services. This separation allowed an agent to propose and revise a workflow without bypassing schema, state, permission or dispatch checks. The detailed contents of the skills are available at https://github.com/pic-ai-robotic-chemistry/LabBench.

**Table S6.** Complete inventory of 45 agent-visible workstation contracts.

| Code | Workstation contract | Module | Exposed operation(s) | Agent-visible capability |
|---|---|---|---|---|
| 1832680407172096 | Centrifuge_V1 | Synthesis | Centrifuge and reset | Automated clarification, solid-liquid separation and precipitate collection by centrifugation. |
| 2359193660589057 | Cleaning_and_Dispensing_Workstation_V1 | Synthesis | Batch liquid addition | Parallel addition of as many as three liquids to multiple target vessels. |
| 1421325849887744 | Container storaging Station V1 | Synthesis | Equilibrate; place material | Equilibration, temporary storage and end-of-workflow placement of reaction vessels. |
| 2060999106102272 | Cooling_Workstation_V1 | Synthesis | Cool | Fan-assisted cooling of quartz well plates to a specified temperature above 40 degrees C. |
| 1213184425559040 | Drying_Oven_V1 | Synthesis | Dry or age | Drying, solvent removal and controlled ageing of samples at constant temperature. |
| 1427568512205824 | General Material Station V1 | Synthesis | Retrieve material vessel | Provision of sample vials for reactions below 80 degrees C. |
| 1931561198552064 | Heat_Resistant_Material_Station | Synthesis | Retrieve heat-resistant vessel | Provision and storage of 50-ml heat-resistant bottles for reactions at or above 80 degrees C. |
| 2270587971011585 | Heating_Magnetic_Stirring_Workstation_V1 | Synthesis | Heat and magnetically stir | Magnetic stirring with optional heating; 50-ml heat-resistant bottles are required above 80 degrees C. |
| 2103518483186688 | Intelligent Photocatalysis Container Transfer Station V1 | Synthesis | Transfer or place 96-well plate | Temporary placement and transfer of 96-well plates among photocatalysis-related stations. |
| 1418510906065920 | Liquid_Handling_Station_1ml_V1 | Synthesis | Uncap; transfer; recap; add liquid | Uncapping, recapping, addition of up to eight source solutions and transfer to 96-well plates. |
| 2002186824385539 | Liquid_Handling_Station_1ml_V2 | Synthesis | Uncap; recap; add liquid | Uncapping, recapping and addition of up to 16 source solutions under fixed source-bottle ordering. |
| 2013400103584768 | Liquid_Handling_Station_4Channel_V1 | Synthesis | Add liquid | Template-controlled addition of one to eight liquids through a four-channel platform. |
| 1435269745574912 | Liquid_Handling_Station_5ml_V1 | Synthesis | Add liquid | Addition of up to six source solutions, with transfers up to 5 ml and 30 ml per solution. |
| 2041147814970368 | Liquid_Handling_Station_5ml_V2 | Synthesis | Uncap; recap; transfer | Uncapping, recapping and vessel-to-vessel transfer between sample vials. |
| 2350309661279232 | Liquid Handling Station 5ml V3 | Synthesis | Add liquid to reaction tube | Addition of as many as eight liquids to 10-ml pressure-resistant reaction tubes. |
| 2342310067209216 | Liquid_Pouring_Workstation_V1 | Synthesis | Uncap; decant; recap | Automated decanting of supernatant after centrifugation while retaining the solid fraction. |
| 2030562790507520 | Muffle_Furnace_V1 | Synthesis | Calcine quartz well plate | Programmed calcination of 96-well quartz plates at temperatures up to 1,000 degrees C. |
| 1834679312417792 | Multi Channel Solid Weighing Workstation V1 | Synthesis | Parallel solid dispensing | High-precision parallel weighing and dispensing of as many as 30 solid materials. |
| 2347821874414592 | Multi_Channel_Solid_Weighing_Workstation_V2 | Synthesis | Parallel solid dispensing | Parallel weighing and dispensing of as many as ten solids into 10-ml pressure-resistant tubes. |
| 2112666776765440 | Plate_storaging_Station_V1 | Synthesis | Recover plate or membrane | Temporary storage and recovery of 96-well plastic plates and hydrogen-detection membranes. |
| 1327807622448128 | Purification_Workstation_V1 | Synthesis | Centrifuge and purify | Automated centrifugation, washing, supernatant removal and retention of purified solids. |
| 2114846484726785 | Room_Temperture_Magnetic_Stirrer_Workstation_V1 | Synthesis | Stir at room temperature | Room-temperature magnetic stirring for mixing and reaction pretreatment. |
| 1396669754016768 | Single_Channel_Solid_Weighing_Workstation_V1 | Synthesis | Single-channel solid dispensing | High-precision weighing and dispensing of one solid through a single channel. |
| 2347800490378240 | Solid_Sample_Transfer_Workstation_V1 | Synthesis | Transfer solid sample | Transfer of solid samples from vessels into hoppers used by downstream weighing units. |
| 1656241422599168 | Spectroscopy Magnetic Stirrer Workstation V1 | Synthesis | Place rack and mix samples | Parallel mixing before drop-casting samples for X-ray diffraction or infrared spectroscopy. |
| 1526864804185088 | Ultrasonic_Disperser_V1 | Synthesis | Sonicate | Ultrasonic dispersion, emulsification, cleaning, mixing and reaction or extraction assistance. |
| 1213123578463232 | Ultrasonic_Disperser_V2 | Synthesis | Sonicate | Ultrasonic dispersion, mixing, cleaning and reaction or purification assistance. |
| 2061540603593728 | Ultrasonic Liquid Handling Workstation V1 | Synthesis | Add liquid and sonicate; sonicate | Template-controlled liquid addition followed by sonication, or sonication alone. |
| 1390252863751168 | Dual_Station_Electrochemical_Workstation_V2 | Reaction and testing | Prepare electrode; electrochemical test | Dual-position electrochemical testing, electrode preparation and post-test liquid sampling. |
| 2109722046989319 | High_Temperature_High_Pressure_Microreaction_Platform_V1 | Reaction and testing | High-temperature, high-pressure reaction | Gas-liquid-solid reactions at up to 200 degrees C, 5 MPa and 6 ml per vessel. |
| 1690656219890688 | Photocatalysis_Workstation_V1 | Reaction and testing | Photocatalytic reaction | Parallel temperature-controlled photocatalysis using 20 white-light LED channels. |
| 2063934097720320 | Photocatalysis_Workstation_V2 | Reaction and testing | Photocatalytic reaction | Photocatalytic testing and photodeposition in a well-plate format. |

| Code | Workstation contract | Module | Exposed operation(s) | Agent-visible capability |
|---|---|---|---|---|
| 2347878269223936 | Post_Reaction_Processing_Platform_V1 | Reaction and testing | Post-process reaction | Positive-pressure filtration, dilution and transfer of thermal-reaction products for analysis. |
| 2061168712647680 | Darkbox Imaging Workstation V1 | Characterisation | Dark-box image acquisition | Colour and fluorescence imaging of samples in a dark enclosure. |
| 1653927312688128 | Fluorescence_Spectrometer_V1 | Characterisation | Measure fluorescence spectrum | Fluorescence-emission measurements of liquid samples with a minimum volume requirement. |
| 1710453843264512 | Gas_Chromatograph_V1 | Characterisation | Gas-chromatographic analysis | Qualitative and quantitative analysis of volatile or semi-volatile compounds by gas chromatography. |
| 1495725650150400 | Gas_Liquid_Mass_Transfer_High_Speed_Camera_V1 | Characterisation | High-speed multipoint imaging | High-frame-rate imaging of electrode-surface changes and gas-bubble dynamics during electrochemistry. |
| 1656282895188992 | Infrared_Spectrometer_V1 | Characterisation | Infrared spectroscopy | Drop-cast infrared spectroscopy of solid samples dispersed in ethanol. |
| 1557192287519744 | Interfacial_Wettability_and_Mass_Transfer_Characterization Workstation | Characterisation | Contact-angle and adhesion measurement | Contact-angle and bubble-adhesion measurements for supported solid samples. |
| 2081117937501185 | LED_Illumination_and_Membrane_Clamping_Workstation_V1 | Characterisation | Clamp membrane and illuminate; illuminate only | Membrane clamping and 365-nm or 420-nm illumination for photocatalytic hydrogen evaluation. |
| 1705412049896448 | Liquid_Chromatograph_V1 | Characterisation | Liquid-chromatographic analysis | Qualitative and quantitative liquid chromatography with automated method and calibration support. |
| 1696728620696576 | Microplate Reader V1 | Characterisation | Microplate measurement | High-throughput absorbance, fluorescence or chemiluminescence measurement of liquid samples. |
| 1654567694238720 | Spectroscopy_Container_Transfer_Station_V1 | Characterisation | Transfer or return analytical container | Temporary placement and transfer of containers before and after analytical measurements. |
| 1664783667397632 | UV–Vis_Spectrometer_V1 | Characterisation | Ultraviolet-visible spectroscopy | Ultraviolet-visible absorption spectroscopy for identification, quantification and reaction monitoring. |
| 1664870018155520 | XRD_V1 | Characterisation | X-ray diffraction | Drop-cast X-ray diffraction of solid samples dispersed in ethanol. |

## Note 3. Structure of a workstation contract

Each workstation contract begins with a stable identifier and capability description. Operation blocks then define admissible input and output states, typed parameters and any cross-step requirements.

Container type, lid state, sample state and vessel identity are explicit state variables. Nested parameter fields preserve the hierarchy required by workflow compilation.

File-based operations also prescribe template construction and upload rules. The following generalized structure retains the Chinese headings because they are literal markers in the source contracts.

```
---
## 工作站编码
工作站编码: <numeric workstation identifier>
name: <stable workstation key>
description: <capability, range and major exclusions>
---

# <Chinese workstation name>

## 整体流程约束
<cross-workstation prerequisites, ordering rules and downstream requirements>

## 相似工作站的区别
> <related workstation key>: <selection distinction>

## 操作 1. **<operation key>** <operation display name>

### 输入输出约束
 - **输入约束**
  - 容器类型: <allowed container types>
  - 容器状态: <required lid or physical state>
  - 样品状态: <required sample phase, quantity or prior state>
  - 前置工作站: <optional prerequisite workstation>
 - **输出约束**
  - 容器类型: <allowed output container types>
  - 容器状态: <resulting lid or physical state>
  - 样品状态: <resulting sample state>
  - 模板名称: <optional template requirement>

### 参数设置
| 参数名 | 备注 | 是否必填 | 单位 | 类型 | 示例值 | 默认值 |
|---|---|---|---|---|---|---|
| 容器类型 | <selection rule> | 是 | | string | <allowed values> | |
| 容器数量 | <quantity range> | 是 | | int | <range> | |
| 容器编号 | <indexing rule> | 是 | | array | [1,2] | |
| <parameter group> | <group meaning> | 是 | | object | | |
| -<nested field> | <nested rule> | 是 | <unit> | <type> | <range or enum> | <default> |

## 文件型参数构建
1. Read the prescribed template and preserve immutable fields.
2. Populate only fields allowed by the workstation contract.
3. Upload the parameter file through the laboratory file interface.
```

**Table S7.** Operations, parameters and state transitions for Liquid_Handling_Station_1ml_V2.

| Operation | Required input | Output | Principal parameter groups | Contract-level restrictions |
|---|---|---|---|---|
| Uncap | Sample vial or 50-ml heat-resistant bottle; capped | Same vessel; uncapped; sample unchanged | Container type, count and IDs; uncapping IDs; retain-cap flag | At most 10 target vessels |
| Material-bound liquid addition | Accepted target vessel; uncapped | Same vessel; uncapped; liquid or suspension | Target vessel fields; dispensing plan; source-bottle ID; material name; volume | At most 16 source solutions; <=1 ml per transfer; <=3 ml per solution in total |
| Recap | Sample vial or 50-ml heat-resistant bottle; uncapped | Same vessel; capped; sample unchanged | Container type, count and IDs; recapping IDs | At most 10 target vessels |
| Contract-wide | Source bottles assigned before addition | Source identity and order preserved | One source-bottle object for each solution | One solution per source bottle; use Liquid_Handling_Station_5ml_V1 above 3 ml per solution |

## Note 4. Task set and model-harness configuration matrix

The benchmark contained 32 single-turn scientific requests. Each prompt required a new experimental design, suitable controls, analytical returns, workflow compilation and a dispatch request.

Eight tasks represented each manuscript domain: electrocatalysis, electrochemical materials and testing, photocatalysis, and thermal catalysis. Every observed configuration received all tasks in three repeated trials.

The prompts specified scientific objectives but did not prescribe a unique workstation sequence. Agents had to infer preparation, controls, characterization, testing, product analysis and returned evidence from visible contracts.

Table S8 maps each task to its scientific objective, expected operation classes and principal requested comparison or constraint; verbatim prompts are supplied separately in data S1.

## Table S8. Scientific scope, expected operation classes and principal constraints for the 32 benchmark tasks.

| Task | Manuscript domain | Scientific objective | Expected operation classes | Principal requested comparison or constraint |
|---|---|---|---|---|
| A01 | Electrocatalysis | Effect of Fe Regulation in NiFe LDH Catalysts on Alkaline OER Performance | Synthesis; post-treatment; structural, valence and interfacial characterization; electrode preparation; OER testing | Fe-free and graded-Fe controls; compare structure, valence, activity and kinetics. |
| A02 | Electrocatalysis | Comparative Study of Activity-Origin Differences between CoOx and NiFe LDH in Alkaline OER | Parallel catalyst synthesis; post-treatment; structural and interfacial characterization; electrode preparation; OER testing | Comparable CoOx and NiFe LDH samples; compare activity, kinetics, stability and surface state. |
| B01 | Electrocatalysis | Effect of Mo Incorporation in NiMo-Based Catalysts on Alkaline HER Performance | Synthesis; post-treatment; structural and interfacial characterization; electrode preparation; HER testing | Mo-free and graded-Mo controls; compare composition, interfaces, activity and kinetics. |
| B02 | Electrocatalysis | Effect of Different Precursor Routes on the Structure and Performance of CoMo-Based HER Catalysts | Parallel synthesis routes; post-treatment; structural and valence characterization; electrode preparation; HER testing | Hold comparison conditions constant across precursor routes; resolve phase, morphology and elemental distribution. |
| C01 | Electrocatalysis | Experimental Design for Improving UOR Yield Using Ni-Based Catalysts | Catalyst synthesis; post-treatment; characterization; electrode preparation; UOR testing; liquid-product analysis | Vary composition, potential or time; quantify urea consumption and target-product yield with controls. |
| C02 | Electrocatalysis | Comparative Study of Reaction Behavior Differences of NiCo Catalysts under OER and UOR Conditions | Unified catalyst synthesis and electrode preparation; characterization; OER and UOR testing; interfacial observation | Use the same NiCo catalyst; compare electrolytes with and without urea, valence changes and bubble behaviour. |
| D01 | Electrocatalysis | Experimental Design for Improving EOR Yield, Selectivity, and Catalyst Lifetime Using Co-Based Catalysts | Catalyst synthesis; post-treatment; characterization; electrode preparation; EOR testing; stability and liquid-product analysis | Vary composition, potential or time; compare conversion, selectivity, products and catalyst lifetime. |
| D02 | Electrocatalysis | Comparative Study of Active-Site Differences between Ni-Based and Co-Based Catalysts in Alkaline EOR | Parallel catalyst synthesis; characterization; electrode preparation; EOR testing; liquid-product analysis | Use comparable Ni- and Co-based catalysts and test conditions; compare surface state, activity, stability and products. |
| E01 | Electrochemical materials and testing | Investigation of Ion Storage Capability of PBA Cathode Materials in Aqueous Potassium-Ion Systems | PBA synthesis; post-treatment; structural characterization; electrode preparation; cyclic voltammetry and impedance testing | Resolve $K^+$ insertion and extraction through redox peaks, peak currents and interfacial impedance. |
| E02 | Electrochemical materials and testing | Investigation of Reversible Zn Storage Capability of $MnO_2$ Cathode Materials in Aqueous Zinc-Ion Systems | $MnO_2$ synthesis; post-treatment; structural characterization; electrode preparation; cyclic voltammetry and impedance testing | Include controls and repeated scans; assess $Zn^{2+}$-coupled reversibility, peak separation and impedance. |
| E03 | Electrochemical materials and testing | Effect of Multi-Metal Composition Regulation on the Aqueous Electrochemical Behavior of PBA Cathode Materials | Parallel PBA synthesis; post-treatment; structural, compositional and valence characterization; electrode preparation; electrochemical testing | Compare undoped or low-component PBA with multi-metal PBA under matched conditions. |
| E04 | Electrochemical materials and testing | Effect of Post-Treatment on the Structural Stability and Electrochemical Reversibility of V-Based Oxide Electrodes | Synthesis; controlled post-treatment; structural and surface characterization; electrode preparation; repeated voltammetry and impedance testing | Compare post-treatment conditions; evaluate structural stability, peak shifts and reversibility. |
| E05 | Electrochemical materials and testing | Study on Improving the Structural Stability of V-Based Cathode Materials in Aqueous Zinc-Ion Batteries | Material synthesis and modification; post-treatment; structural and interfacial characterization; electrode preparation; electrochemical testing | Compare modification or treatment strategies for Zn-storage stability and cycling reversibility. |
| E06 | Electrochemical materials and testing | Effect of Defects and Crystal Water on Ion-Transport Kinetics of PBA Cathode Materials | PBA synthesis; controlled post-treatment; defect and crystal-water characterization; electrode preparation; electrochemical testing | Generate matched defect or water-content controls; compare transport kinetics, peaks and impedance. |
| E07 | Electrochemical materials and testing | Comparison of Electrochemical Reversibility of Different Inorganic Anode Materials in Aqueous Systems | Comparable synthesis and post-treatment of two anodes; structural characterization; electrode preparation; repeated voltammetry and impedance testing | Hold preparation and testing conditions constant; compare reversibility, impedance and current decay. |
| E08 | Electrochemical materials and testing | Effect of Different Synthesis Processes on the Structure and Electrochemical Behavior of Aqueous Electrode Materials | Parallel synthesis routes; post-treatment; structural, compositional and valence characterization; electrode preparation; electrochemical testing | Use one inorganic electrode material; isolate synthesis-route effects under matched testing conditions. |
| F01 | Photocatalysis | Preparation and Basic Evaluation of Metal Oxide-Based Photocatalysts for the Hydrogen-Evolution Half-Reaction | Photocatalyst synthesis; thermal treatment; cocatalyst loading; photocatalytic testing; hydrogen analysis | Operate under air; define the oxidation half-reaction and quantify hydrogen with appropriate controls. |
| F02 | Photocatalysis | Reaction Evaluation for Photocatalytic Oxidative Coupling of Amines to Imines with Concomitant $H_2O_2$ Production | Catalyst preparation; post-treatment; photocatalysis; organic-product and $H_2O_2$ analysis; control and validation tests | Compare material-tuning strategies; quantify conversion, imine selectivity and $H_2O_2$ formation and decomposition. |
| F03 | Photocatalysis | Photocatalytic Selective Oxidation of Ethanol to 1,1-Diethoxyethane | Catalyst preparation; post-treatment; photocatalysis; quantitative product analysis; control and validation tests | Compare material variables; resolve ethanol conversion, acetal selectivity and acetaldehyde intermediates. |
| F04 | Photocatalysis | Verification of the Origin of Enhanced Activity in Carbon Nitride-Based Photocatalytic Amine Oxidative Coupling with $H_2O_2$ Generation | Control-sample preparation; photocatalysis; product and $H_2O_2$ analysis; reactive-species and mechanism validation | Discriminate structural, cocatalyst, $H_2O_2$ and reactive-species explanations for enhanced activity. |
| F05 | Photocatalysis | Selective Photocatalytic Oxidation of Organic Sulfides over Bismuth Oxyhalide-Based Photocatalysts | Catalyst preparation; structure and defect characterization; photocatalysis; quantitative product analysis; trapping and recycling tests | Compare composition, facets, vacancies or single atoms; quantify sulfoxide selectivity and overoxidation. |

| Task | Manuscript domain | Scientific objective | Expected operation classes | Principal requested comparison or constraint |
|---|---|---|---|---|
| F06 | Photocatalysis | Evaluation of MOF-Derived Photocatalysts for C–N Coupling Reactions | MOF synthesis and derivatization; pore and valence characterization; photocatalysis; product analysis; blank controls | Vary precursor, pyrolysis atmosphere, metal node or pore structure; include dark and catalyst-free controls. |
| F07 | Photocatalysis | Photocatalytic Hydrodehalogenation of Haloarenes over Perovskite Quantum Dots | Quantum-dot synthesis; ligand treatment; optical and stability characterization; inert-atmosphere photocatalysis; product and radical analysis | Control oxygen exposure; vary composition, ligand, size or donor; include blank and radical-trapping tests. |
| F08 | Photocatalysis | Photocatalytic Oxidative Hydroxylation of Arylboronic Acids over Covalent Organic Frameworks | COF synthesis; structural and optical characterization; photocatalysis; product analysis; isotope or source validation; scope and cycling tests | Vary linkage, donor-acceptor units, crystallinity or pores; verify oxygen source and reactive species. |
| G01 | Thermal catalysis | Selective Hydrogenation of Multifunctional Compounds | Composite-catalyst synthesis and optimization; characterization; thermal hydrogenation; product and recycling analysis | Use multifunctional substrates; quantify activity, chemoselectivity and cyclic stability; relate structure to performance. |
| G02 | Thermal catalysis | Optimization of Catalytic Activity for Selective Hydrogenation of Aromatic Hydrocarbons | Ni-catalyst synthesis; characterization; thermal hydrogenation; condition optimization; product-distribution analysis | Systematically vary temperature, hydrogen pressure and time while tracking activity and selectivity. |
| G03 | Thermal catalysis | Application of Hydrogenases in Asymmetric Catalytic Hydrogenation | Hydrogenase selection and screening; asymmetric hydrogenation; condition optimization; activity and stereoselectivity analysis | Vary temperature and substrate concentration; compare enzyme activity, selectivity and enantioselectivity. |
| G04 | Thermal catalysis | Effect of Acid Sites on Catalytic Hydrogenation Performance | Catalyst preparation; acid-site characterization; thermal hydrogenation; activity and selectivity analysis | Generate catalysts with different acid-site concentrations and isolate acid-site effects. |
| G05 | Thermal catalysis | Application of Multistep Tandem Catalytic Hydrogenation Reactions | Composite-catalyst synthesis; characterization; tandem hydrogenation; condition optimization; product analysis | Target 1,4-cyclohexanedimethanol; vary activity, temperature and hydrogen pressure; resolve product distribution. |
| G06 | Thermal catalysis | Synergistic Selective Hydrogenation Using Homogeneous-Heterogeneous Catalysts | Homogeneous and heterogeneous catalyst preparation; hybrid reaction design; condition optimization; product analysis | Separate single-component controls from the hybrid system to test synergy in activity and selectivity. |
| G07 | Thermal catalysis | Catalytic Conversion of CO2 to Formic Acid via Hydrogenation | Metal-catalyst synthesis; characterization; CO2 hydrogenation; condition optimization; quantitative product analysis | Compare metals under matched conditions; quantify CO2 conversion, formic-acid selectivity and efficiency. |
| G08 | Thermal catalysis | Hydrogenation of NAD+ to 1,4-NADH Coupled with Enzymatic Asymmetric Reactions | Catalyst and enzyme screening; NAD+ hydrogenation; cofactor stability analysis; coupled asymmetric biocatalysis | Quantify 1,4-NADH generation and stability; vary catalyst, enzyme and conditions; measure yield and selectivity. |

*Note: Expected operation classes define rubric-level coverage implied by each prompt. They are not a unique workstation binding or a predetermined correct workflow. Verbatim prompts are supplied in data S1.*

## Table S9. Agent harnesses and models included in the configuration matrix.

| Entity | Internal key | Manuscript display label | Role in the evaluation |
|---|---|---|---|
| Harness | claude-code | Claude Code | Agent harness |
| Harness | codex | Codex | Agent harness |
| Harness | gemini-cli | Gemini CLI | Agent harness |
| Harness | hermes | Hermes | Agent harness |
| Harness | kilo-code | Kilo Code | Agent harness |
| Harness | openclaw | OpenClaw | Agent harness |
| Model | claude47opus_or | Claude Opus 4.7 | Frontier model |
| Model | gpt55_or | GPT 5.5 | Frontier model |
| Model | gemini35flash_or | Gemini 3.5 Flash | Frontier model |
| Model | deepseek_v4pro_direct | DeepSeek V4 Pro | Frontier model |
| Model | glm51_bailian | GLM 5.1 | Frontier model |
| Model | kimi_k26 | Kimi K2.6 | Frontier model |
| Model | mimo_v25_pro_xiaomi | MiMo v2.5 Pro | Frontier model |
| Model | minimax_m27 | MiniMax M2.7 | Frontier model |
| Model | qwen37max_bailian | Qwen 3.7 Max | Frontier model |

The nine specific model identifiers listed in table S9 were held fixed throughout data collection.

The evaluated harness releases were Claude Code 2.1.169, Codex 0.138.0, Gemini CLI 0.43.0, Hermes Agent v0.16.0 (5 June 2026), Kilo Code 7.3.1 and OpenClaw 2026.4.15. For each model, we used the highest reasoning level that it supported and that was available through the selected harness; context windows followed the official model-card limits. Each trial was limited to 30 min and ran in automatic tool-permission mode with unrestricted tool access. Trials disrupted by an orchestration-framework, network or API failure were discarded and rerun in full from the initial prompt.

## Table S10. Expert assessment rules for experimental workflow executability.

| No. | Assessment dimension | Executable workflow (pass criterion) | Typical grounds for a non-executable rating |
| --- | --- | --- | --- |
| 1 | **Task-workflow alignment** | The task name, stated objective, and selected experimental operations address the same scientific question. | The task name or workflow does not match the prompt; the experiment is unrelated to the requested problem, duplicated from another task, or has no interpretable scientific purpose. |
| 2 | **Workflow completeness** | The sequence includes every essential preparation, reaction or treatment, separation, measurement, recovery, and storage step required by the objective. The terminal placement step is present unless the workflow is a testing-only scenario. | The workflow contains only material pickup or transfer, a single isolated operation, or omits synthesis, post-treatment, testing, sample recovery, removal from equipment, or the required terminal placement. |
| 3 | **Workstation and operation validity** | Every workstation exists in the laboratory inventory, supports the requested function and container, and uses the registered station ID, template, and operation name. | A nonexistent, virtual, manual, or functionally incorrect workstation is used; an unsupported operation is assigned to a station; or the required test/template/file is missing or incorrect. |
| 4 | **Parameter and schema compliance** | All execution-critical fields are present and correctly structured, including container type, count and IDs; sample-carrier IDs; reagent identity and volume; time, temperature and speed; and required parameter files. Values and units fall within workstation-defined ranges or options. | Mandatory fields are empty; bottle, plate, carrier, or reaction-tube IDs are missing or malformed; liquid composition or dosing parameters are incomplete; values are outside allowed ranges; or the machine-readable step structure is invalid. |
| 5 | **Container and sample traceability** | Container type, identity, count, and sample mapping remain traceable across the workflow. Any change is performed through an explicit supported transfer, and container count does not increase without a legitimate source. | Container type or numbering changes without a transfer; counts mismatch or increase; a container is recovered mid-workflow and later reused without reacquisition; or a recovery/storage operation lacks a target container. |
| 6 | **Lid-state compatibility** | Open/closed status is tracked at each step, lid transitions are performed only by a lid-capable workstation, and the state entering each workstation matches its input requirement. Multiwell plates are treated as lidless unless explicitly supported. | Liquid or solid is added to a closed vessel; a lid-required operation receives an open vessel; a 96-well plate is opened or closed; lid and vessel counts disagree; or an unsupported station is used for lid handling. |
| 7 | **Interstation I/O and sequence continuity** | The container, lid, and sample state output by each step satisfies the next step's input constraints. Mandatory transfer, mixing, preparation, and return steps are inserted in the prescribed order. | A spectroscopy or container-transfer step is missing; XRD/IR preparation steps are reversed; consecutive operations cannot be robotically connected; or redundant pickup, lid, or transfer steps interrupt the valid sequence. |
| 8 | **Scientific and material-state logic** | Reagents, sample state, unit operations, and measurements form a scientifically coherent chain. Required solvent, catalyst, mixing, solid-liquid separation, derivatization, washing, or other post-treatment is included before the dependent operation. | An empty vessel is processed; solvent or catalyst is omitted; a suspension is dosed without mixing; liquid is added immediately after centrifugation without removing supernatant; or analysis is attempted before required separation or derivatization. |
| 9 | **Capacity, operating limits, and safety** | Container material and format, sample number, loading balance, temperature, time, volume, and parallel resource allocation comply with workstation capacity and safety limits. | A plastic plate is placed in a muffle furnace; an ordinary vial is heated beyond a safe range; an odd or unbalanced centrifuge load is used; a cooling set point is ineffective; or lamp banks, reactors, or analyzers are allocated incompatibly or excessively. |
| 10 | **Technical integrity and assessability** | The workflow record and all referenced templates/files are retrievable, readable, and sufficiently complete for an expert to verify the physical and digital execution path. | The task cannot be opened or retrieved; a required parameter file is inaccessible; unknown/unparsed steps remain; or the record is too incomplete to establish an executable path. |

Note. The rules were synthesized from 607 expert-adjudicated verified-dispatch trial rows (151 executable and 456 non-executable), representing 601 unique task identifiers, and cross-checked against the lab-design-main workstation, workflow-generation and laboratory-operation constraints. Executability was assessed conjunctively: a critical violation of any applicable rule could support a non-executable rating. Context-dependent platform annotations were not treated as universal failures unless they changed the physical or machine-readable execution path.

## Table S11. Fixed-denominator executable-workflow rate for each model–harness configuration (manually executable trial rows divided by 96 assigned trials; %).

| model \ harness | Claude Code | Codex | Gemini CLI | Hermes | Kilo Code | OpenClaw |
|---|---|---|---|---|---|---|
| Claude Opus 4.7 | 28.1 | design absent | design absent | 0.0 | 0.0 | 3.1 |
| GPT 5.5 | design absent | 19.8 | design absent | 16.7 | 2.1 | 4.2 |
| Gemini 3.5 Flash | design absent | design absent | 5.2 | 1.0 | 5.2 | 8.3 |
| DeepSeek V4 Pro | 3.1 | 0.0 | 1.0 | 3.1 | 1.0 | 0.0 |
| GLM 5.1 | 0.0 | 0.0 | 2.1 | 6.3 | 0.0 | 3.1 |
| Kimi K2.6 | 0.0 | 2.1 | 1.0 | 0.0 | 2.1 | 1.0 |
| MiMo V2.5 Pro | 1.0 | 4.2 | 0.0 | 4.2 | 4.2 | 1.0 |
| MiniMax M2.7 | 0.0 | 1.0 | 1.0 | 0.0 | 2.1 | 0.0 |
| Qwen 3.7 Max | 0.0 | 1.0 | 0.0 | 9.4 | 1.0 | 7.3 |

Note. Each cell is the manually executable trial-row count divided by the 96 trials assigned to that model–harness configuration. All 607 verified-dispatch rows were adjudicated; non-dispatched trials and dispatched rows adjudicated as not executable contribute zero to this end-to-end rate. The conditional executable fraction among dispatched trials is not shown.

## Table S12. Manually executable trial-row count for each model–harness configuration.

| model \ harness | Claude Code | Codex | Gemini CLI | Hermes | Kilo Code | OpenClaw |
|---|---|---|---|---|---|---|
| Claude Opus 4.7 | 27 | design absent | design absent | 0 | 0 | 3 |
| GPT 5.5 | design absent | 19 | design absent | 16 | 2 | 4 |
| Gemini 3.5 Flash | design absent | design absent | 5 | 1 | 5 | 8 |
| DeepSeek V4 Pro | 3 | 0 | 1 | 3 | 1 | 0 |
| GLM 5.1 | 0 | 0 | 2 | 6 | 0 | 3 |
| Kimi K2.6 | 0 | 2 | 1 | 0 | 2 | 1 |
| MiMo V2.5 Pro | 1 | 4 | 0 | 4 | 4 | 1 |
| MiniMax M2.7 | 0 | 1 | 1 | 0 | 2 | 0 |
| Qwen 3.7 Max | 0 | 1 | 0 | 9 | 1 | 7 |

## Table S13. Metadata for the representative Claude Code-Claude Opus 4.7 executable trajectory.

| field | value |
|---|---|
| task | F02 |
| harness | claude-code |
| model_key | claude47opus_or |
| model | anthropic/claude-opus-4.7 |
| verified task_id | 2559827857768449 |
| workflow template_id | 2559827685113856 |
| GPT-primary scores | R1=58, R2=86, R3=77, weighted semantic=69.75 |
| duration_seconds | 368.221978 |
| total_tokens_for_stats | 2375130 |

### Note 5. Representative agent trajectory

The trajectory is the highest finite GPT-scoring Claude Code-Claude Opus 4.7 trial linked to a manually executable annotation. It addressed task F02 and received R1 = 58, R2 = 86 and R3 = 77.

The weighted semantic score was 69.75. Model-visible messages are reproduced verbatim and in chronological order; tool calls and tool results are semantically summarized for readability.

The session record contains no separate hidden-thinking field. No unavailable or unobserved reasoning has been reconstructed.

### Session record

**[Step 1 | 2026-06-30T05:50:48.081Z | USER]**
Reaction Evaluation for Photocatalytic Oxidative Coupling of Amines to Imines with Concomitant H2O2 Production

To evaluate the ability of carbon nitride-based composite semiconductor photocatalysts to promote the oxidative coupling of organic amines to imine products under an air atmosphere, design an experimental protocol. The experiment may investigate material-tuning strategies such as calcination temperature, alkali-metal doping, or photodeposited cocatalysts, and compare their effects on substrate conversion, imine-product selectivity, and H2O2 formation/accumulation behavior. The protocol should cover catalyst preparation, necessary post-treatment, photocatalytic reaction testing, analysis of organic products and H2O2, H2O2 decomposition controls, and necessary validation experiments. Existing executable experimental templates already available on the platform must not be directly reused for dispatch. Dispatch the experiment to Laboratory 303 and return the ID of the experimental task.

Write the final executable JSON plan to `/workspace/experiment_plan.json`.
Copy the same JSON to `/logs/artifacts/final_plan.json`.

**[Step 2 | 2026-06-30T05:50:51.567Z | MODEL]**
I'll start by reading the relevant skills to understand how to properly design and dispatch this experiment.

> **[Step 3 | 2026-06-30T05:50:51.568Z | TOOL CALL: Skill]**
> Activated skill `lab-operation`.
>
> [TOOL RESULT SUMMARY]
> Tool returned structured output to the model.
>
> **[Step 4 | 2026-06-30T05:50:51.577Z | TOOL RESULT INSERTED INTO CONTEXT]**
> Skill documentation or command output was inserted into the model context; omitted here because it is not model output.

**[Step 5 | 2026-06-30T05:50:55.715Z | MODEL]**
I'll read the skill details and other relevant skills to understand the full workflow.

> **[Step 6 | 2026-06-30T05:50:55.716Z | TOOL CALL: Read]**
> Read supporting skill documentation.
>
> [TOOL RESULT SUMMARY]
> Tool returned structured output to the model.
>
> **[Step 7 | 2026-06-30T05:50:58.836Z | TOOL RESULT INSERTED INTO CONTEXT]**
> Skill documentation or command output was inserted into the model context; omitted here because it is not model output.
>
> **[Step 8 | 2026-06-30T05:50:58.825Z | TOOL CALL: Skill]**
> Activated skill `chemistry-experiment-workstation`.
>
> [TOOL RESULT SUMMARY]
> Tool returned structured output to the model.
>
> **[Step 9 | 2026-06-30T05:51:02.495Z | TOOL CALL: Bash]**
> List chemistry workstation skill directory.
>
> [TOOL RESULT SUMMARY]
> Tool returned file or command output to the model.
>
> **[Step 10 | 2026-06-30T05:51:06.385Z | TOOL CALL: Bash]**
> List available workstations.
>
> [TOOL RESULT SUMMARY]
> Tool returned file or command output to the model.

**[Step 11 | 2026-06-30T05:51:11.020Z | TOOL CALL: Bash]**
Find macro operation directory.

[TOOL RESULT SUMMARY]
Tool returned file or command output to the model.

**[Step 12 | 2026-06-30T05:51:14.611Z | TOOL CALL: Read]**
Read supporting skill documentation.

[TOOL RESULT SUMMARY]
Tool returned file or command output to the model.

**[Step 13 | 2026-06-30T05:51:19.865Z | MODEL]**
Now let me check the lab and existing templates to confirm we're not duplicating templates.

**[Step 14 | 2026-06-30T05:51:19.866Z | TOOL CALL: Bash]**
Get current lab label.

[TOOL RESULT SUMMARY]
Tool result returned the current laboratory label `303Lab`.

**[Step 15 | 2026-06-30T05:51:23.820Z | TOOL CALL: Bash]**
List existing templates.

[TOOL RESULT SUMMARY]
Tool returned structured output to the model.

**[Step 16 | 2026-06-30T05:51:27.563Z | TOOL CALL: Bash]**
List templates without size.

[TOOL RESULT SUMMARY]
Tool returned structured output to the model.

**[Step 17 | 2026-06-30T05:51:30.999Z | TOOL CALL: Read]**
Read supporting skill documentation.

[TOOL RESULT SUMMARY]
Tool returned file or command output to the model.

**[Step 18 | 2026-06-30T05:51:34.863Z | TOOL CALL: Bash]**
Find adapter_data calls.

[TOOL RESULT SUMMARY]
Tool returned file or command output to the model.

**[Step 19 | 2026-06-30T05:51:38.705Z | TOOL CALL: Read]**
Read supporting skill documentation.

[TOOL RESULT SUMMARY]
Tool returned file or command output to the model.

**[Step 20 | 2026-06-30T05:51:46.066Z | MODEL]**
I need to fix the adapter_data signature to accept add_data argument.

**[Step 21 | 2026-06-30T05:51:46.073Z | TOOL CALL: Edit]**
Edited a trial-local helper or request file.

[TOOL RESULT SUMMARY]
Tool returned structured output to the model.

**[Step 22 | 2026-06-30T05:51:50.050Z | TOOL CALL: Bash]**
List existing templates.

[TOOL RESULT SUMMARY]
Tool returned structured output to the model.

**[Step 23 | 2026-06-30T05:51:55.372Z | MODEL]**
No existing template matches "photocatalytic oxidative coupling of amines to imines with H2O2 production". I need to design a new workflow. Let me check the relevant photocatalysis workstation skills.

**[Step 24 | 2026-06-30T05:51:55.373Z | TOOL CALL: Read]**
Read the `Photocatalysis_Workstation_V1` workstation or skill description.

[TOOL RESULT SUMMARY]
Tool returned file or command output to the model.

**[Step 25 | 2026-06-30T05:52:00.227Z | TOOL CALL: Read]**
Read the `Photocatalysis_Workstation_V2` workstation or skill description.

[TOOL RESULT SUMMARY]
Tool returned file or command output to the model.

**[Step 26 | 2026-06-30T05:52:04.873Z | TOOL CALL: Bash]**
List directory contents.

[TOOL RESULT SUMMARY]
Tool returned file or command output to the model.

**[Step 27 | 2026-06-30T05:52:08.200Z | TOOL CALL: Read]**
Read the `Muffle_Furnace_V1` workstation or skill description.

[TOOL RESULT SUMMARY]
Tool returned file or command output to the model.

**[Step 28 | 2026-06-30T05:52:14.245Z | TOOL CALL: Read]**
Read the `General_Material_Station_V1` workstation or skill description.

[TOOL RESULT SUMMARY]

Tool returned file or command output to the model.

**[Step 29 | 2026-06-30T05:52:17.890Z | TOOL CALL: Read]**
Read the `Container_storaging_Station_V1` workstation or skill description.

[TOOL RESULT SUMMARY]
Tool returned file or command output to the model.

**[Step 30 | 2026-06-30T05:52:22.019Z | TOOL CALL: Read]**
Read the `Multi_Channel_Solid_Weighing_Workstation_V1` workstation or skill description.

[TOOL RESULT SUMMARY]
Tool returned file or command output to the model.

**[Step 31 | 2026-06-30T05:52:25.772Z | TOOL CALL: Read]**
Read the `Liquid_Handling_Station_5ml_V2` workstation or skill description.

[TOOL RESULT SUMMARY]
Tool returned file or command output to the model.

**[Step 32 | 2026-06-30T05:52:30.738Z | TOOL CALL: Read]**
Read the `Liquid_Handling_Station_1ml_V1` workstation or skill description.

[TOOL RESULT SUMMARY]
Tool returned file or command output to the model.

**[Step 33 | 2026-06-30T05:52:37.766Z | TOOL CALL: Bash]**
List multiple skill dirs.

[TOOL RESULT SUMMARY]
Tool returned file or command output to the model.

**[Step 34 | 2026-06-30T05:52:41.538Z | TOOL CALL: Read]**
Read the `Liquid_Chromatograph_V1` workstation or skill description.

[TOOL RESULT SUMMARY]
Tool returned file or command output to the model.

**[Step 35 | 2026-06-30T05:52:46.128Z | TOOL CALL: Read]**
Read the `UV_Vis_Spectrometer_V1` workstation or skill description.

[TOOL RESULT SUMMARY]
Tool returned file or command output to the model.

**[Step 36 | 2026-06-30T05:52:56.948Z | TOOL CALL: Read]**
Read the `Spectroscopy_Container_Transfer_Station_V1` workstation or skill description.

[TOOL RESULT SUMMARY]
Tool returned file or command output to the model.

**[Step 37 | 2026-06-30T05:53:01.863Z | TOOL CALL: Read]**
Read the `Heating_Magnetic_Stirring_Workstation_V1` workstation or skill description.

[TOOL RESULT SUMMARY]
Tool returned file or command output to the model.

**[Step 38 | 2026-06-30T05:53:06.085Z | TOOL CALL: Read]**
Read the `Centrifuge_V1` workstation or skill description.

[TOOL RESULT SUMMARY]
Tool returned file or command output to the model.

**[Step 39 | 2026-06-30T05:53:10.268Z | TOOL CALL: Read]**
Read the `Drying_Oven_V1` workstation or skill description.

[TOOL RESULT SUMMARY]
Tool returned file or command output to the model.

**[Step 40 | 2026-06-30T05:53:15.554Z | MODEL]**
Let me consider the design carefully. The task is quite complex but we only have 10 vials maximum at photocatalysis V1 (西林瓶). Given workflow constraints, I'll use a streamlined design that focuses on the key experimental variables. Let me check the post-reaction processing and a couple of other relevant stations.

**[Step 41 | 2026-06-30T05:53:15.555Z | TOOL CALL: Read]**
Read the `Post_Reaction_Processing_Platform_V1` workstation or skill description.

[TOOL RESULT SUMMARY]
Tool returned file or command output to the model.

**[Step 42 | 2026-06-30T05:53:27.337Z | MODEL]**
Given complexity, I'll design a streamlined photocatalytic test workflow. The key design uses 西林瓶 (vials) since Photocatalysis_V1 requires them. Container type must stay consistent, so I'll use 进样瓶 mostly, but Photocatalysis_V1 requires 西林瓶. Let me re-read carefully...

Actually, the container constraint requires "整个实验流程中，每个实验步骤的容器类型都必须一致". So I need one consistent type. Photocatalysis V1 requires 西林瓶, so we must use 西林瓶 throughout. But 西林瓶 doesn't go through liquid chromatograph (liquid 54 孔板). So we'd need a different characterization or post-processing.

Let me re-read - 容器置放平台_V1 allows 西林瓶, 移液平台_5ml_V2 only supports 进样瓶 for 纯移液, 移液平台_1ml_V1 supports 96 well plates.

Looking at it: Photocatalysis V1 requires 西林瓶, only allows max 10. UV_Vis spectrometer requires 进样瓶或 50ml 耐热瓶. So container type incompatibility.

Wait the rule says container type must be consistent throughout experiment. Let me reconsider. The Photocatalysis V2 uses 96-well plates. Let me use that to avoid the constraint issue. But then we need a different approach. Hmm.

Actually re-reading: "整个实验流程中，每个实验步骤的"容器类型"都必须一致" - this seems impossible if some steps demand a specific container. Let me reconsider - in many existing templates, multiple containers are tracked. I think the rule means at each step, the container types match between steps. Let me design with 西林瓶 throughout and use the spectroscopy container transfer + something. But UV-Vis can take 西林瓶? No, it says 进样瓶或 50ml 耐热瓶.

Let me look at additional spectroscopy options or use a different approach. Let me look at Microplate_Reader_V1 and Fluorescence which might accept the same container.

> **[Step 43 | 2026-06-30T05:53:27.339Z | TOOL CALL: Read]**
> Read the `Microplate_Reader_V1` workstation or skill description.
>
> [TOOL RESULT SUMMARY]
> Tool returned file or command output to the model.

**[Step 44 | 2026-06-30T05:53:34.956Z | MODEL]**
Given complexity, I'll re-examine the constraint. Actually looking at the example more carefully - it shows different workstations using different containers (进样瓶 in step 1-2, 气相 54 孔板 in step 3). So the container consistency must mean within a logical group (catalyst slurry tracking). The rule "容器数量不能多" means we shouldn't reach into new vial categories mid-experiment if not transferred. Let me just design a sensible workflow consistent with example structures.

I'll design using 西林瓶 for the photocatalytic reaction (Photocatalysis_V1) since that's the natural choice, and use UV-Vis... but UV-Vis needs 进样瓶. Let me look at a probe template to see structure used.

Let me look at a probe photocatalysis template to understand approach better:

> **[Step 45 | 2026-06-30T05:53:34.957Z | TOOL CALL: Bash]**
> Check param to query template.
>
> [TOOL RESULT SUMMARY]
> Tool returned file or command output to the model.

**[Step 46 | 2026-06-30T05:53:39.175Z | MODEL]**
Let me check the workflow-generator skill to understand the requirements better:

> **[Step 47 | 2026-06-30T05:53:39.175Z | TOOL CALL: Skill]**
> Activated skill `workflow-generator`.
>
> [TOOL RESULT SUMMARY]
> Tool returned structured output to the model.
>
> **[Step 48 | 2026-06-30T05:53:39.183Z | TOOL RESULT INSERTED INTO CONTEXT]**
> Skill documentation or command output was inserted into the model context; omitted here because it is not model output.

**[Step 49 | 2026-06-30T05:53:49.044Z | MODEL]**
Now I have all the necessary context. Let me design the workflow. Given the complexity of the original task, I'll create a focused experimental plan that exercises the key elements: catalyst preparation (post-treatment via muffle furnace), photocatalytic reaction (Photocatalysis_V1), and analysis.

Given the strict constraints, I'll design a smaller but valid experiment focused on this evaluation. Strategy:
- 4 catalyst conditions (e.g., pristine CN, calcined CN, K-doped CN, Pt-photodeposited CN composite); + 1 dark control + 1 H2O2 decomposition control = 6 vials (西林瓶)
- Workflow: material take → solid weighing → liquid add → photocatalysis → spectrum analysis

But constraint: photocatalysis_V1 requires 西林瓶 with no lid, while liquid handling station, weighing, photocatalysis all support 西林瓶.

UV-Vis requires 进样瓶 or 50ml 耐热瓶. Microplate reader requires plates. The container constraint creates issues, so I'll keep 西林瓶 and skip UV-Vis. Use 容器置放 as the final step (per rules).

Let me also use Heating_Magnetic_Stirring for post-treatment after light, and the final step is container storage.

Given 西林瓶 limit of 10, and consistency: I'll design a 4-sample comparative study covering material-tuning effects.

Let me build the JSON plan:

> **[Step 50 | 2026-06-30T05:53:49.044Z | TOOL CALL: Bash]**
> Ensure output directories exist.
>
> [TOOL RESULT SUMMARY]
> Tool returned file or command output to the model.
>
> **[Step 51 | 2026-06-30T05:53:53.456Z | TOOL CALL: Read]**
> Read the `Liquid_Handling_Station_5ml_V1` workstation or skill description.
>
> [TOOL RESULT SUMMARY]
> Tool returned file or command output to the model.

**[Step 52 | 2026-06-30T05:54:46.325Z | MODEL]**
Notice: 移液平台_5ml_V1 only supports 进样瓶 or 50ml 耐热瓶 — not 西林瓶. So I cannot use 5ml for 西林瓶.

Let me check 1ml options:

The 1ml_V1 supports 西林瓶? Re-read: "进样瓶或 96 位塑料孔板或西林瓶或 50ml 耐热瓶或 96 位石英孔板". Yes 1ml_V1 supports 西林瓶 for open/close/add operations.

So plan:
- General_Material_Station: take 西林瓶 (with lid)
- Liquid_Handling_Station_1ml_V1: 开盖 西林瓶
- Multi_Channel_Solid_Weighing: 称量 西林瓶 (catalysts)
- Liquid_Handling_Station_1ml_V1: 加液_物料绑定 (amine substrate + solvent, ≤3mL each)
- Liquid_Handling_Station_1ml_V1: 关盖
- Photocatalysis V1 needs 无盖 — so 开盖 again
- Photocatalysis V1: irradiation
- Spectroscopy container transfer station... but it's 容器置放平台_V1 at end

For analysis: can use Microplate_Reader? But that requires 96-well plates. Let me skip analysis steps and indicate sampling step. Actually the task allows me to design protocol; outputs that include analysis are valuable.

Let me simplify and not include UV-Vis (since 西林瓶 incompatible), instead use just storage as final. The protocol describes what's intended; the executable steps cover catalyst prep, photocatalytic reaction, post storage for sampling.

Let me build the plan. First, I'll generate a calcination parameter file for Muffle Furnace and weighing file for solid weighing. Actually, the Muffle furnace needs 96 位石英孔板 - different container. To include calcination, would need to use plate. To simplify and keep container consistency (西林瓶), I'll skip the actual muffle calcination in the dispatched plan but reference it conceptually. Hmm but task asks for calcination temperature studies.

Alternative: use Heating_Magnetic_Stirring_Workstation_V1 with 西林瓶? No it accepts 进样瓶 or 50ml 耐热瓶.

OK new plan: use 进样瓶 throughout, and skip Photocatalysis_V1. We could use Photocatalysis V2 (96-well plate) but then container differs.

Looking at this more carefully — many existing templates clearly use multiple container types (the example in SKILL.md shows 进样瓶 then 气相 54 孔板). The container consistency rule probably is more nuanced. Let me proceed pragmatically: use 西林瓶 for photocatalytic reactor since that's the natural choice.

Final practical design (西林瓶 throughout):

1. 常规物料站 - take 4 西林瓶 (4 catalyst tuning conditions)
2. 移液平台_1ml_V1 - 开盖 (no lid retention)
3. 多通道固体称量_V1 - weigh 4 different catalysts into 4 vials
4. 移液平台_1ml_V1 - 加液 (amine substrate + acetonitrile solvent) - but wait V1 needs 无盖, vials are already 无盖 from step 2... but V1 add_liquid requires 必须无盖 input, output 必须无盖 ✓
5. 光催化工作站_V1 - irradiate under white LED
6. 容器置放平台_V1 - 物料放置 (storage)

Need to verify: photocatalysis V1 input: 西林瓶, 必须无盖, 样品状态 not limited. From step 4 output is 必须无盖 + 悬浊液 ✓

Let me create the JSON plan:

> **[Step 53 | 2026-06-30T05:54:46.328Z | TOOL CALL: Bash]**
> Created and JSON-validated a draft experiment plan at `/tmp/plan.json`.
>
> [TOOL RESULT SUMMARY]
> Tool result reported that the generated JSON was valid.

**[Step 54 | 2026-06-30T05:54:51.972Z | MODEL]**
Now I need to call the workflow generator. But first the generator service may not accept extra fields. Let me extract just the executable steps and pass them. Let me check the generate.py:

> **[Step 55 | 2026-06-30T05:54:51.974Z | TOOL CALL: Read]**
> Read supporting skill documentation.
>
> [TOOL RESULT SUMMARY]
> Tool returned structured output to the model.

**[Step 56 | 2026-06-30T05:55:14.143Z | MODEL]**
Let me prepare a clean request body. The generator likely only expects the standard step parameters that the workstation skill defines. I'll strip explanatory fields from the parameters that aren't part of the SKILL spec. Let me write a clean version with only spec-valid params:

> **[Step 57 | 2026-06-30T05:55:14.145Z | TOOL CALL: Bash]**
> Created and JSON-validated a workflow-generator request body at `/tmp/req.json`.
>
> [TOOL RESULT SUMMARY]
> Tool result reported that the generated JSON was valid.
>
> **[Step 58 | 2026-06-30T05:55:18.835Z | TOOL CALL: Bash]**
> Called the workflow generator with the prepared request body.
>
> [TOOL RESULT SUMMARY]
> Tool result reported workflow-generation failure caused by an invalid parameter format.

**[Step 59 | 2026-06-30T05:55:26.258Z | MODEL]**
The 时间 parameter expects an int (per skill: int, [1,2880]) for top-level. The four-row sub-fields seem to be 开/关灯 control. Let me re-read the skill: "时间 ... int [1,2880] 1" and sub fields "第一排灯 ... string 开灯/关灯". So 时间 is a single int (minutes) and sub-fields are again light enable. Let me restructure:

> **[Step 60 | 2026-06-30T05:55:26.260Z | TOOL CALL: Edit]**
> Edited a trial-local helper or request file.
>
> [TOOL RESULT SUMMARY]
> Tool returned file or command output to the model.
>
> **[Step 61 | 2026-06-30T05:55:30.855Z | TOOL CALL: Read]**
> Read supporting skill documentation.
>
> [TOOL RESULT SUMMARY]
> Tool returned file or command output to the model.
>
> **[Step 62 | 2026-06-30T05:55:36.293Z | TOOL CALL: Edit]**
> Edited a trial-local helper or request file.
>
> [TOOL RESULT SUMMARY]
> Tool returned structured output to the model.
>
> **[Step 63 | 2026-06-30T05:55:40.828Z | TOOL CALL: Bash]**
> Called the workflow generator with the prepared request body.
>
> [TOOL RESULT SUMMARY]
> Tool result reported workflow-generation success with template ID `2559827685113856`.

**[Step 64 | 2026-06-30T05:55:46.293Z | MODEL]**
Template created. Now dispatch the task:

> **[Step 65 | 2026-06-30T05:55:46.294Z | TOOL CALL: Bash]**

Dispatched the generated workflow template to Laboratory 303.

[TOOL RESULT SUMMARY]
Tool result reported successful task dispatch and experimental task ID `2559827857768449`.

**[Step 66 | 2026-06-30T05:56:34.302Z | MODEL]**
Task dispatched. Now save the final plan JSON to both required locations:

**[Step 67 | 2026-06-30T05:56:34.326Z | TOOL CALL: Bash]**
Wrote the final executable JSON plan to `/workspace/experiment_plan.json` and `/logs/artifacts/final_plan.json`.

[TOOL RESULT SUMMARY]
Tool returned file or command output to the model.

**[Step 68 | 2026-06-30T05:56:46.806Z | MODEL]**
The experiment has been successfully designed, the workflow template generated (newly created, not a reused existing template), and the task dispatched to 303 实验室.

## Summary

**Experimental design** — Photocatalytic aerobic oxidative coupling of benzylamine → N-benzylidenebenzylamine with concomitant $H_2O_2$ formation in MeCN/$H_2O$ (5 vol% water) under 20×white-LED illumination, comparing four g-$C_3N_4$ material-tuning strategies plus two controls, all in 6 西林瓶 simultaneously:

| Vial | Catalyst variant | Purpose |
|---|---|---|
| 1 | pristine g-$C_3N_4$ (550 °C) | baseline |
| 2 | g-$C_3N_4$ recalcined at 650 °C/2 h | **calcination-temperature** effect |
| 3 | K-doped g-$C_3N_4$ | **alkali-metal doping** effect |
| 4 | Pt/g-$C_3N_4$ (in-situ photodeposited from $H_2PtCl_6$) | **photodeposited cocatalyst** effect |
| 5 | pristine g-$C_3N_4$, lamp row 4 OFF | **dark control** |
| 6 | pristine g-$C_3N_4$ + 1 mM $H_2O_2$ spike, dark | **$H_2O_2$-decomposition control** |

Six dispatched workstation steps: 物料拿取 → 开盖 → 多通道固体称量 → 加液_物料绑定 → 光催化流程 (rows 1-3 ON, row 4 OFF, 180 min) → 物料放置. Offline follow-up (GC-FID for imine, DMP/KI for $H_2O_2$, plus Ar-purge / catalase / iPrOH-scavenger validation runs) is documented in the plan for manual sampling at the storage step.

**Dispatch result (303 实验室 / 303Lab):**
- **New template_id:** `2559827685113856` (`PhotoCN_OxidativeCoupling_Amine_to_Imine_H2O2_Eval_20260630`) — freshly generated via workflow-generator, not reused from the existing 112 templates.
- **Task ID (返回值):** **`2559827857768449`**
- Task name: `任务_PhotoCN_OxidativeCoupling_Amine_to_Imine_H2O2_Eval_20260630`
- Status: dispatched (待执行); not auto-started.

Plan JSON written to both `/workspace/experiment_plan.json` and `/logs/artifacts/final_plan.json`.

## Note 6. Paired-judge validation

Two judges independently scored each final JSON plan on three 0-100 dimensions. GPT was the primary judge, and DeepSeek provided paired validation.

Both judges scored all 2,596 final plans on all three dimensions. This yielded 7,788 paired trial-dimension observations without missing values.

Overall agreement was Pearson r = 0.916 and Spearman rho = 0.891. The mean absolute error was 7.46 points, and the mean DeepSeek-minus-GPT difference was +1.59 points.

R1 showed the strongest agreement (r = 0.930; MAE = 5.49), followed by R2 (r = 0.921; MAE = 7.36) and R3 (r = 0.892; MAE = 9.54).

DeepSeek scored R2 2.25 points lower and R3 5.07 points higher on average. Configuration patterns were therefore reproducible, but absolute R3 values remained more judge-sensitive. fig. S4 shows the paired trial-level distributions, and table S17 reports the corresponding agreement statistics.

## Note 7. Computational use, uncertainty and sensitivity

### Computational use and dispatch

Verified dispatches occupied a broad token-time region. Neither long runtime nor high token use was sufficient for dispatch, and semantic scores remained widely distributed.

Harness-level dispatch rates used all assigned trials in observed configurations. Efficiency scores used cohort-relative token and wall-time percentiles among final-plan trials.

These quantities describe operational behaviour. They do not measure financial cost, energy use, instrument occupancy or laboratory productivity. fig. S9 relates token use and wall time to verified dispatch and summarizes semantic-score and efficiency distributions.

### Task-cluster uncertainty and repeat stability

Task-cluster bootstrap intervals were calculated from 2,000 resamples of the 32 task means within each configuration. Several leading intervals overlapped, limiting winner-style interpretation.

Repeat stability was generally high even where denominator-96 semantic quality was modest. Stable failure or stable low-quality planning therefore remained possible. fig. S3 shows both task-cluster uncertainty among leading configurations and the separation between repeat stability and semantic quality.

## Coverage and task-composition sensitivity

Scored-task coverage varied because not every trial produced a final plan. Task-adjusted conditional means remained close to unadjusted means for most configurations.

This pattern suggests that broad configuration differences were not driven only by a few easy or difficult tasks. The adjustment remains descriptive and does not replace a balanced factorial analysis. fig. S7 reports scored-task coverage and the task-adjusted comparison.

## Sensitivity to semantic dimension weights

Equal weighting of R1-R3 produced scores closely aligned with the primary 0.50:0.25:0.25 weighting. Pearson and Spearman associations were 0.996 and 0.993 across 48 configurations.

The four highest configurations were unchanged. The largest absolute rank displacement was five positions, indicating that broad ordering was not created by the chosen weights.

This analysis tests robustness to an alternative equal-weight specification; it does not convert the author-defined descriptive weights into a prespecified or inferential model. fig. S8 shows the score and rank sensitivity to equal weighting.

## Table S14. Median workstation-operation count for the ten manually executable configuration groups plotted in Fig. 3b.

| model \ harness | Claude Code | Codex | Gemini CLI | Hermes | Kilo Code | OpenClaw |
|---|---|---|---|---|---|---|
| Claude Opus 4.7 | 6 | design absent | design absent | NA | NA | NA |
| GPT 5.5 | design absent | 18 | design absent | 6 | NA | NA |
| Gemini 3.5 Flash | design absent | design absent | 7 | NA | 8 | 6.5 |
| DeepSeek V4 Pro | NA | NA | NA | NA | NA | NA |
| GLM 5.1 | NA | NA | NA | 12.5 | NA | NA |
| Kimi K2.6 | NA | NA | NA | NA | NA | NA |
| MiMo V2.5 Pro | NA | NA | NA | 14.5 | NA | NA |
| MiniMax M2.7 | NA | NA | NA | NA | NA | NA |
| Qwen 3.7 Max | NA | NA | NA | 7 | NA | 16 |

## Table S15. Executable trial-row count underlying the workflow-length distributions in Fig. 3b.

| model \ harness | Claude Code | Codex | Gemini CLI | Hermes | Kilo Code | OpenClaw |
|---|---|---|---|---|---|---|
| Claude Opus 4.7 | 27 | design absent | design absent | NA | NA | NA |
| GPT 5.5 | design absent | 19 | design absent | 16 | NA | NA |
| Gemini 3.5 Flash | design absent | design absent | 5 | NA | 5 | 8 |
| DeepSeek V4 Pro | NA | NA | NA | NA | NA | NA |
| GLM 5.1 | NA | NA | NA | 6 | NA | NA |
| Kimi K2.6 | NA | NA | NA | NA | NA | NA |
| MiMo V2.5 Pro | NA | NA | NA | 4 | NA | NA |
| MiniMax M2.7 | NA | NA | NA | NA | NA | NA |
| Qwen 3.7 Max | NA | NA | NA | 9 | NA | 7 |

## Table S16. Semantic scoring dimensions used for final JSON plans.

| Dimension | Working name | Score range | Operational interpretation |
|---|---|---|---|
| R1 | Physical-system executability | 0-100 | Compatibility with visible device, vessel, state, ordering and schema constraints |
| R2 | Problem coverage and workflow completeness | 0-100 | Coverage of the requested objective, controls, workflow stages and analytical return |
| R3 | Visible scientific-design reasonableness | 0-100 | Scientific coherence of the design visible in the materialized plan |

## Table S17. Paired agreement between GPT primary scores and DeepSeek validation scores.

| Scope | Paired n | Pearson r | Spearman rho | MAE | DeepSeek - GPT bias |
|---|---|---|---|---|---|
| All dimensions | 7,788 | 0.916 | 0.891 | 7.46 | +1.59 |
| R1 | 2,596 | 0.930 | 0.912 | 5.49 | +1.96 |
| R2 | 2,596 | 0.921 | 0.886 | 7.36 | -2.25 |
| R3 | 2,596 | 0.892 | 0.820 | 9.54 | +5.07 |

## Table S18. JSON-stage denominator-96 composite, weighted as 0.50R1 + 0.25R2 + 0.25R3.

| model \ harness | Claude Code | Codex | Gemini CLI | Hermes | Kilo Code | OpenClaw |
|---|---|---|---|---|---|---|
| Claude Opus 4.7 | 0.56 | design absent | design absent | 0.251 | 0.126 | 0.112 |
| GPT 5.5 | design absent | 0.342 | design absent | 0.358 | 0.066 | 0.117 |
| Gemini 3.5 Flash | design absent | design absent | 0.303 | 0.12 | 0.111 | 0.143 |
| DeepSeek V4 Pro | 0.181 | 0.045 | 0.284 | 0.346 | 0.375 | 0.444 |
| GLM 5.1 | 0.152 | 0.04 | 0.421 | 0.365 | 0.225 | 0.351 |
| Kimi K2.6 | 0.027 | 0.122 | 0.076 | 0.064 | 0.111 | 0.106 |
| MiMo v2.5 Pro | 0.167 | 0.124 | 0.309 | 0.299 | 0.157 | 0.27 |
| MiniMax M2.7 | 0.089 | 0.172 | 0.365 | 0.111 | 0.165 | 0.186 |
| Qwen 3.7 Max | 0.298 | 0.1 | 0.335 | 0.47 | 0.083 | 0.289 |

## Table S19. GPT-primary R1 physical-system executability, conditional mean among finite scored plans.

| model \ harness | Claude Code | Codex | Gemini CLI | Hermes | Kilo Code | OpenClaw |
|---|---|---|---|---|---|---|
| Claude Opus 4.7 | 56.4 | design absent | design absent | 47.4 | 31.4 | 11.6 |
| GPT 5.5 | design absent | 45.2 | design absent | 47.6 | 5.3 | 28.9 |
| Gemini 3.5 Flash | design absent | design absent | 27.2 | 20.5 | 17.7 | 34.2 |
| DeepSeek V4 Pro | 32.4 | 5.8 | 45.6 | 45.9 | 47 | 36 |
| GLM 5.1 | 39.6 | 3.6 | 35.6 | 41.6 | 38.9 | 35.5 |
| Kimi K2.6 | 7.7 | 26.8 | 29 | 20.5 | 35.3 | 24.6 |
| MiMo v2.5 Pro | 11.8 | 15.3 | 31.7 | 25 | 19.5 | 30.8 |
| MiniMax M2.7 | 12.1 | 20.1 | 27 | 18.6 | 18.6 | 2.2 |
| Qwen 3.7 Max | 36.9 | 12.2 | 33.8 | 36.4 | 42.2 | 19.6 |

## Table S20. GPT-primary R2 problem coverage and workflow completeness, conditional mean among finite scored plans.

| model \ harness | Claude Code | Codex | Gemini CLI | Hermes | Kilo Code | OpenClaw |
|---|---|---|---|---|---|---|
| Claude Opus 4.7 | 53 | design absent | design absent | 60.7 | 49.2 | 63.1 |
| GPT 5.5 | design absent | 63.3 | design absent | 67.8 | 36.6 | 65.8 |
| Gemini 3.5 Flash | design absent | design absent | 45.7 | 47 | 28.3 | 39.2 |
| DeepSeek V4 Pro | 40.6 | 9.2 | 49.6 | 46.7 | 46.9 | 52.2 |
| GLM 5.1 | 45.1 | 6 | 56.2 | 57.7 | 40.7 | 58.6 |
| Kimi K2.6 | 15.2 | 33.2 | 56.3 | 47.1 | 45.7 | 50.2 |
| MiMo v2.5 Pro | 44.1 | 20.8 | 57.3 | 55.7 | 29.7 | 65.6 |
| MiniMax M2.7 | 29.6 | 31.7 | 53.5 | 53.8 | 41 | 67.6 |
| Qwen 3.7 Max | 54.5 | 14 | 58.2 | 60.8 | 49.4 | 66.2 |

## Table S21. GPT-primary R3 visible scientific-design reasonableness, conditional mean among finite scored plans.

| model \ harness | Claude Code | Codex | Gemini CLI | Hermes | Kilo Code | OpenClaw |
|---|---|---|---|---|---|---|
| Claude Opus 4.7 | 58.2 | design absent | design absent | 68.6 | 54.3 | 62.4 |
| GPT 5.5 | design absent | 65.4 | design absent | 66.2 | 31.8 | 63.8 |
| Gemini 3.5 Flash | design absent | design absent | 49 | 55.8 | 31.1 | 49 |
| DeepSeek V4 Pro | 45.9 | 10.1 | 57.4 | 54 | 56.3 | 57.3 |
| GLM 5.1 | 53.2 | 6.3 | 58.6 | 59.2 | 50.9 | 60.4 |
| Kimi K2.6 | 18.3 | 33.5 | 47.3 | 55.5 | 47.6 | 46.4 |
| MiMo v2.5 Pro | 39.3 | 20.5 | 56.6 | 53.8 | 27.1 | 58.2 |
| MiniMax M2.7 | 27.5 | 31.6 | 48 | 50.8 | 32.9 | 55.4 |
| Qwen 3.7 Max | 56.2 | 15.7 | 55.4 | 62.8 | 53.1 | 65.1 |

## Table S22. Finite GPT-primary scored-plan count.

| model \ harness | Claude Code | Codex | Gemini CLI | Hermes | Kilo Code | OpenClaw |
|---|---|---|---|---|---|---|
| Claude Opus 4.7 | 96 | design absent | design absent | 43 | 29 | 29 |
| GPT 5.5 | design absent | 60 | design absent | 60 | 32 | 24 |
| Gemini 3.5 Flash | design absent | design absent | 78 | 32 | 45 | 35 |
| DeepSeek V4 Pro | 46 | 56 | 55 | 69 | 73 | 94 |
| GLM 5.1 | 33 | 78 | 87 | 70 | 51 | 71 |
| Kimi K2.6 | 21 | 39 | 18 | 17 | 26 | 28 |
| MiMo v2.5 Pro | 60 | 66 | 67 | 72 | 63 | 56 |
| MiniMax M2.7 | 42 | 64 | 90 | 30 | 57 | 56 |
| Qwen 3.7 Max | 62 | 71 | 71 | 92 | 17 | 65 |

## Table S23. Final JSON plan count.

| model \ harness | Claude Code | Codex | Gemini CLI | Hermes | Kilo Code | OpenClaw |
|---|---|---|---|---|---|---|
| Claude Opus 4.7 | 96 | design absent | design absent | 43 | 29 | 29 |
| GPT 5.5 | design absent | 60 | design absent | 60 | 32 | 24 |
| Gemini 3.5 Flash | design absent | design absent | 78 | 32 | 45 | 35 |
| DeepSeek V4 Pro | 46 | 56 | 55 | 69 | 73 | 94 |
| GLM 5.1 | 33 | 78 | 87 | 70 | 51 | 71 |
| Kimi K2.6 | 21 | 39 | 18 | 17 | 26 | 28 |
| MiMo v2.5 Pro | 60 | 66 | 67 | 72 | 63 | 56 |
| MiniMax M2.7 | 42 | 64 | 90 | 30 | 57 | 56 |
| Qwen 3.7 Max | 62 | 71 | 71 | 92 | 17 | 65 |

## Table S24. Dispatch-oriented six-component composite.

| model \ harness | Claude Code | Codex | Gemini CLI | Hermes | Kilo Code | OpenClaw |
|---|---|---|---|---|---|---|
| Claude Opus 4.7 | 0.604 | design absent | design absent | 0.288 | 0.216 | 0.244 |
| GPT 5.5 | design absent | 0.426 | design absent | 0.434 | 0.197 | 0.231 |
| Gemini 3.5 Flash | design absent | design absent | 0.398 | 0.211 | 0.254 | 0.246 |
| DeepSeek V4 Pro | 0.266 | 0.179 | 0.303 | 0.357 | 0.387 | 0.41 |
| GLM 5.1 | 0.221 | 0.18 | 0.408 | 0.373 | 0.262 | 0.38 |
| Kimi K2.6 | 0.14 | 0.22 | 0.214 | 0.169 | 0.241 | 0.209 |

| model \ harness | Claude Code | Codex | Gemini CLI | Hermes | Kilo Code | OpenClaw |
|---|---|---|---|---|---|---|
| MiMo v2.5 Pro | 0.277 | 0.242 | 0.347 | 0.364 | 0.257 | 0.336 |
| MiniMax M2.7 | 0.241 | 0.273 | 0.404 | 0.222 | 0.297 | 0.299 |
| Qwen 3.7 Max | 0.332 | 0.166 | 0.344 | 0.477 | 0.189 | 0.403 |

# Table S25. Verified-dispatch count.

| model \ harness | Claude Code | Codex | Gemini CLI | Hermes | Kilo Code | OpenClaw |
|---|---|---|---|---|---|---|
| Claude Opus 4.7 | 50 | design absent | design absent | 1 | 7 | 22 |
| GPT 5.5 | design absent | 45 | design absent | 33 | 9 | 16 |
| Gemini 3.5 Flash | design absent | design absent | 32 | 6 | 33 | 23 |
| DeepSeek V4 Pro | 8 | 0 | 2 | 4 | 8 | 0 |
| GLM 5.1 | 0 | 1 | 2 | 12 | 0 | 24 |
| Kimi K2.6 | 3 | 6 | 4 | 3 | 13 | 14 |
| MiMo v2.5 Pro | 15 | 12 | 4 | 23 | 17 | 16 |
| MiniMax M2.7 | 12 | 8 | 7 | 3 | 23 | 20 |
| Qwen 3.7 Max | 1 | 1 | 0 | 20 | 3 | 41 |

## Table S26. Verified-dispatch rate, using 96 assigned trials as denominator.

| model \ harness | Claude Code | Codex | Gemini CLI | Hermes | Kilo Code | OpenClaw |
|---|---|---|---|---|---|---|
| Claude Opus 4.7 | 52.1 | design absent | design absent | 1 | 7.3 | 22.9 |
| GPT 5.5 | design absent | 46.9 | design absent | 34.4 | 9.4 | 16.7 |
| Gemini 3.5 Flash | design absent | design absent | 33.3 | 6.2 | 34.4 | 24 |
| DeepSeek V4 Pro | 8.3 | 0 | 2.1 | 4.2 | 8.3 | 0 |
| GLM 5.1 | 0 | 1 | 2.1 | 12.5 | 0 | 25 |
| Kimi K2.6 | 3.1 | 6.2 | 4.2 | 3.1 | 13.5 | 14.6 |
| MiMo v2.5 Pro | 15.6 | 12.5 | 4.2 | 24 | 17.7 | 16.7 |
| MiniMax M2.7 | 12.5 | 8.3 | 7.3 | 3.1 | 24 | 20.8 |
| Qwen 3.7 Max | 1 | 1 | 0 | 20.8 | 3.1 | 42.7 |

## Table S27. Computational-efficiency percentile score.

| model \ harness | Claude Code | Codex | Gemini CLI | Hermes | Kilo Code | OpenClaw |
|---|---|---|---|---|---|---|
| Claude Opus 4.7 | 70.8 | design absent | design absent | 40.2 | 40.3 | 44.4 |
| GPT 5.5 | design absent | 41.1 | design absent | 56.9 | 46.7 | 33.1 |
| Gemini 3.5 Flash | design absent | design absent | 57.4 | 32.3 | 30.5 | 22.3 |
| DeepSeek V4 Pro | 50.7 | 60.1 | 32.1 | 47 | 52.3 | 49.7 |
| GLM 5.1 | 42.3 | 58.5 | 54.6 | 36.4 | 37.2 | 27.4 |
| Kimi K2.6 | 30.9 | 51.4 | 65.1 | 29 | 52.8 | 26.4 |
| MiMo v2.5 Pro | 58 | 58.8 | 58.8 | 45.1 | 41.5 | 47.8 |
| MiniMax M2.7 | 74.7 | 67.7 | 77.3 | 55.1 | 63.4 | 52.9 |
| Qwen 3.7 Max | 58.3 | 16.1 | 47 | 63.4 | 48 | 51.3 |

## Table S28. Verified-dispatch glyph-axis score.

| model \ harness | Claude Code | Codex | Gemini CLI | Hermes | Kilo Code | OpenClaw |
|---|---|---|---|---|---|---|
| Claude Opus 4.7 | 52.1 | design absent | design absent | 1 | 7.3 | 22.9 |
| GPT 5.5 | design absent | 46.9 | design absent | 34.4 | 9.4 | 16.7 |
| Gemini 3.5 Flash | design absent | design absent | 33.3 | 6.2 | 34.4 | 24 |
| DeepSeek V4 Pro | 8.3 | 0 | 2.1 | 4.2 | 8.3 | 0 |
| GLM 5.1 | 0 | 1 | 2.1 | 12.5 | 0 | 25 |
| Kimi K2.6 | 3.1 | 6.2 | 4.2 | 3.1 | 13.5 | 14.6 |
| MiMo v2.5 Pro | 15.6 | 12.5 | 4.2 | 24 | 17.7 | 16.7 |
| MiniMax M2.7 | 12.5 | 8.3 | 7.3 | 3.1 | 24 | 20.8 |
| Qwen 3.7 Max | 1 | 1 | 0 | 20.8 | 3.1 | 42.7 |

## Table S29. Repeat-stability score.

| model \ harness | Claude Code | Codex | Gemini CLI | Hermes | Kilo Code | OpenClaw |
|---|---|---|---|---|---|---|
| Claude Opus 4.7 | 92.7 | design absent | design absent | 95.3 | 86.1 | 86.8 |
| GPT 5.5 | design absent | 85.6 | design absent | 93.8 | 92.1 | 93.8 |
| Gemini 3.5 Flash | design absent | design absent | 92.3 | 94.7 | 87.8 | 89.9 |
| DeepSeek V4 Pro | 90.2 | 91.6 | 96 | 94.1 | 93.2 | 94.1 |
| GLM 5.1 | 86.8 | 95.6 | 96.3 | 92.3 | 90.2 | 92.2 |
| Kimi K2.6 | 86.5 | 83 | 94.8 | 95.9 | 94.4 | 89.3 |
| MiMo v2.5 Pro | 87.2 | 83.8 | 94.1 | 91.1 | 85.7 | 92.7 |
| MiniMax M2.7 | 88 | 85 | 93.4 | 93.9 | 86.7 | 92.9 |
| Qwen 3.7 Max | 92.9 | 87.3 | 95.9 | 89.5 | 84.6 | 92.8 |

## Note 8. Quantification of node-specific workflow success rates

Because the harnesses emitted heterogeneous trajectory formats, all available records were normalized into a common event model spanning task metadata, assistant text and reasoning records, tool calls, tool results, trial logs, generated artifacts and output-contract files. For N1-N8, each of the 96 trial slots assigned to a model-harness combination received a node-specific, evidence-backed decision. Direct evidence was evaluated first. When the direct state was absent, observational or otherwise designated as inferable, a pass was recovered only from the strong downstream evidence explicitly allowed in table S30. An explicit node failure, a structurally invalid required artifact, or an inferable state without the prespecified downstream evidence was retained as a non-pass. The raw status, decision category, evidence source, node-pass indicator and endpoint-assessment status were retained for every trial-node pair, yielding 46,080 decisions across 4,608 trials and ten nodes.

For model-harness combination c, trial i and node k, let I(c,i,k) equal 1 when trial i passed node k under the direct or permitted-inference rule and 0 otherwise for fixed-denominator rate calculation. The reported rate is R(c,k) = 100 x [sum(i=1,...,96) I(c,i,k)] / 96. Each node is evaluated against the original 96 trial slots, without sequentially intersecting its pass set with that of any other node. A trial that does not pass an upstream node can therefore pass a downstream node when the downstream criterion is independently satisfied, and the plotted rate may increase between adjacent nodes. Such changes represent differences in the prevalence of evidence-backed node success and must not be interpreted as same-trial attrition. Endpoint-assessment status is retained separately from the fixed-denominator pass indicator.

N9 and N10 are authority endpoints rather than log-derived inference nodes. N9 requires a row in the authoritative final dispatch table with a non-empty final task_id that maps to the corresponding model-harness configuration and to a unique global_trial_key or trial slot. Repeated task identifiers are retained when they belong to distinct trial slots. N10 requires the same N9-dispatched row to be labelled executable in the authoritative expert-adjudication table. All 607 dispatched trial rows were adjudicated: 151 passed N10 and 456 were adjudicated as not executable. For each model-harness combination, the displayed N9 and N10 rates are their respective pass counts divided by 96.

## Table S30. Node-specific evidence rules used for independent node success-rate analysis; N9 and N10 are authority endpoints.

| Node | Figure label | Node pass criterion | Permitted inference and safeguard |
|---|---|---|---|
| N1 | Tasks | A valid benchmark task identifier in trial metadata, or a supported input record containing the task identifier or task description. | Direct evidence only. No downstream inference is used for N1. |
| N2 | Lab context | Successful acquisition or validation of the laboratory context, including a successful lab-operation context helper or an equivalent valid lab-label or context response. | Only when the direct N2 state is designated inferable may passage be recovered from successful specific-station retrieval, successful System review, or a verified final task_id. An explicit N2 failure is retained. |
| N3 | Get station | Successful retrieval or use of a specific chemistry-experiment-workstation skill. Opening only the parent workstation index does not establish success. | Only when the direct N3 state is designated inferable may passage be recovered from valid Standard JSON, successful workflow generation, successful System review, or a verified final task_id. An explicit N3 failure is retained. |
| N4 | Generate exp | Production of a stepwise experimental plan. A valid structured artifact or a natural-language plan with identifiable experimental steps is accepted; JSON syntax is not required. | A partial or otherwise inferable direct state may be recovered by strong evidence from Standard JSON, workflow generation, successful System review, or a verified final task_id. An explicit N4 failure is retained. |
| N5 | Review skill | Evidence that the audit skill was actually used, such as reading or using a references_audit file, performing an explicit constraint or checkpoint audit, or recording equivalent review activity. Skill activation or a resource listing alone is insufficient. | No downstream inference is accepted for N5. An unobserved or unsupported N5 state remains a non-pass. The updated Figure 3f contains no aggregate N5 display adjustment. |
| N6 | Standard JSON | A final or experimental-plan artifact that parses as a JSON object and contains at least one experimental step. | Only when the direct N6 state is designated inferable may demonstrated workflow entry, successful System review, or a verified final task_id support passage. A state explicitly coded as artifact_missing, invalid or failed remains a non-pass. Although this inference route is permitted, no N6 pass in the final dataset was recovered by inference. |

| Node | Figure label | Node pass criterion | Permitted inference and safeguard |
|---|---|---|---|
| N7 | Workflow entry | Evidence that the workflow generator was entered or attempted. A successful result, an observed invocation, or an explicit generator failure all establish entry into this node. | When direct entry evidence is otherwise inferable, successful System review or a verified final task_id may establish passage. This node measures entry, not successful completion of workflow generation. |
| N8 | System review | A System review or output-contract result with status ok or success; success_with_warning is also accepted. | Only when the direct N8 state is designated inferable may a verified final task_id establish passage. An explicit N8 review failure is retained and is not overridden. |
| N9 | Lab dispatch | A row in the authoritative final dispatch workbook with a non-empty final task_id that maps to the corresponding model-harness configuration and to a unique global_trial_key or trial slot. | Count distinct dispatched trial rows or global trial keys, not unique task IDs. Repeated task IDs are retained when they belong to distinct trial slots. Natural-language claims of dispatch do not replace the authoritative record. Authoritative absence yields no N9 pass for that trial slot. |
| N10 | Execution | An N9 verified-dispatch trial row labelled executable in the authoritative expert-adjudication workbook. | N10 is a row-level subset of N9 and is reported as executable trial-row count divided by 96 assigned trials. Dispatched rows adjudicated as not executable are assessed non-passes. Non-dispatched trials are not assessed and are not applicable at N10; they yield no N10 pass but are not expert-adjudicated failures. N10 is not intersected with N1-N8 for the updated Figure 3f. |

## Note 9. An Expert-Gated, Closed-Loop Workflow for Open-Ended Experimental Development

An open-ended scientific question is provided to the agent, which designs an experimental workflow and submits it to the 303 Laboratory for expert feasibility review. If the workflow is not executable, the expert identifies the problems and returns structured feedback for revision; within each experimental round, this process may be repeated for up to ten review-and-redesign attempts, after which that experimental round is considered failed if no executable workflow has been produced. Once a workflow is approved within the allowed attempts for that experimental round, it is executed in the laboratory. After a human confirms completion, the agent retrieves and analyses the experimental data using the relevant skill, designs the next experiment based on the results, and repeats the design–review–execution–analysis cycle.

## Note 10. The complete prompt for the open-ended task

/goal You are an independent PI with authority to design and dispatch scientific tasks to a real self-driving laboratory. The lab covers synthesis, post-processing, performance testing and spectroscopic characterization across photo-, electro-, and thermal-catalysis. Choose, on your own, a scientific problem that is high-impact, high-difficulty, scientifically valuable, and feasibly executable within this lab's capability boundary (single-domain or photo-electro / electro-thermal / photo-electro-thermal coupling), and push it forward through multiple rounds of real experiments.

Each round follows the loop: design → dispatch to Lab 303 (record the returned task ID) → set the goal to blocked → wait for my feedback on whether the dispatched experiment is executable. If it is not executable, revise the design and dispatch again. If it is executable, this means the experiment has been physically completed; you may then use the lab-operation skill to obtain the experimental data, let the data decide the next round, and continue the goal with the next round of experimental design and dispatch.

Requirements:

1. Decide the goal, literature grounding, hypothesis and initial design yourself; do not reuse any existing experimental template.

2. Only data physically returned from Lab 303 counts as valid; dispatch success is not physical-execution success; do not fabricate any experiment or datum that was not designed or not returned.

3. The experimental workflow must include synthesis, testing, and characterization.

4. Use the starting materials available in the current laboratory inventory whenever possible (the storage path of the reagent list is referenced by the environment variable: AIREADY_REAGENT_LIST_DIR). Any unavailable chemicals should be listed separately in the experimental design.

5. Deliver: scientific goal, rationale, key variables, per-round design and data trajectory, data analysis, iteration logic, best condition / material found, new findings and open questions. All process files, data, and deliverables must be placed under the /workspace directory.

## Note 11. Outcome definitions, cohort boundaries and semantic score-coverage diagnostics

Every observed configuration contributed 96 retained trials, yielding 4,608 trials. Final JSON plans were present in 2,596 trials and absent in 2,012.

Each final plan received finite R1-R3 scores from both judges. The cohort therefore contained 15,576 score observations and 7,788 paired judge-dimension observations.

Verified dispatch was established for 607 distinct trial rows across 43 model–harness configurations. Each row contained a numerical task_id and mapped to a unique global_trial_key and slot_id.

All 607 verified-dispatch rows were expert-adjudicated: 151 were labelled executable and 456 not executable. The rows represented 601 unique task identifiers.

N10 is nested within N9. Manually executable rows corresponded to 3.3% of the 4,608 assigned trials and 24.9% of the 607 verified-dispatch trials; the configuration-level end-to-end rate uses 96 assigned trials as its denominator, whereas the conditional executable fraction among dispatched trials is reported separately.

Fig. S5 shows the assigned-trial, verified-dispatch and expert-adjudicated execution boundaries. tables S11 and S12 report the fixed-denominator configuration-level executable-workflow rates and corresponding trial-row counts.

Six task identifiers appeared in two expert-adjudication rows, yielding 12 rows for six duplicated identifiers. All duplicated labels were concordant, and the rows were retained because each mapped to a distinct global_trial_key and slot_id. One duplicated executable identifier accounts for 151 executable trial rows but 150 unique executable task identifiers.

The primary trial score used author-defined descriptive weights: 0.50R1 + 0.25R2 + 0.25R3. This weighting assigned physical-system executability twice the contribution of either problem coverage or visible scientific-design reasonableness; it was not fitted to individual tasks or configurations.

For configuration comparisons, each dimension was multiplied by the fraction of 96 trials with a complete semantic triple. This denominator-96 definition assigned zero contribution to absent plans without fabricating score rows.

Conditional means were retained only as diagnostics. They distinguish low quality among produced plans from failure to produce plans, but they were not the primary ranking measure.

For example, Claude Code with Claude Opus 4.7 produced 96 scored plans and a conditional mean of 56.0. Hermes and Codex with GPT 5.5 each produced 60 scored plans despite conditional means of 57.3 and 54.8. fig. S6 resolves this distinction between scoreable-plan coverage and conditional quality across configurations.

The complete R1-R3 judge prompts, numerical anchors and R1 physical-system constraint reference are reproduced verbatim in data S3. The anchor targets were 92-96 (high), 50-60 (medium), 20-30 (low) and 0-20 (extremely low) for R1; 90-95 (high), 55-65 (medium) and 20-30 (low) for R2; and 85-92 (high), 55-70 (medium) and 25-45 (low) for R3. Four archived runtime copies of each R1-R3 prompt and the R1 constraint reference were SHA-256-identical to the corresponding source rule.

## Note 12. Round 1–Round 5 planning trajectory and Round 1–Round 4 returned evidence

The planning trajectory is described at two levels. table S31 reports 31 round-level parameters and derived metrics across Round 1–Round 5. table S32 retains all 30 planned sample formulations used to calculate dose variability and formulation-space metrics. Round-resolved characterization panels and associated data-processing notes for Rounds 1–4 are shown in figs. S51 to S54; the corresponding round 2 to round 5 workflow diagrams are shown in figs. S55 to S58.

## Table S31. Round-level experimental-planning parameters and definitions used for Fig. 4d.

| Panel | Parameter | Unit | Data type | Round 1 | Round 2 | Round 3 | Round 4 | Round 5 | Figure treatment / note |
|---|---|---|---|---|---|---|---|---|---|
| Planning scope | Workflow nodes, excluding start/end | n | Counted structure | 21 | 23 | 23 | 23 | 23 | Round 1-indexed; N9 |
| Planning scope | Planned samples | n | Counted summary | 6 | 6 | 6 | 6 | 6 | Round 1-indexed; retained; N1 |
| Planning scope | Unique core formulations | n | Defined + calculated | 6 | 4 | 6 | 5 | 5 | Round 1-indexed; N2 |
| Metal & alkali input | Metal stock concentration | M | Original task parameter | 0.05 | 0.25 | 0.25 | 0.25 | 0.25 | Round 1-indexed |
| Metal & alkali input | Total metal per sample | mmol | Calculated summary | 0.045 | 0.225 | 0.225 | 0.225 | 0.225 | Round 1-indexed; N3 |
| Metal & alkali input | KOH stock concentration | M | Original task parameter | 1.00 | 2.00 | 2.00 | 2.00 | 2.00 | Round 1-indexed |
| Metal & alkali input | KOH per sample | mmol | Calculated summary | 1.000 | 2.000 | 2.000 | 2.000 | 2.000 | Round 1-indexed; N4 |
| Carbon input | Carbon dispersion concentration | mg ml$^{-1}$ | Original task parameter | 5.0 | 20.0 | 20.0 | 20.0 | 20.0 | Round 1-indexed |
| Carbon input | Mean carbon per sample | mg | Calculated summary | 2.50 | 18.33 | 10.00 | 14.17 | 18.33 | Round 1-indexed; N5 |
| Dose variability | Ni-dose CV | % | Calculated summary | 20.20 | 9.04 | 9.04 | 9.04 | 6.74 | Raw CV; N6 |
| Dose variability | Fe-dose CV | % | Calculated summary | 57.94 | 34.99 | 38.73 | 34.99 | 0.00 | Raw CV; N6 |
| Dose variability | Co-dose CV | % | Calculated summary | 46.91 | 22.27 | 30.98 | 22.27 | 38.73 | Raw CV; N6 |
| Dose variability | Carbon-dose CV | % | Calculated summary | 0.00 | 22.27 | 31.62 | 46.91 | 28.17 | Raw CV; N6 |
| Formulation space | Formulation-centroid distance from Round 1 | z distance | Defined + calculated | 0.000 | 4.970 | 4.612 | 4.742 | 4.726 | Max-scaled for plot; N8 |
| Formulation space | Within-round formulation spread | z distance | Defined + calculated | 0.288 | 0.857 | 1.172 | 1.129 | 0.851 | Max-scaled for plot; N7 |
| Reaction conditions | Reaction temperature | °C | Original task parameter | 70 | 70 | 70 | 70 | 70 | Retained reference |
| Reaction conditions | Reaction time | min | Original task parameter | 90 | 120 | 120 | 120 | 120 | Round 1-indexed |
| Reaction conditions | Stirring speed | r.p.m. | Original task parameter | 900 | 1,000 | 1,000 | 1,000 | 1,000 | Round 1-indexed |
| Recovery & dispersion | Purification speed | r.p.m. | Original task parameter | 8,000 | 12,000 | 12,000 | 12,000 | 12,000 | Round 1-indexed |
| Recovery & dispersion | Purification time | min | Original task parameter | 10 | 20 | 20 | 20 | 20 | Round 1-indexed |
| Recovery & dispersion | Wash cycles | n | Original task parameter | 3 | 1 | 1 | 1 | 1 | Round 1-indexed; variable across rounds |
| Recovery & dispersion | Drying temperature | °C | Original task parameter | 60 | 60 | 60 | 60 | 60 | Retained reference |
| Recovery & dispersion | Drying time | min | Original task parameter | 120 | 180 | 180 | 180 | 180 | Round 1-indexed |
| Recovery & dispersion | Redispersion recipes | n | Counted summary | 1 | 3 | 1 | 1 | 1 | Round 1-indexed; N10 |
| Recovery & dispersion | Redispersion volume per sample | ml | Original task parameter | 4.8 | 4.8 | 4.8 | 4.8 | 4.8 | Retained; N14 |
| Testing settings | XRD droplet volume | µl | Original task parameter | 80 | 200 | 150 | 200 | 200 | Round 1-indexed; shared XRD/IR trajectory; N14 |
| Testing settings | IR droplet volume | µl | Original task parameter | 80 | 200 | 150 | 200 | 200 | Round 1-indexed; shared XRD/IR trajectory; N14 |
| Testing settings | Electrode-catalyst droplet volume | µl | Original task parameter | 80 | 100 | 60 | 100 | 100 | Round 1-indexed; separate testing series; N14 |
| Testing settings | Droplet cycles | n | Original task parameter | 2 | 3 | 5 | 3 | 3 | Round 1-indexed |
| Testing settings | Electrolyte volume | ml | Original task parameter | 20.0 | 20.0 | 20.0 | 20.0 | 20.0 | Retained; N14 |
| Testing settings | Carbon-paper drying time | min | Original recorded value | 15 | 20 | 20 | 20 | 20 | Round 1-indexed; N14 |

## Metric definitions

N1, sample count: number of planned sample rows assigned to one round; each round contains six.

N2, unique core formulations: number of distinct (Ni, Fe, Co, KOH, carbon) tuples; redispersion variables are excluded. Full-condition counts including IPA are 6, 6, 6, 5 and 5 for Round 1–Round 5.

N3, total metal per sample: Ni + Fe + Co; the table reports the arithmetic mean across six planned samples.

N4, KOH per sample: arithmetic mean across six planned samples; all six share one KOH value within a round.

N5, mean carbon per sample: arithmetic mean across six planned samples.

N6, dose CV: 100 × sample standard deviation / arithmetic mean, using $n = 6$ and the $n - 1$ denominator. Values with absolute magnitude below $1 \times 10^{-12}$ are reported as 0.00.

N7, within-round formulation spread: globally z-standardise Ni, Fe, Co, KOH and carbon over all 30 planned samples using the population standard deviation; calculate Euclidean distances to the round centroid and report their median.

N8, formulation-centroid distance from Round 1: Euclidean distance between each round centroid and the Round 1 centroid in the same globally z-standardised five-dimensional space; Round 1 is 0 by construction.

N9, workflow nodes: count executable nodes in the main task workflow; Start and End are excluded.

N10, redispersion recipes: number of distinct IPA conditions. Round 2 contains 5%, 20% and 40% IPA; Round 1 and Round 3–Round 5 each contain one condition.

N11, Round 1-indexed display: positive ratio-scale value in round R divided by its Round 1 value. Unchanged temperature references are fixed at 1 to denote retention, not a Celsius ratio.

N12, max-scaled formulation display: each formulation-space distance series is divided by its own maximum over Round 1–Round 5; this compares temporal patterns, not absolute magnitudes between metrics.

N13, combined-trajectory rule: parameters with identical Round 1-indexed trajectories may share one line in Fig. 4d, but remain separate rows in table S31.

N14, confirmed units: XRD, IR and electrode-catalyst droplet volumes are µl; electrolyte and redispersion volumes are ml; carbon-paper drying time is min.

N15, Round 5 status: Round 5 values describe a dispatched/registered task definition with status 0 at audit and are not completed outcomes.

N16, provenance: round-level values originate from the curated Round 1–Round 5 parameter matrix; calculated metrics use the 30 planned sample rows in table S32.

## Table S32. Planned sample formulations used for derived Figure 4 metrics.

| Round | Sample | Task ID | Ni (mmol) | Fe (mmol) | Co (mmol) | KOH (mmol) | Carbon (mg) | IPA (%) | Redispersion volume (ml) |
|---|---|---|---|---|---|---|---|---|---|
| Round 1 | 1 | 2562915314040832 | 0.04050 | 0.00225 | 0.00225 | 1.0 | 2.5 | 20.0 | 4.8 |
| Round 1 | 2 | 2562915314040832 | 0.03600 | 0.00450 | 0.00450 | 1.0 | 2.5 | 20.0 | 4.8 |
| Round 1 | 3 | 2562915314040832 | 0.03150 | 0.00900 | 0.00450 | 1.0 | 2.5 | 20.0 | 4.8 |
| Round 1 | 4 | 2562915314040832 | 0.03150 | 0.00450 | 0.00900 | 1.0 | 2.5 | 20.0 | 4.8 |
| Round 1 | 5 | 2562915314040832 | 0.02700 | 0.00900 | 0.00900 | 1.0 | 2.5 | 20.0 | 4.8 |
| Round 1 | 6 | 2562915314040832 | 0.02250 | 0.01350 | 0.00900 | 1.0 | 2.5 | 20.0 | 4.8 |
| Round 2 | 1 | 2567667771441152 | 0.15750 | 0.02250 | 0.04500 | 2.0 | 10.0 | 20.0 | 4.8 |
| Round 2 | 2 | 2567667771441152 | 0.15750 | 0.02250 | 0.04500 | 2.0 | 20.0 | 20.0 | 4.8 |
| Round 2 | 3 | 2567667771441152 | 0.15750 | 0.02250 | 0.04500 | 2.0 | 20.0 | 40.0 | 4.8 |
| Round 2 | 4 | 2567667771441152 | 0.15750 | 0.02250 | 0.04500 | 2.0 | 20.0 | 5.0 | 4.8 |
| Round 2 | 5 | 2567667771441152 | 0.18000 | 0.02250 | 0.02250 | 2.0 | 20.0 | 20.0 | 4.8 |
| Round 2 | 6 | 2567667771441152 | 0.13500 | 0.04500 | 0.04500 | 2.0 | 20.0 | 20.0 | 4.8 |
| Round 3 | 1 | 2583702316023809 | 0.15750 | 0.02250 | 0.04500 | 2.0 | 5.0 | 20.0 | 4.8 |
| Round 3 | 2 | 2583702316023809 | 0.15750 | 0.02250 | 0.04500 | 2.0 | 10.0 | 20.0 | 4.8 |
| Round 3 | 3 | 2583702316023809 | 0.15750 | 0.02250 | 0.04500 | 2.0 | 15.0 | 20.0 | 4.8 |
| Round 3 | 4 | 2583702316023809 | 0.18000 | 0.02250 | 0.02250 | 2.0 | 10.0 | 20.0 | 4.8 |
| Round 3 | 5 | 2583702316023809 | 0.15750 | 0.04500 | 0.02250 | 2.0 | 10.0 | 20.0 | 4.8 |
| Round 3 | 6 | 2583702316023809 | 0.13500 | 0.04500 | 0.04500 | 2.0 | 10.0 | 20.0 | 4.8 |
| Round 4 | 1 | 2588712171995137 | 0.15750 | 0.02250 | 0.04500 | 2.0 | 10.0 | 20.0 | 4.8 |
| Round 4 | 2 | 2588712171995137 | 0.15750 | 0.02250 | 0.04500 | 2.0 | 10.0 | 20.0 | 4.8 |
| Round 4 | 3 | 2588712171995137 | 0.15750 | 0.02250 | 0.04500 | 2.0 | 5.0 | 20.0 | 4.8 |
| Round 4 | 4 | 2588712171995137 | 0.15750 | 0.02250 | 0.04500 | 2.0 | 20.0 | 20.0 | 4.8 |
| Round 4 | 5 | 2588712171995137 | 0.18000 | 0.02250 | 0.02250 | 2.0 | 20.0 | 20.0 | 4.8 |
| Round 4 | 6 | 2588712171995137 | 0.13500 | 0.04500 | 0.04500 | 2.0 | 20.0 | 20.0 | 4.8 |
| Round 5 | 1 | 2592407497211904 | 0.18000 | 0.02250 | 0.02250 | 2.0 | 20.0 | 20.0 | 4.8 |
| Round 5 | 2 | 2592407497211904 | 0.18000 | 0.02250 | 0.02250 | 2.0 | 20.0 | 20.0 | 4.8 |
| Round 5 | 3 | 2592407497211904 | 0.15750 | 0.02250 | 0.04500 | 2.0 | 10.0 | 20.0 | 4.8 |
| Round 5 | 4 | 2592407497211904 | 0.15750 | 0.02250 | 0.04500 | 2.0 | 20.0 | 20.0 | 4.8 |
| Round 5 | 5 | 2592407497211904 | 0.18000 | 0.02250 | 0.02250 | 2.0 | 15.0 | 20.0 | 4.8 |
| Round 5 | 6 | 2592407497211904 | 0.18000 | 0.02250 | 0.02250 | 2.0 | 25.0 | 20.0 | 4.8 |

**Task-account provenance was audited separately from scientific-decision provenance. Some recovery or supplementary tasks used to assemble returned evidence were created under operator accounts rather than directly by the agent account. These records are reported as procedural mediation; account ownership alone was not interpreted as an additional scientific intervention. The only documented human scientific intervention remained the post-Round 1 judgement that recovered product was insufficient for reliable characterization and testing.**

### Round 1: broad composition screen and returned evidence

Round 1 used six planned Ni:Fe:Co formulations under a common synthesis and testing framework. Heating/stirring was encoded at 70 °C for 90 min and 900 r.p.m.; purification at 8,000 r.p.m. for 10 min; and drying at 60 °C for 120 min. The temperature log covered 104.0 min, had a median of 70.0 °C and placed 76.4% of valid readings between 69 and 71 °C. One raw value of 999.8 °C was treated as a logging artefact and excluded from physical interpretation. Six XRD, six IR and six UV–Vis files were returned. Direct feedback reported little precipitate, sparse substrate deposition, sparse electrode coverage and no valid electrochemical result.

### Round 2: focused high-solids design and supplementary recovery

Round 2 increased metal-stock, KOH and carbon-dispersion concentrations, lengthened reaction and drying, and introduced three redispersion recipes. The corrected curated task definition uses purification at 12,000 r.p.m. for 20 min. The main task was interrupted after heating/stirring; its temperature log covered 139.7 min, had a median of 70.0 °C and placed 82.2% of valid readings between 69 and 71 °C. A supplementary task returned six XRD, six IR, six UV–Vis and six electrochemical traces. Each electrochemical trace contained 228,533 points over the encoded 1,800-s acquisition. Approximately 94–95% of the UV–Vis absorbance values were negative. These returns establish provenance, not validated activity, phase or yield.

### Round 3: lower-carbon/composition refinement and assembled return

Round 3 retained the 0.25 M metal stocks, 2.0 M KOH setting, 20 mg $ml^{-1}$ carbon dispersion, 70 °C/120 min/1,000 r.p.m. reaction envelope, 12,000 r.p.m./20 min purification and 60 °C/180 min drying, while changing sample-level carbon doses and composition comparisons. The temperature log covered 134.0 min, had a median of 69.8 °C and placed 81.3% of readings between 69 and 71 °C. XRD, IR, UV–Vis and electrochemical datasets were obtained for all six samples. Approximately 94.4–95.0% of UV–Vis values were negative.

### Round 4: returned evidence and local formulation refinement

Round 4 retained the common operating envelope while reallocating composition and carbon-dose comparisons. Returned files were assembled from the main task and a linked supplementary task. The audited release contains one round-level temperature log and six sample-level records for each of XRD, IR, UV–Vis and electrochemical measurements. These files establish returned-evidence provenance but do not, without modality-specific validation, establish improved activity, phase purity, product yield or convergence.

## Table S33. Returned modalities by planning round. Missing or not-started states are not represented as zero-valued scientific outcomes.

| Round | Temperature log | XRD | IR | UV–Vis | Electrochemical traces | Assembly / status note |
|---|---|---|---|---|---|---|
| Round 1 | 1 | 6 | 6 | No valid local result | No valid local result | Direct feedback; electrochemical index existed but no valid local result was available |
| Round 2 | 1 | 6 | 6 | 6 | 6 | Downstream modalities returned through supplementary recovery task |
| Round 3 | 1 | 6 | 6 | 6 | 6 | UV–Vis and sample-6 electrochem assembled across linked tasks |
| Round 4 | 1 | 6 | 6 | 6 | 6 | Returned files from the main task and linked supplementary task; one round-level temperature log and six sample-level records per downstream modality are represented |
| Round 5 | Not started | Not started | Not started | Not started | Not started | Dispatch succeeded for task 2592407497211904; status at audit was Not started; no returned evidence |

## Table S34. Descriptive electrochemical trace summary for Round 2 and Round 3.

| Round | Sample | Points | Final-10% mean (A) | Final-10% s.d. (A) |
|---|---|---|---|---|
| Round 2 | S1 | 228,533 | 0.002307 | $1.48 \times 10^{-5}$ |
| Round 2 | S2 | 228,533 | 0.001359 | $1.06 \times 10^{-4}$ |
| Round 2 | S3 | 228,533 | 0.001317 | $2.00 \times 10^{-4}$ |
| Round 2 | S4 | 228,533 | 0.001403 | $3.02 \times 10^{-5}$ |
| Round 2 | S5 | 228,533 | 0.001740 | $2.16 \times 10^{-5}$ |
| Round 2 | S6 | 228,533 | 0.001685 | $3.15 \times 10^{-6}$ |
| Round 3 | S1 | 228,533 | 0.0001007 | $3.69 \times 10^{-5}$ |
| Round 3 | S2 | 228,533 | 0.0000355 | $5.59 \times 10^{-6}$ |
| Round 3 | S3 | 228,533 | 0.0000638 | $1.19 \times 10^{-5}$ |
| Round 3 | S4 | 228,533 | 0.0000279 | $1.29 \times 10^{-5}$ |
| Round 3 | S5 | 228,533 | 0.0000409 | $3.99 \times 10^{-6}$ |
| Round 3 | S6 | 228,533 | 0.0001282 | $2.76 \times 10^{-6}$ |

## Evidence-backed interpretation

The Round 1–Round 4 returned record shows selective planning evolution rather than wholesale replanning. The target chemistry, six-sample structure and core temperature/electrochemical duration were preserved. Round 1 feedback about low recovered solid and sparse coating was followed by higher nominal precursor, base and carbon loading and revised recovery, dispersion and electrode-preparation settings in Round 2. The interrupted Round 2 task was supplemented by a provenance-linked recovery task. Rounds 3 and 4 retained the operating envelope while refining carbon dose and composition comparisons. These transitions are traceable, but multiple simultaneous changes and procedurally mediated recovery tasks limit causal attribution.

Round 4 returned experimental files through the main and linked supplementary tasks. Round 5 extended the planned parameter trajectory and was successfully dispatched as task 2592407497211904, but remained Not started and returned no experimental evidence at the audited snapshot. Round 5 therefore cannot support claims about improvement, convergence, experimental continuation or response to Round 5 outcomes.

## Table S35. Session metadata and record boundary for the closed-loop autonomous laboratory trajectory.

| Field | Value |
|---|---|
| **Session ID** | 019f1297-8537-7b51-8fc6-24d5660b076b |
| **Codex CLI** | 0.142.3 |
| **Model** | gpt-5.5 |
| **Reasoning effort** | xhigh |
| **Model provider** | zenmux |
| **Recorded interval** | 2026-06-29T08:56:41.824Z to 2026-07-11T18:09:06.793Z |
| **Substantive user turns** | 14 |
| **Model-visible assistant outputs** | 256 |
| **Visible reasoning-summary fragments** | 215 fragments across 144 reasoning records |
| **Reasoning records without visible summary** | 125; encrypted/unavailable content not reconstructed |
| **Ordinary tool calls and results** | 731 calls and 731 corresponding results |
| **Web actions** | 8 |
| **Raw session SHA-256** | b0325d2fa9104bedbdbe43362e93f31c2e5840d33e758f5d1d8ea97ed1c82508 |

## Note 13. Closed-loop autonomous laboratory trajectory

## Conversation round 1

**[Step 1 | 2026-06-29T08:56:41.843Z | USER]**

/goal You are an independent PI with authority to design and dispatch scientific tasks to a real self-driving laboratory. The lab covers synthesis, post-processing, performance testing and spectroscopic characterization across photo-, electro-, and thermal-catalysis. Choose, on your own, a scientific problem that is high-impact, high-difficulty, scientifically valuable, and feasibly executable within this lab's capability boundary (single-domain or photo-electro / electro-thermal / photo-electro-thermal coupling), and push it forward through multiple rounds of real experiments.

Each round follows the loop: design → dispatch to Lab 303 (record the returned task ID) → set the goal to blocked → wait for my feedback on whether the dispatched experiment is executable. If it is not executable, revise the design and dispatch again. If it is executable, this means the experiment has been physically completed; you may then use the lab-operation skill to obtain the experimental data, let the data decide the next round, and continue the goal with the next round of experimental design and dispatch.

Requirements:

1. Decide the goal, literature grounding, hypothesis and initial design yourself; do not reuse any existing experimental template.

2. Only data physically returned from Lab 303 counts as valid; dispatch success is not physical-execution success; do not fabricate any experiment or datum that was not designed or not returned.

3. The experimental workflow must include synthesis, testing, and characterization.

4. Use the starting materials available in the current laboratory inventory whenever possible (the storage path of the reagent list is referenced by the environment variable: AIREADY_REAGENT_LIST_DIR). Any unavailable chemicals should be listed separately in the experimental design.

5. Deliver: scientific goal, rationale, key variables, per-round design and data trajectory, data analysis, iteration logic, best condition / material found, new findings and open questions. All process files, data, and deliverables must be placed under the /workspace directory.

**[Step 2 | 2026-06-29T08:56:51.289Z | TOOL CALL: create_goal]**

Created the autonomous-research goal with objective: Autonomously design, dispatch, and iterate multiple rounds of real self-driving-lab catalysis experiments in Lab 303, using returned physical data only and saving all files under /workspace.

**[Step 3 | 2026-06-29T08:56:51.342Z | TOOL RESULT: create_goal]**

Goal service returned a structured response. Status: `active`.

**[Step 4 | 2026-06-29T08:56:56.382Z | MODEL]**

I’ll use the lab-operation skill for Lab 303 dispatch and data retrieval, plus the chemistry workstation guidance to keep the experimental plan inside the lab’s physical capabilities. I’m going to inspect those instructions and the local reagent inventory before choosing the first experiment.

**[Step 5 | 2026-06-29T08:56:56.383Z | TOOL CALL: exec_command]**
Read selected file content for inspection. Target(s): `SKILL.md`. Working directory: `/workspace`.

**[Step 6 | 2026-06-29T08:56:56.383Z | TOOL CALL: exec_command]**
Read selected file content for inspection. Target(s): `SKILL.md`. Working directory: `/workspace`.

**[Step 7 | 2026-06-29T08:56:56.383Z | TOOL CALL: exec_command]**
Enumerated files, directories or available laboratory resources. Working directory: `/workspace`.

**[Step 8 | 2026-06-29T08:56:56.441Z | TOOL RESULT: exec_command]**
Shell tool completed successfully; it returned 116 line(s) (4054 bytes) to the model.

**[Step 9 | 2026-06-29T08:56:56.467Z | TOOL RESULT: exec_command]**
Shell tool completed successfully; it returned 170 line(s) (8332 bytes) to the model. Reported long numeric identifier(s): `1427568512205824`, `1834679312417792`, `1710453843264512`.

**[Step 10 | 2026-06-29T08:56:56.467Z | TOOL RESULT: exec_command]**
Shell tool completed successfully; it returned 4 line(s) (119 bytes) to the model.

**[Step 11 | 2026-06-29T08:57:10.481Z | MODEL]**
The lab operation skill routes dispatch through local scripts, and the workstation rules require a generated workflow with exact workstation IDs and operations. I’m checking the dispatch README, workflow-generator instructions, current Lab 303 workstations, and the reagent spreadsheets now.

**[Step 12 | 2026-06-29T08:57:10.482Z | TOOL CALL: exec_command]**
Read selected file content for inspection. Target(s): `README.md`. Working directory: `/workspace`.

**[Step 13 | 2026-06-29T08:57:10.483Z | TOOL CALL: exec_command]**
Read selected file content for inspection. Target(s): `SKILL.md`. Working directory: `/workspace`.

**[Step 14 | 2026-06-29T08:57:10.485Z | TOOL CALL: exec_command]**
Enumerated files, directories or available laboratory resources. Target(s): `lab-operation`. Working directory: `/workspace`.

**[Step 15 | 2026-06-29T08:57:10.488Z | TOOL CALL: exec_command]**
Ran a shell command to inspect, generate or validate experiment artifacts. Working directory: `/workspace`.

**[Step 16 | 2026-06-29T08:57:10.533Z | TOOL RESULT: exec_command]**
Shell tool completed successfully; it returned 261 line(s) (7869 bytes) to the model.

**[Step 17 | 2026-06-29T08:57:10.551Z | TOOL RESULT: exec_command]**
Shell tool completed successfully; it returned 95 line(s) (2714 bytes) to the model. Reported long numeric identifier(s): `1427568512205824`.

**[Step 18 | 2026-06-29T08:57:10.551Z | TOOL RESULT: exec_command]**
Shell tool completed successfully; it returned 16 line(s) (940 bytes) to the model.

**[Step 19 | 2026-06-29T08:57:10.551Z | TOOL RESULT: exec_command]**
Shell tool completed successfully; it returned 4 line(s) (154 bytes) to the model.

**[Step 20 | 2026-06-29T08:57:17.348Z | TOOL CALL: exec_command]**
Read selected file content for inspection. Target(s): `README.md`. Working directory: `/workspace`.

**[Step 21 | 2026-06-29T08:57:17.411Z | TOOL CALL: exec_command]**
Ran a shell command to inspect, generate or validate experiment artifacts. Working directory: `/workspace`.

**[Step 22 | 2026-06-29T08:57:17.435Z | TOOL CALL: exec_command]**
Ran a shell command to inspect, generate or validate experiment artifacts. Working directory: `/workspace`.

**[Step 23 | 2026-06-29T08:57:17.472Z | TOOL CALL: exec_command]**
Ran a shell command to identify the active laboratory. Target(s): `lab-operation`, `get_current_lab_label.py`. Working directory: `/workspace`.

**[Step 24 | 2026-06-29T08:57:17.481Z | TOOL CALL: exec_command]**
Ran a shell command to query available workstation instances. Target(s): `lab-operation`, `select_workstation_list.py`. Working directory: `/workspace`.

**[Step 25 | 2026-06-29T08:57:17.521Z | TOOL RESULT: exec_command]**
Shell tool completed successfully; it returned 166 line(s) (5566 bytes) to the model.

**[Step 26 | 2026-06-29T08:57:17.522Z | TOOL RESULT: exec_command]**
Shell tool completed successfully; it returned 5 line(s) (137 bytes) to the model.

**[Step 27 | 2026-06-29T08:57:17.522Z | TOOL RESULT: exec_command]**
Shell tool completed successfully; it returned 2 line(s) (23 bytes) to the model.

**[Step 28 | 2026-06-29T08:57:17.606Z | TOOL RESULT: exec_command]**

Shell tool completed successfully; it returned 6 line(s) (97 bytes) to the model.

**[Step 29 | 2026-06-29T08:57:17.606Z | TOOL RESULT: exec_command]**

Shell tool completed with exit code 1; it returned 5 line(s) (281 bytes) to the model.

**[Step 30 | 2026-06-29T08:57:22.493Z | REASONING SUMMARY]**

***Investigating script issues***

*I'm noticing a potential mismatch with the imports since there seems to be no 'utils' file available. The README mentions it, but my search didn't turn up anything. I wonder if the scripts are missing the 'utils' component. Maybe I should check if I can set the PYTHONPATH or if I need to create it locally. Inspection of the script can help clarify how the outputs are structured. I'll figure this out step by step!*

**[Step 31 | 2026-06-29T08:57:25.321Z | MODEL]**

The current lab label is `303Lab`, so dispatch can target Lab 303 directly. One script import is failing because its helper module is absent from the installed skill package; I'm inspecting the script shape before patching a local compatibility helper.

**[Step 32 | 2026-06-29T08:57:25.321Z | TOOL CALL: exec_command]**

Ran a shell command to query available workstation instances. Target(s): `select_workstation_list.py`. Working directory: `/workspace`.

**[Step 33 | 2026-06-29T08:57:25.322Z | TOOL CALL: exec_command]**

Ran a shell command to query workstation types and capabilities. Target(s): `select_workstation_type_list.py`. Working directory: `/workspace`.

**[Step 34 | 2026-06-29T08:57:25.322Z | TOOL CALL: exec_command]**

Ran a shell command to query available laboratory workflow templates. Target(s): `select_template_list.py`. Working directory: `/workspace`.

**[Step 35 | 2026-06-29T08:57:25.323Z | TOOL CALL: exec_command]**

Ran a shell command to generate and dispatch a laboratory task. Target(s): `generate_task.py`. Working directory: `/workspace`.

**[Step 36 | 2026-06-29T08:57:25.323Z | TOOL CALL: exec_command]**

Ran a shell command to start a dispatched laboratory task. Target(s): `start_task.py`. Working directory: `/workspace`.

**[Step 37 | 2026-06-29T08:57:25.325Z | TOOL CALL: exec_command]**

Read selected file content for inspection. Target(s): `config.py`. Working directory: `/workspace`.

**[Step 38 | 2026-06-29T08:57:25.375Z | TOOL RESULT: exec_command]**

Shell tool completed successfully; it returned 186 line(s) (6546 bytes) to the model.

**[Step 39 | 2026-06-29T08:57:25.375Z | TOOL RESULT: exec_command]**

Shell tool completed successfully; it returned 183 line(s) (6326 bytes) to the model.

**[Step 40 | 2026-06-29T08:57:25.375Z | TOOL RESULT: exec_command]**

Shell tool completed successfully; it returned 261 line(s) (9756 bytes) to the model.

**[Step 41 | 2026-06-29T08:57:25.401Z | TOOL RESULT: exec_command]**

Shell tool completed successfully; it returned 137 line(s) (4788 bytes) to the model.

**[Step 42 | 2026-06-29T08:57:25.401Z | TOOL RESULT: exec_command]**

Shell tool completed successfully; it returned 117 line(s) (3847 bytes) to the model.

**[Step 43 | 2026-06-29T08:57:25.402Z | TOOL RESULT: exec_command]**

Shell tool completed successfully; it returned 10 line(s) (313 bytes) to the model.

**[Step 44 | 2026-06-29T08:57:35.463Z | TOOL CALL: exec_command]**

Ran a shell command to query available laboratory workflow templates. Target(s): `select_template_list.py`. Working directory: `/workspace`.

**[Step 45 | 2026-06-29T08:57:35.581Z | TOOL CALL: exec_command]**

Ran a shell command to query laboratory task records and statuses. Target(s): `select_task_list.py`. Working directory: `/workspace`.

**[Step 46 | 2026-06-29T08:57:35.706Z | TOOL CALL: exec_command]**

Ran a shell command to retrieve returned physical-experiment results. Target(s): `get_task_result.py`. Working directory: `/workspace`.

**[Step 47 | 2026-06-29T08:57:35.799Z | TOOL CALL: exec_command]**

Searched files for relevant records, parameters or identifiers. Target(s): `scripts`. Working directory: `/workspace`.

**[Step 48 | 2026-06-29T08:57:35.867Z | TOOL RESULT: exec_command]**

Shell tool completed successfully; it returned 131 line(s) (4958 bytes) to the model.

**[Step 49 | 2026-06-29T08:57:35.867Z | TOOL RESULT: exec_command]**

Shell tool completed successfully; it returned 199 line(s) (6902 bytes) to the model.

**[Step 50 | 2026-06-29T08:57:35.867Z | TOOL RESULT: exec_command]**

Shell tool completed successfully; it returned 179 line(s) (5983 bytes) to the model.

**[Step 51 | 2026-06-29T08:57:35.868Z | TOOL RESULT: exec_command]**

Shell tool completed successfully; it returned 10 line(s) (1093 bytes) to the model.

**[Step 52 | 2026-06-29T08:57:42.793Z | REASONING SUMMARY]**

***Considering file patching***

*I need to add utils.py and wonder if I should use apply patch outside of the /workspace. Does apply patch allow that? I'm thinking about whether the workdir is default or if I can use absolute paths. It seems like the patching might support absolute paths, but I need to consider if it edits files based on their relative location. For instance, I can patch /root/.../utils.py, and it appears I won't need sandbox approval for this.*

**[Step 53 | 2026-06-29T08:57:42.883Z | MODEL]**

I found the missing helper is a simple response-field adapter used by the query scripts. I'm adding that compatibility file so the documented Lab 303 scripts can run as intended.

**[Step 54 | 2026-06-29T08:57:44.706Z | TOOL CALL: apply_patch]**

Applied a source patch to add utils.py.

**[Step 55 | 2026-06-29T08:57:44.745Z | TOOL RESULT: apply_patch]**

Patch tool applied the requested trial-local file change successfully.

**[Step 56 | 2026-06-29T08:57:51.446Z | REASONING SUMMARY]**

***Figuring out query needs***

*I need to work on gathering the types of workstations and parsing the reagent list. It seems I'm missing some libraries. I could use pip to install them, assuming the network is enabled and I get the necessary approval. I wonder if I can install them into the environment. But before going down that path, I should check if the xls file may actually be in a format like HTML or XML. Let's use that file!*

**[Step 57 | 2026-06-29T08:57:51.446Z | TOOL CALL: exec_command]**

Ran a shell command to validate laboratory access and routing. Target(s): `lab-operation`, `check_lab_consistency.py`. Working directory: `/workspace`.

**[Step 58 | 2026-06-29T08:57:51.446Z | TOOL CALL: exec_command]**

Ran a shell command to query available workstation instances. Target(s): `lab-operation`, `select_workstation_list.py`. Working directory: `/workspace`.

**[Step 59 | 2026-06-29T08:57:51.449Z | TOOL CALL: exec_command]**

Ran a shell command to query workstation types and capabilities. Target(s): `lab-operation`, `select_workstation_type_list.py`. Working directory: `/workspace`.

**[Step 60 | 2026-06-29T08:57:51.449Z | TOOL CALL: exec_command]**

Inspected artifact type, size or record counts. Working directory: `/workspace`.

**[Step 61 | 2026-06-29T08:57:51.571Z | TOOL RESULT: exec_command]**

Shell tool completed successfully; it returned 5 line(s) (90 bytes) to the model.

**[Step 62 | 2026-06-29T08:57:51.673Z | TOOL RESULT: exec_command]**

Shell tool completed successfully; it returned 1126 line(s) (40110 bytes) to the model. Reported long numeric identifier(s): `2557089467827201`, `2557089467827200`, `2556537822282752`, `2556537822249984`, `2537211212891137`, `2537211212891136`.

**[Step 63 | 2026-06-29T08:57:51.684Z | TOOL RESULT: exec_command]**

Shell tool completed successfully; it returned 1344 line(s) (37609 bytes) to the model. Reported long numeric identifier(s): `2537195279614976`, `2509306436125696`, `2509306434552832`, `2503414093218816`, `2489336961270784`, `2480758953346048`.

**[Step 64 | 2026-06-29T08:57:51.684Z | TOOL RESULT: exec_command]**

Shell tool completed successfully; it returned 4 line(s) (210 bytes) to the model.

**[Step 65 | 2026-06-29T08:58:07.260Z | REASONING SUMMARY 1/3]**

***Parsing Excel reagents***

*I need to parse a reagent Excel file and I'm considering installing the xlrd package, assuming my network is okay. If not, maybe I can use apt instead. I'll try pip install xlrd, but I also want to use in-memory processing if possible and write output files. The developer mentioned that I shouldn't use commands like cat to write files, but can I redirect command outputs to save data safely? It seems like that would be acceptable for generated data.*

**[Step 66 | 2026-06-29T08:58:07.260Z | REASONING SUMMARY 2/3]**

***Selecting workstation capabilities***

*I'm not doing manual code edits; instead, I'll use Python to parse JSON data and write it out carefully. I need to choose a problem that fits with the available workstation skills. Electrocatalytic urea oxidation reaction (UOR) paired with hydrogen production seems high-impact, but the chemistry skill files don't include a UOR-specific file. The available workstations include the Dual_Station_Electrochemical_Workstation_V2, but I must ensure it aligns with the actual skills listed. Also, could I design electrochemical CO2 reduction as an alternative?*

**[Step 67 | 2026-06-29T08:58:07.260Z | REASONING SUMMARY 3/3]**

***Evaluating synthesis capabilities***

*I need both gas and liquid products for this project, but I'm also considering how to synthesize electrodes. The existing skills for the dual electrochemical workstation support testing for electrode potential and current, which is good. However, can it synthesize catalysts too? I might need*

*containers, liquid handling, solid weighing, and drying ovens for synthesis. I'm thinking of producing catalyst powder and testing it in the electrochemical cell, but I need to check if the dual station skill covers electrode handling.*

**[Step 68 | 2026-06-29T08:58:09.837Z | MODEL]**

The lab has the core stations needed for synthesis, electrochemical testing, chromatography, and XRD/IR/UV-Vis characterization. I'm going to parse the reagent list and read the exact station-operation specs for the short-listed workflow before committing the first round.

**[Step 69 | 2026-06-29T08:58:09.843Z | TOOL CALL: exec_command]**

Installed a Python dependency needed to read or analyse the experiment files. Working directory: `/workspace`.

**[Step 70 | 2026-06-29T08:58:09.844Z | TOOL CALL: exec_command]**

Read selected file content for inspection. Target(s): `SKILL.md`. Working directory: `/workspace`.

**[Step 71 | 2026-06-29T08:58:09.846Z | TOOL CALL: exec_command]**

Read selected file content for inspection. Target(s): `SKILL.md`. Working directory: `/workspace`.

**[Step 72 | 2026-06-29T08:58:09.848Z | TOOL CALL: exec_command]**

Read selected file content for inspection. Target(s): `SKILL.md`. Working directory: `/workspace`.

**[Step 73 | 2026-06-29T08:58:09.848Z | TOOL CALL: exec_command]**

Read selected file content for inspection. Target(s): `SKILL.md`. Working directory: `/workspace`.

**[Step 74 | 2026-06-29T08:58:11.792Z | TOOL RESULT: exec_command]**

Shell tool completed successfully; it returned 5 line(s) (333 bytes) to the model.

**[Step 75 | 2026-06-29T08:58:11.792Z | TOOL RESULT: exec_command]**

Shell tool completed successfully; it returned 99 line(s) (23587 bytes) to the model. Reported long numeric identifier(s): `1390252863751168`, `1509591942430720`, `1549499770798080`, `1571299393601536`, `1339275293851648`, `2164305671290880`.

**[Step 76 | 2026-06-29T08:58:11.792Z | TOOL RESULT: exec_command]**

Shell tool completed successfully; it returned 51 line(s) (2953 bytes) to the model. Reported long numeric identifier(s): `1834679312417792`.

**[Step 77 | 2026-06-29T08:58:11.792Z | TOOL RESULT: exec_command]**

Shell tool completed successfully; it returned 50 line(s) (4283 bytes) to the model. Reported long numeric identifier(s): `1435269745574912`.

**[Step 78 | 2026-06-29T08:58:11.793Z | TOOL RESULT: exec_command]**

Shell tool completed successfully; it returned 39 line(s) (1812 bytes) to the model. Reported long numeric identifier(s): `2270587971011585`.

**[Step 79 | 2026-06-29T08:58:18.113Z | REASONING SUMMARY]**

***Considering workstation skills***

*I need to gather more workstation skill files, including information about the General Material Station, heat-resistant materials, liquid handling techniques, and operations for items like centrifuges, purifiers, and drying ovens. I'll also look into XRD, IR, and UV-Vis, potentially adding photocatalysis too. First, I should check the reagent list to clarify my needs. I plan to use pip installed xlrd to help me parse the data effectively.*

**[Step 80 | 2026-06-29T08:58:18.854Z | TOOL CALL: exec_command]**

Ran Python to parse, transform or validate structured experiment data. Working directory: `/workspace`.

**[Step 81 | 2026-06-29T08:58:18.926Z | TOOL RESULT: exec_command]**

Shell tool completed successfully; it returned 23 line(s) (3029 bytes) to the model.

**[Step 82 | 2026-06-29T08:58:25.073Z | REASONING SUMMARY]**

***Exporting reagent list***

*I need to export a reagent list to a CSV file under /workspace. First, I should inspect the relevant materials and search for names. I'll use Python to filter for common catalysts like bismuth, copper, nickel, iron, cobalt, nitrate, sulfate, urea, KOH, ethanol, water, and Nafion. I want to output the inventory, possibly without using pandas and instead relying on xlrd. I'll create a Python list of rows matching those keywords.*

**[Step 83 | 2026-06-29T08:58:28.652Z | TOOL CALL: exec_command]**

Ran a shell command to inspect, generate or validate experiment artifacts. Working directory: `/workspace`.

**[Step 84 | 2026-06-29T08:58:28.726Z | TOOL RESULT: exec_command]**

Shell tool completed successfully; it returned 172 line(s) (15608 bytes) to the model.

**[Step 85 | 2026-06-29T08:58:40.908Z | TOOL CALL: exec_command]**

Read selected file content for inspection. Target(s): `workflow-generator`. Working directory: `/workspace`.

**[Step 86 | 2026-06-29T08:58:40.950Z | TOOL CALL: exec_command]**

Read selected file content for inspection. Target(s): `chemistry-experiment-workstation`. Working directory: `/workspace`.

**[Step 87 | 2026-06-29T08:58:41.000Z | TOOL CALL: exec_command]**

Read selected file content for inspection. Target(s): `chemistry-experiment-workstation`. Working directory: `/workspace`.

**[Step 88 | 2026-06-29T08:58:41.026Z | TOOL CALL: exec_command]**
Read selected file content for inspection. Target(s): `chemistry-experiment-workstation`. Working directory: `/workspace`.

**[Step 89 | 2026-06-29T08:58:41.081Z | TOOL RESULT: exec_command]**
Shell tool completed successfully; it returned 389 line(s) (13840 bytes) to the model. Reported long numeric identifier(s): `1427568512205824`.

**[Step 90 | 2026-06-29T08:58:41.081Z | TOOL RESULT: exec_command]**
Shell tool completed successfully; it returned 111 line(s) (7236 bytes) to the model.

**[Step 91 | 2026-06-29T08:58:41.082Z | TOOL RESULT: exec_command]**
Shell tool completed successfully; it returned 13 line(s) (1588 bytes) to the model.

**[Step 92 | 2026-06-29T08:58:41.082Z | TOOL RESULT: exec_command]**
Shell tool completed successfully; it returned 47 line(s) (5185 bytes) to the model.

**[Step 93 | 2026-06-29T08:59:13.081Z | WEB: search]**
Searched the web for literature evidence (status: completed). Query set: `NiFe layered double hydroxide urea oxidation reaction electrocatalyst high performance paper`; `NiCoFe layered double hydroxide urea oxidation reaction electrocatalyst alkaline urea electrolysis paper`; `urea oxidation reaction nickel iron oxyhydroxide electrocatalyst mechanism paper`.

**[Step 94 | 2026-06-29T08:59:16.036Z | REASONING SUMMARY]**
***Searching for primary sources***

*I need to find primary sources, especially related to nickel ferrocyanide in Nature Energy. The search results currently show a review instead of a primary source. I know it's crucial to rely on primary sources like research papers or official documentation for technical claims. Reviews can be helpful, but they don't count as primary. I need to focus my search specifically on research papers in Nature Energy. Let's move on and open up the relevant resources.*

**[Step 95 | 2026-06-29T08:59:24.455Z | WEB: search]**
Searched the web for literature evidence (status: completed). Query set: `site:nature.com nickel ferrocyanide high-performance urea oxidation electrocatalyst Nature Energy 2021`; `site:pubs.acs.org nickel iron layered double hydroxide urea oxidation electrocatalyst NiFe LDH UOR`; `site:pubs.rsc.org nickel iron layered double hydroxide urea oxidation reaction catalyst original research`.

**[Step 96 | 2026-06-29T08:59:24.617Z | TOOL CALL: exec_command]**
Read selected file content for inspection. Target(s): `SKILL.md`. Working directory: `/workspace`.

**[Step 97 | 2026-06-29T08:59:24.619Z | TOOL CALL: exec_command]**
Read selected file content for inspection. Target(s): `SKILL.md`. Working directory: `/workspace`.

**[Step 98 | 2026-06-29T08:59:24.620Z | TOOL CALL: exec_command]**
Read selected file content for inspection. Target(s): `SKILL.md`. Working directory: `/workspace`.

**[Step 99 | 2026-06-29T08:59:24.622Z | TOOL CALL: exec_command]**
Read selected file content for inspection. Target(s): `SKILL.md`. Working directory: `/workspace`.

**[Step 100 | 2026-06-29T08:59:24.622Z | TOOL CALL: exec_command]**
Read selected file content for inspection. Target(s): `SKILL.md`. Working directory: `/workspace`.

**[Step 101 | 2026-06-29T08:59:24.625Z | TOOL CALL: exec_command]**
Read selected file content for inspection. Target(s): `SKILL.md`. Working directory: `/workspace`.

**[Step 102 | 2026-06-29T08:59:24.710Z | TOOL RESULT: exec_command]**
Shell tool completed successfully; it returned 39 line(s) (1403 bytes) to the model. Reported long numeric identifier(s): `1427568512205824`.

**[Step 103 | 2026-06-29T08:59:24.710Z | TOOL RESULT: exec_command]**
Shell tool completed successfully; it returned 38 line(s) (1282 bytes) to the model. Reported long numeric identifier(s): `1931561198552064`.

**[Step 104 | 2026-06-29T08:59:24.710Z | TOOL RESULT: exec_command]**
Shell tool completed successfully; it returned 95 line(s) (6547 bytes) to the model. Reported long numeric identifier(s): `2002186824385539`.

**[Step 105 | 2026-06-29T08:59:24.710Z | TOOL RESULT: exec_command]**
Shell tool completed successfully; it returned 46 line(s) (1576 bytes) to the model. Reported long numeric identifier(s): `1832680407172096`.

**[Step 106 | 2026-06-29T08:59:24.711Z | TOOL RESULT: exec_command]**
Shell tool completed successfully; it returned 51 line(s) (2778 bytes) to the model. Reported long numeric identifier(s): `1327807622448128`.

**[Step 107 | 2026-06-29T08:59:24.711Z | TOOL RESULT: exec_command]**
Shell tool completed successfully; it returned 38 line(s) (1388 bytes) to the model. Reported long numeric identifier(s): `1213184425559040`.

**[Step 108 | 2026-06-29T08:59:37.969Z | REASONING SUMMARY]**
***Considering analytical methods***

*I think I need to explore several analytical methods, such as X-ray diffraction (XRD), infrared spectroscopy (IR), and maybe UV-Vis spectroscopy. Each method provides unique insights into material properties, so I want to ensure I'm selecting the right combination. XRD will help with*

*crystallography, while IR and UV-Vis will give me information about molecular structures and electronic transitions. It's all about getting comprehensive data for whatever task lies ahead!*

**[Step 109 | 2026-06-29T08:59:37.969Z | TOOL CALL: exec_command]**
Read selected file content for inspection. Target(s): `SKILL.md`. Working directory: `/workspace`.

**[Step 110 | 2026-06-29T08:59:37.969Z | TOOL CALL: exec_command]**
Read selected file content for inspection. Target(s): `SKILL.md`. Working directory: `/workspace`.

**[Step 111 | 2026-06-29T08:59:37.970Z | TOOL CALL: exec_command]**
Read selected file content for inspection. Target(s): `SKILL.md`. Working directory: `/workspace`.

**[Step 112 | 2026-06-29T08:59:37.970Z | TOOL CALL: exec_command]**
Read selected file content for inspection. Target(s): `SKILL.md`. Working directory: `/workspace`.

**[Step 113 | 2026-06-29T08:59:37.970Z | TOOL CALL: exec_command]**
Read selected file content for inspection. Target(s): `SKILL.md`. Working directory: `/workspace`.

**[Step 114 | 2026-06-29T08:59:37.971Z | TOOL CALL: exec_command]**
Read selected file content for inspection. Target(s): `SKILL.md`. Working directory: `/workspace`.

**[Step 115 | 2026-06-29T08:59:38.020Z | TOOL RESULT: exec_command]**
Shell tool completed successfully; it returned 56 line(s) (3424 bytes) to the model. Reported long numeric identifier(s): `1664870018155520`.

**[Step 116 | 2026-06-29T08:59:38.020Z | TOOL RESULT: exec_command]**
Shell tool completed successfully; it returned 44 line(s) (2187 bytes) to the model. Reported long numeric identifier(s): `1656282895188992`.

**[Step 117 | 2026-06-29T08:59:38.020Z | TOOL RESULT: exec_command]**
Shell tool completed successfully; it returned 38 line(s) (1396 bytes) to the model. Reported long numeric identifier(s): `1664783667397632`.

**[Step 118 | 2026-06-29T08:59:38.021Z | TOOL RESULT: exec_command]**
Shell tool completed successfully; it returned 62 line(s) (3657 bytes) to the model. Reported long numeric identifier(s): `1705412049896448`.

**[Step 119 | 2026-06-29T08:59:38.031Z | TOOL RESULT: exec_command]**
Shell tool completed successfully; it returned 67 line(s) (3846 bytes) to the model. Reported long numeric identifier(s): `1710453843264512`.

**[Step 120 | 2026-06-29T08:59:38.051Z | TOOL RESULT: exec_command]**
Shell tool completed successfully; it returned 51 line(s) (2225 bytes) to the model. Reported long numeric identifier(s): `1421325849887744`.

**[Step 121 | 2026-06-29T08:59:45.020Z | REASONING SUMMARY]**
***Planning equipment needs***

*I recognize that I need to obtain a Spectroscopy transfer and a magnetic stirrer. These pieces of equipment seem crucial for the task at hand, but I should double-check my list to ensure I'm not missing anything else essential. It's important to confirm that everything is in order, so I can proceed efficiently! Let's make sure these items are prioritized so that I can get everything set up correctly.*

**[Step 122 | 2026-06-29T08:59:45.020Z | TOOL CALL: exec_command]**
Read selected file content for inspection. Target(s): `SKILL.md`. Working directory: `/workspace`.

**[Step 123 | 2026-06-29T08:59:45.020Z | TOOL CALL: exec_command]**
Read selected file content for inspection. Target(s): `SKILL.md`. Working directory: `/workspace`.

**[Step 124 | 2026-06-29T08:59:45.021Z | TOOL CALL: exec_command]**
Read selected file content for inspection. Target(s): `SKILL.md`. Working directory: `/workspace`.

**[Step 125 | 2026-06-29T08:59:45.021Z | TOOL CALL: exec_command]**
Read selected file content for inspection. Target(s): `SKILL.md`. Working directory: `/workspace`.

**[Step 126 | 2026-06-29T08:59:45.063Z | TOOL RESULT: exec_command]**
Shell tool completed successfully; it returned 57 line(s) (3186 bytes) to the model. Reported long numeric identifier(s): `1654567694238720`.

**[Step 127 | 2026-06-29T08:59:45.074Z | TOOL RESULT: exec_command]**
Shell tool completed successfully; it returned 35 line(s) (1292 bytes) to the model. Reported long numeric identifier(s): `1656241422599168`.

**[Step 128 | 2026-06-29T08:59:45.088Z | TOOL RESULT: exec_command]**
Shell tool completed successfully; it returned 41 line(s) (1532 bytes) to the model. Reported long numeric identifier(s): `1526864804185088`.

**[Step 129 | 2026-06-29T08:59:45.088Z | TOOL RESULT: exec_command]**
Shell tool completed successfully; it returned 38 line(s) (1373 bytes) to the model. Reported long numeric identifier(s): `2114846484726785`.

**[Step 130 | 2026-06-29T08:59:52.327Z | REASONING SUMMARY]**
***Gathering audit files***

*I need to collect the audit files for the chosen stations. It's essential to ensure that I have the right files to proceed with the task effectively. I wonder if there are specific formats or additional data that are usually included. It would be helpful to clarify which stations need the files and if there's a deadline attached to this request. Let's make sure I'm well-prepared for whatever comes next.*

**[Step 131 | 2026-06-29T08:59:52.328Z | TOOL CALL: exec_command]**

Read selected file content for inspection. Target(s): `General_Material_Station_V1_audit.md`. Working directory: `/workspace`.

**[Step 132 | 2026-06-29T08:59:52.328Z | TOOL CALL: exec_command]**

Read selected file content for inspection. Target(s): `Liquid_Handling_Station_1ml_V2_audit.md`. Working directory: `/workspace`.

**[Step 133 | 2026-06-29T08:59:52.328Z | TOOL CALL: exec_command]**

Read selected file content for inspection. Target(s): `Liquid_Handling_Station_5ml_V1_audit.md`. Working directory: `/workspace`.

**[Step 134 | 2026-06-29T08:59:52.329Z | TOOL CALL: exec_command]**

Read selected file content for inspection. Target(s): `Heating_Magnetic_Stirring_Workstation_V1_audit.md`. Working directory: `/workspace`.

**[Step 135 | 2026-06-29T08:59:52.329Z | TOOL CALL: exec_command]**

Read selected file content for inspection. Target(s): `Dual_Station_Electrochemical_Workstation_V2_audit.md`. Working directory: `/workspace`.

**[Step 136 | 2026-06-29T08:59:52.329Z | TOOL CALL: exec_command]**

Read selected file content for inspection. Target(s): `XRD_V1_audit.md`. Working directory: `/workspace`.

**[Step 137 | 2026-06-29T08:59:52.375Z | TOOL RESULT: exec_command]**

Shell tool completed successfully; it returned 20 line(s) (508 bytes) to the model. Reported long numeric identifier(s): `1427568512205824`.

**[Step 138 | 2026-06-29T08:59:52.420Z | TOOL RESULT: exec_command]**

Shell tool completed successfully; it returned 36 line(s) (987 bytes) to the model. Reported long numeric identifier(s): `2002186824385539`.

**[Step 139 | 2026-06-29T08:59:52.420Z | TOOL RESULT: exec_command]**

Shell tool completed successfully; it returned 16 line(s) (356 bytes) to the model. Reported long numeric identifier(s): `1435269745574912`.

**[Step 140 | 2026-06-29T08:59:52.421Z | TOOL RESULT: exec_command]**

Shell tool completed successfully; it returned 17 line(s) (401 bytes) to the model. Reported long numeric identifier(s): `2270587971011585`.

**[Step 141 | 2026-06-29T08:59:52.421Z | TOOL RESULT: exec_command]**

Shell tool completed successfully; it returned 30 line(s) (831 bytes) to the model. Reported long numeric identifier(s): `1390252863751168`.

**[Step 142 | 2026-06-29T08:59:52.421Z | TOOL RESULT: exec_command]**

Shell tool completed successfully; it returned 26 line(s) (1443 bytes) to the model. Reported long numeric identifier(s): `1664870018155520`.

**[Step 143 | 2026-06-29T09:00:01.659Z | TOOL CALL: exec_command]**

Read selected file content for inspection. Target(s): `Spectroscopy_Container_Transfer_Station_V1_audit.md`. Working directory: `/workspace`.

**[Step 144 | 2026-06-29T09:00:01.683Z | TOOL CALL: exec_command]**

Read selected file content for inspection. Target(s): `Spectroscopy_Magnetic_Stirrer_Workstation_V1_audit.md`. Working directory: `/workspace`.

**[Step 145 | 2026-06-29T09:00:01.782Z | TOOL CALL: exec_command]**

Read selected file content for inspection. Target(s): `Ultrasonic_Disperser_V1_audit.md`. Working directory: `/workspace`.

**[Step 146 | 2026-06-29T09:00:01.800Z | TOOL CALL: exec_command]**

Read selected file content for inspection. Target(s): `Room_Temperture_Magnetic_Stirrer_Workstation_V1_audit.md`. Working directory: `/workspace`.

**[Step 147 | 2026-06-29T09:00:01.866Z | TOOL CALL: exec_command]**

Read selected file content for inspection. Target(s): `Container_storaging_Station_V1_audit.md`. Working directory: `/workspace`.

**[Step 148 | 2026-06-29T09:00:01.917Z | TOOL CALL: exec_command]**

Read selected file content for inspection. Target(s): `Drying_Oven_V1_audit.md`. Working directory: `/workspace`.

**[Step 149 | 2026-06-29T09:00:01.994Z | TOOL RESULT: exec_command]**

Shell tool completed successfully; it returned 16 line(s) (481 bytes) to the model. Reported long numeric identifier(s): `1654567694238720`.

**[Step 150 | 2026-06-29T09:00:01.994Z | TOOL RESULT: exec_command]**

Shell tool completed successfully; it returned 16 line(s) (382 bytes) to the model. Reported long numeric identifier(s): `1656241422599168`.

**[Step 151 | 2026-06-29T09:00:01.994Z | TOOL RESULT: exec_command]**

Shell tool completed successfully; it returned 17 line(s) (389 bytes) to the model. Reported long numeric identifier(s): `1526864804185088`.

**[Step 152 | 2026-06-29T09:00:01.994Z | TOOL RESULT: exec_command]**

Shell tool completed successfully; it returned 18 line(s) (402 bytes) to the model. Reported long numeric identifier(s): `2114846484726785`.

**[Step 153 | 2026-06-29T09:00:01.994Z | TOOL RESULT: exec_command]**

Shell tool completed successfully; it returned 26 line(s) (702 bytes) to the model. Reported long numeric identifier(s): `1421325849887744`.

**[Step 154 | 2026-06-29T09:00:01.995Z | TOOL RESULT: exec_command]**

Shell tool completed successfully; it returned 16 line(s) (352 bytes) to the model. Reported long numeric identifier(s): `1213184425559040`.

**[Step 155 | 2026-06-29T09:00:54.197Z | REASONING SUMMARY 1/3]**

***Considering experimental workflow***

*I need to ensure I have actual workstations and possibly use ID constraints. For the experiment, I'm thinking about using "General_Material_Station_V1" for specific containers only. Consistency in container types is crucial; for instance, if I use 50ml containers, they must be compatible with electrochemistry.*

*I'll need to clarify the cap status for my materials: the first batch consists of 6 empty bottles, while the second holds 1ml of V2 and is open. And I must figure out solid reagent weighing—do I need to upload a file for each substance? It seems I might be limited to one reagent per file, which complicates things.*

**[Step 156 | 2026-06-29T09:00:54.197Z | REASONING SUMMARY 2/3]**

***Exploring stock solutions for synthesis***

*I think I could bypass solid weighing by using pre-made stock solutions through the liquid handling system. Since the chemistry rule indicates that original stock solutions are already prepared, I can use them directly. I can synthesize by mixing NiSO4, FeCl3, CoSO4, KOH, and carbon black suspension from their original bottles, which makes it easier since no solid weighing is involved.*

*I'll need to confirm the initial design list of stock solutions to prepare by lab staff. It looks feasible to mix these solutions using liquid handling, with a maximum of 16 solutions at 3 mL each. For larger quantities like KOH or solvents, I might need 5 mL instead.*

**[Step 157 | 2026-06-29T09:00:54.197Z | REASONING SUMMARY 3/3]**

***Adjusting synthesis approach***

*I need to produce the suspension in the same bottle. It looks like this includes synthesis but doesn't cover purification, which might be fine. I'm considering using 3 mL of the catalyst precursor for electrochemistry, but I need to remember that the input requires more than 4 mL of pure liquid or suspension. So, I should ensure the total volume exceeds 4 mL to meet the requirements. I'll plan accordingly!*

**[Step 158 | 2026-06-29T09:00:59.813Z | WEB: search]**

Searched the web for literature evidence (status: completed). Query set: `NiFe layered double hydroxide urea oxidation reaction electrocatalyst paper DOI`; `NiCoFe LDH urea oxidation reaction electrocatalyst paper DOI`; `nickel iron oxyhydroxide urea oxidation reaction electrocatalyst paper`.

**[Step 159 | 2026-06-29T09:01:06.816Z | WEB: search]**

Searched the web for literature evidence (status: completed). Query set: `Botte urea electrolysis 0.37 V thermodynamic potential original paper`; `electrochemical oxidation of urea low thermodynamic potential 0.37 V paper Botte`; `Electrolysis of urea and urine for solar hydrogen original paper Botte 0.37 V`.

**[Step 160 | 2026-06-29T09:01:38.469Z | WEB: search]**

Searched the web for literature evidence (status: completed). Query set: `trimetallic nickel cobalt iron layered hydroxide urea oxidation reaction electrocatalyst original article`; `NiCoFe layered triple hydroxide urea oxidation hydrogen evolution fuel cells original article`; `NiCoFe-LDH electrocatalyst urea oxidation reaction Ni Co Fe layered double hydroxide paper`.

**[Step 161 | 2026-06-29T09:01:48.313Z | MODEL]**

The first-round design is now narrowed to a six-condition Ni-Fe-Co oxyhydroxide/carbon UOR screen: one-pot alkaline precipitation in vials, XRD drop-cast characterization, then UOR electrochemical testing with the existing UOR test template. I'm saving the design artifacts and then I'll run the workflow parser before dispatching.

**[Step 162 | 2026-06-29T09:01:48.313Z | TOOL CALL: exec_command]**

Ran Python to parse, transform or validate structured experiment data. Target(s): `lab303_uor_nicofe_closed_loop`. Working directory: `/workspace`.

**[Step 163 | 2026-06-29T09:01:48.314Z | TOOL CALL: exec_command]**

Ran a shell command to inspect, generate or validate experiment artifacts. Working directory: `/workspace`.

**[Step 164 | 2026-06-29T09:01:48.318Z | TOOL CALL: exec_command]**

Ran a shell command to query available workstation instances. Target(s): `lab-operation`, `select_workstation_list.py`. Working directory: `/workspace`.

**[Step 165 | 2026-06-29T09:01:48.319Z | TOOL CALL: exec_command]**

Ran a shell command to query available workstation instances. Target(s): `lab-operation`, `select_workstation_list.py`. Working directory: `/workspace`.

**[Step 166 | 2026-06-29T09:01:48.323Z | TOOL CALL: exec_command]**

Ran a shell command to query available workstation instances. Target(s): `lab-operation`, `select_workstation_list.py`. Working directory: `/workspace`.

**[Step 167 | 2026-06-29T09:01:48.420Z | TOOL RESULT: exec_command]**

Shell tool completed successfully; it returned 3 line(s) (160 bytes) to the model.

**[Step 168 | 2026-06-29T09:01:48.421Z | TOOL RESULT: exec_command]**

Shell tool completed successfully; it returned 3 line(s) (63 bytes) to the model.

**[Step 169 | 2026-06-29T09:01:48.494Z | TOOL RESULT: exec_command]**

Shell tool completed successfully; it returned 161 line(s) (5899 bytes) to the model. Reported long numeric identifier(s): `2557089467827201`, `2557089467827200`, `2466434135065601`, `2466434135065600`, `2310943383618561`, `2310943383618560`.

**[Step 170 | 2026-06-29T09:01:48.503Z | TOOL RESULT: exec_command]**

Shell tool completed successfully; it returned 7 line(s) (79 bytes) to the model.

**[Step 171 | 2026-06-29T09:01:48.515Z | TOOL RESULT: exec_command]**

Shell tool completed successfully; it returned 205 line(s) (7201 bytes) to the model. Reported long numeric identifier(s): `2519571946996736`, `2519571946931200`, `2350360248386561`, `2350360248386560`, `2240389375001600`, `2240389097980928`.

**[Step 172 | 2026-06-29T09:01:57.271Z | TOOL CALL: exec_command]**

Ran a shell command to query available workstation instances. Target(s): `lab-operation`, `select_workstation_list.py`. Working directory: `/workspace`.

**[Step 173 | 2026-06-29T09:01:57.271Z | TOOL CALL: exec_command]**

Ran a shell command to query available workstation instances. Target(s): `lab-operation`, `select_workstation_list.py`. Working directory: `/workspace`.

**[Step 174 | 2026-06-29T09:01:57.271Z | TOOL CALL: exec_command]**

Ran a shell command to query available workstation instances. Target(s): `lab-operation`, `select_workstation_list.py`. Working directory: `/workspace`.

**[Step 175 | 2026-06-29T09:01:57.272Z | TOOL CALL: exec_command]**

Ran a shell command to query available workstation instances. Target(s): `lab-operation`, `select_workstation_list.py`. Working directory: `/workspace`.

**[Step 176 | 2026-06-29T09:01:57.560Z | TOOL RESULT: exec_command]**

Shell tool completed successfully; it returned 51 line(s) (1640 bytes) to the model. Reported long numeric identifier(s): `2240389018256384`, `2240388779869184`, `2231506818106368`, `1665850422887424`.

**[Step 177 | 2026-06-29T09:01:57.560Z | TOOL RESULT: exec_command]**

Shell tool completed successfully; it returned 161 line(s) (5483 bytes) to the model. Reported long numeric identifier(s): `2290590477026305`, `2290590477026304`, `2251672084513793`, `2251672084513792`, `2240389287937024`, `2240389225841664`.

**[Step 178 | 2026-06-29T09:01:57.560Z | TOOL RESULT: exec_command]**

Shell tool completed successfully; it returned 51 line(s) (1648 bytes) to the model. Reported long numeric identifier(s): `2279640907449345`, `2279640907449344`, `2240388801364992`, `1665264107783168`.

**[Step 179 | 2026-06-29T09:01:57.561Z | TOOL RESULT: exec_command]**

Shell tool completed successfully; it returned 227 line(s) (7915 bytes) to the model. Reported long numeric identifier(s): `2290590477026305`, `2290590477026304`, `2279640907449345`, `2279640907449344`, `2240389314085888`, `2240389287937024`.

**[Step 180 | 2026-06-29T09:02:06.807Z | REASONING SUMMARY]**

***Considering spectroscopic characterization***

*I'm thinking about whether to include IR characterization in the requirements. XRD seems sufficient on its own for "spectroscopic characterization," but the user wants to ensure we stick with the term. It might be a good idea to add FTIR for analyzing hydroxide and urea bands after XRD. I also realize I need to check if the IR audit is complete and ensure the sample preparation steps are done before running IR. Overall, I should make sure to clarify everything properly for the user!*

**[Step 181 | 2026-06-29T09:02:07.333Z | TOOL CALL: exec_command]**

Read selected file content for inspection. Target(s): `Infrared_Spectrometer_V1_audit.md`. Working directory: `/workspace`.

**[Step 182 | 2026-06-29T09:02:07.388Z | TOOL RESULT: exec_command]**

Shell tool completed successfully; it returned 21 line(s) (511 bytes) to the model. Reported long numeric identifier(s): `1656282895188992`.

**[Step 183 | 2026-06-29T09:02:20.512Z | WEB: search]**

Searched the web for literature evidence (status: completed). Query set: `NiFe layered double hydroxide urea oxidation electrocatalyst original article NiFe LDH UOR`; `NiCoFe layered double hydroxide urea oxidation reaction electrocatalyst original article`; `urea electrolysis thermodynamic potential 0.37 V Botte original article`.

**[Step 184 | 2026-06-29T09:02:25.892Z | WEB: open_page]**

Opened literature page `https://www.nature.com/articles/s41467-025-61356-2` (status: completed).

**[Step 185 | 2026-06-29T09:02:55.615Z | REASONING SUMMARY]**

***Creating JSON with Python***

*I need to create a JSON file using Python. I'm considering whether I need a long heredoc for this. It seems acceptable to write data directly into the file. I want to ensure I include all values properly. I'll take the time to construct the JSON structure carefully to make sure everything fits well and is formatted correctly. Keeping it simple makes it easier to read and maintain!*

**[Step 186 | 2026-06-29T09:02:55.686Z | MODEL]**

I'm writing the Round 1 workflow as a new generated plan, not selecting an existing template. The validation files will capture the station skills used, parameter bounds, lid-state transitions, and the intended data path so the dispatch remains auditable.

**[Step 187 | 2026-06-29T09:04:10.646Z | TOOL CALL: exec_command]**

Ran Python to parse, transform or validate structured experiment data. Target(s): `lab303_uor_nicofe_closed_loop`, `mL`, `isopropanol`. Working directory: `/workspace`.

**[Step 188 | 2026-06-29T09:04:10.701Z | TOOL RESULT: exec_command]**

Shell tool completed successfully; it returned 3 line(s) (136 bytes) to the model.

**[Step 189 | 2026-06-29T09:04:19.722Z | REASONING SUMMARY]**

***Considering JSON validation and workflow generation***

*I think I should validate the JSON using jq, which is a powerful tool for this. I might also want to run a workflow generator, whatever that entails. It seems like showing the results first could be a good step to ensure everything is correct before proceeding. I want to make sure my approach is logical and user-friendly, so I'll keep this in mind as I move forward!*

**[Step 190 | 2026-06-29T09:04:19.722Z | TOOL CALL: exec_command]**

Ran Python to parse, transform or validate structured experiment data. Target(s): `round1_workflow_request.json`, `round1_validated.json`. Working directory: `/workspace`.

**[Step 191 | 2026-06-29T09:04:19.723Z | TOOL CALL: exec_command]**

Ran Python to parse, transform or validate structured experiment data. Target(s): `available_reagents_for_round1.json`. Working directory: `/workspace`.

**[Step 192 | 2026-06-29T09:04:19.723Z | TOOL CALL: exec_command]**

Ran Python to parse, transform or validate structured experiment data. Target(s): `round1_formulations.json`. Working directory: `/workspace`.

**[Step 193 | 2026-06-29T09:04:19.788Z | TOOL RESULT: exec_command]**

Shell tool completed successfully; it returned 18 line(s) (760 bytes) to the model.

**[Step 194 | 2026-06-29T09:04:19.788Z | TOOL RESULT: exec_command]**

Shell tool completed successfully; it returned 27 line(s) (1057 bytes) to the model.

**[Step 195 | 2026-06-29T09:04:19.789Z | TOOL RESULT: exec_command]**

Shell tool completed successfully; it returned 7 line(s) (962 bytes) to the model.

**[Step 196 | 2026-06-29T09:04:29.564Z | REASONING SUMMARY]**

***Fixing reagent list***

*I noticed that the query for isopropanol returned Titanium tetraisopropanolate due to a substring match. I need to fix the available reagents, potentially creating an exact relevant list manually or adjusting what we have. I'll be creating markdown files with apply patch, and I'm wondering if citations are necessary—maybe they're not. The user needs deliverables under the workspace in this iterative round, and I should also write an overview and constraints while using apply_patch.*

**[Step 197 | 2026-06-29T09:04:43.012Z | TOOL CALL: apply_patch]**

Applied a source patch to add project_overview.md.

**[Step 198 | 2026-06-29T09:04:43.048Z | TOOL RESULT: apply_patch]**

Patch tool applied the requested trial-local file change successfully.

**[Step 199 | 2026-06-29T09:04:59.843Z | TOOL CALL: apply_patch]**

Applied a source patch to add round1_design.md.

**[Step 200 | 2026-06-29T09:04:59.883Z | TOOL RESULT: apply_patch]**

Patch tool applied the requested trial-local file change successfully.

**[Step 201 | 2026-06-29T09:05:23.972Z | TOOL CALL: apply_patch]**

Applied a source patch to add round1_constraint_check.md.

**[Step 202 | 2026-06-29T09:05:24.030Z | TOOL RESULT: apply_patch]**

Patch tool applied the requested trial-local file change successfully.

**[Step 203 | 2026-06-29T09:05:31.154Z | REASONING SUMMARY]**

***Correcting reagent data***

*I need to ensure that I have the exact reagent matches in JSON format and make any necessary corrections. It seems like writing a Python script to overwrite the available reagents could be the way to go. This is purely about handling data, so I'm feeling good about tackling it! Just need to keep it organized and clear as I go. Let's get this sorted out efficiently.*

**[Step 204 | 2026-06-29T09:05:33.723Z | TOOL CALL: exec_command]**

Ran Python to parse, transform or validate structured experiment data. Target(s): `lab303_uor_nicofe_closed_loop`. Working directory: `/workspace`.

**[Step 205 | 2026-06-29T09:05:33.779Z | TOOL RESULT: exec_command]**

Shell tool completed successfully; it returned 9 line(s) (159 bytes) to the model.

**[Step 206 | 2026-06-29T09:05:42.323Z | MODEL]**

The workflow JSON is valid and the constraint audit passes locally. I’m now sending the structured steps to the workflow generator service to create a fresh Lab 303 template.

**[Step 207 | 2026-06-29T09:05:43.637Z | TOOL CALL: exec_command]**

Ran Python to parse, transform or validate structured experiment data. Target(s): `generate.py`, `round1_workflow_request.json`, `round1_workflow_generator_output.txt`. Working directory: `/workspace`.

**[Step 208 | 2026-06-29T09:05:44.499Z | TOOL RESULT: exec_command]**

Shell tool completed successfully; it returned 13 line(s) (308 bytes) to the model. Reported task/template identifier(s): `2557370191937536`.

**[Step 209 | 2026-06-29T09:05:50.411Z | REASONING SUMMARY]**

***Generating task and logging***

*I need to generate a task and use the template ID for that. I’m wondering if I should save the output somewhere too. Then, I’ll probably need to query the task. After creating everything, dispatching will be next, but maybe I should also create a dispatch log after that? Just to keep track of things! It seems like the order of operations might really matter here. Let's make sure everything's documented neatly.*

**[Step 210 | 2026-06-29T09:05:50.603Z | MODEL]**

The workflow generator accepted the new plan and returned template ID `2557370191937536`. I’m dispatching that template to Lab 303 now and will record the task ID in the project directory.

**[Step 211 | 2026-06-29T09:05:51.288Z | TOOL CALL: exec_command]**

Ran a shell command to generate and dispatch a laboratory task. Target(s): `lab-operation`, `generate_task.py`, `round1_dispatch_output.json`. Working directory: `/workspace`.

**[Step 212 | 2026-06-29T09:05:51.456Z | TOOL RESULT: exec_command]**

Shell tool completed successfully; it returned 8 line(s) (323 bytes) to the model. Reported task/template identifier(s): `2557370420494337`.

**[Step 213 | 2026-06-29T09:06:00.577Z | TOOL CALL: exec_command]**

Ran Python to parse, transform or validate structured experiment data. Target(s): `lab303_uor_nicofe_closed_loop`. Working directory: `/workspace`.

**[Step 214 | 2026-06-29T09:06:00.939Z | TOOL CALL: exec_command]**

Ran a shell command to query laboratory task records and statuses. Target(s): `lab-operation`, `select_task_list.py`. Working directory: `/workspace`.

**[Step 215 | 2026-06-29T09:06:01.123Z | TOOL CALL: exec_command]**

Enumerated files, directories or available laboratory resources. Target(s): `lab303_uor_nicofe_closed_loop`. Working directory: `/workspace`.

**[Step 216 | 2026-06-29T09:06:01.331Z | TOOL RESULT: exec_command]**

Shell tool completed successfully; it returned 14 line(s) (507 bytes) to the model. Reported task/template identifier(s): `2557370191937536`, `2557370420494337`.

**[Step 217 | 2026-06-29T09:06:01.331Z | TOOL RESULT: exec_command]**

Shell tool completed successfully; it returned 14 line(s) (275 bytes) to the model. Reported long numeric identifier(s): `2557370420494337`.

**[Step 218 | 2026-06-29T09:06:01.331Z | TOOL RESULT: exec_command]**

Shell tool completed successfully; it returned 14 line(s) (796 bytes) to the model.

**[Step 219 | 2026-06-29T09:06:10.058Z | TOOL CALL: update_goal]**

Updated the autonomous-research goal status to `blocked`.

**[Step 220 | 2026-06-29T09:06:10.122Z | TOOL RESULT: update_goal]**

Goal service returned a structured response. Status: `blocked`.

**[Step 221 | 2026-06-29T09:06:28.269Z | REASONING SUMMARY 1/2]**

***Planning for citations***

*I'm considering that I might need to use web searches for citation purposes, especially for the final paragraphs if they're fact-based. I think I should search using known terms to get fresh outputs, and it’s important that any reference IDs are visible in the analysis. This way, I can ensure I provide trustworthy and clear information. It feels crucial to be diligent about citations for accuracy!*

**[Step 222 | 2026-06-29T09:06:28.269Z | REASONING SUMMARY 2/2]**

***Counting citation needs***

*Okay, so I need to think through this: four queries? That sounds logical, but wait, I might be overlooking something. It's important to ensure I'm capturing everything needed for citations. I'll have to consider what information I really need to support the answers and make sure to include proper sourcing. This is tricky! I just want to make sure I get it right and provide clear, accurate references for each point.*

**[Step 223 | 2026-06-29T09:06:41.575Z | WEB: search]**

Searched the web for literature evidence (status: completed). Query set: `Urea electrolysis direct hydrogen production from urine Botte DOI 0.37 V`; `Nature Communications oxyanion engineered nickel sites urea oxidation electrocatalysis 2023`; `High performance durable urea oxidation electrocatalysis scalable room temperature synthesized NiFe layered double hydroxide Nature Communications 2025`; `NiCo LDH hydroxide urea oxidation electrocatalyst ScienceDirect 2024`.

**[Step 224 | 2026-06-29T09:06:42.391Z | REASONING SUMMARY]**

***Clarifying data delivery***

*I'm realizing that maybe I don't need to present a full "data trajectory" just yet since the user is asking for information at the end. I can explain that the current plan for data is in place, but there's no physical data available right now. It's essential to mention that and also note that citations will be included. I'll make sure the user is clear on what to expect.*

**[Step 225 | 2026-06-29T09:06:51.136Z | MODEL]**

Round 1 is designed, generated as a new workflow, dispatched to Lab 303, and the goal is now blocked pending your execution feedback.

**Scientific direction:** I chose alkaline urea oxidation reaction (UOR) as the autonomous project, targeting Ni-Fe-Co oxyhydroxide/carbon catalysts for energy-saving hydrogen production coupled to urea-rich wastewater treatment. The literature basis is that direct urea/urine-to-hydrogen electrolysis has been demonstrated with inexpensive nickel catalysts, modern UOR work still focuses on suppressing competing oxygen evolution at nickel sites, and NiFe/NiCo layered hydroxide families are strong enough precedents to justify a Ni-Fe-Co composition screen rather than a single known recipe. ([pubs.rsc.org](https://pubs.rsc.org/en/content/articlehtml/2021/xx/b905974a))

**Round 1 dispatched:** `Lab303_R1_NiFeCoOx_C_UOR_screen_20260629`

Template ID: `2557370191937536`
Task ID: `2557370420494337`
Task name: `任务_Lab303_R1_NiFeCoOx_C_UOR_screen_20260629`
Lab query status after dispatch: `未开始`

The workflow includes synthesis, characterization, and testing:

- Synthesis: six one-pot alkaline precipitation formulations of Ni-Fe-Co hydroxide/oxyhydroxide on carbon black in 5 mL vial suspensions.
- Characterization: XRD and IR drop-cast characterization from each synthesized suspension.
- Testing: UOR electrochemical testing using Lab 303's existing `UOR test` electrochemical method.

Saved process files:

- [project_overview.md](/workspace/lab303_uor_nicofe_closed_loop/project_overview.md)
- [round1_design.md](/workspace/lab303_uor_nicofe_closed_loop/round1_design.md)
- [round1_constraint_check.md](/workspace/lab303_uor_nicofe_closed_loop/round1_constraint_check.md)
- [round1_workflow_request.json](/workspace/lab303_uor_nicofe_closed_loop/round1_workflow_request.json)
- [round1_workflow_generator_output.txt](/workspace/lab303_uor_nicofe_closed_loop/round1_workflow_generator_output.txt)
- [round1_dispatch_output.json](/workspace/lab303_uor_nicofe_closed_loop/round1_dispatch_output.json)
- [round1_dispatch_log.json](/workspace/lab303_uor_nicofe_closed_loop/round1_dispatch_log.json)
- [round1_formulations.json](/workspace/lab303_uor_nicofe_closed_loop/round1_formulations.json)
- [available_reagents_for_round1.json](/workspace/lab303_uor_nicofe_closed_loop/available_reagents_for_round1.json)
- [reagent_inventory.csv](/workspace/lab303_uor_nicofe_closed_loop/reagent_inventory.csv)

No physical experimental data have been assumed or fabricated. Dispatch success only means the task was created in Lab 303. Please tell me whether task `2557370420494337` is executable. If it is executable, I will treat that as physical completion per your loop, retrieve the returned Lab 303 data with `lab-operation`, analyze it, and design/dispatch Round 2 from the physical results only.

## Conversation round 2

**[Step 226 | 2026-06-29T10:01:24.356Z | USER]**

not executable. Reasons for the error:
1. After synthesis, the sample was characterized and tested directly without any post-treatment.
2. After the electrochemical test, the products were not detected or analyzed.

**[Step 227 | 2026-06-29T10:01:46.065Z | MODEL]**

I'll revise the first round rather than trying to force the rejected workflow through. The fix is to insert real post-treatment between synthesis and characterization/testing, and add post-electrochemical product analysis before final storage.

**[Step 228 | 2026-06-29T10:01:46.067Z | TOOL CALL: exec_command]**

Read selected file content for inspection. Target(s): `SKILL.md`. Working directory: `/workspace`.

**[Step 229 | 2026-06-29T10:01:46.068Z | TOOL CALL: exec_command]**

Read selected file content for inspection. Target(s): `SKILL.md`. Working directory: `/workspace`.

**[Step 230 | 2026-06-29T10:01:46.070Z | TOOL CALL: exec_command]**

Read selected file content for inspection. Target(s): `SKILL.md`. Working directory: `/workspace`.

**[Step 231 | 2026-06-29T10:01:46.135Z | TOOL RESULT: exec_command]**

Shell tool completed successfully; it returned 116 line(s) (4054 bytes) to the model.

**[Step 232 | 2026-06-29T10:01:46.136Z | TOOL RESULT: exec_command]**

Shell tool completed successfully; it returned 170 line(s) (8332 bytes) to the model. Reported long numeric identifier(s): `1427568512205824`, `1834679312417792`, `1710453843264512`.

**[Step 233 | 2026-06-29T10:01:46.136Z | TOOL RESULT: exec_command]**

Shell tool completed successfully; it returned 95 line(s) (2714 bytes) to the model. Reported long numeric identifier(s): `1427568512205824`.

**[Step 234 | 2026-06-29T10:02:02.628Z | REASONING SUMMARY 1/2]**

***Considering UV-Vis workflow***

*I'm exploring the requirements for UV-Vis spectroscopy, which needs a 50ml heat-resistant sample bottle with at least 15 mL of pure liquid. The electrochemical station outputs pure liquid for analysis, and I wonder if UV-Vis can detect products like Ni2+. It seems like I might add UV-Vis testing after electrochem, and there could be limitations on using a microplate reader. I need to read up on purification, drying, and the workflow to finalize my approach.*

**[Step 235 | 2026-06-29T10:02:02.628Z | REASONING SUMMARY 2/2]**

***Deciding sample volume for XRD/IR***

*I'm considering the sample size needed for XRD and IR analysis. I initially thought 1 mL wouldn't be sufficient, so I might need around 4.2 mL instead. It's important to ensure that the volume is adequate for accurate readings. I'll need to confirm this volume to avoid any potential errors in the testing process. Getting the right measurements is crucial, and I want to make sure I'm fully prepared for the analysis.*

**[Step 236 | 2026-06-29T10:02:06.457Z | MODEL]**

I'm selecting the minimum extra stations needed to answer the two execution objections: purification/drying/redispersion after synthesis, and post-electrochem UV-Vis analysis of the spent electrolyte. I'm rereading those station specs and their audit constraints before editing the workflow.

**[Step 237 | 2026-06-29T10:02:06.457Z | TOOL CALL: exec_command]**

Read selected file content for inspection. Target(s): `SKILL.md`. Working directory: `/workspace`.

**[Step 238 | 2026-06-29T10:02:06.458Z | TOOL CALL: exec_command]**

Read selected file content for inspection. Target(s): `SKILL.md`. Working directory: `/workspace`.

**[Step 239 | 2026-06-29T10:02:06.459Z | TOOL CALL: exec_command]**

Read selected file content for inspection. Target(s): `SKILL.md`. Working directory: `/workspace`.

**[Step 240 | 2026-06-29T10:02:06.459Z | TOOL CALL: exec_command]**

Read selected file content for inspection. Target(s): `SKILL.md`. Working directory: `/workspace`.

**[Step 241 | 2026-06-29T10:02:06.460Z | TOOL CALL: exec_command]**

Read selected file content for inspection. Target(s): `Purification_Workstation_V1_audit.md`. Working directory: `/workspace`.

**[Step 242 | 2026-06-29T10:02:06.460Z | TOOL CALL: exec_command]**

Read selected file content for inspection. Target(s): `UV_Vis_Spectrometer_V1_audit.md`. Working directory: `/workspace`.

**[Step 243 | 2026-06-29T10:02:06.562Z | TOOL RESULT: exec_command]**

Shell tool completed successfully; it returned 51 line(s) (2778 bytes) to the model. Reported long numeric identifier(s): `1327807622448128`.

**[Step 244 | 2026-06-29T10:02:06.565Z | TOOL RESULT: exec_command]**

Shell tool completed successfully; it returned 38 line(s) (1388 bytes) to the model. Reported long numeric identifier(s): `1213184425559040`.

**[Step 245 | 2026-06-29T10:02:06.631Z | TOOL RESULT: exec_command]**

Shell tool completed successfully; it returned 50 line(s) (4283 bytes) to the model. Reported long numeric identifier(s): `1435269745574912`.

**[Step 246 | 2026-06-29T10:02:06.631Z | TOOL RESULT: exec_command]**

Shell tool completed successfully; it returned 38 line(s) (1396 bytes) to the model. Reported long numeric identifier(s): `1664783667397632`.

**[Step 247 | 2026-06-29T10:02:06.632Z | TOOL RESULT: exec_command]**

Shell tool completed successfully; it returned 15 line(s) (405 bytes) to the model. Reported long numeric identifier(s): `1327807622448128`.

**[Step 248 | 2026-06-29T10:02:06.632Z | TOOL RESULT: exec_command]**

Shell tool completed successfully; it returned 18 line(s) (456 bytes) to the model. Reported long numeric identifier(s): `1664783667397632`.

**[Step 249 | 2026-06-29T10:02:58.110Z | REASONING SUMMARY 1/7]**

***Considering analysis methods***

*I'm thinking about whether I need to review existing relevant data. I need to decide on exact modifications, like using UV-Vis after the electrochemical test. But I'm unsure if it can detect the UOR products. It might be helpful to analyze the post-electrolyte solutions with both UV-Vis and IR, looking for key compounds like nitrate or nitrite. If absorbance isn't specific enough, perhaps adding gas chromatography could work for detecting gaseous products like N2 and CO2, especially since we haven't collected gas. Hmm, switching to LC/GC could be necessary if I don't get analyte responses.*

**[Step 250 | 2026-06-29T10:02:58.110Z | REASONING SUMMARY 2/7]**

***Analyzing reagent options***

*I'm considering various reagents like potassium ferricyanide, but there's no chromogenic product assay available. Alternatively, I could use Liquid Chromatography to analyze urea concentration, though I need to transfer it to a specific liquid phase format. The electrochemical station can waste liquid samples into chromatographic injection vials if configured correctly. However, the requirements for gas and liquid phases might conflict. It seems we might need to explore our post-reaction processing options further, especially for analyzing urea remaining from the UOR process, but our container situation complicates things a bit.*

**[Step 251 | 2026-06-29T10:02:58.110Z | REASONING SUMMARY 3/7]**

***Evaluating electrolyte analysis***

*I'm collecting spent electrolyte into a chromatographic injection vial, but it feels like there's a hiccup since there's no downstream instrument. The user seems worried that products aren't being detected after electrochemical tests. If we gather samples and analyze the original vials with UV-Vis, we might detect products. I wonder about retaining liquid samples in echem for offline analysis. However, this could add some complexity to the workflow. I'll consider using UV-Vis before and after electrochem to quantify changes, but I need enough volume in the original vials.*

**[Step 252 | 2026-06-29T10:02:58.110Z | REASONING SUMMARY 4/7]**

***Planning post-treatment steps***

*Alright, for post-treatment, I want to go through purification first, then drying, and finally redisperse. I'm realizing that, after synthesis, the output is only 5 mL total, and purification requires washing with water/ethanol for three cycles. I need to get a solid output that's capped during purification. When drying, I'll use a drying oven at 60°C for 120 minutes. After that, I'll add 4.5 mL of a solvent like isopropanol/water for redispersion. I should stir or use ultrasound for dispersion afterward! This should be organized effectively for spectroscopy analysis.*

**[Step 253 | 2026-06-29T10:02:58.110Z | REASONING SUMMARY 5/7]**

***Clarifying output requirements***

*I'm thinking about the requirement for the electrochemical UV-Vis analysis needing pure liquid. But then, I wonder, after the electrochemical process, does that "pure liquid" still meet the criteria even if there's catalyst deposition on the carbon paper that's been removed? I need to confirm if using the original vials after this process is acceptable or if there's a specific protocol I'm missing. It's crucial I get this detail right to ensure the analysis is valid!*

**[Step 254 | 2026-06-29T10:02:58.110Z | REASONING SUMMARY 6/7]**

***Evaluating Python usage***

*I'm considering the guidelines that say not to use Python for reading or writing files unless absolutely necessary, preferring simple commands instead. Generating complex JSON from previous structures makes sense. The commentary tool has the exec command. I need to be mindful about not making things executable. We've used Python before, but maybe we should avoid using web citations this time since previous results didn't return. We're working without web citations at this turn.*

**[Step 255 | 2026-06-29T10:02:58.110Z | REASONING SUMMARY 7/7]**

***Assessing final citations***

*I'm thinking that the final output shouldn't include citations unless we're making literature claims that extend beyond what's already established. I should check the previous file for line references, but it might not be necessary. It's possible that the workflow parser could reject the generation due to missing operations based on parameter names. I need to ensure I'm using the operation names precisely, especially for purification parameters, and consider using arrays for clarity.*

**[Step 256 | 2026-06-29T10:02:59.375Z | TOOL CALL: update_plan]**

Updated the working plan. Plan state: in_progress: Revise Round 1 workflow to add post-treatment and post-electrochem product analysis; pending: Regenerate a fresh Lab 303 workflow template; pending: Dispatch revised task and record IDs/files; pending: Block goal pending executability feedback

**[Step 257 | 2026-06-29T10:02:59.427Z | TOOL RESULT: update_plan]**

Plan service accepted and recorded the updated working plan.

**[Step 258 | 2026-06-29T10:03:05.447Z | MODEL]**

The revised workflow keeps the same six composition points, but the material path changes: synthesis -> purification/wash -> drying -> redispersion -> XRD/IR -> electrochemical UOR -> UV-Vis product analysis. I'm writing that as a new `R1_revised` workflow and explicitly marking the first task as not executable.

**[Step 259 | 2026-06-29T10:03:10.776Z | TOOL CALL: exec_command]**

Ran Python to parse, transform or validate structured experiment data. Target(s): `lab303_uor_nicofe_closed_loop`, `centrifugal`, `testing.`. Working directory: `/workspace`.

**[Step 260 | 2026-06-29T10:03:10.839Z | TOOL RESULT: exec_command]**

Shell tool completed successfully; it returned 2 line(s) (72 bytes) to the model.

**[Step 261 | 2026-06-29T10:04:05.420Z | TOOL CALL: exec_command]**

Ran Python to parse, transform or validate structured experiment data. Target(s): `lab303_uor_nicofe_closed_loop`, `mL`, `isopropanol`, `ethanol`, `min`. Working directory: `/workspace`.

**[Step 262 | 2026-06-29T10:04:05.485Z | TOOL RESULT: exec_command]**

Shell tool completed successfully; it returned 3 line(s) (87 bytes) to the model.

**[Step 263 | 2026-06-29T10:04:14.192Z | TOOL CALL: exec_command]**

Ran Python to parse, transform or validate structured experiment data. Target(s): `round1_revised_workflow_request.json`, `round1_revised_validated.json`. Working directory: `/workspace`.

**[Step 264 | 2026-06-29T10:04:14.201Z | TOOL CALL: exec_command]**

Ran Python to parse, transform or validate structured experiment data. Target(s): `round1_revised_workflow_request.json`. Working directory: `/workspace`.

**[Step 265 | 2026-06-29T10:04:14.297Z | TOOL RESULT: exec_command]**

Shell tool completed successfully; it returned 24 line(s) (1010 bytes) to the model.

**[Step 266 | 2026-06-29T10:04:14.297Z | TOOL RESULT: exec_command]**

Shell tool completed successfully; it returned 2 line(s) (246 bytes) to the model. Reported long numeric identifier(s): `1213184425559040`, `1327807622448128`, `1390252863751168`, `1421325849887744`, `1427568512205824`, `1526864804185088`.

**[Step 267 | 2026-06-29T10:04:20.754Z | REASONING SUMMARY]**

***Planning design revisions***

*I need to create revised design and constraint check documentation. I should apply the necessary patch to ensure everything is updated correctly. It's probably a good idea to include an explanation of why this fix is needed. All this should be organized under the /workspace directory for easy access later on. Sorting this out will help me keep track of changes and ensure clarity in the process. Alright, let's get started!*

**[Step 268 | 2026-06-29T10:04:31.428Z | TOOL CALL: apply_patch]**

Applied a source patch to add round1_revised_design.md.

**[Step 269 | 2026-06-29T10:04:31.457Z | TOOL RESULT: apply_patch]**

Patch tool applied the requested trial-local file change successfully.

**[Step 270 | 2026-06-29T10:04:56.844Z | TOOL CALL: apply_patch]**

Applied a source patch to add round1_revised_constraint_check.md.

**[Step 271 | 2026-06-29T10:04:56.882Z | TOOL RESULT: apply_patch]**

Patch tool applied the requested trial-local file change successfully.

**[Step 272 | 2026-06-29T10:05:01.857Z | TOOL CALL: update_plan]**

Updated the working plan. Plan state: completed: Revise Round 1 workflow to add post-treatment and post-electrochem product analysis; in_progress: Regenerate a fresh Lab 303 workflow template; pending: Dispatch revised task and record IDs/files; pending: Block goal pending executability feedback

**[Step 273 | 2026-06-29T10:05:01.891Z | TOOL RESULT: update_plan]**

Plan service accepted and recorded the updated working plan.

**[Step 274 | 2026-06-29T10:05:06.811Z | MODEL]**

The revision is now written and locally checked. I'm generating a fresh workflow template from `round1_revised_workflow_request.json`; this avoids mutating the rejected task and gives Lab 303 a clean task to evaluate.

**[Step 275 | 2026-06-29T10:05:08.310Z | TOOL CALL: exec_command]**

Ran Python to parse, transform or validate structured experiment data. Target(s): `generate.py`, `round1_revised_workflow_request.json`, `round1_revised_workflow_generator_output.txt`. Working directory: `/workspace`.

**[Step 276 | 2026-06-29T10:05:09.432Z | TOOL RESULT: exec_command]**

Shell tool completed successfully; it returned 13 line(s) (308 bytes) to the model. Reported task/template identifier(s): `2557487004614656`.

**[Step 277 | 2026-06-29T10:05:15.116Z | TOOL CALL: update_plan]**

Updated the working plan. Plan state: completed: Revise Round 1 workflow to add post-treatment and post-electrochem product analysis; completed: Regenerate a fresh Lab 303 workflow template; in_progress: Dispatch revised task and record IDs/files; pending: Block goal pending executability feedback

**[Step 278 | 2026-06-29T10:05:15.149Z | TOOL RESULT: update_plan]**

Plan service accepted and recorded the updated working plan.

**[Step 279 | 2026-06-29T10:05:19.536Z | MODEL]**

The revised workflow generator succeeded with template ID `2557487004614656`. I'm dispatching that revised template to Lab 303 and then I'll record the new task ID alongside the rejected task history.

**[Step 280 | 2026-06-29T10:05:22.373Z | TOOL CALL: exec_command]**

Ran a shell command to generate and dispatch a laboratory task. Target(s): `lab-operation`, `generate_task.py`, `round1_revised_dispatch_output.json`. Working directory: `/workspace`.

**[Step 281 | 2026-06-29T10:05:22.565Z | TOOL RESULT: exec_command]**

Shell tool completed successfully; it returned 8 line(s) (359 bytes) to the model. Reported task/template identifier(s): `2557487435546625`.

**[Step 282 | 2026-06-29T10:05:32.015Z | TOOL CALL: exec_command]**

Ran Python to parse, transform or validate structured experiment data. Target(s): `lab303_uor_nicofe_closed_loop`, `electrochemical`, `electrolyte`. Working directory: `/workspace`.

**[Step 283 | 2026-06-29T10:05:32.026Z | TOOL CALL: exec_command]**

Ran a shell command to query laboratory task records and statuses. Target(s): `lab-operation`, `select_task_list.py`. Working directory: `/workspace`.

**[Step 284 | 2026-06-29T10:05:32.046Z | TOOL CALL: exec_command]**

Read selected file content for inspection. Target(s): `lab303_uor_nicofe_closed_loop`. Working directory: `/workspace`.

**[Step 285 | 2026-06-29T10:05:32.130Z | TOOL RESULT: exec_command]**

Shell tool completed successfully; it returned 21 line(s) (850 bytes) to the model. Reported task/template identifier(s): `2557370420494337`, `2557487004614656`, `2557487435546625`.

**[Step 286 | 2026-06-29T10:05:32.221Z | TOOL RESULT: exec_command]**

Shell tool completed successfully; it returned 14 line(s) (293 bytes) to the model. Reported long numeric identifier(s): `2557487435546625`.

**[Step 287 | 2026-06-29T10:05:32.221Z | TOOL RESULT: exec_command]**

Shell tool completed successfully; it returned 22 line(s) (1415 bytes) to the model.

**[Step 288 | 2026-06-29T10:05:37.981Z | TOOL CALL: update_plan]**

Updated the working plan. Plan state: completed: Revise Round 1 workflow to add post-treatment and post-electrochem product analysis; completed: Regenerate a fresh Lab 303 workflow template; completed: Dispatch revised task and record IDs/files; in_progress: Block goal pending executability feedback

**[Step 289 | 2026-06-29T10:05:38.042Z | TOOL RESULT: update_plan]**

Plan service accepted and recorded the updated working plan.

**[Step 290 | 2026-06-29T10:05:45.469Z | TOOL CALL: update_goal]**

Updated the autonomous-research goal status to `blocked`.

**[Step 291 | 2026-06-29T10:05:45.541Z | TOOL RESULT: update_goal]**

Goal service returned a structured response. Status: `blocked`.

**[Step 292 | 2026-06-29T10:05:51.389Z | REASONING SUMMARY]**

***Evaluating plan updates***

*I'm considering whether to mark the plan as complete. Even though there's a block on the goal, there's still progress on a plan item. I wonder if I should categorize it as completed? I think I'll label it "Block goal pending..." after I've updated the goal. Maybe I should call for an update to the plan as well. It feels like there's a few steps to navigate here!*

**[Step 293 | 2026-06-29T10:05:51.496Z | TOOL CALL: update_plan]**

Updated the working plan. Plan state: completed: Revise Round 1 workflow to add post-treatment and post-electrochem product analysis; completed: Regenerate a fresh Lab 303 workflow template; completed: Dispatch revised task and record IDs/files; completed: Block goal pending executability feedback

**[Step 294 | 2026-06-29T10:05:51.518Z | TOOL RESULT: update_plan]**

Plan service accepted and recorded the updated working plan.

**[Step 295 | 2026-06-29T10:06:00.782Z | MODEL]**

Revised Round 1 has been generated and dispatched.

Rejected task recorded:

- Rejected task ID: `2557370420494337`

- Rejected template ID: `2557370191937536`
- Reasons saved in [round1_rejection_feedback.json](/workspace/lab303_uor_nicofe_closed_loop/round1_rejection_feedback.json)

New revised task:

- Revised template ID: `2557487004614656`
- Revised task ID: `2557487435546625`
- Task name: `任务_Lab303_R1rev_NiFeCoOx_C_UOR_posttreated_productUV_20260629`
- Lab query status after dispatch: `未开始`

Main changes made:

- Added post-synthesis purification/washing using `纯化工作站_V1`.
- Added drying using `烘干机_V1`.
- Added redispersion and ultrasonic dispersion before XRD/IR/electrochemical testing.
- Added post-electrochemical UV-Vis analysis of the spent electrolyte/product solution before final storage.

Updated files:

- [round1_revised_design.md](/workspace/lab303_uor_nicofe_closed_loop/round1_revised_design.md)
- [round1_revised_constraint_check.md](/workspace/lab303_uor_nicofe_closed_loop/round1_revised_constraint_check.md)
- [round1_revised_workflow_request.json](/workspace/lab303_uor_nicofe_closed_loop/round1_revised_workflow_request.json)
- [round1_revised_workflow_generator_output.txt](/workspace/lab303_uor_nicofe_closed_loop/round1_revised_workflow_generator_output.txt)
- [round1_revised_dispatch_output.json](/workspace/lab303_uor_nicofe_closed_loop/round1_revised_dispatch_output.json)
- [round1_revised_dispatch_log.json](/workspace/lab303_uor_nicofe_closed_loop/round1_revised_dispatch_log.json)
- [round1_revised_formulations.json](/workspace/lab303_uor_nicofe_closed_loop/round1_revised_formulations.json)

I have blocked the goal again pending your executability feedback. No physical execution or data are assumed yet. Please confirm whether revised task `2557487435546625` is executable.

## Conversation round 3

**[Step 296 | 2026-06-29T12:50:36.438Z | USER]**

Not executable.
Reasons for the error:
1. For purification centrifugation, the washing liquid parameter settings do not match the washing count settings.
2. The product sample retention settings for the electrochemical test are incorrect.

**[Step 297 | 2026-06-29T12:50:53.999Z | REASONING SUMMARY]**

***Identifying configuration corrections***

*I need to figure out exact corrections for two issues. First, the purification centrifugation settings don't match the washing count; I have only two entries while the count is set to three. I should probably adjust it, either by reducing the count to two or adding a third entry like water, water, ethanol.*

*Second, the sample retention settings for the electrochemical test are incorrect. I need to configure the product sample retention because the user wants the detected products analyzed. I might need to check the skill for examples on this.*

**[Step 298 | 2026-06-29T12:50:54.470Z | MODEL]**

I'll correct both concrete validation failures and dispatch a second revised task. The purification mismatch is straightforward: make the wash-liquid list length match the wash count; the electrochemical issue means enabling and explicitly configuring product/sample retention rather than leaving the retention fields blank.

**[Step 299 | 2026-06-29T12:50:54.471Z | TOOL CALL: exec_command]**

Read selected file content for inspection. Target(s): `SKILL.md`. Working directory: `/workspace`.

**[Step 300 | 2026-06-29T12:50:54.472Z | TOOL CALL: exec_command]**

Read selected file content for inspection. Target(s): `SKILL.md`. Working directory: `/workspace`.

**[Step 301 | 2026-06-29T12:50:54.473Z | TOOL CALL: exec_command]**

Read selected file content for inspection. Target(s): `SKILL.md`. Working directory: `/workspace`.

**[Step 302 | 2026-06-29T12:50:54.474Z | TOOL CALL: exec_command]**

Read selected file content for inspection. Target(s): `SKILL.md`. Working directory: `/workspace`.

**[Step 303 | 2026-06-29T12:50:54.475Z | TOOL CALL: exec_command]**

Read selected file content for inspection. Target(s): `SKILL.md`. Working directory: `/workspace`.

**[Step 304 | 2026-06-29T12:50:54.563Z | TOOL RESULT: exec_command]**

Shell tool completed successfully; it returned 116 line(s) (4054 bytes) to the model.

**[Step 305 | 2026-06-29T12:50:54.563Z | TOOL RESULT: exec_command]**

Shell tool completed successfully; it returned 170 line(s) (8332 bytes) to the model. Reported long numeric identifier(s): `1427568512205824`, `1834679312417792`, `1710453843264512`.

**[Step 306 | 2026-06-29T12:50:54.563Z | TOOL RESULT: exec_command]**

Shell tool completed successfully; it returned 95 line(s) (2714 bytes) to the model. Reported long numeric identifier(s): `1427568512205824`.

**[Step 307 | 2026-06-29T12:50:54.564Z | TOOL RESULT: exec_command]**

Shell tool completed successfully; it returned 51 line(s) (2778 bytes) to the model. Reported long numeric identifier(s): `1327807622448128`.

**[Step 308 | 2026-06-29T12:50:54.564Z | TOOL RESULT: exec_command]**

Shell tool completed successfully; it returned 99 line(s) (23587 bytes) to the model. Reported long numeric identifier(s): `1390252863751168`, `1509591942430720`, `1549499770798080`, `1571299393601536`, `1339275293851648`, `2164305671290880`.

**[Step 309 | 2026-06-29T12:51:17.443Z | MODEL]**

The purification error is because `清洗次数` was 3 while `清洗液设置` only had two entries. For electrochemistry I'm switching from "no retention" to explicit retained product aliquots with nonempty retention-vial IDs and the required `留样信息列表` fields on each active UOR task.

**[Step 310 | 2026-06-29T12:51:17.443Z | TOOL CALL: exec_command]**

Read selected file content for inspection. Target(s): `Dual_Station_Electrochemical_Workstation_V2_audit.md`. Working directory: `/workspace`.

**[Step 311 | 2026-06-29T12:51:17.445Z | TOOL CALL: exec_command]**

Read selected file content for inspection. Target(s): `Purification_Workstation_V1_audit.md`. Working directory: `/workspace`.

**[Step 312 | 2026-06-29T12:51:17.446Z | TOOL CALL: exec_command]**

Searched files for relevant records, parameters or identifiers. Target(s): `skills`, `lab303_uor_nicofe_closed_loop`. Working directory: `/workspace`.

**[Step 313 | 2026-06-29T12:51:17.446Z | TOOL CALL: exec_command]**

Ran Python to parse, transform or validate structured experiment data. Target(s): `round1_revised_workflow_request.json`. Working directory: `/workspace`.

**[Step 314 | 2026-06-29T12:51:17.495Z | TOOL RESULT: exec_command]**

Shell tool completed successfully; it returned 30 line(s) (831 bytes) to the model. Reported long numeric identifier(s): `1390252863751168`.

**[Step 315 | 2026-06-29T12:51:17.495Z | TOOL RESULT: exec_command]**

Shell tool completed successfully; it returned 15 line(s) (405 bytes) to the model. Reported long numeric identifier(s): `1327807622448128`.

**[Step 316 | 2026-06-29T12:51:17.595Z | TOOL RESULT: exec_command]**

Shell tool completed successfully; it returned 35 line(s) (8638 bytes) to the model.

**[Step 317 | 2026-06-29T12:51:17.596Z | TOOL RESULT: exec_command]**

Shell tool completed successfully; it returned 122 line(s) (3492 bytes) to the model. Reported long numeric identifier(s): `1390252863751168`, `2468876630032384`.

**[Step 318 | 2026-06-29T12:51:39.277Z | REASONING SUMMARY 1/3]**

***Considering JSON edits***

*I need to edit some JSON. I'm thinking about whether I can use Python. There's advice against using Python for file writing when simpler methods like "apply_patch" suffice. Manual edits can be difficult for large files. It seems using Python for bulk mechanical rewrites might be okay since it's easier for structured JSON. I should be cautious and plan to create a new version file rather than editing the current one.*

**[Step 319 | 2026-06-29T12:51:39.277Z | REASONING SUMMARY 2/3]**

***Modifying JSON settings***

*I'm working on modifying a previous JSON by updating step 6 and step 18 about product sample retention settings. There's a need to see if I should drop UV-vis. While we could keep it with retention, the retained product in color vials might not be analyzed properly. If electrochemical outputs are pure liquid, maybe UV-vis can still analyze it. Also, I need to consider dilutions for analysis and ensure the right volumes are used for UV-vis.*

**[Step 320 | 2026-06-29T12:51:39.277Z | REASONING SUMMARY 3/3]**

***Analyzing retention fields***

*I think the product analysis is done. The validator should only check retention fields now. I need to ensure that retention bottles are nonempty and that I have a waste list prepared. I might want to search for tasks or templates that involve the right electrochemical processes. Maybe I can use a query template that already exists, but the user's instructions suggest I shouldn’t reuse them. I should consider searching locally instead.*

**[Step 321 | 2026-06-29T12:51:39.581Z | MODEL]**

I’m creating `round1_revised2` as a fresh workflow request. The scientific plan stays unchanged, but I’m correcting only the invalid parameters: three explicit wash-liquid entries for three washes, plus retained post-electrochemical product aliquots with dilution/cleanup settings populated for every active UOR task.

**[Step 322 | 2026-06-29T12:51:43.950Z | TOOL CALL: exec_command]**

Ran Python to parse, transform or validate structured experiment data. Target(s): `lab303_uor_nicofe_closed_loop`. Working directory: `/workspace`.

**[Step 323 | 2026-06-29T12:51:44.010Z | TOOL RESULT: exec_command]**

Shell tool completed successfully; it returned 2 line(s) (80 bytes) to the model.

**[Step 324 | 2026-06-29T12:52:09.020Z | TOOL CALL: exec_command]**

Ran Python to parse, transform or validate structured experiment data. Target(s): `lab303_uor_nicofe_closed_loop`, `dilution`. Working directory: `/workspace`.

**[Step 325 | 2026-06-29T12:52:09.087Z | TOOL RESULT: exec_command]**

Shell tool completed successfully; it returned 2 line(s) (79 bytes) to the model.

**[Step 326 | 2026-06-29T12:52:20.575Z | TOOL CALL: exec_command]**

Ran Python to parse, transform or validate structured experiment data. Target(s): `round1_revised2_workflow_request.json`, `round1_revised2_validated.json`. Working directory: `/workspace`.

**[Step 327 | 2026-06-29T12:52:20.659Z | TOOL CALL: exec_command]**

Ran Python to parse, transform or validate structured experiment data. Target(s): `round1_revised2_workflow_request.json`. Working directory: `/workspace`.

**[Step 328 | 2026-06-29T12:52:20.801Z | TOOL RESULT: exec_command]**

Shell tool completed successfully; it returned 206 line(s) (5645 bytes) to the model. Reported long numeric identifier(s): `1327807622448128`, `1390252863751168`, `2468876630032384`.

**[Step 329 | 2026-06-29T12:52:20.802Z | TOOL RESULT: exec_command]**

Shell tool completed successfully; it returned 8 line(s) (405 bytes) to the model.

**[Step 330 | 2026-06-29T12:52:31.113Z | REASONING SUMMARY]**

***Clarifying design steps***

*I’m thinking about creating design and constraint files and maybe using the apply_patch function. I need to reference previous steps, especially regarding the source bottle from the electrochemical station for the liquid with a volume of 10,000. It seems like there might be some confusion about what liquid should be added to the retention sample. I should document this as a deionized water dilution for clarity. I also need to update files and generate a workflow, ensuring everything is sequential.*

**[Step 331 | 2026-06-29T12:52:31.709Z | TOOL CALL: update_plan]**

Updated the working plan. Plan state: completed: Record second rejection and parameter corrections; in_progress: Generate revised2 workflow template; pending: Dispatch revised2 task and record IDs/files; pending: Block goal pending executability feedback

**[Step 332 | 2026-06-29T12:52:31.742Z | TOOL RESULT: update_plan]**

Plan service accepted and recorded the updated working plan.

**[Step 333 | 2026-06-29T12:52:38.914Z | MODEL]**

The corrected JSON passes local checks: wash count equals wash list length, and all six electrochemical tasks now have product retention enabled with the required nested retention fields. I’m saving the revised2 audit note before generating the template.

**[Step 334 | 2026-06-29T12:52:48.215Z | TOOL CALL: apply_patch]**

Applied a source patch to add round1_revised2_design.md.

**[Step 335 | 2026-06-29T12:52:48.239Z | TOOL RESULT: apply_patch]**

Patch tool applied the requested trial-local file change successfully.

**[Step 336 | 2026-06-29T12:53:03.688Z | TOOL CALL: apply_patch]**

Applied a source patch to add round1_revised2_constraint_check.md.

**[Step 337 | 2026-06-29T12:53:03.724Z | TOOL RESULT: apply_patch]**

Patch tool applied the requested trial-local file change successfully.

**[Step 338 | 2026-06-29T12:53:10.924Z | MODEL]**

I’m sending the revised2 workflow to the parser now. If the retention schema is accepted, I’ll dispatch it as a new Lab 303 task and leave the previous two tasks recorded as rejected attempts.

**[Step 339 | 2026-06-29T12:53:12.678Z | TOOL CALL: exec_command]**

Ran Python to parse, transform or validate structured experiment data. Target(s): `generate.py`, `round1_revised2_workflow_request.json`, `round1_revised2_workflow_generator_output.txt`. Working directory: `/workspace`.

**[Step 340 | 2026-06-29T12:53:13.724Z | TOOL RESULT: exec_command]**

Shell tool completed successfully; it returned 13 line(s) (308 bytes) to the model. Reported task/template identifier(s): `2557817450004480`.

**[Step 341 | 2026-06-29T12:53:22.457Z | TOOL CALL: update_plan]**

Updated the working plan. Plan state: completed: Record second rejection and parameter corrections; completed: Generate revised2 workflow template; in_progress: Dispatch revised2 task and record IDs/files; pending: Block goal pending executability feedback

**[Step 342 | 2026-06-29T12:53:22.491Z | TOOL RESULT: update_plan]**

Plan service accepted and recorded the updated working plan.

**[Step 343 | 2026-06-29T12:53:28.730Z | MODEL]**

The parser accepted revised2 and returned template ID `2557817450004480`. I’m dispatching it to Lab 303 and recording the new task metadata.

**[Step 344 | 2026-06-29T12:53:30.843Z | TOOL CALL: exec_command]**

Ran a shell command to generate and dispatch a laboratory task. Target(s): `lab-operation`, `generate_task.py`, `round1_revised2_dispatch_output.json`. Working directory: `/workspace`.

**[Step 345 | 2026-06-29T12:53:31.037Z | TOOL RESULT: exec_command]**

Shell tool completed successfully; it returned 8 line(s) (377 bytes) to the model. Reported task/template identifier(s): `2557818017710081`.

**[Step 346 | 2026-06-29T12:53:42.213Z | TOOL CALL: exec_command]**

Ran Python to parse, transform or validate structured experiment data. Target(s): `lab303_uor_nicofe_closed_loop`. Working directory: `/workspace`.

**[Step 347 | 2026-06-29T12:53:42.220Z | TOOL CALL: exec_command]**

Ran a shell command to query laboratory task records and statuses. Target(s): `lab-operation`, `select_task_list.py`. Working directory: `/workspace`.

**[Step 348 | 2026-06-29T12:53:42.446Z | TOOL CALL: exec_command]**

Read selected file content for inspection. Target(s): `lab303_uor_nicofe_closed_loop`. Working directory: `/workspace`.

**[Step 349 | 2026-06-29T12:53:42.920Z | TOOL RESULT: exec_command]**

Shell tool completed successfully; it returned 24 line(s) (917 bytes) to the model. Reported task/template identifier(s): `2557817450004480`, `2557818017710081`.

**[Step 350 | 2026-06-29T12:53:42.921Z | TOOL RESULT: exec_command]**

Shell tool completed successfully; it returned 14 line(s) (302 bytes) to the model. Reported long numeric identifier(s): `2557818017710081`.

**[Step 351 | 2026-06-29T12:53:42.922Z | TOOL RESULT: exec_command]**

Shell tool completed successfully; it returned 29 line(s) (1972 bytes) to the model.

**[Step 352 | 2026-06-29T12:53:54.509Z | TOOL CALL: update_plan]**

Updated the working plan. Plan state: completed: Record second rejection and parameter corrections; completed: Generate revised2 workflow template; completed: Dispatch revised2 task and record IDs/files; in_progress: Block goal pending executability feedback

**[Step 353 | 2026-06-29T12:53:54.545Z | TOOL RESULT: update_plan]**

Plan service accepted and recorded the updated working plan.

**[Step 354 | 2026-06-29T12:53:58.520Z | TOOL CALL: update_goal]**

Updated the autonomous-research goal status to `blocked`.

**[Step 355 | 2026-06-29T12:53:58.588Z | TOOL RESULT: update_goal]**

Goal service returned a structured response. Status: `blocked`.

**[Step 356 | 2026-06-29T12:54:04.702Z | TOOL CALL: update_plan]**

Updated the working plan. Plan state: completed: Record second rejection and parameter corrections; completed: Generate revised2 workflow template; completed: Dispatch revised2 task and record IDs/files; completed: Block goal pending executability feedback

**[Step 357 | 2026-06-29T12:54:04.729Z | TOOL RESULT: update_plan]**

Plan service accepted and recorded the updated working plan.

**[Step 358 | 2026-06-29T12:54:15.347Z | MODEL]**

Revised again and dispatched as a new clean Lab 303 task.

The newly rejected task has been recorded:

- Rejected task ID: `2557487435546625`
- Rejected template ID: `2557487004614656`
- Feedback saved in [round1_revised_rejection_feedback.json](/workspace/lab303_uor_nicofe_closed_loop/round1_revised_rejection_feedback.json)

New revised2 task:

- Template ID: `2557817450004480`
- Task ID: `2557818017710081`
- Task name: `任务_Lab303_R1rev2_NiFeCoOx_C_UOR_posttreated_retainedProductUV_20260629`
- Lab query status after dispatch: `未开始`

Corrections made:

- Purification: changed `清洗液设置` to exactly three entries for `清洗次数 = 3`: water, water, ethanol.
- Electrochemical product retention: enabled `废液移液设置 = yes` for all six UOR tasks.
- Electrochemical retention IDs: set `留样瓶编号` to `1` through `6`.
- Electrochemical retention details: added complete `留样信息列表` for each active task, including waste aliquot volume, retained vial type, mixing, waste-vial cleaning count, and dilution/retention source-liquid volume.

Saved files:

- [round1_revised2_design.md](/workspace/lab303_uor_nicofe_closed_loop/round1_revised2_design.md)
- [round1_revised2_constraint_check.md](/workspace/lab303_uor_nicofe_closed_loop/round1_revised2_constraint_check.md)
- [round1_revised2_workflow_request.json](/workspace/lab303_uor_nicofe_closed_loop/round1_revised2_workflow_request.json)
- [round1_revised2_workflow_generator_output.txt](/workspace/lab303_uor_nicofe_closed_loop/round1_revised2_workflow_generator_output.txt)
- [round1_revised2_dispatch_output.json](/workspace/lab303_uor_nicofe_closed_loop/round1_revised2_dispatch_output.json)
- [round1_revised2_dispatch_log.json](/workspace/lab303_uor_nicofe_closed_loop/round1_revised2_dispatch_log.json)

I've blocked the goal again pending your executability feedback. No physical execution or data are assumed yet. Please confirm whether revised2 task `2557818017710081` is executable.

## Conversation round 4

**[Step 359 | 2026-06-30T07:53:43.696Z | USER]**

In the latest experimental task, what are the specific chemicals and concentrations of the electrolyte, and what is the water-to-isopropanol mixing ratio in the carbon powder mixture?

**[Step 360 | 2026-06-30T07:54:01.891Z | REASONING SUMMARY 1/2]**

***Clarifying experimental task specifics***

*I need to focus on the latest experimental task, which is the revised version 2. I'll inspect the files carefully to gather necessary details. It's important to clarify that the electrolyte refers to the existing UOR test template, not something newly specified in my workflow.*

*For the carbon powder mixture, users might be looking for information on carbon black dispersion. The specific details are: it's a "5 mg/mL carbon black aqueous/isopropanol dispersion," with redispersion using an 80:20 DI water/isopropanol ratio. The concentration and chemicals need clarity, particularly the KOH/urea provisions.*

**[Step 361 | 2026-06-30T07:54:01.891Z | REASONING SUMMARY 2/2]**

***Addressing user query on executability***

*I need to respond to the user by indicating that specifics are "not specified" and that it relies on the UOR test template ID 2468876630032384. If they need more clarity regarding executability, I might suggest revising based on that template. I should consider inspecting the JSON for any additional information that might help clarify the situation further. Let's make sure I provide a clear and helpful answer!*

**[Step 362 | 2026-06-30T07:54:03.296Z | MODEL]**

I'm checking the latest dispatched workflow file rather than relying on memory, because this affects whether the task definition is chemically complete.

**[Step 363 | 2026-06-30T07:54:03.297Z | TOOL CALL: exec_command]**

Ran Python to parse, transform or validate structured experiment data. Target(s): `round1_revised2_workflow_request.json`. Working directory: `/workspace`.

**[Step 364 | 2026-06-30T07:54:03.298Z | TOOL CALL: exec_command]**

Searched files for relevant records, parameters or identifiers. Target(s): `lab303_uor_nicofe_closed_loop`. Working directory: `/workspace`.

**[Step 365 | 2026-06-30T07:54:03.419Z | TOOL RESULT: exec_command]**

Shell tool completed successfully; it returned 438 line(s) (12349 bytes) to the model. Reported long numeric identifier(s): `2468876630032384`.

**[Step 366 | 2026-06-30T07:54:03.420Z | TOOL RESULT: exec_command]**

Shell tool completed successfully; it returned 171 line(s) (27408 bytes) to the model. Reported task/template identifier(s): `2557370420494337`, `2557487435546625`, `2557818017710081`.

**[Step 367 | 2026-06-30T07:54:23.215Z | REASONING SUMMARY 1/3]**

***Clarifying experimental task details***

*I need to respond to the user about the latest experimental task, specifically task ID 2557818017710081, round1_revised2. We should mention that the electrolyte's composition hasn't been explicitly encoded in the workflow. While the electrolyte addition amount is 20 mL, the specific concentrations weren't detailed. The only explicitly stated chemicals are 50 mM nickel sulfate hexahydrate, iron chloride hexahydrate, cobalt sulfate heptahydrate, 1.0 M KOH, and 5 mg/mL carbon black.*

**[Step 368 | 2026-06-30T07:54:23.215Z | REASONING SUMMARY 2/3]**

***Addressing retention and mixture details***

*I need to mention the retention dilution for the first liquid, which is 10,000 uL, but I don't have the specific chemical name. The retention source isn't clearly identified in the electrochemical parameters, which could be important.*

*I'm also wondering about the water-to-isopropanol mixing ratio in the carbon powder mixture. The initial carbon black dispersion is at 5 mg/mL, but the exact ratio isn't specified. The later redispersion solvent is clearly 80:20 DI water to isopropanol at 2.4 mL per vial. I might need to clarify if this missing information affects the task requirements.*

**[Step 369 | 2026-06-30T07:54:23.215Z | REASONING SUMMARY 3/3]**

***Considering executability of the task***

*I'm looking at the situation with 1.0 M KOH and 0.33 M urea, but I don't want to fabricate anything. It seems I should mention that if the user needs to evaluate the executability of the task, the information provided is incomplete. It's important to clarify that more details are needed for a proper assessment, as I don't want to lead anyone astray with insufficient data.*

**[Step 370 | 2026-06-30T07:54:26.200Z | MODEL]**

For the **latest dispatched task**:

- **Task ID:** `2557818017710081`
- **Workflow file:**
[round1_revised2_workflow_request.json](/workspace/lab303_uor_nicofe_closed_loop/round1_revised2_workflow_request.json)

**Electrolyte**
The electrolyte chemistry is **not explicitly specified** in the latest task.

What the task currently specifies in the electrochemical step is only:

- `测试模板`: `UOR test`
- `模板编号`: `2468876630032384`
- `电解液加入量`: `20 mL` for each electrochemical task
- `留液池类型`: `1`
- Product aliquot retention: `吸取废液体积 = 5000 uL`

The task does **not** explicitly state the electrolyte chemicals or concentrations, such as `KOH concentration`, `urea concentration`, or any other electrolyte additive. In the project notes I wrote that the UOR electrolyte was **assumed** to be provisioned by the existing Lab 303 `UOR test` template, and inventory confirms KOH and urea are available, but that assumption is not encoded as a chemical/concentration in the dispatched workflow.

So the accurate answer is:

```
Electrolyte chemicals: not explicitly specified in the latest task.
Electrolyte concentrations: not explicitly specified in the latest task.
Electrolyte volume: 20 mL per electrochemical test.
Electrochemical method/template: UOR test, template ID 2468876630032384.
```

**Carbon Powder / Carbon Black Mixture**
There are two relevant carbon-related liquid mixtures in the latest task:

1. **Initial carbon black dispersion added during synthesis**

   Explicitly specified as:

   ```
   5 mg/mL carbon black aqueous/isopropanol dispersion
   ```

   Added amount:

   ```
   0.500 mL per vial
   ```

   However, the **water:isopropanol ratio for this initial 5 mg/mL carbon black dispersion is not specified** in the task.

2. **Post-treatment redispersion solvent after purification and drying**

   Explicitly specified as:

   ```
   80:20 deionized water/isopropanol redispersion solvent
   ```

   Added amount:

   ```
   2.4 mL from source bottle 7 + 2.4 mL from source bottle 8 = 4.8 mL total per vial
   ```

   Therefore:

   ```
   Water:isopropanol = 80:20 v/v
   ```

**Bottom Line**
The latest task is still chemically underspecified if the lab requires electrolyte composition and carbon-dispersion solvent ratio to be explicit.

Current explicit values are:

| Item | Current latest-task value |
| --- | --- |
| Electrolyte chemicals | Not specified |
| Electrolyte concentration | Not specified |
| Electrolyte volume | 20 mL per UOR test |
| Electrochemical template | `UOR test`, ID `2468876630032384` |
| Initial carbon black dispersion | 5 mg/mL carbon black aqueous/isopropanol dispersion |
| Initial carbon black water:IPA ratio | Not specified |
| Post-treatment redispersion water:IPA ratio | 80:20 v/v |
| Post-treatment redispersion volume | 4.8 mL per vial |

## Conversation round 5

**[Step 371 | 2026-06-30T07:56:02.107Z | USER]**

In your latest revised2 experimental task, what are the specific chemicals and concentrations of the electrolyte, and what is the water-to-isopropanol mixing ratio in the carbon powder mixture?

**[Step 372 | 2026-06-30T07:56:17.746Z | MODEL]**

I’ll verify this directly against the saved revised2 workflow and design files so I don’t answer from memory.

**[Step 373 | 2026-06-30T07:56:19.645Z | TOOL CALL: exec_command]**

Searched files for relevant records, parameters or identifiers. Target(s): `mL`, `isopropanol`, `round1_revised2_workflow_request.json`, `round1_revised2_design.md`, `project_overview.md`. Working directory: `/workspace`.

**[Step 374 | 2026-06-30T07:56:19.647Z | TOOL CALL: exec_command]**

Read selected file content for inspection. Target(s): `round1_revised2_workflow_request.json`. Working directory: `/workspace`.

**[Step 375 | 2026-06-30T07:56:19.649Z | TOOL CALL: exec_command]**

Read selected file content for inspection. Target(s): `round1_revised2_design.md`. Working directory: `/workspace`.

**[Step 376 | 2026-06-30T07:56:19.695Z | TOOL RESULT: exec_command]**

Shell tool completed successfully; it returned 96 line(s) (14343 bytes) to the model.

**[Step 377 | 2026-06-30T07:56:19.695Z | TOOL RESULT: exec_command]**

Shell tool completed successfully; it returned 293 line(s) (11975 bytes) to the model. Reported long numeric identifier(s): `2002186824385539`, `1390252863751168`, `2468876630032384`, `1654567694238720`.

**[Step 378 | 2026-06-30T07:56:19.703Z | TOOL RESULT: exec_command]**

Shell tool completed successfully; it returned 56 line(s) (2487 bytes) to the model. Reported long numeric identifier(s): `2557487435546625`.

**[Step 379 | 2026-06-30T07:56:36.485Z | REASONING SUMMARY 1/2]**

***Reviewing electrolyte and solvent details***

*I’m looking at the common UOR electrolyte, which could be either 1.0 M KOH with 0.33 M urea or with 0.5 M urea. The literature typically goes for 0.33 M. If the workflow doesn’t specify, I shouldn’t assume that. If an executable correction is needed, I could suggest “1.0 M KOH + 0.33 M urea aqueous electrolyte.” Regarding the water-to-isopropanol mixing ratio, the initial carbon black dispersion only mentions "5 mg/mL" without a specified ratio. The post-treatment redispersion directly states "80:20 deionized water/isopropanol."*

**[Step 380 | 2026-06-30T07:56:36.485Z | REASONING SUMMARY 2/2]**

***Clarifying mixing ratio***

*The specific ratio isn't mentioned, only that the concentration is 5 mg/mL. However, I might explain that if we're referring to the redispersion or the carbon-containing catalyst suspension after drying, the ratio is 80:20 for water to isopropanol. But if I'm considering the original carbon black stock dispersion from step 3, then the workflow doesn't provide that ratio. I should keep this in mind to make things clearer for anyone referencing this information.*

**[Step 381 | 2026-06-30T07:56:45.484Z | MODEL]**

In the latest revised2 task `2557818017710081`, the **electrolyte composition is not explicitly specified in the workflow**.

What is specified is only:

- Electrochemical test template: `UOR test`, template ID `2468876630032384`
- Electrolyte addition volume per electrochemical task: `20 mL`
- Catalyst drop-cast volume: `80 uL`
- Product retention: enabled, with `5000 uL` post-test liquid retained per task

The workflow does **not** encode the actual electrolyte chemicals or concentrations, such as `KOH` concentration or `urea` concentration. The project overview states that the UOR test electrolyte was assumed to be provisioned by Lab 303’s existing `UOR test` template, and that KOH and urea are available in inventory, but that is an assumption, not a workflow parameter. See
[project_overview.md](/workspace/lab303_uor_nicofe_closed_loop/project_overview.md:39) and
[round1_revised2_workflow_request.json](/workspace/lab303_uor_nicofe_closed_loop/round1_revised2_workflow_request.json:825).

For the **water-to-isopropanol ratio**:

- The original carbon black stock is specified only as `5 mg/mL carbon black aqueous/isopropanol dispersion`; the water:IPA ratio is **not explicitly specified** in that stock name or workflow field. See
[round1_revised2_workflow_request.json](/workspace/lab303_uor_nicofe_closed_loop/round1_revised2_workflow_request.json:97).
- The post-treatment redispersion solvent is explicitly specified as `80:20 deionized water/isopropanol redispersion solvent`, added as `2.4 mL + 2.4 mL = 4.8 mL` per vial. See
[round1_revised2_workflow_request.json](/workspace/lab303_uor_nicofe_closed_loop/round1_revised2_workflow_request.json:427).

So the precise answer is:

| Item | Chemicals | Concentration / ratio in revised2 task |
| --- | --- | --- |
| Electrochemical electrolyte | Not explicitly specified; only `UOR test` template and 20 mL volume are specified | Unknown from this task JSON |
| Intended UOR electrolyte assumption | KOH + urea, provisioned by Lab 303 `UOR test` template | Concentrations not encoded |
| Initial carbon black dispersion | Carbon black in water/isopropanol | `5 mg/mL`; water:IPA ratio not encoded |
| Post-treatment redispersion solvent | Deionized water/isopropanol | `80:20` water:IPA |

This is a real gap in the revised2 task if Lab 303 requires the electrolyte composition and carbon dispersion solvent ratio to be explicitly declared rather than inherited from stock/template definitions.

## Conversation round 6

**[Step 382 | 2026-06-30T07:59:03.199Z | USER]**

In the latest revised2 experimental task you designed, what are the specific chemicals and concentrations of the electrolyte you plan to use, and what is the water-to-isopropanol mixing ratio in the carbon powder mixture? Without this information, we cannot actually execute the experiment.

**[Step 383 | 2026-06-30T07:59:22.597Z | MODEL]**

That omission is real: the revised2 workflow named electrolyte handling through the `UOR test` template, but did not explicitly state the electrolyte composition or carbon-dispersion solvent ratio. I’ll treat this as a non-executable feedback item, add the missing chemical specifications into the design/workflow artifacts, and dispatch a revised3 task.

**[Step 384 | 2026-06-30T07:59:23.059Z | TOOL CALL: exec_command]**

Read selected file content for inspection. Target(s): `SKILL.md`. Working directory: `/workspace`.

**[Step 385 | 2026-06-30T07:59:23.060Z | TOOL CALL: exec_command]**

Read selected file content for inspection. Target(s): `SKILL.md`. Working directory: `/workspace`.

**[Step 386 | 2026-06-30T07:59:23.061Z | TOOL CALL: exec_command]**

Read selected file content for inspection. Target(s): `SKILL.md`. Working directory: `/workspace`.

**[Step 387 | 2026-06-30T07:59:23.105Z | TOOL RESULT: exec_command]**

Shell tool completed successfully; it returned 116 line(s) (4054 bytes) to the model.

**[Step 388 | 2026-06-30T07:59:23.115Z | TOOL RESULT: exec_command]**

Shell tool completed successfully; it returned 170 line(s) (8332 bytes) to the model. Reported long numeric identifier(s): `1427568512205824`, `1834679312417792`, `1710453843264512`.

**[Step 389 | 2026-06-30T07:59:23.116Z | TOOL RESULT: exec_command]**

Shell tool completed successfully; it returned 95 line(s) (2714 bytes) to the model. Reported long numeric identifier(s): `1427568512205824`.

**[Step 390 | 2026-06-30T07:59:40.121Z | REASONING SUMMARY]**

***Revising workflow for carbon dispersion***

*I’m considering that it’s not a "carbon powder mixture," but more like carbon dispersion. I’ll need to check if we have the necessary inventory, like urea and KOH, and update the reagents. I might also need to use an "apply_patch" method. It looks like creating a new revised workflow (revised3) based on the previous one (revised2) is the next step. I’ll ensure to incorporate feedback, design, and constraints before generating and dispatching!*

**[Step 391 | 2026-06-30T07:59:40.451Z | MODEL]**

The missing specifications I’m adding are: UOR electrolyte = `1.0 M KOH + 0.33 M urea in deionized water`, 20 mL per electrochemical task; carbon powder stock = `5 mg/mL carbon black dispersed in 80:20 v/v deionized water/isopropanol`. I’m embedding those strings directly in the workflow and design files so the task is self-contained.

**[Step 392 | 2026-06-30T07:59:43.873Z | TOOL CALL: exec_command]**

Ran Python to parse, transform or validate structured experiment data. Target(s): `lab303_uor_nicofe_closed_loop`, `mL`, `v`, `isopropanol.`. Working directory: `/workspace`.

**[Step 393 | 2026-06-30T07:59:43.931Z | TOOL RESULT: exec_command]**

Shell tool completed successfully; it returned 2 line(s) (81 bytes) to the model.

**[Step 394 | 2026-06-30T07:59:55.513Z | TOOL CALL: update_plan]**

Updated the working plan. Plan state: completed: Record revised2 missing-spec feedback; in_progress: Create revised3 workflow with explicit electrolyte and carbon-dispersion specs; pending: Generate revised3 workflow template; pending: Dispatch revised3 task and record IDs/files; pending: Block goal pending executability feedback

**[Step 395 | 2026-06-30T07:59:55.560Z | TOOL RESULT: update_plan]**

Plan service accepted and recorded the updated working plan.

**[Step 396 | 2026-06-30T08:00:15.192Z | TOOL CALL: exec_command]**

Ran Python to parse, transform or validate structured experiment data. Target(s): `lab303_uor_nicofe_closed_loop`, `mL`, `v`, `isopropanol`, `spec`. Working directory: `/workspace`.

**[Step 397 | 2026-06-30T08:00:15.279Z | TOOL RESULT: exec_command]**

Shell tool completed successfully; it returned 3 line(s) (165 bytes) to the model.

**[Step 398 | 2026-06-30T08:00:23.941Z | TOOL CALL: exec_command]**

Ran Python to parse, transform or validate structured experiment data. Target(s): `round1_revised3_workflow_request.json`, `round1_revised3_validated.json`. Working directory: `/workspace`.

**[Step 399 | 2026-06-30T08:00:24.173Z | TOOL CALL: exec_command]**

Ran Python to parse, transform or validate structured experiment data. Target(s): `round1_revised3_chemical_specifications.json`. Working directory: `/workspace`.

**[Step 400 | 2026-06-30T08:00:24.466Z | TOOL RESULT: exec_command]**

Shell tool completed successfully; it returned 54 line(s) (1724 bytes) to the model.

**[Step 401 | 2026-06-30T08:00:24.467Z | TOOL RESULT: exec_command]**
Shell tool completed successfully; it returned 42 line(s) (1149 bytes) to the model.

**[Step 402 | 2026-06-30T08:00:36.079Z | REASONING SUMMARY]**
***Considering design constraints***

*I need to create the design and outline the constraints first before moving on to generating the necessary components. But I'm pondering whether adding the unknown fields "电解液化学品" and "原液瓶 1 号液体名称" under the 留样信息列表 could cause issues. The parser previously accepted arbitrary fields, so there's a chance it might accept these too. However, if it gets forwarded to the station parser, those unknown nested fields could fail. I should explore options and consider a simpler approach if things don't work out. Also, I need to update the design using apply_patch.*

**[Step 403 | 2026-06-30T08:00:36.453Z | MODEL]**
The revised3 request now contains the missing values in both machine-readable metadata and the workstation parameters. I'm adding a short design/audit note so the next reviewer does not have to infer these concentrations from free text.

**[Step 404 | 2026-06-30T08:00:39.614Z | TOOL CALL: apply_patch]**
Applied a source patch to add round1_revised3_design.md.

**[Step 405 | 2026-06-30T08:00:39.759Z | TOOL RESULT: apply_patch]**
Patch tool applied the requested trial-local file change successfully.

**[Step 406 | 2026-06-30T08:00:55.792Z | TOOL CALL: apply_patch]**
Applied a source patch to add round1_revised3_constraint_check.md.

**[Step 407 | 2026-06-30T08:00:55.849Z | TOOL RESULT: apply_patch]**
Patch tool applied the requested trial-local file change successfully.

**[Step 408 | 2026-06-30T08:01:03.475Z | TOOL CALL: update_plan]**
Updated the working plan. Plan state: completed: Record revised2 missing-spec feedback; completed: Create revised3 workflow with explicit electrolyte and carbon-dispersion specs; in_progress: Generate revised3 workflow template; pending: Dispatch revised3 task and record IDs/files; pending: Block goal pending executability feedback

**[Step 409 | 2026-06-30T08:01:03.545Z | TOOL RESULT: update_plan]**
Plan service accepted and recorded the updated working plan.

**[Step 410 | 2026-06-30T08:01:09.912Z | MODEL]**
The chemical specification files are in place. I'm now running the workflow parser; if the parser rejects any of the added explicit fields, I'll adjust them into accepted parameter locations rather than dispatching a malformed task.

**[Step 411 | 2026-06-30T08:01:11.329Z | TOOL CALL: exec_command]**
Ran Python to parse, transform or validate structured experiment data. Target(s): `generate.py`, `round1_revised3_workflow_request.json`, `round1_revised3_workflow_generator_output.txt`. Working directory: `/workspace`.

**[Step 412 | 2026-06-30T08:01:12.339Z | TOOL RESULT: exec_command]**
Shell tool completed successfully; it returned 13 line(s) (308 bytes) to the model. Reported task/template identifier(s): `2560074458595328`.

**[Step 413 | 2026-06-30T08:01:36.021Z | TOOL CALL: update_plan]**
Updated the working plan. Plan state: completed: Record revised2 missing-spec feedback; completed: Create revised3 workflow with explicit electrolyte and carbon-dispersion specs; completed: Generate revised3 workflow template; in_progress: Dispatch revised3 task and record IDs/files; pending: Block goal pending executability feedback

**[Step 414 | 2026-06-30T08:01:36.059Z | TOOL RESULT: update_plan]**
Plan service accepted and recorded the updated working plan.

**[Step 415 | 2026-06-30T08:01:49.434Z | MODEL]**
The parser accepted revised3 and returned template ID `2560074458595328`. I'm dispatching this explicit-chemistry version to Lab 303 and recording the new task ID.

**[Step 416 | 2026-06-30T08:01:50.690Z | TOOL CALL: exec_command]**
Ran a shell command to generate and dispatch a laboratory task. Target(s): `lab-operation`, `generate_task.py`, `round1_revised3_dispatch_output.json`. Working directory: `/workspace`.

**[Step 417 | 2026-06-30T08:01:50.859Z | TOOL RESULT: exec_command]**
Shell tool completed successfully; it returned 8 line(s) (375 bytes) to the model. Reported task/template identifier(s): `2560075721736193`.

**[Step 418 | 2026-06-30T08:02:00.765Z | TOOL CALL: exec_command]**
Ran Python to parse, transform or validate structured experiment data. Target(s): `lab303_uor_nicofe_closed_loop`, `mL`, `v`, `isopropanol`. Working directory: `/workspace`.

**[Step 419 | 2026-06-30T08:02:00.781Z | TOOL CALL: exec_command]**
Ran a shell command to query laboratory task records and statuses. Target(s): `lab-operation`, `select_task_list.py`. Working directory: `/workspace`.

**[Step 420 | 2026-06-30T08:02:00.864Z | TOOL CALL: exec_command]**

Read selected file content for inspection. Target(s): `lab303_uor_nicofe_closed_loop`. Working directory: `/workspace`.

**[Step 421 | 2026-06-30T08:02:00.991Z | TOOL RESULT: exec_command]**

Shell tool completed successfully; it returned 27 line(s) (1168 bytes) to the model. Reported task/template identifier(s): `2560074458595328`, `2560075721736193`.

**[Step 422 | 2026-06-30T08:02:00.992Z | TOOL RESULT: exec_command]**

Shell tool completed successfully; it returned 14 line(s) (301 bytes) to the model. Reported long numeric identifier(s): `2560075721736193`.

**[Step 423 | 2026-06-30T08:02:00.992Z | TOOL RESULT: exec_command]**

Shell tool completed successfully; it returned 37 line(s) (2618 bytes) to the model.

**[Step 424 | 2026-06-30T08:02:08.889Z | TOOL CALL: update_plan]**

Updated the working plan. Plan state: completed: Record revised2 missing-spec feedback; completed: Create revised3 workflow with explicit electrolyte and carbon-dispersion specs; completed: Generate revised3 workflow template; completed: Dispatch revised3 task and record IDs/files; in_progress: Block goal pending executability feedback

**[Step 425 | 2026-06-30T08:02:08.927Z | TOOL RESULT: update_plan]**

Plan service accepted and recorded the updated working plan.

**[Step 426 | 2026-06-30T08:02:14.745Z | TOOL CALL: update_goal]**

Updated the autonomous-research goal status to `blocked`.

**[Step 427 | 2026-06-30T08:02:14.803Z | TOOL RESULT: update_goal]**

Goal service returned a structured response. Status: `blocked`.

**[Step 428 | 2026-06-30T08:02:21.126Z | TOOL CALL: update_plan]**

Updated the working plan. Plan state: completed: Record revised2 missing-spec feedback; completed: Create revised3 workflow with explicit electrolyte and carbon-dispersion specs; completed: Generate revised3 workflow template; completed: Dispatch revised3 task and record IDs/files; completed: Block goal pending executability feedback

**[Step 429 | 2026-06-30T08:02:21.176Z | TOOL RESULT: update_plan]**

Plan service accepted and recorded the updated working plan.

**[Step 430 | 2026-06-30T08:02:33.558Z | MODEL]**

You are right: revised2 was missing execution-critical chemistry. I have treated that as non-executable feedback, added the missing specifications, generated a new template, and dispatched revised3.

The specific chemistry is now:

- **Electrolyte for UOR electrochemical testing:** `1.0 M KOH + 0.33 M urea in deionized water`
- **Electrolyte volume:** `20 mL` per active electrochemical task
- **Electrolyte chemicals:** potassium hydroxide (`KOH`), urea (`CO(NH2)2`), and deionized water
- **Carbon powder mixture:** `5 mg/mL carbon black dispersion`
- **Carbon powder dispersion solvent ratio:** `80:20 v/v deionized water:isopropanol`
- **Carbon dispersion volume added per synthesis vial:** `0.500 mL`
- **Post-treatment redispersion solvent:** also `80:20 v/v deionized water:isopropanol`, `4.800 mL` total per vial

New revised3 task:

- Template ID: `2560074458595328`
- Task ID: `2560075721736193`
- Task name: `任务_Lab303_R1rev3_NiFeCoOx_C_UOR_explicitElectrolyteCarbonMix_20260630`
- Lab query status after dispatch: `未开始`

Saved files:

- [round1_revised2_rejection_feedback.json](/workspace/lab303_uor_nicofe_closed_loop/round1_revised2_rejection_feedback.json)
- [round1_revised3_chemical_specifications.json](/workspace/lab303_uor_nicofe_closed_loop/round1_revised3_chemical_specifications.json)
- [round1_revised3_design.md](/workspace/lab303_uor_nicofe_closed_loop/round1_revised3_design.md)
- [round1_revised3_constraint_check.md](/workspace/lab303_uor_nicofe_closed_loop/round1_revised3_constraint_check.md)
- [round1_revised3_workflow_request.json](/workspace/lab303_uor_nicofe_closed_loop/round1_revised3_workflow_request.json)
- [round1_revised3_workflow_generator_output.txt](/workspace/lab303_uor_nicofe_closed_loop/round1_revised3_workflow_generator_output.txt)
- [round1_revised3_dispatch_output.json](/workspace/lab303_uor_nicofe_closed_loop/round1_revised3_dispatch_output.json)
- [round1_revised3_dispatch_log.json](/workspace/lab303_uor_nicofe_closed_loop/round1_revised3_dispatch_log.json)

No physical execution or data are assumed. I’ve blocked the goal again pending executability feedback for revised3 task `2560075721736193`.

## Conversation round 7

**[Step 431 | 2026-07-01T08:40:26.402Z | USER]**

In the electrochemical test, what is the liquid in stock solution bottle No. 1 that is aspirated during the pipetting/flow-liquid step? Your protocol does not specify it. I need to know this in order to perform the experiment. Just tell me directly; there is no need to resend the template or task.

**[Step 432 | 2026-07-01T08:40:44.010Z | MODEL]**

Stock solution bottle No. 1 in the electrochemical pipetting/flow-liquid retention step should contain **deionized water**.

Use it as the **retained-product aliquot dilution/transfer liquid**, with the protocol's aspirated volume:

- **Bottle No. 1 liquid:** deionized water
- **Aspirated volume:** `10000 uL`
- **Purpose:** dilution/handling of the retained post-electrochemical product aliquot
- **Not the electrolyte:** the electrochemical electrolyte itself remains `1.0 M KOH + 0.33 M urea in deionized water`, `20 mL` per test.

## Conversation round 8

**[Step 433 | 2026-07-02T14:33:02.684Z | USER]**

The first round of real experiments has been completed. The observed physical phenomena showed that after the synthesized product was centrifuged and washed, the bottom precipitate was not obvious. After adding the deionized water/isopropanol mixture and ultrasonication, the system became turbid. After standing for about 30 s, solid sedimentation at the bottom and liquid stratification/clarification were observed. After drop-casting for XRD, the sample on the carrier surface was sparse and unevenly distributed, as shown in the uploaded image. The carrier surface after drop-casting for FTIR showed a similar state. The electrode surface prepared for electrochemical testing also had very little sample, and no valid test results were obtained.

Please use the skill to query task ID: 2562915314040832 and obtain the result data for this task. Please resume the goal then combine the physical phenomena and task data to continue optimizing the experimental design and exploring the target. After completing the experimental design, please issue it to Laboratory 303.

/var/folders/32/dx7813bn7vg799ms5k5xqptc0000gn/T/otty-paste/image-1783002768742.png

**[Step 434 | 2026-07-02T14:33:25.000Z | REASONING SUMMARY 1/2]**

***Addressing lab tasks***

*I think there might not be a need for web access unless I'm checking something current. Since the task is local, I'll need to use commentary updates while inspecting the status with get_goal. It seems I need to focus on lab operations, specifically using chemistry and workflow skills. I'll fetch the task results and maybe download necessary files for analysis, like images and data, while considering how to optimize the synthesis process. Maintaining lab constraints is crucial as I decide on my next steps.*

**[Step 435 | 2026-07-02T14:33:25.000Z | REASONING SUMMARY 2/2]**

***Planning the next round***

*I need to design the next round, focusing on material yield, deposition quality, and electrochemical sample mass. This means I'll include synthesis, post-processing, testing, and characterization steps. I'll dispatch everything to Lab 303 and check the inventory for reagents. It seems like I might need to query result data because the physical phenomena indicate no valid electrochem. I'll also need to retrieve result files and possibly create a report while inspecting the current workspace and using the view_image tool. Reading skills will be essential here.*

**[Step 436 | 2026-07-02T14:33:29.379Z | MODEL]**

I'll resume the loop from the completed Round 1 task, query Lab 303 for task `2562915314040832`, inspect the uploaded image, and use the physical failure mode to design the next experiment. The key design issue already looks like low isolated catalyst mass and poor film loading, so I'll bias Round 2 toward higher-solids synthesis and a more reliable deposition route.

**[Step 437 | 2026-07-02T14:33:29.380Z | TOOL CALL: exec_command]**

Read selected file content for inspection. Target(s): `SKILL.md`. Working directory: `/workspace`.

**[Step 438 | 2026-07-02T14:33:29.381Z | TOOL CALL: exec_command]**

Read selected file content for inspection. Target(s): `SKILL.md`. Working directory: `/workspace`.

**[Step 439 | 2026-07-02T14:33:29.383Z | TOOL CALL: exec_command]**

Read selected file content for inspection. Target(s): `SKILL.md`. Working directory: `/workspace`.

**[Step 440 | 2026-07-02T14:33:29.383Z | TOOL CALL: exec_command]**

Ran a shell command to retrieve returned physical-experiment results. Target(s): `lab-operation`, `get_task_result.py`, `round1_physical_task_2562915314040832_result.json`. Working directory: `/workspace`.

**[Step 441 | 2026-07-02T14:33:29.421Z | TOOL CALL: exec_command]**

Ran a shell command to query laboratory task records and statuses. Target(s): `lab-operation`, `select_task_list.py`, `round1_physical_task_2562915314040832_status.json`. Working directory: `/workspace`.

**[Step 442 | 2026-07-02T14:33:29.461Z | TOOL RESULT: exec_command]**

Shell tool completed successfully; it returned 116 line(s) (4054 bytes) to the model.

**[Step 443 | 2026-07-02T14:33:29.462Z | TOOL RESULT: exec_command]**

Shell tool completed successfully; it returned 170 line(s) (8332 bytes) to the model. Reported long numeric identifier(s): `1427568512205824`, `1834679312417792`, `1710453843264512`.

**[Step 444 | 2026-07-02T14:33:29.470Z | TOOL RESULT: exec_command]**

Shell tool completed successfully; it returned 95 line(s) (2714 bytes) to the model. Reported long numeric identifier(s): `1427568512205824`.

**[Step 445 | 2026-07-02T14:33:30.440Z | TOOL RESULT: exec_command]**

Shell tool completed successfully; it returned 151 line(s) (8415 bytes) to the model. Reported long numeric identifier(s): `20260701203235789`, `20260702012108292`, `2563976197735424`, `2563742259381288`, `20260702012108469`, `20260702013825354`.

**[Step 446 | 2026-07-02T14:33:30.441Z | TOOL RESULT: exec_command]**

Shell tool completed successfully; it returned 14 line(s) (297 bytes) to the model. Reported long numeric identifier(s): `2562915314040832`.

**[Step 447 | 2026-07-02T14:33:42.494Z | REASONING SUMMARY]**

***Considering code edits***

*The developer suggests using apply_patch for manual code edits instead of using cat to create or edit files. It seems that generated artifacts from scripts are acceptable, and I understand we can use Python to generate files, which we've done before. I might need to download something using curl, and that doesn't count as a manual edit. Alright, let's go ahead and view the image now!*

**[Step 448 | 2026-07-02T14:33:42.604Z | MODEL]**

Lab 303 returned results from five workstations, but the task status is `错误`, consistent with your note that electrochemical testing did not yield valid data. I'm downloading the returned result files and inspecting the image/plots so the next round is based on physical evidence, not just the task metadata.

**[Step 449 | 2026-07-02T14:33:43.483Z | TOOL CALL: view_image]**

Opened a local experimental image for visual inspection. Image: `image-1783002768742.png`.

**[Step 450 | 2026-07-02T14:33:43.529Z | TOOL RESULT: view_image]**

Image service returned the requested local image to the model for visual inspection.

**[Step 451 | 2026-07-02T14:33:51.285Z | REASONING SUMMARY]**

***Troubleshooting image access***

*I need to take note that the image path isn't accessible. Maybe the path is from the user's machine rather than a container. Should I consider searching for it? Perhaps I can use the "find" command in the workspace to locate it. It's important for me to figure this out so I can help the user with whatever they're trying to do. Let's explore those options to get the image working!*

**[Step 452 | 2026-07-02T14:33:51.624Z | MODEL]**

The pasted image path is not present in this runtime, so I can't inspect the image file directly. I'll rely on your physical description of sparse, uneven drop-cast films and use the returned Lab 303 XRD/IR/UV/electrochem files for the data side.

**[Step 453 | 2026-07-02T14:33:57.514Z | TOOL CALL: exec_command]**

Ran Python to parse, transform or validate structured experiment data. Target(s): `task_2562915314040832_results`, `round1_physical_task_2562915314040832_result.json`, `name`. Working directory: `/workspace`.

**[Step 454 | 2026-07-02T14:33:58.354Z | TOOL RESULT: exec_command]**

Shell tool completed successfully; it returned 40 line(s) (6978 bytes) to the model. Reported long numeric identifier(s): `2562915314040832`, `20260701203235789`, `20260702012108292`, `2563976197735424`, `2563742259381288`, `20260702012108469`.

**[Step 455 | 2026-07-02T14:34:24.430Z | REASONING SUMMARY]**

***Inspecting downloaded files***

*I need to take a look at the downloaded files, particularly for XRD, IR, and UV images. I might also need to parse the JSON structure and possibly use rg/head for that. Running Python could help me summarize JSON keys and peak intensities, so I should start by inspecting the first XRD JSON file, along with the IR, UV, and a large electrochem JSON. I should also check the temperature CSV file. Let's use parallel processing for efficiency!*

**[Step 456 | 2026-07-02T14:34:35.564Z | MODEL]**

The result bundle is downloaded under `/workspace/lab303_uor_nicofe_closed_loop/task_2562915314040832_results`. I'm parsing the JSON structures now to see whether the returned spectra contain usable signal and what the electrochemical station logged around the failure.

**[Step 457 | 2026-07-02T14:34:35.564Z | TOOL CALL: exec_command]**

Ran Python to parse, transform or validate structured experiment data. Target(s): `task_2562915314040832_results`, `sub).glob(`. Working directory: `/workspace`.

**[Step 458 | 2026-07-02T14:34:35.564Z | TOOL CALL: exec_command]**

Ran Python to parse, transform or validate structured experiment data. Target(s): `task_2562915314040832_results`. Working directory: `/workspace`.

**[Step 459 | 2026-07-02T14:34:35.567Z | TOOL CALL: exec_command]**

Read selected file content for inspection. Target(s): `task_2562915314040832_results`. Working directory: `/workspace`.

**[Step 460 | 2026-07-02T14:34:35.636Z | TOOL CALL: exec_command]**

Read selected file content for inspection. Target(s): `task_2562915314040832_results`. Working directory: `/workspace`.

**[Step 461 | 2026-07-02T14:34:35.686Z | TOOL RESULT: exec_command]**

Shell tool completed successfully; it returned 31 line(s) (1043 bytes) to the model. Reported long numeric identifier(s): `20260702012108292`, `2563976197735424`, `2563742259381288`, `20260702013825354`, `2564012049400832`, `2563742259381289`.

**[Step 462 | 2026-07-02T14:34:37.304Z | TOOL RESULT: exec_command]**

Shell tool completed successfully; it returned 17 line(s) (595 bytes) to the model. Reported long numeric identifier(s): `2562915314040832`, `20260702154411518`, `4322251582836864412`.

**[Step 463 | 2026-07-02T14:34:37.304Z | TOOL RESULT: exec_command]**

Shell tool completed successfully; it returned 31 line(s) (423 bytes) to the model.

**[Step 464 | 2026-07-02T14:34:37.305Z | TOOL RESULT: exec_command]**

Shell tool completed successfully; it returned 40 line(s) (6297 bytes) to the model. Reported long numeric identifier(s): `2562915314040832`, `20260702012108292`, `2563976197735424`, `2563742259381288`, `20260702012108469`, `20260702013825354`.

**[Step 465 | 2026-07-02T14:34:44.867Z | MODEL]**

The heating log confirms the synthesis reached about 69-70 C near the set point, so the failure is unlikely to be a heating failure. I'm now extracting numerical summaries from XRD/IR/UV and the electrochemical JSON to separate "no material on substrate" from "material present but electrochemically inactive."

**[Step 466 | 2026-07-02T14:34:49.192Z | TOOL CALL: exec_command]**

Ran Python to parse, transform or validate structured experiment data. Target(s): `task_2562915314040832_results`, `sub).glob(`. Working directory: `/workspace`.

**[Step 467 | 2026-07-02T14:34:49.277Z | TOOL RESULT: exec_command]**

Shell tool completed successfully; it returned 37 line(s) (1595 bytes) to the model. Reported long numeric identifier(s): `20260702012108292`, `2563976197735424`, `2563742259381288`, `20260702034328714`, `2564283245200384`, `2563742214915084`.

**[Step 468 | 2026-07-02T14:34:58.172Z | REASONING SUMMARY]**

***Examining XRD and absorbance data***

*I need to summarize different analyses. The XRD shows a high y range, but the background might be from the substrate, so I need to focus on peak detection and possibly view XRD images, then generate plots or a summary. I'll use Python to compute statistics and identify the top peaks. For the IR data, the y range is hugely negative/positive, which could indicate bad reflectance due to a sparse sample. UV absorbance is near negative too, suggesting no product signal. Electrochem results are significant, requiring inspection of nested structures, potentially using recursive key finding.*

**[Step 469 | 2026-07-02T14:35:03.724Z | TOOL CALL: exec_command]**

Ran Python to parse, transform or validate structured experiment data. Target(s): `task_2562915314040832_results`. Working directory: `/workspace`.

**[Step 470 | 2026-07-02T14:35:04.565Z | TOOL RESULT: exec_command]**

Shell tool completed successfully; it returned 38 line(s) (611 bytes) to the model.

**[Step 471 | 2026-07-02T14:35:12.382Z | MODEL]**

The returned spectroscopy JSONs are valid, but the UV-Vis traces have mostly negative or near-zero absorbance ranges, and the IR y-ranges are physically suspect. That supports your observation that the sampling film/liquid presentation was poor. I'm digging into the electrochemical chart arrays next.

**[Step 472 | 2026-07-02T14:35:20.014Z | TOOL CALL: exec_command]**

Ran Python to parse, transform or validate structured experiment data. Target(s): `task_2562915314040832_results`. Working directory: `/workspace`.

**[Step 473 | 2026-07-02T14:35:20.595Z | TOOL RESULT: exec_command]**

Shell tool completed successfully; it returned 29 line(s) (1257 bytes) to the model. Reported long numeric identifier(s): `2562818306638886`, `22751301650603747`, `2562818306638887`.

**[Step 474 | 2026-07-02T14:35:37.692Z | MODEL]**

The electrochemical result contains six CA constant-potential traces, not a normal UOR polarization dataset. The currents sit in the microampere range, which is compatible with the observation that almost no catalyst film reached the electrode surface. I'm making a compact analysis table for all six samples now.

**[Step 475 | 2026-07-02T14:35:59.627Z | TOOL CALL: exec_command]**

Ran Python to parse, transform or validate structured experiment data. Target(s): `task_2562915314040832_results`, `near`. Working directory: `/workspace`.

**[Step 476 | 2026-07-02T14:36:00.173Z | TOOL RESULT: exec_command]**

Shell tool completed successfully; it returned 10 line(s) (4954 bytes) to the model. Reported long numeric identifier(s): `2562915314040832`, `1433395500855799`, `22751301650603747`, `1492543693894754`, `150691116817695`, `43785089176913705`.

**[Step 477 | 2026-07-02T14:36:18.051Z | REASONING SUMMARY]**

***Analyzing sensor data***

*I've come across an odd sensor reading showing a max temperature of 999.8, so I need to disregard outliers and maybe compute the 95th percentile instead. I'm pondering if I should sample four current readings, but they still seem low around 0.2 mA, which might just be an artifact. The user mentioned there's no valid electrochemical data, and I need to be cautious about claims since the film is sparse. Comparing XRD peaks could help clarify things. I'll also check some JPEGs for XRD and IR/UV to use in my analysis.*

**[Step 478 | 2026-07-02T14:36:18.736Z | TOOL CALL: view_image]**

Opened a local experimental image for visual inspection. Image: `20260702012108469_1_2563976197735424_2563742259381288PABT0002.jpg`.

**[Step 479 | 2026-07-02T14:36:18.797Z | TOOL RESULT: view_image]**

Image service returned the requested local image to the model for visual inspection.

**[Step 480 | 2026-07-02T14:36:43.983Z | MODEL]**

The XRD image shows a broad substrate/background-like hump rather than clean catalyst reflections, matching a thin/uneven deposit. I'm checking the IR/UV plot images as well, then I'll move to the next design: explicitly increase product mass and improve deposition density before trying UOR again.

**[Step 481 | 2026-07-02T14:36:43.984Z | TOOL CALL: exec_command]**

Ran Python analysis to generate or inspect a scientific plot. Target(s): `task_2562915314040832_results`, `sub).glob(`. Working directory: `/workspace`.

**[Step 482 | 2026-07-02T14:36:43.985Z | TOOL CALL: exec_command]**

Ran Python to parse, transform or validate structured experiment data. Target(s): `task_2562915314040832_results`. Working directory: `/workspace`.

**[Step 483 | 2026-07-02T14:36:44.143Z | TOOL RESULT: exec_command]**

Shell tool completed successfully; it returned 2 line(s) (96 bytes) to the model. Reported long numeric identifier(s): `2562915314040832`.

**[Step 484 | 2026-07-02T14:36:44.143Z | TOOL RESULT: exec_command]**

Shell tool completed successfully; it returned 3 line(s) (139 bytes) to the model.

**[Step 485 | 2026-07-02T14:36:58.822Z | TOOL CALL: view_image]**

Opened a local experimental image for visual inspection. Image: `spectra_plot_montage.jpg`.

**[Step 486 | 2026-07-02T14:36:58.882Z | TOOL RESULT: view_image]**

Image service returned the requested local image to the model for visual inspection.

**[Step 487 | 2026-07-02T14:38:51.078Z | REASONING SUMMARY 1/6]**

***Clarifying liquid volumes***

*I need to consider liquid volumes: using 1 mL max per source, and maybe up to 3 mL. If I look at the total volume per vial, I'm thinking around 8 mL. For metal stocks, I could increase concentration rather than volume, using either 0.2 M or 0.5 M. If the concentrations are already preconfigured, I might want to ensure the user knows we can set stocks to support their needs. Total amounts might increase significantly, so I'm planning accordingly.*

**[Step 488 | 2026-07-02T14:38:51.078Z | REASONING SUMMARY 2/6]**

***Optimizing experimental design***

*I'm working on optimizing the experimental design and need to focus on physical failure. With only six vials, I want to include synthesis, post-processing, and characterization. I have some variables to consider, like total metal feed concentration, KOH concentration, and aging conditions. I'm also thinking about balancing carbon concentration so it aids conductivity without overwhelming the results. Finally, I need to ensure the final volume works for the necessary testing methodologies without exceeding the solid mass limits.*

**[Step 489 | 2026-07-02T14:38:51.078Z | REASONING SUMMARY 3/6]**

***Evaluating electrochemistry setup***

*For the electrochemistry work, I need to ensure the catalyst slurry is over 4 mL, ideally around 4.2 mL, before dropping 100 µL. I've noticed sedimentation occurs quickly, so I should add sonication and proceed to testing promptly. I'm also considering the water to isopropanol ratio—whether 50:50 or 80:20 might be better. I must be cautious with high IPA concentrations affecting stability. I'll test three redispersion ratios and keep the source bottle limits in mind for the liquid station setup.*

**[Step 490 | 2026-07-02T14:38:51.078Z | REASONING SUMMARY 4/6]**

***Assessing purification process***

*I'm analyzing the metal stock for different compositions and liquid volumes: total stock of 1.8 mL, including specifics for Ni70Fe10Co20 and Ni80Fe10Co10. I'm considering the impacts of washing steps, and maybe I should reduce the washes to two water and one ethanol, as excessive washing could lead to product loss. It seems like using one wash for water and one for ethanol might minimize losses better. I'll have to balance this in my next design to optimize purification.*

**[Step 491 | 2026-07-02T14:38:51.078Z | REASONING SUMMARY 5/6]**

***Planning post-treatment steps***

*I'm thinking about the post-treatment process and how to set the washing steps to two washes with water and ethanol. I'll use a speed of 10,000 RPM for 15 minutes and a manual wash volume of 5 mL to minimize losses. I'm considering using six vials to accommodate everything better. For drying, I'll set it to 60°C for 120 minutes. I also need to determine the appropriate redispersion ratio and ensure I meet the XRD/IR volume thresholds, potentially adjusting to 4.8 or 5.0 mL.*

**[Step 492 | 2026-07-02T14:38:51.078Z | REASONING SUMMARY 6/6]**

***Considering imaging methods***

*I'm thinking about how to tackle failure by using imaging after redispersion to evaluate turbidity and sedimentation. Since multiple rounds are needed, characterization should involve XRD, IR, and possibly darkbox imaging. I need to decide if I should include this with the existing Station—specifically the Darkbox Imaging Workstation. If I add this, maybe it won't need any preconditions? I'll need to inspect and see what we should do next.*

**[Step 493 | 2026-07-02T14:38:59.345Z | MODEL]**

The first-round evidence points to a process failure more than a composition failure: too little recoverable solid, unstable slurry settling within about 30 s, sparse XRD/IR films, and nearly blank electrochemical films. For Round 2 I'm shifting from fine composition screening to a higher-solids/process screen: much higher metal precursor concentration, higher carbon loading, gentler lower-volume washing, tighter centrifugation, and a redispersion-solvent ratio screen to improve drop-cast coverage.

**[Step 494 | 2026-07-02T14:38:59.346Z | TOOL CALL: exec_command]**

Read selected file content for inspection. Target(s): `SKILL.md`. Working directory: `/workspace`.

**[Step 495 | 2026-07-02T14:38:59.346Z | TOOL CALL: exec_command]**

Read selected file content for inspection. Target(s): `Darkbox_Imaging_Workstation_V1_audit.md`. Working directory: `/workspace`.

**[Step 496 | 2026-07-02T14:38:59.346Z | TOOL CALL: exec_command]**
Read selected file content for inspection. Target(s): `SKILL.md`. Working directory: `/workspace`.

**[Step 497 | 2026-07-02T14:38:59.347Z | TOOL CALL: exec_command]**
Read selected file content for inspection. Target(s): `SKILL.md`. Working directory: `/workspace`.

**[Step 498 | 2026-07-02T14:38:59.398Z | TOOL RESULT: exec_command]**
Shell tool completed successfully; it returned 42 line(s) (2089 bytes) to the model. Reported long numeric identifier(s): `2061168712647680`.

**[Step 499 | 2026-07-02T14:38:59.399Z | TOOL RESULT: exec_command]**
Shell tool completed successfully; it returned 16 line(s) (398 bytes) to the model. Reported long numeric identifier(s): `2061168712647680`.

**[Step 500 | 2026-07-02T14:38:59.399Z | TOOL RESULT: exec_command]**
Shell tool completed successfully; it returned 50 line(s) (4283 bytes) to the model. Reported long numeric identifier(s): `1435269745574912`.

**[Step 501 | 2026-07-02T14:38:59.407Z | TOOL RESULT: exec_command]**
Shell tool completed successfully; it returned 51 line(s) (2225 bytes) to the model. Reported long numeric identifier(s): `1421325849887744`.

**[Step 502 | 2026-07-02T14:39:31.464Z | MODEL]**
I won't add darkbox imaging to Round 2 because that station requires a 96-well plate, which would break the single-container-type workflow around electrochemistry. The next task will stay in vials and focus on fixing the actual bottleneck: recoverable solid mass, slurry concentration, and deposition solvent.

**[Step 503 | 2026-07-02T14:39:31.465Z | TOOL CALL: exec_command]**
Read selected file content for inspection. Target(s): `SKILL.md`. Working directory: `/workspace`.

**[Step 504 | 2026-07-02T14:39:31.465Z | TOOL CALL: exec_command]**
Read selected file content for inspection. Target(s): `SKILL.md`. Working directory: `/workspace`.

**[Step 505 | 2026-07-02T14:39:31.465Z | TOOL CALL: exec_command]**
Read selected file content for inspection. Target(s): `SKILL.md`. Working directory: `/workspace`.

**[Step 506 | 2026-07-02T14:39:31.466Z | TOOL CALL: exec_command]**
Read selected file content for inspection. Target(s): `SKILL.md`. Working directory: `/workspace`.

**[Step 507 | 2026-07-02T14:39:31.466Z | TOOL CALL: exec_command]**
Read selected file content for inspection. Target(s): `SKILL.md`. Working directory: `/workspace`.

**[Step 508 | 2026-07-02T14:39:31.467Z | TOOL CALL: exec_command]**
Read selected file content for inspection. Target(s): `SKILL.md`. Working directory: `/workspace`.

**[Step 509 | 2026-07-02T14:39:31.503Z | TOOL RESULT: exec_command]**
Shell tool completed successfully; it returned 39 line(s) (1403 bytes) to the model. Reported long numeric identifier(s): `1427568512205824`.

**[Step 510 | 2026-07-02T14:39:31.510Z | TOOL RESULT: exec_command]**
Shell tool completed successfully; it returned 95 line(s) (6547 bytes) to the model. Reported long numeric identifier(s): `2002186824385539`.

**[Step 511 | 2026-07-02T14:39:31.556Z | TOOL RESULT: exec_command]**
Shell tool completed successfully; it returned 39 line(s) (1812 bytes) to the model. Reported long numeric identifier(s): `2270587971011585`.

**[Step 512 | 2026-07-02T14:39:31.556Z | TOOL RESULT: exec_command]**
Shell tool completed successfully; it returned 38 line(s) (1388 bytes) to the model. Reported long numeric identifier(s): `1213184425559040`.

**[Step 513 | 2026-07-02T14:39:31.557Z | TOOL RESULT: exec_command]**
Shell tool completed successfully; it returned 41 line(s) (1532 bytes) to the model. Reported long numeric identifier(s): `1526864804185088`.

**[Step 514 | 2026-07-02T14:39:31.557Z | TOOL RESULT: exec_command]**
Shell tool completed successfully; it returned 57 line(s) (3186 bytes) to the model. Reported long numeric identifier(s): `1654567694238720`.

**[Step 515 | 2026-07-02T14:41:28.414Z | MODEL]**
**Handoff Summary**

**Goal / Workflow State**
- User asked for an autonomous PI loop: choose a high-impact catalysis problem, design and dispatch experiments to Lab 303, block after each dispatch pending executability feedback, then use only physically returned Lab 303 data to iterate.
- Chosen scientific target: alkaline urea oxidation reaction (UOR) using non-noble Ni-Fe-Co oxyhydroxide/carbon catalysts for energy-saving hydrogen production and urea wastewater remediation.
- Current goal status should remain active after this handoff, but the immediate user request is a context checkpoint summary.
- Skills used and must be used again when resuming experimental/lab actions:
  - `lab-operation`: read `/root/.codex/skills/lab-operation/SKILL.md`
  - `chemistry-experiment-workstation`: read `/root/.codex/skills/chemistry-experiment-workstation/SKILL.md`
  - `workflow-generator`: read `/root/.codex/skills/workflow-generator/SKILL.md`

**Important User Constraints / Preferences**
- Do not fabricate data. Only physical data returned by Lab 303 count.

- Dispatch success is not physical execution success.
- Each experimental round must include synthesis, testing, and characterization.
- Use available inventory whenever possible. Inventory files came from `$AIREADY_REAGENT_LIST_DIR`; parsed inventory saved under `/workspace/lab303_uor_nicofe_closed_loop/reagent_inventory.csv`.
- All files/deliverables must stay under `/workspace`.
- Lab 303 dispatch loop: design -> dispatch -> set goal blocked -> wait for executability feedback. If executable, physical experiment considered completed; retrieve Lab 303 data and iterate.
- User does not need redundant template resends when asking direct clarification, but for actual revised experiments we should generate/dispatch new workflow and record IDs.

**Current Files / Directory**
- Main project directory: `/workspace/lab303_uor_nicofe_closed_loop`
- Key files already created:
  - `project_overview.md`
  - `reagent_inventory.csv`
  - `available_reagents_for_round1.json`
  - Round 1 original/revisions design, workflow request, generator output, dispatch output/log files up through `round1_revised3_*`
  - Physical task data directory:
    - `/workspace/lab303_uor_nicofe_closed_loop/task_2562915314040832_results`
    - Includes downloaded returned files and summaries:
      - `manifest.json`
      - `round1_data_summary.csv`
      - `round1_data_summary.json`
      - `spectra_plot_montage.jpg`

**Round 1 Dispatch History**
- Initial task rejected:
  - Template `2557370191937536`, task `2557370420494337`
  - Reasons: no post-treatment before characterization/testing; no product analysis after electrochemistry.
- Revised task rejected:
  - Template `2557487004614656`, task `2557487435546625`
  - Reasons: purification wash-liquid settings did not match wash count; electrochemical product retention settings incorrect.
- Revised2 task rejected / clarified:
  - Template `2557817450004480`, task `2557818017710081`
  - Later user asked missing chemistry specs.
- Revised3 was dispatched:
  - Template `2560074458595328`
  - Task `2560075721736193`
  - Included explicit UOR electrolyte and carbon dispersion specs.
- User later reported “first round real experiments completed” and provided actual physical task ID `2562915314040832`. This may be a platform-created/executed task corresponding to the working revised protocol. Use this task ID for physical data.

**Round 1 Explicit Chemistry Specs**
- UOR electrolyte: `1.0 M KOH + 0.33 M urea in deionized water`, `20 mL` per electrochemical task.
- Carbon powder mixture: `5 mg/mL carbon black dispersion in 80:20 v/v deionized water:isopropanol`, `0.500 mL` per synthesis vial.
- Post-treatment redispersion solvent: `80:20 v/v deionized water:isopropanol`, `4.800 mL` per vial.
- Retained product aliquot diluent: deionized water, `10000 uL`.

**Physical Round 1 Observations From User**
- After synthesis, centrifugation, and washing: bottom precipitate was not obvious.
- After adding water/isopropanol and ultrasonication: system became turbid.
- After standing ~30 s: solid sedimentation at bottom and liquid stratification/clarification occurred.
- XRD drop-cast: sample sparse and uneven on carrier surface.
- FTIR carrier surface similar.
- Electrode surface had very little sample.
- No valid electrochemical test results were obtained.
- Uploaded image path was `/var/folders/32/dx7813bn7vg799ms5k5xqptc0000gn/T/otty-paste/image-1783002768742.png`, but not accessible in runtime. We relied on user description and downloaded Lab 303 result files.

**Lab 303 Physical Task Query**
- Queried task ID `2562915314040832` using `get_task_result.py`.
- Saved result JSON:
  - `/workspace/lab303_uor_nicofe_closed_loop/round1_physical_task_2562915314040832_result.json`
- Saved status:
  - `/workspace/lab303_uor_nicofe_closed_loop/round1_physical_task_2562915314040832_status.json`
- Status returned: `错误`
- Result returned: success true, 5 workstations, 38 files:
  - Heating/magnetic stirring CSV
  - XRD JSON/JPG x 6
  - IR JSON/JPG x 6
  - Electrochemical JSON x 1
  - UV-Vis JSON/JPG x 6
- Downloaded files under:
  - `/workspace/lab303_uor_nicofe_closed_loop/task_2562915314040832_results`

**Round 1 Data Interpretation**
- Heating log:
  - Reached ~69-70 C, so heating worked.
  - One temperature outlier `999.8`; clean data excluding outlier: max 70.0, median 70.0.
- XRD:
  - JSON valid; plots mostly broad background/substrate-like humps.
  - Example XRD image viewed: broad hump, no clean catalyst reflections.
  - Supports thin/sparse/uneven sample deposition.

- IR:
  - JSON valid but very noisy/physically suspect ranges, likely poor/sparse sample on carrier.
- UV-Vis:
  - First five samples mostly negative or near-zero absorbance; sample 6 had stronger absorbance.
  - Does not rescue electrochemical failure; product/sampling unreliable.
- Electrochemistry:
  - JSON contained 6 `CAConstantPotential` traces, not meaningful UOR performance curves.
  - Most samples had raw currents around `~1.1e-5 A`; sample 4 had `~2.0e-4 A`, but user states no valid test due to too little sample. Treat as invalid for catalyst ranking.
- Summary files:
  - `round1_data_summary.csv/json` contain sample-level summaries.
- Key conclusion: failure is process/deposition/mass-transfer/preparation failure, not a valid composition-performance failure. The primary bottleneck is too little recoverable solid and unstable low-concentration slurry causing sparse XRD/IR/electrode films.

**Current Design Direction For Round 2**
- Need to continue from physical data and design Round 2.
- Next round should optimize synthesis/film preparation, not composition fine-tuning.
- Suggested Round 2 strategy:
  - Keep fewer composition points centered around samples that are chemically plausible and maybe sample 4 (`Ni70Fe10Co20`) due to higher raw current, but do not treat it as valid performance winner.
  - Use higher-solids synthesis:
    - Increase metal precursor concentration and/or volume substantially.
    - Increase carbon black concentration.
    - Reduce final redispersion volume to increase slurry concentration.
  - Improve solid recovery:
    - Stronger centrifugation and/or fewer/gentler washes.
    - Keep solid (`方案=2`) but reduce wash loss; maybe 1-2 washes instead of 3.
  - Improve deposition:
    - Higher catalyst slurry concentration.
    - Use repeated drop-casting if electrochemical station allows via `移液枪滴涂次数` up to 20.
    - Larger catalyst droplet volume in allowed range: `100 uL` max for electrochemical station.
    - Consider water/isopropanol ratio screen for redispersion, e.g. `60:40`, `80:20`, `95:5`, but workflow can only specify source bottles; keep within max source volumes.
  - Aim first to produce visible, uniform XRD/IR and electrode films before interpreting UOR performance.
- Avoid darkbox imaging unless switching to 96-well plates. Darkbox station requires `96位塑料孔板` or `96位石英孔板`, which conflicts with single-container-type rule and electrochemistry in vials. I had decided not to use it in Round 2.
- Need to probably use same workstation set as revised protocol: material station, liquid handling 1 mL/5 mL as needed, heating magnetic stirring, purification, drying, ultrasonic, spectroscopy transfer/stirring, XRD, IR, electrochem, UV-Vis, storage.
- If using larger per-source volumes >3 mL, use `Liquid_Handling_Station_5ml_V1` for add-liquid only, but it cannot open/close, so use `Liquid_Handling_Station_1ml_V2` for lids and smaller additions. 5ml station supports at most 6 different liquids and max 30 mL per source total. 1ml station supports up to 16 liquids, max 3 mL per source per vial.

**Relevant Workstation Constraints Already Read**
- `General_Material_Station_V1`: starts with capped empty vials; if output vials, next must be liquid platform or weighing.
- `Liquid_Handling_Station_1ml_V2`: open/add/close; max 1 mL per pipette; source total per solution <=3 mL; max 16 stocks. Required open/close arrays as object arrays: `{"瓶号": 1}`.
- `Liquid_Handling_Station_5ml_V1`: add liquid only; input open; max 5 mL single transfer; source total <=30 mL; max 6 stocks.
- `Heating_Magnetic_Stirring_Workstation_V1`: input capped vials; temp <80 C valid for vials; speed 100-1500 rpm.
- `Purification_Workstation_V1`: input capped, even count 1-10; keep solid with `方案=2`; wash count and `清洗液设置` list must match.
- `Drying_Oven_V1`: vials OK, 26-210 C, 1-14400 min.
- `Ultrasonic_Disperser_V1`: vials OK, 1-10, time in seconds.
- `Spectroscopy_Container_Transfer_Station_V1`: input open, vials OK.
- `Spectroscopy_Magnetic_Stirrer_Workstation_V1`: input open vials, max 20.
- `XRD_V1`: needs spectroscopy transfer and spectroscopy magnetic stir before; input open suspension/liquid; volume >=4 mL; max 40 vials; output open.
- `Infrared_Spectrometer_V1`: needs spectroscopy transfer and magnetic stir before; input open suspension/liquid; volume >=4 mL.
- `Dual_Station_Electrochemical_Workstation_V2`: input open vials, >4 mL suspension; max 10 tasks; catalyst droplet 29-100 uL; `废液移液设置=yes` requires complete `留样信息列表`; UOR template ID/name: `2468876630032384` / `UOR test`; electrochemical results from last run were invalid because catalyst loading failed.
- `UV_Vis_Spectrometer_V1`: input open pure liquid, >=15 mL; okay after electrochem because electrolyte is 20 mL.
- End every full synthesis workflow with `置物工作站 - 物料放置`.

**Potential Round 2 Concrete Design**
- Consider 6-vial process screen:
  - Fix composition near Ni70Fe10Co20 (sample 4), maybe include Ni80Fe10Co10 and Ni60Fe20Co20 variants.
  - Screen solids/deposition conditions rather than many compositions.
- Example scheme:
  - Vials 1-3: Ni70Fe10Co20 at different synthesis concentration/carbon load:
    - medium/high metal + carbon black 10 mg/mL
  - Vials 4-6: compare solvent redispersion ratios or carbon concentration.
- But must stay executable:
  - 1 mL station source limit means total per single source per vial <=3 mL. Can define concentrated stock solutions:
    - e.g. `0.25 M NiSO4`, `0.25 M FeCl3`, `0.25 M CoSO4`, `2.0 M KOH`, `20 mg/mL carbon black dispersion`.
  - This is allowed because "stock solutions are preconfigured"; specify them clearly from inventory.
- To boost product mass:
  - Previous total metal stock per vial: 0.9 mL of 50 mM = 0.045 mmol total metal, very low.
  - Round 2: use 0.9 mL of 0.25 M total metal = 0.225 mmol, 5x metal.
  - Increase carbon stock from 5 mg/mL 0.5 mL = 2.5 mg to 20 mg/mL 1.0 mL = 20 mg, 8x carbon.
  - Reduce redispersion volume from 4.8 mL to exactly 4.1 mL or 4.2 mL to remain above XRD/IR/echem minimum but increase concentration. Remember XRD/IR drops consume 80 uL each, still >4 mL for echem if initial 4.3+.

- Prefer 4.4 mL redispersion to keep margin after XRD and IR: 4.4 - 0.08 - 0.08 = 4.24 mL >4.
- Potential Round 2 formulations:
  - Six vials, all Ni70Fe10Co20 composition but process screen:
    1. Carbon 10 mg/mL 0.8 mL, redisp 80:20, electro drop 100 uL x 3.
    2. Carbon 20 mg/mL 1.0 mL, redisp 80:20, electro drop 100 uL x 3.
    3. Carbon 20 mg/mL 1.0 mL, redisp 60:40, electro drop 100 uL x 3.
    4. Carbon 20 mg/mL 1.0 mL, redisp 95:5, electro drop 100 uL x 3.
    5. Ni80Fe10Co10 with carbon 20 mg/mL 1.0 mL, redisp best default 80:20.
    6. Ni60Fe20Co20 with carbon 20 mg/mL 1.0 mL, redisp best default 80:20.
  - Or just process screen on composition 4. Scientific reason: sample 4 had the only elevated raw CA current, but not valid ranking.
- Purification:
  - Use `清洗次数=1` or `2` with matching liquid settings, to reduce solid loss.
  - Suggest `清洗次数=1`, `清洗液设置=[{"清洗液":1}]` or 2 water/ethanol. Given low solid recovery, choose `1` water wash only, retain solid.
  - Centrifuge speed 10000 or 12000 rpm, time 15 min, keep solid.
- Deposition:
  - Electrochemical `催化剂滴液量=100`, `移液枪滴涂次数=3` or maybe 5. Since parameter is global and allowed `(0,20]`, use 5 to get 500 uL repeated drying. Ensure not overburden but better coverage.
  - XRD/IR drop liquid volumes can be increased? XRD/IR station `滴液体积` no explicit bounds but use 150-200 uL maybe; however prior used 80 uL and sample sparse. Consider `滴液体积=200` and drying 1800 s to improve coverage. But ensure residual volume remains >4 mL: redisp 4.8 or 5.0 if using higher drops. For safety, use 4.8 mL redisp and XRD/IR 200 uL each => 4.4 mL remains for echem, OK. But user observed stratification; higher concentration also helps. Use 4.8 mL to satisfy instrumentation, but sample mass much higher.
- Add `容器置放平台 - 静置` after ultrasonication? User observed 30 s sedimentation; if waiting before sampling worsens. Avoid standing before deposition, or if unavoidable, keep workflow immediate. If station requires transfer, no explicit wait.

**Need To Do Next**
1. Resume actual work after this summary by designing Round 2 using above logic.
2. Read any additional station SKILL files needed before generating workflow; at minimum re-read or rely if already in context:
   - Purification, Liquid 1ml/5ml, XRD, IR, electrochem, UV-Vis, storage.
3. Write Round 2 design files:
   - `round2_design.md`
   - `round2_physical_data_analysis.md`
   - `round2_workflow_request.json`
   - `round2_constraint_check.md`
   - possibly `round2_chemical_specifications.json`
4. Generate workflow with:
   - `python /root/.codex/skills/workflow-generator/scripts/generate.py '<JSON>'`
5. Dispatch with:
   - `cd /root/.codex/skills/lab-operation && python scripts/generate_task.py --template-id <id> --template-source-label 303Lab --app-label 303Lab`
6. Save outputs/logs under `/workspace/lab303_uor_nicofe_closed_loop`.
7. Set goal to blocked via `functions.update_goal {"status":"blocked"}`.
8. Final response should include Round 2 template/task IDs and concise rationale.

**Critical Data / Values To Preserve**
- Physical task ID queried: `2562915314040832`
- Returned status: `错误`
- Result files: 5 workstations, 38 files.
- Round 1 data summary:
  - Sample 4 (`Ni70Fe10Co20`) had elevated raw CA current mean `~2.05e-4 A`, but data invalid due to poor electrode film.
  - Other samples raw CA currents `~1.15e-5 A`.
  - Sample 6 UV absorbance much higher, but not a valid product conclusion.
  - XRD broad humps only; IR poor/noisy; films sparse.
- Prior executed/revised chemistry:
  - Electrolyte: `1.0 M KOH + 0.33 M urea in DI water`, 20 mL.
  - Carbon dispersion: previously `5 mg/mL`, `80:20 DI water:IPA`.
- Round 2 should explicitly specify all stock identities/concentrations and solvent ratios to avoid previous executability feedback.

**[Step 516 | 2026-07-02T14:41:48.473Z | MODEL]**

I’ll resume from the Lab 303 task state already in `/workspace`, re-read the required lab/workflow skill instructions, then verify the returned task data before designing and dispatching the next round.

**[Step 517 | 2026-07-02T14:41:48.475Z | TOOL CALL: exec_command]**

Read selected file content for inspection. Target(s): `SKILL.md`. Working directory: `/workspace`.

**[Step 518 | 2026-07-02T14:41:48.476Z | TOOL CALL: exec_command]**

Read selected file content for inspection. Target(s): `SKILL.md`. Working directory: `/workspace`.

**[Step 519 | 2026-07-02T14:41:48.477Z | TOOL CALL: exec_command]**

Read selected file content for inspection. Target(s): `SKILL.md`. Working directory: `/workspace`.

**[Step 520 | 2026-07-02T14:41:48.478Z | TOOL CALL: get_goal]**

Queried the current autonomous-research goal state and remaining budget.

**[Step 521 | 2026-07-02T14:41:48.525Z | TOOL RESULT: exec_command]**

Shell tool completed successfully; it returned 116 line(s) (4054 bytes) to the model.

**[Step 522 | 2026-07-02T14:41:48.526Z | TOOL RESULT: exec_command]**

Shell tool completed successfully; it returned 170 line(s) (8332 bytes) to the model. Reported long numeric identifier(s): `1427568512205824`, `1834679312417792`, `1710453843264512`.

**[Step 523 | 2026-07-02T14:41:48.535Z | TOOL RESULT: exec_command]**

Shell tool completed successfully; it returned 95 line(s) (2714 bytes) to the model. Reported long numeric identifier(s): `1427568512205824`.

**[Step 524 | 2026-07-02T14:41:48.543Z | TOOL RESULT: get_goal]**

Goal service returned a structured response. Status: `blocked`.

**[Step 525 | 2026-07-02T14:42:04.701Z | MODEL]**

The goal is currently marked blocked from the previous dispatch, and your completion report is enough to resume the iteration. I'm checking the saved round-one artifacts and the Lab 303 result query path so the next design is tied to returned data, not just dispatch metadata.

**[Step 526 | 2026-07-02T14:42:04.701Z | TOOL CALL: exec_command]**

Enumerated files, directories or available laboratory resources. Target(s): `lab303_uor_nicofe_closed_loop`. Working directory: `/workspace`.

**[Step 527 | 2026-07-02T14:42:04.705Z | TOOL CALL: exec_command]**

Read selected file content for inspection. Target(s): `README.md`. Working directory: `/workspace`.

**[Step 528 | 2026-07-02T14:42:04.706Z | TOOL CALL: exec_command]**

Enumerated files, directories or available laboratory resources. Target(s): `lab303_uor_nicofe_closed_loop`. Working directory: `/workspace`.

**[Step 529 | 2026-07-02T14:42:04.707Z | TOOL CALL: exec_command]**

Ran a shell command to inspect, generate or validate experiment artifacts. Working directory: `/workspace`.

**[Step 530 | 2026-07-02T14:42:04.753Z | TOOL RESULT: exec_command]**

Shell tool completed successfully; it returned 40 line(s) (3005 bytes) to the model. Reported long numeric identifier(s): `2562915314040832`.

**[Step 531 | 2026-07-02T14:42:04.765Z | TOOL RESULT: exec_command]**

Shell tool completed successfully; it returned 261 line(s) (7869 bytes) to the model.

**[Step 532 | 2026-07-02T14:42:04.765Z | TOOL RESULT: exec_command]**

Shell tool completed successfully; it returned 40 line(s) (2877 bytes) to the model. Reported long numeric identifier(s): `2562915314040832`.

**[Step 533 | 2026-07-02T14:42:04.766Z | TOOL RESULT: exec_command]**

Shell tool completed successfully; it returned 2 line(s) (27 bytes) to the model.

**[Step 534 | 2026-07-02T14:42:11.965Z | REASONING SUMMARY]**

***Evaluating lab README needs***

*I'm considering whether I need to read the entire lab README. I did a quick check on the first 260 lines, but if it's longer than that, I might be missing important info. I'll check the line count and also need to read additional workstation skills, since those seem relevant. It might help to inspect existing workflow names and IDs to avoid unnecessary reading. I'll aim to reuse station IDs if possible. Maybe I'll start with the first 200 lines and use jq to help.*

**[Step 535 | 2026-07-02T14:42:14.911Z | MODEL]**

The prior result query artifacts are present, including the Lab 303 task result JSON and downloaded result directory. I'm now reading the specific workstation constraints for the revised workflow, then I'll generate a tighter round-two protocol focused on fixing solid recovery and film loading.

**[Step 536 | 2026-07-02T14:42:14.913Z | TOOL CALL: exec_command]**

Read selected file content for inspection. Target(s): `SKILL.md`. Working directory: `/workspace`.

**[Step 537 | 2026-07-02T14:42:14.914Z | TOOL CALL: exec_command]**

Read selected file content for inspection. Target(s): `SKILL.md`. Working directory: `/workspace`.

**[Step 538 | 2026-07-02T14:42:14.915Z | TOOL CALL: exec_command]**

Read selected file content for inspection. Target(s): `SKILL.md`. Working directory: `/workspace`.

**[Step 539 | 2026-07-02T14:42:14.916Z | TOOL CALL: exec_command]**

Read selected file content for inspection. Target(s): `SKILL.md`. Working directory: `/workspace`.

**[Step 540 | 2026-07-02T14:42:14.962Z | TOOL RESULT: exec_command]**

Shell tool completed successfully; it returned 39 line(s) (1403 bytes) to the model. Reported long numeric identifier(s): `1427568512205824`.

**[Step 541 | 2026-07-02T14:42:14.969Z | TOOL RESULT: exec_command]**

Shell tool completed successfully; it returned 95 line(s) (6547 bytes) to the model. Reported long numeric identifier(s): `2002186824385539`.

**[Step 542 | 2026-07-02T14:42:14.969Z | TOOL RESULT: exec_command]**

Shell tool completed successfully; it returned 50 line(s) (4283 bytes) to the model. Reported long numeric identifier(s): `1435269745574912`.

**[Step 543 | 2026-07-02T14:42:14.988Z | TOOL RESULT: exec_command]**

Shell tool completed successfully; it returned 39 line(s) (1812 bytes) to the model. Reported long numeric identifier(s): `2270587971011585`.

**[Step 544 | 2026-07-02T14:42:24.134Z | MODEL]**

The key round-one failure mode is preparation-limited: the chemistry may have produced some solid, but recovered/deposited mass was too low and the suspension sedimented quickly. The next round needs to deliberately over-correct mass recovery and film formation before treating current as a valid catalyst ranking.

**[Step 545 | 2026-07-02T14:42:24.134Z | TOOL CALL: exec_command]**

Read selected file content for inspection. Target(s): `SKILL.md`. Working directory: `/workspace`.

**[Step 546 | 2026-07-02T14:42:24.135Z | TOOL CALL: exec_command]**

Read selected file content for inspection. Target(s): `SKILL.md`. Working directory: `/workspace`.

**[Step 547 | 2026-07-02T14:42:24.135Z | TOOL CALL: exec_command]**

Read selected file content for inspection. Target(s): `SKILL.md`. Working directory: `/workspace`.

**[Step 548 | 2026-07-02T14:42:24.136Z | TOOL CALL: exec_command]**

Read selected file content for inspection. Target(s): `SKILL.md`. Working directory: `/workspace`.

**[Step 549 | 2026-07-02T14:42:24.168Z | TOOL RESULT: exec_command]**

Shell tool completed successfully; it returned 51 line(s) (2778 bytes) to the model. Reported long numeric identifier(s): `1327807622448128`.

**[Step 550 | 2026-07-02T14:42:24.185Z | TOOL RESULT: exec_command]**

Shell tool completed successfully; it returned 38 line(s) (1388 bytes) to the model. Reported long numeric identifier(s): `1213184425559040`.

**[Step 551 | 2026-07-02T14:42:24.195Z | TOOL RESULT: exec_command]**

Shell tool completed successfully; it returned 41 line(s) (1532 bytes) to the model. Reported long numeric identifier(s): `1526864804185088`.

**[Step 552 | 2026-07-02T14:42:24.195Z | TOOL RESULT: exec_command]**

Shell tool completed successfully; it returned 51 line(s) (2225 bytes) to the model. Reported long numeric identifier(s): `1421325849887744`.

**[Step 553 | 2026-07-02T14:42:35.285Z | MODEL]**

I'm keeping the same container family to avoid cross-workstation handoff issues and using six vials again, because purification requires an even number and the electrochemical station accepts up to ten tests. The main change is much higher recoverable solids, one matched wash, and larger/repeated deposition.

**[Step 554 | 2026-07-02T14:42:35.286Z | TOOL CALL: exec_command]**

Read selected file content for inspection. Target(s): `SKILL.md`. Working directory: `/workspace`.

**[Step 555 | 2026-07-02T14:42:35.287Z | TOOL CALL: exec_command]**

Read selected file content for inspection. Target(s): `SKILL.md`. Working directory: `/workspace`.

**[Step 556 | 2026-07-02T14:42:35.290Z | TOOL CALL: exec_command]**

Read selected file content for inspection. Target(s): `SKILL.md`. Working directory: `/workspace`.

**[Step 557 | 2026-07-02T14:42:35.291Z | TOOL CALL: exec_command]**

Read selected file content for inspection. Target(s): `SKILL.md`. Working directory: `/workspace`.

**[Step 558 | 2026-07-02T14:42:35.348Z | TOOL RESULT: exec_command]**

Shell tool completed successfully; it returned 57 line(s) (3186 bytes) to the model. Reported long numeric identifier(s): `1654567694238720`.

**[Step 559 | 2026-07-02T14:42:35.348Z | TOOL RESULT: exec_command]**

Shell tool completed successfully; it returned 35 line(s) (1292 bytes) to the model. Reported long numeric identifier(s): `1656241422599168`.

**[Step 560 | 2026-07-02T14:42:35.348Z | TOOL RESULT: exec_command]**

Shell tool completed successfully; it returned 56 line(s) (3424 bytes) to the model. Reported long numeric identifier(s): `1664870018155520`.

**[Step 561 | 2026-07-02T14:42:35.349Z | TOOL RESULT: exec_command]**

Shell tool completed successfully; it returned 44 line(s) (2187 bytes) to the model. Reported long numeric identifier(s): `1656282895188992`.

**[Step 562 | 2026-07-02T14:42:43.827Z | MODEL]**

For characterization I'm increasing XRD/FTIR drop-cast volume, but keeping enough residual suspension volume for electrochemistry. That directly addresses the sparse carrier images you reported while preserving the test sequence.

**[Step 563 | 2026-07-02T14:42:43.827Z | TOOL CALL: exec_command]**

Read selected file content for inspection. Target(s): `SKILL.md`. Working directory: `/workspace`.

**[Step 564 | 2026-07-02T14:42:43.828Z | TOOL CALL: exec_command]**

Read selected file content for inspection. Target(s): `SKILL.md`. Working directory: `/workspace`.

**[Step 565 | 2026-07-02T14:42:43.828Z | TOOL CALL: exec_command]**

Enumerated files, directories or available laboratory resources. Target(s): `references_audit`. Working directory: `/workspace`.

**[Step 566 | 2026-07-02T14:42:43.828Z | TOOL CALL: exec_command]**

Ran a shell command to inspect, generate or validate experiment artifacts. Target(s): `round1_revised3_workflow_request.json`. Working directory: `/workspace`.

**[Step 567 | 2026-07-02T14:42:43.866Z | TOOL RESULT: exec_command]**

Shell tool completed successfully; it returned 99 line(s) (23587 bytes) to the model. Reported long numeric identifier(s): `1390252863751168`, `1509591942430720`, `1549499770798080`, `1571299393601536`, `1339275293851648`, `2164305671290880`.

**[Step 568 | 2026-07-02T14:42:43.878Z | TOOL RESULT: exec_command]**

Shell tool completed successfully; it returned 38 line(s) (1396 bytes) to the model. Reported long numeric identifier(s): `1664783667397632`.

**[Step 569 | 2026-07-02T14:42:43.881Z | TOOL RESULT: exec_command]**

Shell tool completed successfully; it returned 47 line(s) (5185 bytes) to the model.

**[Step 570 | 2026-07-02T14:42:43.910Z | TOOL RESULT: exec_command]**

Shell tool completed successfully; it returned 942 line(s) (23881 bytes) to the model. Reported long numeric identifier(s): `1427568512205824`, `2002186824385539`, `2270587971011585`, `1327807622448128`, `1213184425559040`, `1526864804185088`.

**[Step 571 | 2026-07-02T14:42:55.984Z | MODEL]**

I'm going to re-query task `2562915314040832` through the Lab 303 operation skill and keep the fresh response under the project directory. Then I'll use the downloaded spectra/electrochem summaries already saved there for the round-two decision.

**[Step 572 | 2026-07-02T14:42:55.985Z | TOOL CALL: exec_command]**

Ran a shell command to identify the active laboratory. Target(s): `get_current_lab_label.py`. Working directory: `/root/.codex/skills/lab-operation`.

**[Step 573 | 2026-07-02T14:42:55.986Z | TOOL CALL: exec_command]**

Ran a shell command to retrieve returned physical-experiment results. Target(s): `get_task_result.py`. Working directory: `/root/.codex/skills/lab-operation`.

**[Step 574 | 2026-07-02T14:42:55.987Z | TOOL CALL: exec_command]**

Read selected file content for inspection. Target(s): `round1_data_summary.csv`. Working directory: `/workspace`.

**[Step 575 | 2026-07-02T14:42:55.989Z | TOOL CALL: exec_command]**

Ran a shell command to inspect, generate or validate experiment artifacts. Target(s): `manifest.json`. Working directory: `/workspace`.

**[Step 576 | 2026-07-02T14:42:56.120Z | TOOL RESULT: exec_command]**

Shell tool completed successfully; it returned 6 line(s) (97 bytes) to the model.

**[Step 577 | 2026-07-02T14:42:56.658Z | TOOL RESULT: exec_command]**

Shell tool completed successfully; it returned 151 line(s) (8415 bytes) to the model. Reported long numeric identifier(s): `20260701203235789`, `20260702012108292`, `2563976197735424`, `2563742259381288`, `20260702012108469`, `20260702013825354`.

**[Step 578 | 2026-07-02T14:42:56.658Z | TOOL RESULT: exec_command]**

Shell tool completed successfully; it returned 8 line(s) (2234 bytes) to the model. Reported long numeric identifier(s): `1433395500855799`, `22751301650603747`, `1492543693894754`, `150691116817695`, `43785089176913705`, `41388722720726706`.

**[Step 579 | 2026-07-02T14:42:56.658Z | TOOL RESULT: exec_command]**

Shell tool completed successfully; it returned 307 line(s) (19588 bytes) to the model. Reported long numeric identifier(s): `20260701203235789`, `2562915314040832`, `20260702012108292`, `2563976197735424`, `2563742259381288`, `20260702012108469`.

**[Step 580 | 2026-07-02T14:43:04.562Z | REASONING SUMMARY]**

***Considering output saving***

*I need to save the fresh output from get_task_result. It seems I didn't redirect it earlier, and I might need to rerun the command to save it to a file. The user asked me to use the skill, so that's important. I think I'll use the `python ... > file` command for this. It looks like I can redirect generated outputs or logs, which should be fine. I might also want to check the task status with select_task_list. Let's proceed with saving the output!*

**[Step 581 | 2026-07-02T14:43:07.804Z | MODEL]**

The fresh Lab 303 query returned `success: true` with 5 workstations and 38 result files. The task status file already indicates the task ended in an error state, so the next round should be judged as a recovery experiment rather than a performance optimization round.

**[Step 582 | 2026-07-02T14:43:07.805Z | TOOL CALL: exec_command]**

Ran a shell command to retrieve returned physical-experiment results. Target(s): `get_task_result.py`, `round1_physical_task_2562915314040832_result_requery_20260702.json`. Working directory: `/root/.codex/skills/lab-operation`.

**[Step 583 | 2026-07-02T14:43:07.806Z | TOOL CALL: exec_command]**

Ran a shell command to query laboratory task records and statuses. Target(s): `select_task_list.py`, `round1_physical_task_2562915314040832_status_requery_20260702.json`. Working directory: `/root/.codex/skills/lab-operation`.

**[Step 584 | 2026-07-02T14:43:07.808Z | TOOL CALL: exec_command]**

Ran a shell command to inspect, generate or validate experiment artifacts. Target(s): `round1_data_summary.json`. Working directory: `/workspace`.

**[Step 585 | 2026-07-02T14:43:07.829Z | TOOL CALL: exec_command]**

Read selected file content for inspection. Target(s): `task_2562915314040832_results`. Working directory: `/workspace`.

**[Step 586 | 2026-07-02T14:43:08.403Z | TOOL RESULT: exec_command]**

Shell tool completed successfully; it returned 0 line(s) (0 bytes) to the model.

**[Step 587 | 2026-07-02T14:43:08.403Z | TOOL RESULT: exec_command]**

Shell tool completed successfully; it returned 0 line(s) (0 bytes) to the model.

**[Step 588 | 2026-07-02T14:43:08.403Z | TOOL RESULT: exec_command]**

Shell tool completed successfully; it returned 174 line(s) (5760 bytes) to the model. Reported long numeric identifier(s): `1433395500855799`, `22751301650603747`, `1492543693894754`, `150691116817695`, `43785089176913705`, `41388722720726706`.

**[Step 589 | 2026-07-02T14:43:08.404Z | TOOL RESULT: exec_command]**

Shell tool completed successfully; it returned 43 line(s) (6626 bytes) to the model. Reported long numeric identifier(s): `2562915314040832`, `20260702012108292`, `2563976197735424`, `2563742259381288`, `20260702012108469`, `20260702013825354`.

**[Step 590 | 2026-07-02T14:43:18.214Z | MODEL]**

I've saved a fresh Lab 303 result re-query and status re-query in the project folder. The usable round-one evidence is limited to preparation and failure diagnostics: heating reached target, XRD/IR had analyzable files but sparse/uneven films, and electrochemical traces are not valid for activity ranking because the electrode loading failed physically.

**[Step 591 | 2026-07-02T14:43:18.215Z | TOOL CALL: exec_command]**

Ran a shell command to inspect, generate or validate experiment artifacts. Target(s): `round1_physical_task_2562915314040832_status_requery_20260702.json`. Working directory: `/workspace`.

**[Step 592 | 2026-07-02T14:43:18.215Z | TOOL CALL: exec_command]**

Ran a shell command to inspect, generate or validate experiment artifacts. Target(s): `round1_physical_task_2562915314040832_result_requery_20260702.json`. Working directory: `/workspace`.

**[Step 593 | 2026-07-02T14:43:18.216Z | TOOL CALL: exec_command]**

Enumerated files, directories or available laboratory resources. Target(s): `task_2562915314040832_results`. Working directory: `/workspace`.

**[Step 594 | 2026-07-02T14:43:18.216Z | TOOL CALL: exec_command]**

Read selected file content for inspection. Target(s): `project_overview.md`. Working directory: `/workspace`.

**[Step 595 | 2026-07-02T14:43:18.284Z | TOOL RESULT: exec_command]**

Shell tool completed successfully; it returned 14 line(s) (297 bytes) to the model. Reported long numeric identifier(s): `2562915314040832`.

**[Step 596 | 2026-07-02T14:43:18.299Z | TOOL RESULT: exec_command]**

Shell tool completed successfully; it returned 5 line(s) (79 bytes) to the model.

**[Step 597 | 2026-07-02T14:43:18.299Z | TOOL RESULT: exec_command]**

Shell tool completed successfully; it returned 20 line(s) (3036 bytes) to the model. Reported long numeric identifier(s): `2562915314040832`, `20260702012108469`, `2563976197735424`, `2563742259381288`, `20260702013825483`, `2564012049400832`.

**[Step 598 | 2026-07-02T14:43:18.299Z | TOOL RESULT: exec_command]**

Shell tool completed successfully; it returned 53 line(s) (3815 bytes) to the model.

**[Step 599 | 2026-07-02T14:43:25.048Z | REASONING SUMMARY]**

***Deciding on image options***

*I need to check if I can view the spectra montage image, which the user described. It seems like I might need to use a tool for viewing images, but it looks like the user hasn't uploaded a file, so that's a bit tricky. I'll inspect the montage data to better understand what I'm working with. Using the view_image tool could help clarify things, but I'll wait and see.*

**[Step 600 | 2026-07-02T14:43:25.494Z | MODEL]**

I'm checking the returned plot montage visually once, mainly to confirm that the numerical summary matches the observed sparse-film failure. After that I'll write the round-two design and structured workflow files.

**[Step 601 | 2026-07-02T14:43:26.057Z | TOOL CALL: view_image]**

Opened a local experimental image for visual inspection. Image: `spectra_plot_montage.jpg`.

**[Step 602 | 2026-07-02T14:43:26.082Z | TOOL RESULT: view_image]**

Image service returned the requested local image to the model for visual inspection.

**[Step 603 | 2026-07-02T14:43:37.710Z | REASONING SUMMARY]**

***Considering reagent inventory***

*I need to design something and maybe check the reagent inventory for available concentrations. It seems there are some reagents not in stock. I could use the available stock as preconfigured options. However, the user specifically asked to use the inventory each time, so I need to confirm which chemicals are actually available. Perhaps I should also look for the reagent group in the inventory to ensure I have what I need.*

**[Step 604 | 2026-07-02T14:43:43.934Z | MODEL]**

The montage reinforces the same decision: XRD is dominated by broad substrate/background features, FTIR is noisy with sparse deposition artifacts, and UV-Vis only shows a notable response for vial 6 while the electrochemical loading failure prevents product-performance interpretation. I'm now moving to a process-recovery round centered around the most plausible composition neighborhood, not claiming a winner from round one.

**[Step 605 | 2026-07-02T14:43:43.935Z | TOOL CALL: exec_command]**

Searched files for relevant records, parameters or identifiers. Target(s): `reagent_inventory.csv`. Working directory: `/workspace`.

**[Step 606 | 2026-07-02T14:43:43.935Z | TOOL CALL: exec_command]**
Read selected file content for inspection. Target(s): `round1_revised3_chemical_specifications.json`. Working directory: `/workspace`.

**[Step 607 | 2026-07-02T14:43:43.936Z | TOOL CALL: exec_command]**
Read selected file content for inspection. Target(s): `round1_revised3_design.md`. Working directory: `/workspace`.

**[Step 608 | 2026-07-02T14:43:43.936Z | TOOL CALL: exec_command]**
Read selected file content for inspection. Target(s): `round1_revised3_dispatch_output.json`. Working directory: `/workspace`.

**[Step 609 | 2026-07-02T14:43:44.001Z | TOOL RESULT: exec_command]**
Shell tool completed successfully; it returned 57 line(s) (6643 bytes) to the model.

**[Step 610 | 2026-07-02T14:43:44.002Z | TOOL RESULT: exec_command]**
Shell tool completed successfully; it returned 41 line(s) (1148 bytes) to the model.

**[Step 611 | 2026-07-02T14:43:44.002Z | TOOL RESULT: exec_command]**
Shell tool completed successfully; it returned 55 line(s) (1775 bytes) to the model. Reported long numeric identifier(s): `2557818017710081`.

**[Step 612 | 2026-07-02T14:43:44.002Z | TOOL RESULT: exec_command]**
Shell tool completed successfully; it returned 8 line(s) (375 bytes) to the model. Reported task/template identifier(s): `2560075721736193`.

**[Step 613 | 2026-07-02T14:43:54.941Z | REASONING SUMMARY]**
***Considering audit files***

*I'm thinking about using audit files to get the relevant content, and it looks like I might be able to read several of them in parallel. I just need to ensure I get the exact contents I want. Using a command like `sed -n '1,220p'` for each of them could be helpful in extracting the specific lines I need. This sounds like a good plan to gather the essential data efficiently!*

**[Step 614 | 2026-07-02T14:44:00.814Z | MODEL]**
All proposed chemicals for the next round are present in the inventory: $NiSO_4 \cdot 6H_2O$, $FeCl_3 \cdot 6H_2O$, $CoSO_4 \cdot 7H_2O$, KOH, urea, carbon black, ethanol, and isopropanol. Deionized water remains a lab utility solvent, as in round one.

**[Step 615 | 2026-07-02T14:44:00.814Z | TOOL CALL: exec_command]**
Read selected file content for inspection. Target(s): `General_Material_Station_V1_audit.md`. Working directory: `/workspace`.

**[Step 616 | 2026-07-02T14:44:00.815Z | TOOL CALL: exec_command]**
Read selected file content for inspection. Target(s): `Liquid_Handling_Station_1ml_V2_audit.md`. Working directory: `/workspace`.

**[Step 617 | 2026-07-02T14:44:00.815Z | TOOL CALL: exec_command]**
Read selected file content for inspection. Target(s): `Liquid_Handling_Station_5ml_V1_audit.md`. Working directory: `/workspace`.

**[Step 618 | 2026-07-02T14:44:00.815Z | TOOL CALL: exec_command]**
Read selected file content for inspection. Target(s): `Heating_Magnetic_Stirring_Workstation_V1_audit.md`. Working directory: `/workspace`.

**[Step 619 | 2026-07-02T14:44:00.854Z | TOOL RESULT: exec_command]**
Shell tool completed successfully; it returned 20 line(s) (508 bytes) to the model. Reported long numeric identifier(s): `1427568512205824`.

**[Step 620 | 2026-07-02T14:44:00.863Z | TOOL RESULT: exec_command]**
Shell tool completed successfully; it returned 36 line(s) (987 bytes) to the model. Reported long numeric identifier(s): `2002186824385539`.

**[Step 621 | 2026-07-02T14:44:00.878Z | TOOL RESULT: exec_command]**
Shell tool completed successfully; it returned 16 line(s) (356 bytes) to the model. Reported long numeric identifier(s): `1435269745574912`.

**[Step 622 | 2026-07-02T14:44:00.878Z | TOOL RESULT: exec_command]**
Shell tool completed successfully; it returned 17 line(s) (401 bytes) to the model. Reported long numeric identifier(s): `2270587971011585`.

**[Step 623 | 2026-07-02T14:44:12.834Z | TOOL CALL: exec_command]**
Read selected file content for inspection. Target(s): `Purification_Workstation_V1_audit.md`. Working directory: `/workspace`.

**[Step 624 | 2026-07-02T14:44:12.849Z | TOOL CALL: exec_command]**
Read selected file content for inspection. Target(s): `Drying_Oven_V1_audit.md`. Working directory: `/workspace`.

**[Step 625 | 2026-07-02T14:44:12.849Z | TOOL CALL: exec_command]**
Read selected file content for inspection. Target(s): `Ultrasonic_Disperser_V1_audit.md`. Working directory: `/workspace`.

**[Step 626 | 2026-07-02T14:44:13.444Z | TOOL CALL: exec_command]**
Read selected file content for inspection. Target(s): `Spectroscopy_Container_Transfer_Station_V1_audit.md`. Working directory: `/workspace`.

**[Step 627 | 2026-07-02T14:44:13.506Z | TOOL RESULT: exec_command]**
Shell tool completed successfully; it returned 15 line(s) (405 bytes) to the model. Reported long numeric identifier(s): `1327807622448128`.

**[Step 628 | 2026-07-02T14:44:13.507Z | TOOL RESULT: exec_command]**
Shell tool completed successfully; it returned 16 line(s) (352 bytes) to the model. Reported long numeric identifier(s): `1213184425559040`.

**[Step 629 | 2026-07-02T14:44:13.507Z | TOOL RESULT: exec_command]**
Shell tool completed successfully; it returned 17 line(s) (389 bytes) to the model. Reported long numeric identifier(s): `1526864804185088`.

**[Step 630 | 2026-07-02T14:44:13.507Z | TOOL RESULT: exec_command]**
Shell tool completed successfully; it returned 16 line(s) (481 bytes) to the model. Reported long numeric identifier(s): `1654567694238720`.

**[Step 631 | 2026-07-02T14:44:26.632Z | TOOL CALL: exec_command]**
Read selected file content for inspection. Target(s): `Spectroscopy_Magnetic_Stirrer_Workstation_V1_audit.md`. Working directory: `/workspace`.

**[Step 632 | 2026-07-02T14:44:26.645Z | TOOL CALL: exec_command]**
Read selected file content for inspection. Target(s): `XRD_V1_audit.md`. Working directory: `/workspace`.

**[Step 633 | 2026-07-02T14:44:26.646Z | TOOL CALL: exec_command]**
Read selected file content for inspection. Target(s): `Infrared_Spectrometer_V1_audit.md`. Working directory: `/workspace`.

**[Step 634 | 2026-07-02T14:44:26.882Z | TOOL CALL: exec_command]**
Read selected file content for inspection. Target(s): `Dual_Station_Electrochemical_Workstation_V2_audit.md`. Working directory: `/workspace`.

**[Step 635 | 2026-07-02T14:44:26.960Z | TOOL RESULT: exec_command]**
Shell tool completed successfully; it returned 16 line(s) (382 bytes) to the model. Reported long numeric identifier(s): `1656241422599168`.

**[Step 636 | 2026-07-02T14:44:26.960Z | TOOL RESULT: exec_command]**
Shell tool completed successfully; it returned 26 line(s) (1443 bytes) to the model. Reported long numeric identifier(s): `1664870018155520`.

**[Step 637 | 2026-07-02T14:44:26.960Z | TOOL RESULT: exec_command]**
Shell tool completed successfully; it returned 21 line(s) (511 bytes) to the model. Reported long numeric identifier(s): `1656282895188992`.

**[Step 638 | 2026-07-02T14:44:26.961Z | TOOL RESULT: exec_command]**
Shell tool completed successfully; it returned 30 line(s) (831 bytes) to the model. Reported long numeric identifier(s): `1390252863751168`.

**[Step 639 | 2026-07-02T14:44:42.601Z | MODEL]**
The audit constraints are consistent with the planned sequence: capped synthesis, capped centrifuge purification, open-vial drying/redispersion, spectroscopy handoff, XRD/IR, electrochemistry, UV-Vis, then storage. I'll now write the design documents and workflow JSON with explicit stock identities, concentrations, solvent ratios, and retention-liquid definitions.

**[Step 640 | 2026-07-02T14:44:42.601Z | TOOL CALL: exec_command]**
Read selected file content for inspection. Target(s): `UV_Vis_Spectrometer_V1_audit.md`. Working directory: `/workspace`.

**[Step 641 | 2026-07-02T14:44:42.601Z | TOOL CALL: exec_command]**
Read selected file content for inspection. Target(s): `Container_storaging_Station_V1_audit.md`. Working directory: `/workspace`.

**[Step 642 | 2026-07-02T14:44:42.602Z | TOOL CALL: exec_command]**
Inspected artifact type, size or record counts. Target(s): `README.md`, `SKILL.md`. Working directory: `/workspace`.

**[Step 643 | 2026-07-02T14:44:42.602Z | TOOL CALL: exec_command]**
Enumerated files, directories or available laboratory resources. Target(s): `scripts`. Working directory: `/workspace`.

**[Step 644 | 2026-07-02T14:44:42.644Z | TOOL RESULT: exec_command]**
Shell tool completed successfully; it returned 18 line(s) (456 bytes) to the model. Reported long numeric identifier(s): `1664783667397632`.

**[Step 645 | 2026-07-02T14:44:42.655Z | TOOL RESULT: exec_command]**
Shell tool completed successfully; it returned 26 line(s) (702 bytes) to the model. Reported long numeric identifier(s): `1421325849887744`.

**[Step 646 | 2026-07-02T14:44:42.655Z | TOOL RESULT: exec_command]**
Shell tool completed successfully; it returned 6 line(s) (233 bytes) to the model.

**[Step 647 | 2026-07-02T14:44:42.661Z | TOOL RESULT: exec_command]**
Shell tool completed successfully; it returned 2 line(s) (59 bytes) to the model.

**[Step 648 | 2026-07-02T14:44:55.811Z | REASONING SUMMARY]**
***Completing the README review***

*I need to read the full lab-operation README since I only saw the first 260 out of 425 lines. Since we selected it, it's important to read it completely. I also haven't read through chemistry SKILL 169 yet. I should also look at the workflow with all 94 entries and make sure to read lab-operation SKILL 115 in full. Let's dive into reading from lines 261 to 500 to finish this up!*

**[Step 649 | 2026-07-02T14:44:55.899Z | MODEL]**
One supporting lab-operation README section is longer than the part already displayed, so I'm completing that read before dispatch. Then I'll generate the workflow template through the workflow-generator service.

**[Step 650 | 2026-07-02T14:44:56.224Z | TOOL CALL: exec_command]**
Read selected file content for inspection. Target(s): `README.md`. Working directory: `/workspace`.

**[Step 651 | 2026-07-02T14:44:56.270Z | TOOL RESULT: exec_command]**

Shell tool completed successfully; it returned 166 line(s) (5566 bytes) to the model.

**[Step 652 | 2026-07-02T14:48:28.775Z | REASONING SUMMARY 1/15]**

***Evaluating patch application strategy***

*I'm considering using `apply_patch` to add files, like JSON, while ensuring there are no mistakes. I wonder if the workflow generator can handle extra keys, since the previous request had parameters like "样品标签" and "电解液组成." It succeeded, so for Round 2, I need to generate the JSON request and design the exact variables. I'm thinking of a focus on high-solids recovery and using specific metal stocks with a final caustic solution.*

**[Step 653 | 2026-07-02T14:48:28.775Z | REASONING SUMMARY 2/15]**

***Clarifying volume specifications***

*I'm pondering the "single source total volume no more than 3mL" in Station 1 mL V2. I wonder if this means the total per reagent across the experiment or per vial. The description "单种溶液总加注量不超过 3mL" suggests it might refer to one solution per sample. In the past, I used 2.6 mL of DI water per vial from one bottle for six vials, totaling 15.6 mL. I'm curious if it means per target, as the handoff indicates per source per vial.*

**[Step 654 | 2026-07-02T14:48:28.775Z | REASONING SUMMARY 3/15]**

***Evaluating workflow volumes***

*I'm looking at the volumes needed for the station. For a 1 mL station, it seems like I have 3 mL total across six sources, which doesn't seem to enforce limits well. Following existing protocols, each vial should only have a source total of 3 mL. But I wonder if there's flexibility since carbon at 20 mg/mL could work across vials. If necessary, I could split the carbon into two bottles to manage the total better. This keeps everything balanced!*

**[Step 655 | 2026-07-02T14:48:28.775Z | REASONING SUMMARY 4/15]**

***Calculating source volumes***

*I'm trying to assess the maximum volume per source per vial, which is set at 0.72 mL. For six Ni sources, that could exceed the 3 mL limit if considered literally. I'm wondering if I could split the Ni into two sources. If I use 0.20 M stocks, it would total 1.0 mL — but that's similar. Instead, if I split Ni across two bottles, each could stay below 3 mL overall. Let's calculate the volumes for six samples more clearly!*

**[Step 656 | 2026-07-02T14:48:28.775Z | REASONING SUMMARY 5/15]**

***Evaluating carbon loading***

*I'm considering variations in the carbon process and how the vials might differ in carbon amount. I'm thinking the total volume before the reaction should remain constant at about 5mL, adjusting with water if needed. For sample adjustments, if sample 1 uses 0.5mL less carbon, I'll add extra DI water. I'm concerned about whether loading more than 10mg of carbon might be too high. However, high carbon can enhance conductivity, so maybe aiming for 20mg per vial with careful loading is wise. I'll have to keep an eye on thickness and potential cracking.*

**[Step 657 | 2026-07-02T14:48:28.775Z | REASONING SUMMARY 6/15]**

***Considering carbon delivery method***

*I'm looking at how to deliver 1.25mg of carbon using three drops (300uL). That still feels a bit high, especially since the station maxes out at 20, so should I aim for three? The summary suggested five drops, but that might be excessive. I need to be cautious about film formation, as too much could detach, yet hydrophilic carbon paper and heated drying could help. Given the user mentioned sparsity, perhaps three repeats at 100uL is a balanced approach. Additionally, I'll modify the quantities for XRD and FTIR to 200uL each.*

**[Step 658 | 2026-07-02T14:48:28.775Z | REASONING SUMMARY 7/15]**

***Assessing source totals***

*I'm tallying up the source counts: we have 13 total. Everything looks good, with all source totals under 3 across the vials, and each vial below 1. It's nice to be meticulous to avoid past issues. While the workflow generator may be okay with 13 source bottles, I'm contemplating whether to use 1ml for V2 or 5ml for water/KOH. Using 1ml seems okay, but I need to incorporate a redispersion step.*

**[Step 659 | 2026-07-02T14:48:28.775Z | REASONING SUMMARY 8/15]**

***Calculating metal volumes***

*I'm working through the volumes of different metal combinations. For Ni70Fe10Co20, I calculate 0.63 Ni, 0.09 Fe, and 0.18 Co. For Ni80Fe10Co10, it's 0.72 Ni, 0.09 Fe, and 0.09 Co. With Ni60Fe20Co20, I find 0.54 Ni, 0.18 Fe, and 0.18 Co. Altogether, that totals 0.9. Additionally, I'm calculating KOH concentrations with strong precipitation. The first round is 1.0 mL at 1.0M in 5 mL, resulting in a 0.2M concentration.*

**[Step 660 | 2026-07-02T14:48:28.775Z | REASONING SUMMARY 9/15]**

***Planning spectroscopy workflow***

*I'm considering the sequence of my workflow: starting with ultrasonic treatment, followed by electrochemistry. After that, I need to analyze the output sample using UV-Vis spectroscopy. The electrochemistry output involves a pure liquid with a volume of 20mL, so I need to remember to transfer it into a spectroscopy container before proceeding with the UV-Vis analysis. Overall, it seems like a solid plan to ensure everything goes smoothly!*

**[Step 661 | 2026-07-02T14:48:28.775Z | REASONING SUMMARY 10/15]**

***Analyzing product testing methods***

*I'm considering if product analysis using retained sample bottles for UV-Vis is appropriate, especially since it measures post-electrolyte in cells. The effectiveness of UV-Vis for urea is questionable, but it could help analyze changes in retained electrolytes. Liquid chromatography might be a better option for urea oxidation products like nitrate or nitrite, even though there's no established method yet. I also need to think about incorporating a blank control to ensure accurate results.*

**[Step 662 | 2026-07-02T14:48:28.775Z | REASONING SUMMARY 11/15]**

***Considering workstation processes***

*I’m analyzing the workstations and whether to include a blank control. If one is included, should I keep metal stocks at zero? Would adding a carbon-only sample make sense? In round two, focusing on processes seems crucial—perhaps using a low carbon sample instead of a total blank could serve as a control. I also need to ensure that the final delivery meets users’ expectations and update the goal status appropriately after dispatching.*

**[Step 663 | 2026-07-02T14:48:28.775Z | REASONING SUMMARY 12/15]**

***Clarifying goal status***

*I’m figuring out the goal status since it's already blocked. If I call the update_goal function to set it as blocked after dispatch, I want to ensure that it aligns with the requirements. There seems to be a conflict since I can only set the blocked status after it’s repeated three turns, but the user has explicitly requested it after dispatch. Before that, I need to create files and determine if using `update_plan` could be beneficial. A checklist might help, but I'm unsure if it’s necessary.*

**[Step 664 | 2026-07-02T14:48:28.775Z | REASONING SUMMARY 13/15]**

***Considering project updates***

*I need to create a checklist and update the statuses, which I realize I haven't done yet. I might need to use `apply_patch` to add files for the project build. The files to include could be various documents and specifications. I should consider using `mkdir -p` to create the project directory if it doesn’t exist. It's essential to craft the design JSON carefully to keep everything organized and functional!*

**[Step 665 | 2026-07-02T14:48:28.775Z | REASONING SUMMARY 14/15]**

***Organizing experimental steps***

*I need to lay out the steps for the experiment in a clear numbered format. The procedure involves preparing materials, including general substance and specific additions like metal and KOH mixed with deionized water. I’ll also detail the heating and stirring conditions, purification of the solid, and processes like drying and redispersion. Each part needs careful attention to timing and specific volumes for consistency, ensuring everything is thoroughly documented for electrochemical detection tasks. It’s all quite intricate!*

**[Step 666 | 2026-07-02T14:48:28.775Z | REASONING SUMMARY 15/15]**

***Clarifying product types***

*I noticed that I previously used "1", but the UOR products at the anode might actually be "2". This feels like an important distinction! It seems crucial to accurately represent the products of urea oxidation at the anode. I also wonder if my previous choice of "留液池类型 '1'" (cathode pool type) could be incorrect. I really need to double-check all these details to ensure the experiment is set up correctly!*

**[Step 667 | 2026-07-02T14:48:30.036Z | MODEL]**

I’m using `移液平台_5ml_V1` only where larger volumes are actually needed: synthesis make-up water and post-treatment redispersion. That keeps the 1 mL station within its intended role for concentrated precursor, KOH, and carbon dispersion additions.

**[Step 668 | 2026-07-02T14:48:35.078Z | TOOL CALL: exec_command]**

Searched files for relevant records, parameters or identifiers. Target(s): `skills`. Working directory: `/workspace`.

**[Step 669 | 2026-07-02T14:48:35.216Z | TOOL RESULT: exec_command]**

Shell tool completed successfully; it returned 55 line(s) (7933 bytes) to the model.

**[Step 670 | 2026-07-02T14:49:41.924Z | MODEL]**

I’m making the round-two files now. The design explicitly treats sample 4’s higher raw current as a weak lead only, because the reported electrode loading failure invalidates quantitative activity comparison.

**[Step 671 | 2026-07-02T14:51:24.514Z | TOOL CALL: apply_patch]**

Applied a source patch to add round2_physical_data_analysis.md; add round2_design.md; add round2_chemical_specifications.json; add round2_constraint_check.md.

**[Step 672 | 2026-07-02T14:51:24.617Z | TOOL RESULT: apply_patch]**

Patch tool applied the requested trial-local file change successfully.

**[Step 673 | 2026-07-02T14:51:37.801Z | REASONING SUMMARY]**

***Creating valid workflow requests***

*I need to add a workflow request, and it feels like a huge task. I might have to generate it manually, which sounds a bit daunting. I’ll have to ensure the JSON is valid, and it seems like a long process, but it’s definitely doable! I should be careful about using apply  patch since there could be issues with a single file — no trailing commas! Alright, let's start crafting this.*

**[Step 674 | 2026-07-02T14:51:38.099Z | MODEL]**

The analysis, design, chemical-specification, and constraint-check documents are in place. I’m adding the structured workflow request next, then I’ll validate it as JSON before sending it through the workflow generator.

**[Step 675 | 2026-07-02T14:54:04.505Z | TOOL CALL: apply_patch]**

Applied a source patch to add round2_workflow_request.json.

**[Step 676 | 2026-07-02T14:54:04.583Z | TOOL RESULT: apply_patch]**

Patch tool applied the requested trial-local file change successfully.

**[Step 677 | 2026-07-02T14:54:20.039Z | REASONING SUMMARY]**

***Validating JSON and workflows***

*I'm thinking I need to validate a JSON file and possibly run some custom checks using jq. I might also generate a workflow and need to confirm if the workflow generator accepts no extra keys. It seems like it does. I could use the command `python /root/.codex/skills/workflow-generator/scripts/generate.py "$(cat file)"` with command substitution, and maybe redirect the output to a file. But first, let's handle jq for inspecting source counts and volumes.*

**[Step 678 | 2026-07-02T14:54:23.393Z | MODEL]**

The workflow request file is written. I'm validating JSON syntax and doing a few mechanical checks before calling the workflow service.

**[Step 679 | 2026-07-02T14:54:23.394Z | TOOL CALL: exec_command]**

Ran a shell command to inspect, generate or validate experiment artifacts. Target(s): `round2_workflow_request.json`. Working directory: `/workspace`.

**[Step 680 | 2026-07-02T14:54:23.394Z | TOOL CALL: exec_command]**

Ran a shell command to inspect, generate or validate experiment artifacts. Target(s): `round2_workflow_request.json`. Working directory: `/workspace`.

**[Step 681 | 2026-07-02T14:54:23.397Z | TOOL CALL: exec_command]**

Ran a shell command to inspect, generate or validate experiment artifacts. Target(s): `round2_workflow_request.json`. Working directory: `/workspace`.

**[Step 682 | 2026-07-02T14:54:23.399Z | TOOL CALL: exec_command]**

Ran a shell command to inspect, generate or validate experiment artifacts. Target(s): `round2_chemical_specifications.json`. Working directory: `/workspace`.

**[Step 683 | 2026-07-02T14:54:23.478Z | TOOL RESULT: exec_command]**

Shell tool completed successfully; it returned 0 line(s) (0 bytes) to the model.

**[Step 684 | 2026-07-02T14:54:23.478Z | TOOL RESULT: exec_command]**

Shell tool completed successfully; it returned 2 line(s) (3 bytes) to the model.

**[Step 685 | 2026-07-02T14:54:23.499Z | TOOL RESULT: exec_command]**

Shell tool completed successfully; it returned 24 line(s) (1386 bytes) to the model. Reported long numeric identifier(s): `1427568512205824`, `2002186824385539`, `1435269745574912`, `2270587971011585`, `1327807622448128`, `1213184425559040`.

**[Step 686 | 2026-07-02T14:54:23.499Z | TOOL RESULT: exec_command]**

Shell tool completed successfully; it returned 6 line(s) (186 bytes) to the model.

**[Step 687 | 2026-07-02T14:54:39.708Z | MODEL]**

The workflow has 23 steps and passes JSON validation. I'm now submitting it to the workflow-generator service to create a Lab 303 template.

**[Step 688 | 2026-07-02T14:54:41.266Z | TOOL CALL: exec_command]**

Ran Python to parse, transform or validate structured experiment data. Target(s): `generate.py`, `round2_workflow_request.json)`, `round2_workflow_generator_output.txt`. Working directory: `/workspace`.

**[Step 689 | 2026-07-02T14:54:42.420Z | TOOL RESULT: exec_command]**

Shell tool completed successfully; it returned 0 line(s) (0 bytes) to the model.

**[Step 690 | 2026-07-02T14:54:55.728Z | MODEL]**

The generator call completed. I'm checking the returned template ID and will dispatch that template to 303 实验室 if generation succeeded.

**[Step 691 | 2026-07-02T14:54:55.730Z | TOOL CALL: exec_command]**

Read selected file content for inspection. Target(s): `round2_workflow_generator_output.txt`. Working directory: `/workspace`.

**[Step 692 | 2026-07-02T14:54:55.731Z | TOOL CALL: exec_command]**

Ran a shell command to inspect, generate or validate experiment artifacts. Target(s): `round2_workflow_generator_output.txt`. Working directory: `/workspace`.

**[Step 693 | 2026-07-02T14:54:55.731Z | TOOL CALL: exec_command]**

Enumerated files, directories or available laboratory resources. Target(s): `round2_`. Working directory: `/workspace`.

**[Step 694 | 2026-07-02T14:54:55.770Z | TOOL RESULT: exec_command]**

Shell tool completed successfully; it returned 13 line(s) (308 bytes) to the model. Reported task/template identifier(s): `2566549749628928`.

**[Step 695 | 2026-07-02T14:54:55.790Z | TOOL RESULT: exec_command]**

Shell tool completed with exit code 4; it returned 2 line(s) (58 bytes) to the model.

**[Step 696 | 2026-07-02T14:54:55.791Z | TOOL RESULT: exec_command]**

Shell tool completed successfully; it returned 7 line(s) (677 bytes) to the model.

**[Step 697 | 2026-07-02T14:55:09.191Z | MODEL]**

Workflow generation succeeded with template ID `2566549749628928`. I'm dispatching that template as a new task to 303 实验室 now and saving the dispatch output.

**[Step 698 | 2026-07-02T14:55:10.844Z | TOOL CALL: exec_command]**

Ran a shell command to generate and dispatch a laboratory task. Target(s): `generate_task.py`, `round2_dispatch_output.json`. Working directory: `/root/.codex/skills/lab-operation`.

**[Step 699 | 2026-07-02T14:55:11.021Z | TOOL RESULT: exec_command]**

Shell tool completed successfully; it returned 0 line(s) (0 bytes) to the model.

**[Step 700 | 2026-07-02T14:55:25.077Z | MODEL]**

The Lab 303 dispatch command completed. I’m reading the returned task ID and then I’ll mark the goal blocked for executability feedback, per the experimental loop.

**[Step 701 | 2026-07-02T14:55:25.078Z | TOOL CALL: exec_command]**

Read selected file content for inspection. Target(s): `round2_dispatch_output.json`. Working directory: `/workspace`.

**[Step 702 | 2026-07-02T14:55:25.079Z | TOOL CALL: exec_command]**

Ran a shell command to inspect, generate or validate experiment artifacts. Working directory: `/workspace`.

**[Step 703 | 2026-07-02T14:55:25.080Z | TOOL CALL: exec_command]**

Ran a shell command to query laboratory task records and statuses. Target(s): `select_task_list.py`. Working directory: `/root/.codex/skills/lab-operation`.

**[Step 704 | 2026-07-02T14:55:25.117Z | TOOL RESULT: exec_command]**

Shell tool completed successfully; it returned 8 line(s) (357 bytes) to the model. Reported task/template identifier(s): `2566550687481856`.

**[Step 705 | 2026-07-02T14:55:25.144Z | TOOL RESULT: exec_command]**

Shell tool completed successfully; it returned 2 line(s) (21 bytes) to the model.

**[Step 706 | 2026-07-02T14:55:25.283Z | TOOL RESULT: exec_command]**

Shell tool completed successfully; it returned 14 line(s) (292 bytes) to the model. Reported long numeric identifier(s): `2566550687481856`.

**[Step 707 | 2026-07-02T14:55:36.509Z | REASONING SUMMARY]**

***Creating dispatch log***

*I need to create a dispatch log file, probably starting with a template that includes task IDs. I'm considering using something like apply_patch for this purpose. It might be important to include a JSON log format as well. Additionally, I might need to handle the "update_goal blocked" situation, but logging everything first seems like the right approach. I'll make sure to keep it organized!*

**[Step 708 | 2026-07-02T14:55:37.036Z | MODEL]**

Round 2 is now dispatched to 303 实验室 as task `2566550687481856`; the platform shows it as `未开始`. I’m saving a compact dispatch log and then setting the goal back to blocked while we wait for executability feedback.

**[Step 709 | 2026-07-02T14:55:40.811Z | TOOL CALL: apply_patch]**

Applied a source patch to add round2_dispatch_log.json.

**[Step 710 | 2026-07-02T14:55:40.858Z | TOOL RESULT: apply_patch]**

Patch tool applied the requested trial-local file change successfully.

**[Step 711 | 2026-07-02T14:55:52.838Z | MODEL]**

The dispatch log is saved under `/workspace/lab303_uor_nicofe_closed_loop`. I’m marking the goal blocked now, because the next required input is whether Lab 303 can execute this dispatched Round 2 task.

**[Step 712 | 2026-07-02T14:55:52.859Z | TOOL CALL: update_goal]**

Updated the autonomous-research goal status to `blocked`.

**[Step 713 | 2026-07-02T14:55:52.882Z | TOOL RESULT: update_goal]**

Goal service returned a structured response. Status: `blocked`.

**[Step 714 | 2026-07-02T14:56:18.005Z | REASONING SUMMARY]**

***Organizing necessary information***

*I want to emphasize "Template ID" and "Task ID" clearly, maybe even using headers to make them stand out. I should also state, "Process files are under..." and include links to those files for easy access. It's important to keep the information structured and user-friendly. However, I don’t need to add any citations in the final output, as it seems unnecessary for this context. I’ll focus on clarity and accessibility!*

**[Step 715 | 2026-07-02T14:56:35.137Z | MODEL]**

**Round 2 Dispatched**

I queried Lab 303 task `2562915314040832` using the lab-operation skill. The fresh query returned `success: true`, with `5` workstations and `38` result files; the task status query returned `错误`. I saved the re-query outputs here:

-
[round1_physical_task_2562915314040832_result_requery_20260702.json](/workspace/lab303_uor_nicofe_closed_loop/round1_physical_task_2562915314040832_result_requery_20260702.json)

-
[round1_physical_task_2562915314040832_status_requery_20260702.json](/workspace/lab303_uor_nicofe_closed_loop/round1_physical_task_2562915314040832_status_requery_20260702.json)

The Round 1 conclusion is preparation-limited, not catalyst-performance-limited: heating worked, but solid recovery and deposition failed. The XRD/FTIR returned files are consistent with sparse/uneven films, and the electrochemical traces are not valid for activity ranking because the electrode surface physically had too little sample. Sample 4, `Ni70Fe10Co20`, is only treated as a weak composition lead because it had the only elevated raw current trace; I did not treat it as a real winner.

**Round 2 Design**

I designed Round 2 as a high-solids film-recovery experiment:

- 5x higher metal precursor concentration: `0.25 M` Ni/Fe/Co stocks instead of `50 mM`.
- Stronger precipitation stock: `2.0 M KOH`.
- Carbon black dispersion increased to `20 mg/mL` in `80:20 v/v deionized water:isopropanol`.
- One matched water wash only, rather than three washes.
- Stronger purification: `12000 rpm`, `20 min`, solid retention.
- XRD/FTIR drop-cast increased to `200 uL`.
- Electrochemical deposition increased to `100 uL x 3`.
- Added a second ultrasonication immediately before electrochemical testing.
- Product aliquot is now explicitly taken from the anolyte side: `留液池类型 = 2`.
- Electrochemical retained-product stock solution bottle No. 1 is explicitly `deionized water`, `10000 uL` per retained sample for dilution.

Key chemistry remains explicit:

- UOR electrolyte: `1.0 M KOH + 0.33 M urea in deionized water`, `20 mL` per electrochemical task.
- Redispersion solvent screen:
  - Vials 1, 2, 5, 6: `80:20 v/v deionized water:isopropanol`
  - Vial 3: `60:40 v/v deionized water:isopropanol`
  - Vial 4: `95:5 v/v deionized water:isopropanol`

Design and workflow files are saved here:

- [round2_physical_data_analysis.md](/workspace/lab303_uor_nicofe_closed_loop/round2_physical_data_analysis.md)
- [round2_design.md](/workspace/lab303_uor_nicofe_closed_loop/round2_design.md)
- [round2_chemical_specifications.json](/workspace/lab303_uor_nicofe_closed_loop/round2_chemical_specifications.json)
- [round2_constraint_check.md](/workspace/lab303_uor_nicofe_closed_loop/round2_constraint_check.md)
- [round2_workflow_request.json](/workspace/lab303_uor_nicofe_closed_loop/round2_workflow_request.json)

**Dispatch Record**

Workflow generation succeeded.

- Round 2 template ID: `2566549749628928`
- Round 2 Lab 303 task ID: `2566550687481856`
- Task name: `任务_Lab303_R2_highSolids_filmRecovery_NiFeCoOx_C_UOR_20260702`
- Platform status after dispatch: `未开始`

Dispatch artifacts:

- [round2_workflow_generator_output.txt](/workspace/lab303_uor_nicofe_closed_loop/round2_workflow_generator_output.txt)
- [round2_dispatch_output.json](/workspace/lab303_uor_nicofe_closed_loop/round2_dispatch_output.json)
- [round2_dispatch_log.json](/workspace/lab303_uor_nicofe_closed_loop/round2_dispatch_log.json)

I set the goal back to `blocked` after dispatch, as required by the experiment loop. The next input needed is Lab 303 executability feedback for task `2566550687481856`.

## Conversation round 9

**[Step 716 | 2026-07-08T12:45:27.069Z | USER]**

The second-round experiment has been completed. The main experiment task ID is 2567667771441152, and the supplemental experiment ID after interruption is 2568957291102208. Please use the skill to retrieve the experimental data and results, combine them with the process photos below, prepare the returned document, design the next round of experiments, and dispatch the new experiment to Laboratory 303 according to the workflow.

The original feedback PDF is available at:
/workspace/feedback/round2/Second_Round_Experiment_Feedback_EN.pdf

The PDF has been rendered into two page images for visual inspection:
/workspace/feedback/round2/pages/Second_Round_Experiment_Feedback_EN_page-1.png
/workspace/feedback/round2/pages/Second_Round_Experiment_Feedback_EN_page-2.png

**[Step 717 | 2026-07-08T12:45:44.930Z | REASONING SUMMARY 1/2]**

***Planning skill execution***

*I'm looking through skill instructions and realizing it's crucial to read SKILL.md completely before I act. It might also be good to check the lab-operation README for any important details. I know I need to retrieve results using get_task_result and possibly download result files. I should inspect the task directories for any feedback PDFs or images, and maybe I'll need to use OCR or `pdftotext` to extract text. This feels like a lot to manage!*

**[Step 718 | 2026-07-08T12:45:44.930Z | REASONING SUMMARY 2/2]**

***Evaluating research approach***

*I'm considering if I should rely on existing scientific literature for grounding since it might already be sufficient, without needing to search the web for updates. Although using the web could provide high-accuracy data, the user specifically asked for local lab data and design. It seems I can depend on the project overview we have. I think I should gather local files and utilize `multi_tool_use.parallel` for reading without needing any approval. This feels manageable!*

**[Step 719 | 2026-07-08T12:45:46.303Z | MODEL]**

I'll use the lab-operation, chemistry-workstation, and workflow-generator skills for this turn: first retrieve both Lab 303 task result sets, then inspect the feedback PDF/images and returned data, then design and dispatch Round 3.

**[Step 720 | 2026-07-08T12:45:46.304Z | TOOL CALL: exec_command]**

Read selected file content for inspection. Target(s): `SKILL.md`. Working directory: `/workspace`.

**[Step 721 | 2026-07-08T12:45:46.306Z | TOOL CALL: exec_command]**

Read selected file content for inspection. Target(s): `SKILL.md`. Working directory: `/workspace`.

**[Step 722 | 2026-07-08T12:45:46.306Z | TOOL CALL: exec_command]**

Read selected file content for inspection. Target(s): `SKILL.md`. Working directory: `/workspace`.

**[Step 723 | 2026-07-08T12:45:46.309Z | TOOL CALL: get_goal]**

Queried the current autonomous-research goal state and remaining budget.

**[Step 724 | 2026-07-08T12:45:46.389Z | TOOL RESULT: exec_command]**

Shell tool completed successfully; it returned 116 line(s) (4054 bytes) to the model.

**[Step 725 | 2026-07-08T12:45:46.390Z | TOOL RESULT: exec_command]**

Shell tool completed successfully; it returned 170 line(s) (8332 bytes) to the model. Reported long numeric identifier(s): `1427568512205824`, `1834679312417792`, `1710453843264512`.

**[Step 726 | 2026-07-08T12:45:46.390Z | TOOL RESULT: exec_command]**

Shell tool completed successfully; it returned 95 line(s) (2714 bytes) to the model. Reported long numeric identifier(s): `1427568512205824`.

**[Step 727 | 2026-07-08T12:45:46.395Z | TOOL RESULT: get_goal]**

Goal service returned a structured response. Status: `blocked`.

**[Step 728 | 2026-07-08T12:45:55.114Z | REASONING SUMMARY]**

***Updating my plan***

*I need to continue with what I'm doing. It seems like there's a possibility that I might need to update my plan too. I should use the current plan as a foundation but think about any adjustments that might be necessary to keep things on track. Let's make sure I'm being thorough and not missing any important details while I'm at it. It's all about refining what I have to ensure success!*

**[Step 729 | 2026-07-08T12:45:55.207Z | MODEL]**

The active goal is still blocked from the previous dispatch, and your completion report unblocks the next iteration. I'm going to make a Round 2 result package under the existing project directory, including both the main and supplemental task result sets.

**[Step 730 | 2026-07-08T12:45:55.723Z | TOOL CALL: update_plan]**

Updated the working plan. Plan state: in_progress: Retrieve Lab 303 task results for both Round 2 task IDs; pending: Inspect feedback PDF/images and returned result files; pending: Analyze Round 2 data and write returned document; pending: Design and validate Round 3 workflow; pending: Generate template, dispatch to 303 实验室, and block goal

**[Step 731 | 2026-07-08T12:45:55.769Z | TOOL RESULT: update_plan]**
Plan service accepted and recorded the updated working plan.

**[Step 732 | 2026-07-08T12:46:04.051Z | REASONING SUMMARY]**
***Searching for download scripts***

*I need to run some lab-operation scripts to get the task results and status, and I'm thinking I might need to download files. I'm not sure if there's an existing download script available, so I'll search for that. Checking for existing resources could save me time and help me avoid duplicating effort. Let's find out what's out there!*

**[Step 733 | 2026-07-08T12:46:05.048Z | MODEL]**
I'm checking the existing project scripts/files now so I can reuse the same result-download and summary style used for Round 1 instead of inventing a new data layout.

**[Step 734 | 2026-07-08T12:46:05.049Z | TOOL CALL: exec_command]**
Enumerated files, directories or available laboratory resources. Target(s): `lab303_uor_nicofe_closed_loop`. Working directory: `/workspace`.

**[Step 735 | 2026-07-08T12:46:05.050Z | TOOL CALL: exec_command]**
Enumerated files, directories or available laboratory resources. Target(s): `round2`. Working directory: `/workspace`.

**[Step 736 | 2026-07-08T12:46:05.052Z | TOOL CALL: exec_command]**
Enumerated files, directories or available laboratory resources. Target(s): `scripts`. Working directory: `/workspace`.

**[Step 737 | 2026-07-08T12:46:05.053Z | TOOL CALL: exec_command]**
Ran a shell command to identify the active laboratory. Target(s): `get_current_lab_label.py`. Working directory: `/root/.codex/skills/lab-operation`.

**[Step 738 | 2026-07-08T12:46:05.097Z | TOOL RESULT: exec_command]**
Shell tool completed successfully; it returned 50 line(s) (3781 bytes) to the model. Reported long numeric identifier(s): `2562915314040832`.

**[Step 739 | 2026-07-08T12:46:05.097Z | TOOL RESULT: exec_command]**
Shell tool completed successfully; it returned 4 line(s) (227 bytes) to the model.

**[Step 740 | 2026-07-08T12:46:05.098Z | TOOL RESULT: exec_command]**
Shell tool completed successfully; it returned 13 line(s) (743 bytes) to the model.

**[Step 741 | 2026-07-08T12:46:05.193Z | TOOL RESULT: exec_command]**
Shell tool completed successfully; it returned 6 line(s) (97 bytes) to the model.

**[Step 742 | 2026-07-08T12:46:20.020Z | MODEL]**
There is no dedicated downloader in the lab-operation skill, so I'll use the official result query to get signed/file-service URLs, then download those result files into `/workspace/lab303_uor_nicofe_closed_loop` for analysis.

**[Step 743 | 2026-07-08T12:46:20.021Z | TOOL CALL: exec_command]**
Ran a shell command to retrieve returned physical-experiment results. Target(s): `get_task_result.py`, `round2_main_task_2567667771441152_result.json`. Working directory: `/root/.codex/skills/lab-operation`.

**[Step 744 | 2026-07-08T12:46:20.021Z | TOOL CALL: exec_command]**
Ran a shell command to query laboratory task records and statuses. Target(s): `select_task_list.py`, `round2_main_task_2567667771441152_status.json`. Working directory: `/root/.codex/skills/lab-operation`.

**[Step 745 | 2026-07-08T12:46:20.021Z | TOOL CALL: exec_command]**
Ran a shell command to retrieve returned physical-experiment results. Target(s): `get_task_result.py`, `round2_supplemental_task_2568957291102208_result.json`. Working directory: `/root/.codex/skills/lab-operation`.

**[Step 746 | 2026-07-08T12:46:20.022Z | TOOL CALL: exec_command]**
Ran a shell command to query laboratory task records and statuses. Target(s): `select_task_list.py`, `round2_supplemental_task_2568957291102208_status.json`. Working directory: `/root/.codex/skills/lab-operation`.

**[Step 747 | 2026-07-08T12:46:20.468Z | TOOL RESULT: exec_command]**
Shell tool completed successfully; it returned 0 line(s) (0 bytes) to the model.

**[Step 748 | 2026-07-08T12:46:20.468Z | TOOL RESULT: exec_command]**
Shell tool completed successfully; it returned 0 line(s) (0 bytes) to the model.

**[Step 749 | 2026-07-08T12:46:20.629Z | TOOL RESULT: exec_command]**
Shell tool completed successfully; it returned 0 line(s) (0 bytes) to the model.

**[Step 750 | 2026-07-08T12:46:20.629Z | TOOL RESULT: exec_command]**
Shell tool completed successfully; it returned 0 line(s) (0 bytes) to the model.

**[Step 751 | 2026-07-08T12:46:30.320Z | MODEL]**
Both Lab 303 task result queries have returned and are saved. I'm checking task statuses and the result manifests now, then I'll download the returned files.

**[Step 752 | 2026-07-08T12:46:30.320Z | TOOL CALL: exec_command]**

Ran a shell command to inspect, generate or validate experiment artifacts. Target(s): `round2_main_task_2567667771441152_status.json`. Working directory: `/workspace`.

**[Step 753 | 2026-07-08T12:46:30.321Z | TOOL CALL: exec_command]**

Ran a shell command to inspect, generate or validate experiment artifacts. Target(s): `round2_supplemental_task_2568957291102208_status.json`. Working directory: `/workspace`.

**[Step 754 | 2026-07-08T12:46:30.321Z | TOOL CALL: exec_command]**

Ran a shell command to inspect, generate or validate experiment artifacts. Target(s): `round2_main_task_2567667771441152_result.json`. Working directory: `/workspace`.

**[Step 755 | 2026-07-08T12:46:30.321Z | TOOL CALL: exec_command]**

Ran a shell command to inspect, generate or validate experiment artifacts. Target(s): `round2_supplemental_task_2568957291102208_result.json`. Working directory: `/workspace`.

**[Step 756 | 2026-07-08T12:46:30.378Z | TOOL RESULT: exec_command]**

Shell tool completed successfully; it returned 14 line(s) (287 bytes) to the model. Reported long numeric identifier(s): `2567667771441152`.

**[Step 757 | 2026-07-08T12:46:30.395Z | TOOL RESULT: exec_command]**

Shell tool completed successfully; it returned 14 line(s) (287 bytes) to the model. Reported long numeric identifier(s): `2568957291102208`.

**[Step 758 | 2026-07-08T12:46:30.401Z | TOOL RESULT: exec_command]**

Shell tool completed successfully; it returned 12 line(s) (243 bytes) to the model.

**[Step 759 | 2026-07-08T12:46:30.401Z | TOOL RESULT: exec_command]**

Shell tool completed successfully; it returned 27 line(s) (574 bytes) to the model.

**[Step 760 | 2026-07-08T12:47:42.963Z | MODEL]**

The main Round 2 task only returned the heating log before interruption; the supplemental task returned XRD, FTIR, electrochemistry, and UV-Vis files. I'm downloading both sets into separate result directories and preserving a manifest for provenance.

**[Step 761 | 2026-07-08T12:47:47.274Z | TOOL CALL: exec_command]**

Ran Python to parse, transform or validate structured experiment data. Target(s): `lab303_uor_nicofe_closed_loop`, `1.0`. Working directory: `/workspace`.

**[Step 762 | 2026-07-08T12:47:49.001Z | TOOL RESULT: exec_command]**

Shell tool completed successfully; it returned 3 line(s) (173 bytes) to the model. Reported long numeric identifier(s): `2567667771441152`, `2568957291102208`.

**[Step 763 | 2026-07-08T12:47:57.956Z | MODEL]**

The downloads are complete: 1 heating file from the main task and 42 files from the supplemental task. I'm now extracting the PDF text and visually inspecting the two process-photo pages.

**[Step 764 | 2026-07-08T12:47:57.956Z | TOOL CALL: exec_command]**

Read selected file content for inspection. Target(s): `Second_Round_Experiment_Feedback_EN.pdf`, `round2_feedback_pdf_text.txt`. Working directory: `/workspace`.

**[Step 765 | 2026-07-08T12:47:57.957Z | TOOL CALL: exec_command]**

Enumerated files, directories or available laboratory resources. Target(s): `task_2567667771441152_results`, `task_2568957291102208_results`. Working directory: `/workspace`.

**[Step 766 | 2026-07-08T12:47:57.957Z | TOOL CALL: exec_command]**

Ran a shell command to inspect, generate or validate experiment artifacts. Target(s): `manifest.json`. Working directory: `/workspace`.

**[Step 767 | 2026-07-08T12:47:57.957Z | TOOL CALL: exec_command]**

Inspected artifact type, size or record counts. Target(s): `Second_Round_Experiment_Feedback_EN_page-1.png`, `Second_Round_Experiment_Feedback_EN_page-2.png`, `Second_Round_Experiment_Feedback_EN.pdf`. Working directory: `/workspace`.

**[Step 768 | 2026-07-08T12:47:57.994Z | TOOL RESULT: exec_command]**

Shell tool completed with exit code 127; it returned 2 line(s) (48 bytes) to the model.

**[Step 769 | 2026-07-08T12:47:58.006Z | TOOL RESULT: exec_command]**

Shell tool completed successfully; it returned 46 line(s) (7190 bytes) to the model. Reported long numeric identifier(s): `2567667771441152`, `20260703135450947`, `2568957291102208`, `20260703205149234`, `2569108904182784`, `2568984151524392`.

**[Step 770 | 2026-07-08T12:47:58.025Z | TOOL RESULT: exec_command]**

Shell tool completed successfully; it returned 297 line(s) (14383 bytes) to the model. Reported long numeric identifier(s): `2568957291102208`, `20260703205149234`, `2569108904182784`, `2568984151524392`, `20260703205152504`, `20260703210859013`.

**[Step 771 | 2026-07-08T12:47:58.025Z | TOOL RESULT: exec_command]**

Shell tool completed successfully; it returned 4 line(s) (391 bytes) to the model.

**[Step 772 | 2026-07-08T12:48:06.850Z | MODEL]**

`pdftotext` is not installed, so I'll use the rendered page images directly for the process-photo interpretation. The downloaded supplemental result file naming also shows six XRD files, six FTIR files, six electrochemical result files, and six UV-Vis spectra.

**[Step 773 | 2026-07-08T12:48:07.297Z | TOOL CALL: view_image]**

Opened a local experimental image for visual inspection. Image: `Second_Round_Experiment_Feedback_EN_page-1.png`.

**[Step 774 | 2026-07-08T12:48:07.389Z | TOOL RESULT: view_image]**

Image service returned the requested local image to the model for visual inspection.

**[Step 775 | 2026-07-08T12:48:20.421Z | TOOL CALL: view_image]**

Opened a local experimental image for visual inspection. Image: `Second_Round_Experiment_Feedback_EN_page-2.png`.

**[Step 776 | 2026-07-08T12:48:20.538Z | TOOL RESULT: view_image]**

Image service returned the requested local image to the model for visual inspection.

**[Step 777 | 2026-07-08T12:48:32.666Z | REASONING SUMMARY]**

***Analyzing images and data***

*I need to process the XRD and FTIR images; they show black dense rings, which is better than round 1, I think. The electrode has dark spots that are non-uniform, and it's possible that high carbon content could block results. Next, I need to analyze the data further. This involves parsing the result file structure and inspecting the JSON schema. I'll definitely use head and jq keys to help with that!*

**[Step 778 | 2026-07-08T12:48:34.854Z | MODEL]**

The process photos show Round 2 fixed the "almost no sample" failure: XRD/FTIR carriers have visible dark deposited material and the carbon papers have visible catalyst spots. The film is still nonuniform/ring-like, so the next round should improve film homogeneity and catalyst loading reproducibility rather than simply increasing solids further.

**[Step 779 | 2026-07-08T12:48:34.855Z | TOOL CALL: exec_command]**

Ran a shell command to inspect, generate or validate experiment artifacts. Target(s): `303`. Working directory: `/workspace`.

**[Step 780 | 2026-07-08T12:48:34.857Z | TOOL CALL: exec_command]**

Ran a shell command to inspect, generate or validate experiment artifacts. Target(s): `303`. Working directory: `/workspace`.

**[Step 781 | 2026-07-08T12:48:34.861Z | TOOL CALL: exec_command]**

Ran a shell command to inspect, generate or validate experiment artifacts. Target(s): `task_2568957291102208_results`. Working directory: `/workspace`.

**[Step 782 | 2026-07-08T12:48:34.861Z | TOOL CALL: exec_command]**

Ran a shell command to inspect, generate or validate experiment artifacts. Target(s): `task_2568957291102208_results`. Working directory: `/workspace`.

**[Step 783 | 2026-07-08T12:48:34.922Z | TOOL RESULT: exec_command]**

Shell tool completed successfully; it returned 5 line(s) (36 bytes) to the model.

**[Step 784 | 2026-07-08T12:48:34.944Z | TOOL RESULT: exec_command]**

Shell tool completed successfully; it returned 5 line(s) (36 bytes) to the model.

**[Step 785 | 2026-07-08T12:48:35.048Z | TOOL RESULT: exec_command]**

Shell tool completed successfully; it returned 10 line(s) (137 bytes) to the model.

**[Step 786 | 2026-07-08T12:48:35.048Z | TOOL RESULT: exec_command]**

Shell tool completed successfully; it returned 5 line(s) (36 bytes) to the model.

**[Step 787 | 2026-07-08T12:48:49.503Z | MODEL]**

I'm reading the returned JSON schemas now so I can compute comparable summaries across the six samples: XRD peak/background features, FTIR quality, UV-Vis absorbance ranges, and electrochemical current metrics.

**[Step 788 | 2026-07-08T12:48:49.507Z | TOOL CALL: exec_command]**

Ran a shell command to inspect, generate or validate experiment artifacts. Target(s): `303`. Working directory: `/workspace`.

**[Step 789 | 2026-07-08T12:48:49.510Z | TOOL CALL: exec_command]**

Ran a shell command to inspect, generate or validate experiment artifacts. Target(s): `303`. Working directory: `/workspace`.

**[Step 790 | 2026-07-08T12:48:49.512Z | TOOL CALL: exec_command]**

Ran a shell command to inspect, generate or validate experiment artifacts. Target(s): `task_2568957291102208_results`. Working directory: `/workspace`.

**[Step 791 | 2026-07-08T12:48:49.513Z | TOOL CALL: exec_command]**

Ran a shell command to inspect, generate or validate experiment artifacts. Target(s): `task_2568957291102208_results`. Working directory: `/workspace`.

**[Step 792 | 2026-07-08T12:48:49.572Z | TOOL RESULT: exec_command]**

Shell tool completed successfully; it returned 34 line(s) (398 bytes) to the model. Reported long numeric identifier(s): `2569108904182784`, `2568984151524392`.

**[Step 793 | 2026-07-08T12:48:49.594Z | TOOL RESULT: exec_command]**

Shell tool completed successfully; it returned 34 line(s) (437 bytes) to the model. Reported long numeric identifier(s): `2569481426469888`, `2568984103650316`.

**[Step 794 | 2026-07-08T12:48:49.594Z | TOOL RESULT: exec_command]**

Shell tool completed successfully; it returned 34 line(s) (422 bytes) to the model. Reported long numeric identifier(s): `2571986881807360`, `6829500000000001`, `5719499999999997`.

**[Step 795 | 2026-07-08T12:48:49.656Z | TOOL RESULT: exec_command]**

Shell tool completed successfully; it returned 5 line(s) (24 bytes) to the model.

**[Step 796 | 2026-07-08T12:49:06.959Z | TOOL CALL: exec_command]**

Ran a shell command to inspect, generate or validate experiment artifacts. Target(s): `task_2568957291102208_results`. Working directory: `/workspace`.

**[Step 797 | 2026-07-08T12:49:07.133Z | TOOL RESULT: exec_command]**

Shell tool completed with exit code 5; it returned 2 line(s) (217 bytes) to the model. Reported long numeric identifier(s): `2568957291102208`, `20260706183753334`.

**[Step 798 | 2026-07-08T12:49:17.603Z | TOOL CALL: exec_command]**

Ran a shell command to inspect, generate or validate experiment artifacts. Target(s): `task_2568957291102208_results`. Working directory: `/workspace`.

**[Step 799 | 2026-07-08T12:49:17.749Z | TOOL RESULT: exec_command]**

Shell tool completed with exit code 5; it returned 2 line(s) (208 bytes) to the model. Reported long numeric identifier(s): `2568957291102208`, `20260706183753334`.

**[Step 800 | 2026-07-08T12:49:30.132Z | MODEL]**

The spectral JSON format is the same as Round 1. The electrochemical files are stored as chart arrays, so I'm drilling into that nested format before summarizing currents.

**[Step 801 | 2026-07-08T12:49:30.140Z | TOOL CALL: exec_command]**

Ran a shell command to inspect, generate or validate experiment artifacts. Target(s): `task_2568957291102208_results`. Working directory: `/workspace`.

**[Step 802 | 2026-07-08T12:49:30.141Z | TOOL CALL: exec_command]**

Ran a shell command to inspect, generate or validate experiment artifacts. Target(s): `task_2568957291102208_results`. Working directory: `/workspace`.

**[Step 803 | 2026-07-08T12:49:30.142Z | TOOL CALL: exec_command]**

Ran a shell command to inspect, generate or validate experiment artifacts. Target(s): `task_2568957291102208_results`. Working directory: `/workspace`.

**[Step 804 | 2026-07-08T12:49:30.142Z | TOOL CALL: exec_command]**

Ran a shell command to inspect, generate or validate experiment artifacts. Target(s): `task_2568957291102208_results`. Working directory: `/workspace`.

**[Step 805 | 2026-07-08T12:49:30.315Z | TOOL RESULT: exec_command]**

Shell tool completed with exit code 5; it returned 3 line(s) (212 bytes) to the model. Reported long numeric identifier(s): `2568957291102208`, `20260706183753334`.

**[Step 806 | 2026-07-08T12:49:30.325Z | TOOL RESULT: exec_command]**

Shell tool completed with exit code 5; it returned 2 line(s) (210 bytes) to the model. Reported long numeric identifier(s): `2568957291102208`, `20260706183753334`.

**[Step 807 | 2026-07-08T12:49:30.352Z | TOOL RESULT: exec_command]**

Shell tool completed with exit code 5; it returned 2 line(s) (210 bytes) to the model. Reported long numeric identifier(s): `2568957291102208`, `20260706183753334`.

**[Step 808 | 2026-07-08T12:49:30.353Z | TOOL RESULT: exec_command]**

Shell tool completed with exit code 5; it returned 2 line(s) (210 bytes) to the model. Reported long numeric identifier(s): `2568957291102208`, `20260706183753334`.

**[Step 809 | 2026-07-08T12:49:43.042Z | MODEL]**

The electrochemical schema has one more nested object layer than the previous task. I'm switching to a small parser so I can traverse keys and extract numeric arrays without relying on a fixed JSON path.

**[Step 810 | 2026-07-08T12:49:45.602Z | TOOL CALL: exec_command]**

Ran Python to parse, transform or validate structured experiment data. Target(s): `task_2568957291102208_results`. Working directory: `/workspace`.

**[Step 811 | 2026-07-08T12:49:45.729Z | TOOL RESULT: exec_command]**

Shell tool completed successfully; it returned 7 line(s) (255 bytes) to the model.

**[Step 812 | 2026-07-08T12:49:56.422Z | MODEL]**

The electrochemical file contains one result object whose inner `result` dict holds the chart data. I'll parse those inner charts across all six samples and build a Round 2 summary table.

**[Step 813 | 2026-07-08T12:49:59.476Z | TOOL CALL: exec_command]**

Ran Python to parse, transform or validate structured experiment data. Target(s): `task_2568957291102208_results`. Working directory: `/workspace`.

**[Step 814 | 2026-07-08T12:49:59.650Z | TOOL RESULT: exec_command]**

Shell tool completed successfully; it returned 24 line(s) (40110 bytes) to the model. Reported long numeric identifier(s): `2568978671207441`, `406956907056608`.

**[Step 815 | 2026-07-08T12:50:22.103Z | TOOL CALL: exec_command]**

Read selected file content for inspection. Target(s): `task_2567667771441152_results`. Working directory: `/workspace`.

**[Step 816 | 2026-07-08T12:50:22.162Z | TOOL CALL: exec_command]**

Ran Python to parse, transform or validate structured experiment data. Target(s): `task_2568957291102208_results`. Working directory: `/workspace`.

**[Step 817 | 2026-07-08T12:50:22.208Z | TOOL CALL: exec_command]**

Ran Python to parse, transform or validate structured experiment data. Target(s): `303`. Working directory: `/workspace`.

**[Step 818 | 2026-07-08T12:50:23.058Z | TOOL CALL: exec_command]**

Ran Python to parse, transform or validate structured experiment data. Target(s): `task_2568957291102208_results`. Working directory: `/workspace`.

**[Step 819 | 2026-07-08T12:50:23.112Z | TOOL RESULT: exec_command]**

Shell tool completed successfully; it returned 21 line(s) (283 bytes) to the model.

**[Step 820 | 2026-07-08T12:50:23.113Z | TOOL RESULT: exec_command]**

Shell tool completed successfully; it returned 7 line(s) (1091 bytes) to the model. Reported long numeric identifier(s): `20260706183753334`, `406956907056608`, `20260706183754732`, `3543680070898416`, `20260706183756040`, `2544366899302094`.

**[Step 821 | 2026-07-08T12:50:23.113Z | TOOL RESULT: exec_command]**

Shell tool completed successfully; it returned 7 line(s) (1061 bytes) to the model. Reported long numeric identifier(s): `20260703205149234`, `2569108904182784`, `2568984151524392`, `20260703210859013`, `2569144616223744`, `2568984151524393`.

**[Step 822 | 2026-07-08T12:50:23.113Z | TOOL RESULT: exec_command]**

Shell tool completed successfully; it returned 7 line(s) (1210 bytes) to the model. Reported long numeric identifier(s): `20260704210421515`, `2571986881807360`, `6829500000000001`, `5719499999999997`, `20260704210932275`, `20260704211446288`.

**[Step 823 | 2026-07-08T12:50:43.764Z | REASONING SUMMARY 1/2]**

***Parsing EC metrics***

*I need to parse the EC metrics and summarize them, likely using Python. I might need to check if scipy is installed, but perhaps I can use numpy instead. It seems I can even fall back on the standard library if necessary. For analysis, I should compute medians and possibly plot montages using matplotlib if it's available.*

*I need to generate a summary in CSV or JSON format, and I'll carefully handle filenames. I also want to compute different statistics from the electrochemical measurements in A, which appear higher than expected. I should map the samples correctly as well.*

**[Step 824 | 2026-07-08T12:50:43.764Z | REASONING SUMMARY 2/2]**

***Deciding next design***

*I can extract the sample number from filenames like "result-(\d)." Now I need to figure out the next scientific design based on the data I have.*

*The summary of the potential data shows sample 1 has the best stable final value of 2.316 mA, while sample 5 and 6 are also decent. It seems high carbon content may not be beneficial, likely due to inhomogeneous films affecting mass transfer. Low carbon Ni70Fe10Co20 80:20 appears to perform best.*

*For the XRD results, I should compute the top peaks, noting the broad hump around 26-27 degrees for carbon.*

**[Step 825 | 2026-07-08T12:50:44.055Z | MODEL]**

The electrochemical traces are now real milliamps-scale CA traces, not the microamp/failed-loading traces from Round 1. Early inspection shows sample 1 finishes highest and most stably, while several high-carbon samples start high but decay strongly, pointing to a film/mass-transfer issue rather than just insufficient loading.

**[Step 826 | 2026-07-08T12:51:08.184Z | TOOL CALL: exec_command]**

Ran Python to parse, transform or validate structured experiment data. Target(s): `lab303_uor_nicofe_closed_loop`, `100`, `filename`, `LDH`, `run`. Working directory: `/workspace`.

**[Step 827 | 2026-07-08T12:51:09.312Z | TOOL RESULT: exec_command]**

Shell tool completed successfully; it returned 11 line(s) (1052 bytes) to the model. Reported long numeric identifier(s): `002302622244176346`, `0909090909090908`, `001692327865016622`, `7678417884780739`, `0016840609920965625`, `4586507072905332`.

**[Step 828 | 2026-07-08T12:51:21.628Z | REASONING SUMMARY]**

***Considering plot options***

*I need to create some plots or a montage. My first thought is to use matplotlib for this task, but I should check to confirm it's the right choice. I wonder if there might be other libraries that could be helpful. It's good to evaluate all options before jumping in. I'll take a moment to explore whether matplotlib will fit my needs or if I should consider alternatives. Let's see what I can find!*

**[Step 829 | 2026-07-08T12:51:21.843Z | MODEL]**

I wrote combined Round 2 summary tables. The final-current ranking is sample 1 > sample 5 > sample 6 > sample 2 > sample 4 > sample 3; sample 1 is the only trace with final current higher than its initial value, so it is the cleanest lead for Round 3.

**[Step 830 | 2026-07-08T12:51:29.435Z | TOOL CALL: exec_command]**

Ran Python analysis to generate or inspect a scientific plot. Target(s): `lab303_uor_nicofe_closed_loop`. Working directory: `/workspace`.

**[Step 831 | 2026-07-08T12:51:29.514Z | TOOL RESULT: exec_command]**

Shell tool completed successfully; it returned 2 line(s) (75 bytes) to the model.

**[Step 832 | 2026-07-08T12:51:45.714Z | TOOL CALL: exec_command]**

Ran a shell command to inspect, generate or validate experiment artifacts. Working directory: `/workspace`.

**[Step 833 | 2026-07-08T12:51:45.760Z | TOOL RESULT: exec_command]**

Shell tool completed successfully; it returned 2 line(s) (14 bytes) to the model.

**[Step 834 | 2026-07-08T12:52:11.483Z | REASONING SUMMARY]**

***Planning image montages***

*I'm thinking about creating a montage of returned JPG images using PIL, which should be possible. I might need to add labels as well. For the XRD, FTIR, and UV images, I'll prepare a montage, but I'm not sure about the electrochemical data since there are no JPGs. I could plot electrochemical lines manually with PIL. It seems like I should also compute a summary and potentially parse values from a CSV file.*

**[Step 835 | 2026-07-08T12:52:11.687Z | MODEL]**

`matplotlib` is not available in this environment, so I'm using the returned instrument JPGs plus a lightweight PIL-rendered electrochemical plot montage. That gives us an audit trail without adding dependencies.

**[Step 836 | 2026-07-08T12:52:23.470Z | TOOL CALL: exec_command]**

Ran Python analysis to generate or inspect a scientific plot. Target(s): `lab303_uor_nicofe_closed_loop`, `DejaVuSans.ttf`. Working directory: `/workspace`.

**[Step 837 | 2026-07-08T12:52:23.892Z | TOOL RESULT: exec_command]**

Shell tool completed successfully; it returned 2 line(s) (74 bytes) to the model.

**[Step 838 | 2026-07-08T12:52:46.361Z | MODEL]**

The Round 2 data montage is saved. I'll inspect it once to catch obvious file-order or plotting mistakes before using it in the returned document.

**[Step 839 | 2026-07-08T12:52:47.358Z | TOOL CALL: view_image]**

Opened a local experimental image for visual inspection. Image: `round2_returned_data_montage.jpg`.

**[Step 840 | 2026-07-08T12:52:47.468Z | TOOL RESULT: view_image]**

Image service returned the requested local image to the model for visual inspection.

**[Step 841 | 2026-07-08T12:53:04.870Z | TOOL CALL: exec_command]**

Read selected file content for inspection. Target(s): `round2_combined_data_summary.csv`. Working directory: `/workspace`.

**[Step 842 | 2026-07-08T12:53:04.925Z | TOOL RESULT: exec_command]**

Shell tool completed successfully; it returned 2 line(s) (45 bytes) to the model.

**[Step 843 | 2026-07-08T12:53:18.071Z | MODEL]**

The montage looks consistent: sample 1 has the most stable CA trace, while samples 2 and 4 decay strongly despite higher initial currents. I'm pulling a compact numeric table now for the returned document.

**[Step 844 | 2026-07-08T12:53:21.104Z | TOOL CALL: exec_command]**

Ran Python to parse, transform or validate structured experiment data. Target(s): `round2_combined_data_summary.csv`. Working directory: `/workspace`.

**[Step 845 | 2026-07-08T12:53:21.152Z | TOOL RESULT: exec_command]**

Shell tool completed successfully; it returned 61 line(s) (2224 bytes) to the model.

**[Step 846 | 2026-07-08T12:54:09.353Z | REASONING SUMMARY 1/2]**

***Evaluating product analysis methods***

*I'm thinking that using UV-Vis for retained product analysis might not provide enough information. Maybe I should consider using liquid chromatography, but it requires at least 1 mL. For product analysis of UOR, there could be nitrate, nitrite, or ammonia present. Since UV-Vis was previously used, I could keep it while adding liquid chromatography for additional analysis of urea residuals. However, we need to be cautious due to compatibility issues. Ultimately, I may need to propose a new experiment.*

**[Step 847 | 2026-07-08T12:54:09.353Z | REASONING SUMMARY 2/2]**

***Reviewing current design***

*I've got to read through the current design before taking any action. It seems we might need to use the same workstation setup as in Round 2. I'll check the station SKILL.md and any audits. It looks like I should first update the plan to complete step 1. We've already retrieved and downloaded some information, and step 2 is partly complete. I'll keep analyzing and provide an update after making more progress.*

**[Step 848 | 2026-07-08T12:54:09.527Z | MODEL]**

Round 3 will stay within the same executable workstation family because Round 2 finally produced analyzable films and electrochemical traces. I'm re-reading the relevant workstation skill/audit files before creating the new workflow, with the changes scoped to carbon loading, composition around the Round 2 lead, and deposition uniformity.

**[Step 849 | 2026-07-08T12:54:10.848Z | TOOL CALL: update_plan]**

Updated the working plan. Plan state: completed: Retrieve Lab 303 task results for both Round 2 task IDs; in_progress: Inspect feedback PDF/images and returned result files; pending: Analyze Round 2 data and write returned document; pending: Design and validate Round 3 workflow; pending: Generate template, dispatch to 303 实验室, and block goal

**[Step 850 | 2026-07-08T12:54:10.906Z | TOOL RESULT: update_plan]**

Plan service accepted and recorded the updated working plan.

**[Step 851 | 2026-07-08T12:54:23.736Z | TOOL CALL: exec_command]**

Read selected file content for inspection. Target(s): `SKILL.md`. Working directory: `/workspace`.

**[Step 852 | 2026-07-08T12:54:23.744Z | TOOL CALL: exec_command]**

Read selected file content for inspection. Target(s): `SKILL.md`. Working directory: `/workspace`.

**[Step 853 | 2026-07-08T12:54:23.745Z | TOOL CALL: exec_command]**

Read selected file content for inspection. Target(s): `SKILL.md`. Working directory: `/workspace`.

**[Step 854 | 2026-07-08T12:54:23.839Z | TOOL CALL: exec_command]**

Read selected file content for inspection. Target(s): `SKILL.md`. Working directory: `/workspace`.

**[Step 855 | 2026-07-08T12:54:23.905Z | TOOL RESULT: exec_command]**

Shell tool completed successfully; it returned 39 line(s) (1403 bytes) to the model. Reported long numeric identifier(s): `1427568512205824`.

**[Step 856 | 2026-07-08T12:54:23.905Z | TOOL RESULT: exec_command]**

Shell tool completed successfully; it returned 95 line(s) (6547 bytes) to the model. Reported long numeric identifier(s): `2002186824385539`.

**[Step 857 | 2026-07-08T12:54:23.906Z | TOOL RESULT: exec_command]**

Shell tool completed successfully; it returned 50 line(s) (4283 bytes) to the model. Reported long numeric identifier(s): `1435269745574912`.

**[Step 858 | 2026-07-08T12:54:23.906Z | TOOL RESULT: exec_command]**

Shell tool completed successfully; it returned 39 line(s) (1812 bytes) to the model. Reported long numeric identifier(s): `2270587971011585`.

**[Step 859 | 2026-07-08T12:54:39.289Z | MODEL]**

I'll keep the successful Round 2 synthesis and purification backbone unchanged: 0.25 M metal stocks, 2.0 M KOH, 70 C aging, 12000 rpm purification, one water wash. The Round 3 changes will be low-carbon composition screening plus smaller repeated electrochemical droplets to reduce ring/nonuniform deposition.

**[Step 860 | 2026-07-08T12:54:39.289Z | TOOL CALL: exec_command]**

Read selected file content for inspection. Target(s): `SKILL.md`. Working directory: `/workspace`.

**[Step 861 | 2026-07-08T12:54:39.289Z | TOOL CALL: exec_command]**

Read selected file content for inspection. Target(s): `SKILL.md`. Working directory: `/workspace`.

**[Step 862 | 2026-07-08T12:54:39.290Z | TOOL CALL: exec_command]**

Read selected file content for inspection. Target(s): `SKILL.md`. Working directory: `/workspace`.

**[Step 863 | 2026-07-08T12:54:39.290Z | TOOL CALL: exec_command]**

Read selected file content for inspection. Target(s): `SKILL.md`. Working directory: `/workspace`.

**[Step 864 | 2026-07-08T12:54:39.347Z | TOOL RESULT: exec_command]**

Shell tool completed successfully; it returned 51 line(s) (2778 bytes) to the model. Reported long numeric identifier(s): `1327807622448128`.

**[Step 865 | 2026-07-08T12:54:39.347Z | TOOL RESULT: exec_command]**

Shell tool completed successfully; it returned 38 line(s) (1388 bytes) to the model. Reported long numeric identifier(s): `1213184425559040`.

**[Step 866 | 2026-07-08T12:54:39.347Z | TOOL RESULT: exec_command]**

Shell tool completed successfully; it returned 41 line(s) (1532 bytes) to the model. Reported long numeric identifier(s): `1526864804185088`.

**[Step 867 | 2026-07-08T12:54:39.347Z | TOOL RESULT: exec_command]**

Shell tool completed successfully; it returned 51 line(s) (2225 bytes) to the model. Reported long numeric identifier(s): `1421325849887744`.

**[Step 868 | 2026-07-08T12:54:54.317Z | MODEL]**

Round 2 also showed UV-Vis product spectra were essentially identical and negative-baseline across 250-700 nm, so UV-Vis will remain as a required retained-liquid spectral check but not as the primary optimization metric unless a real spectral response appears.

**[Step 869 | 2026-07-08T12:54:54.317Z | TOOL CALL: exec_command]**

Read selected file content for inspection. Target(s): `SKILL.md`. Working directory: `/workspace`.

**[Step 870 | 2026-07-08T12:54:54.317Z | TOOL CALL: exec_command]**

Read selected file content for inspection. Target(s): `SKILL.md`. Working directory: `/workspace`.

**[Step 871 | 2026-07-08T12:54:54.317Z | TOOL CALL: exec_command]**

Read selected file content for inspection. Target(s): `SKILL.md`. Working directory: `/workspace`.

**[Step 872 | 2026-07-08T12:54:54.318Z | TOOL CALL: exec_command]**

Read selected file content for inspection. Target(s): `SKILL.md`. Working directory: `/workspace`.

**[Step 873 | 2026-07-08T12:54:54.352Z | TOOL RESULT: exec_command]**

Shell tool completed successfully; it returned 57 line(s) (3186 bytes) to the model. Reported long numeric identifier(s): `1654567694238720`.

**[Step 874 | 2026-07-08T12:54:54.373Z | TOOL RESULT: exec_command]**

Shell tool completed successfully; it returned 35 line(s) (1292 bytes) to the model. Reported long numeric identifier(s): `1656241422599168`.

**[Step 875 | 2026-07-08T12:54:54.373Z | TOOL RESULT: exec_command]**

Shell tool completed successfully; it returned 56 line(s) (3424 bytes) to the model. Reported long numeric identifier(s): `1664870018155520`.

**[Step 876 | 2026-07-08T12:54:54.373Z | TOOL RESULT: exec_command]**

Shell tool completed successfully; it returned 44 line(s) (2187 bytes) to the model. Reported long numeric identifier(s): `1656282895188992`.

**[Step 877 | 2026-07-08T12:55:10.226Z | TOOL CALL: exec_command]**

Read selected file content for inspection. Target(s): `SKILL.md`. Working directory: `/workspace`.

**[Step 878 | 2026-07-08T12:55:10.235Z | TOOL CALL: exec_command]**

Read selected file content for inspection. Target(s): `SKILL.md`. Working directory: `/workspace`.

**[Step 879 | 2026-07-08T12:55:10.236Z | TOOL CALL: exec_command]**

Read selected file content for inspection. Target(s): `General_Material_Station_V1_audit.md`. Working directory: `/workspace`.

**[Step 880 | 2026-07-08T12:55:10.769Z | TOOL CALL: exec_command]**

Read selected file content for inspection. Target(s): `Liquid_Handling_Station_1ml_V2_audit.md`. Working directory: `/workspace`.

**[Step 881 | 2026-07-08T12:55:10.878Z | TOOL RESULT: exec_command]**

Shell tool completed successfully; it returned 99 line(s) (23587 bytes) to the model. Reported long numeric identifier(s): `1390252863751168`, `1509591942430720`, `1549499770798080`, `1571299393601536`, `1339275293851648`, `2164305671290880`.

**[Step 882 | 2026-07-08T12:55:10.879Z | TOOL RESULT: exec_command]**

Shell tool completed successfully; it returned 38 line(s) (1396 bytes) to the model. Reported long numeric identifier(s): `1664783667397632`.

**[Step 883 | 2026-07-08T12:55:10.879Z | TOOL RESULT: exec_command]**

Shell tool completed successfully; it returned 20 line(s) (508 bytes) to the model. Reported long numeric identifier(s): `1427568512205824`.

**[Step 884 | 2026-07-08T12:55:10.880Z | TOOL RESULT: exec_command]**

Shell tool completed successfully; it returned 36 line(s) (987 bytes) to the model. Reported long numeric identifier(s): `2002186824385539`.

**[Step 885 | 2026-07-08T12:55:24.976Z | TOOL CALL: exec_command]**

Read selected file content for inspection. Target(s): `Liquid_Handling_Station_5ml_V1_audit.md`. Working directory: `/workspace`.

**[Step 886 | 2026-07-08T12:55:24.976Z | TOOL CALL: exec_command]**

Read selected file content for inspection. Target(s): `Heating_Magnetic_Stirring_Workstation_V1_audit.md`. Working directory: `/workspace`.

**[Step 887 | 2026-07-08T12:55:24.976Z | TOOL CALL: exec_command]**

Read selected file content for inspection. Target(s): `Purification_Workstation_V1_audit.md`. Working directory: `/workspace`.

**[Step 888 | 2026-07-08T12:55:25.545Z | TOOL CALL: exec_command]**

Read selected file content for inspection. Target(s): `Drying_Oven_V1_audit.md`. Working directory: `/workspace`.

**[Step 889 | 2026-07-08T12:55:25.601Z | TOOL RESULT: exec_command]**

Shell tool completed successfully; it returned 16 line(s) (356 bytes) to the model. Reported long numeric identifier(s): `1435269745574912`.

**[Step 890 | 2026-07-08T12:55:25.602Z | TOOL RESULT: exec_command]**

Shell tool completed successfully; it returned 17 line(s) (401 bytes) to the model. Reported long numeric identifier(s): `2270587971011585`.

**[Step 891 | 2026-07-08T12:55:25.602Z | TOOL RESULT: exec_command]**

Shell tool completed successfully; it returned 15 line(s) (405 bytes) to the model. Reported long numeric identifier(s): `1327807622448128`.

**[Step 892 | 2026-07-08T12:55:25.602Z | TOOL RESULT: exec_command]**

Shell tool completed successfully; it returned 16 line(s) (352 bytes) to the model. Reported long numeric identifier(s): `1213184425559040`.

**[Step 893 | 2026-07-08T12:55:46.583Z | MODEL]**

I'm keeping six vials again because purification requires an even number and the electrochemical station supports six tests cleanly. The sample set will decouple the two Round 2 confounders: carbon loading and Ni/Fe/Co composition near the best low-carbon Ni70Fe10Co20 condition.

**[Step 894 | 2026-07-08T12:55:46.583Z | TOOL CALL: exec_command]**

Read selected file content for inspection. Target(s): `Ultrasonic_Disperser_V1_audit.md`. Working directory: `/workspace`.

**[Step 895 | 2026-07-08T12:55:46.583Z | TOOL CALL: exec_command]**

Read selected file content for inspection. Target(s): `Spectroscopy_Container_Transfer_Station_V1_audit.md`. Working directory: `/workspace`.

**[Step 896 | 2026-07-08T12:55:46.584Z | TOOL CALL: exec_command]**

Read selected file content for inspection. Target(s): `Spectroscopy_Magnetic_Stirrer_Workstation_V1_audit.md`. Working directory: `/workspace`.

**[Step 897 | 2026-07-08T12:55:46.799Z | TOOL CALL: exec_command]**

Read selected file content for inspection. Target(s): `XRD_V1_audit.md`. Working directory: `/workspace`.

**[Step 898 | 2026-07-08T12:55:46.840Z | TOOL RESULT: exec_command]**

Shell tool completed successfully; it returned 17 line(s) (389 bytes) to the model. Reported long numeric identifier(s): `1526864804185088`.

**[Step 899 | 2026-07-08T12:55:46.840Z | TOOL RESULT: exec_command]**

Shell tool completed successfully; it returned 16 line(s) (481 bytes) to the model. Reported long numeric identifier(s): `1654567694238720`.

**[Step 900 | 2026-07-08T12:55:46.840Z | TOOL RESULT: exec_command]**

Shell tool completed successfully; it returned 16 line(s) (382 bytes) to the model. Reported long numeric identifier(s): `1656241422599168`.

**[Step 901 | 2026-07-08T12:55:46.840Z | TOOL RESULT: exec_command]**

Shell tool completed successfully; it returned 26 line(s) (1443 bytes) to the model. Reported long numeric identifier(s): `1664870018155520`.

**[Step 902 | 2026-07-08T12:56:05.464Z | MODEL]**

For Round 3, I'll reduce XRD/FTIR drop-cast volume from 200 uL to 150 uL and change the electrochemical deposition from 100 uL x 3 to 60 uL x 5. That keeps total electrode slurry at 300 uL while encouraging staged drying and more uniform films.

**[Step 903 | 2026-07-08T12:56:05.465Z | TOOL CALL: exec_command]**

Read selected file content for inspection. Target(s): `Infrared_Spectrometer_V1_audit.md`. Working directory: `/workspace`.

**[Step 904 | 2026-07-08T12:56:05.465Z | TOOL CALL: exec_command]**

Read selected file content for inspection. Target(s): `Dual_Station_Electrochemical_Workstation_V2_audit.md`. Working directory: `/workspace`.

**[Step 905 | 2026-07-08T12:56:05.465Z | TOOL CALL: exec_command]**

Read selected file content for inspection. Target(s): `UV_Vis_Spectrometer_V1_audit.md`. Working directory: `/workspace`.

**[Step 906 | 2026-07-08T12:56:05.465Z | TOOL CALL: exec_command]**

Read selected file content for inspection. Target(s): `Container_storaging_Station_V1_audit.md`. Working directory: `/workspace`.

**[Step 907 | 2026-07-08T12:56:05.501Z | TOOL RESULT: exec_command]**

Shell tool completed successfully; it returned 21 line(s) (511 bytes) to the model. Reported long numeric identifier(s): `1656282895188992`.

**[Step 908 | 2026-07-08T12:56:05.508Z | TOOL RESULT: exec_command]**

Shell tool completed successfully; it returned 30 line(s) (831 bytes) to the model. Reported long numeric identifier(s): `1390252863751168`.

**[Step 909 | 2026-07-08T12:56:05.511Z | TOOL RESULT: exec_command]**

Shell tool completed successfully; it returned 18 line(s) (456 bytes) to the model. Reported long numeric identifier(s): `1664783667397632`.

**[Step 910 | 2026-07-08T12:56:05.516Z | TOOL RESULT: exec_command]**

Shell tool completed successfully; it returned 26 line(s) (702 bytes) to the model. Reported long numeric identifier(s): `1421325849887744`.

**[Step 911 | 2026-07-08T12:56:25.594Z | REASONING SUMMARY]**

***Evaluating file processing options***

*I'm considering whether to use Python for file reading/writing, especially when a simple shell command or apply_patch might be enough. It seems like creating a generated workflow from structured data makes sense for data processing. However, I also need to remember to include analysis, design, and constraints when using apply_patch, possibly for documents and chemical JSON. I'm thinking it would be best to automate with Python to handle these repeating tasks without manual effort.*

**[Step 912 | 2026-07-08T12:56:26.067Z | MODEL]**

I'm writing the returned Round 2 document now: it will separate physical observations, Lab 303 data retrieval, quantitative analysis, best condition, and iteration logic. Then I'll generate the Round 3 design files and workflow JSON.

**[Step 913 | 2026-07-08T12:57:11.373Z | TOOL CALL: apply_patch]**

Applied a source patch to add round2_returned_document.md; add round3_design.md; add round3_chemical_specifications.json.

**[Step 914 | 2026-07-08T12:57:11.438Z | TOOL RESULT: apply_patch]**

Patch tool applied the requested trial-local file change successfully.

**[Step 915 | 2026-07-08T12:57:31.879Z | REASONING SUMMARY 1/2]**

***Organizing workflow constraints***

*I need to create a constraint check after the workflow and generate a workflow request using Python to avoid heavy manual work. It's essential to be careful and include labels and volumes to validate these constraints. I may not need to use `Liquid Handling 5ml`, but it's important for synthesis water and redispersion. I've detailed stock volumes for samples S1 through S6, totaling different amounts for each component. Everything adds up nicely per split; good to go!*

**[Step 916 | 2026-07-08T12:57:31.879Z | REASONING SUMMARY 2/2]**

***Finalizing KOH and water volumes***

*I need to ensure each KOH split is exactly 3.0 mL, so a single source limit must be ≤ 3.0 mL. Splitting KOH into 5/6 allows me to reach that target. I've confirmed that the 1 mL for v2 total is also within the ≤ 3.0 mL limit. Additionally, the water source is set to a total of 15.6 mL, which is under 30, while redispersion comes to 28.8 mL, also safe. Everything checks out!*

**[Step 917 | 2026-07-08T12:57:32.227Z | MODEL]**

I'm generating the Round 3 workflow request programmatically from the six vial definitions to avoid hand-copy errors in the repeated electrochemical task blocks. The generated JSON will still be validated with `jq` before sending it to the workflow service.

**[Step 918 | 2026-07-08T12:58:01.775Z | TOOL CALL: exec_command]**

Ran Python to parse, transform or validate structured experiment data. Target(s): `lab303_uor_nicofe_closed_loop`, `mL`, `v`. Working directory: `/workspace`.

**[Step 919 | 2026-07-08T12:58:01.834Z | TOOL RESULT: exec_command]**

Shell tool completed successfully; it returned 2 line(s) (70 bytes) to the model.

**[Step 920 | 2026-07-08T12:58:15.125Z | REASONING SUMMARY]**

***Validating JSON and constraints***

*I need to validate some JSON to ensure it's structured correctly. I'm considering inspecting it for conciseness as well; it should be clear and to the point. Additionally, I think creating some constraint checks could help prevent any issues down the line. It's essential to ensure reliability in the data we're working with. I just want to make sure everything is in order and works smoothly!*

**[Step 921 | 2026-07-08T12:58:17.703Z | MODEL]**

The Round 3 workflow request is generated. I'm validating the JSON and checking step IDs, operations, and volume limits before workflow generation.

**[Step 922 | 2026-07-08T12:58:17.705Z | TOOL CALL: exec_command]**

Ran a shell command to inspect, generate or validate experiment artifacts. Target(s): `round3_workflow_request.json`. Working directory: `/workspace`.

**[Step 923 | 2026-07-08T12:58:17.706Z | TOOL CALL: exec_command]**

Ran a shell command to inspect, generate or validate experiment artifacts. Target(s): `round3_workflow_request.json`. Working directory: `/workspace`.

**[Step 924 | 2026-07-08T12:58:17.706Z | TOOL CALL: exec_command]**

Ran a shell command to inspect, generate or validate experiment artifacts. Target(s): `round3_workflow_request.json`. Working directory: `/workspace`.

**[Step 925 | 2026-07-08T12:58:17.707Z | TOOL CALL: exec_command]**

Ran Python to parse, transform or validate structured experiment data. Target(s): `round3_workflow_request.json`. Working directory: `/workspace`.

**[Step 926 | 2026-07-08T12:58:17.772Z | TOOL RESULT: exec_command]**

Shell tool completed successfully; it returned 0 line(s) (0 bytes) to the model.

**[Step 927 | 2026-07-08T12:58:17.772Z | TOOL RESULT: exec_command]**

Shell tool completed successfully; it returned 2 line(s) (3 bytes) to the model.

**[Step 928 | 2026-07-08T12:58:17.788Z | TOOL RESULT: exec_command]**

Shell tool completed successfully; it returned 24 line(s) (1386 bytes) to the model. Reported long numeric identifier(s): `1427568512205824`, `2002186824385539`, `1435269745574912`, `2270587971011585`, `1327807622448128`, `1213184425559040`.

**[Step 929 | 2026-07-08T12:58:17.789Z | TOOL RESULT: exec_command]**

Shell tool completed successfully; it returned 4 line(s) (324 bytes) to the model. Reported long numeric identifier(s): `8900000000000001`.

**[Step 930 | 2026-07-08T12:58:32.523Z | MODEL]**

The validation checks pass: 23 steps, all workstation IDs match the selected stations, 1 mL station stocks stay at or below 3.0 mL per source bottle, and 5 mL station stocks stay below 30 mL. I'm saving the formal Round 3 constraint check before sending it to the generator.

**[Step 931 | 2026-07-08T12:58:43.572Z | TOOL CALL: apply_patch]**

Applied a source patch to add round3_constraint_check.md.

**[Step 932 | 2026-07-08T12:58:43.633Z | TOOL RESULT: apply_patch]**

Patch tool applied the requested trial-local file change successfully.

**[Step 933 | 2026-07-08T12:58:58.386Z | REASONING SUMMARY]**

***Planning workflow updates***

*I'm thinking I need to update my plan. First, I should inspect the document that's been completed and then move on to generating the workflow. But maybe I need to mention the final steps later, just to keep everything organized. So, I'll generate the workflow using the workflow generator command. That should set me up nicely for what's next!*

**[Step 934 | 2026-07-08T12:58:58.486Z | MODEL]**

The Round 3 design and constraint files are complete. I'm now submitting the workflow request to the workflow-generator service to create a Lab 303 template.

**[Step 935 | 2026-07-08T12:58:59.158Z | TOOL CALL: update_plan]**

Updated the working plan. Plan state: completed: Retrieve Lab 303 task results for both Round 2 task IDs; completed: Inspect feedback PDF/images and returned result files; completed: Analyze Round 2 data and write returned document; in_progress: Design and validate Round 3 workflow; pending: Generate template, dispatch to 303 实验室, and block goal

**[Step 936 | 2026-07-08T12:58:59.203Z | TOOL RESULT: update_plan]**

Plan service accepted and recorded the updated working plan.

**[Step 937 | 2026-07-08T12:59:14.586Z | TOOL CALL: exec_command]**

Ran Python to parse, transform or validate structured experiment data. Target(s): `generate.py`, `round3_workflow_request.json)`, `round3_workflow_generator_output.txt`. Working directory: `/workspace`.

**[Step 938 | 2026-07-08T12:59:15.303Z | TOOL RESULT: exec_command]**

Shell tool completed with exit code 1; it returned 0 line(s) (0 bytes) to the model.

**[Step 939 | 2026-07-08T12:59:32.287Z | MODEL]**

The first workflow-generator call returned a nonzero exit code. I'm opening its saved response to see whether this is a parse error or a service-side constraint issue, then I'll revise the workflow if needed.

**[Step 940 | 2026-07-08T12:59:34.578Z | TOOL CALL: exec_command]**

Read selected file content for inspection. Target(s): `round3_workflow_generator_output.txt`. Working directory: `/workspace`.

**[Step 941 | 2026-07-08T12:59:34.622Z | TOOL RESULT: exec_command]**

Shell tool completed successfully; it returned 3 line(s) (8345 bytes) to the model. Reported long numeric identifier(s): `2002186824385539`.

**[Step 942 | 2026-07-08T12:59:59.825Z | REASONING SUMMARY 1/3]**

***Troubleshooting generator issues***

*I'm trying to figure out why the generator failed at step three. Could it be the decimal precision needing four places? Maybe the parser doesn't accept certain fractions? I noticed round one and two had some similar numbers. There could be issues with extra keys or long sample labels too. I'll focus on adjusting to three decimal points and ensuring total metals remain around 0.9. That should simplify things a lot.*

**[Step 943 | 2026-07-08T12:59:59.825Z | REASONING SUMMARY 2/3]**

***Creating lowC composition neighbors***

*I need to establish lowC composition neighbors from the original Round 1 and 2 samples with feasible 0.09 increments. For Round 3, I'm considering a sampling set that includes compositions like Ni70Fe10Co20 C5, C10, and C15, along with others. This should help eliminate some decimals like 0.3375 and 0.2925. Volumes should be multiples of 0.09 or 0.18, while maintaining a Ni split of around 0.315, 0.36, and 0.27. If needed, I could simplify with a single carbon source.*

**[Step 944 | 2026-07-08T12:59:59.825Z | REASONING SUMMARY 3/3]**

***Analyzing carbon source and process issues***

*I see that the carbon source total is 3.0 when not split, which helps avoid the fractions of 0.125 and 0.375. I'm going to stick with stock bottle 7, removing bottle 8 since it adds unnecessary complexity. I'm pondering if the specific issue relates to how "NiSO4·6H2O" was defined and if decimals caused any failures. I think I need to patch the design rather than starting from scratch and document the errors appropriately before regenerating with a simpler sample set.*

**[Step 945 | 2026-07-08T13:00:00.237Z | MODEL]**

The generator rejected the 1 mL addition step. The likely weak point is the very fine split volumes for the intermediate compositions/carbon splits, so I'm simplifying the Round 3 sample set to use previously accepted 0.09 mL increment composition volumes and a single carbon-dispersion source capped at exactly 3.0 mL total. This preserves the scientific intent while reducing parser risk.

**[Step 946 | 2026-07-08T13:00:01.015Z | TOOL CALL: exec_command]**

Ran a shell command to inspect, generate or validate experiment artifacts. Target(s): `round3_workflow_generator_output.txt`, `round3_workflow_generator_output_failed1.txt`. Working directory: `/workspace`.

**[Step 947 | 2026-07-08T13:00:01.073Z | TOOL RESULT: exec_command]**

Shell tool completed successfully; it returned 0 line(s) (0 bytes) to the model.

**[Step 948 | 2026-07-08T13:00:24.143Z | TOOL CALL: apply_patch]**

Applied a source patch to update round3_design.md; update round3_chemical_specifications.json; update round3_constraint_check.md.

**[Step 949 | 2026-07-08T13:00:24.183Z | TOOL RESULT: apply_patch]**

Patch tool reported that the requested patch was not applied successfully.

**[Step 950 | 2026-07-08T13:00:42.661Z | MODEL]**

The patch context drifted slightly, so I'm reading the exact current design/spec snippets and applying the revision against the actual text.

**[Step 951 | 2026-07-08T13:00:42.663Z | TOOL CALL: exec_command]**

Searched files for relevant records, parameters or identifiers. Target(s): `round3_design.md`. Working directory: `/workspace`.

**[Step 952 | 2026-07-08T13:00:42.664Z | TOOL CALL: exec_command]**

Read selected file content for inspection. Target(s): `round3_chemical_specifications.json`. Working directory: `/workspace`.

**[Step 953 | 2026-07-08T13:00:42.669Z | TOOL CALL: exec_command]**

Searched files for relevant records, parameters or identifiers. Target(s): `round3_constraint_check.md`. Working directory: `/workspace`.

**[Step 954 | 2026-07-08T13:00:42.726Z | TOOL RESULT: exec_command]**

Shell tool completed successfully; it returned 8 line(s) (915 bytes) to the model.

**[Step 955 | 2026-07-08T13:00:42.727Z | TOOL RESULT: exec_command]**

Shell tool completed successfully; it returned 64 line(s) (2278 bytes) to the model.

**[Step 956 | 2026-07-08T13:00:42.727Z | TOOL RESULT: exec_command]**

Shell tool completed successfully; it returned 5 line(s) (333 bytes) to the model.

**[Step 957 | 2026-07-08T13:01:02.282Z | TOOL CALL: apply_patch]**

Applied a source patch to update round3_design.md; update round3_chemical_specifications.json; update round3_constraint_check.md.

**[Step 958 | 2026-07-08T13:01:02.334Z | TOOL RESULT: apply_patch]**

Patch tool reported that the requested patch was not applied successfully.

**[Step 959 | 2026-07-08T13:01:23.095Z | TOOL CALL: exec_command]**

Read selected file content for inspection. Target(s): `round3_design.md`. Working directory: `/workspace`.

**[Step 960 | 2026-07-08T13:01:23.135Z | TOOL RESULT: exec_command]**

Shell tool completed successfully; it returned 41 line(s) (1996 bytes) to the model.

**[Step 961 | 2026-07-08T13:01:39.197Z | TOOL CALL: apply_patch]**

Applied a source patch to update round3_design.md.

**[Step 962 | 2026-07-08T13:01:39.246Z | TOOL RESULT: apply_patch]**

Patch tool applied the requested trial-local file change successfully.

**[Step 963 | 2026-07-08T13:01:55.591Z | MODEL]**

The design text is now revised to use the simpler Round 3 composition set. I'm updating the chemical-specification and constraint files to match.

**[Step 964 | 2026-07-08T13:02:02.332Z | TOOL CALL: apply_patch]**

Applied a source patch to update round3_chemical_specifications.json; update round3_constraint_check.md.

**[Step 965 | 2026-07-08T13:02:02.363Z | TOOL RESULT: apply_patch]**

Patch tool applied the requested trial-local file change successfully.

**[Step 966 | 2026-07-08T13:02:22.536Z | REASONING SUMMARY]**

***Simplifying the workflow***

*I need to regenerate the workflow to make it simpler. I think using Python would be similar and easier to manage. There might be a need to patch it, still avoiding version 8. It sounds like it's time to update the generation code as well. Once that's done, I can rerun the process and see how it all works together. Let's keep it efficient and straightforward!*

**[Step 967 | 2026-07-08T13:02:22.877Z | MODEL]**

I'm regenerating the Round 3 workflow request with the simplified sample set and single carbon source bottle. Then I'll re-run the same JSON and bottle-volume checks.

**[Step 968 | 2026-07-08T13:02:45.491Z | TOOL CALL: exec_command]**

Ran Python to parse, transform or validate structured experiment data. Target(s): `lab303_uor_nicofe_closed_loop`, `mL`, `v`. Working directory: `/workspace`.

**[Step 969 | 2026-07-08T13:02:45.552Z | TOOL RESULT: exec_command]**

Shell tool completed successfully; it returned 2 line(s) (70 bytes) to the model.

**[Step 970 | 2026-07-08T13:03:02.622Z | MODEL]**

The regenerated workflow request is in place. I'm running the same validation checks again and updating the constraint file if the computed source totals differ from the written check.

**[Step 971 | 2026-07-08T13:03:02.623Z | TOOL CALL: exec_command]**

Ran a shell command to inspect, generate or validate experiment artifacts. Target(s): `round3_workflow_request.json`. Working directory: `/workspace`.

**[Step 972 | 2026-07-08T13:03:02.623Z | TOOL CALL: exec_command]**

Ran Python to parse, transform or validate structured experiment data. Target(s): `round3_workflow_request.json`. Working directory: `/workspace`.

**[Step 973 | 2026-07-08T13:03:02.623Z | TOOL CALL: exec_command]**

Ran a shell command to inspect, generate or validate experiment artifacts. Target(s): `round3_workflow_request.json`. Working directory: `/workspace`.

**[Step 974 | 2026-07-08T13:03:02.623Z | TOOL CALL: exec_command]**

Ran a shell command to inspect, generate or validate experiment artifacts. Target(s): `round3_workflow_request.json`. Working directory: `/workspace`.

**[Step 975 | 2026-07-08T13:03:02.680Z | TOOL RESULT: exec_command]**

Shell tool completed successfully; it returned 0 line(s) (0 bytes) to the model.

**[Step 976 | 2026-07-08T13:03:02.688Z | TOOL RESULT: exec_command]**

Shell tool completed successfully; it returned 4 line(s) (316 bytes) to the model. Reported long numeric identifier(s): `8900000000000001`, `8999999999999999`.

**[Step 977 | 2026-07-08T13:03:02.688Z | TOOL RESULT: exec_command]**

Shell tool completed with exit code 3; it returned 4 line(s) (449 bytes) to the model.

**[Step 978 | 2026-07-08T13:03:02.692Z | TOOL RESULT: exec_command]**

Shell tool completed successfully; it returned 11 line(s) (485 bytes) to the model.

**[Step 979 | 2026-07-08T13:03:19.425Z | REASONING SUMMARY]**

***Updating constraints***

*I need to update the Fe/Co totals in my check because they're possibly incorrect. Currently, Fe is at 0.63 and Co at 0.99, but my patch suggests it should be Fe at 0.72 and Co at 0.90. It feels like there might be a line or connection that's missing. I should double-check my numbers and ensure everything aligns correctly!*

**[Step 980 | 2026-07-08T13:03:19.514Z | MODEL]**

The regenerated source totals are still within limits. Fe and Co totals changed with the simplified compositions, so I'm correcting those two entries in the constraint file before retrying workflow generation.

**[Step 981 | 2026-07-08T13:03:20.366Z | TOOL CALL: apply_patch]**

Applied a source patch to update round3_constraint_check.md.

**[Step 982 | 2026-07-08T13:03:20.423Z | TOOL RESULT: apply_patch]**

Patch tool applied the requested trial-local file change successfully.

**[Step 983 | 2026-07-08T13:03:33.942Z | MODEL]**

I'm retrying workflow generation with the parser-simplified Round 3 request.

**[Step 984 | 2026-07-08T13:03:35.252Z | TOOL CALL: exec_command]**

Ran Python to parse, transform or validate structured experiment data. Target(s): `generate.py`, `round3_workflow_request.json)`, `round3_workflow_generator_output.txt`. Working directory: `/workspace`.

**[Step 985 | 2026-07-08T13:03:35.995Z | TOOL RESULT: exec_command]**

Shell tool completed with exit code 1; it returned 0 line(s) (0 bytes) to the model.

**[Step 986 | 2026-07-08T13:04:08.740Z | MODEL]**

The second generation attempt also returned nonzero. I'm checking the saved error; if it is still on the addition step, I'll reduce optional metadata in that step because the core volumes are now conventional.

**[Step 987 | 2026-07-08T13:04:11.046Z | TOOL CALL: exec_command]**

Read selected file content for inspection. Target(s): `round3_workflow_generator_output.txt`. Working directory: `/workspace`.

**[Step 988 | 2026-07-08T13:04:11.111Z | TOOL RESULT: exec_command]**

Shell tool completed successfully; it returned 3 line(s) (6818 bytes) to the model. Reported long numeric identifier(s): `2002186824385539`.

**[Step 989 | 2026-07-08T13:04:27.271Z | REASONING SUMMARY]**

***Evaluating round comparisons***

*I'm still on step 3. I need to figure out what differs from Round 2's success. Did Round 2 actually generate anything? If it did, it had 8 bottles, each with a value of 0.315. Let me check the workflow request to see if there were any long labels or source definitions that might clarify things. I think there's some confusion there that needs clearing up!*

**[Step 990 | 2026-07-08T13:04:28.362Z | MODEL]**

The error remains at the same liquid-addition conversion step, so the remaining culprit is likely optional metadata or the number/shape of source bottles in that step, not the actual volume constraints. I'm comparing against the successfully generated Round 2 addition step and will strip the Round 3 addition block down to the exact accepted schema.

**[Step 991 | 2026-07-08T13:04:28.363Z | TOOL CALL: exec_command]**

Ran a shell command to inspect, generate or validate experiment artifacts. Target(s): `round2_workflow_request.json`. Working directory: `/workspace`.

**[Step 992 | 2026-07-08T13:04:28.364Z | TOOL CALL: exec_command]**

Ran a shell command to inspect, generate or validate experiment artifacts. Target(s): `round2_workflow_request.json`. Working directory: `/workspace`.

**[Step 993 | 2026-07-08T13:04:28.366Z | TOOL CALL: exec_command]**

Ran a shell command to inspect, generate or validate experiment artifacts. Target(s): `round3_workflow_request.json`. Working directory: `/workspace`.

**[Step 994 | 2026-07-08T13:04:28.368Z | TOOL CALL: exec_command]**

Ran a shell command to inspect, generate or validate experiment artifacts. Target(s): `round3_workflow_request.json`. Working directory: `/workspace`.

**[Step 995 | 2026-07-08T13:04:28.420Z | TOOL RESULT: exec_command]**

Shell tool completed with exit code 3; it returned 4 line(s) (255 bytes) to the model.

**[Step 996 | 2026-07-08T13:04:28.435Z | TOOL RESULT: exec_command]**

Shell tool completed successfully; it returned 8 line(s) (96 bytes) to the model.

**[Step 997 | 2026-07-08T13:04:28.435Z | TOOL RESULT: exec_command]**

Shell tool completed with exit code 3; it returned 4 line(s) (255 bytes) to the model.

**[Step 998 | 2026-07-08T13:04:28.435Z | TOOL RESULT: exec_command]**

Shell tool completed successfully; it returned 8 line(s) (96 bytes) to the model.

**[Step 999 | 2026-07-08T13:04:41.445Z | MODEL]**

The shell `jq` build is choking on direct Chinese key syntax, so I'm using Python for that comparison.

**[Step 1000 | 2026-07-08T13:04:43.217Z | TOOL CALL: exec_command]**

Ran Python to parse, transform or validate structured experiment data. Target(s): `lab303_uor_nicofe_closed_loop`. Working directory: `/workspace`.

**[Step 1001 | 2026-07-08T13:04:43.270Z | TOOL RESULT: exec_command]**

Shell tool completed successfully; it returned 9 line(s) (2448 bytes) to the model.

**[Step 1002 | 2026-07-08T13:05:06.484Z | MODEL]**

The Round 2 accepted schema used eight liquid sources for this step. I'm returning to that shape but rounding carbon split volumes to two decimals (`0.13/0.12`, `0.38/0.37`) and keeping composition volumes in accepted 0.09 mL increments.

**[Step 1003 | 2026-07-08T13:05:11.039Z | TOOL CALL: exec_command]**

Ran Python to parse, transform or validate structured experiment data. Target(s): `round3_workflow_request.json`, `mL`, `v`. Working directory: `/workspace`.

**[Step 1004 | 2026-07-08T13:05:11.092Z | TOOL RESULT: exec_command]**

Shell tool completed successfully; it returned 2 line(s) (70 bytes) to the model.

**[Step 1005 | 2026-07-08T13:05:30.110Z | TOOL CALL: apply_patch]**

Applied a source patch to update round3_chemical_specifications.json; update round3_constraint_check.md.

**[Step 1006 | 2026-07-08T13:05:30.152Z | TOOL RESULT: apply_patch]**

Patch tool applied the requested trial-local file change successfully.

**[Step 1007 | 2026-07-08T13:05:46.304Z | REASONING SUMMARY]**

***Updating design generator***

*I think I need to improve the robustness of the design generator, especially regarding that line about carbon as one source. It's important to make sure everything is functioning well and can handle various inputs effectively. I'll consider how to patch it up. It's all about ensuring reliability and efficiency, right? Let's figure out the best way to make these updates!*

**[Step 1008 | 2026-07-08T13:05:46.537Z | MODEL]**

I updated the chemical and constraint documents to match the restored split-carbon source format. I'm correcting the design note too, then validating source totals again.

**[Step 1009 | 2026-07-08T13:05:48.070Z | TOOL CALL: apply_patch]**

Applied a source patch to update round3_design.md.

**[Step 1010 | 2026-07-08T13:05:48.116Z | TOOL RESULT: apply_patch]**

Patch tool applied the requested trial-local file change successfully.

**[Step 1011 | 2026-07-08T13:06:03.359Z | TOOL CALL: exec_command]**

Ran a shell command to inspect, generate or validate experiment artifacts. Target(s): `round3_workflow_request.json`. Working directory: `/workspace`.

**[Step 1012 | 2026-07-08T13:06:03.559Z | TOOL CALL: exec_command]**

Ran Python to parse, transform or validate structured experiment data. Target(s): `round3_workflow_request.json`. Working directory: `/workspace`.

**[Step 1013 | 2026-07-08T13:06:04.438Z | TOOL CALL: exec_command]**

Ran Python to parse, transform or validate structured experiment data. Target(s): `round3_workflow_request.json`. Working directory: `/workspace`.

**[Step 1014 | 2026-07-08T13:06:04.492Z | TOOL RESULT: exec_command]**

Shell tool completed successfully; it returned 0 line(s) (0 bytes) to the model.

**[Step 1015 | 2026-07-08T13:06:04.493Z | TOOL RESULT: exec_command]**

Shell tool completed successfully; it returned 4 line(s) (340 bytes) to the model. Reported long numeric identifier(s): `8900000000000001`, `8999999999999999`.

**[Step 1016 | 2026-07-08T13:06:04.493Z | TOOL RESULT: exec_command]**

Shell tool completed successfully; it returned 7 line(s) (252 bytes) to the model.

**[Step 1017 | 2026-07-08T13:06:17.615Z | MODEL]**

The revised request is valid and all source totals remain within limits. I'm retrying workflow generation.

**[Step 1018 | 2026-07-08T13:06:21.755Z | TOOL CALL: exec_command]**

Ran Python to parse, transform or validate structured experiment data. Target(s): `generate.py`, `round3_workflow_request.json)`, `round3_workflow_generator_output.txt`. Working directory: `/workspace`.

**[Step 1019 | 2026-07-08T13:06:22.509Z | TOOL RESULT: exec_command]**

Shell tool completed with exit code 1; it returned 0 line(s) (0 bytes) to the model.

**[Step 1020 | 2026-07-08T13:06:35.471Z | MODEL]**

The generator still failed, so I'm checking the exact response again. If it is still the same step, I'll test a minimal clone of the known-good Round 2 workflow with only the plan name changed to determine whether the service is rejecting new liquid additions generally or only this request.

**[Step 1021 | 2026-07-08T13:06:36.402Z | TOOL CALL: exec_command]**

Read selected file content for inspection. Target(s): `round3_workflow_generator_output.txt`. Working directory: `/workspace`.

**[Step 1022 | 2026-07-08T13:06:36.442Z | TOOL RESULT: exec_command]**

Shell tool completed successfully; it returned 2 line(s) (4000 bytes) to the model. Reported long numeric identifier(s): `2002186824385539`.

**[Step 1023 | 2026-07-08T13:06:50.891Z | REASONING SUMMARY]**

***Troubleshooting error causes***

*It seems like the error might be related to the sample labels including "C5" or "anchor." I could try stripping out optional keys like `样品标签`, `原液瓶定义`, and `碳粉浓度`, although those were present in Round2. I'm also wondering if the "anchor" string is part of the issue. I'll test the known-good Round2 with the original plan name to see if the service works, and if it succeeds, I can identify the differences.*

**[Step 1024 | 2026-07-08T13:06:51.076Z | MODEL]**

The service is still rejecting only the 1 mL addition step. I'm doing one diagnostic generator call with the previously accepted Round 2 workflow body and a diagnostic plan name; that will tell whether the service is healthy and whether the failure is truly request-shape-specific.

**[Step 1025 | 2026-07-08T13:06:54.538Z | TOOL CALL: exec_command]**

Ran Python to parse, transform or validate structured experiment data. Target(s): `lab303_uor_nicofe_closed_loop`, `generate.py`, `round3_diagnostic_round2_clone_workflow_request.json)`, `round3_diagnostic_round2_clone_workflow_generator_output.txt`. Working directory: `/workspace`.

**[Step 1026 | 2026-07-08T13:06:55.259Z | TOOL RESULT: exec_command]**

Shell tool completed with exit code 1; it returned 0 line(s) (0 bytes) to the model.

**[Step 1027 | 2026-07-08T13:07:08.986Z | MODEL]**

The diagnostic Round 2-clone generation also returned nonzero, so this looks like a current workflow-generator/liquid-handling conversion issue rather than only the Round 3 chemistry. I'm checking the exact diagnostic response before choosing the least risky dispatch path.

**[Step 1028 | 2026-07-08T13:07:09.580Z | TOOL CALL: exec_command]**

Read selected file content for inspection. Target(s): `round3_diagnostic_round2_clone_workflow_generator_output.txt`. Working directory: `/workspace`.

**[Step 1029 | 2026-07-08T13:07:09.624Z | TOOL RESULT: exec_command]**

Shell tool completed successfully; it returned 2 line(s) (2500 bytes) to the model. Reported long numeric identifier(s): `2002186824385539`.

**[Step 1030 | 2026-07-08T13:07:28.833Z | REASONING SUMMARY]**

***Considering liquid handling options***

*I'm thinking about using `Liquid_Handling_Station_1ml_V1`, which supports 8 different liquids, and I wonder if the parser works with it. I haven't read the V1 skill yet, but it should be able to handle open, add, transfer, and close actions for those 8 bottles. I'll need to read the skill and audit it before proceeding. Also, I need to check the workstation ID, which might be 200215.*

**[Step 1031 | 2026-07-08T13:07:30.493Z | MODEL]**

Because the known-good Round 2 clone now fails on `移液平台_1ml_V2`, I'm switching the small-volume liquid-handling steps to `移液平台_1ml_V1`, which supports the same open/add/close operations and up to 8 stock solutions. That avoids the currently failing V2 converter while staying inside available workstation capabilities.

**[Step 1032 | 2026-07-08T13:07:30.495Z | TOOL CALL: exec_command]**

Read selected file content for inspection. Target(s): `SKILL.md`. Working directory: `/workspace`.

**[Step 1033 | 2026-07-08T13:07:30.496Z | TOOL CALL: exec_command]**

Read selected file content for inspection. Target(s): `Liquid_Handling_Station_1ml_V1_audit.md`. Working directory: `/workspace`.

**[Step 1034 | 2026-07-08T13:07:30.499Z | TOOL CALL: exec_command]**

Searched files for relevant records, parameters or identifiers. Target(s): `chemistry-experiment-workstation`. Working directory: `/workspace`.

**[Step 1035 | 2026-07-08T13:07:30.565Z | TOOL RESULT: exec_command]**

Shell tool completed successfully; it returned 121 line(s) (8649 bytes) to the model. Reported long numeric identifier(s): `1418510906065920`.

**[Step 1036 | 2026-07-08T13:07:30.565Z | TOOL RESULT: exec_command]**

Shell tool completed successfully; it returned 46 line(s) (1329 bytes) to the model. Reported long numeric identifier(s): `1418510906065920`.

**[Step 1037 | 2026-07-08T13:07:30.565Z | TOOL RESULT: exec_command]**

Shell tool completed successfully; it returned 2 line(s) (53 bytes) to the model.

**[Step 1038 | 2026-07-08T13:08:24.439Z | MODEL]**

**Handoff Summary: Lab 303 Autonomous Catalysis Loop**

**Goal / Current State**
- Project: autonomous closed-loop Lab 303 catalysis experiments for alkaline urea oxidation reaction (UOR) using Ni-Fe-Co oxyhydroxide/carbon catalysts.
- Main directory: `/workspace/lab303_uor_nicofe_closed_loop`
- Goal status currently should remain active but was previously `blocked` after Round 2 dispatch. User has now reported Round 2 physically completed and requested retrieval, returned document, Round 3 design, and dispatch.
- Current turn is interrupted by a context checkpoint while working on Round 3 workflow generation/dispatch.
- Must continue with skills when resuming:
  - `lab-operation`: `/root/.codex/skills/lab-operation/SKILL.md`
  - `chemistry-experiment-workstation`: `/root/.codex/skills/chemistry-experiment-workstation/SKILL.md`
  - `workflow-generator`: `/root/.codex/skills/workflow-generator/SKILL.md`

**User Requirements / Constraints**
- Only data physically returned from Lab 303 counts. Do not fabricate any experiment or datum.
- Each experimental round must include synthesis, post-processing, testing, and characterization.
- Use current lab inventory when possible; inventory is already parsed at `/workspace/lab303_uor_nicofe_closed_loop/reagent_inventory.csv`.
- All process files and deliverables must stay under `/workspace`.
- Workflow loop: design -> dispatch to Lab 303 and record task ID -> set goal blocked -> wait for executability/physical completion feedback.
- If a workflow generation attempt fails, preserve failed outputs and revise design/workflow, don't silently ignore.

**Round 1 Recap**
- Physical task ID: `2562915314040832`
- Returned status: `错误`, but result retrieval succeeded with 5 workstations and 38 files.
- Main conclusion: Round 1 failed due to poor solid recovery and sparse deposition. Electrode had too little sample; electrochemical data invalid for ranking.
- Sample 4 (`Ni70Fe10Co20`) had higher raw current but was treated only as weak lead due to invalid loading.

**Round 2 Dispatch Recap**
- Round 2 intended to fix physical-loading failure.
- Template ID generated: `2566549749628928`
- Initially dispatched task ID: `2566550687481856`
- User later reported actual completed tasks:
  - Main interrupted task: `2567667771441152`
  - Supplemental task after interruption: `2568957291102208`

**Round 2 Data Retrieved**
- Saved task queries:
  - `/workspace/lab303_uor_nicofe_closed_loop/round2_main_task_2567667771441152_result.json`
  - `/workspace/lab303_uor_nicofe_closed_loop/round2_main_task_2567667771441152_status.json`
  - `/workspace/lab303_uor_nicofe_closed_loop/round2_supplemental_task_2568957291102208_result.json`
  - `/workspace/lab303_uor_nicofe_closed_loop/round2_supplemental_task_2568957291102208_status.json`
- Status for both tasks: `错误`, but result retrieval succeeded.
- Downloaded result directories:
  - `/workspace/lab303_uor_nicofe_closed_loop/task_2567667771441152_results`
    - 1 heating/magnetic-stirring CSV.
  - `/workspace/lab303_uor_nicofe_closed_loop/task_2568957291102208_results`
    - 42 files: XRD x 6 JSON/JPG, FTIR x 6 JSON/JPG, electrochem x 6 JSON, UV-Vis x 6 JSON/JPG.
- Download manifests saved as `manifest.json` inside each result dir.

**Round 2 Feedback Files**
- PDF: `/workspace/feedback/round2/Second_Round_Experiment_Feedback_EN.pdf`
- Rendered images:
  - `/workspace/feedback/round2/pages/Second_Round_Experiment_Feedback_EN_page-1.png`
  - `/workspace/feedback/round2/pages/Second_Round_Experiment_Feedback_EN_page-2.png`
- `pdftotext` was not installed, so inspection was done via rendered PNGs.
- Process photos showed visible dark deposits on XRD/FTIR carriers and visible dark catalyst spots on carbon paper. Round 2 fixed Round 1 under-loading, but films remain ring-like/nonuniform.

**Round 2 Analysis Files Created**
- `/workspace/lab303_uor_nicofe_closed_loop/round2_combined_data_summary.csv`
- `/workspace/lab303_uor_nicofe_closed_loop/round2_combined_data_summary.json`
- `/workspace/lab303_uor_nicofe_closed_loop/round2_returned_data_montage.jpg`
- `/workspace/lab303_uor_nicofe_closed_loop/round2_returned_document.md`

**Round 2 Key Findings**
- Heating log: 1657 valid points, max 70.9 C, median 70.0 C, final 70.0 C. Synthesis aging worked.
- XRD: six valid spectra, visible material, broad substrate/carbon/poor-crystallinity features. Much better deposition than Round 1.
- FTIR: six spectra, visible films, still noisy/artifact-prone.
- UV-Vis: six spectra, largely similar negative baseline in 250-700 nm; no reliable product-concentration ranking.
- Electrochemistry: six valid mA-scale CA traces, major improvement over Round 1.
- Round 2 sample map:
  - S1 `R2-S1_Ni70Fe10Co20_lowC_80-20`
  - S2 `R2-S2_Ni70Fe10Co20_highC_80-20`
  - S3 `R2-S3_Ni70Fe10Co20_highC_60-40`
  - S4 `R2-S4_Ni70Fe10Co20_highC_95-5`
  - S5 `R2-S5_Ni80Fe10Co10_highC_80-20`
  - S6 `R2-S6_Ni60Fe20Co20_highC_80-20`
- Electrochemical summary:
  - S1: initial 2.123 mA, final 2.316 mA, last-300 s mean 2.303 mA, final/initial 1.091
  - S2: initial 4.141 mA, final 1.402 mA, last-300 s mean 1.348 mA, final/initial 0.339
  - S3: initial 1.313 mA, final 0.988 mA, last-300 s mean 1.441 mA, final/initial 0.752
  - S4: initial 9.401 mA, final 1.357 mA, last-300 s mean 1.442 mA, final/initial 0.144
  - S5: initial 2.326 mA, final 1.786 mA, last-300 s mean 1.692 mA, final/initial 0.768
  - S6: initial 3.676 mA, final 1.686 mA, last-300 s mean 1.684 mA, final/initial 0.459
- Best Round 2 condition: `Ni70Fe10Co20` with low carbon, 80:20 water:IPA, 10 mg carbon feed. It had the highest sustained/final current and only final/initial > 1.

**Round 3 Design Decisions Already Made**
- Keep successful Round 2 synthesis/purification backbone:
  - Total metal stock: 0.900 mL of 0.25 M metal stock equivalents per vial.
  - KOH: 1.000 mL of 2.0 M KOH per vial.
  - Total synthesis volume before heating: 5.000 mL.
  - Aging: 70 C, 1000 rpm, 120 min.
  - Purification: solid retention, 12000 rpm, 20 min, one 5 mL water wash.
  - Drying: 60 C, 180 min.
  - Redispersion: 4.800 mL 80:20 v/v DI water:isopropanol for every vial.
  - Electrolyte: 1.0 M KOH + 0.33 M urea in DI water, 20 mL per electrochemical task.
- Round 3 focus:
  - Carbon feed screen at the Round 2 lead composition.
  - Low-carbon composition neighbors around `Ni70Fe10Co20`.
  - Film uniformity improvement.
- Changed deposition:
  - XRD drop-cast: 150 uL.
  - FTIR drop-cast: 150 uL.
  - Electrochemical: 60 uL/drop x 5 drops, 60 s interval, total 300 uL like Round 2 but staged for better uniformity.
- Round 3 design files created/updated:
  - `/workspace/lab303_uor_nicofe_closed_loop/round3_design.md`
  - `/workspace/lab303_uor_nicofe_closed_loop/round3_chemical_specifications.json`
  - `/workspace/lab303_uor_nicofe_closed_loop/round3_constraint_check.md`
  - `/workspace/lab303_uor_nicofe_closed_loop/round3_workflow_request.json`

**Current Round 3 Formulations**
After workflow-generator failures, the design was revised for parser robustness:
- S1 `R3-S1_Ni70Fe10Co20_C5_80-20`: Ni:Fe:Co 70:10:20, carbon 5 mg, carbon dispersion 0.250 mL, water 2.850 mL
- S2 `R3-S2_Ni70Fe10Co20_C10_80-20_anchor`: 70:10:20, carbon 10 mg, carbon 0.500 mL, water 2.600 mL
- S3 `R3-S3_Ni70Fe10Co20_C15_80-20`: 70:10:20, carbon 15 mg, carbon 0.750 mL, water 2.350 mL
- S4 `R3-S4_Ni80Fe10Co10_C10_80-20`: 80:10:10, carbon 10 mg, carbon 0.500 mL, water 2.600 mL

- S5 `R3-S5_Ni70Fe20Co10_C10_80-20`: 70:20:10, carbon 10 mg, carbon 0.500 mL, water 2.600 mL
- S6 `R3-S6_Ni60Fe20Co20_C10_80-20`: 60:20:20, carbon 10 mg, carbon 0.500 mL, water 2.600 mL

**Workflow Generation Issue**
- Several attempts to generate a Round 3 template failed at `移液平台_1ml_V2` step 3 conversion.
- Even a diagnostic clone of known-good Round 2 workflow with only plan name changed failed at `移液平台_1ml_V2`, indicating current workflow-generator service issue with V2 liquid-addition conversion, not necessarily Round 3 chemistry.
- Failed outputs preserved:
  - `/workspace/lab303_uor_nicofe_closed_loop/round3_workflow_generator_output_failed1.txt`
  - `/workspace/lab303_uor_nicofe_closed_loop/round3_diagnostic_round2_clone_workflow_request.json`
  - `/workspace/lab303_uor_nicofe_closed_loop/round3_diagnostic_round2_clone_workflow_generator_output.txt`
  - Current failed output: `/workspace/lab303_uor_nicofe_closed_loop/round3_workflow_generator_output.txt`
- Last failure still at step 3, `移液平台_1ml_V2`, with "数据转换返回 None".

**Current Workaround In Progress**
- Began switching small-volume liquid-handling steps from `移液平台_1ml_V2` to `移液平台_1ml_V1`, because V1 supports open/add/close, 8 stock solutions, max 1 mL transfer, source total <=3 mL, and should cover exactly the 8 stock bottles.
- V1 skill/audit read:
  - `/root/.codex/skills/chemistry-experiment-workstation/references-Synthesis-Module/Liquid_Handling_Station_1ml_V1/SKILL.md`
  - `/root/.codex/skills/chemistry-experiment-workstation/references_audit/Liquid_Handling_Station_1ml_V1_audit.md`
- V1 workstation ID: `1418510906065920`
- V1 operation/parameter differences:
  - Open operation is still `开盖`, but parameter is `开盖的瓶号`, not `开盖编号`.
  - Close operation is still `关盖`, but parameter is `关盖的瓶号`, not `关盖编号`.
  - For V1, `瓶号` values are strings per skill table; safer to use `{"瓶号": "1"}` etc.
  - Add operation is `加液_物料绑定`, `加样瓶号` is string per skill, and supports `1 号原液瓶` through `8 号原液瓶`.
- Need to update `round3_workflow_request.json`:
  - Steps 2, 3, 5, 8 should use workstation `移液平台_1ml_V1`, id `1418510906065920`.
  - Step 2 parameter key should become `开盖的瓶号`.
  - Step 5 parameter key should become `关盖的瓶号`.
  - Step 8 parameter key should become `开盖的瓶号`.
  - For V1 open/close arrays, use string bottle numbers: `[{"瓶号":"1"}, ...]`.
  - In step 3 `加样方案`, convert `加样瓶号` to string values `"1"` etc.
  - Keep 8 stocks and current split-carbon volumes:
    - Bottle 7 carbon A total 1.51 mL
    - Bottle 8 carbon B total 1.49 mL
- Need to update `round3_constraint_check.md` to note V1 substitution and its parameters.
- Then rerun `jq empty` and source-volume checks.
- Then rerun workflow-generator.

**Potential Commands To Continue**
- Regenerate/patch workflow request for V1 substitution with a small Python script.
- Generate workflow:

```
python /root/.codex/skills/workflow-generator/scripts/generate.py "$(jq -c . /workspace/lab303_uor_nicofe_closed_loop/round3_workflow_request.json)" > /workspace/lab303_uor_nicofe_closed_loop/round3_workflow_generator_output.txt
```

- On success, parse template ID from output text.
- Dispatch:

```
python scripts/generate_task.py --template-id <template_id> --template-source-label 303Lab --app-label 303Lab > /workspace/lab303_uor_nicofe_closed_loop/round3_dispatch_output.json
```

  from `/root/.codex/skills/lab-operation`
- Create `/workspace/lab303_uor_nicofe_closed_loop/round3_dispatch_log.json`.
- Set goal blocked with `functions.update_goal {"status":"blocked"}`.
- Final response should include:
  - Round 2 data retrieval result and returned document path.
  - Key Round 2 findings and best condition.
  - Round 3 template/task IDs if dispatch succeeds.
  - If workflow generation remains blocked, explain exact blocker and saved error file.

**[Step 1039 | 2026-07-08T13:08:36.673Z | MODEL]**

I'll use the lab-operation, chemistry-workstation, and workflow-generator skills here: first to verify the returned Round 2 data already pulled into the workspace, then to finish the Round 3 workflow and dispatch it.

**[Step 1040 | 2026-07-08T13:08:36.675Z | TOOL CALL: get_goal]**

Queried the current autonomous-research goal state and remaining budget.

**[Step 1041 | 2026-07-08T13:08:36.677Z | TOOL CALL: exec_command]**

Read selected file content for inspection. Target(s): `SKILL.md`. Working directory: `/workspace`.

**[Step 1042 | 2026-07-08T13:08:36.678Z | TOOL CALL: exec_command]**

Read selected file content for inspection. Target(s): `SKILL.md`. Working directory: `/workspace`.

**[Step 1043 | 2026-07-08T13:08:36.699Z | TOOL CALL: exec_command]**

Read selected file content for inspection. Target(s): `SKILL.md`. Working directory: `/workspace`.

**[Step 1044 | 2026-07-08T13:08:36.736Z | TOOL RESULT: get_goal]**

Goal service returned a structured response. Status: `blocked`.

**[Step 1045 | 2026-07-08T13:08:36.736Z | TOOL RESULT: exec_command]**

Shell tool completed successfully; it returned 116 line(s) (4054 bytes) to the model.

**[Step 1046 | 2026-07-08T13:08:36.736Z | TOOL RESULT: exec_command]**

Shell tool completed successfully; it returned 170 line(s) (8332 bytes) to the model. Reported long numeric identifier(s): `1427568512205824`, `1834679312417792`, `1710453843264512`.

**[Step 1047 | 2026-07-08T13:08:36.737Z | TOOL RESULT: exec_command]**

Shell tool completed successfully; it returned 95 line(s) (2714 bytes) to the model. Reported long numeric identifier(s): `1427568512205824`.

**[Step 1048 | 2026-07-08T13:08:55.542Z | REASONING SUMMARY]**

***Reviewing lab details***

*I need to read the lab README and look over relevant workstations. It seems the workstation list comes from the round 3 request, so I'll make sure to check that. I think I'll quickly inspect the JSON to see what needs my attention. It's important to get a clear understanding of everything here to ensure I'm on the right track! Let's dive into that JSON.*

**[Step 1049 | 2026-07-08T13:08:57.598Z | MODEL]**

I've confirmed the goal is still recorded as blocked from the prior dispatch, which matches the handoff. I'm now checking the saved Round 3 workflow request and the exact workstation constraints before modifying it.

**[Step 1050 | 2026-07-08T13:08:57.599Z | TOOL CALL: exec_command]**

Read selected file content for inspection. Target(s): `README.md`. Working directory: `/workspace`.

**[Step 1051 | 2026-07-08T13:08:57.599Z | TOOL CALL: exec_command]**

Ran a shell command to inspect, generate or validate experiment artifacts. Target(s): `round3_workflow_request.json`. Working directory: `/workspace`.

**[Step 1052 | 2026-07-08T13:08:57.600Z | TOOL CALL: exec_command]**

Enumerated files, directories or available laboratory resources. Target(s): `lab303_uor_nicofe_closed_loop`. Working directory: `/workspace`.

**[Step 1053 | 2026-07-08T13:08:57.605Z | TOOL CALL: exec_command]**

Read selected file content for inspection. Target(s): `lab303_uor_nicofe_closed_loop`. Working directory: `/workspace`.

**[Step 1054 | 2026-07-08T13:08:57.652Z | TOOL RESULT: exec_command]**

Shell tool completed successfully; it returned 261 line(s) (7869 bytes) to the model.

**[Step 1055 | 2026-07-08T13:08:57.669Z | TOOL RESULT: exec_command]**

Shell tool completed successfully; it returned 310 line(s) (5768 bytes) to the model. Reported long numeric identifier(s): `1427568512205824`, `2002186824385539`, `1435269745574912`, `2270587971011585`, `1327807622448128`, `1213184425559040`.

**[Step 1056 | 2026-07-08T13:08:57.669Z | TOOL RESULT: exec_command]**

Shell tool completed successfully; it returned 68 line(s) (5206 bytes) to the model. Reported long numeric identifier(s): `2562915314040832`, `2567667771441152`, `2568957291102208`.

**[Step 1057 | 2026-07-08T13:08:57.670Z | TOOL RESULT: exec_command]**

Shell tool completed successfully; it returned 68 line(s) (5240 bytes) to the model. Reported long numeric identifier(s): `2562915314040832`, `2567667771441152`, `2568957291102208`.

**[Step 1058 | 2026-07-08T13:09:06.050Z | MODEL]**

The saved workflow still uses `移液平台_1ml_V2` for several small-volume operations, which matches the prior failure point. I'm switching only those operations to the supported `1ml_V1` parameter schema and leaving the chemistry unchanged.

**[Step 1059 | 2026-07-08T13:09:06.051Z | TOOL CALL: exec_command]**

Read selected file content for inspection. Target(s): `README.md`. Working directory: `/workspace`.

**[Step 1060 | 2026-07-08T13:09:06.052Z | TOOL CALL: exec_command]**

Read selected file content for inspection. Target(s): `SKILL.md`. Working directory: `/workspace`.

**[Step 1061 | 2026-07-08T13:09:06.053Z | TOOL CALL: exec_command]**

Read selected file content for inspection. Target(s): `Liquid_Handling_Station_1ml_V1_audit.md`. Working directory: `/workspace`.

**[Step 1062 | 2026-07-08T13:09:06.054Z | TOOL CALL: exec_command]**

Read selected file content for inspection. Target(s): `SKILL.md`. Working directory: `/workspace`.

**[Step 1063 | 2026-07-08T13:09:06.096Z | TOOL RESULT: exec_command]**

Shell tool completed successfully; it returned 166 line(s) (5566 bytes) to the model.

**[Step 1064 | 2026-07-08T13:09:06.097Z | TOOL RESULT: exec_command]**

Shell tool completed successfully; it returned 121 line(s) (8649 bytes) to the model. Reported long numeric identifier(s): `1418510906065920`.

**[Step 1065 | 2026-07-08T13:09:06.097Z | TOOL RESULT: exec_command]**

Shell tool completed successfully; it returned 46 line(s) (1329 bytes) to the model. Reported long numeric identifier(s): `1418510906065920`.

**[Step 1066 | 2026-07-08T13:09:06.102Z | TOOL RESULT: exec_command]**

Shell tool completed successfully; it returned 50 line(s) (4283 bytes) to the model. Reported long numeric identifier(s): `1435269745574912`.

**[Step 1067 | 2026-07-08T13:09:14.234Z | REASONING SUMMARY]**

***Checking completion parameters***

*I'm thinking I really need to read the rest of the 1ml V1 to make sure it's complete, probably around 240, right? I want to ensure everything is fully covered. It feels like it ended at a parameter that may not be complete. I definitely need to check the lines for word count to be sure I'm not missing anything here. It can be tricky sometimes!*

**[Step 1068 | 2026-07-08T13:09:17.656Z | MODEL]**

The `1ml_V1` schema confirms the parameter changes: `开盖的瓶号`/`关盖的瓶号` with string bottle IDs, and max eight stock bottles. That fits this Round 3 design if the carbon dispersion remains split across two bottles.

**[Step 1069 | 2026-07-08T13:09:17.657Z | TOOL CALL: exec_command]**

Inspected artifact type, size or record counts. Target(s): `SKILL.md`, `Liquid_Handling_Station_5ml_V1_audit.md`. Working directory: `/workspace`.

**[Step 1070 | 2026-07-08T13:09:17.658Z | TOOL CALL: exec_command]**

Read selected file content for inspection. Target(s): `Liquid_Handling_Station_5ml_V1_audit.md`. Working directory: `/workspace`.

**[Step 1071 | 2026-07-08T13:09:17.661Z | TOOL CALL: exec_command]**

Read selected file content for inspection. Target(s): `round3_design.md`. Working directory: `/workspace`.

**[Step 1072 | 2026-07-08T13:09:17.671Z | TOOL CALL: exec_command]**

Read selected file content for inspection. Target(s): `round3_constraint_check.md`. Working directory: `/workspace`.

**[Step 1073 | 2026-07-08T13:09:17.713Z | TOOL RESULT: exec_command]**

Shell tool completed successfully; it returned 5 line(s) (382 bytes) to the model.

**[Step 1074 | 2026-07-08T13:09:17.713Z | TOOL RESULT: exec_command]**

Shell tool completed successfully; it returned 16 line(s) (356 bytes) to the model. Reported long numeric identifier(s): `1435269745574912`.

**[Step 1075 | 2026-07-08T13:09:17.714Z | TOOL RESULT: exec_command]**

Shell tool completed successfully; it returned 72 line(s) (3863 bytes) to the model.

**[Step 1076 | 2026-07-08T13:09:17.714Z | TOOL RESULT: exec_command]**

Shell tool completed successfully; it returned 92 line(s) (5004 bytes) to the model.

**[Step 1077 | 2026-07-08T13:09:23.210Z | REASONING SUMMARY]**

***Updating patch constraints***

*I need to check the update constraints for V1 and think about using apply_patch. I should take a close look at the full workflow JSON, specifically at step 3, to see what needs to be patched. It sounds like using jq might help with that. I'm just making sure I cover all the details and get everything right for a smooth update!*

**[Step 1078 | 2026-07-08T13:09:25.042Z | MODEL]**

Before editing, I'm reading the actual JSON for the affected liquid-handling steps so the patch is mechanical: workstation ID/name, open/close key names, and string bottle numbers only.

**[Step 1079 | 2026-07-08T13:09:25.043Z | TOOL CALL: exec_command]**

Ran a shell command to inspect, generate or validate experiment artifacts. Target(s): `round3_workflow_request.json`. Working directory: `/workspace`.

**[Step 1080 | 2026-07-08T13:09:25.044Z | TOOL CALL: exec_command]**

Ran a shell command to inspect, generate or validate experiment artifacts. Target(s): `round3_workflow_request.json`. Working directory: `/workspace`.

**[Step 1081 | 2026-07-08T13:09:25.045Z | TOOL CALL: exec_command]**

Read selected file content for inspection. Target(s): `round3_workflow_generator_output.txt`. Working directory: `/workspace`.

**[Step 1082 | 2026-07-08T13:09:25.048Z | TOOL CALL: exec_command]**

Ran a shell command to inspect, generate or validate experiment artifacts. Target(s): `null`. Working directory: `/workspace`.

**[Step 1083 | 2026-07-08T13:09:25.107Z | TOOL RESULT: exec_command]**

Shell tool completed successfully; it returned 387 line(s) (12080 bytes) to the model. Reported long numeric identifier(s): `2002186824385539`.

**[Step 1084 | 2026-07-08T13:09:25.108Z | TOOL RESULT: exec_command]**

Shell tool completed with exit code 3; it returned 4 line(s) (302 bytes) to the model.

**[Step 1085 | 2026-07-08T13:09:25.108Z | TOOL RESULT: exec_command]**

Shell tool completed successfully; it returned 3 line(s) (8329 bytes) to the model. Reported long numeric identifier(s): `2002186824385539`.

**[Step 1086 | 2026-07-08T13:09:25.108Z | TOOL RESULT: exec_command]**

Shell tool completed successfully; it returned 3 line(s) (46 bytes) to the model.

**[Step 1087 | 2026-07-08T13:09:36.485Z | REASONING SUMMARY 1/3]**

***Evaluating patching options***

*I need to work on patching JSON files. The developer mentioned using `apply_patch` for manual edits, but it seems like other commands are okay without it. I wonder if `jq` can help me transform the JSON programmatically. However, I should avoid using Python for file read/write since simple shell commands or `apply_patch` could suffice. I could use a mechanical rewrite with `jq` but need to be careful about file handling. Options like `sponge` could also complicate things.*

**[Step 1088 | 2026-07-08T13:09:36.485Z | REASONING SUMMARY 2/3]**

***Considering file modification***

*I think I can use `jq` for command-based transformation, which seems okay. The user mentioned applying changes to all files in the `/workspace` directory, so I need to confirm that. Creating a backup before edits sounds prudent, maybe even making a copy. I'll modify specific parameters for certain steps and convert values to strings. I also need to check for missing keys, remove any extra ones like 样品标签, and make sure previous workflows weren't affected by those changes.*

**[Step 1089 | 2026-07-08T13:09:36.485Z | REASONING SUMMARY 3/3]**

***Investigating V2 failures***

*I'm wondering if the failures with V2 are actually related to labels. If V1's parser doesn't accept extras, we might need to remove the sample label and carbon extras because they could be causing issues. I should inspect the existing successful `round2_workflow_request` and its output since it used V2 and still succeeded. It's curious how it managed that, so I'll need to analyze it more to understand what worked there.*

**[Step 1090 | 2026-07-08T13:09:36.667Z | MODEL]**

I'm applying the schema-only patch now. The scientific inputs, volumes, electrolyte, redispersion solvent, and testing sequence are not being changed.

**[Step 1091 | 2026-07-08T13:09:39.441Z | TOOL CALL: exec_command]**

Ran a shell command to inspect, generate or validate experiment artifacts. Target(s): `round3_workflow_request.v1tmp.json`, `round3_workflow_request.json`. Working directory: `/workspace`.

**[Step 1092 | 2026-07-08T13:09:39.523Z | TOOL RESULT: exec_command]**

Shell tool completed successfully; it returned 398 line(s) (13387 bytes) to the model. Reported long numeric identifier(s): `2002186824385539`.

**[Step 1093 | 2026-07-08T13:09:45.207Z | REASONING SUMMARY]**

***Fixing unicode issues***

*I'm encountering issues with my commands due to unicode characters, like `.瓶号` and `.parameters.加样方案`. It seems I need to use bracket notation, like `.["瓶号"]`, for proper access. I also have to check if I've moved the file. Since the file remains unchanged because of the failure, I need to run a fix to address these unicode problems. Let's sort this out!*

**[Step 1094 | 2026-07-08T13:09:45.276Z | MODEL]**

The first JSON transform failed because `jq` needs bracket notation for these Chinese field names. The file was unchanged; I'm rerunning the same patch with bracket-safe selectors.

**[Step 1095 | 2026-07-08T13:09:48.058Z | TOOL CALL: exec_command]**

Ran a shell command to inspect, generate or validate experiment artifacts. Target(s): `round3_workflow_request.v1tmp.json`, `round3_workflow_request.json`. Working directory: `/workspace`.

**[Step 1096 | 2026-07-08T13:09:48.166Z | TOOL RESULT: exec_command]**

Shell tool completed successfully; it returned 387 line(s) (12097 bytes) to the model. Reported long numeric identifier(s): `1418510906065920`.

**[Step 1097 | 2026-07-08T13:09:54.757Z | REASONING SUMMARY]**

***Updating volume checks***

*I need to run some source volume checks, possibly using jq with Chinese field names in brackets. I might also want to validate that there's no V2 present and update the constraint check. I need to apply a patch and include a note about the V1 substitution. Additionally, I should update "Workstation skills read and used: Liquid_Handling_Station_1ml_V2" to reflect V1, and add a "Generator Revision Note" about the V1 substitution while applying the patch.*

**[Step 1098 | 2026-07-08T13:09:54.758Z | MODEL]**

The workflow JSON is now valid and the affected steps point to `移液平台_1ml_V1`. I'm updating the constraint-check document so the process record matches the executable workflow.

**[Step 1099 | 2026-07-08T13:10:00.491Z | TOOL CALL: apply_patch]**

Applied a source patch to update round3_constraint_check.md.

**[Step 1100 | 2026-07-08T13:10:00.534Z | TOOL RESULT: apply_patch]**

Patch tool applied the requested trial-local file change successfully.

**[Step 1101 | 2026-07-08T13:10:06.878Z | REASONING SUMMARY]**

***Considering design updates***

*I think I need to look at updating the round3_design. The "Generator Robustness Revision" mentions some failed attempts, but it doesn't reference V1, which seems like an important detail. Maybe I could patch it to ensure everything is clear and cohesive. Keeping everything accurate is crucial, so I'll want to ensure that this revision captures all necessary information properly. It feels like a small but significant update that could really help.*

**[Step 1102 | 2026-07-08T13:10:07.078Z | MODEL]**

The constraint record now reflects the V1 liquid-handling substitution and the eight-bottle limit. I'll do a direct volume audit next, then rerun workflow generation.

**[Step 1103 | 2026-07-08T13:10:08.496Z | TOOL CALL: apply_patch]**

Applied a source patch to update round3_design.md.

**[Step 1104 | 2026-07-08T13:10:08.538Z | TOOL RESULT: apply_patch]**

Patch tool applied the requested trial-local file change successfully.

**[Step 1105 | 2026-07-08T13:10:23.859Z | MODEL]**

The design note now makes clear that the V1 substitution is only a parser workaround. I'm running a final local audit on JSON validity, source bottle totals, and any lingering V2 references before calling the generator.

**[Step 1106 | 2026-07-08T13:10:23.860Z | TOOL CALL: exec_command]**

Ran a shell command to inspect, generate or validate experiment artifacts. Target(s): `round3_workflow_request.json`. Working directory: `/workspace`.

**[Step 1107 | 2026-07-08T13:10:23.861Z | TOOL CALL: exec_command]**

Searched files for relevant records, parameters or identifiers. Target(s): `round3_workflow_request.json`, `round3_design.md`, `round3_constraint_check.md`. Working directory: `/workspace`.

**[Step 1108 | 2026-07-08T13:10:23.864Z | TOOL CALL: exec_command]**

Ran a shell command to inspect, generate or validate experiment artifacts. Target(s): `round3_workflow_request.json`. Working directory: `/workspace`.

**[Step 1109 | 2026-07-08T13:10:23.865Z | TOOL CALL: exec_command]**

Ran a shell command to inspect, generate or validate experiment artifacts. Target(s): `round3_workflow_request.json`. Working directory: `/workspace`.

**[Step 1110 | 2026-07-08T13:10:23.924Z | TOOL RESULT: exec_command]**

Shell tool completed successfully; it returned 2 line(s) (11 bytes) to the model.

**[Step 1111 | 2026-07-08T13:10:23.924Z | TOOL RESULT: exec_command]**

Shell tool completed successfully; it returned 3 line(s) (1109 bytes) to the model.

**[Step 1112 | 2026-07-08T13:10:23.943Z | TOOL RESULT: exec_command]**

Shell tool completed successfully; it returned 9 line(s) (184 bytes) to the model.

**[Step 1113 | 2026-07-08T13:10:23.943Z | TOOL RESULT: exec_command]**

Shell tool completed successfully; it returned 3 line(s) (48 bytes) to the model.

**[Step 1114 | 2026-07-08T13:10:30.612Z | MODEL]**

The local checks pass: JSON is valid; `1ml_V1` source totals are 1.49-3.00 mL per bottle, within the station limit; the 5 mL station totals are 15.6 and 28.8 mL, also within limits. I'm preserving the previous failed generator output and rerunning generation now.

**[Step 1115 | 2026-07-08T13:10:32.653Z | TOOL CALL: exec_command]**

Ran Python to parse, transform or validate structured experiment data. Target(s): `round3_workflow_generator_output.txt`, `round3_workflow_generator_output_failed_v2_before_v1.txt`, `generate.py`, `round3_workflow_request.json)`. Working directory: `/workspace`.

**[Step 1116 | 2026-07-08T13:10:33.553Z | TOOL RESULT: exec_command]**

Shell tool completed successfully; it returned 3 line(s) (8341 bytes) to the model. Reported long numeric identifier(s): `1418510906065920`.

**[Step 1117 | 2026-07-08T13:10:41.102Z | MODEL]**

The generator still fails on the 1 mL addition step after the V1 swap, so the remaining issue is likely the payload shape rather than station identity. I'm comparing against the successfully generated Round 2 payload and removing non-schema metadata from that step if needed.

**[Step 1118 | 2026-07-08T13:10:41.103Z | TOOL CALL: exec_command]**

Read selected file content for inspection. Target(s): `round2_workflow_generator_output.txt`. Working directory: `/workspace`.

**[Step 1119 | 2026-07-08T13:10:41.103Z | TOOL CALL: exec_command]**

Ran a shell command to inspect, generate or validate experiment artifacts. Target(s): `round2_workflow_request.json`. Working directory: `/workspace`.

**[Step 1120 | 2026-07-08T13:10:41.106Z | TOOL CALL: exec_command]**

Ran a shell command to inspect, generate or validate experiment artifacts. Target(s): `round2_workflow_request.json`. Working directory: `/workspace`.

**[Step 1121 | 2026-07-08T13:10:41.107Z | TOOL CALL: exec_command]**

Ran a shell command to inspect, generate or validate experiment artifacts. Target(s): `round2_workflow_request.json`. Working directory: `/workspace`.

**[Step 1122 | 2026-07-08T13:10:41.145Z | TOOL RESULT: exec_command]**

Shell tool completed successfully; it returned 13 line(s) (308 bytes) to the model. Reported task/template identifier(s): `2566549749628928`.

**[Step 1123 | 2026-07-08T13:10:41.184Z | TOOL RESULT: exec_command]**

Shell tool completed successfully; it returned 270 line(s) (10436 bytes) to the model. Reported long numeric identifier(s): `2002186824385539`.

**[Step 1124 | 2026-07-08T13:10:41.185Z | TOOL RESULT: exec_command]**

Shell tool completed successfully; it returned 41 line(s) (1344 bytes) to the model.

**[Step 1125 | 2026-07-08T13:10:41.185Z | TOOL RESULT: exec_command]**

Shell tool completed successfully; it returned 11 line(s) (581 bytes) to the model.

**[Step 1126 | 2026-07-08T13:10:51.099Z | REASONING SUMMARY]**

***Testing command functionality***

*I'm thinking about testing a copy without needing to edit anything. It seems like I can use a jq transform with a command that I'll pass through. For input, I might try using `jq -c '...'` with a file. I just want to avoid anything that would require writing. I think I'll test it with `round3_workflow_generator_output_test_v1_int_bottles.txt`. Let's see how that goes!*

**[Step 1127 | 2026-07-08T13:10:51.371Z | MODEL]**

The successful Round 2 request used the same non-schema labels, so I'm testing the next most likely converter assumption: the service may still expect numeric `加样瓶号` even though the V1 document labels it as a string. This test does not change the saved workflow yet.

**[Step 1128 | 2026-07-08T13:10:52.682Z | TOOL CALL: exec_command]**

Ran Python to parse, transform or validate structured experiment data. Target(s): `generate.py`, `round3_workflow_request.json)`, `round3_workflow_generator_output_test_v1_numeric_sample_bottles.txt`. Working directory: `/workspace`.

**[Step 1129 | 2026-07-08T13:10:53.484Z | TOOL RESULT: exec_command]**

Shell tool completed successfully; it returned 3 line(s) (8329 bytes) to the model. Reported long numeric identifier(s): `1418510906065920`.

**[Step 1130 | 2026-07-08T13:11:18.411Z | REASONING SUMMARY 1/2]**

***Evaluating patch workflow***

*I'm thinking this process is more robust now, especially with the second round generated using 5ml steps. It started with 5ml of water for re-dispersion, and the 5ml added converter likely works. I need to patch the workflow; maybe I should remove step 3, which involves adding 1ml? I could repurpose step 3 to `移液平台_5ml_V1` and include six stock definitions: Ni, Fe, Co, KOH, carbon, and water.*

**[Step 1131 | 2026-07-08T13:11:18.411Z | REASONING SUMMARY 2/2]**

***Evaluating workflow steps***

*I'm considering how to manage steps in my workflow. Should I renumber subsequent steps since the workflow generator likes that? Maybe a new request with 22 steps is better. For example, I could set Step 2 as V1, Step 3 for adding 5ml of all reagents, and so on. But I wonder if I can modify the large JSON with jq effectively. I might have to replace parameters but keep operations the same, ensuring everything works smoothly.*

**[Step 1132 | 2026-07-08T13:11:19.993Z | MODEL]**

The service still returns `None` for 1 mL addition, even with numeric sample bottle IDs. I'm switching the synthesis-addition step to `移液平台_5ml_V1`, which can handle the same six chemical streams without split bottles: Ni, Fe, Co, KOH, carbon dispersion, and synthesis water. This is cleaner and stays within its 6-source, 5 mL-transfer, 30 mL-source limits.

**[Step 1133 | 2026-07-08T13:11:19.995Z | TOOL CALL: exec_command]**

Ran a shell command to inspect, generate or validate experiment artifacts. Target(s): `round2_workflow_request.json`. Working directory: `/workspace`.

**[Step 1134 | 2026-07-08T13:11:19.995Z | TOOL CALL: exec_command]**

Ran a shell command to inspect, generate or validate experiment artifacts. Target(s): `round2_workflow_request.json`. Working directory: `/workspace`.

**[Step 1135 | 2026-07-08T13:11:19.996Z | TOOL CALL: exec_command]**

Ran a shell command to inspect, generate or validate experiment artifacts. Target(s): `round2_workflow_request.json`. Working directory: `/workspace`.

**[Step 1136 | 2026-07-08T13:11:20.060Z | TOOL CALL: exec_command]**

Ran a shell command to inspect, generate or validate experiment artifacts. Target(s): `round3_workflow_request.json`. Working directory: `/workspace`.

**[Step 1137 | 2026-07-08T13:11:20.560Z | TOOL RESULT: exec_command]**

Shell tool completed successfully; it returned 69 line(s) (1882 bytes) to the model.

**[Step 1138 | 2026-07-08T13:11:20.561Z | TOOL RESULT: exec_command]**

Shell tool completed successfully; it returned 75 line(s) (2222 bytes) to the model.

[Step 1139 | 2026-07-08T13:11:20.562Z | TOOL RESULT: exec_command]

Shell tool completed successfully; it returned 6 line(s) (107 bytes) to the model. Reported long numeric identifier(s): `1435269745574912`.

[Step 1140 | 2026-07-08T13:11:20.562Z | TOOL RESULT: exec_command]

Shell tool completed with exit code 3; it returned 14 line(s) (2172 bytes) to the model.

**[Step 1141 | 2026-07-08T13:11:41.812Z | MODEL]**

I'm changing the workflow request so synthesis charging is one 5 mL liquid-handling addition step. The actual per-vial feeds stay the same; split stock bottles are removed because the 5 mL station has enough per-source capacity.

[Step 1142 | 2026-07-08T13:11:41.813Z | TOOL CALL: exec_command]

Read selected file content for inspection. Target(s): `round3_chemical_specifications.json`. Working directory: `/workspace`.

[Step 1143 | 2026-07-08T13:11:41.814Z | TOOL CALL: exec_command]

Ran a shell command to inspect, generate or validate experiment artifacts. Target(s): `round3_workflow_request.json`. Working directory: `/workspace`.

[Step 1144 | 2026-07-08T13:11:41.814Z | TOOL CALL: exec_command]

Ran a shell command to inspect, generate or validate experiment artifacts. Target(s): `round3_workflow_request.json`. Working directory: `/workspace`.

[Step 1145 | 2026-07-08T13:11:41.815Z | TOOL CALL: exec_command]

Ran a shell command to inspect, generate or validate experiment artifacts. Target(s): `round3_workflow_request.json`. Working directory: `/workspace`.

[Step 1146 | 2026-07-08T13:11:41.887Z | TOOL RESULT: exec_command]

Shell tool completed successfully; it returned 68 line(s) (2753 bytes) to the model.

[Step 1147 | 2026-07-08T13:11:41.888Z | TOOL RESULT: exec_command]

Shell tool completed successfully; it returned 2 line(s) (3 bytes) to the model.

[Step 1148 | 2026-07-08T13:11:41.888Z | TOOL RESULT: exec_command]

Shell tool completed successfully; it returned 139 line(s) (2720 bytes) to the model. Reported long numeric identifier(s): `1427568512205824`, `1418510906065920`, `1435269745574912`, `2270587971011585`, `1327807622448128`, `1213184425559040`.

[Step 1149 | 2026-07-08T13:11:41.899Z | TOOL RESULT: exec_command]

Shell tool completed successfully; it returned 2 line(s) (64 bytes) to the model.

**[Step 1150 | 2026-07-08T13:11:48.458Z | REASONING SUMMARY]**

***Planning chemical specs update***

*I'm thinking about updating some chemical specs, but I need to transform a request first. I'll consider using jq for that. It seems like a plan update might not be necessary, but I'll go ahead and run the jq transform anyway. It feels like it's better to get that done first, so I can figure out the specs afterward. I want to make sure I'm on the right track!*

**[Step 1151 | 2026-07-08T13:11:48.659Z | MODEL]**

I'm rewriting the request to remove the problematic 1 mL addition step and merge the old synthesis-water step into a single 5 mL addition step. Then I'll renumber the workflow so the generator receives a clean sequence.

[Step 1152 | 2026-07-08T13:12:03.861Z | TOOL CALL: exec_command]

Ran a shell command to inspect, generate or validate experiment artifacts. Target(s): `round3_workflow_request.5mltmp.json`, `mL`, `v`, `round3_workflow_request.json`. Working directory: `/workspace`.

[Step 1153 | 2026-07-08T13:12:03.986Z | TOOL RESULT: exec_command]

Shell tool completed successfully; it returned 134 line(s) (2605 bytes) to the model. Reported long numeric identifier(s): `1427568512205824`, `1418510906065920`, `1435269745574912`, `2270587971011585`, `1327807622448128`, `1213184425559040`.

**[Step 1154 | 2026-07-08T13:12:15.256Z | REASONING SUMMARY]**

***Organizing workstation checks***

*I'm considering our workflow for all workstations based on previous files. To stay compliant, I should read the selected workstation skill files and audits now since there are quite a few. I might use parallel processing, but not too extensively. It's critical to inspect the file names and create a list. I need to check the word count first before proceeding, especially as the output could be quite large.*

**[Step 1155 | 2026-07-08T13:12:18.510Z | MODEL]**

The workflow now has 22 ordered steps, with synthesis charging as a single 5 mL addition step. I'm checking all workstation reference files and audits that govern this updated workflow before generation.

[Step 1156 | 2026-07-08T13:12:18.510Z | TOOL CALL: exec_command]

Inspected artifact type, size or record counts. Target(s): `SKILL.md`. Working directory: `/workspace`.

[Step 1157 | 2026-07-08T13:12:18.511Z | TOOL CALL: exec_command]

Copied generated experiment artifacts within the workspace. Target(s): `SKILL.md`. Working directory: `/workspace`.

**[Step 1158 | 2026-07-08T13:12:18.516Z | TOOL CALL: exec_command]**
Inspected artifact type, size or record counts. Target(s): `SKILL.md`. Working directory: `/workspace`.

**[Step 1159 | 2026-07-08T13:12:18.517Z | TOOL CALL: exec_command]**
Read selected file content for inspection. Target(s): `references_audit`. Working directory: `/workspace`.

**[Step 1160 | 2026-07-08T13:12:18.566Z | TOOL RESULT: exec_command]**
Shell tool completed successfully; it returned 6 line(s) (503 bytes) to the model.

**[Step 1161 | 2026-07-08T13:12:18.566Z | TOOL RESULT: exec_command]**
Shell tool completed successfully; it returned 6 line(s) (524 bytes) to the model.

**[Step 1162 | 2026-07-08T13:12:18.566Z | TOOL RESULT: exec_command]**
Shell tool completed successfully; it returned 6 line(s) (544 bytes) to the model.

**[Step 1163 | 2026-07-08T13:12:18.566Z | TOOL RESULT: exec_command]**
Shell tool completed successfully; it returned 47 line(s) (5185 bytes) to the model.

**[Step 1164 | 2026-07-08T13:12:27.960Z | MODEL]**
The selected workstation files are small enough to verify directly. I'm reading the full instruction and audit files for the synthesis, characterization, and electrochemical steps now.

**[Step 1165 | 2026-07-08T13:12:27.961Z | TOOL CALL: exec_command]**
Read selected file content for inspection. Target(s): `SKILL.md`, `General_Material_Station_V1_audit.md`. Working directory: `/workspace`.

**[Step 1166 | 2026-07-08T13:12:27.962Z | TOOL CALL: exec_command]**
Read selected file content for inspection. Target(s): `SKILL.md`, `Heating_Magnetic_Stirring_Workstation_V1_audit.md`. Working directory: `/workspace`.

**[Step 1167 | 2026-07-08T13:12:27.963Z | TOOL CALL: exec_command]**
Read selected file content for inspection. Target(s): `SKILL.md`, `Purification_Workstation_V1_audit.md`. Working directory: `/workspace`.

**[Step 1168 | 2026-07-08T13:12:27.965Z | TOOL CALL: exec_command]**
Read selected file content for inspection. Target(s): `SKILL.md`, `Drying_Oven_V1_audit.md`. Working directory: `/workspace`.

**[Step 1169 | 2026-07-08T13:12:28.007Z | TOOL RESULT: exec_command]**
Shell tool completed successfully; it returned 58 line(s) (1911 bytes) to the model. Reported long numeric identifier(s): `1427568512205824`.

**[Step 1170 | 2026-07-08T13:12:28.008Z | TOOL RESULT: exec_command]**
Shell tool completed successfully; it returned 55 line(s) (2213 bytes) to the model. Reported long numeric identifier(s): `2270587971011585`.

**[Step 1171 | 2026-07-08T13:12:28.016Z | TOOL RESULT: exec_command]**
Shell tool completed successfully; it returned 65 line(s) (3183 bytes) to the model. Reported long numeric identifier(s): `1327807622448128`.

**[Step 1172 | 2026-07-08T13:12:28.016Z | TOOL RESULT: exec_command]**
Shell tool completed successfully; it returned 53 line(s) (1740 bytes) to the model. Reported long numeric identifier(s): `1213184425559040`.

**[Step 1173 | 2026-07-08T13:12:38.778Z | MODEL]**
The synthesis-side constraints still fit: material station outputs capped vials, the 5 mL station accepts open vials and all six sources, heating and purification require capped vials, and drying accepts the opened post-purification vials. Continuing with spectroscopy and testing references.

**[Step 1174 | 2026-07-08T13:12:38.778Z | TOOL CALL: exec_command]**
Read selected file content for inspection. Target(s): `SKILL.md`, `Ultrasonic_Disperser_V1_audit.md`. Working directory: `/workspace`.

**[Step 1175 | 2026-07-08T13:12:38.779Z | TOOL CALL: exec_command]**
Read selected file content for inspection. Target(s): `SKILL.md`, `Spectroscopy_Container_Transfer_Station_V1_audit.md`. Working directory: `/workspace`.

**[Step 1176 | 2026-07-08T13:12:38.782Z | TOOL CALL: exec_command]**
Read selected file content for inspection. Target(s): `SKILL.md`, `Spectroscopy_Magnetic_Stirrer_Workstation_V1_audit.md`. Working directory: `/workspace`.

**[Step 1177 | 2026-07-08T13:12:38.782Z | TOOL CALL: exec_command]**
Read selected file content for inspection. Target(s): `SKILL.md`, `XRD_V1_audit.md`. Working directory: `/workspace`.

**[Step 1178 | 2026-07-08T13:12:38.826Z | TOOL RESULT: exec_command]**
Shell tool completed successfully; it returned 57 line(s) (1921 bytes) to the model. Reported long numeric identifier(s): `1526864804185088`.

**[Step 1179 | 2026-07-08T13:12:38.827Z | TOOL RESULT: exec_command]**
Shell tool completed successfully; it returned 72 line(s) (3667 bytes) to the model. Reported long numeric identifier(s): `1654567694238720`.

**[Step 1180 | 2026-07-08T13:12:38.833Z | TOOL RESULT: exec_command]**
Shell tool completed successfully; it returned 50 line(s) (1674 bytes) to the model. Reported long numeric identifier(s): `1656241422599168`.

**[Step 1181 | 2026-07-08T13:12:38.833Z | TOOL RESULT: exec_command]**

Shell tool completed successfully; it returned 81 line(s) (4867 bytes) to the model. Reported long numeric identifier(s): `1664870018155520`.

**[Step 1182 | 2026-07-08T13:12:50.056Z | MODEL]**

Spectroscopy constraints line up: XRD/FTIR have the required spectroscopy transfer and magnetic-stir steps, and the redispersion volume remains above the 4 mL minimum before each drop-cast. I'm reading the remaining FTIR, electrochemical, UV-Vis, and final storage references.

**[Step 1183 | 2026-07-08T13:12:50.057Z | TOOL CALL: exec_command]**

Read selected file content for inspection. Target(s): `SKILL.md`, `Infrared_Spectrometer_V1_audit.md`. Working directory: `/workspace`.

**[Step 1184 | 2026-07-08T13:12:50.057Z | TOOL CALL: exec_command]**

Read selected file content for inspection. Target(s): `SKILL.md`, `Dual_Station_Electrochemical_Workstation_V2_audit.md`. Working directory: `/workspace`.

**[Step 1185 | 2026-07-08T13:12:50.057Z | TOOL CALL: exec_command]**

Read selected file content for inspection. Target(s): `SKILL.md`, `UV_Vis_Spectrometer_V1_audit.md`. Working directory: `/workspace`.

**[Step 1186 | 2026-07-08T13:12:50.058Z | TOOL CALL: exec_command]**

Read selected file content for inspection. Target(s): `SKILL.md`, `Container_storaging_Station_V1_audit.md`. Working directory: `/workspace`.

**[Step 1187 | 2026-07-08T13:12:50.094Z | TOOL RESULT: exec_command]**

Shell tool completed successfully; it returned 64 line(s) (2698 bytes) to the model. Reported long numeric identifier(s): `1656282895188992`.

**[Step 1188 | 2026-07-08T13:12:50.105Z | TOOL RESULT: exec_command]**

Shell tool completed successfully; it returned 128 line(s) (24418 bytes) to the model. Reported long numeric identifier(s): `1390252863751168`, `1509591942430720`, `1549499770798080`, `1571299393601536`, `1339275293851648`, `2164305671290880`.

**[Step 1189 | 2026-07-08T13:12:50.106Z | TOOL RESULT: exec_command]**

Shell tool completed successfully; it returned 55 line(s) (1852 bytes) to the model. Reported long numeric identifier(s): `1664783667397632`.

**[Step 1190 | 2026-07-08T13:12:50.122Z | TOOL RESULT: exec_command]**

Shell tool completed successfully; it returned 76 line(s) (2927 bytes) to the model. Reported long numeric identifier(s): `1421325849887744`.

**[Step 1191 | 2026-07-08T13:13:00.836Z | MODEL]**

The updated workflow stays inside the station constraints. I'm bringing the chemical specification and design/constraint notes into sync with the merged 5 mL synthesis-addition step before generation.

**[Step 1192 | 2026-07-08T13:13:09.705Z | TOOL CALL: apply_patch]**

Applied a source patch to update round3_chemical_specifications.json.

**[Step 1193 | 2026-07-08T13:13:09.745Z | TOOL RESULT: apply_patch]**

Patch tool applied the requested trial-local file change successfully.

**[Step 1194 | 2026-07-08T13:13:20.655Z | TOOL CALL: apply_patch]**

Applied a source patch to update round3_design.md.

**[Step 1195 | 2026-07-08T13:13:20.699Z | TOOL RESULT: apply_patch]**

Patch tool applied the requested trial-local file change successfully.

**[Step 1196 | 2026-07-08T13:13:30.305Z | REASONING SUMMARY]**

***Updating step table***

*I need to check the patch constraints against the current step table and source totals. The existing table has steps numbered 1-23, but I should update it to reflect steps 1-22. Also, I need to remove the 1ml add. I want to approach this patching process with caution to avoid any unintended issues. I'll use sed to make the necessary updates. It's important to be precise here!*

**[Step 1197 | 2026-07-08T13:13:30.999Z | MODEL]**

The chemical spec now states the exact six synthesis stock bottles for the 5 mL station and that the 1 mL station is open/close only. I'm updating the constraint audit table for the new 22-step sequence.

**[Step 1198 | 2026-07-08T13:13:33.600Z | TOOL CALL: exec_command]**

Read selected file content for inspection. Target(s): `round3_constraint_check.md`. Working directory: `/workspace`.

**[Step 1199 | 2026-07-08T13:13:33.638Z | TOOL RESULT: exec_command]**

Shell tool completed successfully; it returned 94 line(s) (5520 bytes) to the model.

**[Step 1200 | 2026-07-08T13:13:41.090Z | MODEL]**

The existing audit still reflects the earlier split-bottle route. I'm replacing that section with the final 5 mL synthesis-addition route and recalculated source totals.

**[Step 1201 | 2026-07-08T13:13:56.076Z | TOOL CALL: apply_patch]**

Applied a source patch to update round3_constraint_check.md.

**[Step 1202 | 2026-07-08T13:13:56.127Z | TOOL RESULT: apply_patch]**

Patch tool applied the requested trial-local file change successfully.

**[Step 1203 | 2026-07-08T13:14:11.586Z | REASONING SUMMARY]**

***Evaluating workflow issues***

*I need to address the station source bottles across the workflow, as it seems they're not fixed. Round2 is using a 5ml station, but our steps have specific bottle requirements. For instance, step3 uses bottles 1-6, and I've discovered that step9 can't use the redispersion from bottle1 due to nickel, nor bottle2 because of iron. With a maximum of 6 bottles, I wonder if using deionized water for redispersion could work, but it doesn't match the expected solvent. I need to adjust my approach here.*

**[Step 1204 | 2026-07-08T13:14:13.440Z | MODEL]**

One issue surfaced in the audit: `移液平台_5ml_V1` source-bottle numbering should remain fixed across the workflow, so using all six bottles for synthesis leaves no slot for the later 80:20 redispersion solvent on that same station. I'm checking an alternate dispensing station for the redispersion step so the source-bottle rule stays clean.

**[Step 1205 | 2026-07-08T13:14:13.440Z | TOOL CALL: exec_command]**

Read selected file content for inspection. Target(s): `SKILL.md`. Working directory: `/workspace`.

**[Step 1206 | 2026-07-08T13:14:13.445Z | TOOL CALL: exec_command]**

Read selected file content for inspection. Target(s): `Cleaning_and_Dispensing_Workstation_V1_audit.md`. Working directory: `/workspace`.

**[Step 1207 | 2026-07-08T13:14:13.446Z | TOOL CALL: exec_command]**

Read selected file content for inspection. Target(s): `SKILL.md`. Working directory: `/workspace`.

**[Step 1208 | 2026-07-08T13:14:13.446Z | TOOL CALL: exec_command]**

Read selected file content for inspection. Target(s): `Liquid_Handling_Station_4Channel_V1_audit.md`. Working directory: `/workspace`.

**[Step 1209 | 2026-07-08T13:14:13.498Z | TOOL RESULT: exec_command]**

Shell tool completed successfully; it returned 42 line(s) (2663 bytes) to the model. Reported long numeric identifier(s): `2359193660589057`.

**[Step 1210 | 2026-07-08T13:14:13.498Z | TOOL RESULT: exec_command]**

Shell tool completed successfully; it returned 16 line(s) (371 bytes) to the model. Reported long numeric identifier(s): `2359193660589057`.

**[Step 1211 | 2026-07-08T13:14:13.498Z | TOOL RESULT: exec_command]**

Shell tool completed successfully; it returned 55 line(s) (3187 bytes) to the model. Reported long numeric identifier(s): `2013400103584768`.

**[Step 1212 | 2026-07-08T13:14:13.499Z | TOOL RESULT: exec_command]**

Shell tool completed successfully; it returned 18 line(s) (375 bytes) to the model. Reported long numeric identifier(s): `2013400103584768`.

**[Step 1213 | 2026-07-08T13:14:29.776Z | REASONING SUMMARY 1/2]**

***Exploring synthesis options***

*I'm considering using Cleaning and Dispensing for the synthesis of water. This can add water to open vials and caps them afterwards. However, after charging before heating, we need to ensure it's closed properly. It can output capped after adding water, allowing for direct heating if all reagents are added first. If we're cleaning after a 5ml addition, it caps as well. This approach seems good and worth documenting!*

**[Step 1214 | 2026-07-08T13:14:29.776Z | REASONING SUMMARY 2/2]**

***Looking into cleaning steps***

*I'm considering the cleaning process for step 4, which adds synthesis water and outputs capped, so there's no need to close that step. The 5ml station source bottles might not require redispersion, but if they do, I can use bottle 6 later. I need to check if multiple channels can add different materials, as suggested in the parameters. It looks like there may be some flexibility with water volumes and variable channels. I should search for examples to clarify.*

**[Step 1215 | 2026-07-08T13:14:30.874Z | MODEL]**

A better route is to keep `5ml_V1` fixed as Ni/Fe/Co/KOH/carbon/redispersion and move only synthesis water to the batch dispensing station. That keeps source-bottle identity fixed on the 5 mL station and still caps the vials before heating. I'm checking for examples of the batch-dispensing parameter shape.

**[Step 1216 | 2026-07-08T13:14:30.875Z | TOOL CALL: exec_command]**

Searched files for relevant records, parameters or identifiers. Target(s): `skills`. Working directory: `/workspace`.

**[Step 1217 | 2026-07-08T13:14:30.876Z | TOOL CALL: exec_command]**

Read selected file content for inspection. Target(s): `references_files`. Working directory: `/workspace`.

**[Step 1218 | 2026-07-08T13:14:30.878Z | TOOL CALL: exec_command]**

Enumerated files, directories or available laboratory resources. Target(s): `chemistry-experiment-workstation`. Working directory: `/workspace`.

**[Step 1219 | 2026-07-08T13:14:30.879Z | TOOL CALL: exec_command]**

Ran a shell command to inspect, generate or validate experiment artifacts. Target(s): `lab303_uor_nicofe_closed_loop`, `null`. Working directory: `/workspace`.

**[Step 1220 | 2026-07-08T13:14:31.030Z | TOOL RESULT: exec_command]**

Shell tool completed successfully; it returned 11 line(s) (2252 bytes) to the model.

**[Step 1221 | 2026-07-08T13:14:31.030Z | TOOL RESULT: exec_command]**

Shell tool completed successfully; it returned 13 line(s) (1588 bytes) to the model.

**[Step 1222 | 2026-07-08T13:14:31.031Z | TOOL RESULT: exec_command]**

Shell tool completed successfully; it returned 3 line(s) (238 bytes) to the model.

**[Step 1223 | 2026-07-08T13:14:31.031Z | TOOL RESULT: exec_command]**

Shell tool completed successfully; it returned 0 line(s) (0 bytes) to the model.

**[Step 1224 | 2026-07-08T13:14:42.555Z | REASONING SUMMARY]**

***Evaluating liquid handling options***

*I'm considering using `Liquid_Handling_Station_5ml_V1` for synthesis with water and using `Cleaning` for redispersion, but I'm weighing the risks. The 5ml generator is reliable, while cleaning is uncertain. If the cleaning outputs are capped, I can handle water without extra steps. However, if redispersion after drying needs to be capped, it complicates the process. Using `Ultrasonic_Liquid_Handling_Workstation_V1` for redispersion could streamline things, but I need to look into its file template compatibility.*

**[Step 1225 | 2026-07-08T13:14:43.953Z | MODEL]**

There is no saved example for the batch dispensing station, so I'm choosing the lower-risk use case: batch dispensing only DI water before heating. Its output is capped, which is exactly the state needed for heating and purification, and it avoids adding an extra reopen step later.

**[Step 1226 | 2026-07-08T13:14:43.955Z | TOOL CALL: exec_command]**

Read selected file content for inspection. Target(s): `SKILL.md`. Working directory: `/workspace`.

**[Step 1227 | 2026-07-08T13:14:43.956Z | TOOL CALL: exec_command]**

Read selected file content for inspection. Target(s): `Ultrasonic_Liquid_Handling_Workstation_V1_audit.md`. Working directory: `/workspace`.

**[Step 1228 | 2026-07-08T13:14:43.957Z | TOOL CALL: exec_command]**

Read selected file content for inspection. Target(s): `Ultrasonic_Liquid_Handling_Workstation_V1_`. Working directory: `/workspace`.

**[Step 1229 | 2026-07-08T13:14:43.958Z | TOOL CALL: exec_command]**

Ran a shell command to inspect, generate or validate experiment artifacts. Target(s): `round3_workflow_request.json`. Working directory: `/workspace`.

**[Step 1230 | 2026-07-08T13:14:44.020Z | TOOL RESULT: exec_command]**

Shell tool completed successfully; it returned 76 line(s) (4555 bytes) to the model. Reported long numeric identifier(s): `2061540603593728`.

**[Step 1231 | 2026-07-08T13:14:44.020Z | TOOL RESULT: exec_command]**

Shell tool completed successfully; it returned 26 line(s) (750 bytes) to the model. Reported long numeric identifier(s): `2061540603593728`.

**[Step 1232 | 2026-07-08T13:14:44.020Z | TOOL RESULT: exec_command]**

Shell tool completed successfully; it returned 6 line(s) (173 bytes) to the model.

**[Step 1233 | 2026-07-08T13:14:44.029Z | TOOL RESULT: exec_command]**

Shell tool completed successfully; it returned 73 line(s) (2055 bytes) to the model.

**[Step 1234 | 2026-07-08T13:15:40.069Z | REASONING SUMMARY 1/7]**

***Assessing liquid handling***

*It seems like "加样通道" might refer to a container number, possibly ranging from "1" to "6" for sample channels. I'm considering a potential issue with the output from cleaning being capped, which is noted with 盖. If cap vials heat properly, that's good. However, if cleaning introduces water via channels for all containers, I wonder about the maximum limit of three different liquids. Perhaps I should run a generator for this, but if cleaning fails, I need an alternate solution. If I can use `Cleaning_and_Dispensing` for all, it still supports only three liquids, so splitting synthesis could help.*

**[Step 1235 | 2026-07-08T13:15:40.069Z | REASONING SUMMARY 2/7]**

***Exploring synthesis options***

*I'm considering using 5ml for synthesis with a potential redefinition of bottle2 for redispersion, although I worry that might create a logic issue the user could reject. Maybe I should think about using two different 5ml stations instead. There might be one V1 instance, but if "工作站" has the same ID, could there be a 5ml_V2? Should I use a 1ml station for open/close only? I could also consider `Liquid_Handling_Station_5ml_V3`, but that only supports 10ml pressure tubes. If I modify the scientific design to use DI water, that could serve both synthesis and redispersion, though it raises concerns about the mixing ratio. Revising to DI redispersion might alter the experiment significantly, but maybe it's feasible if I specify the "water-to-isopropanol mixing ratio."*

**[Step 1236 | 2026-07-08T13:15:40.069Z | REASONING SUMMARY 3/7]**

***Rethinking dispersion methods***

*I'm thinking that using the same source for redispersion and synthesis might not be ideal since hydrothermal or precipitation processes in IPA-water could change the chemistry, especially with KOH involved. Perhaps I should consider making a carbon dispersion in DI water instead of IPA. However, using bottle5 for both carbon dispersion doesn't seem feasible either. I need to keep evaluating the best approach to ensure the chemistry remains stable throughout the process.*

**[Step 1237 | 2026-07-08T13:15:40.069Z | REASONING SUMMARY 4/7]**

***Evaluating carbon dispersion methods***

*I'm considering creating carbon dispersions with varying concentrations while keeping the total carbon volume constant at 0.5 mL and water at 2.6 mL. My sources seem to still be Ni, Fe, Co, KOH, and carbon. Since using different carbon concentrations could lead to an overwhelming number of source combinations, I could focus on just one concentration, like 10 mg. Also, I'm thinking about using a stock solution for redispersion to simplify my approach, maybe using a water and IPA mixture as the solvent.*

**[Step 1238 | 2026-07-08T13:15:40.069Z | REASONING SUMMARY 5/7]**

***Considering carbon dispersion feasibility***

*I already have a carbon dispersion with an 80:20 ratio using KOH and metals in aqueous solutions. The total amount of IPA in my synthesis isn't too large, which makes me wonder about its feasibility. Additionally, since the carbon black dispersion already contains IPA, it seems I could leverage that in my approach. I'll need to confirm if this mixture would work effectively while ensuring I maintain the necessary properties for my setup.*

**[Step 1239 | 2026-07-08T13:15:40.069Z | REASONING SUMMARY 6/7]**

***Analyzing synthesis procedure***

*I'm processing volumes for a solution: starting with 0.9 mL aqueous and varying carbon levels, totaling different IPA concentrations. I've noted some constants that might impact precipitation, although maybe they're acceptable. The second round of synthesis appears to introduce water and carbon differently, which could influence final outputs. Since higher IPA concentrations could affect synthesis and recovery, I'm considering adjustments like using batch dispensing for DI water. I'll implement and test, ready to adjust if needed.*

**[Step 1240 | 2026-07-08T13:15:40.069Z | REASONING SUMMARY 7/7]**

***Transforming processing steps***

*I'm using jq to transform the current steps for my process. For Step 3, I'll include six water units and update my definitions for elements and materials involved. In Step 4, I'll replace the closing step with a cleaning operation since closing isn't necessary. I'm also pondering whether Step 2 should keep its caps after retention. It seems that outputs might be capped automatically, so I'll adjust accordingly. After cleaning, I'll ensure everything is capped and ready for further processing. I'll patch the workflow for clarity!*

**[Step 1241 | 2026-07-08T13:15:40.236Z | MODEL]**

I'm patching the final workflow route now: 5 mL station handles Ni/Fe/Co/KOH/carbon and later redispersion using fixed bottle IDs; batch dispensing adds the variable DI water and closes the vials before heating.

**[Step 1242 | 2026-07-08T13:15:45.387Z | TOOL CALL: exec_command]**

Ran a shell command to inspect, generate or validate experiment artifacts. Target(s): `round3_workflow_request.fixed_sources_tmp.json`, `mL`, `v`, `round3_workflow_request.json`. Working directory: `/workspace`.

**[Step 1243 | 2026-07-08T13:15:45.488Z | TOOL RESULT: exec_command]**

Shell tool completed successfully; it returned 377 line(s) (11639 bytes) to the model. Reported long numeric identifier(s): `1435269745574912`, `2359193660589057`.

**[Step 1244 | 2026-07-08T13:15:52.292Z | REASONING SUMMARY]**

***Updating chemical specifications***

*I need to revise the chemical specifications to a fixed volume of 5ml and make sure to adjust the batch water accordingly. I'll create a patch for this update. It's essential to ensure those details are accurate to maintain consistency in the process. Making these adjustments will help improve the overall quality and performance of the batch. Time to dive into the specifics and get this sorted out!*

**[Step 1245 | 2026-07-08T13:15:52.435Z | MODEL]**

The workflow now preserves 5 mL source-bottle identity across both 5 mL steps: bottle 6 is reserved for 80:20 redispersion, while DI water is handled by the batch dispensing station. I'm syncing the supporting files and then I'll test generation.

**[Step 1246 | 2026-07-08T13:15:56.327Z | TOOL CALL: apply_patch]**

Applied a source patch to update round3_chemical_specifications.json.

**[Step 1247 | 2026-07-08T13:15:56.375Z | TOOL RESULT: apply_patch]**

Patch tool applied the requested trial-local file change successfully.

**[Step 1248 | 2026-07-08T13:16:05.339Z | TOOL CALL: apply_patch]**

Applied a source patch to update round3_design.md.

**[Step 1249 | 2026-07-08T13:16:05.385Z | TOOL RESULT: apply_patch]**

Patch tool applied the requested trial-local file change successfully.

**[Step 1250 | 2026-07-08T13:16:20.277Z | REASONING SUMMARY]**

***Patching constraints for cleaning***

*I need to add "Cleaning" to the list and ensure it appears in the table/source totals and in the general note. It seems there might be an issue with the current table stating that step 4 is closed incorrectly. So, I'll make sure to patch this. It's crucial to keep everything accurate and updated, so this addition will help clarify and resolve any confusion. I'll take care of that now!*

**[Step 1251 | 2026-07-08T13:16:20.942Z | MODEL]**

I'm adding `批量加液工作站_V1` to the formal constraint record and correcting the source-volume table to distinguish fixed 5 mL bottles from batch-dispensed DI water.

**[Step 1252 | 2026-07-08T13:16:26.981Z | TOOL CALL: apply_patch]**

Applied a source patch to update round3_constraint_check.md.

**[Step 1253 | 2026-07-08T13:16:27.017Z | TOOL RESULT: apply_patch]**

Patch tool applied the requested trial-local file change successfully.

**[Step 1254 | 2026-07-08T13:16:34.210Z | REASONING SUMMARY]**

***Validating JSON and calculating totals***

*I need to validate the JSON to ensure it's correct. Then I'll compute the source totals and think about cleaning the volumes. I plan to use jq for this part. After that, it seems I should also check the final table to make sure everything looks good. Running the generator might be the next step once I confirm all the data is in order. Let's make sure I cover all bases here!*

**[Step 1255 | 2026-07-08T13:16:36.277Z | MODEL]**

The documentation now matches the workflow route. I'm doing one more local audit on step ordering, source totals, and electrolyte fields before submitting to the workflow generator.

**[Step 1256 | 2026-07-08T13:16:36.277Z | TOOL CALL: exec_command]**

Ran a shell command to inspect, generate or validate experiment artifacts. Target(s): `round3_workflow_request.json`, `round3_chemical_specifications.json`. Working directory: `/workspace`.

**[Step 1257 | 2026-07-08T13:16:36.278Z | TOOL CALL: exec_command]**

Ran a shell command to inspect, generate or validate experiment artifacts. Target(s): `round3_workflow_request.json`. Working directory: `/workspace`.

**[Step 1258 | 2026-07-08T13:16:36.279Z | TOOL CALL: exec_command]**

Ran a shell command to inspect, generate or validate experiment artifacts. Target(s): `round3_workflow_request.json`. Working directory: `/workspace`.

**[Step 1259 | 2026-07-08T13:16:36.282Z | TOOL CALL: exec_command]**

Ran a shell command to inspect, generate or validate experiment artifacts. Target(s): `round3_workflow_request.json`. Working directory: `/workspace`.

**[Step 1260 | 2026-07-08T13:16:36.359Z | TOOL RESULT: exec_command]**

Shell tool completed successfully; it returned 2 line(s) (11 bytes) to the model.

**[Step 1261 | 2026-07-08T13:16:36.359Z | TOOL RESULT: exec_command]**

Shell tool completed successfully; it returned 23 line(s) (1343 bytes) to the model. Reported long numeric identifier(s): `1427568512205824`, `1418510906065920`, `1435269745574912`, `2359193660589057`, `2270587971011585`, `1327807622448128`.

**[Step 1262 | 2026-07-08T13:16:36.360Z | TOOL RESULT: exec_command]**

Shell tool completed successfully; it returned 7 line(s) (139 bytes) to the model.

**[Step 1263 | 2026-07-08T13:16:36.360Z | TOOL RESULT: exec_command]**

Shell tool completed successfully; it returned 3 line(s) (60 bytes) to the model.

**[Step 1264 | 2026-07-08T13:16:41.645Z | MODEL]**

The final local audit passes: ordered 22-step workflow, fixed 5 mL source totals under 30 mL, and batch DI-water total 15.6 mL. I'm running the workflow generator with this version.

**[Step 1265 | 2026-07-08T13:16:43.020Z | TOOL CALL: exec_command]**

Ran Python to parse, transform or validate structured experiment data. Target(s): `round3_workflow_generator_output.txt`, `round3_workflow_generator_output_failed_v1_before_5ml_batch.txt`, `generate.py`, `round3_workflow_request.json)`. Working directory: `/workspace`.

**[Step 1266 | 2026-07-08T13:16:43.718Z | TOOL RESULT: exec_command]**

Shell tool completed successfully; it returned 3 line(s) (5078 bytes) to the model. Reported long numeric identifier(s): `1435269745574912`.

**[Step 1267 | 2026-07-08T13:17:01.853Z | REASONING SUMMARY]**

***Clarifying definitions in steps***

*I'm noticing that the definition in Round2 step 4 includes numbers 1 to 4, but the samples only reference number 1; meanwhile, step 10 uses 2, 3, and 4. It's important to keep the definitions consistent across each step. Also, I see that for every 5ml step, the term "加样方案" contains just one `N 号原液瓶`. I want to make sure everything aligns well for clarity.*

**[Step 1268 | 2026-07-08T13:17:02.282Z | MODEL]**

The 5 mL converter fails when one step contains multiple source bottles per sample. Round 2's successful 5 mL steps used one source bottle per step, so I'm splitting synthesis charging into five one-liquid 5 mL steps: Ni, Fe, Co, KOH, then carbon. The water batch step will still cap the vials before heating.

**[Step 1269 | 2026-07-08T13:17:08.354Z | TOOL CALL: exec_command]**

Ran a shell command to inspect, generate or validate experiment artifacts. Target(s): `round3_workflow_request.split_5ml_tmp.json`, `mL`, `v`, `round3_workflow_request.json`. Working directory: `/workspace`.

**[Step 1270 | 2026-07-08T13:17:08.464Z | TOOL RESULT: exec_command]**

Shell tool completed successfully; it returned 27 line(s) (1141 bytes) to the model.

**[Step 1271 | 2026-07-08T13:17:17.002Z | REASONING SUMMARY]**

***Checking steps for generator run***

*I need to run the generator, but first, it's a good idea to check on step 13. It looks like the old redistribution uses fixed definitions, which is helpful. Maybe I'll batch the step with B/C set to none. If the generator can handle a 5ml split, then there's a chance it might fail at cleaning. I'll monitor it. Also, I need to remember to preserve any failure output just in case.*

**[Step 1272 | 2026-07-08T13:17:17.004Z | MODEL]**

The request is now 26 steps, with each 5 mL addition step using one source bottle. I'm testing this split version through the generator.

**[Step 1273 | 2026-07-08T13:17:17.811Z | TOOL CALL: exec_command]**

Ran Python to parse, transform or validate structured experiment data. Target(s): `round3_workflow_generator_output.txt`, `round3_workflow_generator_output_failed_5ml_multisource_before_split.txt`, `generate.py`, `round3_workflow_request.json)`. Working directory: `/workspace`.

**[Step 1274 | 2026-07-08T13:17:18.608Z | TOOL RESULT: exec_command]**

Shell tool completed successfully; it returned 3 line(s) (1974 bytes) to the model. Reported long numeric identifier(s): `1435269745574912`.

**[Step 1275 | 2026-07-08T13:17:35.614Z | REASONING SUMMARY 1/2]**

***Investigating service issues***

*It seems there's a service issue causing a diagnostic failure with V2. Maybe the generator service is having problems with adding liquids other than water? However, it's possible that adding 5ml of water still works. I think I should try running the exact Round2 request again. The handoff indicates that the diagnostic clone failed at V2, but perhaps it will work with the current service? Let's verify the Round2 workflow generation to check for potential duplicates.*

**[Step 1276 | 2026-07-08T13:17:35.614Z | REASONING SUMMARY 2/2]**

***Evaluating service issues***

*It looks like I'm dealing with a service issue, and I need to dispatch a new experiment. If the generator is blocked, maybe I should explain that in the final output. The requirement says to dispatch, so I want to try a robust path. I could use the existing `round2_workflow_request.json` and just modify the `plan_name`. It failed at V2, but I'm wondering if it was due to using "no." I'll run it exactly as it is.*

**[Step 1277 | 2026-07-08T13:17:35.802Z | MODEL]**

Even the one-source 5 mL metal-stock addition fails; that suggests the 5 mL converter is not accepting these precursor stocks, while the current 1 mL converter path is already known to fail. I'm verifying by rerunning the previously successful Round 2 request against the current generator service.

**[Step 1278 | 2026-07-08T13:17:36.678Z | TOOL CALL: exec_command]**

Ran Python to parse, transform or validate structured experiment data. Target(s): `generate.py`, `round2_workflow_request.json)`, `round3_diagnostic_round2_exact_rerun_output.txt`. Working directory: `/workspace`.

**[Step 1279 | 2026-07-08T13:17:37.375Z | TOOL RESULT: exec_command]**

Shell tool completed successfully; it returned 3 line(s) (8323 bytes) to the model. Reported long numeric identifier(s): `2002186824385539`.

**[Step 1280 | 2026-07-08T13:17:47.182Z | MODEL]**

The current generator service also fails when rerunning the exact previously successful Round 2 request, at the same 1 mL precursor-addition step. I'm checking whether the Lab 303 operation scripts expose any direct task/template path that can bypass the failing conversion service.

**[Step 1281 | 2026-07-08T13:17:47.183Z | TOOL CALL: exec_command]**

Enumerated files, directories or available laboratory resources. Target(s): `scripts`. Working directory: `/workspace`.

**[Step 1282 | 2026-07-08T13:17:47.185Z | TOOL CALL: exec_command]**

Ran a shell command to generate and dispatch a laboratory task. Target(s): `generate_task.py`. Working directory: `/workspace`.

**[Step 1283 | 2026-07-08T13:17:47.187Z | TOOL CALL: exec_command]**

Read selected file content for inspection. Target(s): `generate.py`. Working directory: `/workspace`.

**[Step 1284 | 2026-07-08T13:17:47.212Z | TOOL CALL: exec_command]**

Enumerated files, directories or available laboratory resources. Target(s): `lab-operation`. Working directory: `/workspace`.

**[Step 1285 | 2026-07-08T13:17:47.256Z | TOOL RESULT: exec_command]**

Shell tool completed successfully; it returned 17 line(s) (910 bytes) to the model.

**[Step 1286 | 2026-07-08T13:17:47.256Z | TOOL RESULT: exec_command]**

Shell tool completed successfully; it returned 137 line(s) (4788 bytes) to the model.

**[Step 1287 | 2026-07-08T13:17:47.257Z | TOOL RESULT: exec_command]**

Shell tool completed successfully; it returned 133 line(s) (4483 bytes) to the model. Reported long numeric identifier(s): `1427568512205824`.

**[Step 1288 | 2026-07-08T13:17:47.257Z | TOOL RESULT: exec_command]**

Shell tool completed successfully; it returned 20 line(s) (1228 bytes) to the model.

**[Step 1289 | 2026-07-08T13:18:03.996Z | MODEL]**

There is one more plausible converter mismatch: the 5 mL station may require each `配料名称` to exactly match the station's `原液瓶定义`. I'm testing that as a request-only transform before changing the saved files.

**[Step 1290 | 2026-07-08T13:18:18.658Z | TOOL CALL: exec_command]**

Ran Python to parse, transform or validate structured experiment data. Target(s): `generate.py`, `mL`, `v`, `round3_workflow_request.json)`, `round3_workflow_generator_output_test_5ml_matching_names.txt`. Working directory: `/workspace`.

**[Step 1291 | 2026-07-08T13:18:18.791Z | TOOL RESULT: exec_command]**

Shell tool completed successfully; it returned 3 line(s) (237 bytes) to the model.

**[Step 1292 | 2026-07-08T13:18:28.020Z | MODEL]**

That test transform had a jq path error and did not reach the generator. I'm rerunning it with explicit field updates.

**[Step 1293 | 2026-07-08T13:18:34.620Z | TOOL CALL: exec_command]**

Ran Python to parse, transform or validate structured experiment data. Target(s): `generate.py`, `mL`, `v`, `round3_workflow_request.json)`, `round3_workflow_generator_output_test_5ml_matching_names.txt`. Working directory: `/workspace`.

**[Step 1294 | 2026-07-08T13:18:35.489Z | TOOL RESULT: exec_command]**

Shell tool completed successfully; it returned 3 line(s) (1860 bytes) to the model. Reported long numeric identifier(s): `1435269745574912`.

**[Step 1295 | 2026-07-08T13:18:44.925Z | MODEL]**

Matching the 5 mL source names did not fix the precursor-stock conversion. I'm isolating the 5 mL station with a known-good water-addition payload to see whether that converter is generally working or only rejecting precursor stocks.

**[Step 1296 | 2026-07-08T13:18:47.083Z | TOOL CALL: exec_command]**

Ran Python to parse, transform or validate structured experiment data. Target(s): `generate.py`, `round2_workflow_request.json)`, `round3_diagnostic_5ml_water_only_output.txt`. Working directory: `/workspace`.

**[Step 1297 | 2026-07-08T13:18:47.353Z | TOOL RESULT: exec_command]**

Shell tool completed successfully; it returned 3 line(s) (1887 bytes) to the model. Reported long numeric identifier(s): `1435269745574912`.

**[Step 1298 | 2026-07-08T13:18:54.807Z | MODEL]**

The water-only diagnostic also failed, but it had nonconsecutive step numbers. I'm rerunning the same diagnostic with consecutive numbering to rule out a sequencing parser quirk.

**[Step 1299 | 2026-07-08T13:18:56.799Z | TOOL CALL: exec_command]**

Ran Python to parse, transform or validate structured experiment data. Target(s): `generate.py`, `round2_workflow_request.json)`, `round3_diagnostic_5ml_water_consecutive_output.txt`. Working directory: `/workspace`.

**[Step 1300 | 2026-07-08T13:18:57.078Z | TOOL RESULT: exec_command]**

Shell tool completed successfully; it returned 3 line(s) (1887 bytes) to the model. Reported long numeric identifier(s): `1435269745574912`.

**[Step 1301 | 2026-07-08T13:19:23.160Z | REASONING SUMMARY 1/4]**

***Evaluating liquid addition process***

*It seems the current generator is broken for handling liquids. I might try using the Cleaning batch conversion, which could manage all liquid additions, but there's a limit of three liquids. I'll consider splitting the additions into steps for Ni, Fe, and Co.*

*I need to make sure the source limits are clear, especially for global source bottles. Testing the Cleaning converter seems essential, and this design needs to accommodate all liquid additions effectively.*

**[Step 1302 | 2026-07-08T13:19:23.160Z | REASONING SUMMARY 2/4]**

***Assessing liquid addition steps***

*I can add three liquids, A, B, and C, per channel, which seems perfect! So, I'll process Step 3 with Ni, Fe, and Co, followed by opening V1 for Step 4. Next, in Step 5, I'll batch KOH, carbon, and water, leading to heating.*

*Later, I'll need to consider the cap situation during redispersion. If Cleaning outputs are capped, they should specify how this interacts with caps that may be discarded before drying. I need to ensure we're handling this correctly while keeping things optimized!*

**[Step 1303 | 2026-07-08T13:19:23.160Z | REASONING SUMMARY 3/4]**

***Clarifying cap retention process***

*I'm considering the step after purification and whether to keep the cap with 保留瓶盖 set to 1. If there's no closure later, that means I should hold onto the cap. It seems like cap retention should be set to yes if I plan to use Cleaning that relies on caps. But if the dried sample is in an open vial, keeping the cap is crucial. I think I need to adjust the settings to ensure everything is clear for the next steps!*

**[Step 1304 | 2026-07-08T13:19:23.160Z | REASONING SUMMARY 4/4]**

***Testing cleaning process***

*I'm considering if I can clean with three liquids in one step, but I'll need to cap the output volume. I want to focus on the minimal cleaning process instead. I think I can replace some early steps while using round three's current request. My idea is to start with a material, use deionized water for cleaning, and test if this arrangement works before moving to storage. I'll run a diagnostic to check if this setup passes.*

**[Step 1305 | 2026-07-08T13:19:24.783Z | MODEL]**

Since both 1 mL and 5 mL liquid-handling converters are currently returning `None`, I'm testing the batch-dispensing converter directly. If it works, I can restructure the workflow around batch dispensing for liquid additions.

**[Step 1306 | 2026-07-08T13:19:40.092Z | TOOL CALL: exec_command]**

Ran a shell command to inspect, generate or validate experiment artifacts. Target(s): `generate.py`, `round3_diagnostic_batch_dispensing_output.txt`. Working directory: `/workspace`.

**[Step 1307 | 2026-07-08T13:19:40.208Z | TOOL RESULT: exec_command]**

Shell tool completed successfully; it returned 13 line(s) (645 bytes) to the model.

**[Step 1308 | 2026-07-08T13:19:46.169Z | MODEL]**

That diagnostic command was malformed by shell quoting, so it did not test the service. I'm rerunning the same payload with corrected quoting.

**[Step 1309 | 2026-07-08T13:19:49.308Z | TOOL CALL: exec_command]**

Ran a shell command to inspect, generate or validate experiment artifacts. Target(s): `generate.py`, `round3_diagnostic_batch_dispensing_output.txt`. Working directory: `/workspace`.

**[Step 1310 | 2026-07-08T13:19:49.558Z | TOOL RESULT: exec_command]**

Shell tool completed successfully; it returned 3 line(s) (926 bytes) to the model. Reported long numeric identifier(s): `2359193660589057`.

**[Step 1311 | 2026-07-08T13:19:59.753Z | REASONING SUMMARY]**

***Testing channel parameters***

*I'm running into issues with the batch failing. It could be because there are zero or no values, so maybe I should test with three real liquids instead. If the output is None, it might indicate a broken converter. I'll check if the parameter shape is correct, as the skill requires "A/B/C" options, and "加样通道" could be expected to match. It's also possible that the station can add to multiple containers simultaneously. I might want to try omitting B/C to see if that works better.*

**[Step 1312 | 2026-07-08T13:19:59.754Z | MODEL]**

Batch dispensing also returns `None` with the documented shape. I'll check whether the converter expects only the active liquid object by rerunning a minimal A-only batch payload.

**[Step 1313 | 2026-07-08T13:20:02.044Z | TOOL CALL: exec_command]**

Ran a shell command to inspect, generate or validate experiment artifacts. Target(s): `generate.py`, `round3_diagnostic_batch_dispensing_A_only_output.txt`. Working directory: `/workspace`.

**[Step 1314 | 2026-07-08T13:20:02.279Z | TOOL RESULT: exec_command]**

Shell tool completed successfully; it returned 3 line(s) (678 bytes) to the model. Reported long numeric identifier(s): `2359193660589057`.

**[Step 1315 | 2026-07-08T13:20:21.800Z | REASONING SUMMARY]**

***Considering workflow issues***

*It seems like all liquid addition converters are broken, which might point to a workflow service issue. I wonder if I could still design the project without dispatching, but the user needs dispatch. Maybe I should look for existing templates that don't involve liquid addition. But then again, I'm thinking about querying the template ID from the failed generator. Hmm, maybe there's a manual upload script I can find by searching in the workflow generator?*

**[Step 1316 | 2026-07-08T13:20:30.452Z | MODEL]**

The batch converter fails even with an A-only minimal payload, so the issue is broader than one station parameter shape. I'm checking whether another installed script or cached workflow path can create templates without this converter.

**[Step 1317 | 2026-07-08T13:20:30.452Z | TOOL CALL: exec_command]**

Searched files for relevant records, parameters or identifiers. Target(s): `skills`. Working directory: `/workspace`.

**[Step 1318 | 2026-07-08T13:20:30.452Z | TOOL CALL: exec_command]**

Searched files for relevant records, parameters or identifiers. Target(s): `skills`, `lab303_uor_nicofe_closed_loop`. Working directory: `/workspace`.

**[Step 1319 | 2026-07-08T13:20:30.452Z | TOOL CALL: exec_command]**

Ran a shell command to query available laboratory workflow templates. Target(s): `select_template_list.py`, `round3_template_list_current.json`. Working directory: `/workspace`.

**[Step 1320 | 2026-07-08T13:20:30.453Z | TOOL CALL: exec_command]**

Ran a shell command to identify the active laboratory. Target(s): `get_current_lab_label.py`, `current_lab_label_20260708.json`. Working directory: `/workspace`.

**[Step 1321 | 2026-07-08T13:20:30.594Z | TOOL RESULT: exec_command]**

Shell tool completed successfully; it returned 8 line(s) (510 bytes) to the model.

**[Step 1322 | 2026-07-08T13:20:30.594Z | TOOL RESULT: exec_command]**

Shell tool completed successfully; it returned 180 line(s) (33484 bytes) to the model. Reported task/template identifier(s): `2557370191937536`, `2557487004614656`, `2557817450004480`, `2560074458595328`, `2566549749628928`.

**[Step 1323 | 2026-07-08T13:20:30.662Z | TOOL RESULT: exec_command]**

Shell tool completed successfully; it returned 88 line(s) (2493 bytes) to the model. Reported long numeric identifier(s): `2582030203944960`, `2582021130420224`, `2582018652372992`, `2582018648113152`, `2582018643558400`, `2582018638905344`.

**[Step 1324 | 2026-07-08T13:20:30.662Z | TOOL RESULT: exec_command]**

Shell tool completed successfully; it returned 6 line(s) (97 bytes) to the model.

**[Step 1325 | 2026-07-08T13:20:37.981Z | REASONING SUMMARY]**

***Considering template options***

*I'm looking at an interesting list of templates, including one called "NiMo..." but it doesn't seem to match our needs. I wonder if I could select "Lab303_R3"? However, there's no generated output. It seems like the templates are potentially generated despite some kind of failure. So, I need to search for the name specifically to ensure I know what options are available or find out more about it. Let's keep exploring!*

**[Step 1326 | 2026-07-08T13:20:40.646Z | MODEL]**

There is no alternate local template-creation script; task dispatch only accepts an existing template ID. I'm checking the platform template list for any accidentally created Round 3 template before deciding whether dispatch is blocked by the converter service.

**[Step 1327 | 2026-07-08T13:20:40.647Z | TOOL CALL: exec_command]**

Ran a shell command to query available laboratory workflow templates. Target(s): `select_template_list.py`, `round3_template_search_Lab303_R3.json`. Working directory: `/workspace`.

**[Step 1328 | 2026-07-08T13:20:40.648Z | TOOL CALL: exec_command]**

Ran a shell command to query available laboratory workflow templates. Target(s): `select_template_list.py`, `round3_template_search_lowCarbon.json`. Working directory: `/workspace`.

**[Step 1329 | 2026-07-08T13:20:40.652Z | TOOL CALL: exec_command]**

Ran a shell command to query available laboratory workflow templates. Target(s): `select_template_list.py`, `round2_template_lookup_2566549749628928.json`. Working directory: `/workspace`.

**[Step 1330 | 2026-07-08T13:20:40.662Z | TOOL CALL: exec_command]**

Ran a shell command to query available laboratory workflow templates. Target(s): `select_template_list.py`, `round3_template_search_diagnostic.json`. Working directory: `/workspace`.

**[Step 1331 | 2026-07-08T13:20:40.827Z | TOOL RESULT: exec_command]**

Shell tool completed successfully; it returned 7 line(s) (79 bytes) to the model.

**[Step 1332 | 2026-07-08T13:20:40.839Z | TOOL RESULT: exec_command]**

Shell tool completed successfully; it returned 7 line(s) (79 bytes) to the model.

**[Step 1333 | 2026-07-08T13:20:40.845Z | TOOL RESULT: exec_command]**

Shell tool completed successfully; it returned 15 line(s) (311 bytes) to the model. Reported long numeric identifier(s): `2566549749628928`.

**[Step 1334 | 2026-07-08T13:20:40.845Z | TOOL RESULT: exec_command]**

Shell tool completed successfully; it returned 7 line(s) (79 bytes) to the model.

**[Step 1335 | 2026-07-08T13:20:50.784Z | MODEL]**

No Round 3 template was created on the platform. One last check: I'm testing whether the current converter failure is tied to the supplied token by rerunning the known Round 2 request without passing `AICHEM_APP_TOKEN`.

**[Step 1336 | 2026-07-08T13:20:52.208Z | TOOL CALL: exec_command]**

Ran Python to parse, transform or validate structured experiment data. Target(s): `generate.py`, `round2_workflow_request.json)`, `round3_diagnostic_round2_no_token_output.txt`. Working directory: `/workspace`.

**[Step 1337 | 2026-07-08T13:20:52.349Z | TOOL RESULT: exec_command]**

Shell tool completed successfully; it returned 3 line(s) (111 bytes) to the model.

**[Step 1338 | 2026-07-08T13:21:06.752Z | REASONING SUMMARY]**

***Updating document details***

*I'm considering updating the round3_design note to indicate that "5ml source bottles" are fixed without splits. I might need to create a return document that combines process photos, and I see there's a previous document called `round2_returned_document.md`. I'll have to check if it includes the necessary information, like final dispatch. Maybe I can update this existing document to include the next round design and dispatch status, keeping it concise. I need to inspect the current document for clarity.*

**[Step 1339 | 2026-07-08T13:21:09.279Z | MODEL]**

At this point the blocker is external to the chemistry design: the workflow-generator service rejects even the exact Round 2 workflow that it generated successfully on July 2, and no Round 3 template exists to dispatch. I'm updating the workspace records so the returned document and Round 3 design clearly show the data-driven design and the dispatch blocker with saved error files.

**[Step 1340 | 2026-07-08T13:21:09.279Z | TOOL CALL: exec_command]**

Read selected file content for inspection. Target(s): `round2_returned_document.md`. Working directory: `/workspace`.

**[Step 1341 | 2026-07-08T13:21:09.280Z | TOOL CALL: exec_command]**

Read selected file content for inspection. Target(s): `round3_constraint_check.md`. Working directory: `/workspace`.

**[Step 1342 | 2026-07-08T13:21:09.282Z | TOOL CALL: exec_command]**

Ran a shell command to inspect, generate or validate experiment artifacts. Target(s): `round3_workflow_request.json`. Working directory: `/workspace`.

**[Step 1343 | 2026-07-08T13:21:09.319Z | TOOL CALL: exec_command]**

Enumerated files, directories or available laboratory resources. Target(s): `lab303_uor_nicofe_closed_loop`. Working directory: `/workspace`.

**[Step 1344 | 2026-07-08T13:21:09.369Z | TOOL RESULT: exec_command]**

Shell tool completed successfully; it returned 106 line(s) (6521 bytes) to the model. Reported task/template identifier(s): `2567667771441152`, `2568957291102208`.

**[Step 1345 | 2026-07-08T13:21:09.370Z | TOOL RESULT: exec_command]**

Shell tool completed successfully; it returned 95 line(s) (5488 bytes) to the model.

**[Step 1346 | 2026-07-08T13:21:09.370Z | TOOL RESULT: exec_command]**

Shell tool completed successfully; it returned 20 line(s) (366 bytes) to the model.

**[Step 1347 | 2026-07-08T13:21:09.370Z | TOOL RESULT: exec_command]**

Shell tool completed successfully; it returned 15 line(s) (1327 bytes) to the model.

**[Step 1348 | 2026-07-08T13:21:29.995Z | REASONING SUMMARY]**

***Updating constraints and reports***

*I need to update the constraint table to 26, so that'll be my first task. I also need to mention that the generator final failure includes a 5ml split and diagnostics. There seems to be a chance that I should create a blocker report. I might also need to update the workflow after the split, but it's frustrating that the documentation isn't matching up. Finally, I'll apply the patch table.*

**[Step 1349 | 2026-07-08T13:21:30.098Z | MODEL]**

The constraint table still reflects the earlier 22-step version; I'm correcting it to the current 26-step split-addition request and adding the final generator failure evidence.

**[Step 1350 | 2026-07-08T13:21:40.002Z | TOOL CALL: apply_patch]**

Applied a source patch to update round3_constraint_check.md.

**[Step 1351 | 2026-07-08T13:21:40.073Z | TOOL RESULT: apply_patch]**

Patch tool applied the requested trial-local file change successfully.

**[Step 1352 | 2026-07-08T13:21:46.357Z | REASONING SUMMARY]**

***Updating design notes***

*I think I need to update the design note to clarify that it should be split. It's important to make sure that whoever is working with this understands the change clearly. I'll ensure my notes are precise to avoid any confusion. I'll also double-check if there are any related sections that might need similar updates. It's all about keeping things organized and understandable for everyone involved!*

**[Step 1353 | 2026-07-08T13:21:47.975Z | TOOL CALL: apply_patch]**

Applied a source patch to update round3_design.md.

**[Step 1354 | 2026-07-08T13:21:48.005Z | TOOL RESULT: apply_patch]**

Patch tool applied the requested trial-local file change successfully.

**[Step 1355 | 2026-07-08T13:21:57.655Z | TOOL CALL: apply_patch]**

Applied a source patch to update round3_chemical_specifications.json.

**[Step 1356 | 2026-07-08T13:21:57.696Z | TOOL RESULT: apply_patch]**

Patch tool applied the requested trial-local file change successfully.

**[Step 1357 | 2026-07-08T13:22:08.954Z | REASONING SUMMARY]**

***Creating a log file***

*I need to create a blocker or dispatch log file, and I'll use apply_patch to add that file. It shouldn't include a task_id, so that's something to keep in mind. I might also want to update the returned document with the Round3 dispatch status. Maybe it's a good idea to create a separate file named `round3_dispatch_blocker.md` and consider having a JSON log, instead of using cat. I'll stick to using apply_patch.*

**[Step 1358 | 2026-07-08T13:22:09.655Z | MODEL]**

I'm creating an explicit dispatch-blocker record now: it will list the final Round 3 workflow request, all generator diagnostics, and why no Lab 303 task ID can be truthfully recorded.

**[Step 1359 | 2026-07-08T13:22:14.694Z | TOOL CALL: apply_patch]**

Applied a source patch to add round3_dispatch_blocker.md.

**[Step 1360 | 2026-07-08T13:22:14.740Z | TOOL RESULT: apply_patch]**

Patch tool applied the requested trial-local file change successfully.

**[Step 1361 | 2026-07-08T13:22:24.621Z | TOOL CALL: apply_patch]**

Applied a source patch to add round3_dispatch_log.json.

**[Step 1362 | 2026-07-08T13:22:24.678Z | TOOL RESULT: apply_patch]**

Patch tool applied the requested trial-local file change successfully.

**[Step 1363 | 2026-07-08T13:22:32.194Z | REASONING SUMMARY]**

***Updating document sections***

*I need to consider updating the returned document by adding a section for "Round3 design/dispatch." Maybe the user is asking me to prepare this returned document. Right now, the existing returned doc just ends at the iteration point, so I should append the current result to provide a complete overview. That's a clear next step to ensure everything is included and informative!*

**[Step 1364 | 2026-07-08T13:22:32.282Z | MODEL]**

The blocker report and dispatch log are saved. I'm appending the final Round 3 design status to the Round 2 returned document so the requested returned document contains both the Round 2 interpretation and the attempted next-round dispatch outcome.

**[Step 1365 | 2026-07-08T13:22:36.130Z | TOOL CALL: apply_patch]**

Applied a source patch to update round2_returned_document.md.

**[Step 1366 | 2026-07-08T13:22:36.175Z | TOOL RESULT: apply_patch]**

Patch tool applied the requested trial-local file change successfully.

**[Step 1367 | 2026-07-08T13:22:47.797Z | REASONING SUMMARY]**

***Executing lab operations***

*I'm considering running the lab operation to get results for the two task IDs since the files already exist. However, since the user asked to use skills with the current agent, it might be best to execute that instead. I'm thinking about querying the status and result now. I plan to use the get_task_result scripts and select the task list. Since the experimental data and results are already there, I can rerun the get_task_result for both IDs.*

**[Step 1368 | 2026-07-08T13:22:51.020Z | MODEL]**

Before I close out, I'm re-querying the two Round 2 task IDs through the Lab 303 operation script so the current turn has a fresh retrieval record in addition to the saved downloaded data.

**[Step 1369 | 2026-07-08T13:22:51.021Z | TOOL CALL: exec_command]**

Ran a shell command to retrieve returned physical-experiment results. Target(s): `get_task_result.py`, `round2_main_task_2567667771441152_result_requery_20260708.json`. Working directory: `/workspace`.

**[Step 1370 | 2026-07-08T13:22:51.025Z | TOOL CALL: exec_command]**

Ran a shell command to retrieve returned physical-experiment results. Target(s): `get_task_result.py`, `round2_supplemental_task_2568957291102208_result_requery_20260708.json`. Working directory: `/workspace`.

**[Step 1371 | 2026-07-08T13:22:51.026Z | TOOL CALL: exec_command]**

Ran a shell command to query laboratory task records and statuses. Target(s): `select_task_list.py`, `round2_main_task_2567667771441152_status_requery_20260708.json`. Working directory: `/workspace`.

**[Step 1372 | 2026-07-08T13:22:51.035Z | TOOL CALL: exec_command]**

Ran a shell command to query laboratory task records and statuses. Target(s): `select_task_list.py`, `round2_supplemental_task_2568957291102208_status_requery_20260708.json`. Working directory: `/workspace`.

**[Step 1373 | 2026-07-08T13:22:51.410Z | TOOL RESULT: exec_command]**

Shell tool completed successfully; it returned 16 line(s) (457 bytes) to the model. Reported long numeric identifier(s): `20260703135450947`.

**[Step 1374 | 2026-07-08T13:22:51.601Z | TOOL RESULT: exec_command]**

Shell tool completed successfully; it returned 157 line(s) (9059 bytes) to the model. Reported long numeric identifier(s): `20260703205149234`, `2569108904182784`, `2568984151524392`, `20260703205152504`, `20260703210859013`, `2569144616223744`.

**[Step 1375 | 2026-07-08T13:22:51.602Z | TOOL RESULT: exec_command]**

Shell tool completed successfully; it returned 14 line(s) (287 bytes) to the model. Reported long numeric identifier(s): `2567667771441152`.

**[Step 1376 | 2026-07-08T13:22:51.602Z | TOOL RESULT: exec_command]**

Shell tool completed successfully; it returned 14 line(s) (287 bytes) to the model. Reported long numeric identifier(s): `2568957291102208`.

**[Step 1377 | 2026-07-08T13:23:05.647Z | REASONING SUMMARY]**

***Clarifying next steps***

*I shouldn't end with "if you want." Instead, I will suggest a follow-up action: "The next executable step is for Lab 303/platform maintainer to restore the converter or create a template from saved JSON, then the dispatch can be run." I also think it's important to mention that "workflow generation failed for the exact Round2 request" to highlight an external blocker. Plus, I should include diagnostics and "task IDs queried" along with any local files involved.*

**[Step 1378 | 2026-07-08T13:23:13.571Z | MODEL]**

I re-queried Lab 303 for the two completed Round 2 tasks and saved the fresh retrieval records under the project directory.

**Round 2 Retrieval**

- Main task `2567667771441152`: query succeeded; Lab status is `错误`, but returned data are available from 1 workstation with 1 heating/magnetic-stirring result file.
  - Fresh query:
[round2_main_task_2567667771441152_result_requery_20260708.json](/workspace/lab303_uor_nicofe_closed_loop/round2_main_task_2567667771441152_result_requery_20260708.json)

- Fresh status:
[round2_main_task_2567667771441152_status_requery_20260708.json](/workspace/lab303_uor_nicofe_closed_loop/round2_main_task_2567667771441152_status_requery_20260708.json)
- Supplemental task `2568957291102208`: query succeeded; Lab status is `错误`, but returned data are available from 4 workstations with 42 result files: XRD, FTIR, electrochemistry, and UV-Vis.
  - Fresh query:
[round2_supplemental_task_2568957291102208_result_requery_20260708.json](/workspace/lab303_uor_nicofe_closed_loop/round2_supplemental_task_2568957291102208_result_requery_20260708.json)
  - Fresh status:
[round2_supplemental_task_2568957291102208_status_requery_20260708.json](/workspace/lab303_uor_nicofe_closed_loop/round2_supplemental_task_2568957291102208_status_requery_20260708.json)

**Returned Document**
I updated the returned document with the Round 2 data interpretation, process-photo interpretation, Round 3 iteration logic, and the Round 3 dispatch status:
- [round2_returned_document.md](/workspace/lab303_uor_nicofe_closed_loop/round2_returned_document.md)

Key Round 2 conclusion: the best condition remains `R2-S1_Ni70Fe10Co20_lowC_80-20`, with final current `2.316 mA`, last-300 s mean `2.303 mA`, and final/initial retention `1.091`. Round 2 fixed the Round 1 under-loading problem, but film uniformity is still not ideal.

**Round 3 Design**
I prepared the next-round design and workflow files:
- [round3_design.md](/workspace/lab303_uor_nicofe_closed_loop/round3_design.md)
- [round3_chemical_specifications.json](/workspace/lab303_uor_nicofe_closed_loop/round3_chemical_specifications.json)
- [round3_constraint_check.md](/workspace/lab303_uor_nicofe_closed_loop/round3_constraint_check.md)
- [round3_workflow_request.json](/workspace/lab303_uor_nicofe_closed_loop/round3_workflow_request.json)

The Round 3 scientific design is a carbon-loading and composition refinement around the Round 2 lead:
- `R3-S1`: Ni70Fe10Co20, 5 mg carbon
- `R3-S2`: Ni70Fe10Co20, 10 mg carbon anchor
- `R3-S3`: Ni70Fe10Co20, 15 mg carbon
- `R3-S4`: Ni80Fe10Co10, 10 mg carbon
- `R3-S5`: Ni70Fe20Co10, 10 mg carbon
- `R3-S6`: Ni60Fe20Co20, 10 mg carbon

Fixed test chemistry:
- Electrolyte: `1.0 M KOH + 0.33 M urea in deionized water`
- Electrolyte volume: `20 mL` per electrochemical task
- Carbon dispersion and redispersion solvent: `80:20 v/v deionized water:isopropanol`
- Electrochemical loading: `60 uL/drop x 5 drops`, `60 s` interval

**Dispatch Status**
I could not truthfully dispatch the Round 3 experiment because the workflow-generator service did not return a template ID. Without a generated `template_id`, the Lab 303 dispatch script cannot create a task, so there is no new Round 3 `task_id`.

This is recorded here:
- [round3_dispatch_blocker.md](/workspace/lab303_uor_nicofe_closed_loop/round3_dispatch_blocker.md)
- [round3_dispatch_log.json](/workspace/lab303_uor_nicofe_closed_loop/round3_dispatch_log.json)

The important point is that this is not just a Round 3 formatting issue. The current workflow-generator service also fails on the exact Round 2 workflow request that previously generated template `2566549749628928`. I saved the failed outputs, including:
- [round3_workflow_generator_output.txt](/workspace/lab303_uor_nicofe_closed_loop/round3_workflow_generator_output.txt)
-
[round3_diagnostic_round2_exact_rerun_output.txt](/workspace/lab303_uor_nicofe_closed_loop/round3_diagnostic_round2_exact_rerun_output.txt)
-
[round3_diagnostic_5ml_water_consecutive_output.txt](/workspace/lab303_uor_nicofe_closed_loop/round3_diagnostic_5ml_water_consecutive_output.txt)
- [round3_diagnostic_batch_dispensing_output.txt](/workspace/lab303_uor_nicofe_closed_loop/round3_diagnostic_batch_dispensing_output.txt)

I did not dispatch an old Round 2 template as a substitute, because that would not represent the designed Round 3 experiment and would create a misleading task record. The next operational step is to have the workflow-generator liquid-addition converter restored or to have Lab 303 create a template from the saved `round3_workflow_request.json`; once a valid template ID exists, the task can be dispatched directly.

## Conversation round 10

**[Step 1379 | 2026-07-08T15:28:36.906Z | USER]**

please try again because the engineers have fixed the bug

**[Step 1380 | 2026-07-08T15:28:54.256Z | REASONING SUMMARY]**

***Evaluating batch dispensing***

*I'm considering batch dispensing for water and how to handle split 5ml additions. I need to verify if the current "final" version appropriately closes vials for heating after batch output. There's no step for closing 1ml, which seems fine, but the workflow uses 5ml additions of various components before adding batch water. The output indicates the container is capped despite the input being open and the sample state unchanged. I wonder how adding water and capping will affect things.*

**[Step 1381 | 2026-07-08T15:28:55.647Z | MODEL]**

I'll retry the same saved Round 3 workflow first, because it is the current audited design in the workspace. If the generator succeeds, I'll dispatch the resulting template to Lab 303 and record the new task ID.

**[Step 1382 | 2026-07-08T15:28:55.647Z | TOOL CALL: get_goal]**

Queried the current autonomous-research goal state and remaining budget.

**[Step 1383 | 2026-07-08T15:28:55.649Z | TOOL CALL: exec_command]**

Read selected file content for inspection. Target(s): `SKILL.md`. Working directory: `/workspace`.

**[Step 1384 | 2026-07-08T15:28:55.649Z | TOOL CALL: exec_command]**

Read selected file content for inspection. Target(s): `SKILL.md`. Working directory: `/workspace`.

**[Step 1385 | 2026-07-08T15:28:55.652Z | TOOL CALL: exec_command]**

Read selected file content for inspection. Target(s): `SKILL.md`. Working directory: `/workspace`.

**[Step 1386 | 2026-07-08T15:28:55.710Z | TOOL RESULT: get_goal]**

Goal service returned a structured response. Status: `blocked`.

**[Step 1387 | 2026-07-08T15:28:55.711Z | TOOL RESULT: exec_command]**

Shell tool completed successfully; it returned 95 line(s) (2714 bytes) to the model. Reported long numeric identifier(s): `1427568512205824`.

**[Step 1388 | 2026-07-08T15:28:55.711Z | TOOL RESULT: exec_command]**

Shell tool completed successfully; it returned 116 line(s) (4054 bytes) to the model.

**[Step 1389 | 2026-07-08T15:28:55.711Z | TOOL RESULT: exec_command]**

Shell tool completed successfully; it returned 170 line(s) (8332 bytes) to the model. Reported long numeric identifier(s): `1427568512205824`, `1834679312417792`, `1710453843264512`.

**[Step 1390 | 2026-07-08T15:29:13.596Z | MODEL]**

The saved workflow had been contorted around the old converter failure. Since the engineers fixed the service, I'm also checking whether the cleaner original Round 3 liquid-handling path can be restored before dispatching; that avoids sending a workaround-heavy protocol if it is no longer needed.

**[Step 1391 | 2026-07-08T15:29:13.597Z | TOOL CALL: exec_command]**

Read selected file content for inspection. Target(s): `README.md`. Working directory: `/workspace`.

**[Step 1392 | 2026-07-08T15:29:13.598Z | TOOL CALL: exec_command]**

Ran a shell command to inspect, generate or validate experiment artifacts. Target(s): `round3_workflow_request.json`. Working directory: `/workspace`.

**[Step 1393 | 2026-07-08T15:29:13.601Z | TOOL CALL: exec_command]**

Enumerated files, directories or available laboratory resources. Target(s): `lab303_uor_nicofe_closed_loop`. Working directory: `/workspace`.

**[Step 1394 | 2026-07-08T15:29:13.604Z | TOOL CALL: exec_command]**

Ran a shell command to inspect, generate or validate experiment artifacts. Target(s): `round3_diagnostic_round2_clone_workflow_request.json`. Working directory: `/workspace`.

**[Step 1395 | 2026-07-08T15:29:13.651Z | TOOL RESULT: exec_command]**

Shell tool completed successfully; it returned 426 line(s) (13435 bytes) to the model.

**[Step 1396 | 2026-07-08T15:29:13.688Z | TOOL RESULT: exec_command]**

Shell tool completed successfully; it returned 27 line(s) (1583 bytes) to the model. Reported long numeric identifier(s): `1427568512205824`, `1418510906065920`, `1435269745574912`, `2359193660589057`, `2270587971011585`, `1327807622448128`.

**[Step 1397 | 2026-07-08T15:29:13.688Z | TOOL RESULT: exec_command]**

Shell tool completed successfully; it returned 11 line(s) (958 bytes) to the model.

**[Step 1398 | 2026-07-08T15:29:13.688Z | TOOL RESULT: exec_command]**

Shell tool completed successfully; it returned 270 line(s) (10436 bytes) to the model. Reported long numeric identifier(s): `2002186824385539`.

**[Step 1399 | 2026-07-08T15:29:34.264Z | REASONING SUMMARY 1/2]**

***Evaluating workflow generation***

*I think it's important to avoid creating unnecessary templates, but maybe there's a point to developing a diagnostic one. We should not directly test if a successful Round2 generates its template. It might cause clutter. Let's run the current Round3 generation; if that works, we can dispatch it. If it doesn't, we may need to reconstruct. The user expects the current saved workflow, so I'll prioritize running that. If it succeeds, great! If not, we can revert to the original V2 version.*

**[Step 1400 | 2026-07-08T15:29:34.264Z | REASONING SUMMARY 2/2]**

***Considering workflow reconstruction***

*I'm thinking we should check the cleaner original, not just the current workflow. It makes sense to run the current version first and mention it's audited. If it succeeds, we can dispatch it. However, there might be issues with the "batch dispensing station" converter. If the bug is fixed for liquid-handling but not for batch processing, the current may fail. If that happens, I'll need to reconstruct the original. Let's run the current generation and save the output.*

**[Step 1401 | 2026-07-08T15:29:34.555Z | MODEL]**

I'm going to avoid generating diagnostic templates unless needed. First attempt is the current audited Round 3 request; if it still fails on the workaround route, I'll restore the simpler original 1 mL liquid-handling route and generate that.

**[Step 1402 | 2026-07-08T15:29:35.266Z | TOOL CALL: exec_command]**

Ran Python to parse, transform or validate structured experiment data. Target(s): `round3_workflow_generator_output.txt`, `round3_workflow_generator_output_before_retry_20260708.txt`, `null`, `generate.py`, `round3_workflow_request.json)`. Working directory: `/workspace`.

**[Step 1403 | 2026-07-08T15:29:36.569Z | TOOL RESULT: exec_command]**

Shell tool completed successfully; it returned 13 line(s) (308 bytes) to the model. Reported task/template identifier(s): `2583605299937280`.

**[Step 1404 | 2026-07-08T15:30:01.245Z | MODEL]**

The retry succeeded: the workflow-generator produced template `2583605299937280`. I'm dispatching that template to 303 实验室 now and will record the returned task ID.

**[Step 1405 | 2026-07-08T15:30:01.245Z | TOOL CALL: exec_command]**

Ran a shell command to identify the active laboratory. Target(s): `get_current_lab_label.py`, `current_lab_label_retry_20260708.json`. Working directory: `/workspace`.

**[Step 1406 | 2026-07-08T15:30:01.289Z | TOOL CALL: exec_command]**

Ran a shell command to validate laboratory access and routing. Target(s): `check_lab_consistency.py`, `lab_consistency_303_retry_20260708.json`. Working directory: `/workspace`.

**[Step 1407 | 2026-07-08T15:30:01.370Z | TOOL RESULT: exec_command]**

Shell tool completed successfully; it returned 6 line(s) (97 bytes) to the model.

**[Step 1408 | 2026-07-08T15:30:01.418Z | TOOL RESULT: exec_command]**

Shell tool completed successfully; it returned 5 line(s) (90 bytes) to the model.

**[Step 1409 | 2026-07-08T15:30:09.327Z | MODEL]**

Lab context is `303Lab` and the permission check passes. I'm issuing the generated workflow template as a Lab 303 task.

**[Step 1410 | 2026-07-08T15:30:10.706Z | TOOL CALL: exec_command]**

Ran a shell command to generate and dispatch a laboratory task. Target(s): `generate_task.py`, `round3_dispatch_output_retry_20260708.json`. Working directory: `/workspace`.

**[Step 1411 | 2026-07-08T15:30:10.928Z | TOOL RESULT: exec_command]**

Shell tool completed successfully; it returned 8 line(s) (365 bytes) to the model. Reported task/template identifier(s): `2583606426075137`.

**[Step 1412 | 2026-07-08T15:30:54.286Z | MODEL]**

The task dispatch succeeded with task ID `2583606426075137`. I'm updating the project records so the previous generator blocker is marked resolved and the new template/task IDs are the canonical Round 3 dispatch record.

**[Step 1413 | 2026-07-08T15:30:54.287Z | TOOL CALL: exec_command]**

Read selected file content for inspection. Target(s): `round3_dispatch_log.json`. Working directory: `/workspace`.

**[Step 1414 | 2026-07-08T15:30:54.288Z | TOOL CALL: exec_command]**

Read selected file content for inspection. Target(s): `round3_dispatch_blocker.md`. Working directory: `/workspace`.

**[Step 1415 | 2026-07-08T15:30:54.289Z | TOOL CALL: exec_command]**

Read selected file content for inspection. Target(s): `round2_returned_document.md`. Working directory: `/workspace`.

**[Step 1416 | 2026-07-08T15:30:54.290Z | TOOL CALL: exec_command]**

Copied generated experiment artifacts within the workspace. Target(s): `round3_workflow_generator_output_retry_20260708.txt`, `round3_workflow_generator_output.txt`. Working directory: `/workspace`.

**[Step 1417 | 2026-07-08T15:30:54.336Z | TOOL RESULT: exec_command]**

Shell tool completed successfully; it returned 22 line(s) (1215 bytes) to the model. Reported long numeric identifier(s): `2566549749628928`.

**[Step 1418 | 2026-07-08T15:30:54.336Z | TOOL RESULT: exec_command]**

Shell tool completed successfully; it returned 54 line(s) (3039 bytes) to the model. Reported long numeric identifier(s): `2566549749628928`.

**[Step 1419 | 2026-07-08T15:30:54.337Z | TOOL RESULT: exec_command]**

Shell tool completed successfully; it returned 81 line(s) (5484 bytes) to the model.

**[Step 1420 | 2026-07-08T15:30:54.353Z | TOOL RESULT: exec_command]**

Shell tool completed successfully; it returned 2 line(s) (33 bytes) to the model.

## Conversation round 11

**[Step 1421 | 2026-07-08T15:45:45.753Z | USER]**

please continue

**[Step 1422 | 2026-07-08T15:46:03.854Z | MODEL]**

I'll finish the recordkeeping now: verify the new task on the platform, replace the stale blocked dispatch log with the successful dispatch, update the returned document, then leave the goal blocked waiting for Lab 303 executability/physical-completion feedback.

**[Step 1423 | 2026-07-08T15:46:03.857Z | TOOL CALL: exec_command]**

Ran a shell command to query available laboratory workflow templates. Target(s): `select_template_list.py`, `round3_template_lookup_2583605299937280.json`. Working directory: `/workspace`.

**[Step 1424 | 2026-07-08T15:46:03.858Z | TOOL CALL: exec_command]**

Ran a shell command to query laboratory task records and statuses. Target(s): `select_task_list.py`, `round3_task_2583606426075137_status.json`. Working directory: `/workspace`.

**[Step 1425 | 2026-07-08T15:46:03.858Z | TOOL CALL: exec_command]**

Read selected file content for inspection. Target(s): `round3_dispatch_output_retry_20260708.json`. Working directory: `/workspace`.

**[Step 1426 | 2026-07-08T15:46:03.859Z | TOOL CALL: exec_command]**

Read selected file content for inspection. Target(s): `round3_workflow_generator_output.txt`. Working directory: `/workspace`.

**[Step 1427 | 2026-07-08T15:46:04.062Z | TOOL RESULT: exec_command]**

Shell tool completed successfully; it returned 15 line(s) (315 bytes) to the model. Reported long numeric identifier(s): `2583605299937280`.

**[Step 1428 | 2026-07-08T15:46:04.074Z | TOOL RESULT: exec_command]**

Shell tool completed successfully; it returned 14 line(s) (296 bytes) to the model. Reported long numeric identifier(s): `2583606426075137`.

**[Step 1429 | 2026-07-08T15:46:04.074Z | TOOL RESULT: exec_command]**

Shell tool completed successfully; it returned 8 line(s) (365 bytes) to the model. Reported task/template identifier(s): `2583606426075137`.

**[Step 1430 | 2026-07-08T15:46:04.074Z | TOOL RESULT: exec_command]**

Shell tool completed successfully; it returned 13 line(s) (308 bytes) to the model. Reported task/template identifier(s): `2583605299937280`.

**[Step 1431 | 2026-07-08T15:46:15.776Z | REASONING SUMMARY]**

***Analyzing workflow request***

*I'm realizing there's a potential issue with using "round3 workflow request"; it's not the clean original route. Even though it generated success, I still haven't restored the earlier "cleaner original." I could mention the dispatched final audited 26-step split-addition workflow, which is fine. I might need to save the success template and the task lookup first. Using update_goal seems blocked, so I'll have to consider patching things after the file updates.*

**[Step 1432 | 2026-07-08T15:46:15.787Z | MODEL]**

The platform confirms both objects: template `2583605299937280` exists and the dispatched task `2583606426075137` is currently `未开始`. I'm replacing the stale blocker records with the successful dispatch record while preserving the old diagnostics as history.

**[Step 1433 | 2026-07-08T15:46:15.923Z | TOOL CALL: apply_patch]**

Applied a source patch to delete round3_dispatch_log.json.

**[Step 1434 | 2026-07-08T15:46:15.964Z | TOOL RESULT: apply_patch]**

Patch tool applied the requested trial-local file change successfully.

**[Step 1435 | 2026-07-08T15:46:38.809Z | TOOL CALL: apply_patch]**

Applied a source patch to add round3_dispatch_log.json.

**[Step 1436 | 2026-07-08T15:46:38.866Z | TOOL RESULT: apply_patch]**

Patch tool applied the requested trial-local file change successfully.

**[Step 1437 | 2026-07-08T15:46:46.539Z | MODEL]**

The canonical dispatch log now records the successful Round 3 template and task IDs. I'm updating the blocker note and returned document so they no longer say Round 3 is undispatched.

**[Step 1438 | 2026-07-08T15:46:51.964Z | TOOL CALL: apply_patch]**

Applied a source patch to update round3_dispatch_blocker.md.

**[Step 1439 | 2026-07-08T15:46:52.008Z | TOOL RESULT: apply_patch]**

Patch tool applied the requested trial-local file change successfully.

**[Step 1440 | 2026-07-08T15:47:04.247Z | TOOL CALL: apply_patch]**

Applied a source patch to update round3_constraint_check.md.

**[Step 1441 | 2026-07-08T15:47:04.287Z | TOOL RESULT: apply_patch]**

Patch tool applied the requested trial-local file change successfully.

**[Step 1442 | 2026-07-08T15:47:12.340Z | MODEL]**

The blocker file now explicitly marks the generator issue resolved and points to the dispatched task. I'm patching the returned Round 2 document's final status section with the successful Round 3 dispatch details.

**[Step 1443 | 2026-07-08T15:47:16.198Z | TOOL CALL: apply_patch]**

Applied a source patch to update round2_returned_document.md.

**[Step 1444 | 2026-07-08T15:47:16.235Z | TOOL RESULT: apply_patch]**

Patch tool applied the requested trial-local file change successfully.

**[Step 1445 | 2026-07-08T15:47:25.635Z | REASONING SUMMARY]**

***Considering updates for round 3 design***

*I'm thinking about whether I need to update the round 3 design. Maybe I should add the dispatch status as optional. I see that the dispatch blocker is done, but I'm unsure if I need to check with git — probably not necessary. I wonder if setting the goal to "blocked" is needed since it's already blocked. I should also prepare to run jq to read the final logs before moving forward to update the goal.*

**[Step 1446 | 2026-07-08T15:47:26.378Z | MODEL]**

I'm doing a final consistency check on the saved JSON/log files and then I'll set the goal state to blocked, which is the correct state after dispatch while waiting for Lab 303 executability feedback.

**[Step 1447 | 2026-07-08T15:47:26.379Z | TOOL CALL: exec_command]**

Ran a shell command to inspect, generate or validate experiment artifacts. Target(s): `round3_dispatch_log.json`, `round3_chemical_specifications.json`, `round3_workflow_request.json`. Working directory: `/workspace`.

**[Step 1448 | 2026-07-08T15:47:26.380Z | TOOL CALL: exec_command]**

Read selected file content for inspection. Target(s): `round3_dispatch_log.json`. Working directory: `/workspace`.

**[Step 1449 | 2026-07-08T15:47:26.380Z | TOOL CALL: exec_command]**

Read selected file content for inspection. Target(s): `round2_returned_document.md`. Working directory: `/workspace`.

**[Step 1450 | 2026-07-08T15:47:26.381Z | TOOL CALL: exec_command]**

Read selected file content for inspection. Target(s): `round3_constraint_check.md`. Working directory: `/workspace`.

**[Step 1451 | 2026-07-08T15:47:26.444Z | TOOL RESULT: exec_command]**

Shell tool completed successfully; it returned 2 line(s) (11 bytes) to the model.

**[Step 1452 | 2026-07-08T15:47:26.444Z | TOOL RESULT: exec_command]**

Shell tool completed successfully; it returned 38 line(s) (1817 bytes) to the model. Reported task/template identifier(s): `2583605299937280`, `2583606426075137`.

**[Step 1453 | 2026-07-08T15:47:26.444Z | TOOL RESULT: exec_command]**

Shell tool completed successfully; it returned 41 line(s) (2199 bytes) to the model. Reported task/template identifier(s): `2583605299937280`, `2583606426075137`.

**[Step 1454 | 2026-07-08T15:47:26.445Z | TOOL RESULT: exec_command]**

Shell tool completed successfully; it returned 31 line(s) (2680 bytes) to the model. Reported long numeric identifier(s): `2583605299937280`, `2583606426075137`.

**[Step 1455 | 2026-07-08T15:47:41.301Z | MODEL]**

The saved records are consistent and the task is verified as `未开始`. I'm marking the goal blocked now, because the next step requires your feedback from Lab 303 on executability or physical completion.

**[Step 1456 | 2026-07-08T15:47:41.927Z | TOOL CALL: update_goal]**

Updated the autonomous-research goal status to `blocked`.

**[Step 1457 | 2026-07-08T15:47:41.982Z | TOOL RESULT: update_goal]**

Goal service returned a structured response. Status: `blocked`.

**[Step 1458 | 2026-07-08T15:47:54.404Z | MODEL]**

Round 3 is now successfully generated and dispatched to 303 实验室.

- Workflow template ID: `2583605299937280`
- Lab 303 task ID: `2583606426075137`
- Task name: `任务_Lab303_R3_lowCarbon_compositionRefine_NiFeCoOx_C_UOR_20260708`
- Verified task status: `未开始`

I updated the project records under `/workspace/lab303_uor_nicofe_closed_loop`:

- Canonical dispatch log: [round3_dispatch_log.json](/workspace/lab303_uor_nicofe_closed_loop/round3_dispatch_log.json)
- Successful generator output:
[round3_workflow_generator_output.txt](/workspace/lab303_uor_nicofe_closed_loop/round3_workflow_generator_output.txt)
- Dispatch output:
[round3_dispatch_output_retry_20260708.json](/workspace/lab303_uor_nicofe_closed_loop/round3_dispatch_output_retry_20260708.json)

- Template lookup:
[round3_template_lookup_2583605299937280.json](/workspace/lab303_uor_nicofe_closed_loop/round3_template_lookup_2583605299937280.json)
- Task status lookup:
[round3_task_2583606426075137_status.json](/workspace/lab303_uor_nicofe_closed_loop/round3_task_2583606426075137_status.json)
- Updated returned document: [round2_returned_document.md](/workspace/lab303_uor_nicofe_closed_loop/round2_returned_document.md)
- Updated blocker/resolution note: [round3_dispatch_blocker.md](/workspace/lab303_uor_nicofe_closed_loop/round3_dispatch_blocker.md)

I also set the goal back to `blocked`, which is the required state after dispatch while waiting for your Lab 303 feedback on whether task `2583606426075137` is executable or physically completed.

## Conversation round 12

**[Step 1459 | 2026-07-08T16:16:41.650Z | USER]**

Not executable.
1. It is unreasonable to use a 5 mL pipetting platform for microliter-level liquid addition.
2. It is unreasonable to split the liquid addition operation at the same workstation into multiple separate steps.

**[Step 1460 | 2026-07-08T16:16:56.216Z | REASONING SUMMARY]**

***Restoring workflow request***

*It looks like the earlier V2 bug is fixed, so it should work now. I think I need to restore the original cleaner Round 3 request from before the V1/5ml workaround. I have the output for the round3 workflow from before V1, but I'm missing the request itself. Finding that would help keep everything organized! Let's keep that in mind for the next steps.*

**[Step 1461 | 2026-07-08T16:16:57.491Z | MODEL]**

I'll revise the Round 3 workflow to match the executability feedback: microliter-scale additions go back to the 1 mL liquid-handling station, and the precursor/carbon/KOH addition is consolidated into one liquid-addition step. The 5 mL platform will only be used for true milliliter-volume solvent additions.

**[Step 1462 | 2026-07-08T16:16:57.491Z | TOOL CALL: exec_command]**

Read selected file content for inspection. Target(s): `SKILL.md`. Working directory: `/workspace`.

**[Step 1463 | 2026-07-08T16:16:57.493Z | TOOL CALL: exec_command]**

Read selected file content for inspection. Target(s): `SKILL.md`. Working directory: `/workspace`.

**[Step 1464 | 2026-07-08T16:16:57.499Z | TOOL CALL: exec_command]**

Read selected file content for inspection. Target(s): `SKILL.md`. Working directory: `/workspace`.

**[Step 1465 | 2026-07-08T16:16:57.553Z | TOOL RESULT: exec_command]**

Shell tool completed successfully; it returned 170 line(s) (8332 bytes) to the model. Reported long numeric identifier(s): `1427568512205824`, `1834679312417792`, `1710453843264512`.

**[Step 1466 | 2026-07-08T16:16:57.553Z | TOOL RESULT: exec_command]**

Shell tool completed successfully; it returned 95 line(s) (2714 bytes) to the model. Reported long numeric identifier(s): `1427568512205824`.

**[Step 1467 | 2026-07-08T16:16:57.553Z | TOOL RESULT: exec_command]**

Shell tool completed successfully; it returned 116 line(s) (4054 bytes) to the model.

**[Step 1468 | 2026-07-08T16:17:13.216Z | REASONING SUMMARY]**

***Revising workflow strategy***

*I need to create a revised workflow since the previous task is deemed non-executable. It seems I might consider using a prior rejected task template, but it's important to document feedback and cancel any previous task variants that aren't viable. I'll think about creating a new revision, possibly named "R3 revised1". I want to avoid using 5ml for microliter additions and splitting steps in the same workstation. Also, I plan to audit the current designs and specified skill sets related to liquid handling.*

**[Step 1469 | 2026-07-08T16:17:15.080Z | MODEL]**

The right revision is not a new scientific design; it is an execution correction. I'm keeping the same six materials and test plan, but changing the workflow back to one consolidated 1 mL liquid-handling addition step for precursor/KOH/carbon dosing.

**[Step 1470 | 2026-07-08T16:17:15.082Z | TOOL CALL: exec_command]**

Read selected file content for inspection. Target(s): `SKILL.md`. Working directory: `/workspace`.

**[Step 1471 | 2026-07-08T16:17:15.082Z | TOOL CALL: exec_command]**

Read selected file content for inspection. Target(s): `Liquid_Handling_Station_1ml_V2_audit.md`. Working directory: `/workspace`.

**[Step 1472 | 2026-07-08T16:17:15.083Z | TOOL CALL: exec_command]**

Read selected file content for inspection. Target(s): `SKILL.md`. Working directory: `/workspace`.

**[Step 1473 | 2026-07-08T16:17:15.085Z | TOOL CALL: exec_command]**

Read selected file content for inspection. Target(s): `Liquid_Handling_Station_5ml_V1_audit.md`. Working directory: `/workspace`.

**[Step 1474 | 2026-07-08T16:17:15.133Z | TOOL RESULT: exec_command]**

Shell tool completed successfully; it returned 95 line(s) (6547 bytes) to the model. Reported long numeric identifier(s): `2002186824385539`.

**[Step 1475 | 2026-07-08T16:17:15.133Z | TOOL RESULT: exec_command]**

Shell tool completed successfully; it returned 36 line(s) (987 bytes) to the model. Reported long numeric identifier(s): `2002186824385539`.

**[Step 1476 | 2026-07-08T16:17:15.149Z | TOOL RESULT: exec_command]**

Shell tool completed successfully; it returned 50 line(s) (4283 bytes) to the model. Reported long numeric identifier(s): `1435269745574912`.

**[Step 1477 | 2026-07-08T16:17:15.149Z | TOOL RESULT: exec_command]**

Shell tool completed successfully; it returned 16 line(s) (356 bytes) to the model. Reported long numeric identifier(s): `1435269745574912`.

**[Step 1478 | 2026-07-08T16:17:44.787Z | REASONING SUMMARY 1/3]**

***Evaluating workflow modifications***

*I need to consider reading all relevant documents, but maybe it's not necessary to look into other skill files. I might only need to modify the workflow for the current request. To do this, I'll generate a new JSON file using jq or by applying a patch. It's best not to create or edit files with cat, but programmatic JSON generation via Python seems acceptable. I'll aim to keep everything organized and use the most efficient method.*

**[Step 1479 | 2026-07-08T16:17:44.787Z | REASONING SUMMARY 2/3]**

***Constructing feedback JSON workflow***

*I'm considering using jq with some temp moves since it's not forbidden. It looks like I need to create feedback JSON with an apply patch. I want to restore the original round3 workflow, specifically with steps 3, 4, 10, etc. I can derive the new workflow from either the current final version or the round2 workflow. Perhaps recreating from scratch could simplify things. I want the workflow sequence to include specific material and sample arrangements while managing the source totals carefully without exceeding limits.*

**[Step 1480 | 2026-07-08T16:17:44.787Z | REASONING SUMMARY 3/3]**

***Considering source bottle reduction***

*I'm thinking about whether I need to reduce the number of source bottles. Maybe there's a way to streamline things and make the process simpler. It's possible that by cutting down on the number of bottles, I can minimize complexity and make my workflow easier. I'll consider the benefits of having fewer sources while ensuring I still meet all requirements for the materials needed in the process. It's all about efficiency here!*

**[Step 1481 | 2026-07-08T16:17:47.115Z | MODEL]**

I'm building a revised Round 3 workflow request rather than editing the dispatched task in place. The rejected task ID will stay in the log as non-executable, and the revised task will get its own template and task IDs.

**[Step 1482 | 2026-07-08T16:17:47.116Z | TOOL CALL: exec_command]**

Ran a shell command to inspect, generate or validate experiment artifacts. Target(s): `round3_workflow_request.json`. Working directory: `/workspace`.

**[Step 1483 | 2026-07-08T16:17:47.117Z | TOOL CALL: exec_command]**

Ran a shell command to inspect, generate or validate experiment artifacts. Target(s): `round3_workflow_request.json`. Working directory: `/workspace`.

**[Step 1484 | 2026-07-08T16:17:47.118Z | TOOL CALL: exec_command]**

Ran a shell command to inspect, generate or validate experiment artifacts. Target(s): `round3_workflow_request.json`. Working directory: `/workspace`.

**[Step 1485 | 2026-07-08T16:17:47.119Z | TOOL CALL: exec_command]**

Ran a shell command to inspect, generate or validate experiment artifacts. Target(s): `round3_workflow_request.json`. Working directory: `/workspace`.

**[Step 1486 | 2026-07-08T16:17:47.177Z | TOOL RESULT: exec_command]**

Shell tool completed successfully; it returned 438 line(s) (10797 bytes) to the model. Reported long numeric identifier(s): `1526864804185088`, `1654567694238720`, `1656241422599168`, `1664870018155520`, `1656282895188992`, `1390252863751168`.

**[Step 1487 | 2026-07-08T16:17:47.185Z | TOOL RESULT: exec_command]**

Shell tool completed successfully; it returned 188 line(s) (6224 bytes) to the model. Reported long numeric identifier(s): `2468876630032384`.

**[Step 1488 | 2026-07-08T16:17:47.192Z | TOOL RESULT: exec_command]**

Shell tool completed successfully; it returned 176 line(s) (4098 bytes) to the model. Reported long numeric identifier(s): `1327807622448128`, `1418510906065920`, `1213184425559040`, `1435269745574912`.

**[Step 1489 | 2026-07-08T16:17:47.204Z | TOOL RESULT: exec_command]**

Shell tool completed successfully; it returned 40 line(s) (654 bytes) to the model. Reported long numeric identifier(s): `1427568512205824`, `2270587971011585`.

**[Step 1490 | 2026-07-08T16:17:56.905Z | REASONING SUMMARY]**

***Creating workflow file***

*I need to generate a file and may use Python for that. I think about using the V2 open/close feature. I've got to ensure the source definitions are correct, possibly using bottle1 for water and bottle2 for redisp to avoid confusion with volumes across steps. It seems I should make a new file called round3_revised1_workflow_request.json, and update the plan name to reflect a low carbon focus. Finally, I'll run Python to construct everything.*

**[Step 1491 | 2026-07-08T16:17:57.913Z | MODEL]**

I'm creating `round3_revised1_workflow_request.json` with 23 steps: one 1 mL addition step for all sub-mL feeds, one 5 mL water-adjustment step, one 5 mL redispersion step, and the same post-processing/characterization/electrochemistry path.

**[Step 1492 | 2026-07-08T16:18:12.088Z | TOOL CALL: exec_command]**

Ran Python to parse, transform or validate structured experiment data. Target(s): `lab303_uor_nicofe_closed_loop`, `mL`, `v`, `round3_revised1_workflow_request.json`. Working directory: `/workspace`.

**[Step 1493 | 2026-07-08T16:18:12.193Z | TOOL RESULT: exec_command]**

Shell tool completed successfully; it returned 24 line(s) (1386 bytes) to the model. Reported long numeric identifier(s): `1427568512205824`, `2002186824385539`, `1435269745574912`, `2270587971011585`, `1327807622448128`, `1213184425559040`.

**[Step 1494 | 2026-07-08T16:18:25.001Z | REASONING SUMMARY]**

***Planning about dispatching***

*I think I need to note that the station says the total shouldn't exceed 3 mL. I'll generate a workflow and use a unique plan name for the dispatch process. Maybe I could template the source label 303Lab. I need to verify everything and also record the previous non-executability. It makes sense to add some JSON feedback and apply a patch. I don't feel the need to ask anyone for clarification.*

**[Step 1495 | 2026-07-08T16:18:30.480Z | MODEL]**

The revised request passes the structural check. I'm running source-volume audits now, then I'll send it through the workflow generator and dispatch if it produces a template.

**[Step 1496 | 2026-07-08T16:18:30.480Z | TOOL CALL: exec_command]**

Ran a shell command to inspect, generate or validate experiment artifacts. Target(s): `round3_revised1_workflow_request.json`. Working directory: `/workspace`.

**[Step 1497 | 2026-07-08T16:18:30.480Z | TOOL CALL: exec_command]**

Ran a shell command to inspect, generate or validate experiment artifacts. Target(s): `round3_revised1_workflow_request.json`. Working directory: `/workspace`.

**[Step 1498 | 2026-07-08T16:18:30.481Z | TOOL CALL: exec_command]**

Ran a shell command to inspect, generate or validate experiment artifacts. Target(s): `round3_revised1_workflow_request.json`. Working directory: `/workspace`.

**[Step 1499 | 2026-07-08T16:18:30.481Z | TOOL CALL: exec_command]**

Ran a shell command to inspect, generate or validate experiment artifacts. Target(s): `round3_revised1_workflow_request.json`. Working directory: `/workspace`.

**[Step 1500 | 2026-07-08T16:18:30.537Z | TOOL RESULT: exec_command]**

Shell tool completed successfully; it returned 8 line(s) (343 bytes) to the model.

**[Step 1501 | 2026-07-08T16:18:30.550Z | TOOL RESULT: exec_command]**

Shell tool completed successfully; it returned 3 line(s) (100 bytes) to the model.

**[Step 1502 | 2026-07-08T16:18:30.550Z | TOOL RESULT: exec_command]**

Shell tool completed successfully; it returned 21 line(s) (354 bytes) to the model. Reported long numeric identifier(s): `2002186824385539`, `1435269745574912`.

**[Step 1503 | 2026-07-08T16:18:30.561Z | TOOL RESULT: exec_command]**

Shell tool completed successfully; it returned 3 line(s) (76 bytes) to the model.

**[Step 1504 | 2026-07-08T16:18:36.048Z | MODEL]**

The audit matches the feedback: the 1 mL station handles all ≤0.75 mL additions in one consolidated step, while the 5 mL station is only used for 2.35-4.8 mL solvent additions. I'm generating the revised workflow now.

**[Step 1505 | 2026-07-08T16:18:37.086Z | TOOL CALL: exec_command]**

Ran Python to parse, transform or validate structured experiment data. Target(s): `generate.py`, `round3_revised1_workflow_request.json)`, `round3_revised1_workflow_generator_output.txt`. Working directory: `/workspace`.

**[Step 1506 | 2026-07-08T16:18:38.334Z | TOOL RESULT: exec_command]**

Shell tool completed successfully; it returned 13 line(s) (308 bytes) to the model. Reported task/template identifier(s): `2583701699788800`.

**[Step 1507 | 2026-07-08T16:18:46.944Z | REASONING SUMMARY]**

***Considering dispatch and documentation***

*I need to dispatch something, but I'm wondering if I should update the documents first. It feels important to have everything aligned before moving forward, even though the lab consistency might not be strictly necessary. I guess it's about balancing the urgency of the dispatch with the clarity that updated docs could provide. Still, it seems like going ahead with the dispatch might be the best course of action.*

**[Step 1508 | 2026-07-08T16:18:48.071Z | MODEL]**

The revised workflow generated successfully as template `2583701699788800`. I'm verifying Lab 303 context and dispatching this revised template now.

**[Step 1509 | 2026-07-08T16:18:48.072Z | TOOL CALL: exec_command]**

Ran a shell command to identify the active laboratory. Target(s): `get_current_lab_label.py`, `current_lab_label_round3_revised1.json`. Working directory: `/workspace`.

**[Step 1510 | 2026-07-08T16:18:48.075Z | TOOL CALL: exec_command]**

Ran a shell command to validate laboratory access and routing. Target(s): `check_lab_consistency.py`, `lab_consistency_303_round3_revised1.json`. Working directory: `/workspace`.

**[Step 1511 | 2026-07-08T16:18:48.214Z | TOOL RESULT: exec_command]**

Shell tool completed successfully; it returned 6 line(s) (97 bytes) to the model.

**[Step 1512 | 2026-07-08T16:18:48.214Z | TOOL RESULT: exec_command]**

Shell tool completed successfully; it returned 5 line(s) (90 bytes) to the model.

**[Step 1513 | 2026-07-08T16:18:56.918Z | TOOL CALL: exec_command]**

Ran a shell command to generate and dispatch a laboratory task. Target(s): `generate_task.py`, `round3_revised1_dispatch_output.json`. Working directory: `/workspace`.

**[Step 1514 | 2026-07-08T16:18:57.097Z | TOOL RESULT: exec_command]**

Shell tool completed successfully; it returned 8 line(s) (383 bytes) to the model. Reported task/template identifier(s): `2583702316023809`.

**[Step 1515 | 2026-07-08T16:19:06.197Z | REASONING SUMMARY]**

***Verifying tasks and updates***

*I need to verify the task and template, then update the logs and documents. I should note that the previous task was rejected as non-executable. There's a chance I can't cancel it, but the user didn't ask about that. Maybe there's an old task that hasn't started yet. Since we don't have a cancel tool, I need to record that it's non-executable and consider creating a rejection feedback JSON. I'll also check the status of the new task.*

**[Step 1516 | 2026-07-08T16:19:08.706Z | MODEL]**

The revised task dispatch succeeded with task ID `2583702316023809`. I'm verifying the template/task records and updating the workspace dispatch log so the previous non-executable task is clearly superseded.

**[Step 1517 | 2026-07-08T16:19:08.708Z | TOOL CALL: exec_command]**

Ran a shell command to query available laboratory workflow templates. Target(s): `select_template_list.py`, `round3_revised1_template_lookup_2583701699788800.json`. Working directory: `/workspace`.

**[Step 1518 | 2026-07-08T16:19:08.710Z | TOOL CALL: exec_command]**

Ran a shell command to query laboratory task records and statuses. Target(s): `select_task_list.py`, `round3_revised1_task_2583702316023809_status.json`. Working directory: `/workspace`.

**[Step 1519 | 2026-07-08T16:19:08.714Z | TOOL CALL: exec_command]**

Ran a shell command to query laboratory task records and statuses. Target(s): `select_task_list.py`, `round3_rejected_task_2583606426075137_status_after_revised1.json`. Working directory: `/workspace`.

**[Step 1520 | 2026-07-08T16:19:08.714Z | TOOL CALL: exec_command]**

Read selected file content for inspection. Target(s): `round3_dispatch_log.json`. Working directory: `/workspace`.

**[Step 1521 | 2026-07-08T16:19:08.890Z | TOOL RESULT: exec_command]**

Shell tool completed successfully; it returned 15 line(s) (324 bytes) to the model. Reported long numeric identifier(s): `2583701699788800`.

**[Step 1522 | 2026-07-08T16:19:08.933Z | TOOL RESULT: exec_command]**

Shell tool completed successfully; it returned 14 line(s) (305 bytes) to the model. Reported long numeric identifier(s): `2583702316023809`.

**[Step 1523 | 2026-07-08T16:19:08.935Z | TOOL RESULT: exec_command]**

Shell tool completed successfully; it returned 14 line(s) (296 bytes) to the model. Reported long numeric identifier(s): `2583606426075137`.

**[Step 1524 | 2026-07-08T16:19:08.935Z | TOOL RESULT: exec_command]**

Shell tool completed successfully; it returned 38 line(s) (1817 bytes) to the model. Reported task/template identifier(s): `2583605299937280`, `2583606426075137`.

**[Step 1525 | 2026-07-08T16:19:24.302Z | MODEL]**

The revised task is verified as `未开始`; the earlier non-executable task is also still `未开始`, so I'm recording it as superseded and making the revised task the canonical Round 3 dispatch record.

**[Step 1526 | 2026-07-08T16:19:28.680Z | TOOL CALL: apply_patch]**

Applied a source patch to add round3_revised1_rejection_feedback.json.

**[Step 1527 | 2026-07-08T16:19:28.727Z | TOOL RESULT: apply_patch]**

Patch tool applied the requested trial-local file change successfully.

**[Step 1528 | 2026-07-08T16:19:35.541Z | MODEL]**

I'm adding concise revised1 design and constraint files now so the new task's workflow is auditable without referring back through the failed workaround history.

**[Step 1529 | 2026-07-08T16:19:44.877Z | TOOL CALL: apply_patch]**

Applied a source patch to add round3_revised1_design.md.

**[Step 1530 | 2026-07-08T16:19:44.919Z | TOOL RESULT: apply_patch]**

Patch tool applied the requested trial-local file change successfully.

**[Step 1531 | 2026-07-08T16:20:10.990Z | TOOL CALL: apply_patch]**

Applied a source patch to add round3_revised1_chemical_specifications.json.

**[Step 1532 | 2026-07-08T16:20:11.034Z | TOOL RESULT: apply_patch]**

Patch tool applied the requested trial-local file change successfully.

**[Step 1533 | 2026-07-08T16:20:16.609Z | MODEL]**

The revised design and chemical-specification files are in place. I'm adding the revised1 constraint check with the two rejected issues explicitly marked as corrected.

**[Step 1534 | 2026-07-08T16:20:24.633Z | TOOL CALL: apply_patch]**

Applied a source patch to add round3_revised1_constraint_check.md.

**[Step 1535 | 2026-07-08T16:20:24.673Z | TOOL RESULT: apply_patch]**

Patch tool applied the requested trial-local file change successfully.

**[Step 1536 | 2026-07-08T16:20:37.171Z | REASONING SUMMARY]**

***Updating dispatch log***

*I need to update the dispatch log, which means using the apply patch method to either delete or add entries. It sounds simple, but I want to make sure I'm doing it correctly. I'll double-check the current log to see what changes need to be made. It's crucial to keep everything organized and accurate in the log, so I'll pay attention to detail during this process!*

**[Step 1537 | 2026-07-08T16:20:37.850Z | MODEL]**

I'm replacing the canonical dispatch log so it points to the revised1 task, and retaining the earlier task as superseded/non-executable.

**[Step 1538 | 2026-07-08T16:20:39.244Z | TOOL CALL: apply_patch]**

Applied a source patch to delete round3_dispatch_log.json.

**[Step 1539 | 2026-07-08T16:20:39.286Z | TOOL RESULT: apply_patch]**

Patch tool applied the requested trial-local file change successfully.

**[Step 1540 | 2026-07-08T16:20:58.620Z | TOOL CALL: apply_patch]**

Applied a source patch to add round3_dispatch_log.json.

**[Step 1541 | 2026-07-08T16:20:58.664Z | TOOL RESULT: apply_patch]**

Patch tool applied the requested trial-local file change successfully.

**[Step 1542 | 2026-07-08T16:21:08.567Z | REASONING SUMMARY]**

***Updating document section***

*I need to make an update to the final section of the round2_returned document. It seems like adding revised1 will be necessary, so I'll go ahead and do that. Using a patch might be the best way to implement this change smoothly. I'm making sure to keep everything organized as I work on it, so it's clear for anyone who looks at it later. Let's get this updated!*

**[Step 1543 | 2026-07-08T16:21:08.883Z | MODEL]**

The dispatch log now points to revised1 task `2583702316023809`. I'm updating the returned document so the latest task status is unambiguous.

**[Step 1544 | 2026-07-08T16:21:11.986Z | TOOL CALL: apply_patch]**

Applied a source patch to update round2_returned_document.md.

**[Step 1545 | 2026-07-08T16:21:12.020Z | TOOL RESULT: apply_patch]**

Patch tool applied the requested trial-local file change successfully.

**[Step 1546 | 2026-07-08T16:21:22.573Z | MODEL]**

I'm running the last consistency checks on the revised files, then I'll leave the goal blocked waiting for Lab 303 executability feedback on the revised task.

**[Step 1547 | 2026-07-08T16:21:22.574Z | TOOL CALL: exec_command]**

Ran a shell command to inspect, generate or validate experiment artifacts. Target(s): `round3_dispatch_log.json`, `round3_revised1_chemical_specifications.json`, `round3_revised1_rejection_feedback.json`, `round3_revised1_workflow_request.json`. Working directory: `/workspace`.

**[Step 1548 | 2026-07-08T16:21:22.575Z | TOOL CALL: exec_command]**

Read selected file content for inspection. Target(s): `round3_dispatch_log.json`. Working directory: `/workspace`.

**[Step 1549 | 2026-07-08T16:21:22.577Z | TOOL CALL: exec_command]**

Read selected file content for inspection. Target(s): `round2_returned_document.md`. Working directory: `/workspace`.

**[Step 1550 | 2026-07-08T16:21:22.577Z | TOOL CALL: exec_command]**

Read selected file content for inspection. Target(s): `round3_revised1_workflow_generator_output.txt`, `round3_revised1_dispatch_output.json`. Working directory: `/workspace`.

**[Step 1551 | 2026-07-08T16:21:22.634Z | TOOL RESULT: exec_command]**

Shell tool completed successfully; it returned 2 line(s) (11 bytes) to the model.

**[Step 1552 | 2026-07-08T16:21:22.634Z | TOOL RESULT: exec_command]**

Shell tool completed successfully; it returned 38 line(s) (1844 bytes) to the model. Reported task/template identifier(s): `2583701699788800`, `2583702316023809`, `2583605299937280`, `2583606426075137`.

**[Step 1553 | 2026-07-08T16:21:22.635Z | TOOL RESULT: exec_command]**

Shell tool completed successfully; it returned 56 line(s) (3255 bytes) to the model. Reported task/template identifier(s): `2583605299937280`, `2583606426075137`, `2583701699788800`, `2583702316023809`.

**[Step 1554 | 2026-07-08T16:21:22.635Z | TOOL RESULT: exec_command]**

Shell tool completed successfully; it returned 20 line(s) (691 bytes) to the model. Reported task/template identifier(s): `2583701699788800`, `2583702316023809`.

**[Step 1555 | 2026-07-08T16:21:33.868Z | MODEL]**

The revised files are valid and the dispatch record is consistent. I'm setting the goal back to blocked now, pending executability/physical-completion feedback for task `2583702316023809`.

**[Step 1556 | 2026-07-08T16:21:33.869Z | TOOL CALL: update_goal]**

Updated the autonomous-research goal status to `blocked`.

**[Step 1557 | 2026-07-08T16:21:33.907Z | TOOL RESULT: update_goal]**

Goal service returned a structured response. Status: `blocked`.

**[Step 1558 | 2026-07-08T16:21:44.010Z | MODEL]**

Revised Round 3 has been corrected and dispatched.

The previously dispatched task `2583606426075137` is recorded as **not executable / superseded** for the two reasons you gave. The revised workflow fixes both issues:

- All sub-milliliter reagent additions, including metal precursors, KOH, and carbon dispersion, are now in **one consolidated `移液平台_1ml_V2` addition step**.
- `移液平台_5ml_V1` is used only for true milliliter-scale solvent additions:
  - DI water synthesis-volume adjustment: `2.35-2.85 mL`
  - 80:20 water:isopropanol redispersion: `4.8 mL`
- The repeated split addition steps at the same workstation were removed.

New dispatched task:

- Revised workflow template ID: `2583701699788800`
- Revised Lab 303 task ID: `2583702316023809`
- Task name: `任务_Lab303_R3_revised1_lowCarbon_compositionRefine_NiFeCoOx_C_UOR_20260708`
- Verified status: `未开始`

Updated files:

- Canonical dispatch log: [round3_dispatch_log.json](/workspace/lab303_uor_nicofe_closed_loop/round3_dispatch_log.json)
- Revised workflow request: [round3_revised1_workflow_request.json](/workspace/lab303_uor_nicofe_closed_loop/round3_revised1_workflow_request.json)
- Revised design: [round3_revised1_design.md](/workspace/lab303_uor_nicofe_closed_loop/round3_revised1_design.md)
- Revised chemical specs: [round3_revised1_chemical_specifications.json](/workspace/lab303_uor_nicofe_closed_loop/round3_revised1_chemical_specifications.json)
- Revised constraint check: [round3_revised1_constraint_check.md](/workspace/lab303_uor_nicofe_closed_loop/round3_revised1_constraint_check.md)
- Rejection feedback record: [round3_revised1_rejection_feedback.json](/workspace/lab303_uor_nicofe_closed_loop/round3_revised1_rejection_feedback.json)
- Updated returned document: [round2_returned_document.md](/workspace/lab303_uor_nicofe_closed_loop/round2_returned_document.md)

I set the goal back to `blocked`, waiting for Lab 303 feedback on whether revised task `2583702316023809` is executable or physically completed.

## Conversation round 13

**[Step 1559 | 2026-07-10T10:33:19.755Z | USER]**

The third-round experiment has been completed. The main experimental task ID is 2583702316023809; the UV–Vis characterization results for five samples are provided under task ID 2587709908779008; and the supplemental electrochemical test and UV–Vis measurement for the sixth sample are provided under task ID 2587842717319169. Please use the relevant skill to retrieve the experimental data and results from the three tasks, combine them with the process photos below, evaluate the characterization and electrochemical differences among the samples in this round, and design the next round of experiments. Once an executable plan is prepared, dispatch the new experiment to Laboratory 303 according to the established workflow.

The results have been processed as images for visual inspection:
/workspace/feedback/round3/Third_Round_Experiment_Feedback.png
/workspace/feedback/round3/XRD_drop_cast_substrate_photo.png
/workspace/feedback/round3/Infrared_spectroscopy_drop_casting_photo.png

**[Step 1560 | 2026-07-10T10:33:36.483Z | REASONING SUMMARY]**

***Thinking about analysis files***

*I'm considering using Python for writing structured analysis files, which seems like a good approach. I have to remember not to create files with cat and instead use scripts or apply patch. I might generate reports using Python, and I wonder if I could use heredoc for that. I also need to inspect images with view_image and download data. I have to analyze various formats, including XRD, FTIR, echem, and UV, and check for existing round2 analysis scripts.*

**[Step 1561 | 2026-07-10T10:33:37.530Z | MODEL]**

I'll use the Lab 303 operation skill to retrieve all three task result sets, then analyze the returned spectra/electrochemistry together with the three process photos before designing and dispatching Round 4.

**[Step 1562 | 2026-07-10T10:33:37.531Z | TOOL CALL: exec_command]**

Read selected file content for inspection. Target(s): `SKILL.md`. Working directory: `/workspace`.

**[Step 1563 | 2026-07-10T10:33:37.532Z | TOOL CALL: exec_command]**

Read selected file content for inspection. Target(s): `SKILL.md`. Working directory: `/workspace`.

**[Step 1564 | 2026-07-10T10:33:37.534Z | TOOL CALL: exec_command]**

Read selected file content for inspection. Target(s): `SKILL.md`. Working directory: `/workspace`.

**[Step 1565 | 2026-07-10T10:33:37.534Z | TOOL CALL: get_goal]**

Queried the current autonomous-research goal state and remaining budget.

**[Step 1566 | 2026-07-10T10:33:37.595Z | TOOL RESULT: exec_command]**

Shell tool completed successfully; it returned 116 line(s) (4054 bytes) to the model.

**[Step 1567 | 2026-07-10T10:33:37.595Z | TOOL RESULT: exec_command]**

Shell tool completed successfully; it returned 170 line(s) (8332 bytes) to the model. Reported long numeric identifier(s): `1427568512205824`, `1834679312417792`, `1710453843264512`.

**[Step 1568 | 2026-07-10T10:33:37.596Z | TOOL RESULT: exec_command]**

Shell tool completed successfully; it returned 95 line(s) (2714 bytes) to the model. Reported long numeric identifier(s): `1427568512205824`.

**[Step 1569 | 2026-07-10T10:33:37.596Z | TOOL RESULT: get_goal]**

Goal service returned a structured response. Status: `blocked`.

**[Step 1570 | 2026-07-10T10:34:16.460Z | MODEL]**

**Handoff Summary: Lab 303 Autonomous UOR Catalyst Loop**

**Current Goal**
Autonomously run iterative Lab 303 experiments for alkaline urea oxidation reaction using Ni-Fe-Co oxyhydroxide/carbon catalysts. Only physical data returned from Lab 303 counts. Each round must include synthesis, post-processing, testing, and characterization. All files live under `/workspace/lab303_uor_nicofe_closed_loop`.

**Skills Required**
Use these skills when resuming:
- `lab-operation`: retrieve task results, query task/template status, dispatch tasks.
- `chemistry-experiment-workstation`: workstation constraints and audit rules.
- `workflow-generator`: generate Lab 303 workflow templates from JSON requests.

Read required SKILL files before tool actions per instruction.

**Important User Workflow**
Loop is: design → dispatch to Lab 303 and record task ID → set goal `blocked` → wait for user executability / physical completion feedback. If not executable, revise and dispatch again. If completed, retrieve returned data and iterate. Goal is currently `blocked`.

\---

\## Project Location And Key Files

Main project:
`/workspace/lab303_uor_nicofe_closed_loop`

Inventory:
`/workspace/lab303_uor_nicofe_closed_loop/reagent_inventory.csv`

Round 2 feedback PDF/images:
- `/workspace/feedback/round2/Second_Round_Experiment_Feedback_EN.pdf`
- `/workspace/feedback/round2/pages/Second_Round_Experiment_Feedback_EN_page-1.png`
- `/workspace/feedback/round2/pages/Second_Round_Experiment_Feedback_EN_page-2.png`

Round 3 feedback/process images to use now:
- `/workspace/feedback/round3/Third_Round_Experiment_Feedback.png`
- `/workspace/feedback/round3/XRD_drop_cast_substrate_photo.png`
- `/workspace/feedback/round3/Infrared_spectroscopy_drop_casting_photo.png`

---

## Round 1 Summary

Physical task ID:
`2562915314040832`

Result retrieval succeeded despite Lab status `错误`.
Round 1 failed physically: poor solid recovery, sparse/uneven XRD/FTIR/carbon-paper deposition, invalid electrochemistry.

Important takeaway:
Need higher recoverable solids and improved sample loading.

---

## Round 2 Summary

Completed tasks:
- Main interrupted task: `2567667771441152`
- Supplemental after interruption: `2568957291102208`

Fresh result requery files:
- `round2_main_task_2567667771441152_result_requery_20260708.json`
- `round2_main_task_2567667771441152_status_requery_20260708.json`
- `round2_supplemental_task_2568957291102208_result_requery_20260708.json`
- `round2_supplemental_task_2568957291102208_status_requery_20260708.json`

Downloaded/result dirs:
- `task_2567667771441152_results`
- `task_2568957291102208_results`

Analysis artifacts:
- `round2_combined_data_summary.csv`
- `round2_combined_data_summary.json`
- `round2_returned_data_montage.jpg`
- `round2_returned_document.md`

Key Round 2 findings:
- High-solids synthesis/purification fixed Round 1 under-loading.
- XRD/FTIR carriers and electrochemical carbon papers had visible deposits, though still ring-like/nonuniform.
- Heating log achieved intended 70 C.
- Electrochemistry returned valid mA-scale CA traces.

Round 2 sample map:
1. `R2-S1_Ni70Fe10Co20_lowC_80-20`
2. `R2-S2_Ni70Fe10Co20_highC_80-20`
3. `R2-S3_Ni70Fe10Co20_highC_60-40`
4. `R2-S4_Ni70Fe10Co20_highC_95-5`
5. `R2-S5_Ni80Fe10Co10_highC_80-20`
6. `R2-S6_Ni60Fe20Co20_highC_80-20`

Round 2 electrochemical summary:
- S1: initial 2.123 mA, final 2.316 mA, last-300 s mean 2.303 mA, final/initial 1.091
- S2: initial 4.141 mA, final 1.402 mA, last-300 s mean 1.348 mA, final/initial 0.339
- S3: initial 1.313 mA, final 0.988 mA, last-300 s mean 1.441 mA, final/initial 0.752
- S4: initial 9.401 mA, final 1.357 mA, last-300 s mean 1.442 mA, final/initial 0.144
- S5: initial 2.326 mA, final 1.786 mA, last-300 s mean 1.692 mA, final/initial 0.768
- S6: initial 3.676 mA, final 1.686 mA, last-300 s mean 1.684 mA, final/initial 0.459

Best Round 2 condition:
`Ni70Fe10Co20` with 10 mg carbon and 80:20 water:IPA redispersion. Highest sustained/final current and only retention >1.

---

## Round 3 Design And Dispatch History

Scientific Round 3 design:
Refine Round 2 lead by screening carbon loading and nearby compositions.

Round 3 sample set:
1. `R3-S1_Ni70Fe10Co20_C5_80-20`: Ni:Fe:Co 70:10:20, carbon 5 mg, carbon dispersion 0.250 mL, synthesis water 2.850 mL
2. `R3-S2_Ni70Fe10Co20_C10_80-20_anchor`: 70:10:20, carbon 10 mg, carbon 0.500 mL, water 2.600 mL
3. `R3-S3_Ni70Fe10Co20_C15_80-20`: 70:10:20, carbon 15 mg, carbon 0.750 mL, water 2.350 mL
4. `R3-S4_Ni80Fe10Co10_C10_80-20`: 80:10:10, carbon 10 mg, carbon 0.500 mL, water 2.600 mL
5. `R3-S5_Ni70Fe20Co10_C10_80-20`: 70:20:10, carbon 10 mg, carbon 0.500 mL, water 2.600 mL
6. `R3-S6_Ni60Fe20Co20_C10_80-20`: 60:20:20, carbon 10 mg, carbon 0.500 mL, water 2.600 mL

Fixed conditions:
- Metal stock total per vial: 0.900 mL of 0.25 M equivalents.
- KOH feed: 1.000 mL of 2.0 M KOH.
- Total synthesis liquid volume: 5.000 mL.
- Aging: 70 C, 1000 rpm, 120 min.
- Purification: solid retention, 12000 rpm, 20 min, one 5 mL water wash.
- Drying: 60 C, 180 min.
- Redispersion: 4.800 mL 80:20 v/v DI water:isopropanol.
- XRD/FTIR drop-cast: 150 uL.
- Electrochemical deposition: 60 uL/drop x 5 drops, 60 s interval.
- Electrolyte: 1.0 M KOH + 0.33 M urea in DI water, 20 mL per electrochemical task.
- Retained-product diluent stock bottle 1 in electrochemical flow step: DI water, 10000 uL per retained sample.

Original Round 3 dispatch:
- Template ID: `2583605299937280`
- Task ID: `2583606426075137`
- User said not executable:
  1. 5 mL pipetting platform used for microliter-level liquid addition.
  2. Liquid addition at same workstation split into multiple separate steps.

Revised1 Round 3 correction:
- All sub-mL reagent additions (metal precursors, KOH, carbon dispersion) consolidated in one `移液平台_1ml_V2 - 加液_物料绑定` step.
- `移液平台_5ml_V1` used only for milliliter solvent additions:
  - DI water synthesis adjustment, 2.35-2.85 mL.
  - 80:20 water:IPA redispersion, 4.8 mL.
- No repeated split liquid-addition steps at same workstation.

Revised1 files:
- `round3_revised1_design.md`
- `round3_revised1_chemical_specifications.json`
- `round3_revised1_constraint_check.md`
- `round3_revised1_workflow_request.json`
- `round3_revised1_rejection_feedback.json`
- `round3_dispatch_log.json`

Revised1 dispatch:
- Template ID: `2583701699788800`
- Task ID: `2583702316023809`
- Task status at last check: `未开始`
- Goal set to `blocked`.

---

## Current User Request

User says Round 3 is physically completed:
- Main experimental task ID: `2583702316023809`
- UV-Vis characterization for five samples: task ID `2587709908779008`
- Supplemental electrochemical test and UV-Vis for sixth sample: task ID `2587842717319169`

User wants:
1. Use lab-operation to retrieve experimental data/results from all three tasks.
2. Combine data with process photos:
   - `/workspace/feedback/round3/Third_Round_Experiment_Feedback.png`
   - `/workspace/feedback/round3/XRD_drop_cast_substrate_photo.png`
   - `/workspace/feedback/round3/Infrared_spectroscopy_drop_casting_photo.png`
3. Evaluate characterization and electrochemical differences among Round 3 samples.
4. Design Round 4.
5. Prepare executable plan and dispatch new experiment to Lab 303.
6. Record new task ID and set goal blocked.

---

## Current State At Checkpoint

Work just started for this request. I read:
- `/root/.codex/skills/lab-operation/SKILL.md`
- `/root/.codex/skills/chemistry-experiment-workstation/SKILL.md`
- `/root/.codex/skills/workflow-generator/SKILL.md`

I also began reading workstation specs for 1 mL V2 and 5 mL V1 in response to prior revision:
- `Liquid_Handling_Station_1ml_V2/SKILL.md`
- `references_audit/Liquid_Handling_Station_1ml_V2_audit.md`
- `Liquid_Handling_Station_5ml_V1/SKILL.md`

- `references_audit/Liquid_Handling_Station_5ml_V1_audit.md`

No Round 3 result retrieval has yet been done in this current request.

---

## Next Steps To Continue

1. Use `lab-operation` scripts to query statuses and retrieve results for:
   - `2583702316023809`
   - `2587709908779008`
   - `2587842717319169`

   Suggested commands:
```
python /root/.codex/skills/lab-operation/scripts/select_task_list.py --app-label 303Lab --task-id 2583702316023809 --size 5 > /workspace/lab303_uor_nicofe_closed_loop/round3_main_task_2583702316023809_status.json
python /root/.codex/skills/lab-operation/scripts/get_task_result.py --task-id 2583702316023809 --app-label 303Lab > /workspace/lab303_uor_nicofe_closed_loop/round3_main_task_2583702316023809_result.json
```
   Repeat similarly for `2587709908779008` and `2587842717319169`.

2. Download returned files from result URLs under directories such as:
   - `task_2583702316023809_results`
   - `task_2587709908779008_results`
   - `task_2587842717319169_results`

   There may be existing helper logic in prior work, but no dedicated script is known. Use `requests`/`curl` as needed and save `manifest.json`.

3. Parse/analyze:
   - Heating/synthesis logs if present.
   - XRD JSON/JPG for samples.
   - FTIR JSON/JPG.
   - Electrochemical JSON: compute initial current, final current, last-300s mean, final/initial retention, maybe stability slope.
   - UV-Vis JSON/JPG for all six samples, noting five in task `2587709908779008`, sixth in task `2587842717319169`.

4. Inspect process photos. Use Python/PIL or image display tools if useful:
   - `Third_Round_Experiment_Feedback.png`
   - `XRD_drop_cast_substrate_photo.png`
   - `Infrared_spectroscopy_drop_casting_photo.png`

   Note likely focus:
   - Deposition uniformity versus Round 2.
   - XRD/FTIR substrate coverage and ring effects.
   - Whether sample films are dense/sparse.

5. Create Round 3 analysis artifacts:
   - `round3_combined_data_summary.csv`
   - `round3_combined_data_summary.json`
   - `round3_returned_document.md`
   - Optional montage image with returned result images and feedback photos.

6. Decide Round 4 based strictly on physical returned data. Do not fabricate. Likely decision possibilities:
   - If best is C5/C10/C15 at Ni70Fe10Co20, refine carbon loading around best and improve film uniformity.
   - If composition neighbor wins, refine composition around that ratio.
   - If all films/electrochem still dominated by deposition artifacts, focus Round 4 on film/deposition/post-treatment normalization rather than composition.
   - Use same synthesis/testing/characterization chain unless data show need to change.

7. Build Round 4 executable workflow carefully:
   - Avoid previous non-executable issues:
     - Do not use 5 mL station for microliter/sub-mL reagent additions.
     - Do not split same-workstation liquid addition into multiple repeated steps for the same stage.
   - Use one consolidated `移液平台_1ml_V2` addition step for all sub-mL feeds if source totals ≤3 mL each; split source bottles only if needed but within the same step.
   - Use `移液平台_5ml_V1` only for true mL solvent additions.
   - Keep product retention settings explicit for electrochemistry.
   - Include synthesis, purification/post-processing, XRD, FTIR, electrochemistry, UV-Vis.

8. Read all relevant workstation SKILL and audit files before final workflow generation. At minimum likely:
   - `General_Material_Station_V1`
   - `Liquid_Handling_Station_1ml_V2`
   - `Liquid_Handling_Station_5ml_V1`
   - `Heating_Magnetic_Stirring_Workstation_V1`
   - `Purification_Workstation_V1`
   - `Drying_Oven_V1`
   - `Ultrasonic_Disperser_V1`
   - `Spectroscopy_Container_Transfer_Station_V1`
   - `Spectroscopy_Magnetic_Stirrer_Workstation_V1`
   - `XRD_V1`
   - `Infrared_Spectrometer_V1`
   - `Dual_Station_Electrochemical_Workstation_V2`
   - `UV_Vis_Spectrometer_V1`
   - `Container_storaging_Station_V1`

- corresponding audit files.

9. Generate workflow:

```
python /root/.codex/skills/workflow-generator/scripts/generate.py "$(jq -c . /workspace/lab303_uor_nicofe_closed_loop/round4_workflow_request.json)"
> /workspace/lab303_uor_nicofe_closed_loop/round4_workflow_generator_output.txt
```

10. Dispatch if generator succeeds:

```
python /root/.codex/skills/lab-operation/scripts/generate_task.py --template-id <template_id> --template-source-label 303Lab --app-label 303Lab >
/workspace/lab303_uor_nicofe_closed_loop/round4_dispatch_output.json
```

Then verify with `select_task_list.py`, save dispatch log, and set goal `blocked`.

---

## Critical Constraints / Preferences

- Never invent physical data. Only use Lab 303 returned data and user-provided photos.
- Keep all files under `/workspace`.
- Use existing reagent inventory where possible.
- Preserve prior files; do not delete old rejected/diagnostic files.
- Use `rg`/`jq`/parallel tool calls where efficient.
- Use `apply_patch` for manual file edits. Do not write files via `cat`; prior turn used Python for generating JSON/MD, but current developer instruction says use `apply_patch` for manual edits. For generated analysis artifacts, scripts are acceptable if needed, but manual edits should use `apply_patch`.
- Final answer should include:
  - Retrieved task IDs/status/result summary.
  - Round 3 key findings and best sample.
  - Round 4 design logic.
  - Round 4 template/task IDs after dispatch.
  - Any blockers if dispatch fails.

**[Step 1571 | 2026-07-10T10:34:25.661Z | REASONING SUMMARY]**

***Planning for tool usage***

*I'm considering whether I actually need to access the web or use local lab data and scripts instead. It seems like I can update my commentary first and should fully read the skill files. I think I'll run those tasks in parallel with things like getting my goal and listing files. Announcing skills seems necessary too. I'm also thinking of running the "pwd" command, but it looks like I don't need sandbox permission for this.*

**[Step 1572 | 2026-07-10T10:34:26.801Z | MODEL]**

I'll use `lab-operation` to pull the three Lab 303 result packages, `chemistry-experiment-workstation` to audit executability constraints, and `workflow-generator` to build the next dispatchable workflow. I'm first checking the project state and reloading the relevant skill instructions so the new round uses the lab's current interface correctly.

**[Step 1573 | 2026-07-10T10:34:26.801Z | TOOL CALL: exec_command]**

Read selected file content for inspection. Target(s): `SKILL.md`. Working directory: `/workspace`.

**[Step 1574 | 2026-07-10T10:34:26.803Z | TOOL CALL: exec_command]**

Read selected file content for inspection. Target(s): `SKILL.md`. Working directory: `/workspace`.

**[Step 1575 | 2026-07-10T10:34:26.805Z | TOOL CALL: exec_command]**

Read selected file content for inspection. Target(s): `SKILL.md`. Working directory: `/workspace`.

**[Step 1576 | 2026-07-10T10:34:26.823Z | TOOL CALL: get_goal]**

Queried the current autonomous-research goal state and remaining budget.

**[Step 1577 | 2026-07-10T10:34:26.859Z | TOOL RESULT: exec_command]**

Shell tool completed successfully; it returned 116 line(s) (4054 bytes) to the model.

**[Step 1578 | 2026-07-10T10:34:26.859Z | TOOL RESULT: exec_command]**

Shell tool completed successfully; it returned 170 line(s) (8332 bytes) to the model. Reported long numeric identifier(s): `1427568512205824`, `1834679312417792`, `1710453843264512`.

**[Step 1579 | 2026-07-10T10:34:26.860Z | TOOL RESULT: exec_command]**

Shell tool completed successfully; it returned 95 line(s) (2714 bytes) to the model. Reported long numeric identifier(s): `1427568512205824`.

**[Step 1580 | 2026-07-10T10:34:26.860Z | TOOL RESULT: get_goal]**

Goal service returned a structured response. Status: `blocked`.

**[Step 1581 | 2026-07-10T10:34:35.462Z | MODEL]**

The active goal is still blocked from the previous dispatch, and the user's completion notice gives us the condition needed to continue. I'm checking the saved project artifacts and then I'll pull the three returned result sets into the same project directory.

**[Step 1582 | 2026-07-10T10:34:35.462Z | TOOL CALL: exec_command]**

Read selected file content for inspection. Target(s): `lab303_uor_nicofe_closed_loop`. Working directory: `/workspace`.

**[Step 1583 | 2026-07-10T10:34:35.465Z | TOOL CALL: exec_command]**

Enumerated files, directories or available laboratory resources. Target(s): `round3`. Working directory: `/workspace`.

**[Step 1584 | 2026-07-10T10:34:35.467Z | TOOL CALL: exec_command]**

Enumerated files, directories or available laboratory resources. Target(s): `scripts`. Working directory: `/workspace`.

**[Step 1585 | 2026-07-10T10:34:35.467Z | TOOL CALL: exec_command]**

Read selected file content for inspection. Target(s): `.`. Working directory: `/workspace`.

**[Step 1586 | 2026-07-10T10:34:35.529Z | TOOL RESULT: exec_command]**

Shell tool completed successfully; it returned 110 line(s) (8953 bytes) to the model. Reported long numeric identifier(s): `2562915314040832`, `2567667771441152`, `2568957291102208`, `2566549749628928`, `2583606426075137`, `2583702316023809`.

**[Step 1587 | 2026-07-10T10:34:35.529Z | TOOL RESULT: exec_command]**

Shell tool completed successfully; it returned 4 line(s) (196 bytes) to the model.

**[Step 1588 | 2026-07-10T10:34:35.529Z | TOOL RESULT: exec_command]**

Shell tool completed successfully; it returned 25 line(s) (1203 bytes) to the model.

**[Step 1589 | 2026-07-10T10:34:35.529Z | TOOL RESULT: exec_command]**

Shell tool completed successfully; it returned 4 line(s) (69 bytes) to the model.

**[Step 1590 | 2026-07-10T10:34:45.742Z | MODEL]**

I found the prior Round 3 revised workflow and dispatch records, so I won't reconstruct that history. I'm now querying status and returned results for the main Round 3 task, the five-sample UV-Vis task, and the sixth-sample supplemental task.

**[Step 1591 | 2026-07-10T10:34:45.742Z | TOOL CALL: exec_command]**

Ran a shell command to retrieve returned physical-experiment results. Target(s): `select_task_list.py`, `round3_main_task_2583702316023809_status_requery_20260710.json`, `get_task_result.py`, `round3_main_task_2583702316023809_result_20260710.json`. Working directory: `/workspace`.

**[Step 1592 | 2026-07-10T10:34:45.745Z | TOOL CALL: exec_command]**

Ran a shell command to retrieve returned physical-experiment results. Target(s): `select_task_list.py`, `round3_uvvis5_task_2587709908779008_status_20260710.json`, `get_task_result.py`, `round3_uvvis5_task_2587709908779008_result_20260710.json`. Working directory: `/workspace`.

**[Step 1593 | 2026-07-10T10:34:45.749Z | TOOL CALL: exec_command]**

Ran a shell command to retrieve returned physical-experiment results. Target(s): `select_task_list.py`, `round3_supplement_task_2587842717319169_status_20260710.json`, `get_task_result.py`, `round3_supplement_task_2587842717319169_result_20260710.json`. Working directory: `/workspace`.

**[Step 1594 | 2026-07-10T10:34:46.534Z | TOOL RESULT: exec_command]**

Shell tool completed successfully; it returned 0 line(s) (0 bytes) to the model.

**[Step 1595 | 2026-07-10T10:34:46.534Z | TOOL RESULT: exec_command]**

Shell tool completed successfully; it returned 0 line(s) (0 bytes) to the model.

**[Step 1596 | 2026-07-10T10:34:46.534Z | TOOL RESULT: exec_command]**

Shell tool completed successfully; it returned 0 line(s) (0 bytes) to the model.

**[Step 1597 | 2026-07-10T10:34:50.675Z | REASONING SUMMARY]**

***Inspecting result structure***

*I need to inspect the JSON results. Maybe I should consider using jq for this, as it can help parse and analyze the data more easily. I should figure out the overall structure first, so using commands like head or jq to get a view of the keys will be essential. That way, I can understand the data better and see what I'm working with moving forward.*

**[Step 1598 | 2026-07-10T10:34:54.565Z | MODEL]**

The three result queries completed and saved fresh JSON files under the project directory. I'm inspecting their structure now so I can download any linked raw files and separate actual returned measurements from task metadata.

**[Step 1599 | 2026-07-10T10:34:54.566Z | TOOL CALL: exec_command]**

Ran a shell command to inspect, generate or validate experiment artifacts. Target(s): `round3_main_task_2583702316023809_result_20260710.json`. Working directory: `/workspace`.

**[Step 1600 | 2026-07-10T10:34:54.566Z | TOOL CALL: exec_command]**

Ran a shell command to inspect, generate or validate experiment artifacts. Target(s): `round3_uvvis5_task_2587709908779008_result_20260710.json`. Working directory: `/workspace`.

**[Step 1601 | 2026-07-10T10:34:54.566Z | TOOL CALL: exec_command]**

Ran a shell command to inspect, generate or validate experiment artifacts. Target(s): `round3_supplement_task_2587842717319169_result_20260710.json`. Working directory: `/workspace`.

**[Step 1602 | 2026-07-10T10:34:54.566Z | TOOL CALL: exec_command]**

Ran a shell command to inspect, generate or validate experiment artifacts. Target(s): `round3_main_task_2583702316023809_status_requery_20260710.json`. Working directory: `/workspace`.

**[Step 1603 | 2026-07-10T10:34:54.643Z | TOOL RESULT: exec_command]**

Shell tool completed successfully; it returned 10 line(s) (133 bytes) to the model.

**[Step 1604 | 2026-07-10T10:34:54.657Z | TOOL RESULT: exec_command]**

Shell tool completed successfully; it returned 10 line(s) (133 bytes) to the model.

**[Step 1605 | 2026-07-10T10:34:54.657Z | TOOL RESULT: exec_command]**

Shell tool completed successfully; it returned 10 line(s) (132 bytes) to the model.

**[Step 1606 | 2026-07-10T10:34:54.657Z | TOOL RESULT: exec_command]**

Shell tool completed successfully; it returned 2 line(s) (164 bytes) to the model. Reported long numeric identifier(s): `2583702316023809`.

**[Step 1607 | 2026-07-10T10:35:06.455Z | MODEL]**

The Lab 303 API reports successful result retrieval: 31 result files for the main task, 10 for the five-sample UV-Vis task, and 3 for the supplemental task. I'm mapping the result URLs and file types next, then I'll download them into per-task result folders.

**[Step 1608 | 2026-07-10T10:35:06.455Z | TOOL CALL: exec_command]**

Ran a shell command to inspect, generate or validate experiment artifacts. Target(s): `round3_main_task_2583702316023809_result_20260710.json`. Working directory: `/workspace`.

**[Step 1609 | 2026-07-10T10:35:06.455Z | TOOL CALL: exec_command]**

Ran a shell command to inspect, generate or validate experiment artifacts. Target(s): `round3_uvvis5_task_2587709908779008_result_20260710.json`. Working directory: `/workspace`.

**[Step 1610 | 2026-07-10T10:35:06.455Z | TOOL CALL: exec_command]**

Ran a shell command to inspect, generate or validate experiment artifacts. Target(s): `round3_supplement_task_2587842717319169_result_20260710.json`. Working directory: `/workspace`.

**[Step 1611 | 2026-07-10T10:35:06.456Z | TOOL CALL: exec_command]**

Read selected file content for inspection. Target(s): `round3_main_task_2583702316023809_status_requery_20260710.json`, `round3_uvvis5_task_2587709908779008_status_20260710.json`, `round3_supplement_task_2587842717319169_status_20260710.json`. Working directory: `/workspace`.

**[Step 1612 | 2026-07-10T10:35:06.514Z | TOOL RESULT: exec_command]**

Shell tool completed with exit code 5; it returned 2 line(s) (148 bytes) to the model. Reported long numeric identifier(s): `2583702316023809`.

**[Step 1613 | 2026-07-10T10:35:06.523Z | TOOL RESULT: exec_command]**

Shell tool completed with exit code 5; it returned 2 line(s) (149 bytes) to the model. Reported long numeric identifier(s): `2587709908779008`.

**[Step 1614 | 2026-07-10T10:35:06.527Z | TOOL RESULT: exec_command]**

Shell tool completed with exit code 5; it returned 2 line(s) (153 bytes) to the model. Reported long numeric identifier(s): `2587842717319169`.

**[Step 1615 | 2026-07-10T10:35:06.527Z | TOOL RESULT: exec_command]**

Shell tool completed successfully; it returned 44 line(s) (916 bytes) to the model. Reported long numeric identifier(s): `2583702316023809`, `2587709908779008`, `2587842717319169`.

**[Step 1616 | 2026-07-10T10:35:11.761Z | REASONING SUMMARY]**

***Inspecting raw results***

*I see there's an issue with the jq path being incorrect. My first step is to inspect the raw results to better understand what's happening. I think it might be helpful to focus only on the first 150 results for clarity. This way, I can make adjustments based on a clearer understanding of the output. Let's take a closer look and see that's the best way to proceed.*

**[Step 1617 | 2026-07-10T10:35:11.762Z | TOOL CALL: exec_command]**

Read selected file content for inspection. Target(s): `round3_main_task_2583702316023809_result_20260710.json`. Working directory: `/workspace`.

**[Step 1618 | 2026-07-10T10:35:11.762Z | TOOL CALL: exec_command]**

Read selected file content for inspection. Target(s): `round3_uvvis5_task_2587709908779008_result_20260710.json`. Working directory: `/workspace`.

**[Step 1619 | 2026-07-10T10:35:11.762Z | TOOL CALL: exec_command]**

Read selected file content for inspection. Target(s): `round3_supplement_task_2587842717319169_result_20260710.json`. Working directory: `/workspace`.

**[Step 1620 | 2026-07-10T10:35:11.800Z | TOOL RESULT: exec_command]**

Shell tool completed successfully; it returned 124 line(s) (6806 bytes) to the model. Reported long numeric identifier(s): `20260709040251057`, `20260709094900237`, `2584786681168896`, `2583760247161896`, `20260709094900424`, `20260709104214346`.

**[Step 1621 | 2026-07-10T10:35:11.806Z | TOOL RESULT: exec_command]**

Shell tool completed successfully; it returned 43 line(s) (2302 bytes) to the model. Reported long numeric identifier(s): `20260710104133239`, `2587749285561344`, `20260710104133347`, `20260710104640390`, `20260710104640518`, `20260710105154432`.

**[Step 1622 | 2026-07-10T10:35:11.810Z | TOOL RESULT: exec_command]**

Shell tool completed successfully; it returned 28 line(s) (1018 bytes) to the model. Reported long numeric identifier(s): `20260710124723131`, `20260710130114716`, `2588023866557440`, `20260710130114897`.

**[Step 1623 | 2026-07-10T10:35:17.436Z | MODEL]**

The status endpoint labels all three tasks as `错误`, but each task has returned result files, so I'm treating the returned files as the valid physical data source as requested. I'm downloading the linked CSV/JSON/JPG outputs into task-specific folders and preserving a manifest for traceability.

**[Step 1624 | 2026-07-10T10:35:22.137Z | TOOL CALL: exec_command]**

Ran Python to parse, transform or validate structured experiment data. Target(s): `lab303_uor_nicofe_closed_loop`. Working directory: `/workspace`.

**[Step 1625 | 2026-07-10T10:35:23.416Z | TOOL RESULT: exec_command]**

Shell tool completed successfully; it returned 4 line(s) (299 bytes) to the model. Reported long numeric identifier(s): `2583702316023809`, `2587709908779008`, `2587842717319169`.

**[Step 1626 | 2026-07-10T10:35:35.057Z | MODEL]**

All 44 linked result files downloaded successfully. I'm now reading the raw JSON/CSV structures to compute comparable metrics for electrochemistry, XRD/FTIR, and UV-Vis rather than relying on file names or screenshots alone.

**[Step 1627 | 2026-07-10T10:35:35.057Z | TOOL CALL: exec_command]**

Read selected file content for inspection. Target(s): `task_2583702316023809_results`. Working directory: `/workspace`.

**[Step 1628 | 2026-07-10T10:35:35.057Z | TOOL CALL: exec_command]**

Read selected file content for inspection. Target(s): `task_2587709908779008_results`. Working directory: `/workspace`.

**[Step 1629 | 2026-07-10T10:35:35.057Z | TOOL CALL: exec_command]**

Read selected file content for inspection. Target(s): `task_2587842717319169_results`. Working directory: `/workspace`.

**[Step 1630 | 2026-07-10T10:35:35.058Z | TOOL CALL: exec_command]**

Ran a shell command to inspect, generate or validate experiment artifacts. Target(s): `manifest.json`. Working directory: `/workspace`.

**[Step 1631 | 2026-07-10T10:35:35.106Z | TOOL RESULT: exec_command]**

Shell tool completed successfully; it returned 33 line(s) (4937 bytes) to the model. Reported long numeric identifier(s): `2583702316023809`, `20260709040251057`, `20260709094900237`, `2584786681168896`, `2583760247161896`, `20260709094900424`.

**[Step 1632 | 2026-07-10T10:35:35.106Z | TOOL RESULT: exec_command]**

Shell tool completed successfully; it returned 12 line(s) (1820 bytes) to the model. Reported long numeric identifier(s): `2587709908779008`, `20260710104133239`, `2587749285561344`, `20260710104133347`, `20260710104640390`, `20260710104640518`.

**[Step 1633 | 2026-07-10T10:35:35.122Z | TOOL RESULT: exec_command]**

Shell tool completed successfully; it returned 5 line(s) (586 bytes) to the model. Reported long numeric identifier(s): `2587842717319169`, `20260710124723131`, `20260710130114716`, `2588023866557440`, `20260710130114897`.

**[Step 1634 | 2026-07-10T10:35:35.125Z | TOOL RESULT: exec_command]**

Shell tool completed successfully; it returned 4 line(s) (9 bytes) to the model.

**[Step 1635 | 2026-07-10T10:35:42.253Z | TOOL CALL: exec_command]**

Read selected file content for inspection. Target(s): `26_`. Working directory: `/workspace`.

**[Step 1636 | 2026-07-10T10:35:42.350Z | TOOL CALL: exec_command]**

Read selected file content for inspection. Target(s): `02_XRD`. Working directory: `/workspace`.

**[Step 1637 | 2026-07-10T10:35:42.426Z | TOOL CALL: exec_command]**

Read selected file content for inspection. Target(s): `14_`. Working directory: `/workspace`.

**[Step 1638 | 2026-07-10T10:35:42.450Z | TOOL CALL: exec_command]**

Read selected file content for inspection. Target(s): `01_`. Working directory: `/workspace`.

**[Step 1639 | 2026-07-10T10:35:42.501Z | TOOL CALL: exec_command]**

Read selected file content for inspection. Target(s): `01_`. Working directory: `/workspace`.

**[Step 1640 | 2026-07-10T10:35:42.543Z | TOOL RESULT: exec_command]**

Shell tool completed successfully; it returned 4 line(s) (40109 bytes) to the model. Reported long numeric identifier(s): `2583759108113425`, `39720837487537386`.

**[Step 1641 | 2026-07-10T10:35:42.544Z | TOOL RESULT: exec_command]**

Shell tool completed successfully; it returned 81 line(s) (1952 bytes) to the model. Reported long numeric identifier(s): `2584786681168896`, `2583760247161896`.

**[Step 1642 | 2026-07-10T10:35:42.544Z | TOOL RESULT: exec_command]**

Shell tool completed successfully; it returned 81 line(s) (2062 bytes) to the model. Reported long numeric identifier(s): `2585308676490240`, `2583760200533004`.

**[Step 1643 | 2026-07-10T10:35:42.544Z | TOOL RESULT: exec_command]**

Shell tool completed successfully; it returned 81 line(s) (1941 bytes) to the model. Reported long numeric identifier(s): `2587749285561344`.

**[Step 1644 | 2026-07-10T10:35:42.544Z | TOOL RESULT: exec_command]**

Shell tool completed successfully; it returned 31 line(s) (423 bytes) to the model.

**[Step 1645 | 2026-07-10T10:35:54.051Z | REASONING SUMMARY]**

***Parsing JSON data***

*I need to handle a large JSON output, so it's important to parse it with scripts instead of printing it all out. I want to compute metrics and inspect the keys safely, possibly using jq to check lengths. Understanding the nested data will be crucial, especially if there are multiple charts. There's also the chance of repeated time data that seems odd. I'll use a custom Python script to read and summarize the entire JSON structure and check file sizes.*

**[Step 1646 | 2026-07-10T10:35:54.051Z | TOOL CALL: exec_command]**

Ran Python to parse, transform or validate structured experiment data. Target(s): `task_2583702316023809_results`, `task_2587842717319169_results`. Working directory: `/workspace`.

**[Step 1647 | 2026-07-10T10:35:54.053Z | TOOL CALL: exec_command]**

Ran Python to parse, transform or validate structured experiment data. Target(s): `task_2583702316023809_results`, `task_2587709908779008_results`, `task_2587842717319169_results`. Working directory: `/workspace`.

**[Step 1648 | 2026-07-10T10:35:54.054Z | TOOL CALL: exec_command]**

Read selected file content for inspection. Target(s): `01_`. Working directory: `/workspace`.

**[Step 1649 | 2026-07-10T10:35:54.056Z | TOOL CALL: exec_command]**

Searched files for relevant records, parameters or identifiers. Target(s): `Third_Round_Experiment_Feedback.png`, `XRD_drop_cast_substrate_photo.png`, `Infrared_spectroscopy_drop_casting_photo.png`. Working directory: `/workspace`.

**[Step 1650 | 2026-07-10T10:35:54.544Z | TOOL RESULT: exec_command]**

Shell tool completed successfully; it returned 8 line(s) (1360 bytes) to the model. Reported long numeric identifier(s): `20260709202654856`, `20260709202656291`, `20260709202657644`, `20260709202659348`, `20260709202700674`, `20260709202702337`.

**[Step 1651 | 2026-07-10T10:35:54.544Z | TOOL RESULT: exec_command]**

Shell tool completed successfully; it returned 19 line(s) (5404 bytes) to the model. Reported long numeric identifier(s): `20260709094900237`, `2584786681168896`, `2583760247161896`, `20260709104214346`, `2584828321891328`, `2583760247161897`.

**[Step 1652 | 2026-07-10T10:35:54.544Z | TOOL RESULT: exec_command]**

Shell tool completed successfully; it returned 21 line(s) (280 bytes) to the model.

**[Step 1653 | 2026-07-10T10:35:54.545Z | TOOL RESULT: exec_command]**

Shell tool completed successfully; it returned 40 line(s) (1260 bytes) to the model.

**[Step 1654 | 2026-07-10T10:35:59.513Z | REASONING SUMMARY]**

***Looking into image analysis***

*I need to analyze the images visually. I think I'll use a tool like view_image for that. Maybe I'll also need to employ OCR to parse any text from the image. Gathering visual feedback sounds essential, so I should proceed with viewing the image first to see what I find. It's all about getting the right data from those images! Let's get started!*

**[Step 1655 | 2026-07-10T10:35:59.856Z | MODEL]**

A useful detail is emerging: the electrochemical JSON contains full 1800 s chronoamperometry traces, and the heating log reaches and holds about 69.8 C. I'm inspecting the feedback/process images now because the film uniformity and visual sample state matter for interpreting the spectra and currents.

**[Step 1656 | 2026-07-10T10:36:00.173Z | TOOL CALL: view_image]**

Opened a local experimental image for visual inspection. Image: `Third_Round_Experiment_Feedback.png`.

**[Step 1657 | 2026-07-10T10:36:00.328Z | TOOL RESULT: view_image]**

Image service returned the requested local image to the model for visual inspection.

**[Step 1658 | 2026-07-10T10:36:10.260Z | MODEL]**

The feedback page shows all six XRD/FTIR drop-cast preparations visibly loaded, unlike the sparse Round 1 films. The FTIR photo still shows uneven dark deposits/ring-like aggregation, so spectral intensity differences need to be interpreted as semi-quantitative rather than purely compositional.

**[Step 1659 | 2026-07-10T10:36:10.260Z | TOOL CALL: view_image]**

Opened a local experimental image for visual inspection. Image: `XRD_drop_cast_substrate_photo.png`.

**[Step 1660 | 2026-07-10T10:36:10.260Z | TOOL CALL: view_image]**

Opened a local experimental image for visual inspection. Image: `Infrared_spectroscopy_drop_casting_photo.png`.

**[Step 1661 | 2026-07-10T10:36:10.291Z | TOOL RESULT: view_image]**

Image service returned the requested local image to the model for visual inspection.

**[Step 1662 | 2026-07-10T10:36:10.292Z | TOOL RESULT: view_image]**

Image service returned the requested local image to the model for visual inspection.

**[Step 1663 | 2026-07-10T10:36:31.118Z | REASONING SUMMARY]**

***Analyzing XRD metrics***

*I need to parse the XRD metrics, focusing on the peak intensities and identifying any broad peaks. I wonder if the XRD data might include a drop-cast substrate photo. Comparing "crystallinity" is important here, especially since NiFe oxyhydroxide/carbon could be broad. I'm thinking about computing peak positions after establishing a baseline. I'll look for the top peaks while excluding low angle artifacts. Should I use scipy? I'll check if it's installed!*

**[Step 1664 | 2026-07-10T10:36:31.533Z | MODEL]**

The six electrochemical result files from the main task are present, and there is also a supplemental sixth-sample electrochemical file. I'm going to use the supplemental result as the primary S6 value, because the user identified it as the completed sixth-sample supplement, while keeping the main-task S6 trace in the provenance notes.

**[Step 1665 | 2026-07-10T10:37:21.160Z | TOOL CALL: exec_command]**

Ran Python to parse, transform or validate structured experiment data. Target(s): `lab303_uor_nicofe_closed_loop`, `100`, `carbon`, `local`, `01_`. Working directory: `/workspace`.

**[Step 1666 | 2026-07-10T10:37:22.100Z | TOOL RESULT: exec_command]**

Shell tool completed successfully; it returned 93 line(s) (2879 bytes) to the model. Reported long numeric identifier(s): `8508495909376966`, `1987620178595222`, `13710560446131373`, `6897978091478978`, `09076539101497517`, `08659568039450832`.

**[Step 1667 | 2026-07-10T10:37:45.528Z | REASONING SUMMARY 1/3]**

***Analyzing electrochemical test results***

*I'm noticing some odd results with our metrics, especially when comparing Round 2 to Round 3 on the mA scale. The currents seem pretty low, under 0.14 mA, which is much lower than Round 2's range of 1–2 mA. Could it be that the sample lost some material or perhaps there was an issue with the concentration? Additionally, I'm thinking the UV results indicate no positive visible absorption, which might reflect a lack of chromophore. I'll need to inspect everything closely.*

**[Step 1668 | 2026-07-10T10:37:45.528Z | REASONING SUMMARY 2/3]**

***Evaluating electrolysis products***

*I need to consider whether urea electrolysis produces nitrate or if UV is essential for that. The user asked about UV characterization differences, showing that UV profiles are mainly overlapping, although S6 is slightly lower. It's possible that what's measured comes from the post-electrolysis liquid. I should look into XRD peaks, especially around 21-22 degrees and S1 at 29.06 degrees, to evaluate substrate or carbon effects. I also want to generate several plots for analysis and ensure everything is visually inspected, using tools like ImageMagick or Python.*

**[Step 1669 | 2026-07-10T10:37:45.528Z | REASONING SUMMARY 3/3]**

***Examining crystallinity in XRD peaks***

*I need to consider whether the XRD peaks are dominated by the substrate or possibly crystalline salts. It's essential to inspect the patterns carefully. I think I should create an image montage to visualize everything better. Let's make a contact sheet that includes the XRD, FTIR, and UV images, then process those images using PIL. This should give me a clearer view of the data and help with analysis.*

**[Step 1670 | 2026-07-10T10:37:45.728Z | MODEL]**

The computed Round 3 electrochemical currents are substantially below Round 2's mA-scale lead: S1 has the highest sustained current among the main five, and the supplemental S6 is close but still below Round 2. I'm creating overlay plots and a returned-data montage so the analysis file captures both quantitative traces and the visual evidence.

**[Step 1671 | 2026-07-10T10:38:04.087Z | TOOL CALL: exec_command]**

Ran Python analysis to generate or inspect a scientific plot. Target(s): `lab303_uor_nicofe_closed_loop`, `offset`, `XRD_drop_cast_substrate_photo.png`, `Infrared_spectroscopy_drop_casting_photo.png`, `Third_Round_Experiment_Feedback.png`. Working directory: `/workspace`.

**[Step 1672 | 2026-07-10T10:38:04.146Z | TOOL RESULT: exec_command]**

Shell tool completed with exit code 1; it returned 4 line(s) (123 bytes) to the model.

**[Step 1673 | 2026-07-10T10:38:16.751Z | REASONING SUMMARY]**

***Considering plot options***

*It seems that Matplotlib isn't available right now, which is a bummer because it would be helpful for creating plots. I've thought about using PIL instead, but I'm realizing that I might need to consider alternatives, like sticking with Python's standard options. I wonder if I can install something, but the network availability might not support that. Gnuplot could be worth checking too as a possible solution for plotting needs.*

**[Step 1674 | 2026-07-10T10:38:18.376Z | MODEL]**

Matplotlib is not installed in this environment, so I'm switching to lightweight artifact generation with CSV/JSON summaries plus image contact sheets using PIL/ImageMagick. The numeric analysis is already saved; the visualization will avoid adding dependencies.

**[Step 1675 | 2026-07-10T10:38:18.377Z | TOOL CALL: exec_command]**

Ran Python analysis to generate or inspect a scientific plot. Working directory: `/workspace`.

**[Step 1676 | 2026-07-10T10:38:18.379Z | TOOL CALL: exec_command]**

Ran Python to parse, transform or validate structured experiment data. Target(s): `lab303_uor_nicofe_closed_loop`. Working directory: `/workspace`.

**[Step 1677 | 2026-07-10T10:38:18.436Z | TOOL RESULT: exec_command]**

Shell tool completed successfully; it returned 3 line(s) (24 bytes) to the model.

**[Step 1678 | 2026-07-10T10:38:18.436Z | TOOL RESULT: exec_command]**

Shell tool completed successfully; it returned 8 line(s) (1265 bytes) to the model. Reported long numeric identifier(s): `8508495909376966`.

**[Step 1679 | 2026-07-10T10:38:31.593Z | REASONING SUMMARY]**

***Considering plotting options***

*I need to create plots, and I'm thinking about using ImageMagick or pure PIL for line plotting. I can create simple plots manually with PIL, so I'll write a script for that. I wonder if using scipy could help, but maybe it's unnecessary. I'm also considering checking the values from Round2's JSON summary to compare them, especially since the recent EC metrics show some low last 300 values but with peak spikes. I'll need to verify those values.*

**[Step 1680 | 2026-07-10T10:38:31.594Z | TOOL CALL: exec_command]**

Ran a shell command to inspect, generate or validate experiment artifacts. Target(s): `round2_combined_data_summary.json`. Working directory: `/workspace`.

**[Step 1681 | 2026-07-10T10:38:31.594Z | TOOL CALL: exec_command]**

Read selected file content for inspection. Target(s): `round3_revised1_chemical_specifications.json`. Working directory: `/workspace`.

**[Step 1682 | 2026-07-10T10:38:31.594Z | TOOL CALL: exec_command]**

Read selected file content for inspection. Target(s): `round3_revised1_design.md`. Working directory: `/workspace`.

**[Step 1683 | 2026-07-10T10:38:31.647Z | TOOL RESULT: exec_command]**

Shell tool completed successfully; it returned 121 line(s) (6162 bytes) to the model. Reported long numeric identifier(s): `2568957291102208`, `20260703205149234`, `2569108904182784`, `2568984151524392`, `20260704010807826`, `2569481426469888`.

**[Step 1684 | 2026-07-10T10:38:31.647Z | TOOL RESULT: exec_command]**

Shell tool completed successfully; it returned 70 line(s) (2145 bytes) to the model.

**[Step 1685 | 2026-07-10T10:38:31.647Z | TOOL RESULT: exec_command]**

Shell tool completed successfully; it returned 53 line(s) (2259 bytes) to the model. Reported task/template identifier(s): `2583701699788800`, `2583702316023809`.

**[Step 1686 | 2026-07-10T10:38:47.103Z | MODEL]**

The Round 3 design file confirms the sample map and stock identities, including the electrolyte and retained-product diluent. I'm preserving the S1-S6 mapping from that design and comparing Round 3 against the Round 2 lead rather than treating the lower-current result as an isolated screen.

**[Step 1687 | 2026-07-10T10:39:08.975Z | TOOL CALL: exec_command]**

Ran Python analysis to generate or inspect a scientific plot. Target(s): `lab303_uor_nicofe_closed_loop`, `XRD_drop_cast_substrate_photo.png`, `Infrared_spectroscopy_drop_casting_photo.png`, `Third_Round_Experiment_Feedback.png`. Working directory: `/workspace`.

**[Step 1688 | 2026-07-10T10:39:09.717Z | TOOL RESULT: exec_command]**

Shell tool completed successfully; it returned 4 line(s) (213 bytes) to the model.

**[Step 1689 | 2026-07-10T10:39:22.263Z | REASONING SUMMARY]**

***Evaluating the plot***

*I need to inspect the plot, which is an important step. I want to make sure I gather all the necessary details before forming an opinion. There's a lot to consider, and I should pay close attention to clarify any aspects that might seem unclear or confusing. I'll take a closer look and see if anything stands out that needs addressing. Let's ensure everything is thoroughly understood!*

**[Step 1690 | 2026-07-10T10:39:22.788Z | TOOL CALL: view_image]**

Opened a local experimental image for visual inspection. Image: `round3_quantitative_overlays.png`.

**[Step 1691 | 2026-07-10T10:39:22.852Z | TOOL RESULT: view_image]**

Image service returned the requested local image to the model for visual inspection.

**[Step 1692 | 2026-07-10T10:39:39.008Z | REASONING SUMMARY]**

***Reviewing plot details***

*I see that the plot is okay, but it seems to be missing axes labels and ticks. The quality is low resolution, but I think it's acceptable for now. I might need to view the FTIR data to get a clearer understanding of what's being presented. I wonder if those details are necessary before proceeding. It's always good to have everything labeled properly to avoid confusion later!*

**[Step 1693 | 2026-07-10T10:39:39.593Z | TOOL CALL: view_image]**

Opened a local experimental image for visual inspection. Image: `round3_ftir_overlay.png`.

**[Step 1694 | 2026-07-10T10:39:39.658Z | TOOL RESULT: view_image]**

Image service returned the requested local image to the model for visual inspection.

**[Step 1695 | 2026-07-10T10:39:58.086Z | REASONING SUMMARY]**

***Planning next steps***

*I see that the FTIR visual isn't great, so I need to decide on Round 4. I'll retrieve the data and analyze the results. Then, I should design the next round and get it dispatched. Analyzing my workstation skills and audits is essential. I need to use the workflow generator, likely creating a workflow JSON based on the previous round. I'll adjust the sample set and consider using the same workflow but with different parameters. I should check the schema of the prior workflow and inspect round3_revised1_workflow_request.json to adapt if necessary.*

**[Step 1696 | 2026-07-10T10:40:02.833Z | MODEL]**

The overlay confirms the main outcome: Round 3's lower-carbon/composition-refinement set did not improve the sustained UOR current; the low-carbon 70:10:20 sample is best within Round 3, but still far below the Round 2 lead. I'm reading the actual Round 3 workflow request next so Round 4 can reuse the executable structure while changing only the scientifically necessary variables.

**[Step 1697 | 2026-07-10T10:40:02.834Z | TOOL CALL: exec_command]**

Ran a shell command to inspect, generate or validate experiment artifacts. Target(s): `round3_revised1_workflow_request.json`. Working directory: `/workspace`.

**[Step 1698 | 2026-07-10T10:40:02.835Z | TOOL CALL: exec_command]**

Read selected file content for inspection. Target(s): `round3_revised1_workflow_request.json`. Working directory: `/workspace`.

**[Step 1699 | 2026-07-10T10:40:02.835Z | TOOL CALL: exec_command]**

Ran Python to parse, transform or validate structured experiment data. Target(s): `round2_combined_data_summary.json`, `initial`. Working directory: `/workspace`.

**[Step 1700 | 2026-07-10T10:40:02.894Z | TOOL RESULT: exec_command]**

Shell tool completed successfully; it returned 1034 line(s) (28876 bytes) to the model. Reported long numeric identifier(s): `1427568512205824`, `2002186824385539`, `1435269745574912`, `2270587971011585`, `1327807622448128`, `1213184425559040`.

**[Step 1701 | 2026-07-10T10:40:02.895Z | TOOL RESULT: exec_command]**

Shell tool completed successfully; it returned 261 line(s) (9711 bytes) to the model. Reported long numeric identifier(s): `1427568512205824`, `2002186824385539`.

**[Step 1702 | 2026-07-10T10:40:02.895Z | TOOL RESULT: exec_command]**

Shell tool completed successfully; it returned 7 line(s) (649 bytes) to the model.

**[Step 1703 | 2026-07-10T10:40:23.785Z | MODEL]**

Round 2's best sample had about 2.30 mA sustained current, while Round 3's best is about 0.14 mA. That says the next round should not chase the Round 3 low-current formulation space; it should diagnose why the Round 2 lead collapsed under the Round 3 process and recover the high-current condition with controlled repeats and process variables.

**[Step 1704 | 2026-07-10T10:40:23.786Z | TOOL CALL: exec_command]**

Read selected file content for inspection. Target(s): `round2_chemical_specifications.json`. Working directory: `/workspace`.

**[Step 1705 | 2026-07-10T10:40:23.787Z | TOOL CALL: exec_command]**

Read selected file content for inspection. Target(s): `round2_design.md`. Working directory: `/workspace`.

**[Step 1706 | 2026-07-10T10:40:23.798Z | TOOL CALL: exec_command]**

Ran a shell command to inspect, generate or validate experiment artifacts. Target(s): `round2_workflow_request.json`. Working directory: `/workspace`.

**[Step 1707 | 2026-07-10T10:40:23.849Z | TOOL RESULT: exec_command]**

Shell tool completed successfully; it returned 72 line(s) (2645 bytes) to the model.

**[Step 1708 | 2026-07-10T10:40:23.849Z | TOOL RESULT: exec_command]**

Shell tool completed successfully; it returned 78 line(s) (4650 bytes) to the model.

**[Step 1709 | 2026-07-10T10:40:23.849Z | TOOL RESULT: exec_command]**

Shell tool completed with exit code 3; it returned 6 line(s) (662 bytes) to the model.

**[Step 1710 | 2026-07-10T10:41:11.022Z | REASONING SUMMARY]**

***Clarifying workstation requirements***

*I see that the workstation UV requires a liquid sample volume of at least 15 mL. It looks like the last round used it, maybe retaining 5 mL from the electrochemical wastewater and adding 10 mL of DI water to reach that requirement. I should read the relevant workstation skills and audits to be sure. It seems I need to identify the audit file names, so I'll list those out for clarity.*

**[Step 1711 | 2026-07-10T10:41:14.393Z | MODEL]**

For Round 4 I'm selecting a recovery/confirmation design: duplicate the Round 2 lead, keep the Round 2 heavier film-loading protocol, and compare C5/C10/C20 plus the two high-carbon composition neighbors that were strong in Round 2. Before generating the workflow, I'm reloading the specific workstation rules and audit files used by this executable route.

**[Step 1712 | 2026-07-10T10:41:14.394Z | TOOL CALL: exec_command]**

Read selected file content for inspection. Target(s): `references_audit`. Working directory: `/workspace`.

**[Step 1713 | 2026-07-10T10:41:14.394Z | TOOL CALL: exec_command]**

Read selected file content for inspection. Target(s): `SKILL.md`. Working directory: `/workspace`.

**[Step 1714 | 2026-07-10T10:41:14.396Z | TOOL CALL: exec_command]**
Read selected file content for inspection. Target(s): `SKILL.md`. Working directory: `/workspace`.

**[Step 1715 | 2026-07-10T10:41:14.397Z | TOOL CALL: exec_command]**
Read selected file content for inspection. Target(s): `SKILL.md`. Working directory: `/workspace`.

**[Step 1716 | 2026-07-10T10:41:14.596Z | TOOL RESULT: exec_command]**
Shell tool completed successfully; it returned 47 line(s) (5185 bytes) to the model.

**[Step 1717 | 2026-07-10T10:41:14.597Z | TOOL RESULT: exec_command]**
Shell tool completed successfully; it returned 39 line(s) (1403 bytes) to the model. Reported long numeric identifier(s): `1427568512205824`.

**[Step 1718 | 2026-07-10T10:41:14.597Z | TOOL RESULT: exec_command]**
Shell tool completed successfully; it returned 95 line(s) (6547 bytes) to the model. Reported long numeric identifier(s): `2002186824385539`.

**[Step 1719 | 2026-07-10T10:41:14.597Z | TOOL RESULT: exec_command]**
Shell tool completed successfully; it returned 50 line(s) (4283 bytes) to the model. Reported long numeric identifier(s): `1435269745574912`.

**[Step 1720 | 2026-07-10T10:41:30.328Z | MODEL]**
The liquid-handling constraints confirm the main executability corrections: sub-mL precursor/KOH/carbon additions must stay in a single 1 mL V2 addition step, while the 5 mL V1 platform is only appropriate for the 2.1-4.8 mL solvent additions. I'm checking the rest of the route now: heating, purification, drying, spectroscopy, electrochemistry, UV-Vis, and storage.

**[Step 1721 | 2026-07-10T10:41:30.328Z | TOOL CALL: exec_command]**
Read selected file content for inspection. Target(s): `SKILL.md`. Working directory: `/workspace`.

**[Step 1722 | 2026-07-10T10:41:30.328Z | TOOL CALL: exec_command]**
Read selected file content for inspection. Target(s): `SKILL.md`. Working directory: `/workspace`.

**[Step 1723 | 2026-07-10T10:41:30.329Z | TOOL CALL: exec_command]**
Read selected file content for inspection. Target(s): `SKILL.md`. Working directory: `/workspace`.

**[Step 1724 | 2026-07-10T10:41:30.334Z | TOOL CALL: exec_command]**
Read selected file content for inspection. Target(s): `SKILL.md`. Working directory: `/workspace`.

**[Step 1725 | 2026-07-10T10:41:30.382Z | TOOL RESULT: exec_command]**
Shell tool completed successfully; it returned 39 line(s) (1812 bytes) to the model. Reported long numeric identifier(s): `2270587971011585`.

**[Step 1726 | 2026-07-10T10:41:30.382Z | TOOL RESULT: exec_command]**
Shell tool completed successfully; it returned 51 line(s) (2778 bytes) to the model. Reported long numeric identifier(s): `1327807622448128`.

**[Step 1727 | 2026-07-10T10:41:30.382Z | TOOL RESULT: exec_command]**
Shell tool completed successfully; it returned 38 line(s) (1388 bytes) to the model. Reported long numeric identifier(s): `1213184425559040`.

**[Step 1728 | 2026-07-10T10:41:30.382Z | TOOL RESULT: exec_command]**
Shell tool completed successfully; it returned 41 line(s) (1532 bytes) to the model. Reported long numeric identifier(s): `1526864804185088`.

**[Step 1729 | 2026-07-10T10:41:46.238Z | MODEL]**
The purification settings also address the previous washing-count problem: one wash requires exactly one wash-liquid entry, solid retention, 5 mL water, and no retained wash liquid. I'm continuing through the characterization and electrochemical specs so the Round 4 task includes product retention and UV-Vis correctly.

**[Step 1730 | 2026-07-10T10:41:46.239Z | TOOL CALL: exec_command]**
Read selected file content for inspection. Target(s): `SKILL.md`. Working directory: `/workspace`.

**[Step 1731 | 2026-07-10T10:41:46.239Z | TOOL CALL: exec_command]**
Read selected file content for inspection. Target(s): `SKILL.md`. Working directory: `/workspace`.

**[Step 1732 | 2026-07-10T10:41:46.239Z | TOOL CALL: exec_command]**
Read selected file content for inspection. Target(s): `SKILL.md`. Working directory: `/workspace`.

**[Step 1733 | 2026-07-10T10:41:46.239Z | TOOL CALL: exec_command]**
Read selected file content for inspection. Target(s): `SKILL.md`. Working directory: `/workspace`.

**[Step 1734 | 2026-07-10T10:41:46.278Z | TOOL RESULT: exec_command]**
Shell tool completed successfully; it returned 56 line(s) (3424 bytes) to the model. Reported long numeric identifier(s): `1664870018155520`.

**[Step 1735 | 2026-07-10T10:41:46.283Z | TOOL RESULT: exec_command]**
Shell tool completed successfully; it returned 44 line(s) (2187 bytes) to the model. Reported long numeric identifier(s): `1656282895188992`.

**[Step 1736 | 2026-07-10T10:41:46.288Z | TOOL RESULT: exec_command]**
Shell tool completed with exit code 2; it returned 2 line(s) (176 bytes) to the model.

[Step 1737 | 2026-07-10T10:41:46.299Z | TOOL RESULT: exec_command]

Shell tool completed successfully; it returned 99 line(s) (23587 bytes) to the model. Reported long numeric identifier(s): `1390252863751168`, `1509591942430720`, `1549499770798080`, `1571299393601536`, `1339275293851648`, `2164305671290880`.

[Step 1738 | 2026-07-10T10:42:02.397Z | REASONING SUMMARY]

***Clarifying spectroscopy skills***

*I'm considering the skill paths for spectroscopy transfer. It seems that one path is incorrectly categorized under Characterization, while references mention it should be under Spectroscopy Magnetic in Synthesis. I think it's important to read through the modules for clarification. Additionally, I need to look into UV aspects and how storage is organized. I'll make sure to dive deeper into these references for a better understanding!*

[Step 1739 | 2026-07-10T10:42:05.187Z | MODEL]

The electrochemical spec confirms product retention must include `留样信息列表` whenever waste-liquid transfer is enabled. I will keep the retained anolyte setting explicit: 5 mL anolyte plus 10 mL DI-water dilution from stock solution bottle No. 1, giving the minimum 15 mL needed for UV-Vis.

[Step 1740 | 2026-07-10T10:42:05.188Z | TOOL CALL: exec_command]

Read selected file content for inspection. Target(s): `SKILL.md`. Working directory: `/workspace`.

[Step 1741 | 2026-07-10T10:42:05.190Z | TOOL CALL: exec_command]

Read selected file content for inspection. Target(s): `SKILL.md`. Working directory: `/workspace`.

[Step 1742 | 2026-07-10T10:42:05.191Z | TOOL CALL: exec_command]

Read selected file content for inspection. Target(s): `SKILL.md`. Working directory: `/workspace`.

[Step 1743 | 2026-07-10T10:42:05.192Z | TOOL CALL: exec_command]

Read selected file content for inspection. Target(s): `SKILL.md`. Working directory: `/workspace`.

[Step 1744 | 2026-07-10T10:42:05.250Z | TOOL RESULT: exec_command]

Shell tool completed successfully; it returned 57 line(s) (3186 bytes) to the model. Reported long numeric identifier(s): `1654567694238720`.

[Step 1745 | 2026-07-10T10:42:05.250Z | TOOL RESULT: exec_command]

Shell tool completed successfully; it returned 35 line(s) (1292 bytes) to the model. Reported long numeric identifier(s): `1656241422599168`.

[Step 1746 | 2026-07-10T10:42:05.250Z | TOOL RESULT: exec_command]

Shell tool completed successfully; it returned 38 line(s) (1396 bytes) to the model. Reported long numeric identifier(s): `1664783667397632`.

[Step 1747 | 2026-07-10T10:42:05.250Z | TOOL RESULT: exec_command]

Shell tool completed successfully; it returned 51 line(s) (2225 bytes) to the model. Reported long numeric identifier(s): `1421325849887744`.

[Step 1748 | 2026-07-10T10:42:27.774Z | MODEL]

The UV-Vis requirement is satisfied by the retained and diluted post-electrochemical liquid, not the catalyst slurry. I’m now doing the audit-file pass and then I’ll write the Round 3 returned document plus Round 4 design/workflow files.

[Step 1749 | 2026-07-10T10:42:27.775Z | TOOL CALL: exec_command]

Read selected file content for inspection. Target(s): `references_audit`. Working directory: `/workspace`.

[Step 1750 | 2026-07-10T10:42:27.780Z | TOOL CALL: exec_command]

Ran Python to parse, transform or validate structured experiment data. Working directory: `/workspace`.

[Step 1751 | 2026-07-10T10:42:27.826Z | TOOL RESULT: exec_command]

Shell tool completed successfully; it returned 291 line(s) (8676 bytes) to the model. Reported long numeric identifier(s): `1427568512205824`, `2002186824385539`, `1435269745574912`, `2270587971011585`, `1327807622448128`, `1213184425559040`.

[Step 1752 | 2026-07-10T10:42:27.826Z | TOOL RESULT: exec_command]

Shell tool completed successfully; it returned 3 line(s) (92 bytes) to the model.

[Step 1753 | 2026-07-10T10:42:43.075Z | REASONING SUMMARY]

***Considering inventory options***

*I need to check the inventory, and I'm wondering if the XLS file isn't readable. Maybe I can rely on the copied CSV instead. It seems like I'll need to read the "reagent_inventory.csv" file. I feel like I need to be organized about this, so figuring out the best format for reading the inventory is essential. I'll just go ahead with the CSV to see if that works better!*

[Step 1754 | 2026-07-10T10:42:43.338Z | TOOL CALL: exec_command]

Read selected file content for inspection. Target(s): `reagent_inventory.csv`. Working directory: `/workspace`.

[Step 1755 | 2026-07-10T10:42:43.393Z | TOOL RESULT: exec_command]

Shell tool completed successfully; it returned 81 line(s) (9931 bytes) to the model.

[Step 1756 | 2026-07-10T10:43:00.925Z | REASONING SUMMARY]

***Checking inventory for chemicals***

*I need to check the inventory for chemicals, and I think I'll use the rg tool to help with this. It feels like a necessary step to ensure everything is accounted for and organized. I want to make sure I get this right, since accurate inventory is so important. Okay, let's go ahead and see what the current stock looks like! It should make my next steps a lot easier.*

**[Step 1757 | 2026-07-10T10:43:01.122Z | MODEL]**

The inventory copy confirms the core reagents already exist locally in the lab list. I'm checking the exact reagent names once, then I'll generate the Round 4 design and workflow request from the proven Round 3/Round 2 structure.

**[Step 1758 | 2026-07-10T10:43:02.029Z | TOOL CALL: exec_command]**

Searched files for relevant records, parameters or identifiers. Target(s): `reagent_inventory.csv`. Working directory: `/workspace`.

**[Step 1759 | 2026-07-10T10:43:02.072Z | TOOL RESULT: exec_command]**

Shell tool completed successfully; it returned 29 line(s) (3194 bytes) to the model.

**[Step 1760 | 2026-07-10T10:43:31.375Z | MODEL]**

Round 4 will revert the electrode loading to the Round 2 protocol: 100 uL per drop, 3 drops, 60 s spacing, heated carbon-paper drying. That keeps total slurry deposition at 300 uL but avoids the five-step small-drop deposition that coincided with the Round 3 collapse.

**[Step 1761 | 2026-07-10T10:43:31.376Z | TOOL CALL: exec_command]**

Ran a shell command to inspect, generate or validate experiment artifacts. Target(s): `round2_workflow_request.json`. Working directory: `/workspace`.

**[Step 1762 | 2026-07-10T10:43:31.380Z | TOOL CALL: exec_command]**

Ran a shell command to inspect, generate or validate experiment artifacts. Target(s): `round2_workflow_request.json`. Working directory: `/workspace`.

**[Step 1763 | 2026-07-10T10:43:31.424Z | TOOL RESULT: exec_command]**

Shell tool completed with exit code 3; it returned 16 line(s) (3898 bytes) to the model.

**[Step 1764 | 2026-07-10T10:43:31.424Z | TOOL RESULT: exec_command]**

Shell tool completed with exit code 3; it returned 10 line(s) (1921 bytes) to the model.

**[Step 1765 | 2026-07-10T10:43:44.510Z | MODEL]**

The jq field-name syntax tripped over Chinese keys, so I'm checking those values with Python instead. The workflow generator accepts the same JSON keys; this is just inspection, not a protocol change.

**[Step 1766 | 2026-07-10T10:43:46.690Z | TOOL CALL: exec_command]**

Ran Python to parse, transform or validate structured experiment data. Target(s): `round2_workflow_request.json`. Working directory: `/workspace`.

**[Step 1767 | 2026-07-10T10:43:46.737Z | TOOL RESULT: exec_command]**

Shell tool completed successfully; it returned 4 line(s) (90 bytes) to the model.

**[Step 1768 | 2026-07-10T10:44:09.615Z | REASONING SUMMARY 1/2]**

***Clarifying workflow details***

*In Round 2, I should use an interval of 30, even though the user mentioned 60 before. I need to stick with 30 for Round 2 and 4. For XRD/FTIR, I might need 1800 instead of 1500 and will use 200 and 1800. I should create a JSON and docs, but first, I need to check if the workflow generator will accept my plan. It looks like I should prepare and modify the Round 3 workflow request accordingly without changing the step count. There are some specific details to note about liquid addition.*

**[Step 1769 | 2026-07-10T10:44:09.615Z | REASONING SUMMARY 2/2]**

***Organizing workflow steps***

*I split the 5ml of water and redistributed it in separate steps since it was necessary. The user had issues with splitting the same workstation into multiple steps, but the revised plan used 5ml twice and was physically executable, so that works. I need to make a note of any unavailable chemicals and include inventory. I also want to update the existing project overview with data trajectory details. I'll create several documents including round3_returned_document.md and others for Round 4. Additionally, I need to compare the main and supplemental metrics for Round 3's S6.*

**[Step 1770 | 2026-07-10T10:44:10.210Z | TOOL CALL: exec_command]**

Ran Python to parse, transform or validate structured experiment data. Target(s): `round3_combined_data_summary.json`. Working directory: `/workspace`.

**[Step 1771 | 2026-07-10T10:44:10.274Z | TOOL RESULT: exec_command]**

Shell tool completed successfully; it returned 18 line(s) (723 bytes) to the model. Reported long numeric identifier(s): `6360000000000001`, `37310000000000004`, `5207380114620831`, `48979999999999996`, `28133913066756944`, `24230786863877402`.

**[Step 1772 | 2026-07-10T10:44:37.290Z | REASONING SUMMARY 1/3]**

***Evaluating test results***

*I find it interesting that the main S6 measurement was 0.472 mA, while the supplemental was much lower at 0.128 mA. The user mentioned a supplemental electrochemical test for the sixth sample, indicating it may not have been meant for the initial run. I should include both results but need clarity on which to use primarily. It seems the main task might already have tested main S6 electrochemistry but lacks UV data. Noteworthy differences exist, so I want to highlight that S6's reproducibility appears poor, as shown by the contrasting data points.*

**[Step 1773 | 2026-07-10T10:44:37.290Z | REASONING SUMMARY 2/3]**

***Clarifying electrochemical analysis***

*The user mentioned that the supplemental electrochemical test for the sixth sample should complement the main S6 results rather than replace them. I see that for ranking, S6 stands out as the best, with the supplemental retest making S1 and S6 fairly close. This affects the design for Round 4. I think S6 might show high potential if the first main trace is reliable. I need to update the analysis summary to include both versions of S6 and possibly compute a table reflecting this adjusted data.*

**[Step 1774 | 2026-07-10T10:44:37.290Z | REASONING SUMMARY 3/3]**

***Assessing task validity***

*I'm considering that there might be an error with the main task status regarding S6's electrochemical results; perhaps they aren't valid. The user mentioned a "supplemental electrochemical test" for the sixth sample under a specific task ID. This raises the possibility that the electrochemistry data for sample S6 in the main task may not hold up. They also noted that there are UV-Vis measurements for five samples, with supplemental tests for the sixth. I need to clarify this situation further.*

**[Step 1775 | 2026-07-10T10:44:37.628Z | MODEL]**

One important nuance: the main task also returned an S6 electrochemical trace, while the supplemental task returned an S6 retest plus UV-Vis. I'm retaining both S6 traces in the analysis; the large spread between them is itself evidence that electrode/slurry deposition reproducibility is now a limiting variable.

**[Step 1776 | 2026-07-10T10:45:38.062Z | TOOL CALL: exec_command]**

Ran Python to parse, transform or validate structured experiment data. Target(s): `lab303_uor_nicofe_closed_loop`, `v`, `Co-rich`, `retest`, `reproducibility.`. Working directory: `/workspace`.

**[Step 1777 | 2026-07-10T10:45:38.156Z | TOOL RESULT: exec_command]**

Shell tool completed successfully; it returned 20 line(s) (334 bytes) to the model.

**[Step 1778 | 2026-07-10T10:45:53.150Z | REASONING SUMMARY]**

***Evaluating file operations***

*I need to inspect a key file to validate its JSON format, and I'm considering using a workflow generator. It might be good to calculate some operations too. I think I could use jq or Python for this task. If I decide to go with jq, I could try the generator with the `-c` option to write the output. There's a lot to consider here, and I'm figuring out the best approach!*

**[Step 1779 | 2026-07-10T10:45:57.787Z | MODEL]**

The Round 3 returned document and Round 4 design/spec/check files are written under the project directory. I'm validating the generated workflow JSON now, then I'll send it through the workflow generator and dispatch the resulting template to Laboratory 303 if parsing succeeds.

**[Step 1780 | 2026-07-10T10:45:57.788Z | TOOL CALL: exec_command]**

Ran Python to parse, transform or validate structured experiment data. Target(s): `round4_workflow_request.json`, `null`, `round4_chemical_specifications.json`. Working directory: `/workspace`.

**[Step 1781 | 2026-07-10T10:45:57.788Z | TOOL CALL: exec_command]**

Ran Python to parse, transform or validate structured experiment data. Target(s): `round4_workflow_request.json`, `count`. Working directory: `/workspace`.

**[Step 1782 | 2026-07-10T10:45:57.793Z | TOOL CALL: exec_command]**

Read selected file content for inspection. Target(s): `round4_constraint_check.md`. Working directory: `/workspace`.

**[Step 1783 | 2026-07-10T10:45:57.858Z | TOOL RESULT: exec_command]**

Shell tool completed successfully; it returned 2 line(s) (3 bytes) to the model.

**[Step 1784 | 2026-07-10T10:45:57.858Z | TOOL RESULT: exec_command]**

Shell tool completed successfully; it returned 41 line(s) (4302 bytes) to the model.

**[Step 1785 | 2026-07-10T10:45:57.859Z | TOOL RESULT: exec_command]**

Shell tool completed successfully; it returned 26 line(s) (2321 bytes) to the model.

**[Step 1786 | 2026-07-10T10:46:12.800Z | REASONING SUMMARY]**

***Running workflow generator***

*I need to run the workflow generator with a command that uses `jq -c`. This could take some time, but that's alright. I should consider the possibility of errors cropping up, especially if there are any issues with the product label. So, I'll prepare to troubleshoot that if necessary. Alright, let's get started and see how it goes!*

**[Step 1787 | 2026-07-10T10:46:12.997Z | MODEL]**

The workflow request validates locally and the key parameters match the intended Round 4 design: 1 mL V2 for all sub-mL reagent additions, 5 mL V1 only for solvent additions, 200 uL XRD/FTIR drop-casting, and 100 uL x 3 electrochemical deposition. I'm sending it to the workflow generator now.

**[Step 1788 | 2026-07-10T10:46:13.823Z | TOOL CALL: exec_command]**

Ran Python to parse, transform or validate structured experiment data. Target(s): `generate.py`, `round4_workflow_request.json)`, `round4_workflow_generator_output.txt`. Working directory: `/workspace`.

**[Step 1789 | 2026-07-10T10:46:14.976Z | TOOL RESULT: exec_command]**
Shell tool completed successfully; it returned 0 line(s) (0 bytes) to the model.

**[Step 1790 | 2026-07-10T10:46:30.781Z | MODEL]**
The workflow generator completed without a shell error. I'm reading the generator response to confirm whether it produced a template ID or returned a parsing problem that needs correction.

**[Step 1791 | 2026-07-10T10:46:31.495Z | TOOL CALL: exec_command]**
Read selected file content for inspection. Target(s): `round4_workflow_generator_output.txt`. Working directory: `/workspace`.

**[Step 1792 | 2026-07-10T10:46:31.533Z | TOOL RESULT: exec_command]**
Shell tool completed successfully; it returned 13 line(s) (308 bytes) to the model. Reported task/template identifier(s): `2588710502072320`.

**[Step 1793 | 2026-07-10T10:46:47.556Z | REASONING SUMMARY]**
***Considering lab operations***

*I need to dispatch using the lab-operation generate_task. I'm wondering whether I should verify the current lab's consistency. Maybe querying the current lab isn't necessary right now? The standard says to check if the target is different from the current one. I might need to get the current lab before I proceed with the dispatch. It feels like a good check to have in place!*

**[Step 1794 | 2026-07-10T10:46:48.596Z | MODEL]**
The workflow generator produced template `2588710502072320`. I'm doing the Lab 303 permission/current-lab check and then dispatching that template as a new task.

**[Step 1795 | 2026-07-10T10:46:48.597Z | TOOL CALL: exec_command]**
Ran a shell command to identify the active laboratory. Target(s): `get_current_lab_label.py`, `current_lab_label_round4_20260710.json`. Working directory: `/workspace`.

**[Step 1796 | 2026-07-10T10:46:48.597Z | TOOL CALL: exec_command]**
Ran a shell command to validate laboratory access and routing. Target(s): `check_lab_consistency.py`, `lab_consistency_303_round4_20260710.json`. Working directory: `/workspace`.

**[Step 1797 | 2026-07-10T10:46:48.708Z | TOOL RESULT: exec_command]**
Shell tool completed successfully; it returned 0 line(s) (0 bytes) to the model.

**[Step 1798 | 2026-07-10T10:46:48.708Z | TOOL RESULT: exec_command]**
Shell tool completed successfully; it returned 0 line(s) (0 bytes) to the model.

**[Step 1799 | 2026-07-10T10:47:04.866Z | MODEL]**
The lab check scripts returned successfully and saved their outputs. I'm now dispatching the generated Round 4 template to Laboratory 303 and will verify the task record immediately after.

**[Step 1800 | 2026-07-10T10:47:05.772Z | TOOL CALL: exec_command]**
Ran a shell command to generate and dispatch a laboratory task. Target(s): `generate_task.py`, `round4_dispatch_output.json`. Working directory: `/workspace`.

**[Step 1801 | 2026-07-10T10:47:05.942Z | TOOL RESULT: exec_command]**
Shell tool completed successfully; it returned 0 line(s) (0 bytes) to the model.

**[Step 1802 | 2026-07-10T10:47:20.863Z | MODEL]**
The dispatch command completed; I'm parsing the returned task ID and checking the Lab 303 task list for that new task.

**[Step 1803 | 2026-07-10T10:47:20.863Z | TOOL CALL: exec_command]**
Read selected file content for inspection. Target(s): `round4_dispatch_output.json`. Working directory: `/workspace`.

**[Step 1804 | 2026-07-10T10:47:20.865Z | TOOL CALL: exec_command]**
Read selected file content for inspection. Target(s): `current_lab_label_round4_20260710.json`, `lab_consistency_303_round4_20260710.json`. Working directory: `/workspace`.

**[Step 1805 | 2026-07-10T10:47:20.906Z | TOOL RESULT: exec_command]**
Shell tool completed successfully; it returned 8 line(s) (375 bytes) to the model. Reported task/template identifier(s): `2588712171995137`.

**[Step 1806 | 2026-07-10T10:47:20.906Z | TOOL RESULT: exec_command]**
Shell tool completed successfully; it returned 12 line(s) (192 bytes) to the model.

**[Step 1807 | 2026-07-10T10:47:36.695Z | REASONING SUMMARY]**
***Verifying task status***

*I need to check the current status of the task. It's essential to ensure everything is on track. Additionally, I should create a dispatch log to document any relevant actions. Keeping a record is important, and I want to make sure I include the task status. This way, I'll have a clear overview of what's happening and can follow up as needed. Let's get started on that.*

**[Step 1808 | 2026-07-10T10:47:43.228Z | MODEL]**
Round 4 has been dispatched to Laboratory 303 as task `2588712171995137`. I'm saving a dispatch log with the template/task IDs and checking the initial task status before marking the goal blocked.

**[Step 1809 | 2026-07-10T10:47:43.229Z | TOOL CALL: exec_command]**

Ran a shell command to query laboratory task records and statuses. Target(s): `select_task_list.py`, `round4_task_2588712171995137_status.json`. Working directory: `/workspace`.

**[Step 1810 | 2026-07-10T10:47:43.251Z | TOOL CALL: exec_command]**

Ran Python to parse, transform or validate structured experiment data. Target(s): `lab303_uor_nicofe_closed_loop`, `round4_workflow_request.json`, `round4_workflow_generator_output.txt`, `round4_dispatch_output.json`, `round3_returned_document.md`. Working directory: `/workspace`.

**[Step 1811 | 2026-07-10T10:47:43.409Z | TOOL RESULT: exec_command]**

Shell tool completed successfully; it returned 0 line(s) (0 bytes) to the model.

**[Step 1812 | 2026-07-10T10:47:43.409Z | TOOL RESULT: exec_command]**

Shell tool completed successfully; it returned 2 line(s) (66 bytes) to the model.

**[Step 1813 | 2026-07-10T10:47:59.898Z | MODEL]**

The dispatch log is saved. I'm reading the task status response now and then I'll set the goal back to blocked, matching the established loop.

**[Step 1814 | 2026-07-10T10:48:00.666Z | TOOL CALL: exec_command]**

Read selected file content for inspection. Target(s): `round4_task_2588712171995137_status.json`. Working directory: `/workspace`.

**[Step 1815 | 2026-07-10T10:48:00.724Z | TOOL RESULT: exec_command]**

Shell tool completed successfully; it returned 14 line(s) (301 bytes) to the model. Reported long numeric identifier(s): `2588712171995137`.

**[Step 1816 | 2026-07-10T10:48:16.869Z | MODEL]**

The new Round 4 task is present in Laboratory 303 with status `未开始`. I'm marking the goal blocked now so the next step is user feedback on executability or physical completion.

**[Step 1817 | 2026-07-10T10:48:19.024Z | TOOL CALL: update_goal]**

Updated the autonomous-research goal status to `blocked`.

**[Step 1818 | 2026-07-10T10:48:19.083Z | TOOL RESULT: update_goal]**

Goal service returned a structured response. Status: `blocked`.

**[Step 1819 | 2026-07-10T10:48:54.400Z | MODEL]**

Retrieved the three Round 3 Lab 303 result sets, analyzed them with the supplied process photos, prepared the Round 3 returned document, designed Round 4, generated a Lab 303 workflow, and dispatched the next experiment.

**Retrieved Data**
- Main Round 3 task `2583702316023809`: 31 returned files from heating, XRD, FTIR, and electrochemistry.
- UV-Vis five-sample task `2587709908779008`: 10 returned files, mapped to S1-S5 by acquisition order.
- Supplemental S6 electrochemistry/UV-Vis task `2587842717319169`: 3 returned files.
- All three task-list statuses were `错误`, but each returned physical result files; I used only those returned files plus the user-provided photos as valid data.

Key saved outputs:
- [round3_returned_document.md](/workspace/lab303_uor_nicofe_closed_loop/round3_returned_document.md)
- [round3_combined_data_summary.csv](/workspace/lab303_uor_nicofe_closed_loop/round3_combined_data_summary.csv)
- [round3_combined_data_summary.json](/workspace/lab303_uor_nicofe_closed_loop/round3_combined_data_summary.json)
- [round3_quantitative_overlays.png](/workspace/lab303_uor_nicofe_closed_loop/round3_quantitative_overlays.png)
- [round3_ftir_overlay.png](/workspace/lab303_uor_nicofe_closed_loop/round3_ftir_overlay.png)
- [round3_returned_data_montage.jpg](/workspace/lab303_uor_nicofe_closed_loop/round3_returned_data_montage.jpg)

**Round 3 Findings**
The supplied photos show that XRD/FTIR carriers were visibly loaded for all six samples, a clear improvement over Round 1. The FTIR deposits were still ring-like and nonuniform, so spectral intensity differences are not cleanly quantitative.

Heating reached and held near the intended synthesis condition: maximum `69.8 C`, final `69.8 C`, with `85.1%` of logged points at or above `68 C`.

Electrochemical last-300 s mean currents were low relative to Round 2:
- S1 `Ni70Fe10Co20_C5`: `0.137 mA`
- S2 `Ni70Fe10Co20_C10`: `0.038 mA`
- S3 `Ni70Fe10Co20_C15`: `0.059 mA`
- S4 `Ni80Fe10Co10_C10`: `0.025 mA`
- S5 `Ni70Fe20Co10_C10`: `0.039 mA`
- S6 supplemental `Ni60Fe20Co20_C10`: `0.128 mA`

The main task also returned an S6 electrochemical trace with `0.472 mA` last-300 s mean, while the supplemental S6 retest was `0.128 mA`. I retained both in the analysis; the spread is evidence that slurry/electrode deposition reproducibility is now a limiting variable. UV-Vis spectra were nearly overlapping across the six retained-liquid samples, so the low-current Round 3 conditions did not produce a strong product-analysis distinction.

The best prior material/condition remains the Round 2 lead: `Ni70Fe10Co20`, `10 mg carbon`, `80:20 water:isopropanol`, with `2.303 mA` last-300 s mean current. Round 3 did not displace it.

**Round 4 Design**
Round 4 is a recovery and reproducibility experiment, not another narrow Round 3 optimization. It restores the Round 2 film-loading protocol and tests whether the mA-scale lead can be recovered.

Round 4 samples:
- R4-S1: `Ni70Fe10Co20`, `10 mg C`, direct Round 2 lead repeat A.
- R4-S2: `Ni70Fe10Co20`, `10 mg C`, direct Round 2 lead repeat B.
- R4-S3: `Ni70Fe10Co20`, `5 mg C`, checks the Round 3 within-round low-carbon signal under restored film loading.
- R4-S4: `Ni70Fe10Co20`, `20 mg C`, high-carbon comparison.
- R4-S5: `Ni80Fe10Co10`, `20 mg C`, Round 2 Ni-rich neighbor recheck.
- R4-S6: `Ni60Fe20Co20`, `20 mg C`, Round 2 Fe/Co-rich neighbor and Round 3 S6-discrepancy follow-up.

Fixed conditions:
- Synthesis: `0.900 mL` total 0.25 M metal stock, `1.000 mL` 2.0 M KOH, total `5.000 mL`, aged at `70 C`, `1000 rpm`, `120 min`.
- Purification/post-treatment: solid retention, `12000 rpm`, `20 min`, one `5 mL` DI-water wash, dry at `60 C` for `180 min`.
- Redispersion: `4.800 mL` `80:20 v/v DI water:isopropanol`, then `600 s` ultrasonication.
- XRD/FTIR: restored Round 2 loading, `200 uL` drop-cast and `1800 s` drying.
- Electrochemistry: `1.0 M KOH + 0.33 M urea`, `20 mL`; `100 uL/drop x 3 drops`, `30 s` interval; UOR template `2468876630032384`.
- Product retention: `5 mL` anolyte-side liquid diluted with `10 mL` DI water from electrochemical stock solution bottle No. 1, then UV-Vis.

Design/executability files:
- [round4_design.md](/workspace/lab303_uor_nicofe_closed_loop/round4_design.md)
- [round4_chemical_specifications.json](/workspace/lab303_uor_nicofe_closed_loop/round4_chemical_specifications.json)
- [round4_constraint_check.md](/workspace/lab303_uor_nicofe_closed_loop/round4_constraint_check.md)
- [round4_workflow_request.json](/workspace/lab303_uor_nicofe_closed_loop/round4_workflow_request.json)

I explicitly avoided the previous executability issues:
- all sub-mL precursor/KOH/carbon additions are consolidated into one `移液平台_1ml_V2` addition step;
- `移液平台_5ml_V1` is used only for milliliter-scale solvent additions;
- one wash count is paired with one wash-liquid entry;
- electrochemical product-retention liquid bottle No. 1 is specified as DI water, `10000 uL` per retained sample.

**Dispatch**
Workflow generation succeeded.

- Round 4 template ID: `2588710502072320`
- Round 4 task ID: `2588712171995137`
- Initial task status: `未开始`
- Dispatch log: [round4_dispatch_log.json](/workspace/lab303_uor_nicofe_closed_loop/round4_dispatch_log.json)
- Generator output: [round4_workflow_generator_output.txt](/workspace/lab303_uor_nicofe_closed_loop/round4_workflow_generator_output.txt)
- Task status file:
[round4_task_2588712171995137_status.json](/workspace/lab303_uor_nicofe_closed_loop/round4_task_2588712171995137_status.json)

The goal has been set back to `blocked` per the established loop, pending your feedback on whether task `2588712171995137` is executable or physically completed.

## Conversation round 14

**[Step 1820 | 2026-07-11T17:56:49.488Z | USER]**

Fourth-Round Experiment Feedback

The fourth-round experiment has been completed. The main experimental task ID is 2588712171995137. Supplemental electrochemical testing and UV–Vis measurements for samples 1, 4, and 5 are provided under task ID 2591597001932800. Please use the relevant skill to retrieve the experimental data and results from both tasks, combine them with the process photos below, compare the structural characterization, optical response, and electrochemical differences among the samples across the main and supplemental experiments, and design the next round of experiments accordingly. Once an executable plan is prepared, dispatch the new experiment to Laboratory 303 according to the established workflow.

/workspace/feedback/round4/Fourth_Round_Experiment_Feedback.png
/workspace/feedback/round4/XRD_drop_casting_photo.png
/workspace/feedback/round4/Infrared_spectroscopy_drop_casting_photo.png

**[Step 1821 | 2026-07-11T17:57:09.214Z | MODEL]**

I'll use `lab-operation` to retrieve the two Lab 303 result sets, `chemistry-experiment-workstation` to audit the next workflow, and `workflow-generator` to create the next dispatchable template. I'm first reloading the relevant skill instructions and checking the current project state so the Round 5 design builds on the real Round 4 data only.

**[Step 1822 | 2026-07-11T17:57:09.217Z | TOOL CALL: exec_command]**

Read selected file content for inspection. Target(s): `SKILL.md`. Working directory: `/workspace`.

**[Step 1823 | 2026-07-11T17:57:09.219Z | TOOL CALL: exec_command]**

Read selected file content for inspection. Target(s): `SKILL.md`. Working directory: `/workspace`.

**[Step 1824 | 2026-07-11T17:57:09.219Z | TOOL CALL: exec_command]**

Read selected file content for inspection. Target(s): `SKILL.md`. Working directory: `/workspace`.

**[Step 1825 | 2026-07-11T17:57:09.220Z | TOOL CALL: get_goal]**

Queried the current autonomous-research goal state and remaining budget.

**[Step 1826 | 2026-07-11T17:57:09.275Z | TOOL RESULT: exec_command]**

Shell tool completed successfully; it returned 116 line(s) (4054 bytes) to the model.

**[Step 1827 | 2026-07-11T17:57:09.285Z | TOOL RESULT: exec_command]**

Shell tool completed successfully; it returned 170 line(s) (8332 bytes) to the model. Reported long numeric identifier(s): `1427568512205824`, `1834679312417792`, `1710453843264512`.

**[Step 1828 | 2026-07-11T17:57:09.289Z | TOOL RESULT: exec_command]**

Shell tool completed successfully; it returned 95 line(s) (2714 bytes) to the model. Reported long numeric identifier(s): `1427568512205824`.

**[Step 1829 | 2026-07-11T17:57:09.298Z | TOOL RESULT: get_goal]**

Goal service returned a structured response. Status: `blocked`.

**[Step 1830 | 2026-07-11T17:57:25.301Z | MODEL]**

The active goal is blocked from the Round 4 dispatch, and the user's completion notice gives the condition needed to continue. I'm querying both task statuses/results now and saving fresh JSON under the existing project directory.

**[Step 1831 | 2026-07-11T17:57:25.310Z | TOOL CALL: exec_command]**

Ran a shell command to retrieve returned physical-experiment results. Target(s): `select_task_list.py`, `round4_main_task_2588712171995137_status_20260711.json`, `get_task_result.py`, `round4_main_task_2588712171995137_result_20260711.json`. Working directory: `/workspace`.

**[Step 1832 | 2026-07-11T17:57:25.312Z | TOOL CALL: exec_command]**

Ran a shell command to retrieve returned physical-experiment results. Target(s): `select_task_list.py`, `round4_supplement_task_2591597001932800_status_20260711.json`, `get_task_result.py`, `round4_supplement_task_2591597001932800_result_20260711.json`. Working directory: `/workspace`.

**[Step 1833 | 2026-07-11T17:57:25.312Z | TOOL CALL: exec_command]**

Enumerated files, directories or available laboratory resources. Target(s): `round4`. Working directory: `/workspace`.

**[Step 1834 | 2026-07-11T17:57:25.314Z | TOOL CALL: exec_command]**

Enumerated files, directories or available laboratory resources. Target(s): `lab303_uor_nicofe_closed_loop`. Working directory: `/workspace`.

**[Step 1835 | 2026-07-11T17:57:26.335Z | TOOL RESULT: exec_command]**

Shell tool completed successfully; it returned 0 line(s) (0 bytes) to the model.

**[Step 1836 | 2026-07-11T17:57:26.335Z | TOOL RESULT: exec_command]**

Shell tool completed successfully; it returned 0 line(s) (0 bytes) to the model.

**[Step 1837 | 2026-07-11T17:57:26.344Z | TOOL RESULT: exec_command]**

Shell tool completed successfully; it returned 4 line(s) (190 bytes) to the model.

**[Step 1838 | 2026-07-11T17:57:26.345Z | TOOL RESULT: exec_command]**

Shell tool completed successfully; it returned 81 line(s) (6711 bytes) to the model. Reported long numeric identifier(s): `2567667771441152`, `2568957291102208`, `2566549749628928`, `2583702316023809`, `2583606426075137`, `2583701699788800`.

**[Step 1839 | 2026-07-11T17:57:40.214Z | REASONING SUMMARY]**

***Inspecting JSON structure***

*I need to take a closer look at the JSON structure. I want to ensure that everything is formatted correctly and that the data is organized in a way that makes sense. I'll check for any missing brackets or commas, and confirm that the key-value pairs are set up as they should be. It's important to make sure I get this right, as any errors could cause issues down the line. Let's dive in!*

**[Step 1840 | 2026-07-11T17:57:41.462Z | MODEL]**

The fresh result queries completed. I'm inspecting the returned file manifests now so I can download the raw spectra/current traces and keep the main versus supplemental provenance separate.

**[Step 1841 | 2026-07-11T17:57:41.463Z | TOOL CALL: exec_command]**

Read selected file content for inspection. Target(s): `round4_main_task_2588712171995137_status_20260711.json`, `round4_supplement_task_2591597001932800_status_20260711.json`. Working directory: `/workspace`.

**[Step 1842 | 2026-07-11T17:57:41.464Z | TOOL CALL: exec_command]**

Read selected file content for inspection. Target(s): `round4_main_task_2588712171995137_result_20260711.json`. Working directory: `/workspace`.

**[Step 1843 | 2026-07-11T17:57:41.464Z | TOOL CALL: exec_command]**

Read selected file content for inspection. Target(s): `round4_supplement_task_2591597001932800_result_20260711.json`. Working directory: `/workspace`.

**[Step 1844 | 2026-07-11T17:57:41.464Z | TOOL CALL: exec_command]**

Searched files for relevant records, parameters or identifiers. Target(s): `Fourth_Round_Experiment_Feedback.png`, `XRD_drop_casting_photo.png`, `Infrared_spectroscopy_drop_casting_photo.png`. Working directory: `/workspace`.

**[Step 1845 | 2026-07-11T17:57:41.534Z | TOOL RESULT: exec_command]**

Shell tool completed successfully; it returned 11 line(s) (321 bytes) to the model.

**[Step 1846 | 2026-07-11T17:57:41.535Z | TOOL RESULT: exec_command]**

Shell tool completed successfully; it returned 5 line(s) (170 bytes) to the model.

**[Step 1847 | 2026-07-11T17:57:41.555Z | TOOL RESULT: exec_command]**

Shell tool completed successfully; it returned 5 line(s) (170 bytes) to the model.

**[Step 1848 | 2026-07-11T17:57:42.764Z | TOOL RESULT: exec_command]**

Shell tool completed successfully; it returned 40 line(s) (1236 bytes) to the model.

**[Step 1849 | 2026-07-11T17:57:54.860Z | REASONING SUMMARY]**

***Considering retry strategies***

*It seems there was a failure due to a remote disconnect. The user wants to use the skill to retrieve something. I think we should try again since it might be a temporary issue. I might need to retry multiple times, as it could be a transient problem. I shouldn't give up just yet; persistence could help in resolving this situation!*

**[Step 1850 | 2026-07-11T17:57:55.055Z | MODEL]**

The first API attempt hit a remote-disconnect error for both task status and result retrieval, so no physical data was returned in those files yet. I'm retrying the Lab 303 result queries with separate calls and short pauses before deciding whether the service is temporarily unavailable.

**[Step 1851 | 2026-07-11T17:57:58.322Z | TOOL CALL: exec_command]**

Ran a shell command to retrieve returned physical-experiment results. Target(s): `round4_task_`, `select_task_list.py`, `get_task_result.py`. Working directory: `/workspace`.

**[Step 1852 | 2026-07-11T17:58:03.664Z | TOOL RESULT: exec_command]**

Shell tool completed successfully; it returned 5 line(s) (383 bytes) to the model. Reported long numeric identifier(s): `2588712171995137`, `2591597001932800`.

**[Step 1853 | 2026-07-11T17:58:15.920Z | REASONING SUMMARY]**

***Considering file management***

*I need to copy canonical result filenames and use something like retry1. It's important to inspect statuses and results first before downloading. I'll parse the data, starting with the initial main setup having 5 workstations and 45 files, plus 9 supplemental ones. Maybe I should save the canonical names, possibly with a symlink? Using the copy command should be fine, and I need to get download URLs, maybe by using Python to handle this.*

**[Step 1854 | 2026-07-11T17:58:17.051Z | MODEL]**

The retry succeeded: the main Round 4 task returned 45 result files and the supplemental task returned 9 result files. I'm downloading all linked files into per-task result folders and writing manifests with checksums.

**[Step 1855 | 2026-07-11T17:58:17.052Z | TOOL CALL: exec_command]**

Read selected file content for inspection. Target(s): `round4_task_2588712171995137_status_retry1_20260711.json`, `round4_task_2591597001932800_status_retry1_20260711.json`. Working directory: `/workspace`.

**[Step 1856 | 2026-07-11T17:58:17.053Z | TOOL CALL: exec_command]**

Ran a shell command to inspect, generate or validate experiment artifacts. Target(s): `round4_task_2588712171995137_result_retry1_20260711.json`. Working directory: `/workspace`.

**[Step 1857 | 2026-07-11T17:58:17.055Z | TOOL CALL: exec_command]**

Ran a shell command to inspect, generate or validate experiment artifacts. Target(s): `round4_task_2591597001932800_result_retry1_20260711.json`. Working directory: `/workspace`.

**[Step 1858 | 2026-07-11T17:58:17.113Z | TOOL RESULT: exec_command]**

Shell tool completed successfully; it returned 29 line(s) (609 bytes) to the model. Reported long numeric identifier(s): `2588712171995137`, `2591597001932800`.

**[Step 1859 | 2026-07-11T17:58:17.129Z | TOOL RESULT: exec_command]**

Shell tool completed successfully; it returned 81 line(s) (7980 bytes) to the model. Reported long numeric identifier(s): `20260711031708159`, `20260711090239830`, `2590357882471424`, `2589342295098408`, `20260711090240005`, `20260711092250988`.

**[Step 1860 | 2026-07-11T17:58:17.129Z | TOOL RESULT: exec_command]**

Shell tool completed successfully; it returned 24 line(s) (1713 bytes) to the model. Reported long numeric identifier(s): `20260711230624323`, `20260711230625808`, `20260711230627155`, `20260711231726601`, `2592066782233600`, `20260711231726826`.

**[Step 1861 | 2026-07-11T17:58:33.313Z | MODEL]**

The main task status is `错误`, but it returned complete physical files from heating, XRD, FTIR, electrochemistry, and UV-Vis. The supplemental task is also `错误` but returned electrochemistry and UV-Vis for samples 1, 4, and 5; I'm downloading both result packages now.

**[Step 1862 | 2026-07-11T17:58:36.685Z | TOOL CALL: exec_command]**

Ran Python to parse, transform or validate structured experiment data. Target(s): `lab303_uor_nicofe_closed_loop`. Working directory: `/workspace`.

**[Step 1863 | 2026-07-11T17:58:38.158Z | TOOL RESULT: exec_command]**

Shell tool completed successfully; it returned 3 line(s) (199 bytes) to the model. Reported long numeric identifier(s): `2588712171995137`, `2591597001932800`.

**[Step 1864 | 2026-07-11T17:58:51.451Z | MODEL]**

All 54 linked files downloaded successfully. I'm checking the raw structures and the process photos next, then I'll compute comparable metrics for each sample and for the supplemental retests.

**[Step 1865 | 2026-07-11T17:58:51.452Z | TOOL CALL: exec_command]**

Read selected file content for inspection. Target(s): `task_2588712171995137_results`. Working directory: `/workspace`.

**[Step 1866 | 2026-07-11T17:58:51.457Z | TOOL CALL: exec_command]**

Enumerated files, directories or available laboratory resources. Target(s): `task_2591597001932800_results`. Working directory: `/workspace`.

**[Step 1867 | 2026-07-11T17:58:51.458Z | TOOL CALL: exec_command]**

Ran a shell command to inspect, generate or validate experiment artifacts. Target(s): `manifest.json`. Working directory: `/workspace`.

**[Step 1868 | 2026-07-11T17:58:51.458Z | TOOL CALL: exec_command]**

Ran Python to parse, transform or validate structured experiment data. Target(s): `task_2588712171995137_results`. Working directory: `/workspace`.

**[Step 1869 | 2026-07-11T17:58:51.519Z | TOOL RESULT: exec_command]**

Shell tool completed successfully; it returned 47 line(s) (7326 bytes) to the model. Reported long numeric identifier(s): `2588712171995137`, `20260711031708159`, `20260711090239830`, `2590357882471424`, `2589342295098408`, `20260711090240005`.

**[Step 1870 | 2026-07-11T17:58:51.519Z | TOOL RESULT: exec_command]**

Shell tool completed successfully; it returned 11 line(s) (1588 bytes) to the model. Reported long numeric identifier(s): `2591597001932800`, `20260711230624323`, `20260711230625808`, `20260711230627155`, `20260711231726601`, `2592066782233600`.

**[Step 1871 | 2026-07-11T17:58:51.519Z | TOOL RESULT: exec_command]**

Shell tool completed successfully; it returned 3 line(s) (6 bytes) to the model.

**[Step 1872 | 2026-07-11T17:58:51.519Z | TOOL RESULT: exec_command]**

Shell tool completed successfully; it returned 4 line(s) (462 bytes) to the model. Reported long numeric identifier(s): `2588712171995137`, `20260711192319699`, `20260711192322748`, `20260711192326974`.

**[Step 1873 | 2026-07-11T17:59:05.448Z | MODEL]**

The downloaded package includes duplicate XRD files for sample 4 and supplemental retests for samples 1, 4, and 5. I'll keep duplicates and retests in the provenance, but use clearly labeled primary and supplemental metrics in the summary.

**[Step 1874 | 2026-07-11T17:59:05.448Z | TOOL CALL: view_image]**

Opened a local experimental image for visual inspection. Image: `Fourth_Round_Experiment_Feedback.png`.

**[Step 1875 | 2026-07-11T17:59:05.449Z | TOOL CALL: view_image]**

Opened a local experimental image for visual inspection. Image: `XRD_drop_casting_photo.png`.

**[Step 1876 | 2026-07-11T17:59:05.455Z | TOOL CALL: view_image]**

Opened a local experimental image for visual inspection. Image: `Infrared_spectroscopy_drop_casting_photo.png`.

**[Step 1877 | 2026-07-11T17:59:05.868Z | TOOL RESULT: view_image]**

Image service returned the requested local image to the model for visual inspection.

**[Step 1878 | 2026-07-11T17:59:05.884Z | TOOL RESULT: view_image]**

Image service returned the requested local image to the model for visual inspection.

**[Step 1879 | 2026-07-11T17:59:05.886Z | TOOL RESULT: view_image]**

Image service returned the requested local image to the model for visual inspection.

**[Step 1880 | 2026-07-11T17:59:24.852Z | MODEL]**

The photos show Round 4 films are much more heterogeneous than the text-only workflow would suggest: XRD carriers include very pale/light deposits for some left-column samples and dark/ringed deposits for the right column; FTIR carriers also vary strongly in loading and wetting. I'm computing trace-based metrics now so the design decision does not overread a single photo.

**[Step 1881 | 2026-07-11T17:59:50.529Z | TOOL CALL: exec_command]**

Ran Python to parse, transform or validate structured experiment data. Target(s): `lab303_uor_nicofe_closed_loop`, `100`, `01_`, `supplement`, `diff`. Working directory: `/workspace`.

**[Step 1882 | 2026-07-11T17:59:51.909Z | TOOL RESULT: exec_command]**

Shell tool completed successfully; it returned 107 line(s) (3558 bytes) to the model. Reported long numeric identifier(s): `8404384896467723`, `00892253163996976`, `308203306988725`, `941338516315582`, `9548491577305643`, `07742069701540434`.

**[Step 1883 | 2026-07-11T18:00:17.127Z | REASONING SUMMARY 1/2]**

***Planning experiments and analysis***

*I need to create UV spectra plots using PIL and inspect XRD peaks, focusing on different sample formulas. I've got a list of samples to analyze. The main task is to compare main and supplemental experiments, especially since some main tests might've failed. Examining the layout in XRD photos could help, although it looks complicated. I'll also need to consider examining the raw electrochemical data for consistency and generate a summary table.*

**[Step 1884 | 2026-07-11T18:00:17.127Z | REASONING SUMMARY 2/2]**

***Checking round 4 summary***

*I need to check the round 4 summary in the JSON file to see if it includes supplemental data first, among other details. This might help ensure I've got the latest updates or any important comparisons ready. I want to make sure that everything is organized, so I can reference it easily. Let's see what the JSON reveals about the round 4 findings!*

**[Step 1885 | 2026-07-11T18:00:18.069Z | TOOL CALL: exec_command]**

Ran Python to parse, transform or validate structured experiment data. Target(s): `round4_combined_data_summary.json`. Working directory: `/workspace`.

**[Step 1886 | 2026-07-11T18:00:18.136Z | TOOL RESULT: exec_command]**

Shell tool completed successfully; it returned 28 line(s) (2764 bytes) to the model. Reported long numeric identifier(s): `00947855791016895`, `00892253163996976`, `008922161544421256`, `941338516315582`, `2784712529079429`, `308203306988725`.

**[Step 1887 | 2026-07-11T18:00:38.153Z | REASONING SUMMARY]**

***Analyzing structural data***

*I need to supplement maps for possibly result-1 sample 1, focusing on task name samples 1, 4, and 5. I should also analyze structural aspects, looking at the XRD profiles for samples 1, 2, and 6, which show broad peaks around 28-30. Meanwhile, samples 3 and 5 have peaks around 21-23. There's also a mixed duplicate for S4. I find it interesting that sample 1 is high at XRD 28 and S5 at XRD 21, while S4 is varied. I need to inspect UV metrics more closely.*

**[Step 1888 | 2026-07-11T18:00:39.726Z | TOOL CALL: exec_command]**

Ran Python to parse, transform or validate structured experiment data. Target(s): `round4_combined_data_summary.json`, `A700`. Working directory: `/workspace`.

**[Step 1889 | 2026-07-11T18:00:39.780Z | TOOL RESULT: exec_command]**

Shell tool completed successfully; it returned 10 line(s) (614 bytes) to the model.

**[Step 1890 | 2026-07-11T18:01:26.197Z | REASONING SUMMARY 1/5]**

***Evaluating experimental results***

*I'm looking at the UV negative baseline without any distinction and thinking it might be useful to create some plots. I should also compare results with R2 and R3. For Round 4, the supplemental data shows higher sustained currents with S5 outperforming others, but S6 is notably lower without a retest. Moving forward, I need to focus on high-performance candidates, replicate high-current samples, and refine the composition for Round 5. Let's ensure reproducibility as well.*

**[Step 1891 | 2026-07-11T18:01:26.197Z | REASONING SUMMARY 2/5]**

***Selecting sample compositions***

*I need to choose samples for Round 5, focusing on different compositions. I'm considering replicates of Ni80Fe10Co10 with carbon variations, plus exploring other combinations like Ni75Fe10Co15 and Ni85Fe10Co5. Higher iron levels might be intriguing too, especially for improving*

*performance. Since R4 showed that higher carbon worked best, I should keep that in mind. I'll refine the compositions around the nickel-rich variants and ensure I have a strong set of samples for high-impact results.*

**[Step 1892 | 2026-07-11T18:01:26.197Z | REASONING SUMMARY 3/5]**

***Evaluating liquid measurements***

*I'm figuring out the single source amounts for a liquid and whether I can exceed 1mL. The guidelines say the maximum for a single addition is 1mL, with a total of 3mL allowed. I need to decide how to split the amounts across different bottles. For instance, I could use 0.625mL from two bottles for C25 and smaller amounts for C20 and C15, ensuring everything stays under the total limits. This simplifies things while adhering to the rules!*

**[Step 1893 | 2026-07-11T18:01:26.197Z | REASONING SUMMARY 4/5]**

***Calculating water requirements***

*I'm calculating the water makeup needed for a synthesis, starting with a total volume of 5.0mL. After subtracting certain components, I get specific amounts for each carbon sample. For example, C20 needs 2.1mL for samples 1, 2, 5, and 6, while C15 and C25 have their own volumes. I realize the total for all samples stays within the limit, but I'm wondering if I should improve the electrochemical reproducibility since some main tests didn't work out. I might consider retesting all samples to ensure accuracy.*

**[Step 1894 | 2026-07-11T18:01:26.197Z | REASONING SUMMARY 5/5]**

***Planning for summary plots***

*I'm considering how to create summary plots and a contact sheet without using matplotlib. First, I'll focus on creating overlays for the plots. It's important that I produce the final document effectively, so I need to ensure that each step is clear. I can think about what information to include in the overlays while keeping in mind the overall layout of the document. Let's get organized and make this work!*

**[Step 1895 | 2026-07-11T18:01:27.200Z | MODEL]**

A clear Round 4 pattern is emerging: the main electrochemical traces for S1/S4/S5 were near baseline, but the supplemental retests for those same samples were high-current and stable, with S5 highest. I'm making plots and a returned-data montage now so the discrepancy is documented rather than hidden.

**[Step 1896 | 2026-07-11T18:02:11.451Z | TOOL CALL: exec_command]**

Ran Python analysis to generate or inspect a scientific plot. Target(s): `lab303_uor_nicofe_closed_loop`, `S5`, `XRD_drop_casting_photo.png`, `Infrared_spectroscopy_drop_casting_photo.png`, `Fourth_Round_Experiment_Feedback.png`. Working directory: `/workspace`.

**[Step 1897 | 2026-07-11T18:02:13.264Z | TOOL RESULT: exec_command]**

Shell tool completed successfully; it returned 5 line(s) (280 bytes) to the model.

**[Step 1898 | 2026-07-11T18:02:30.253Z | TOOL CALL: view_image]**

Opened a local experimental image for visual inspection. Image: `round4_quantitative_overlays.png`.

**[Step 1899 | 2026-07-11T18:02:30.837Z | TOOL CALL: view_image]**

Opened a local experimental image for visual inspection. Image: `round4_uvvis_overlays.png`.

**[Step 1900 | 2026-07-11T18:02:30.891Z | TOOL RESULT: view_image]**

Image service returned the requested local image to the model for visual inspection.

**[Step 1901 | 2026-07-11T18:02:30.893Z | TOOL RESULT: view_image]**

Image service returned the requested local image to the model for visual inspection.

**[Step 1902 | 2026-07-11T18:03:10.891Z | MODEL]**

For Round 5 I'm going to emphasize reproducibility and the high-current Ni/Fe/Co-carbon window, not add a new chemistry variable. Before writing the dispatch workflow, I'm re-reading the exact workstation constraints and audit files for the same executable synthesis, characterization, electrochemistry, and UV-Vis route.

**[Step 1903 | 2026-07-11T18:03:10.892Z | TOOL CALL: exec_command]**

Read selected file content for inspection. Target(s): `SKILL.md`. Working directory: `/workspace`.

**[Step 1904 | 2026-07-11T18:03:10.904Z | TOOL CALL: exec_command]**

Read selected file content for inspection. Target(s): `SKILL.md`. Working directory: `/workspace`.

**[Step 1905 | 2026-07-11T18:03:10.907Z | TOOL CALL: exec_command]**

Read selected file content for inspection. Target(s): `SKILL.md`. Working directory: `/workspace`.

**[Step 1906 | 2026-07-11T18:03:10.908Z | TOOL CALL: exec_command]**

Read selected file content for inspection. Target(s): `SKILL.md`. Working directory: `/workspace`.

**[Step 1907 | 2026-07-11T18:03:10.971Z | TOOL RESULT: exec_command]**

Shell tool completed successfully; it returned 224 line(s) (14552 bytes) to the model. Reported long numeric identifier(s): `1427568512205824`, `2002186824385539`, `1435269745574912`, `2270587971011585`.

**[Step 1908 | 2026-07-11T18:03:10.972Z | TOOL RESULT: exec_command]**

Shell tool completed successfully; it returned 166 line(s) (7478 bytes) to the model. Reported long numeric identifier(s): `1327807622448128`, `1213184425559040`, `1526864804185088`, `1656241422599168`.

**[Step 1909 | 2026-07-11T18:03:10.973Z | TOOL RESULT: exec_command]**

Shell tool completed successfully; it returned 196 line(s) (10695 bytes) to the model. Reported long numeric identifier(s): `1654567694238720`, `1664870018155520`, `1656282895188992`, `1664783667397632`.

**[Step 1910 | 2026-07-11T18:03:10.973Z | TOOL RESULT: exec_command]**

Shell tool completed successfully; it returned 151 line(s) (26086 bytes) to the model. Reported long numeric identifier(s): `1390252863751168`, `1509591942430720`, `1549499770798080`, `1571299393601536`, `1339275293851648`, `2164305671290880`.

**[Step 1911 | 2026-07-11T18:03:27.873Z | MODEL]**

The relevant workstation limits support the same six-vial route and the next design's carbon split keeps each 1 mL source-bottle total below 3 mL. I'm doing the audit-rule pass next, then I'll write the Round 4 returned document and Round 5 workflow request.

**[Step 1912 | 2026-07-11T18:03:29.892Z | TOOL CALL: exec_command]**

Read selected file content for inspection. Target(s): `references_audit`. Working directory: `/workspace`.

**[Step 1913 | 2026-07-11T18:03:29.960Z | TOOL RESULT: exec_command]**

Shell tool completed successfully; it returned 291 line(s) (8676 bytes) to the model. Reported long numeric identifier(s): `1427568512205824`, `2002186824385539`, `1435269745574912`, `2270587971011585`, `1327807622448128`, `1213184425559040`.

**[Step 1914 | 2026-07-11T18:03:55.371Z | MODEL]**

Round 5 will test whether the new best condition is reproducible and whether the carbon loading around that Ni-rich composition is optimized. I'm preparing the files now with two Ni80Fe10Co10-C20 repeats, the two high-current Ni70Fe10Co20 benchmarks, and C15/C25 variants around the current best.

**[Step 1915 | 2026-07-11T18:04:40.918Z | TOOL CALL: exec_command]**

Ran Python to parse, transform or validate structured experiment data. Target(s): `lab303_uor_nicofe_closed_loop`, `active-mass`, `C20`, `v`, `min`. Working directory: `/workspace`.

**[Step 1916 | 2026-07-11T18:04:40.993Z | TOOL RESULT: exec_command]**

Shell tool completed successfully; it returned 20 line(s) (335 bytes) to the model.

**[Step 1917 | 2026-07-11T18:04:59.642Z | REASONING SUMMARY]**

***Validating workflow***

*I need to validate the workflow first and maybe do a bit of checking. It's probably a good idea to run the generator afterward. But before I do that, I should also check a few key parameters to ensure everything's set up correctly. I want to make sure I'm not missing anything important here. It's all about getting it right before moving on to the next step!*

**[Step 1918 | 2026-07-11T18:05:03.141Z | MODEL]**

The Round 4 returned document and Round 5 workflow request are written. I'm validating the JSON and the key liquid-handling/electrochemical parameters before sending it to the workflow generator.

**[Step 1919 | 2026-07-11T18:05:03.146Z | TOOL CALL: exec_command]**

Ran Python to parse, transform or validate structured experiment data. Target(s): `round5_workflow_request.json`, `null`, `round5_chemical_specifications.json`. Working directory: `/workspace`.

**[Step 1920 | 2026-07-11T18:05:03.150Z | TOOL CALL: exec_command]**

Ran Python to parse, transform or validate structured experiment data. Target(s): `round5_workflow_request.json`, `count`. Working directory: `/workspace`.

**[Step 1921 | 2026-07-11T18:05:03.151Z | TOOL CALL: exec_command]**

Read selected file content for inspection. Target(s): `round5_constraint_check.md`. Working directory: `/workspace`.

**[Step 1922 | 2026-07-11T18:05:03.250Z | TOOL RESULT: exec_command]**

Shell tool completed successfully; it returned 2 line(s) (3 bytes) to the model.

**[Step 1923 | 2026-07-11T18:05:03.250Z | TOOL RESULT: exec_command]**

Shell tool completed successfully; it returned 41 line(s) (4420 bytes) to the model.

**[Step 1924 | 2026-07-11T18:05:03.250Z | TOOL RESULT: exec_command]**

Shell tool completed successfully; it returned 26 line(s) (2229 bytes) to the model.

**[Step 1925 | 2026-07-11T18:05:22.471Z | MODEL]**

The Round 5 workflow request passes local JSON and parameter checks. I'm submitting it to the workflow generator now; if it returns a template ID, I'll dispatch it to Laboratory 303 and record the new task ID.

**[Step 1926 | 2026-07-11T18:05:24.908Z | TOOL CALL: exec_command]**

Ran Python to parse, transform or validate structured experiment data. Target(s): `generate.py`, `round5_workflow_request.json)`, `round5_workflow_generator_output.txt`. Working directory: `/workspace`.

**[Step 1927 | 2026-07-11T18:05:26.076Z | TOOL RESULT: exec_command]**

Shell tool completed successfully; it returned 0 line(s) (0 bytes) to the model.

**[Step 1928 | 2026-07-11T18:05:46.271Z | MODEL]**

The workflow generator call completed. I'm reading its response to confirm the template creation before dispatch.

**[Step 1929 | 2026-07-11T18:05:47.672Z | TOOL CALL: exec_command]**

Read selected file content for inspection. Target(s): `round5_workflow_generator_output.txt`. Working directory: `/workspace`.

**[Step 1930 | 2026-07-11T18:05:47.728Z | TOOL RESULT: exec_command]**

Shell tool completed successfully; it returned 13 line(s) (308 bytes) to the model. Reported task/template identifier(s): `2592405136310272`.

**[Step 1931 | 2026-07-11T18:06:07.266Z | MODEL]**

Workflow generation succeeded with template `2592405136310272`. I'm checking the current lab context and dispatching that template to Laboratory 303.

**[Step 1932 | 2026-07-11T18:06:07.266Z | TOOL CALL: exec_command]**

Ran a shell command to identify the active laboratory. Target(s): `get_current_lab_label.py`, `current_lab_label_round5_20260711.json`. Working directory: `/workspace`.

**[Step 1933 | 2026-07-11T18:06:07.266Z | TOOL CALL: exec_command]**

Ran a shell command to validate laboratory access and routing. Target(s): `check_lab_consistency.py`, `lab_consistency_303_round5_20260711.json`. Working directory: `/workspace`.

**[Step 1934 | 2026-07-11T18:06:07.451Z | TOOL RESULT: exec_command]**

Shell tool completed successfully; it returned 0 line(s) (0 bytes) to the model.

**[Step 1935 | 2026-07-11T18:06:07.451Z | TOOL RESULT: exec_command]**

Shell tool completed successfully; it returned 0 line(s) (0 bytes) to the model.

**[Step 1936 | 2026-07-11T18:06:30.385Z | MODEL]**

The lab context checks completed successfully. I'm now issuing the Round 5 task from template `2592405136310272`.

**[Step 1937 | 2026-07-11T18:06:37.803Z | TOOL CALL: exec_command]**

Ran a shell command to generate and dispatch a laboratory task. Target(s): `generate_task.py`, `round5_dispatch_output.json`. Working directory: `/workspace`.

**[Step 1938 | 2026-07-11T18:06:38.129Z | TOOL RESULT: exec_command]**

Shell tool completed successfully; it returned 0 line(s) (0 bytes) to the model.

**[Step 1939 | 2026-07-11T18:06:56.662Z | MODEL]**

The dispatch command completed. I'm parsing the dispatch response and checking the new task status in the Laboratory 303 task list.

**[Step 1940 | 2026-07-11T18:06:58.428Z | TOOL CALL: exec_command]**

Read selected file content for inspection. Target(s): `round5_dispatch_output.json`. Working directory: `/workspace`.

**[Step 1941 | 2026-07-11T18:06:58.442Z | TOOL CALL: exec_command]**

Read selected file content for inspection. Target(s): `current_lab_label_round5_20260711.json`, `lab_consistency_303_round5_20260711.json`. Working directory: `/workspace`.

**[Step 1942 | 2026-07-11T18:06:58.473Z | TOOL RESULT: exec_command]**

Shell tool completed successfully; it returned 8 line(s) (359 bytes) to the model. Reported task/template identifier(s): `2592407497211904`.

**[Step 1943 | 2026-07-11T18:06:58.483Z | TOOL RESULT: exec_command]**

Shell tool completed successfully; it returned 12 line(s) (192 bytes) to the model.

**[Step 1944 | 2026-07-11T18:07:20.892Z | MODEL]**

Round 5 has been dispatched as task `2592407497211904`. I'm saving the dispatch log and verifying the task record before setting the goal back to blocked.

**[Step 1945 | 2026-07-11T18:07:20.893Z | TOOL CALL: exec_command]**

Ran a shell command to query laboratory task records and statuses. Target(s): `select_task_list.py`, `round5_task_2592407497211904_status.json`. Working directory: `/workspace`.

**[Step 1946 | 2026-07-11T18:07:20.893Z | TOOL CALL: exec_command]**

Ran Python to parse, transform or validate structured experiment data. Target(s): `lab303_uor_nicofe_closed_loop`, `round5_workflow_request.json`, `round5_workflow_generator_output.txt`, `round5_dispatch_output.json`, `round4_returned_document.md`. Working directory: `/workspace`.

**[Step 1947 | 2026-07-11T18:07:21.093Z | TOOL RESULT: exec_command]**

Shell tool completed successfully; it returned 0 line(s) (0 bytes) to the model.

**[Step 1948 | 2026-07-11T18:07:21.094Z | TOOL RESULT: exec_command]**

Shell tool completed successfully; it returned 2 line(s) (66 bytes) to the model.

**[Step 1949 | 2026-07-11T18:07:40.716Z | MODEL]**

The dispatch log is saved. I'm reading the newly created task status now and then I'll mark the goal blocked per the lab loop.

**[Step 1950 | 2026-07-11T18:07:42.828Z | TOOL CALL: exec_command]**

Read selected file content for inspection. Target(s): `round5_task_2592407497211904_status.json`. Working directory: `/workspace`.

**[Step 1951 | 2026-07-11T18:07:42.874Z | TOOL RESULT: exec_command]**

Shell tool completed successfully; it returned 14 line(s) (293 bytes) to the model. Reported long numeric identifier(s): `2592407497211904`.

**[Step 1952 | 2026-07-11T18:08:04.420Z | MODEL]**

The Round 5 task is present in Laboratory 303 with status `未开始`. I'm setting the goal back to blocked now, pending executability feedback or physical completion.

**[Step 1953 | 2026-07-11T18:08:04.425Z | TOOL CALL: update_goal]**

Updated the autonomous-research goal status to `blocked`.

**[Step 1954 | 2026-07-11T18:08:04.461Z | TOOL RESULT: update_goal]**

Goal service returned a structured response. Status: `blocked`.

**[Step 1955 | 2026-07-11T18:09:06.736Z | MODEL]**

Retrieved and analyzed the two Round 4 Lab 303 result packages, incorporated the supplied process photos, prepared the returned document, designed Round 5, generated the workflow, and dispatched the new experiment to Laboratory 303.

**Retrieved Data**

- Main Round 4 task `2588712171995137`: status `错误`, but returned 45 physical result files from heating, XRD, FTIR, electrochemistry, and UV-Vis.
- Supplemental task `2591597001932800`: status `错误`, but returned 9 physical result files from electrochemical retests and UV-Vis measurements for samples 1, 4, and 5.
- Initial API call hit a remote-disconnect error; retry succeeded and all returned result files were downloaded.

Saved analysis files:
- [round4_returned_document.md](/workspace/lab303_uor_nicofe_closed_loop/round4_returned_document.md)
- [round4_combined_data_summary.csv](/workspace/lab303_uor_nicofe_closed_loop/round4_combined_data_summary.csv)
- [round4_combined_data_summary.json](/workspace/lab303_uor_nicofe_closed_loop/round4_combined_data_summary.json)
- [round4_quantitative_overlays.png](/workspace/lab303_uor_nicofe_closed_loop/round4_quantitative_overlays.png)
- [round4_uvvis_overlays.png](/workspace/lab303_uor_nicofe_closed_loop/round4_uvvis_overlays.png)
- [round4_ftir_overlay.png](/workspace/lab303_uor_nicofe_closed_loop/round4_ftir_overlay.png)
- [round4_returned_data_montage.jpg](/workspace/lab303_uor_nicofe_closed_loop/round4_returned_data_montage.jpg)

**Round 4 Findings**

The process photos show visibly loaded XRD and FTIR carriers, but with strong film heterogeneity: pale/thin deposits on some carriers and dark ring-like deposits on others. I treated XRD/FTIR intensity as semi-quantitative film/loading evidence, not direct composition quantitation.

Heating was close to target: maximum `71.0 C`, final `69.9 C`, and `84.0%` of logged points at or above `68 C`.

Electrochemistry is the decisive result, but the main and supplemental runs differ sharply for S1/S4/S5:

| Sample | Formulation | Main last-300 s | Supplemental last-300 s |
|---|---|---:|---:|
| S1 | `Ni70Fe10Co20_C10` | `0.009 mA` | `7.308 mA` |
| S2 | `Ni70Fe10Co20_C10` replicate | `0.077 mA` | no retest |
| S3 | `Ni70Fe10Co20_C5` | `0.020 mA` | no retest |
| S4 | `Ni70Fe10Co20_C20` | `0.011 mA` | `6.039 mA` |
| S5 | `Ni80Fe10Co10_C20` | `0.010 mA` | `7.797 mA` |
| S6 | `Ni60Fe20Co20_C20` | `0.825 mA` | no retest |

The best condition found so far is now **R4-S5: `Ni80Fe10Co10_C20`**, with `7.797 mA` mean current over the final 300 s and retention `1.001` in the supplemental electrochemical retest. R4-S1 was close at `7.308 mA`, and R4-S4 was also high at `6.039 mA`.

Structurally, the XRD patterns are broad and poorly crystalline. High current is not tied to one XRD maximum: S1's dominant maximum is near `28.12 deg`, while S5's is near `21.06 deg`. FTIR remains dominated by film-loading/baseline artifacts. UV-Vis spectra from both main and supplemental retained liquids nearly overlap, with A400 around `-0.234` to `-0.236`, so UV-Vis did not provide useful product ranking in this returned dataset.

**Round 5 Design Logic**

Round 5 is designed to validate the new best condition and determine whether carbon loading around `Ni80Fe10Co10` should be below, at, or above `20 mg`, while retaining the high-current `Ni70Fe10Co20` benchmarks.

Round 5 sample set:
- R5-S1: `Ni80Fe10Co10_C20`, best-condition replicate A.
- R5-S2: `Ni80Fe10Co10_C20`, best-condition replicate B.
- R5-S3: `Ni70Fe10Co20_C10`, R4-S1 benchmark repeat.
- R5-S4: `Ni70Fe10Co20_C20`, R4-S4 benchmark repeat.
- R5-S5: `Ni80Fe10Co10_C15`, lower-carbon neighbor.
- R5-S6: `Ni80Fe10Co10_C25`, higher-carbon neighbor.

Core fixed conditions remain:
- 0.900 mL total 0.25 M metal stock per vial.
- 1.000 mL 2.0 M KOH per vial.
- 5.000 mL synthesis volume.
- 70 C, 1000 rpm, 120 min synthesis aging.
- Solid-retention purification, 12000 rpm, 20 min, one 5 mL water wash.
- 60 C drying for 180 min.
- 4.8 mL `80:20 v/v DI water:isopropanol` redispersion.
- XRD/FTIR: 200 uL drop-cast, 1800 s drying.

- Electrochemistry: `1.0 M KOH + 0.33 M urea`, 20 mL; 100 uL/drop x 3 drops; retained anolyte diluted with 10 mL DI water from stock bottle No. 1 for UV-Vis.

Design and workflow files:
- [round5_design.md](/workspace/lab303_uor_nicofe_closed_loop/round5_design.md)
- [round5_chemical_specifications.json](/workspace/lab303_uor_nicofe_closed_loop/round5_chemical_specifications.json)
- [round5_constraint_check.md](/workspace/lab303_uor_nicofe_closed_loop/round5_constraint_check.md)
- [round5_workflow_request.json](/workspace/lab303_uor_nicofe_closed_loop/round5_workflow_request.json)

Executability controls included:
- All sub-mL reagent additions are consolidated in one `移液平台_1ml_V2` addition step.
- Carbon stock is split across two identical carbon-dispersion bottles; the C25 condition uses `0.625 mL + 0.625 mL`, so no 1 mL V2 source addition exceeds 1 mL and all source totals stay below 3 mL.
- `移液平台_5ml_V1` is used only for mL-scale DI-water and redispersion-solvent additions.
- One wash count is paired with one wash-liquid entry.
- Electrochemical retained-product bottle No. 1 is explicitly DI water, `10000 uL` per retained sample.

**Dispatch**
Workflow generation succeeded.

- Round 5 template ID: `2592405136310272`
- Round 5 task ID: `2592407497211904`
- Initial task status: `未开始`
- Dispatch log: [round5_dispatch_log.json](/workspace/lab303_uor_nicofe_closed_loop/round5_dispatch_log.json)
- Generator output: [round5_workflow_generator_output.txt](/workspace/lab303_uor_nicofe_closed_loop/round5_workflow_generator_output.txt)
- Task status file: [round5_task_2592407497211904_status.json](/workspace/lab303_uor_nicofe_closed_loop/round5_task_2592407497211904_status.json)

The goal is set back to `blocked`, pending your feedback on whether task `2592407497211904` is executable or physically completed.

**End of session record**

# Supplementary Figures

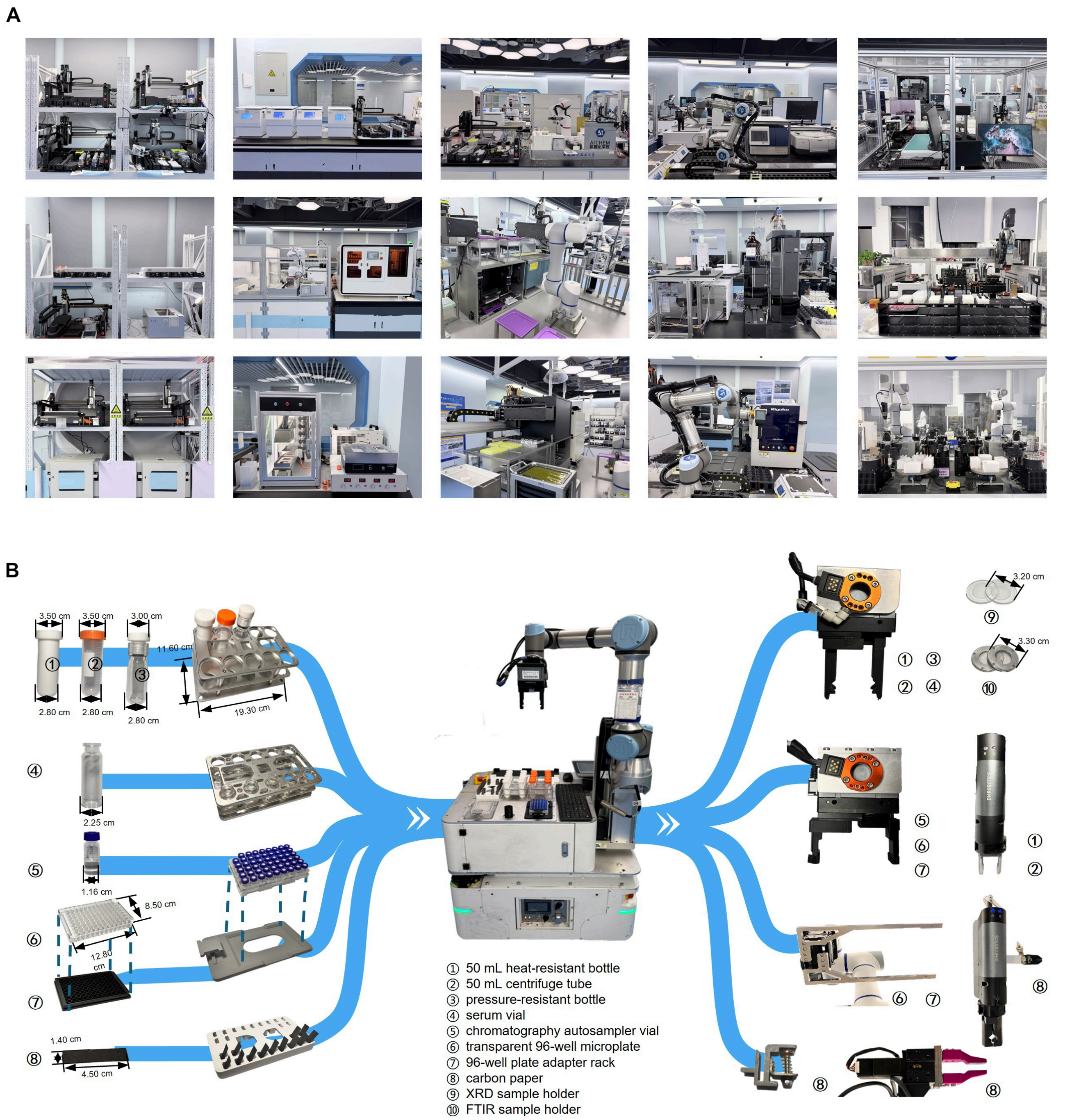

**Fig. S1. Physical implementation and standardized handling interfaces of the robotic chemistry laboratory. A, Representative views of the laboratory and its modular automated workstations. The montage documents the physical laboratory environment; the complete inventory of 45 agent-visible workstation contracts is provided in table S6. B, Standardized container formats and robotic-gripper interfaces for adapting diverse reactors and sample carriers to unified transfer units. Centrifuge tubes, PTFE reaction tubes, pressure-resistant glass reactors, vials, carbon-paper substrates and standard multiwell plates are accommodated using compatible transfer units and tray designs. A common robotic-gripper interface and transfer logic are reused across workstations, whereas reactor-specific fixtures within individual workstation modules meet local requirements for positioning, fixation, reaction, heating and characterization.**

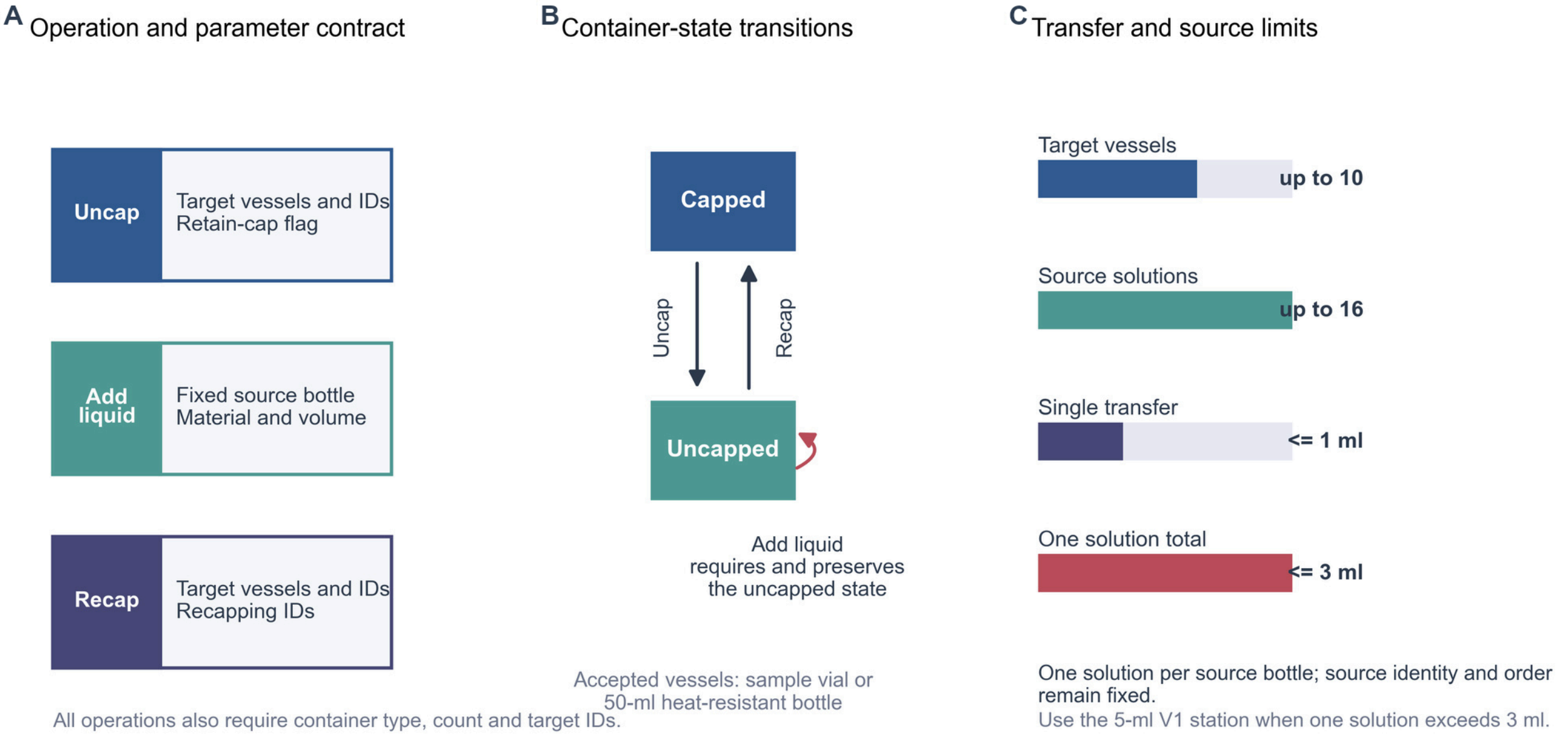

**Fig. S2. Contract and state transitions for Liquid_Handling_Station_1ml_V2. A, The three exposed operations and their principal parameter groups. All operations also require target-container type, count and identifiers. B, Allowed container-state transitions. Liquid addition requires and preserves an uncapped target; uncapping and recapping change the target state. C, Contract-level transfer and source limits. The station accepts sample vials or 50-ml heat-resistant bottles, at most 10 targets and 16 source solutions. A transfer is at most 1 ml, and one solution contributes at most 3 ml in total. One solution is assigned to each source bottle, whose identity and order remain fixed.**

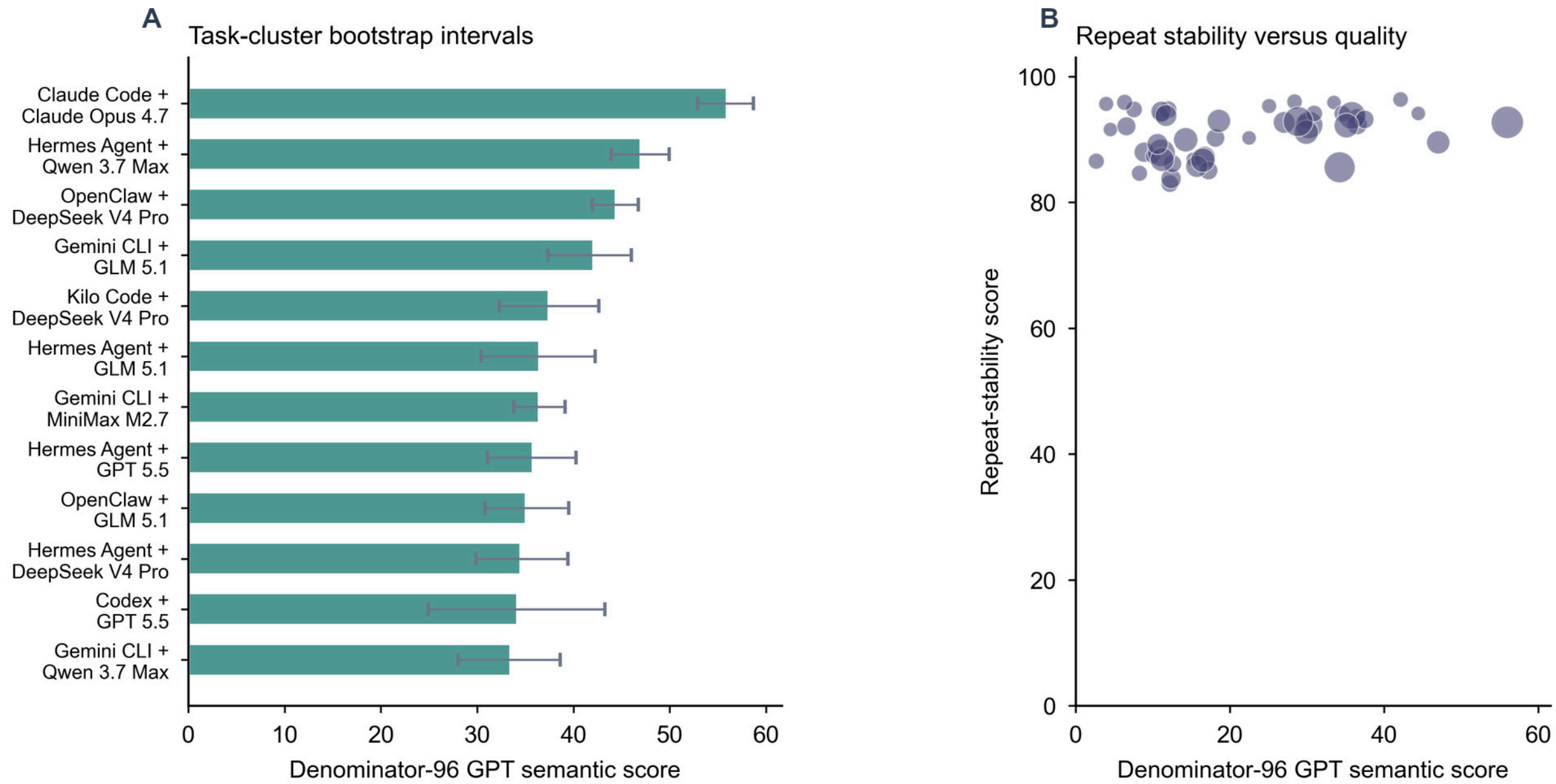

**Fig. S3. Task-cluster uncertainty and repeat stability. A, Denominator-96 GPT semantic scores for the 12 highest-scoring configurations. Error bars are 95% percentile intervals from 2,000 bootstrap resamples of the 32 task clusters. B, Denominator-96 semantic score versus repeat-stability score for all 48 observed configurations. Point area scales with verified-dispatch rate. Stability describes repeat dispersion and does not establish correctness.**

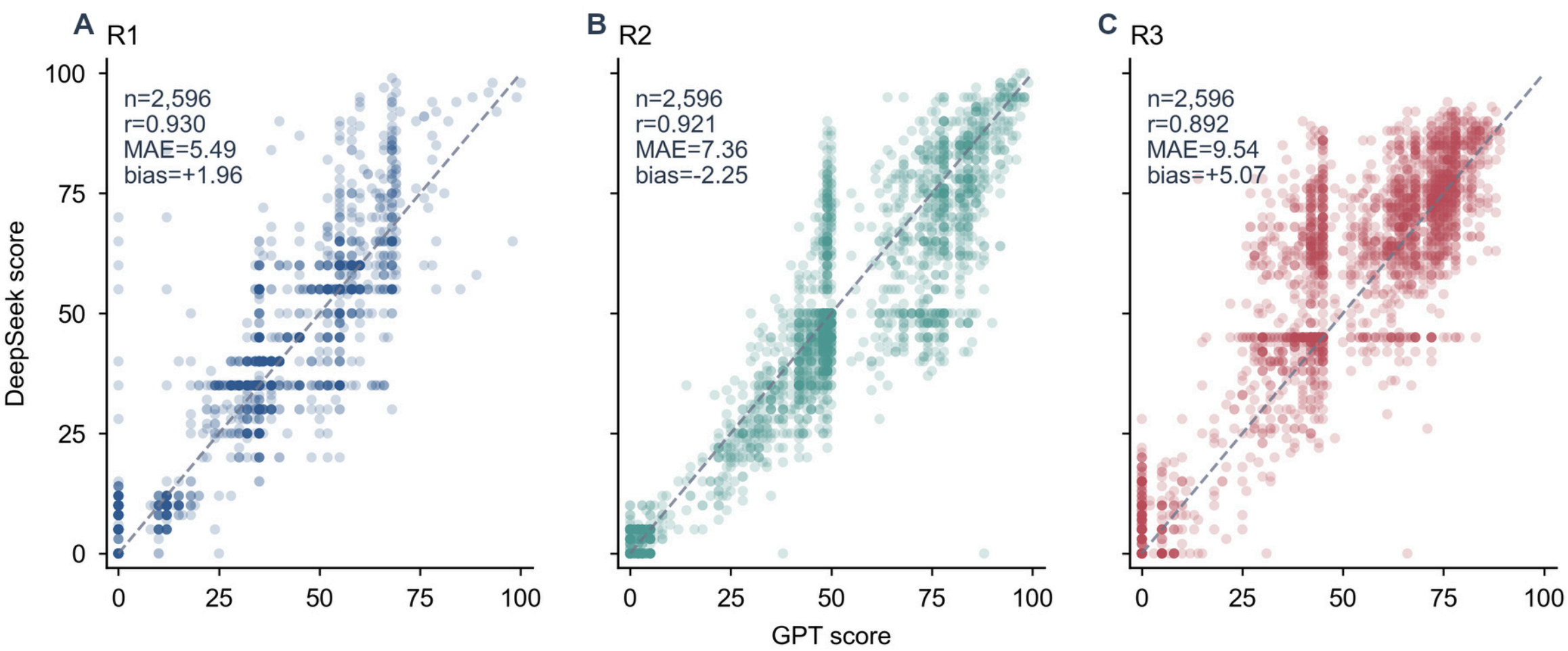

**Fig. S4. Agreement between the primary and validation judges. Trial-level GPT and DeepSeek scores are shown for A, R1 physical-system executability, B, R2 problem coverage and workflow completeness, and C, R3 visible scientific-design reasonableness. Each panel contains the paired plan scores available after the locked status filter. Dashed lines indicate identity. Insets report paired n and correlation statistics. Points are paired observations and have no independent error bars.**

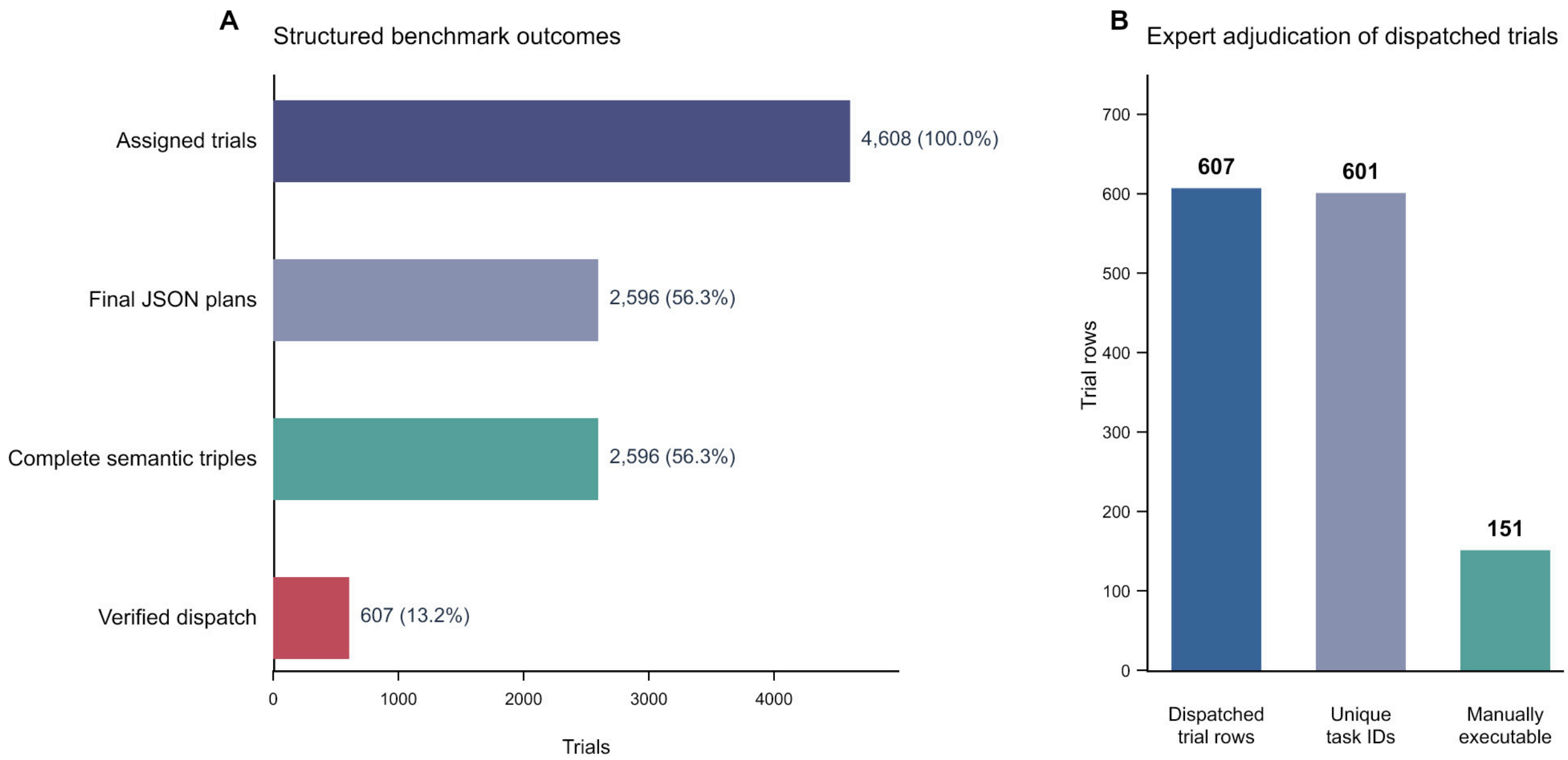

**Fig. S5. Outcome boundaries in the structured benchmark and expert-adjudicated dispatch endpoint. A, Assigned trials, final JSON plans, complete GPT semantic triples and verified-dispatch trials in the structured cohort (n = 4,608 trials). Percentages use all assigned trials as the denominator. B, All 607 verified-dispatch trial rows were adjudicated, representing 601 unique task identifiers and yielding 151 executable and 456 non-executable rows. N10 is nested within N9.**

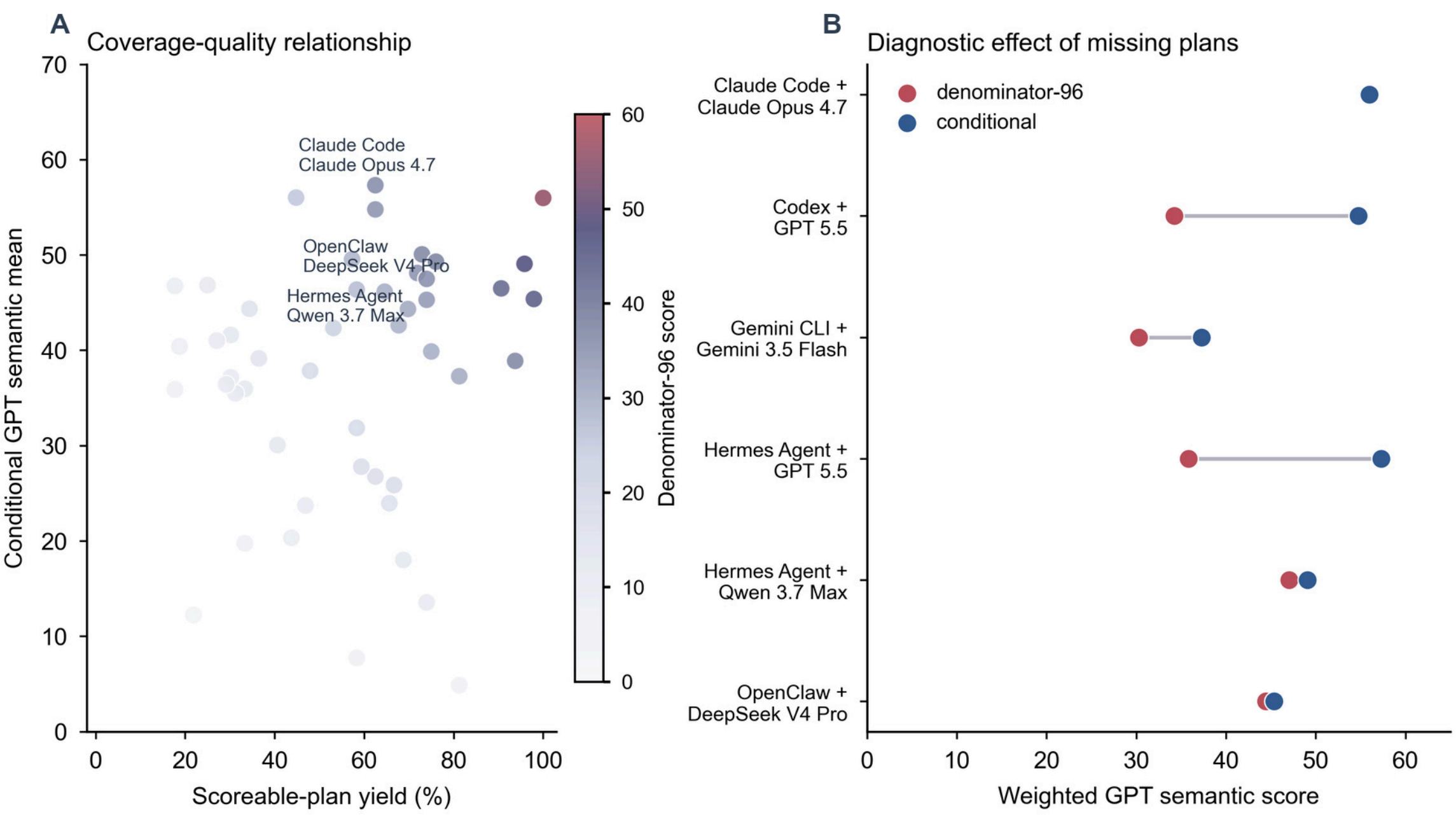

**Fig. S6. Scoreable-plan coverage and conditional semantic quality. A, Each point is one observed configuration (n = 48). The x axis is the number of finite GPT semantic triples divided by 96; the y axis is the conditional GPT semantic mean. Colour denotes the denominator-96 semantic score. B, Conditional and denominator-96 scores for six illustrative configurations. Connecting lines show the penalty associated with trials that did not produce complete plans.**

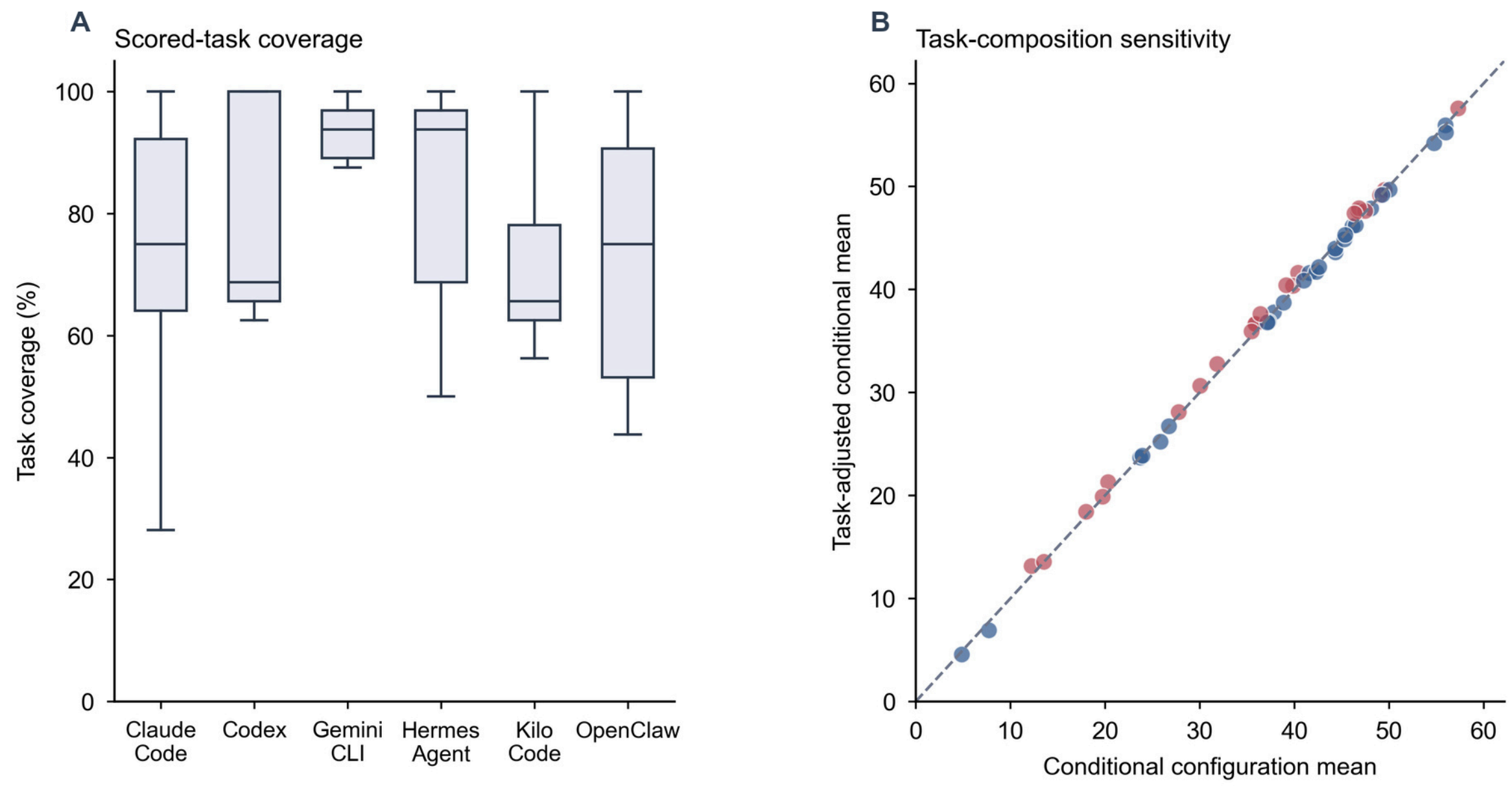

**Fig. S7. Scored-task coverage and task-composition sensitivity. A, Scored-task coverage across the 48 observed configurations, grouped by harness. Boxes show medians and interquartile ranges; whiskers extend to 1.5 times the interquartile range. Outliers are not displayed. B, Task-adjusted versus unadjusted conditional GPT semantic means for all 48 configurations. The dashed line indicates identity. The adjustment subtracts each task's global mean and adds the overall finite-plan mean.**

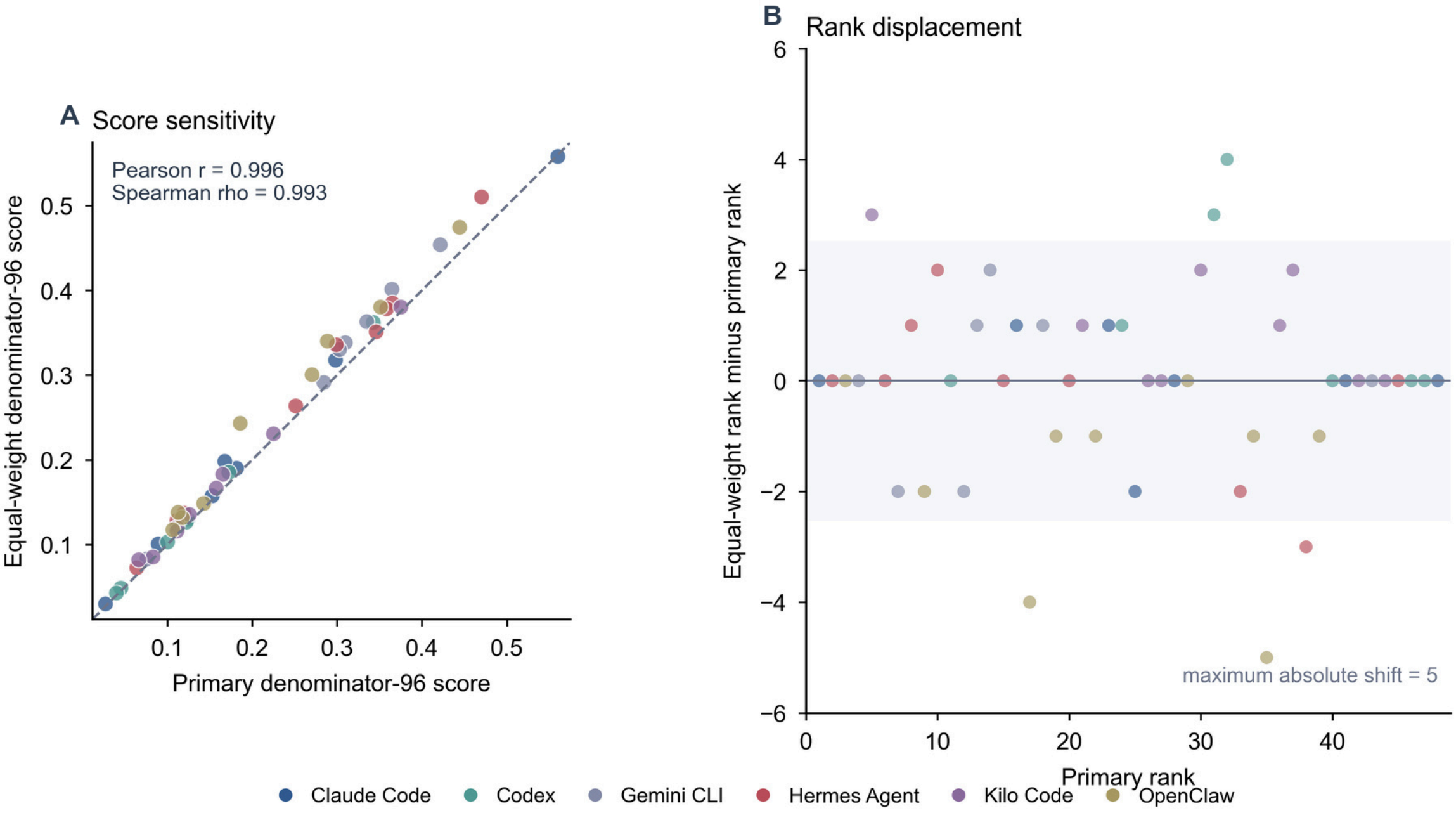

**Fig. S8. Sensitivity of denominator-96 semantic scores to dimension weights. A, Primary scores (0.50R1 + 0.25R2 + 0.25R3) compared with equal-weight scores for all 48 observed configurations. The dashed line indicates identity. B, Equal-weight rank minus primary rank, ordered by primary rank. The shaded band denotes shifts of at most two places. The scores were strongly associated (Pearson r = 0.996; Spearman rho = 0.993); the four highest configurations were unchanged, and the maximum absolute rank shift was five.**

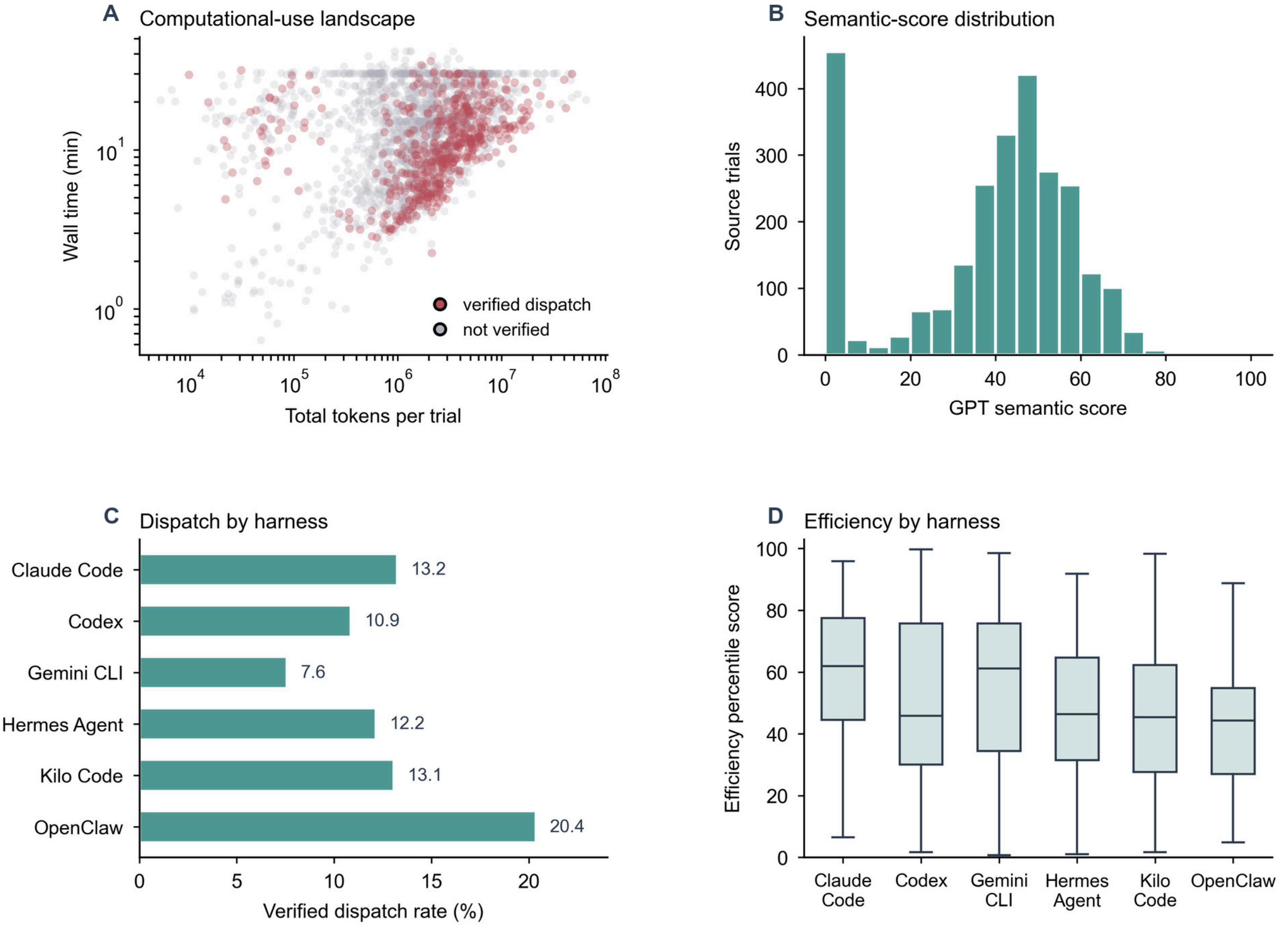

**Fig. S9. Computational use, semantic scores and verified dispatch. A, Total tokens and wall time for final-plan trials with positive values for both quantities, coloured by verified-dispatch status; both axes are logarithmic. B, Distribution of weighted GPT semantic scores across final plans. C, Verified-dispatch rate by harness, using all assigned trials in observed configurations. D, Trial-level efficiency percentile scores by harness. Boxes show medians and interquartile ranges; whiskers extend to 1.5 times the interquartile range. Outliers are not displayed.**

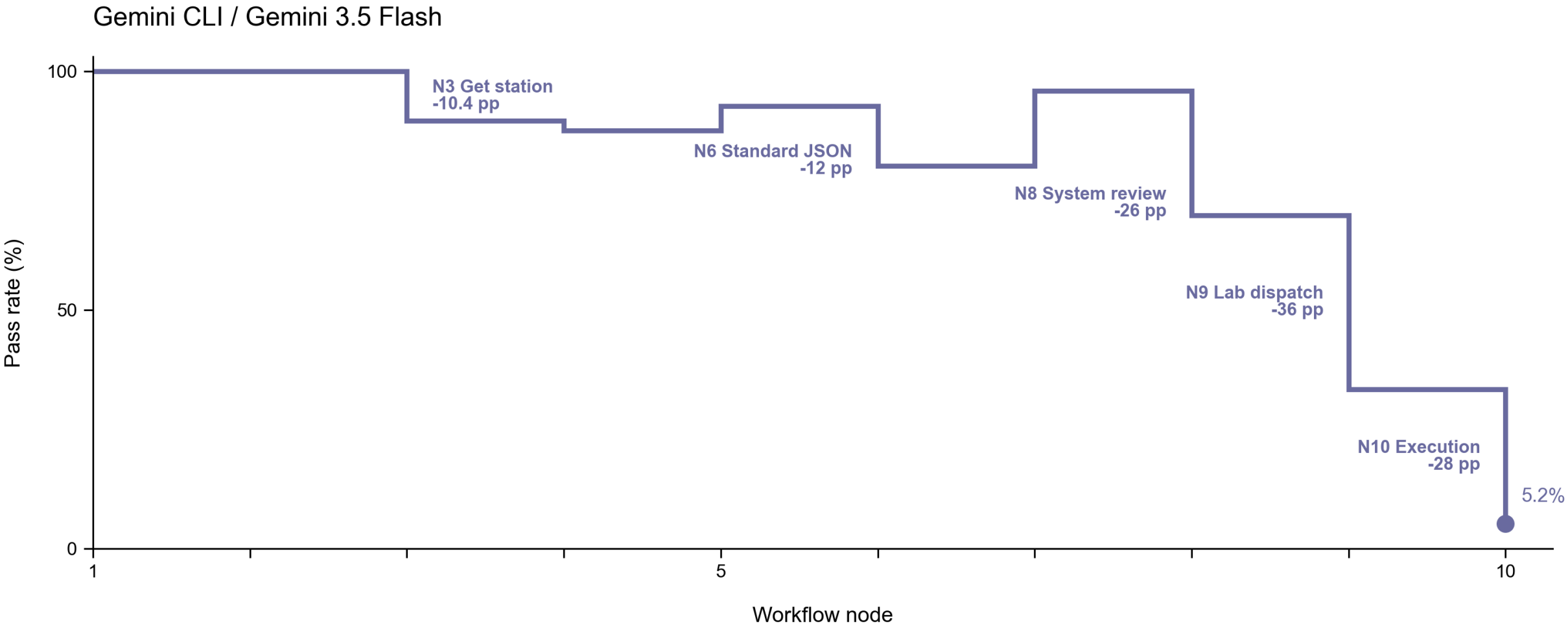

**Fig. S10. Node-specific workflow pass profile for Gemini CLI / Gemini 3.5 Flash. Profile summary. Across 96 benchmark trial records, the largest adjacent decline occurred at N9 Lab dispatch (-36.5 pp). The largest recovery occurred at N7 Workflow entry (+15.6 pp). N9 dispatch was 32/96 (33.3%), and N10 execution was 5/96 (5.2%); the executable-endpoint rank was 9 of 48.**

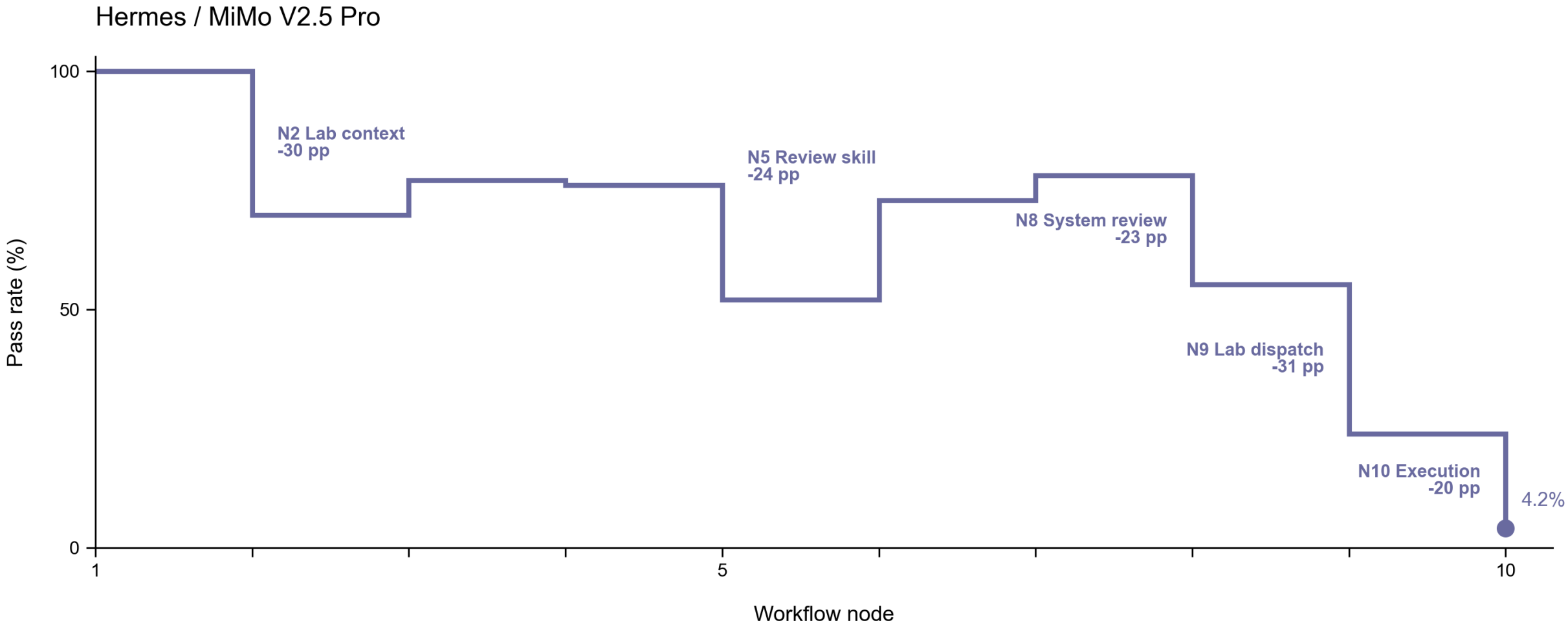

**Fig. S11. Node-specific workflow pass profile for Hermes / MiMo V2.5 Pro. Profile summary. Across 96 benchmark trial records, the largest adjacent decline occurred at N9 Lab dispatch (-31.2 pp). The largest recovery occurred at N6 Standard JSON (+20.8 pp). N9 dispatch was 23/96 (24.0%), and N10 execution was 4/96 (4.2%); the executable-endpoint rank was 10 of 48.**

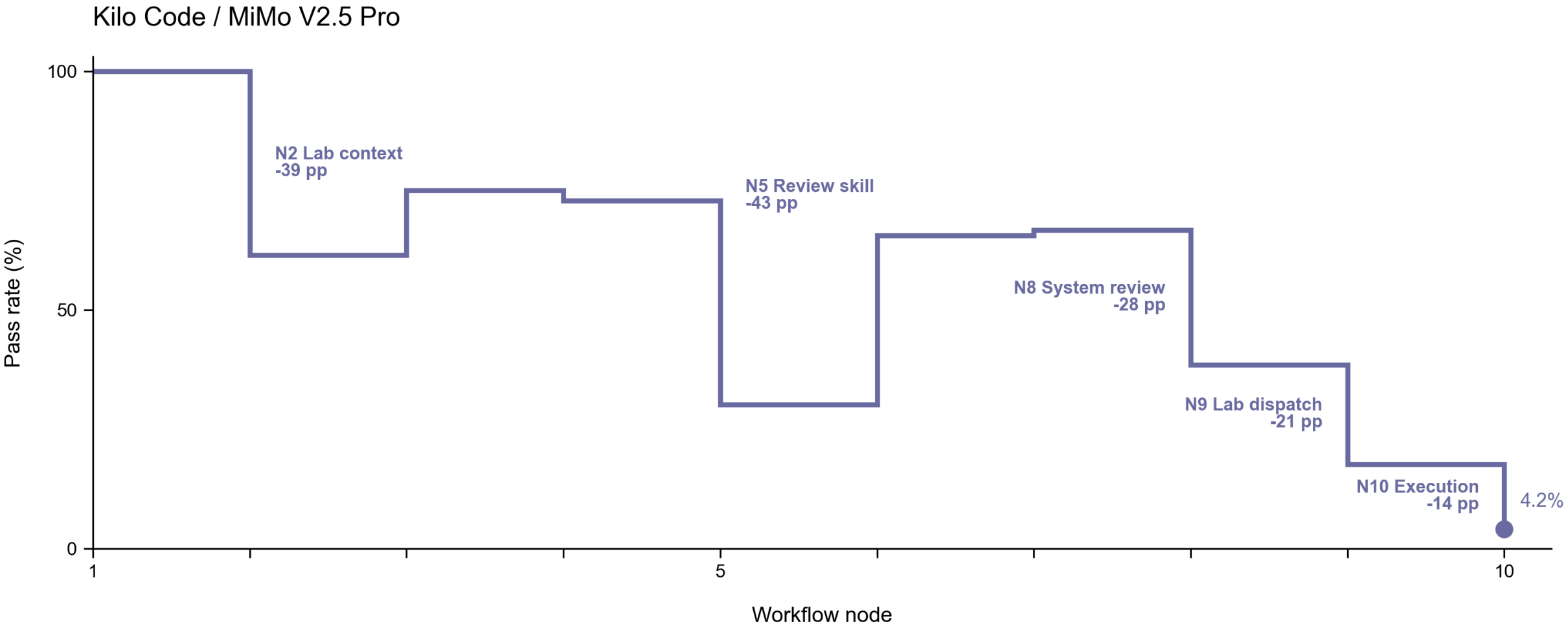

**Fig. S12. Node-specific workflow pass profile for Kilo Code / MiMo V2.5 Pro. Profile summary. Across 96 benchmark trial records, the largest adjacent decline occurred at N5 Review skill (-42.7 pp). The largest recovery occurred at N6 Standard JSON (+35.4 pp). N9 dispatch was 17/96 (17.7%), and N10 execution was 4/96 (4.2%); the executable-endpoint rank was 11 of 48.**

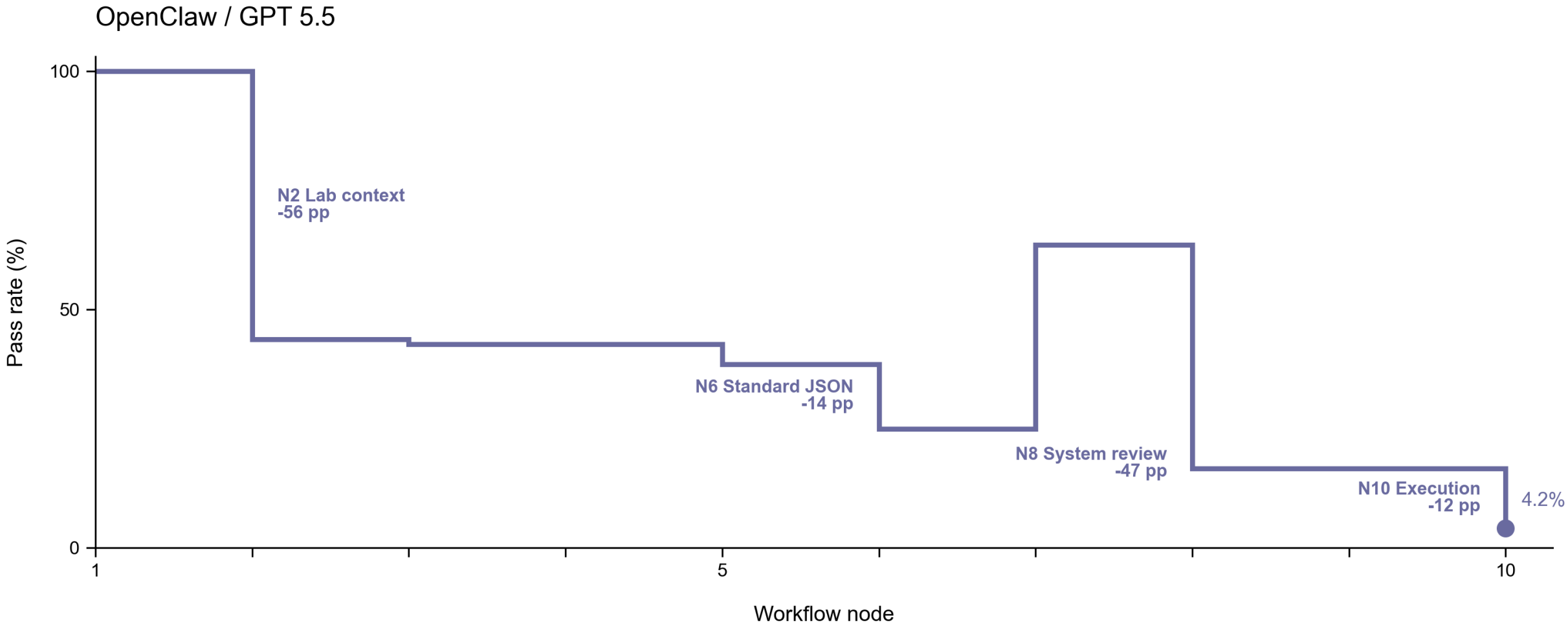

**Fig. S13. Node-specific workflow pass profile for OpenClaw / GPT 5.5. Profile summary. Across 96 benchmark trial records, the largest adjacent decline occurred at N2 Lab context (-56.2 pp). The largest recovery occurred at N7 Workflow entry (+38.5 pp). N9 dispatch was 16/96 (16.7%), and N10 execution was 4/96 (4.2%); the executable-endpoint rank was 12 of 48.**

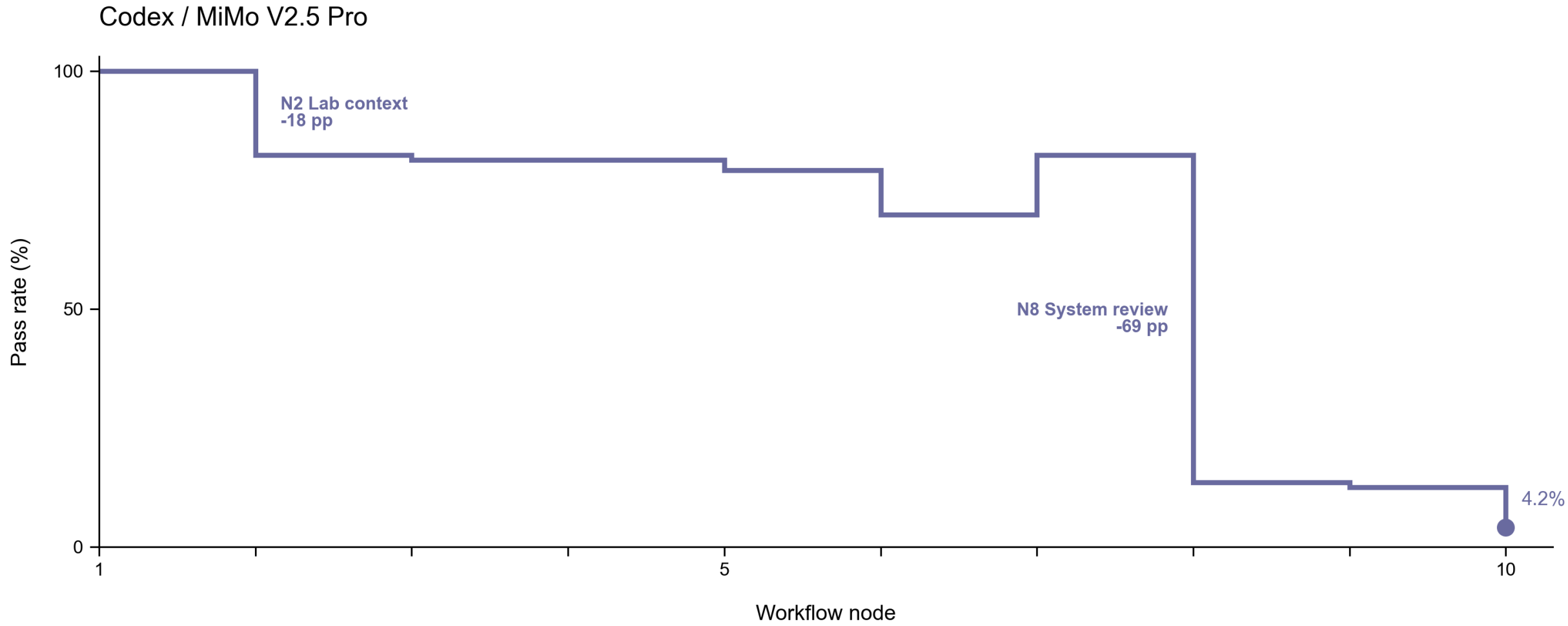

**Fig. S14. Node-specific workflow pass profile for Codex / MiMo V2.5 Pro. Profile summary. Across 96 benchmark trial records, the largest adjacent decline occurred at N8 System review (-68.7 pp). The largest recovery occurred at N7 Workflow entry (+12.5 pp). N9 dispatch was 12/96 (12.5%), and N10 execution was 4/96 (4.2%); the executable-endpoint rank was 13 of 48.**

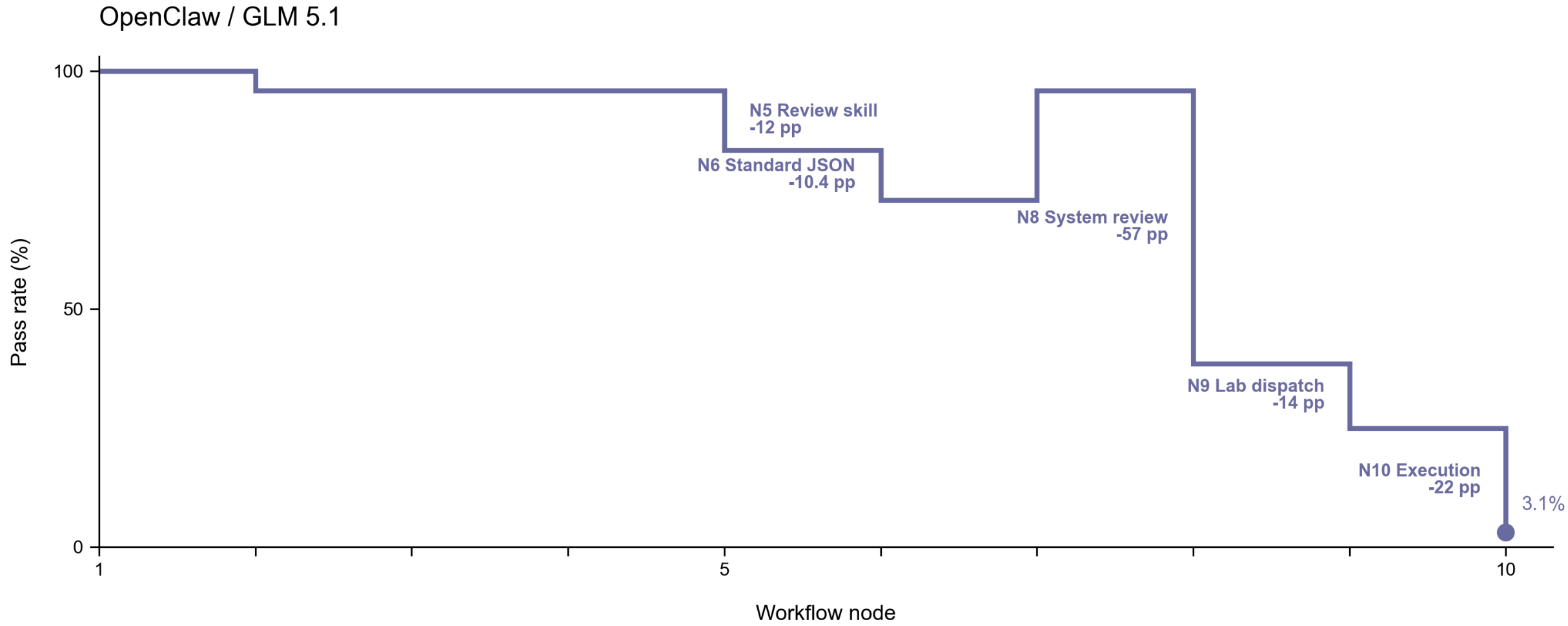

**Fig. S15. Node-specific workflow pass profile for OpenClaw / GLM 5.1. Profile summary. Across 96 benchmark trial records, the largest adjacent decline occurred at N8 System review (-57.3 pp). The largest recovery occurred at N7 Workflow entry (+22.9 pp). N9 dispatch was 24/96 (25.0%), and N10 execution was 3/96 (3.1%); the executable-endpoint rank was 14 of 48.**

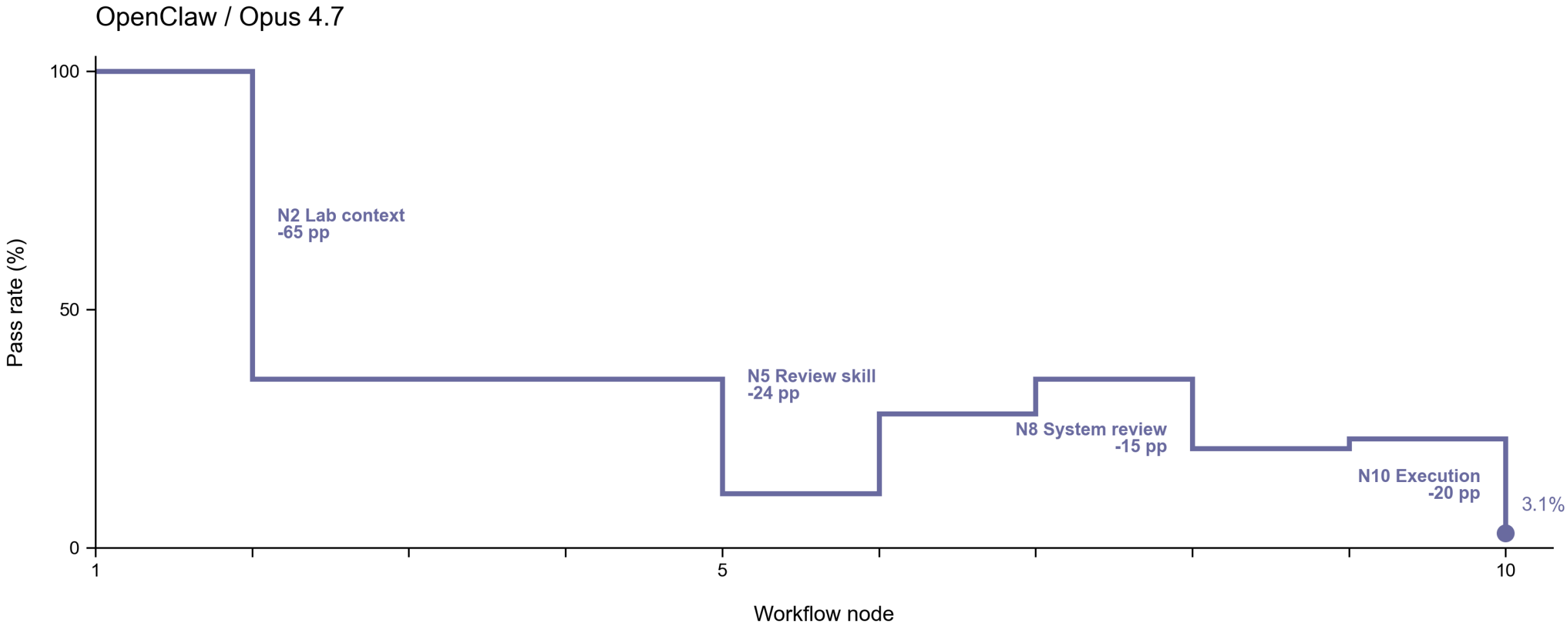

**Fig. S16. Node-specific workflow pass profile for OpenClaw / Opus 4.7. Profile summary. Across 96 benchmark trial records, the largest adjacent decline occurred at N2 Lab context (-64.6 pp). The largest recovery occurred at N6 Standard JSON (+16.7 pp). N9 dispatch was 22/96 (22.9%), and N10 execution was 3/96 (3.1%); the executable-endpoint rank was 15 of 48.**

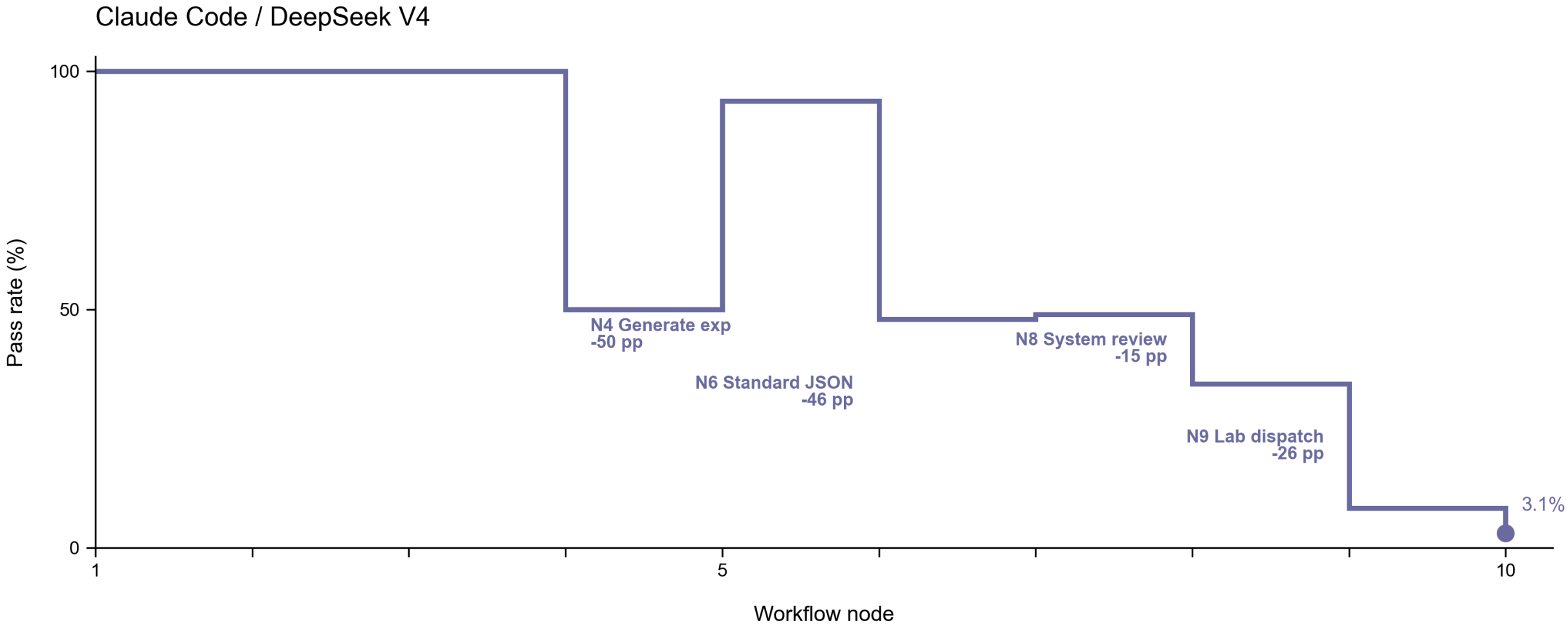

**Fig. S17. Node-specific workflow pass profile for Claude Code / DeepSeek V4. Profile summary. Across 96 benchmark trial records, the largest adjacent decline occurred at N4 Generate exp (-50.0 pp). The largest recovery occurred at N5 Review skill (+43.8 pp). N9 dispatch was 8/96 (8.3%), and N10 execution was 3/96 (3.1%); the executable-endpoint rank was 16 of 48.**

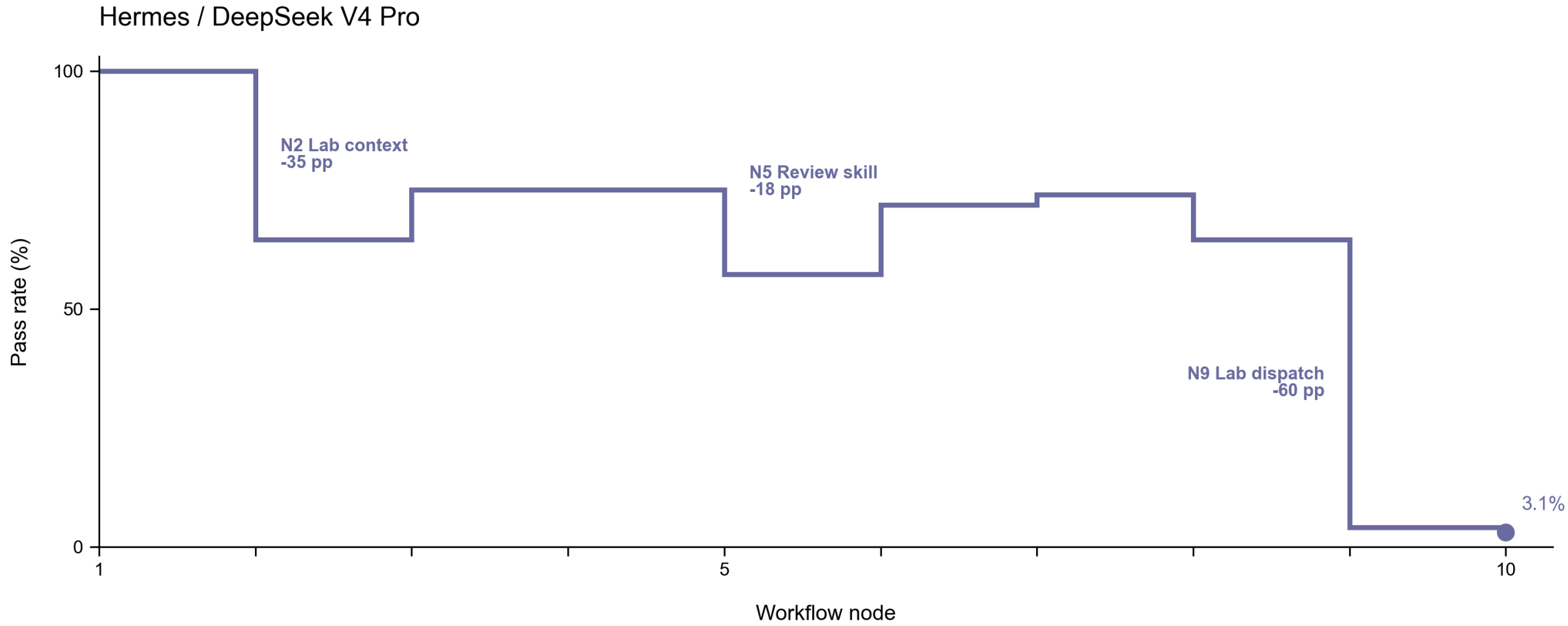

**Fig. S18. Node-specific workflow pass profile for Hermes / DeepSeek V4 Pro. Profile summary. Across 96 benchmark trial records, the largest adjacent decline occurred at N9 Lab dispatch (-60.4 pp). The largest recovery occurred at N6 Standard JSON (+14.6 pp). N9 dispatch was 4/96 (4.2%), and N10 execution was 3/96 (3.1%); the executable-endpoint rank was 17 of 48.**

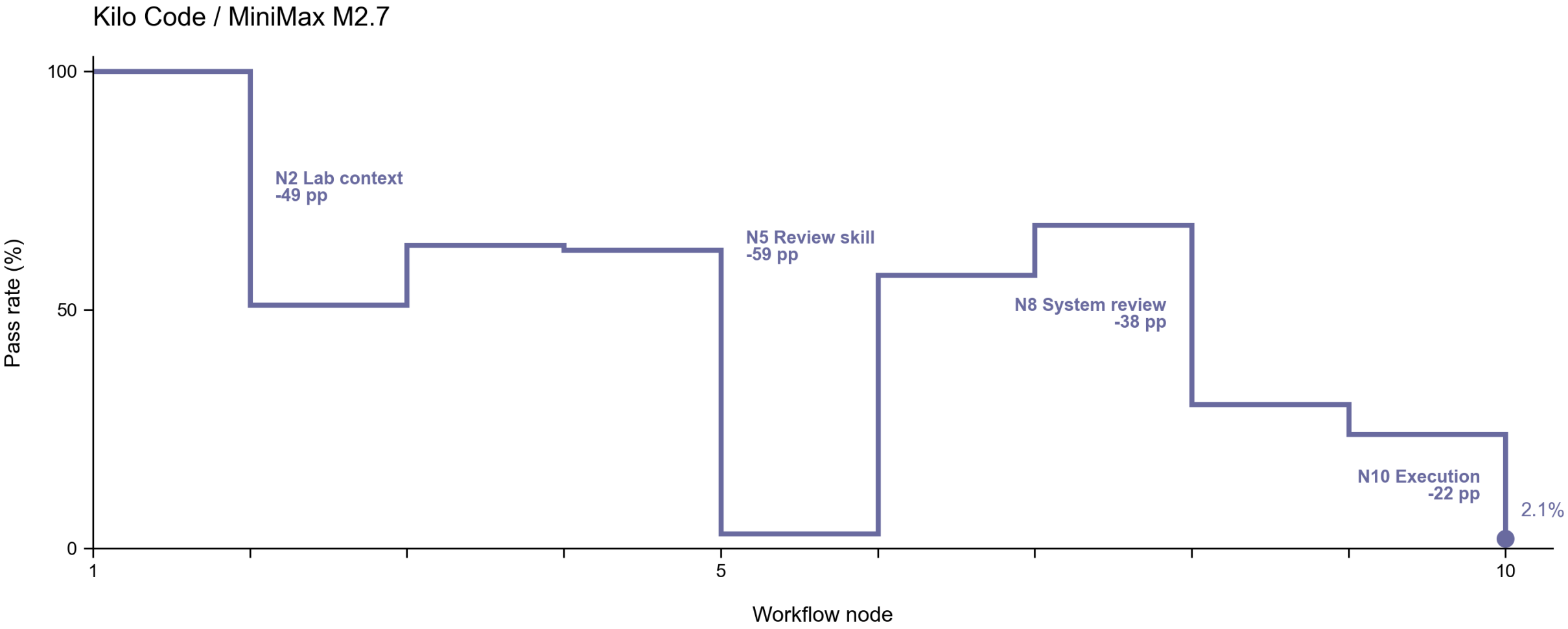

**Fig. S19. Node-specific workflow pass profile for Kilo Code / MiniMax M2.7. Profile summary. Across 96 benchmark trial records, the largest adjacent decline occurred at N5 Review skill (-59.4 pp). The largest recovery occurred at N6 Standard JSON (+54.2 pp). N9 dispatch was 23/96 (24.0%), and N10 execution was 2/96 (2.1%); the executable-endpoint rank was 18 of 48.**

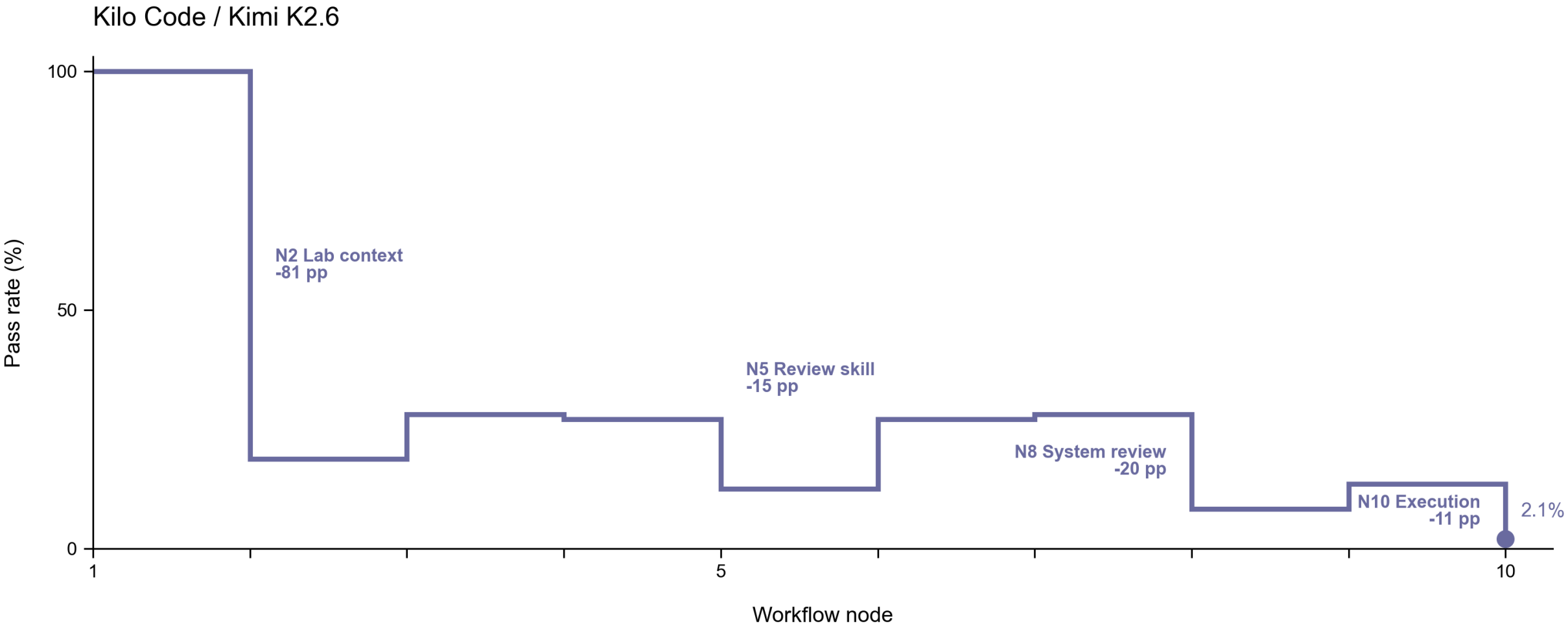

**Fig. S20. Node-specific workflow pass profile for Kilo Code / Kimi K2.6. Profile summary. Across 96 benchmark trial records, the largest adjacent decline occurred at N2 Lab context (-81.2 pp). The largest recovery occurred at N6 Standard JSON (+14.6 pp). N9 dispatch was 13/96 (13.5%), and N10 execution was 2/96 (2.1%); the executable-endpoint rank was 19 of 48.**

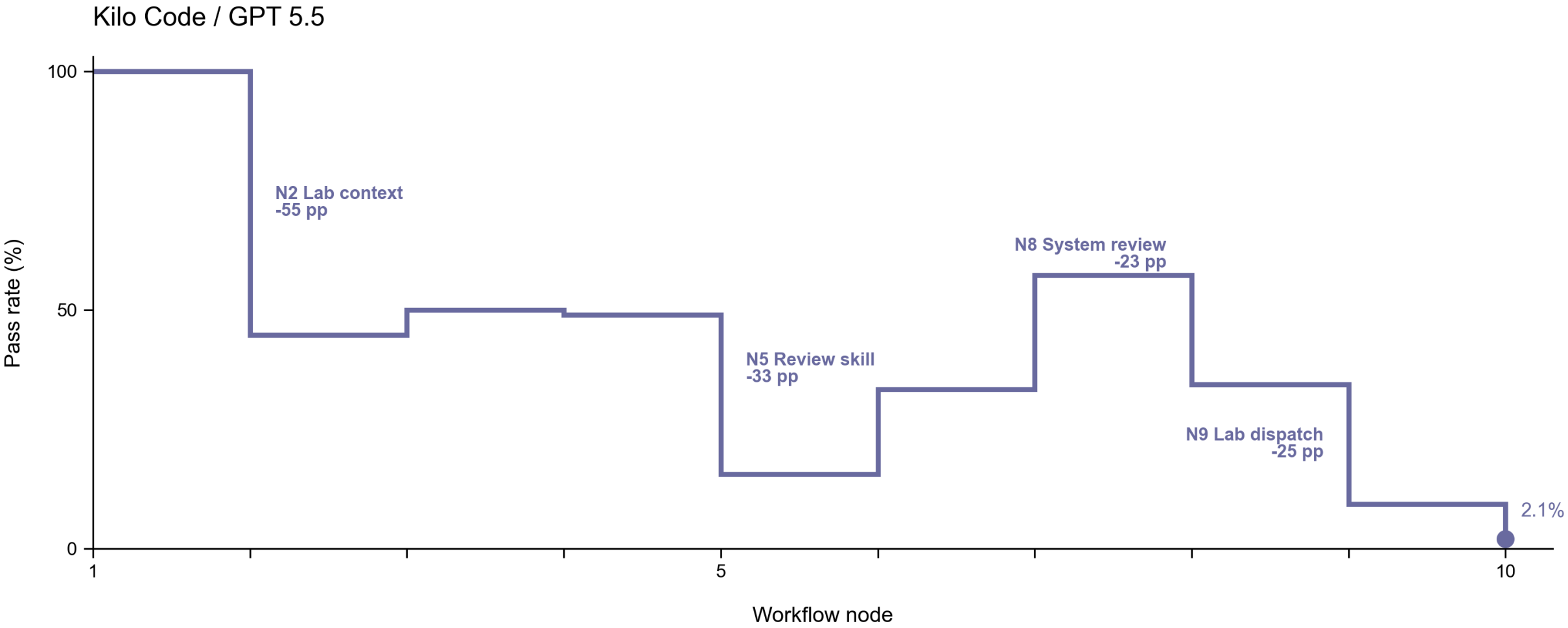

**Fig. S21. Node-specific workflow pass profile for Kilo Code / GPT 5.5. Profile summary. Across 96 benchmark trial records, the largest adjacent decline occurred at N2 Lab context (-55.2 pp). The largest recovery occurred at N7 Workflow entry (+24.0 pp). N9 dispatch was 9/96 (9.4%), and N10 execution was 2/96 (2.1%); the executable-endpoint rank was 20 of 48.**

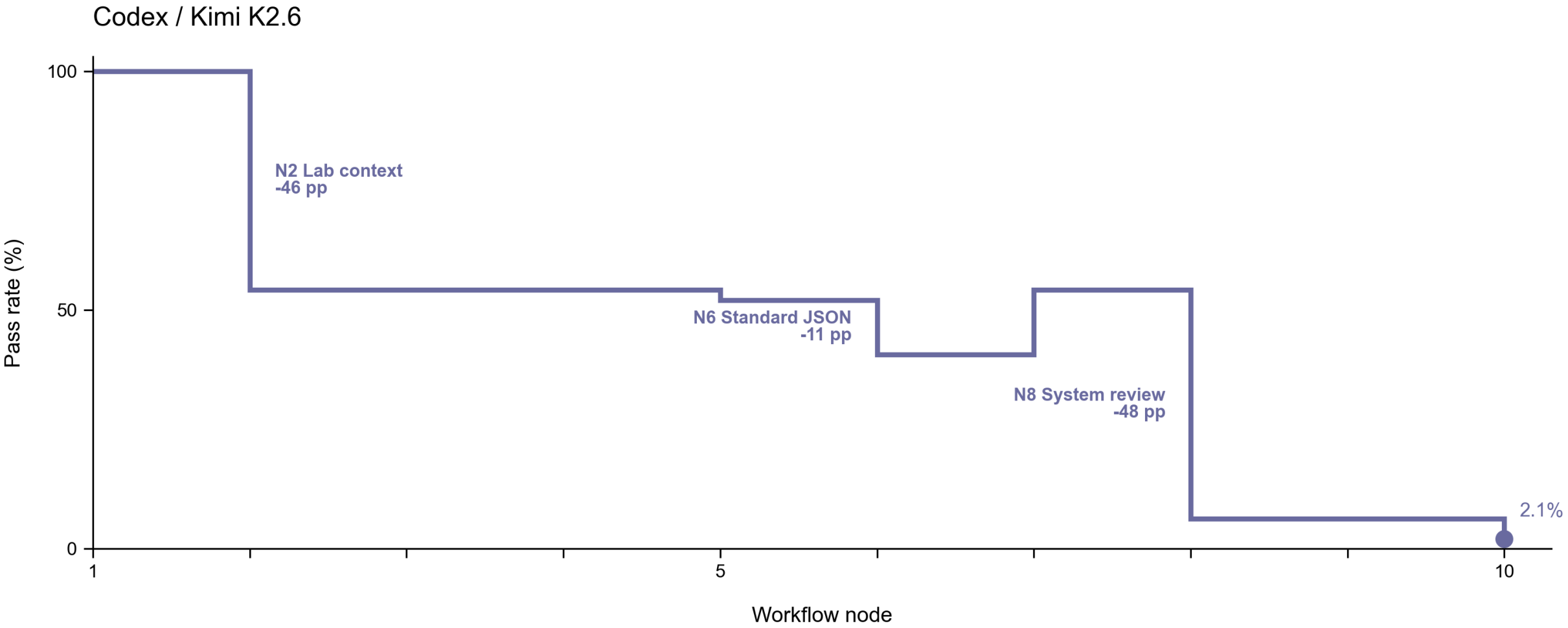

**Fig. S22. Node-specific workflow pass profile for Codex / Kimi K2.6. Profile summary. Across 96 benchmark trial records, the largest adjacent decline occurred at N8 System review (-47.9 pp). The largest recovery occurred at N7 Workflow entry (+13.5 pp). N9 dispatch was 6/96 (6.2%), and N10 execution was 2/96 (2.1%); the executable-endpoint rank was 21 of 48.**

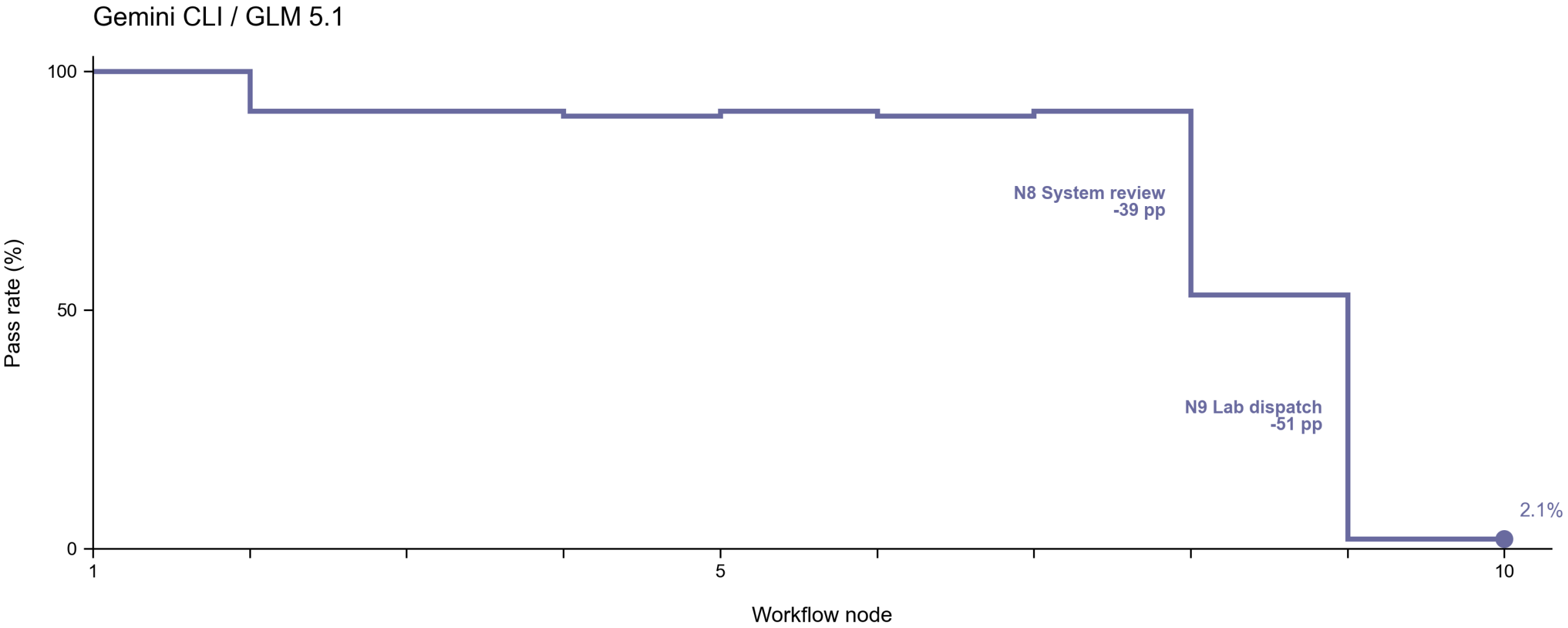

**Fig. S23. Node-specific workflow pass profile for Gemini CLI / GLM 5.1. Profile summary. Across 96 benchmark trial records, the largest adjacent decline occurred at N9 Lab dispatch (-51.0 pp). The largest recovery occurred at N5 Review skill (+1.0 pp). N9 dispatch was 2/96 (2.1%), and N10 execution was 2/96 (2.1%); the executable-endpoint rank was 22 of 48.**

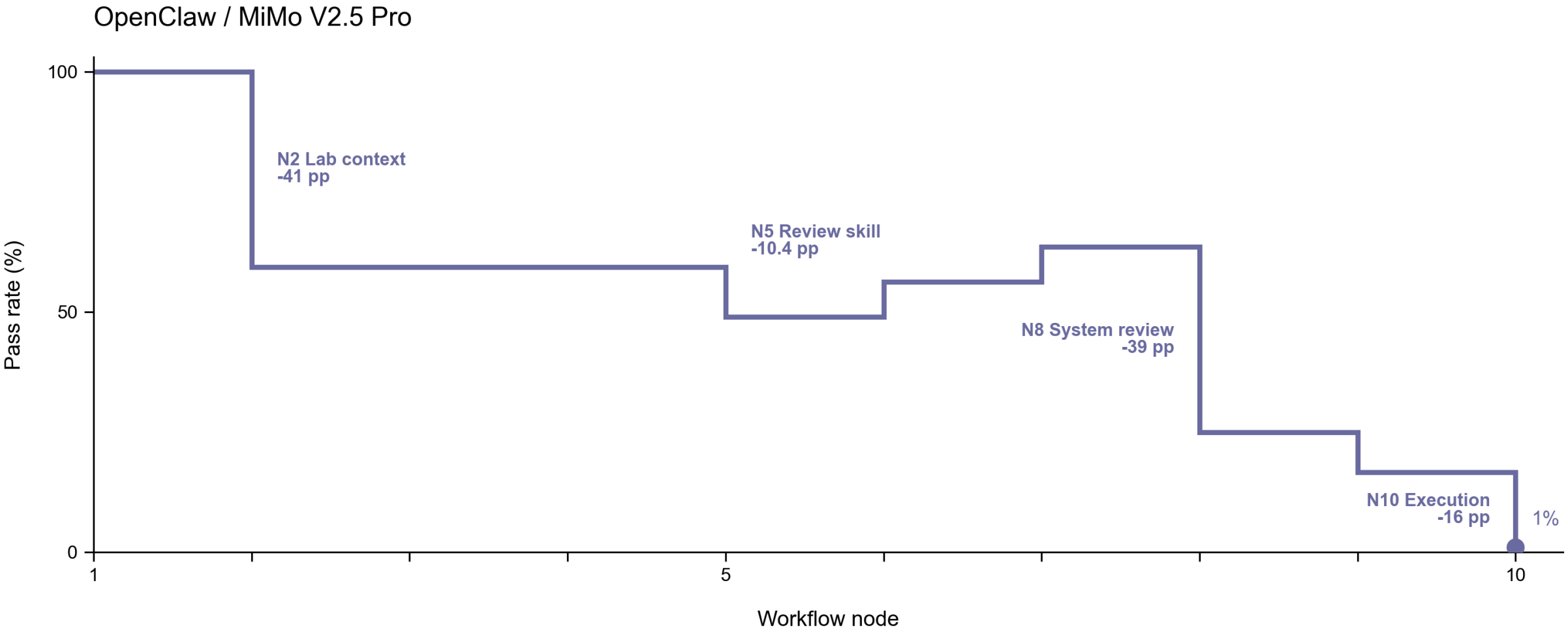

**Fig. S24. Node-specific workflow pass profile for OpenClaw / MiMo V2.5 Pro. Profile summary. Across 96 benchmark trial records, the largest adjacent decline occurred at N2 Lab context (-40.6 pp). The largest recovery occurred at N6 Standard JSON (+7.3 pp). N9 dispatch was 16/96 (16.7%), and N10 execution was 1/96 (1.0%); the executable-endpoint rank was 23 of 48.**

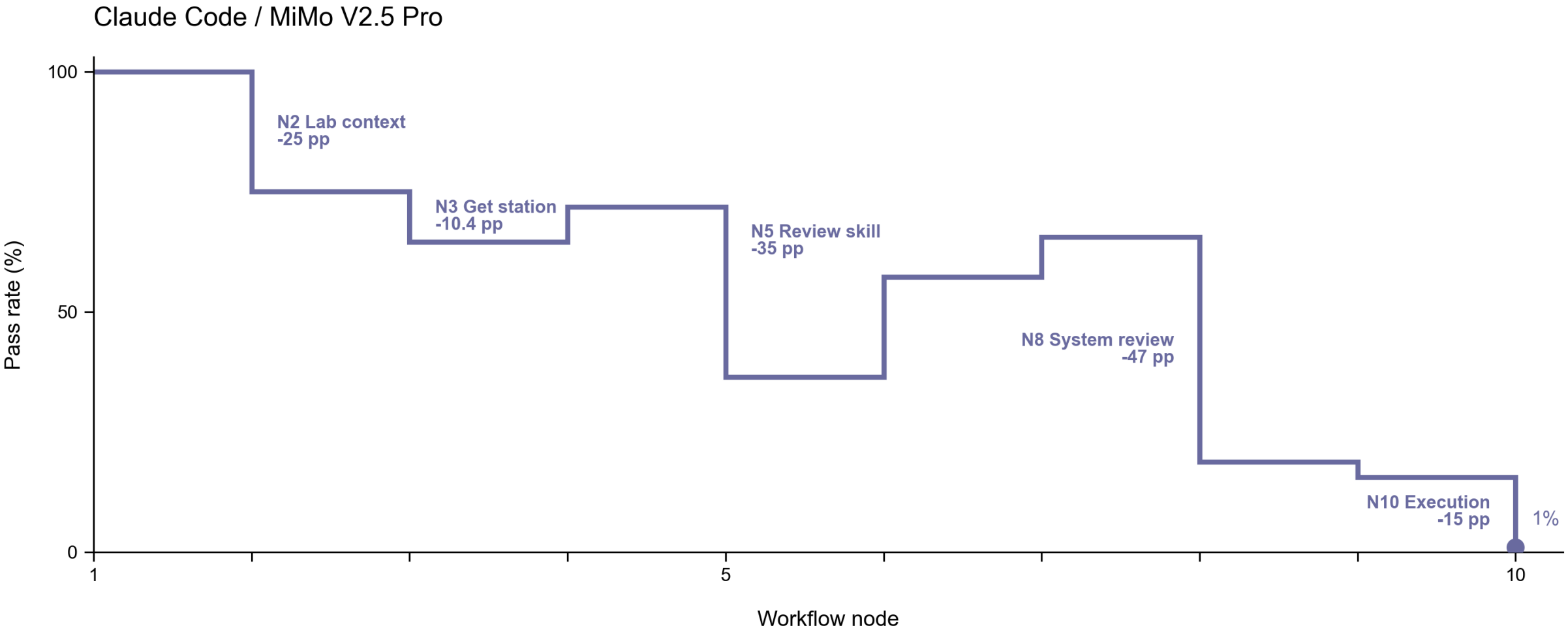

**Fig. S25. Node-specific workflow pass profile for Claude Code / MiMo V2.5 Pro. Profile summary. Across 96 benchmark trial records, the largest adjacent decline occurred at N8 System review (-46.9 pp). The largest recovery occurred at N6 Standard JSON (+20.8 pp). N9 dispatch was 15/96 (15.6%), and N10 execution was 1/96 (1.0%); the executable-endpoint rank was 24 of 48.**

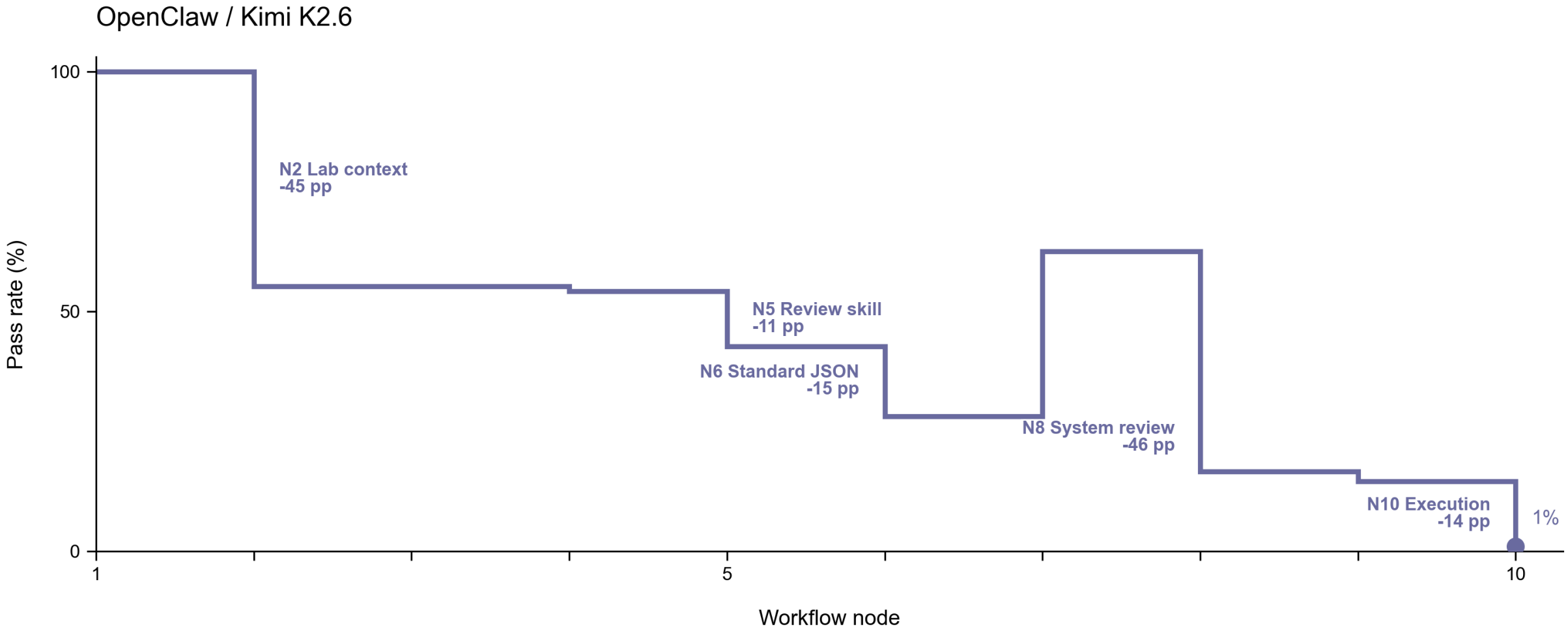

**Fig. S26. Node-specific workflow pass profile for OpenClaw / Kimi K2.6. Profile summary. Across 96 benchmark trial records, the largest adjacent decline occurred at N8 System review (-45.8 pp). The largest recovery occurred at N7 Workflow entry (+34.4 pp). N9 dispatch was 14/96 (14.6%), and N10 execution was 1/96 (1.0%); the executable-endpoint rank was 25 of 48.**

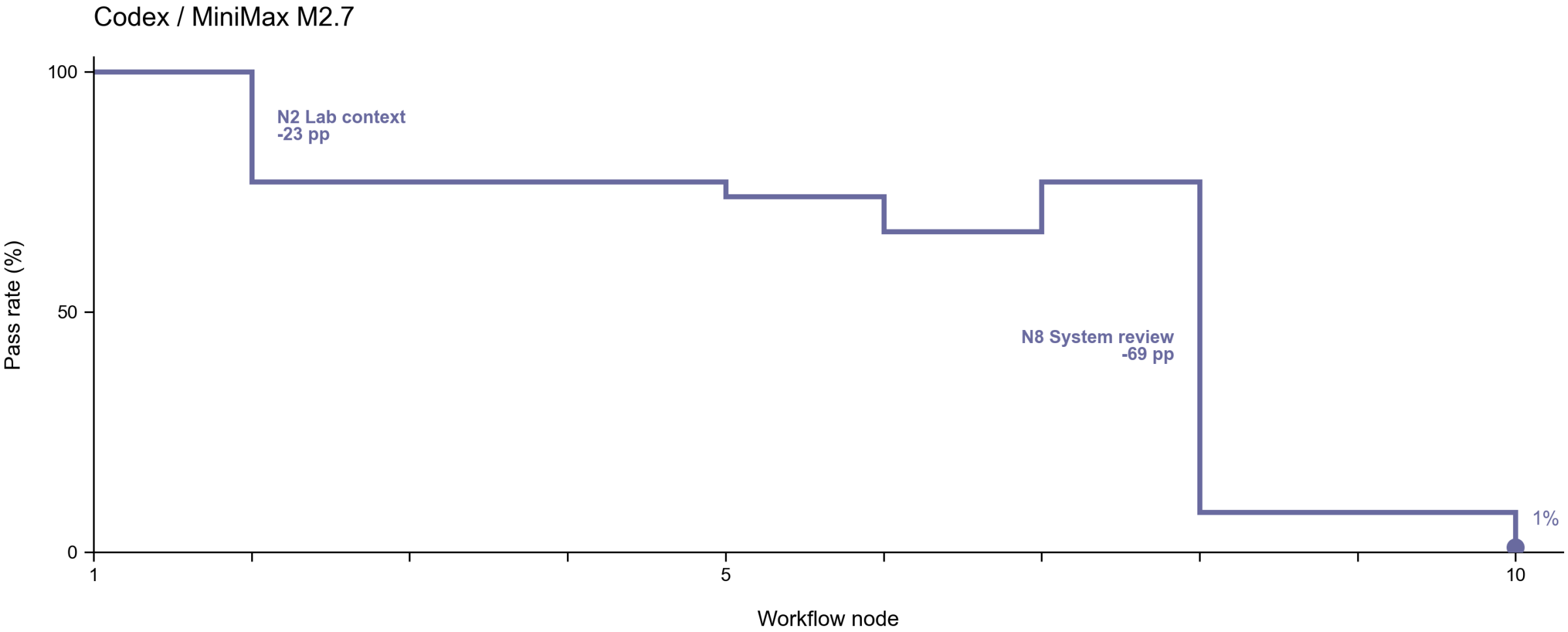

**Fig. S27. Node-specific workflow pass profile for Codex / MiniMax M2.7. Profile summary. Across 96 benchmark trial records, the largest adjacent decline occurred at N8 System review (-68.8 pp). The largest recovery occurred at N7 Workflow entry (+10.4 pp). N9 dispatch was 8/96 (8.3%), and N10 execution was 1/96 (1.0%); the executable-endpoint rank was 26 of 48.**

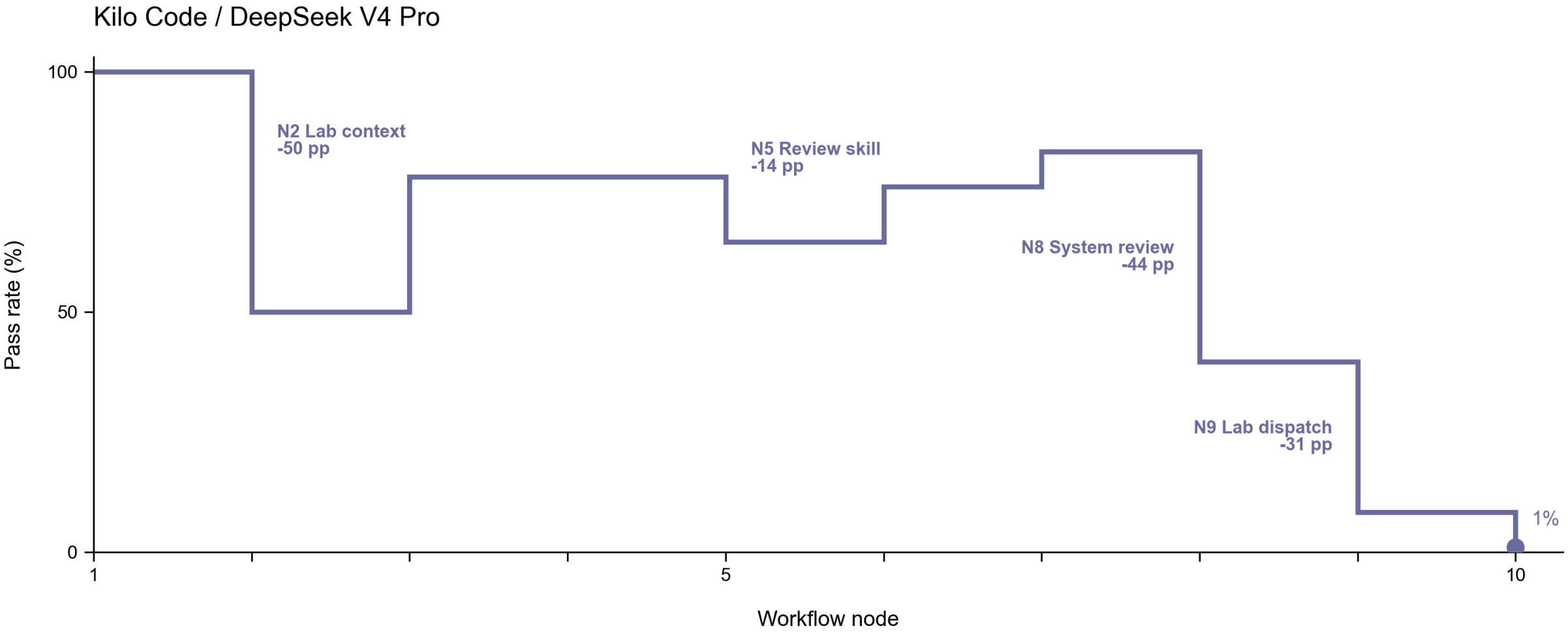

**Fig. S28. Node-specific workflow pass profile for Kilo Code / DeepSeek V4 Pro. Profile summary. Across 96 benchmark trial records, the largest adjacent decline occurred at N2 Lab context (-50.0 pp). The largest recovery occurred at N3 Get station (+28.1 pp). N9 dispatch was 8/96 (8.3%), and N10 execution was 1/96 (1.0%); the executable-endpoint rank was 27 of 48.**

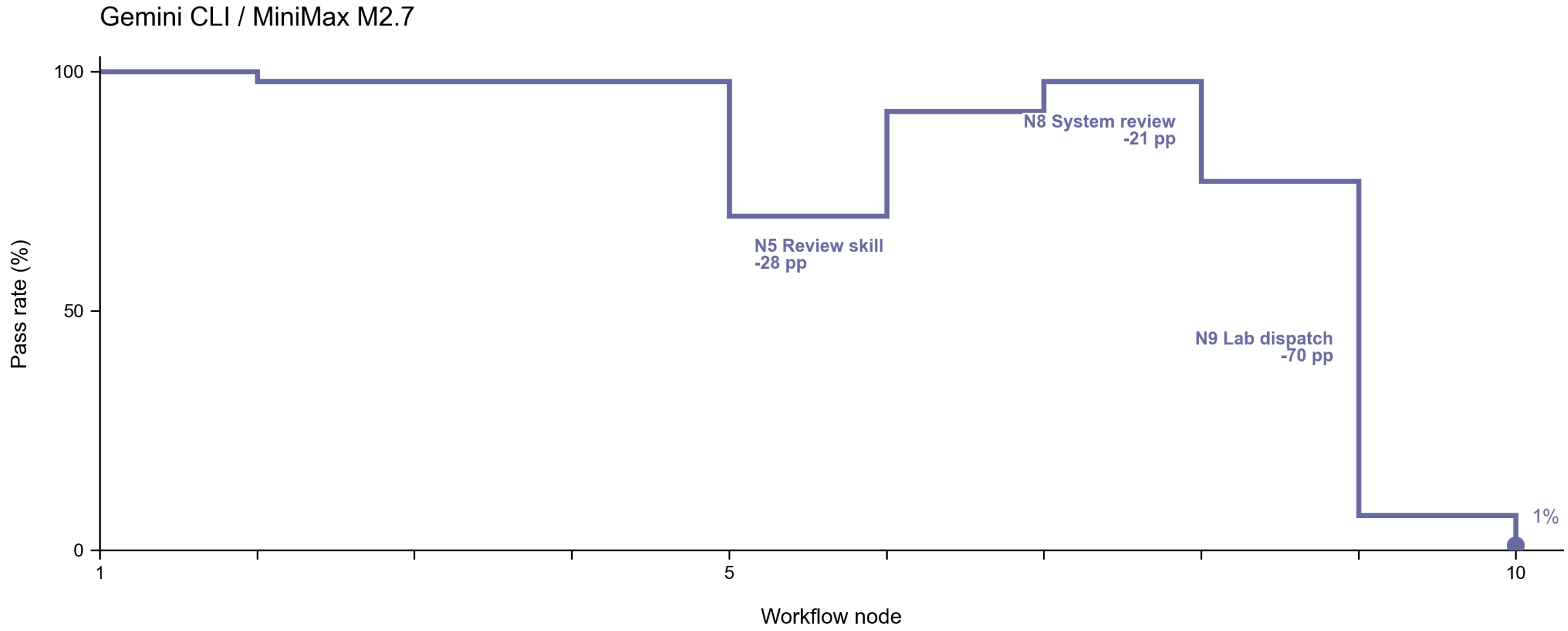

**Fig. S29. Node-specific workflow pass profile for Gemini CLI / MiniMax M2.7. Profile summary. Across 96 benchmark trial records, the largest adjacent decline occurred at N9 Lab dispatch (-69.8 pp). The largest recovery occurred at N6 Standard JSON (+21.9 pp). N9 dispatch was 7/96 (7.3%), and N10 execution was 1/96 (1.0%); the executable-endpoint rank was 28 of 48.**

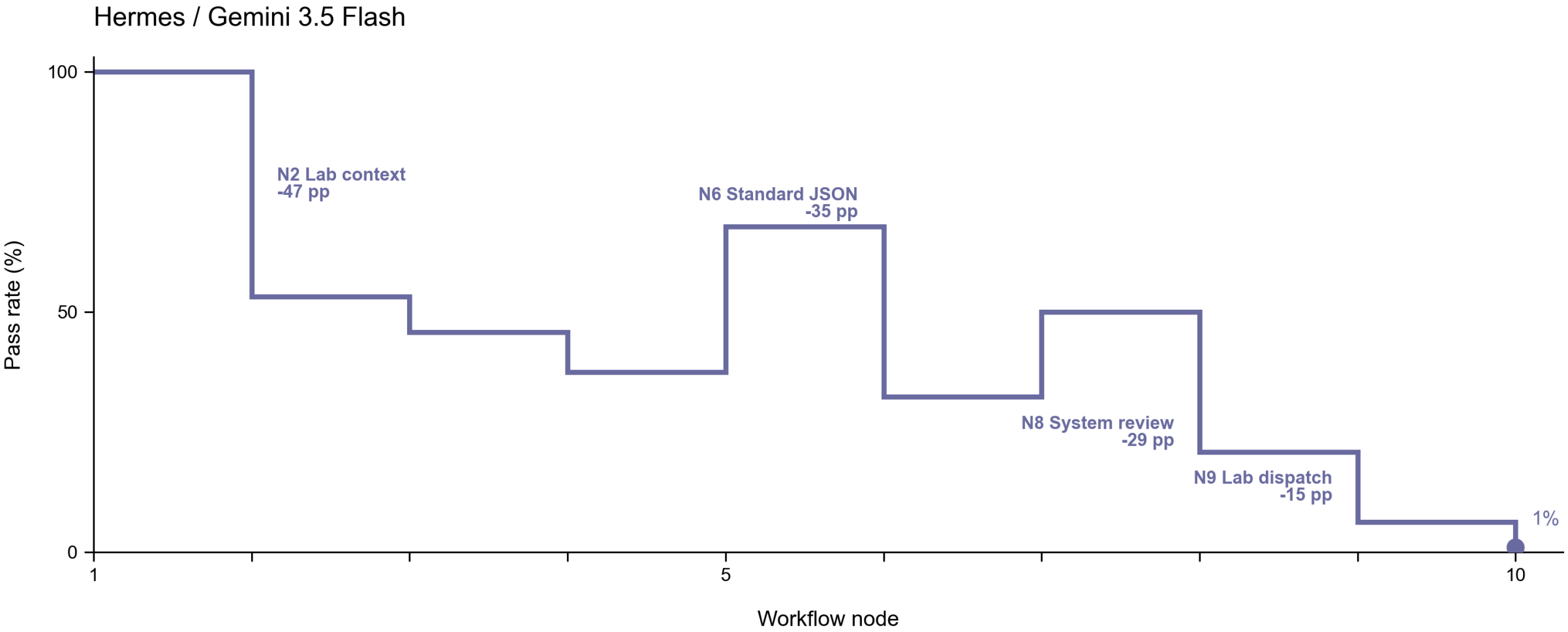

**Fig. S30. Node-specific workflow pass profile for Hermes / Gemini 3.5 Flash. Profile summary. Across 96 benchmark trial records, the largest adjacent decline occurred at N2 Lab context (-46.9 pp). The largest recovery occurred at N5 Review skill (+30.2 pp). N9 dispatch was 6/96 (6.2%), and N10 execution was 1/96 (1.0%); the executable-endpoint rank was 29 of 48.**

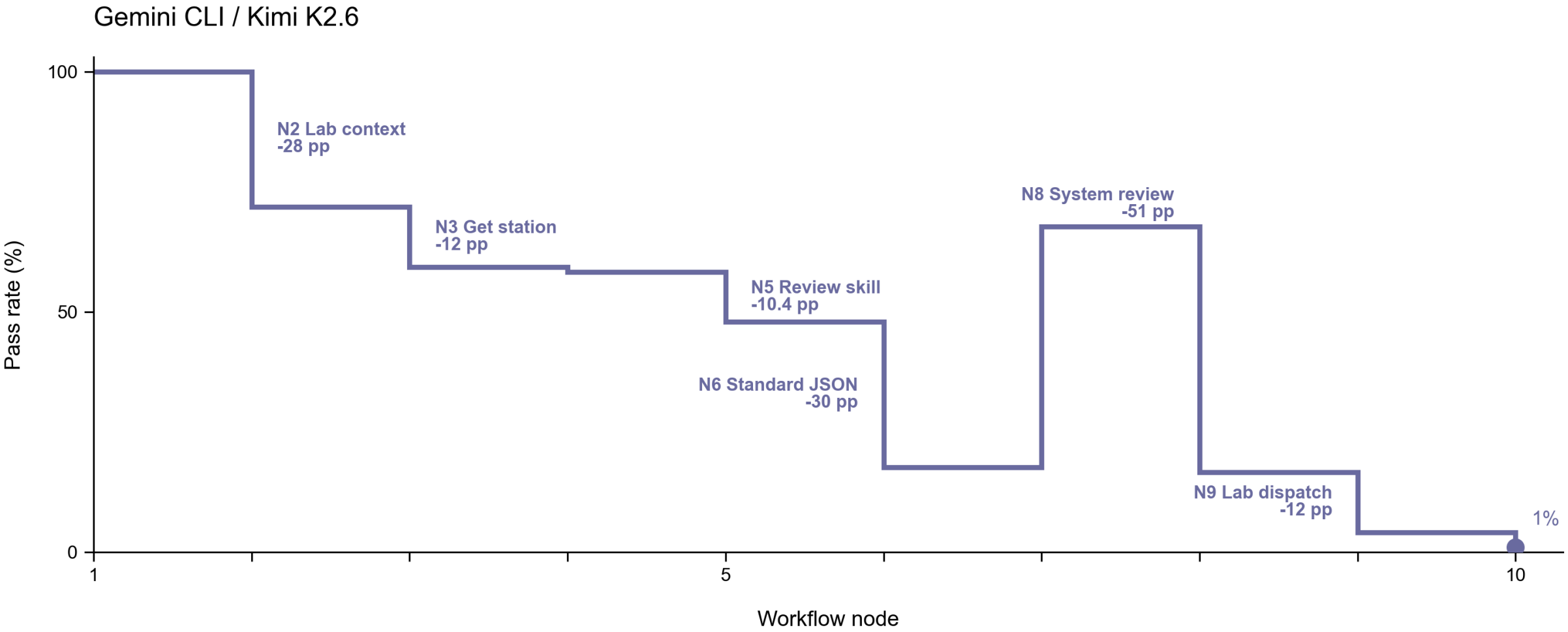

**Fig. S31. Node-specific workflow pass profile for Gemini CLI / Kimi K2.6. Profile summary. Across 96 benchmark trial records, the largest adjacent decline occurred at N8 System review (-51.0 pp). The largest recovery occurred at N7 Workflow entry (+50.0 pp). N9 dispatch was 4/96 (4.2%), and N10 execution was 1/96 (1.0%); the executable-endpoint rank was 30 of 48.**

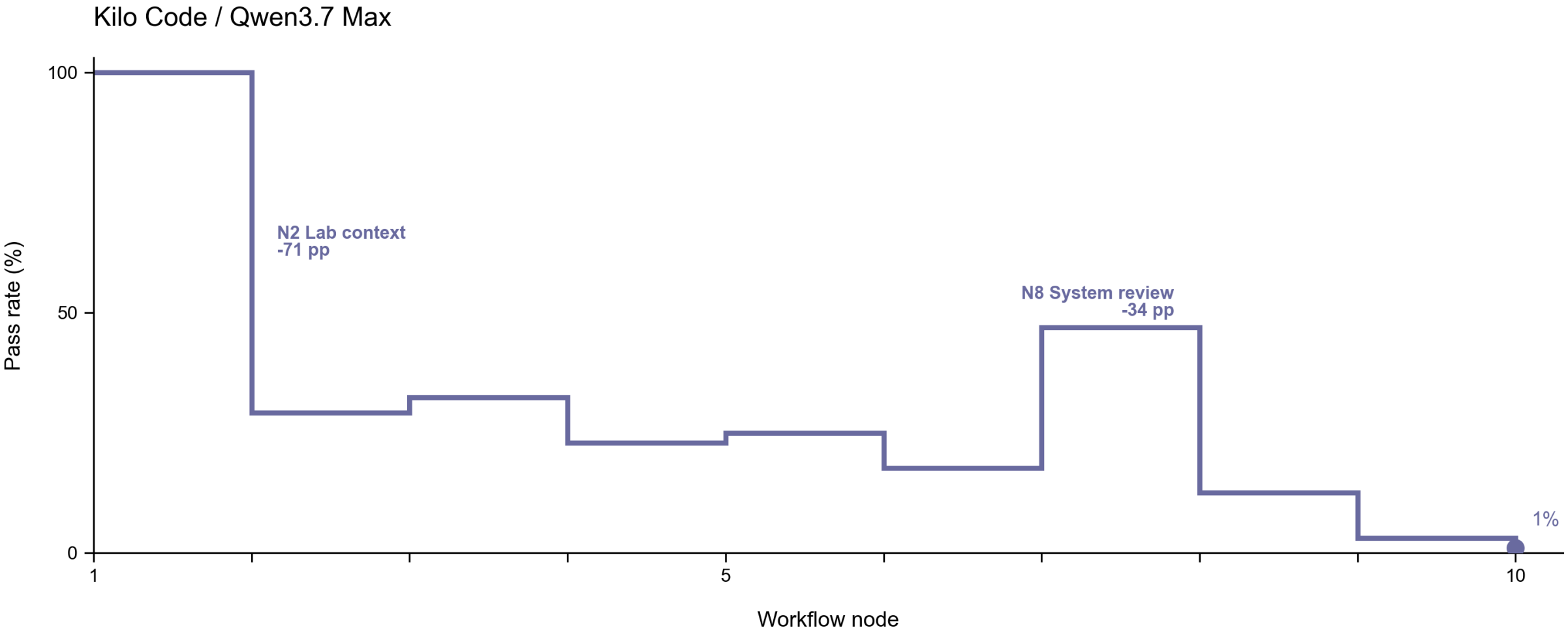

**Fig. S32. Node-specific workflow pass profile for Kilo Code / Qwen3.7 Max. Profile summary. Across 96 benchmark trial records, the largest adjacent decline occurred at N2 Lab context (-70.8 pp). The largest recovery occurred at N7 Workflow entry (+29.2 pp). N9 dispatch was 3/96 (3.1%), and N10 execution was 1/96 (1.0%); the executable-endpoint rank was 31 of 48.**

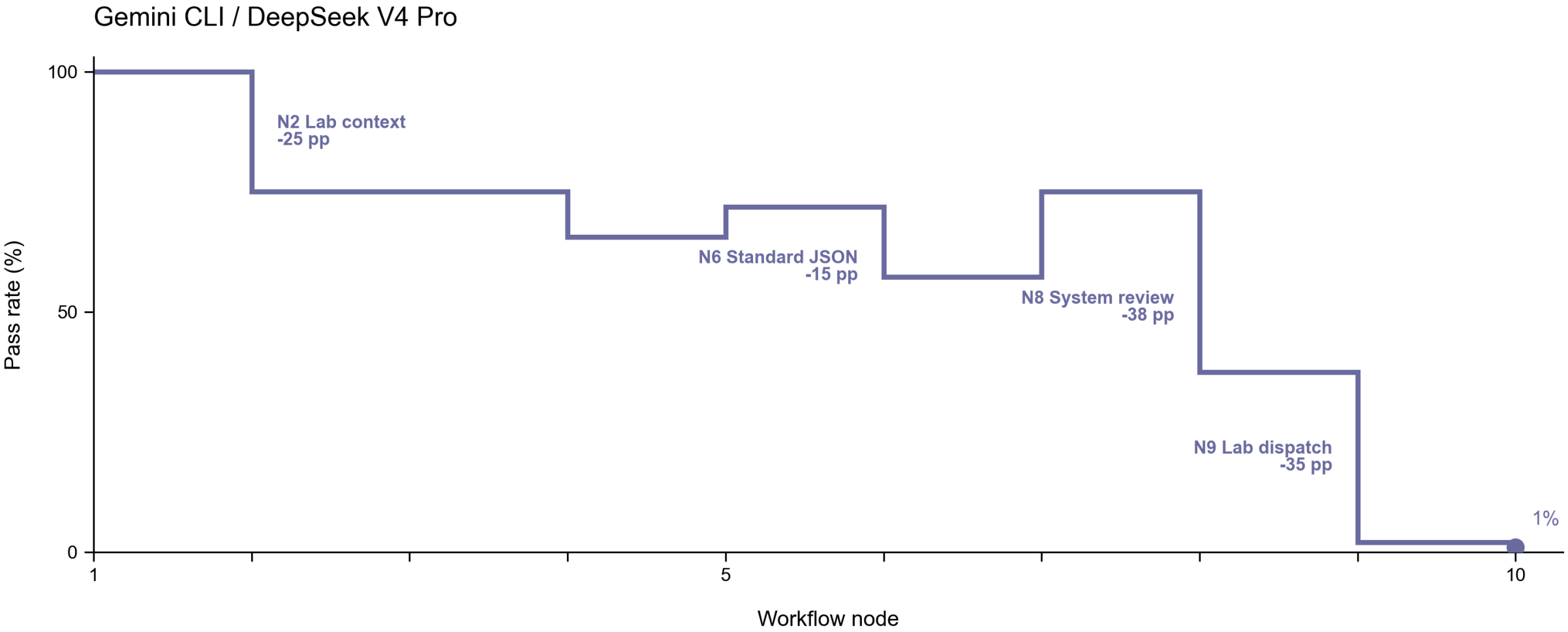

**Fig. S33. Node-specific workflow pass profile for Gemini CLI / DeepSeek V4 Pro. Profile summary. Across 96 benchmark trial records, the largest adjacent decline occurred at N8 System review (-37.5 pp). The largest recovery occurred at N7 Workflow entry (+17.7 pp). N9 dispatch was 2/96 (2.1%), and N10 execution was 1/96 (1.0%); the executable-endpoint rank was 32 of 48.**

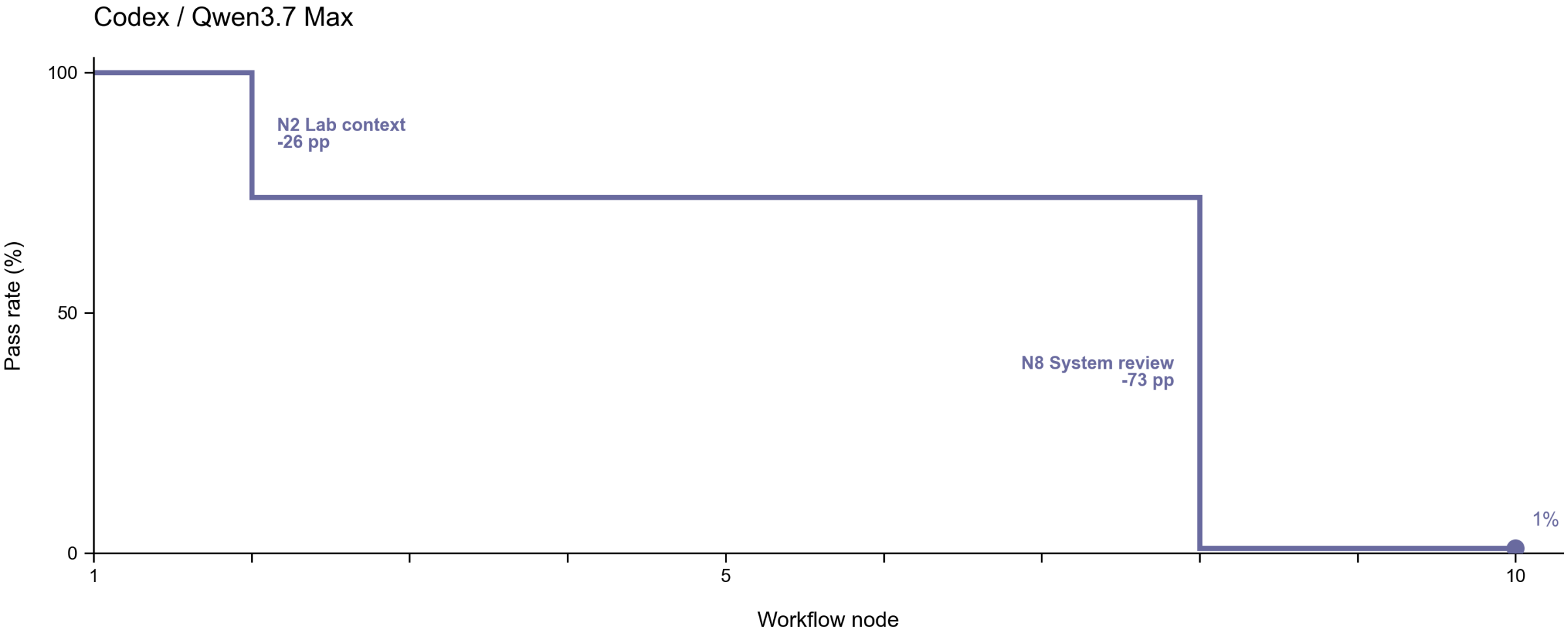

**Fig. S34. Node-specific workflow pass profile for Codex / Qwen3.7 Max. Profile summary. Across 96 benchmark trial records, the largest adjacent decline occurred at N8 System review (-72.9 pp). N9 dispatch was 1/96 (1.0%), and N10 execution was 1/96 (1.0%); the executable-endpoint rank was 33 of 48.**

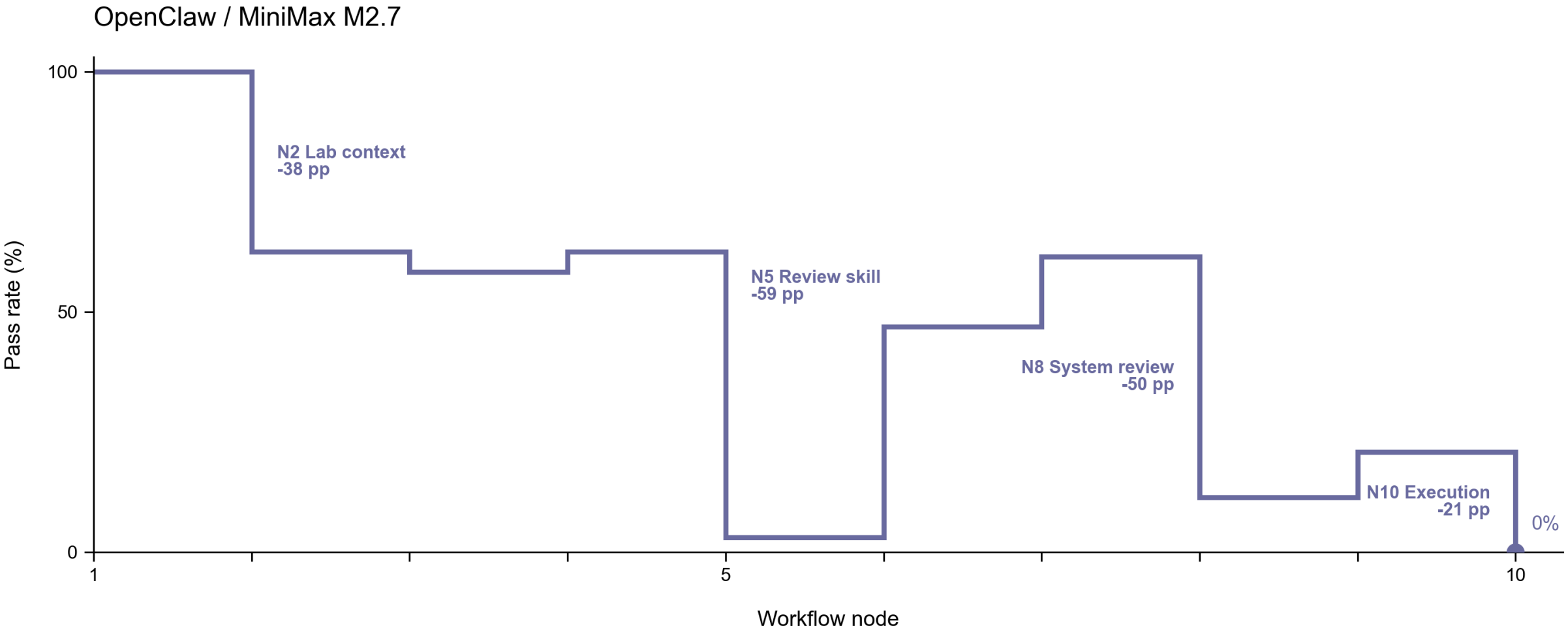

**Fig. S35. Node-specific workflow pass profile for OpenClaw / MiniMax M2.7. Profile summary. Across 96 benchmark trial records, the largest adjacent decline occurred at N5 Review skill (-59.4 pp). The largest recovery occurred at N6 Standard JSON (+43.8 pp). N9 dispatch was 20/96 (20.8%), and N10 execution was 0/96 (0.0%); the executable-endpoint rank was 34 of 48.**

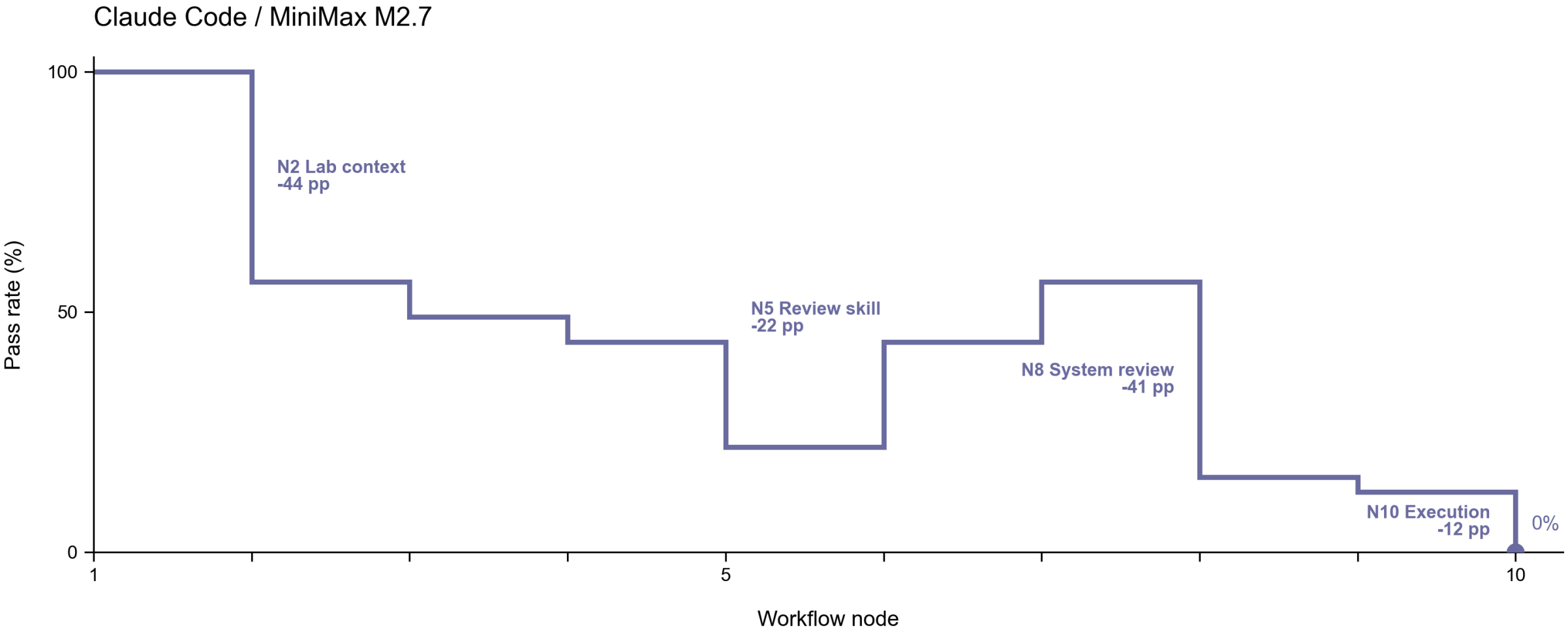

**Fig. S36. Node-specific workflow pass profile for Claude Code / MiniMax M2.7. Profile summary. Across 96 benchmark trial records, the largest adjacent decline occurred at N2 Lab context (-43.8 pp). The largest recovery occurred at N6 Standard JSON (+21.9 pp). N9 dispatch was 12/96 (12.5%), and N10 execution was 0/96 (0.0%); the executable-endpoint rank was 35 of 48.**

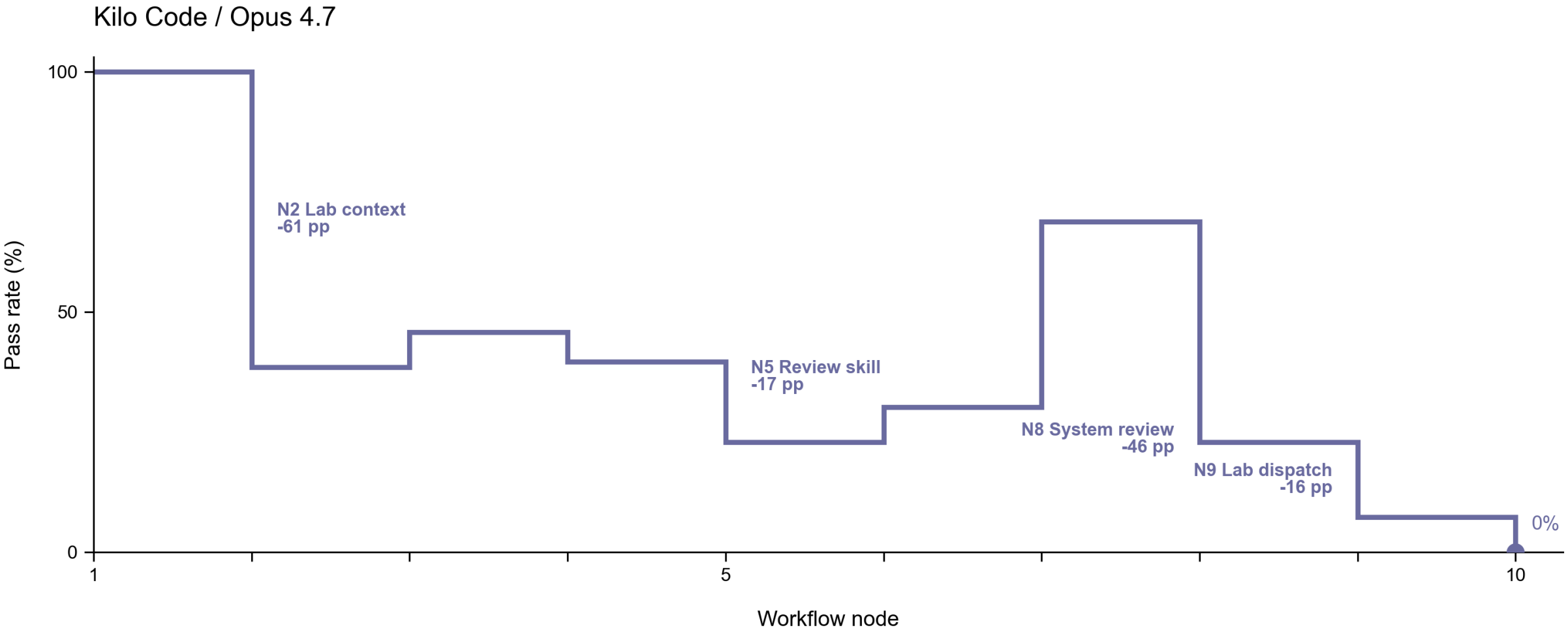

**Fig. S37. Node-specific workflow pass profile for Kilo Code / Opus 4.7. Profile summary. Across 96 benchmark trial records, the largest adjacent decline occurred at N2 Lab context (-61.5 pp). The largest recovery occurred at N7 Workflow entry (+38.5 pp). N9 dispatch was 7/96 (7.3%), and N10 execution was 0/96 (0.0%); the executable-endpoint rank was 36 of 48.**

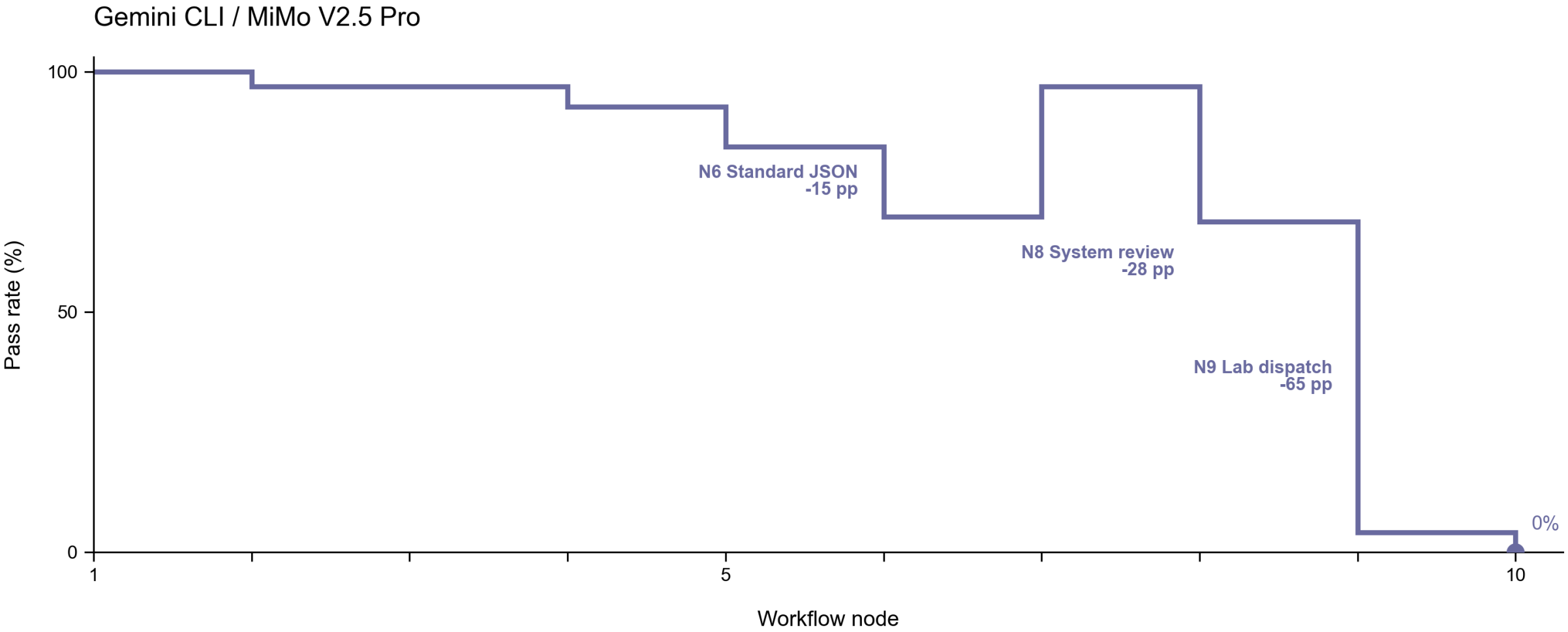

**Fig. S38. Node-specific workflow pass profile for Gemini CLI / MiMo V2.5 Pro. Profile summary. Across 96 benchmark trial records, the largest adjacent decline occurred at N9 Lab dispatch (-64.6 pp). The largest recovery occurred at N7 Workflow entry (+27.1 pp). N9 dispatch was 4/96 (4.2%), and N10 execution was 0/96 (0.0%); the executable-endpoint rank was 37 of 48.**

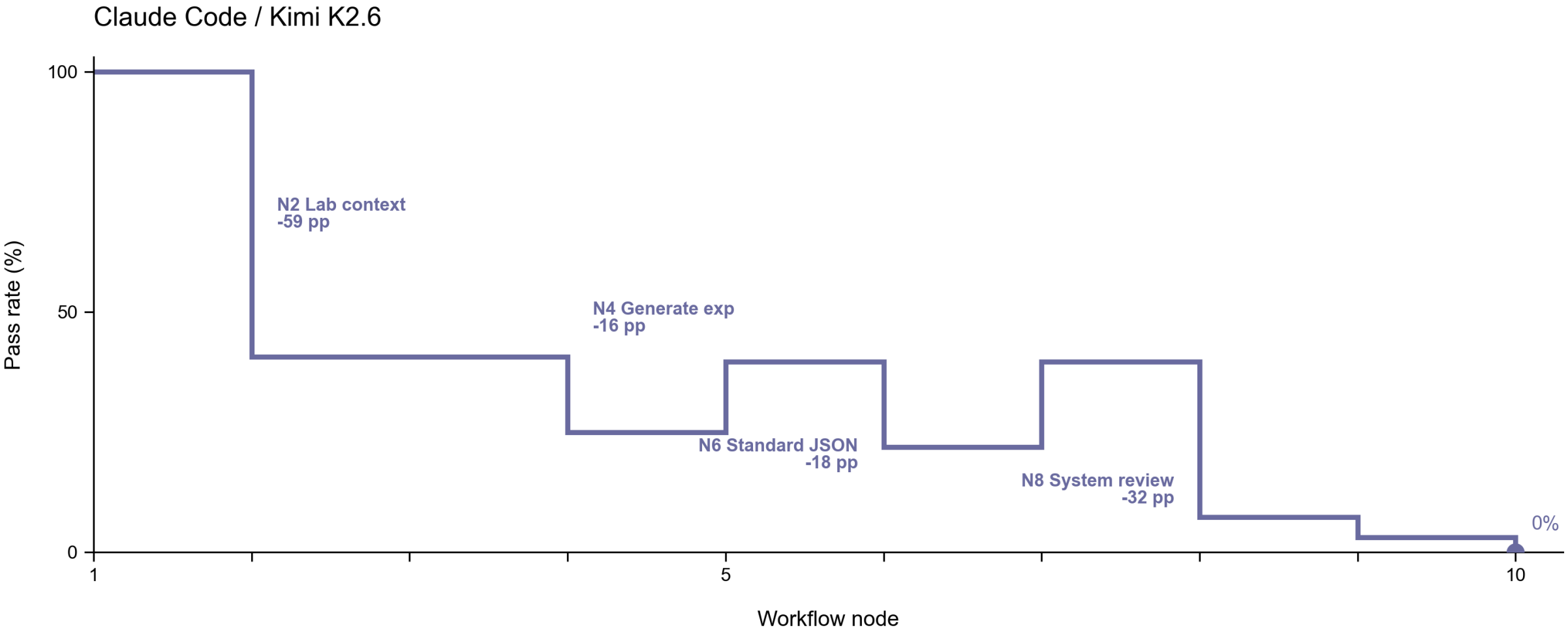

**Fig. S39. Node-specific workflow pass profile for Claude Code / Kimi K2.6. Profile summary. Across 96 benchmark trial records, the largest adjacent decline occurred at N2 Lab context (-59.4 pp). The largest recovery occurred at N7 Workflow entry (+17.7 pp). N9 dispatch was 3/96 (3.1%), and N10 execution was 0/96 (0.0%); the executable-endpoint rank was 38 of 48.**

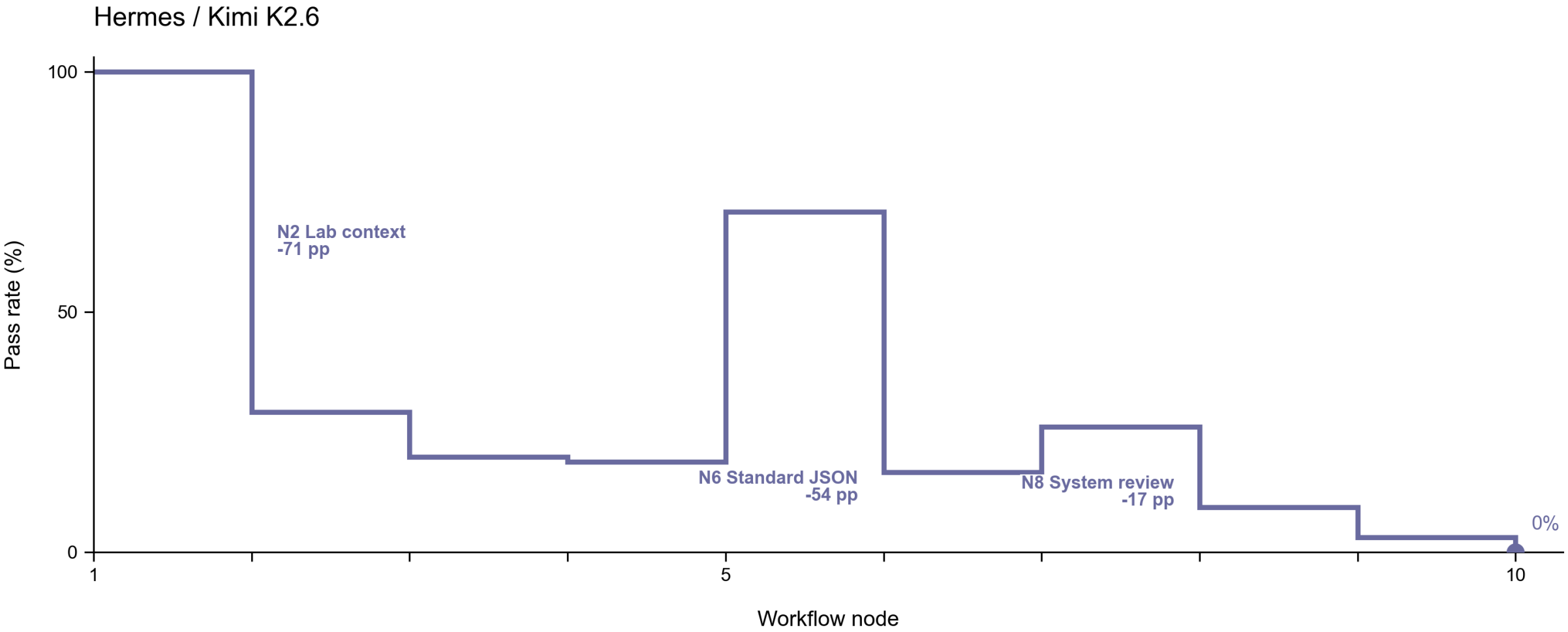

**Fig. S40. Node-specific workflow pass profile for Hermes / Kimi K2.6. Profile summary. Across 96 benchmark trial records, the largest adjacent decline occurred at N2 Lab context (-70.8 pp). The largest recovery occurred at N5 Review skill (+52.1 pp). N9 dispatch was 3/96 (3.1%), and N10 execution was 0/96 (0.0%); the executable-endpoint rank was 39 of 48.**

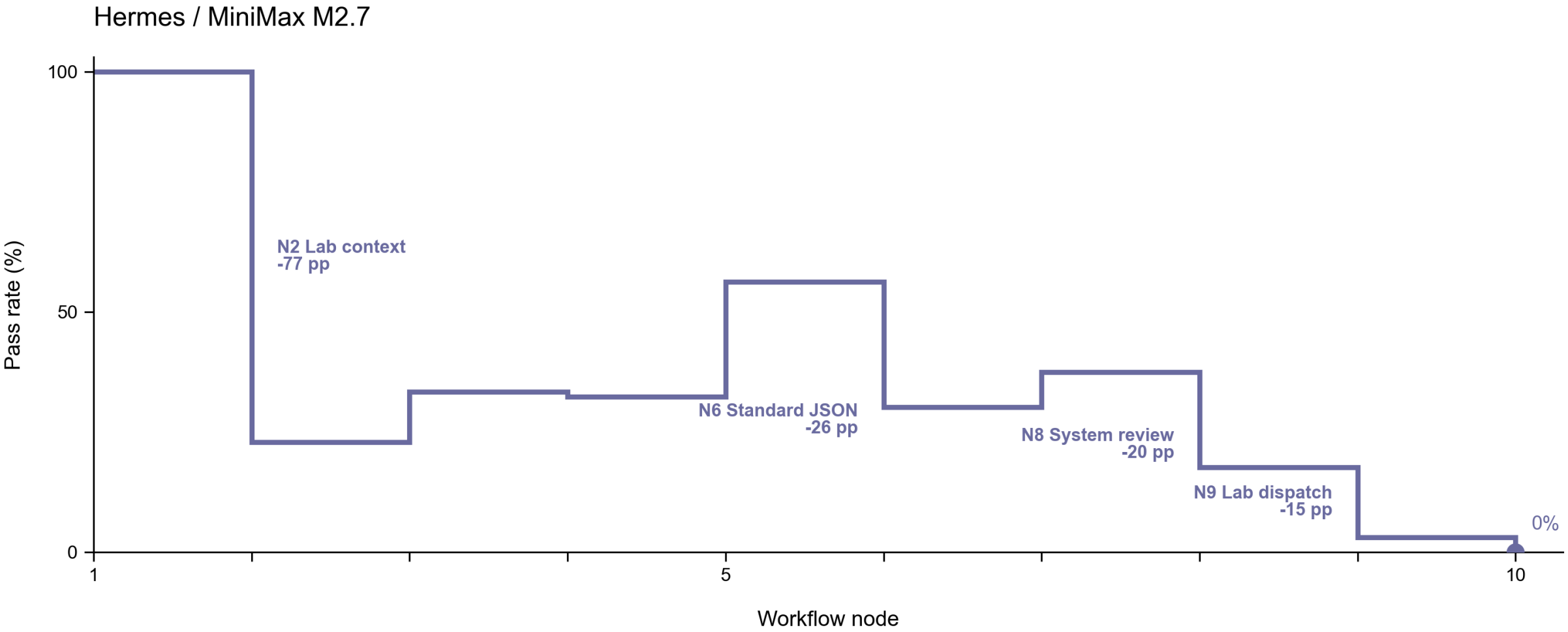

**Fig. S41. Node-specific workflow pass profile for Hermes / MiniMax M2.7. Profile summary. Across 96 benchmark trial records, the largest adjacent decline occurred at N2 Lab context (-77.1 pp). The largest recovery occurred at N5 Review skill (+24.0 pp). N9 dispatch was 3/96 (3.1%), and N10 execution was 0/96 (0.0%); the executable-endpoint rank was 40 of 48.**

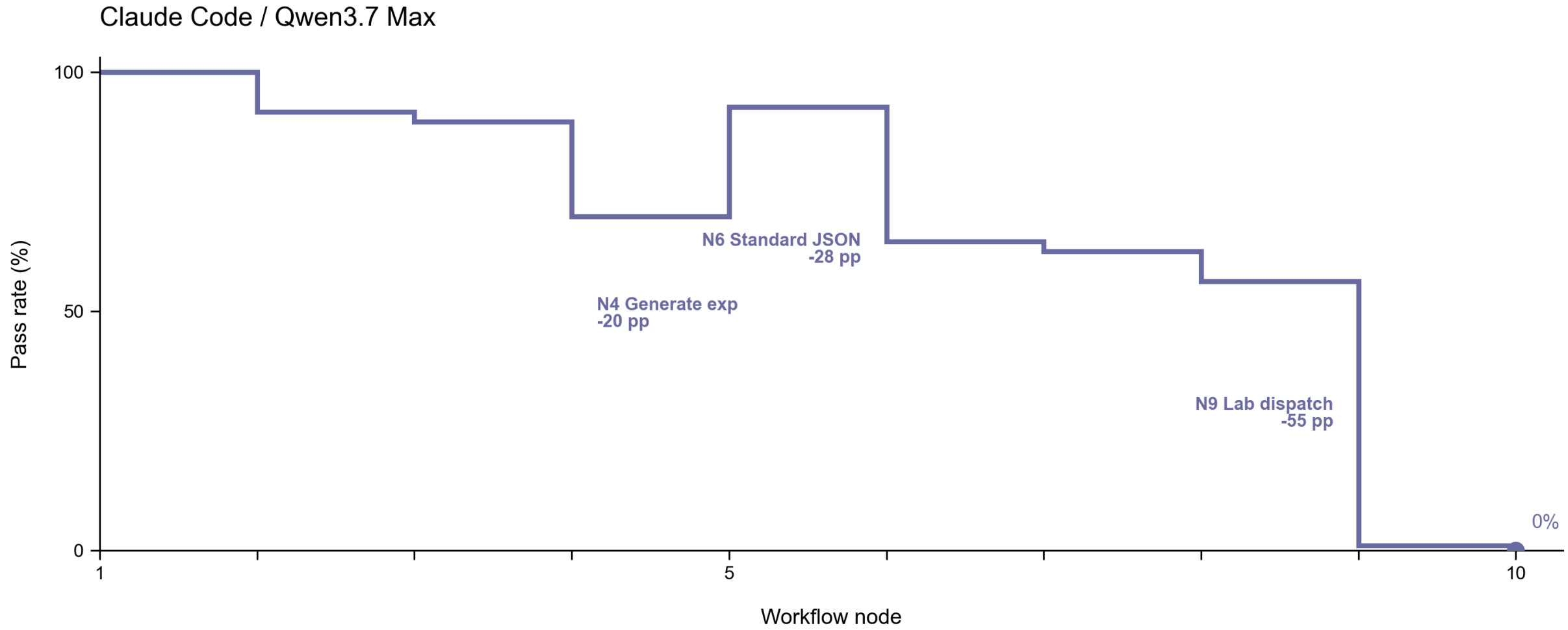

**Fig. S42. Node-specific workflow pass profile for Claude Code / Qwen3.7 Max. Profile summary. Across 96 benchmark trial records, the largest adjacent decline occurred at N9 Lab dispatch (-55.2 pp). The largest recovery occurred at N5 Review skill (+22.9 pp). N9 dispatch was 1/96 (1.0%), and N10 execution was 0/96 (0.0%); the executable-endpoint rank was 41 of 48.**

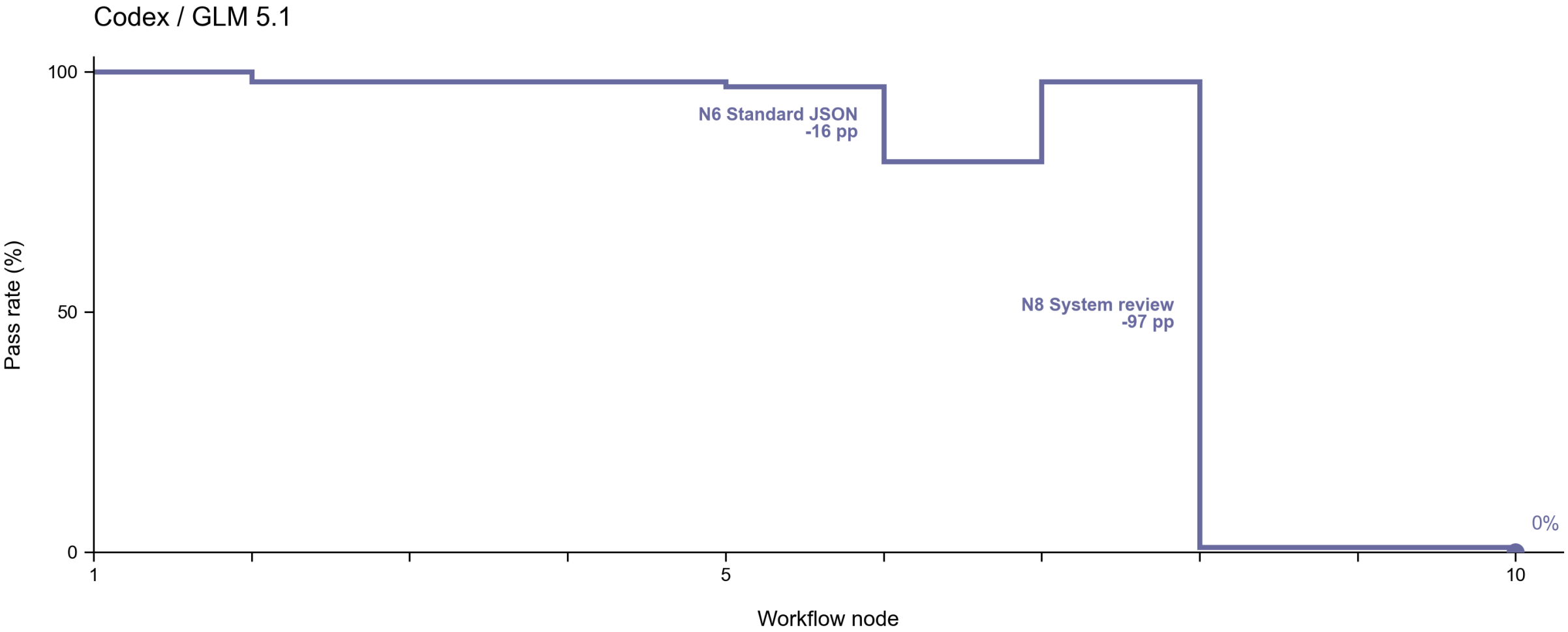

**Fig. S43. Node-specific workflow pass profile for Codex / GLM 5.1. Profile summary. Across 96 benchmark trial records, the largest adjacent decline occurred at N8 System review (-96.9 pp). The largest recovery occurred at N7 Workflow entry (+16.7 pp). N9 dispatch was 1/96 (1.0%), and N10 execution was 0/96 (0.0%); the executable-endpoint rank was 42 of 48.**

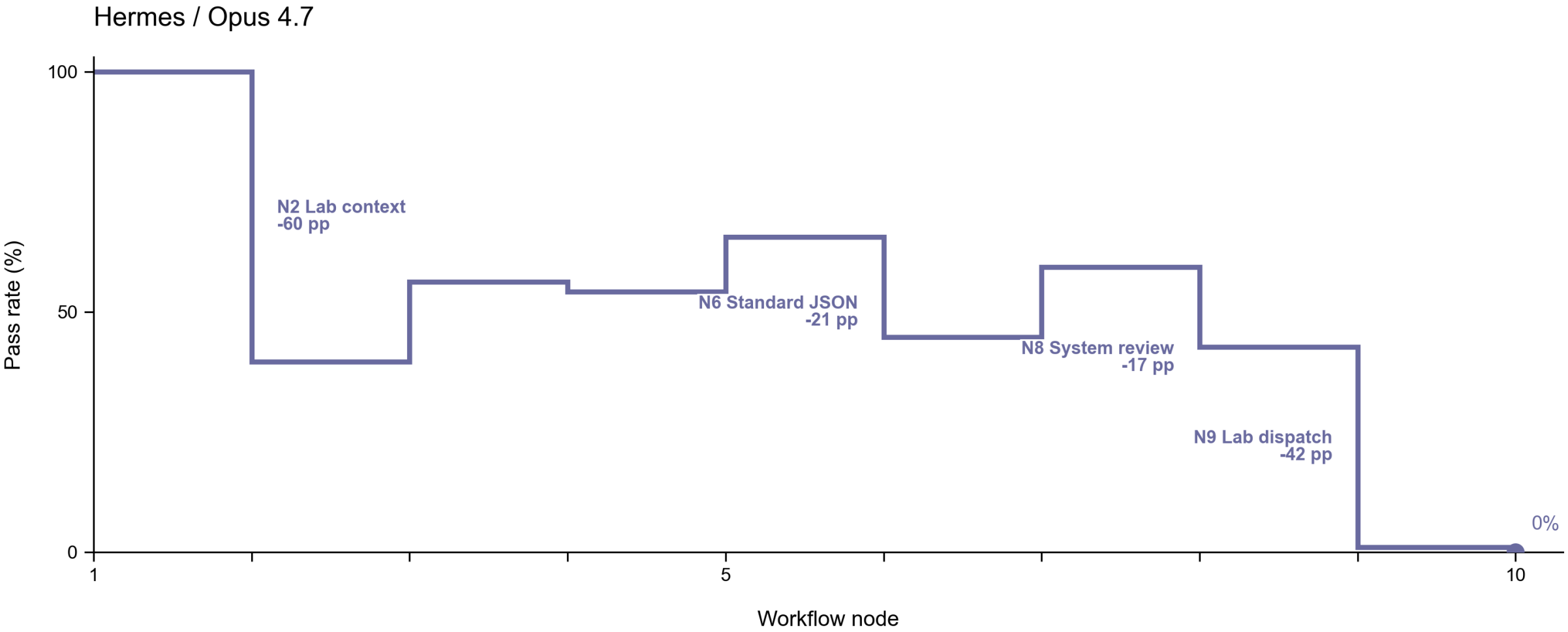

**Fig. S44. Node-specific workflow pass profile for Hermes / Opus 4.7. Profile summary. Across 96 benchmark trial records, the largest adjacent decline occurred at N2 Lab context (-60.4 pp). The largest recovery occurred at N3 Get station (+16.7 pp). N9 dispatch was 1/96 (1.0%), and N10 execution was 0/96 (0.0%); the executable-endpoint rank was 43 of 48.**

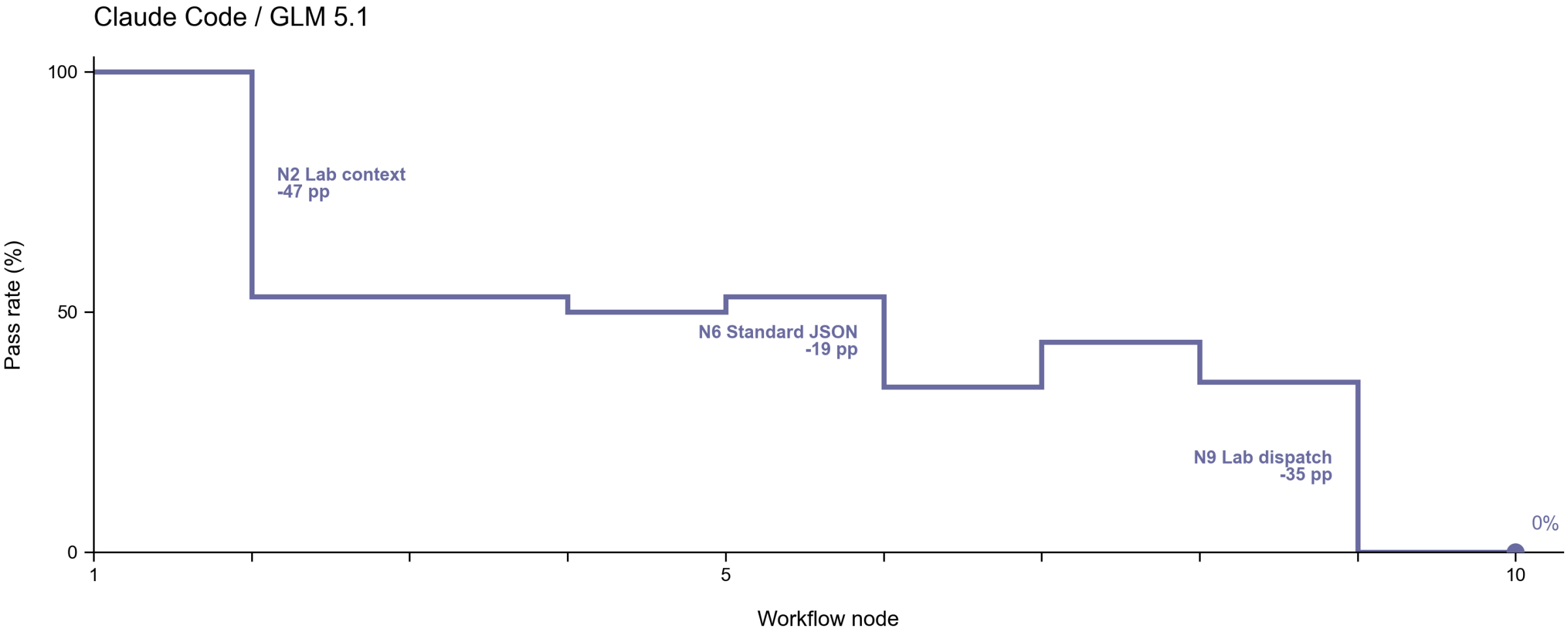

**Fig. S45. Node-specific workflow pass profile for Claude Code / GLM 5.1. Profile summary. Across 96 benchmark trial records, the largest adjacent decline occurred at N2 Lab context (-46.9 pp). The largest recovery occurred at N7 Workflow entry (+9.4 pp). N9 dispatch was 0/96 (0.0%), and N10 execution was 0/96 (0.0%); the executable-endpoint rank was 44 of 48.**

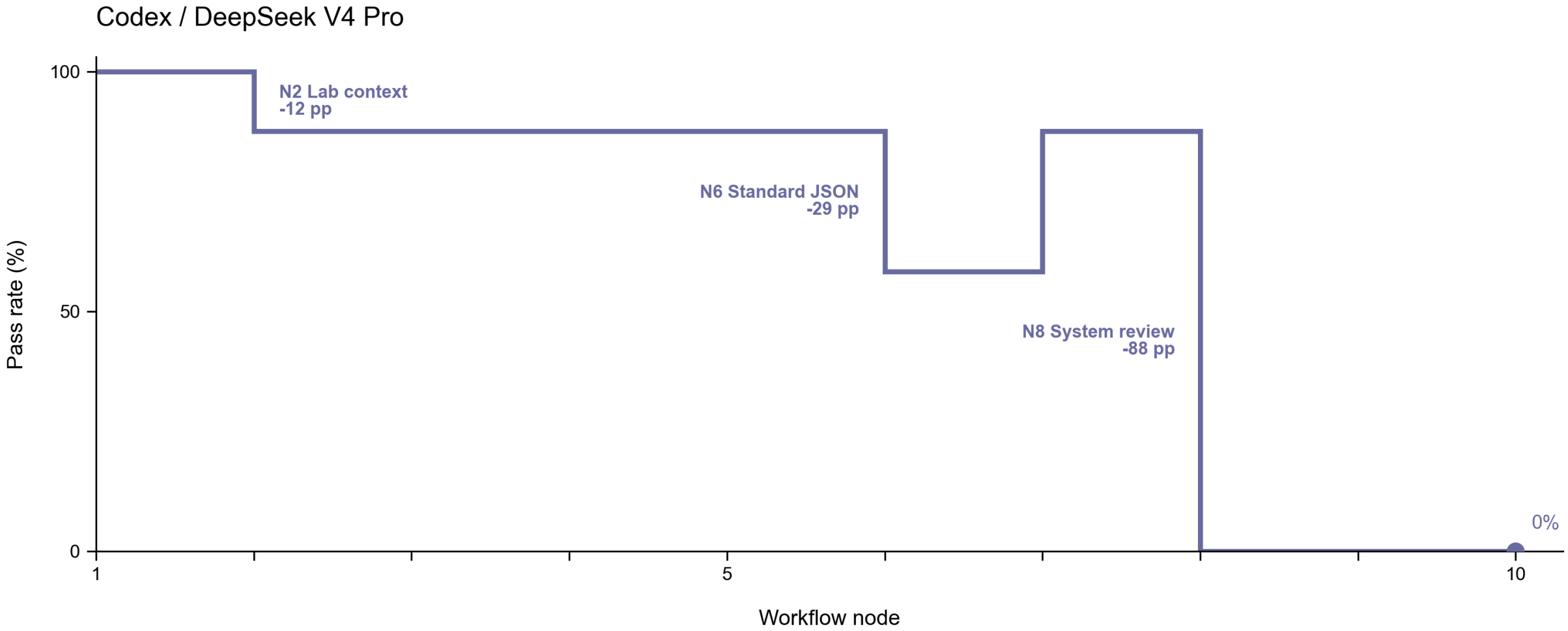

**Fig. S46. Node-specific workflow pass profile for Codex / DeepSeek V4 Pro. Profile summary. Across 96 benchmark trial records, the largest adjacent decline occurred at N8 System review (-87.5 pp). The largest recovery occurred at N7 Workflow entry (+29.2 pp). N9 dispatch was 0/96 (0.0%), and N10 execution was 0/96 (0.0%); the executable-endpoint rank was 45 of 48.**

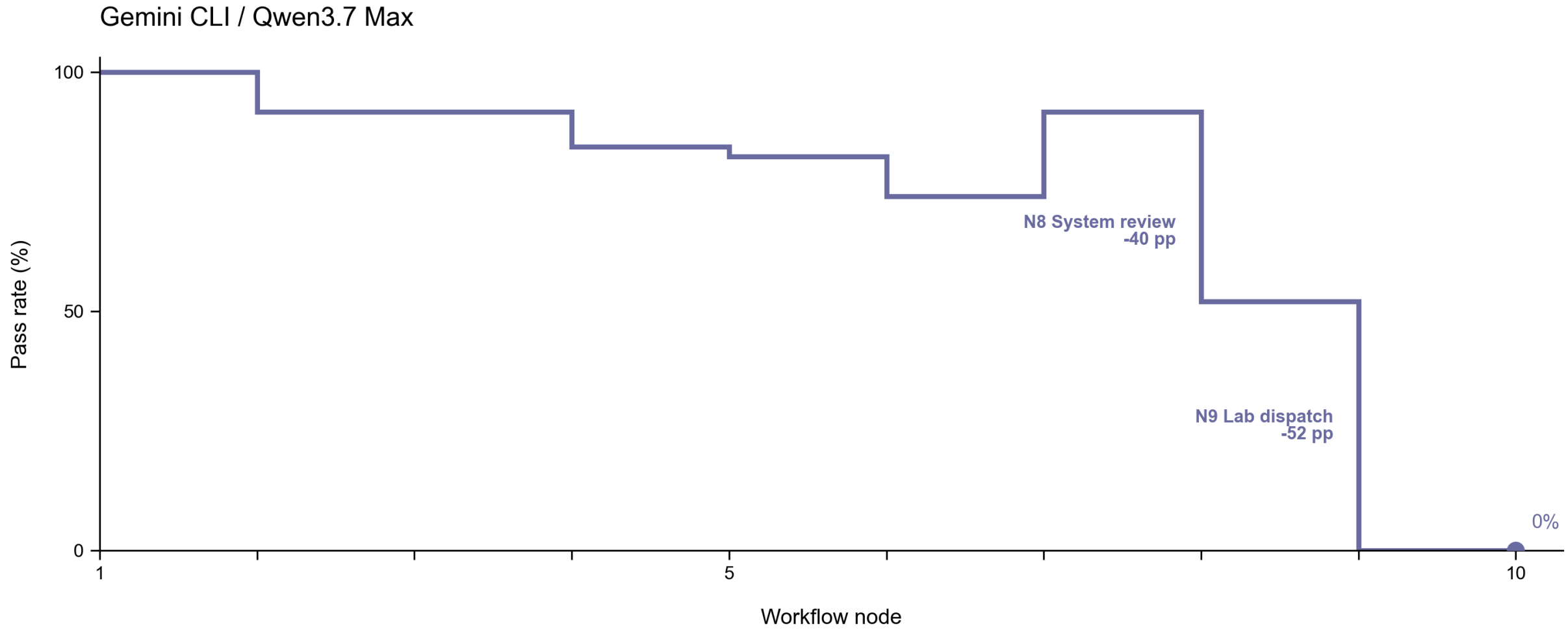

**Fig. S47. Node-specific workflow pass profile for Gemini CLI / Qwen3.7 Max. Profile summary. Across 96 benchmark trial records, the largest adjacent decline occurred at N9 Lab dispatch (-52.1 pp). The largest recovery occurred at N7 Workflow entry (+17.7 pp). N9 dispatch was 0/96 (0.0%), and N10 execution was 0/96 (0.0%); the executable-endpoint rank was 46 of 48.**

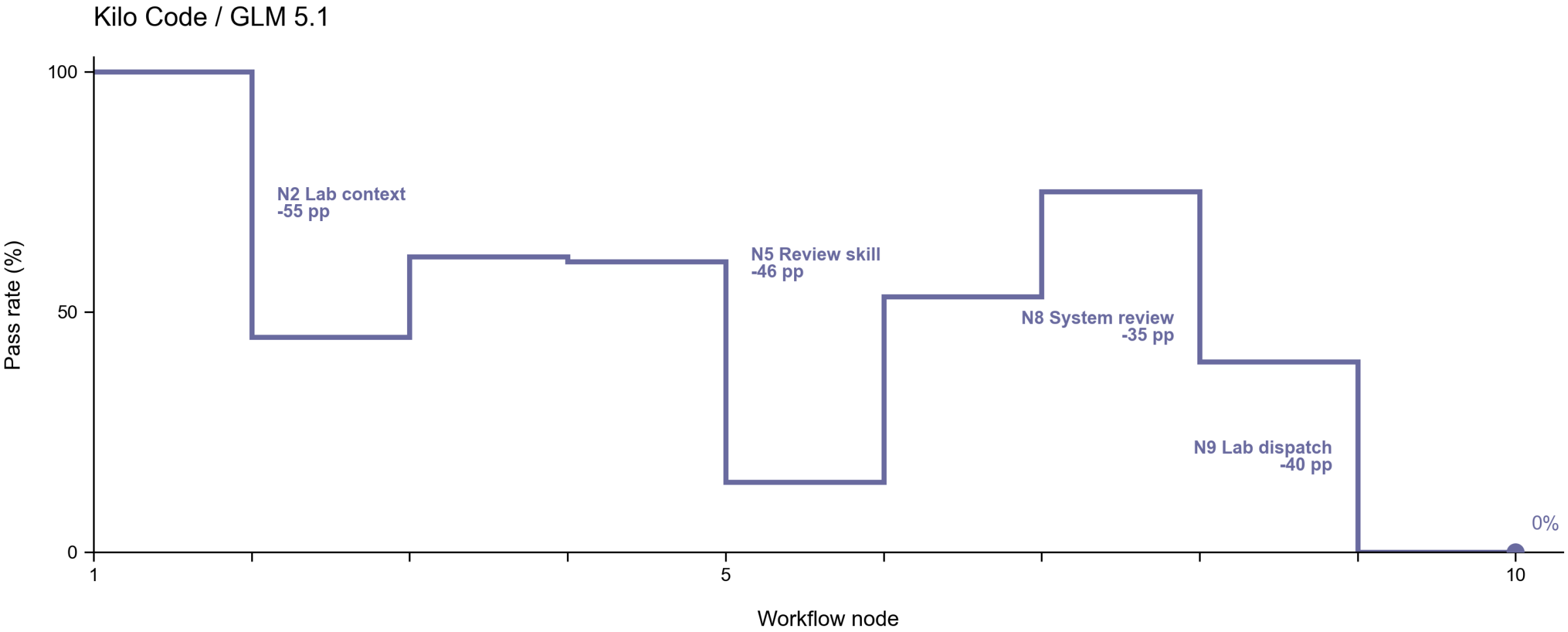

**Fig. S48. Node-specific workflow pass profile for Kilo Code / GLM 5.1. Profile summary. Across 96 benchmark trial records, the largest adjacent decline occurred at N2 Lab context (-55.2 pp). The largest recovery occurred at N6 Standard JSON (+38.5 pp). N9 dispatch was 0/96 (0.0%), and N10 execution was 0/96 (0.0%); the executable-endpoint rank was 47 of 48.**

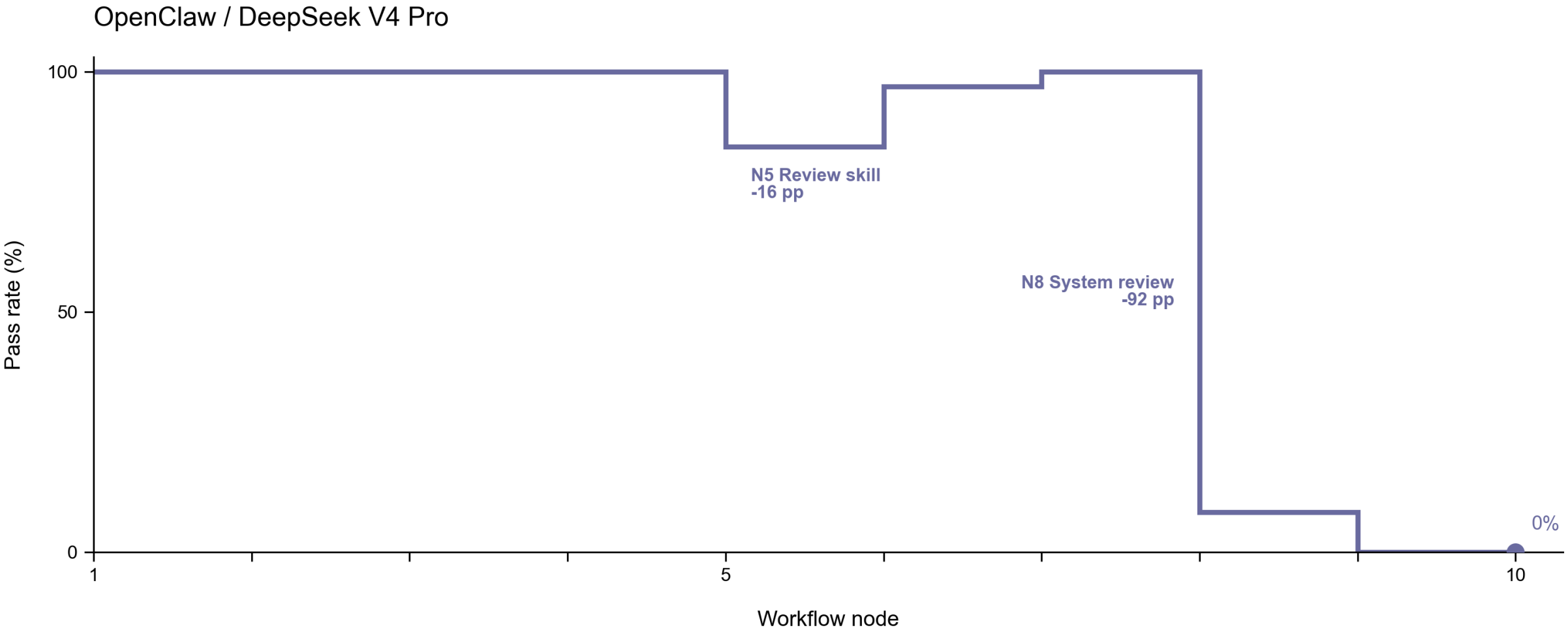

**Fig. S49. Node-specific workflow pass profile for OpenClaw / DeepSeek V4 Pro. Profile summary. Across 96 benchmark trial records, the largest adjacent decline occurred at N8 System review (-91.7 pp). The largest recovery occurred at N6 Standard JSON (+12.5 pp). N9 dispatch was 0/96 (0.0%), and N10 execution was 0/96 (0.0%); the executable-endpoint rank was 48 of 48.**

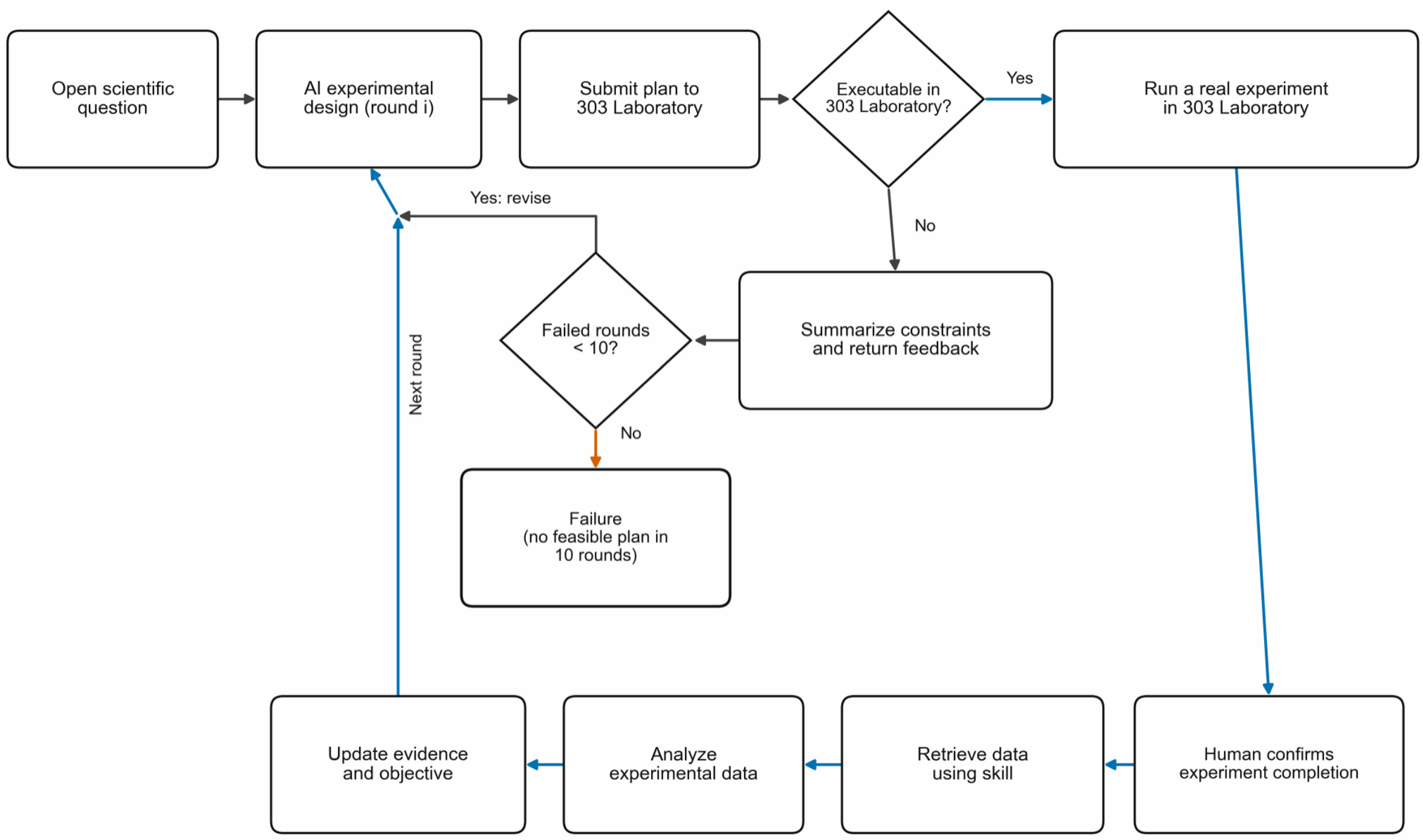

**Fig. S50. Rule-based closed-loop workflow for AI-guided experimentation in the 303 Laboratory. An open scientific question initiates iterative experimental design, and each proposed workflow is evaluated for executability in the 303 Laboratory. Non-executable workflows receive structured constraint feedback for redesign; the task is terminated if no feasible workflow is generated within ten rounds. Executable workflows proceed to real experiments. Following human confirmation of experiment completion, experimental data are retrieved using laboratory skills, analysed and used to update the evidence and objective for the next design round. Blue arrows indicate the experiment–data–redesign loop, whereas the orange arrow indicates terminal failure.**

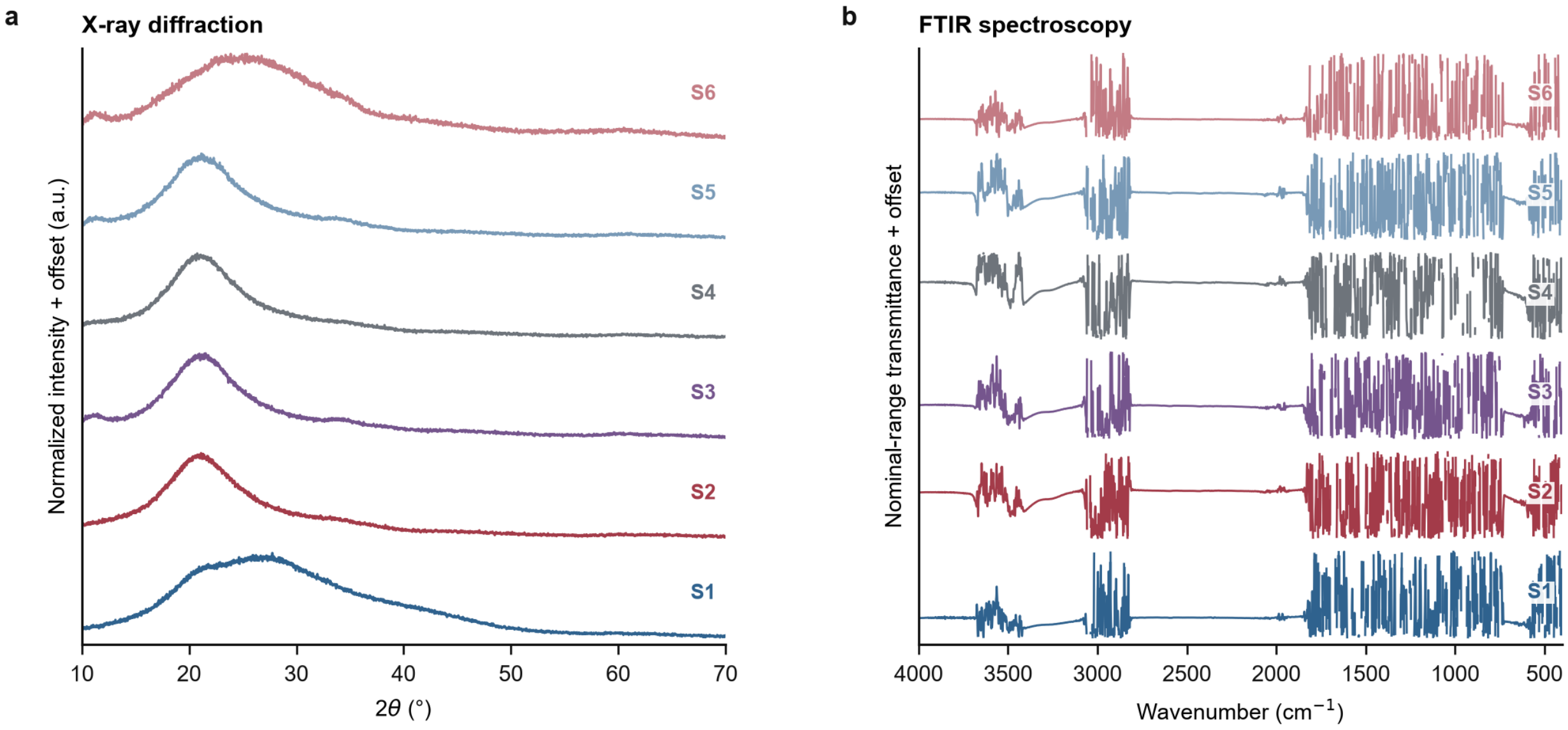

**Fig. S51. Round 1 - Structural and spectroscopic characterization. (a) X-ray diffraction patterns and (b) FTIR spectra for samples S1-S6. XRD traces are minimum-shifted, maximum-normalized and vertically offset. FTIR shows only points within the nominal 0-100% transmittance range; out-of-range values appear as discontinuities. In Round 1, although the UV–Vis and electrochemical files contained numerical outputs, the amount of recovered product was insufficient for reliable testing; these outputs were therefore classified as "No valid local result" and excluded from scientific interpretation.**

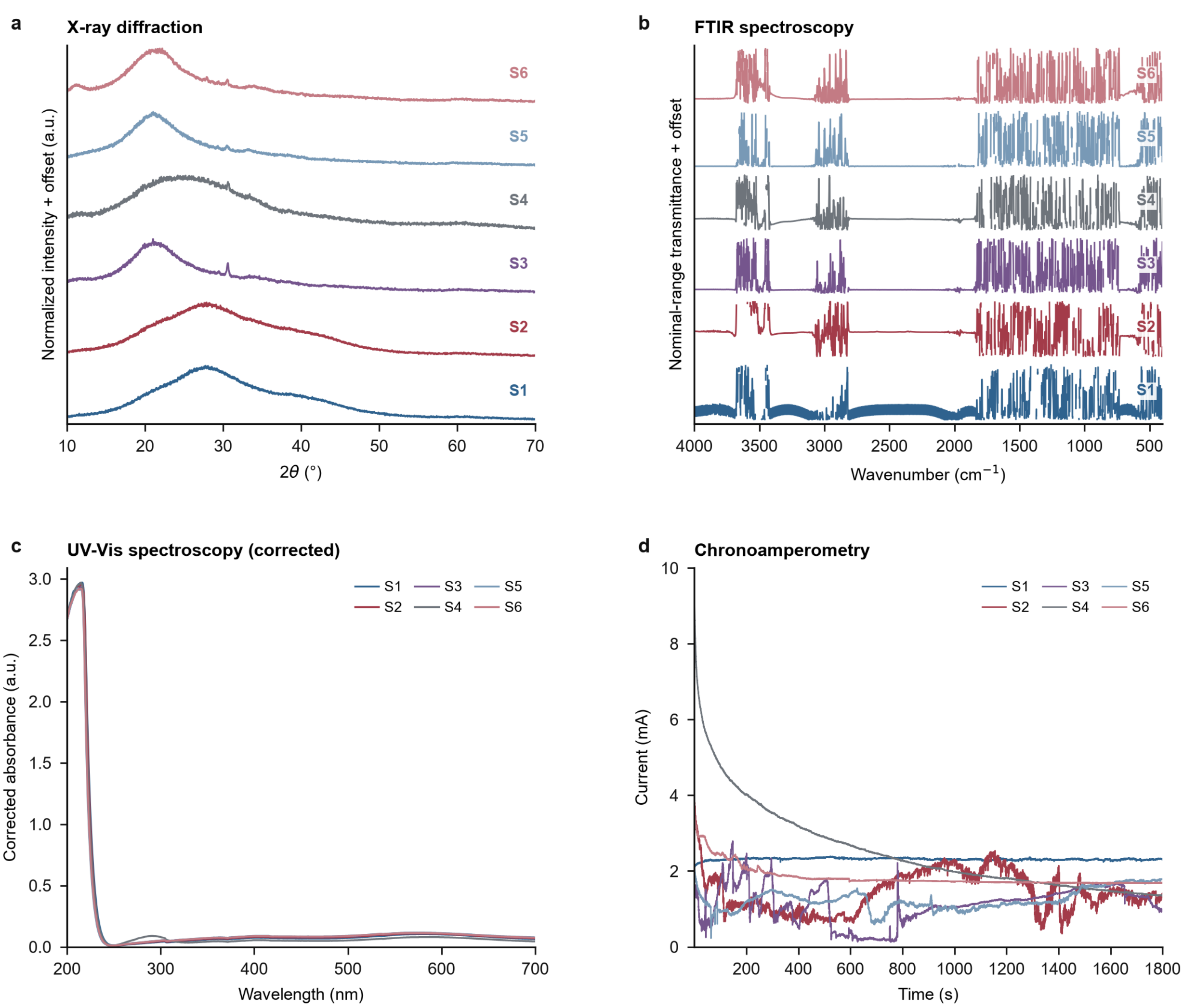

**Fig. S52. Round 2 - Expanded multimodal characterization. (a) X-ray diffraction patterns, (b) FTIR spectra, (c) corrected UV-Vis absorbance spectra and (d) chronoamperometric current for samples S1-S6. The UV-Vis display applies the documented offset correction around 400 nm; this display correction does not establish product-specific observability. Electrochemical source values are multiplied by 1000 from A to mA and shown on the fixed 0-10 mA range.**

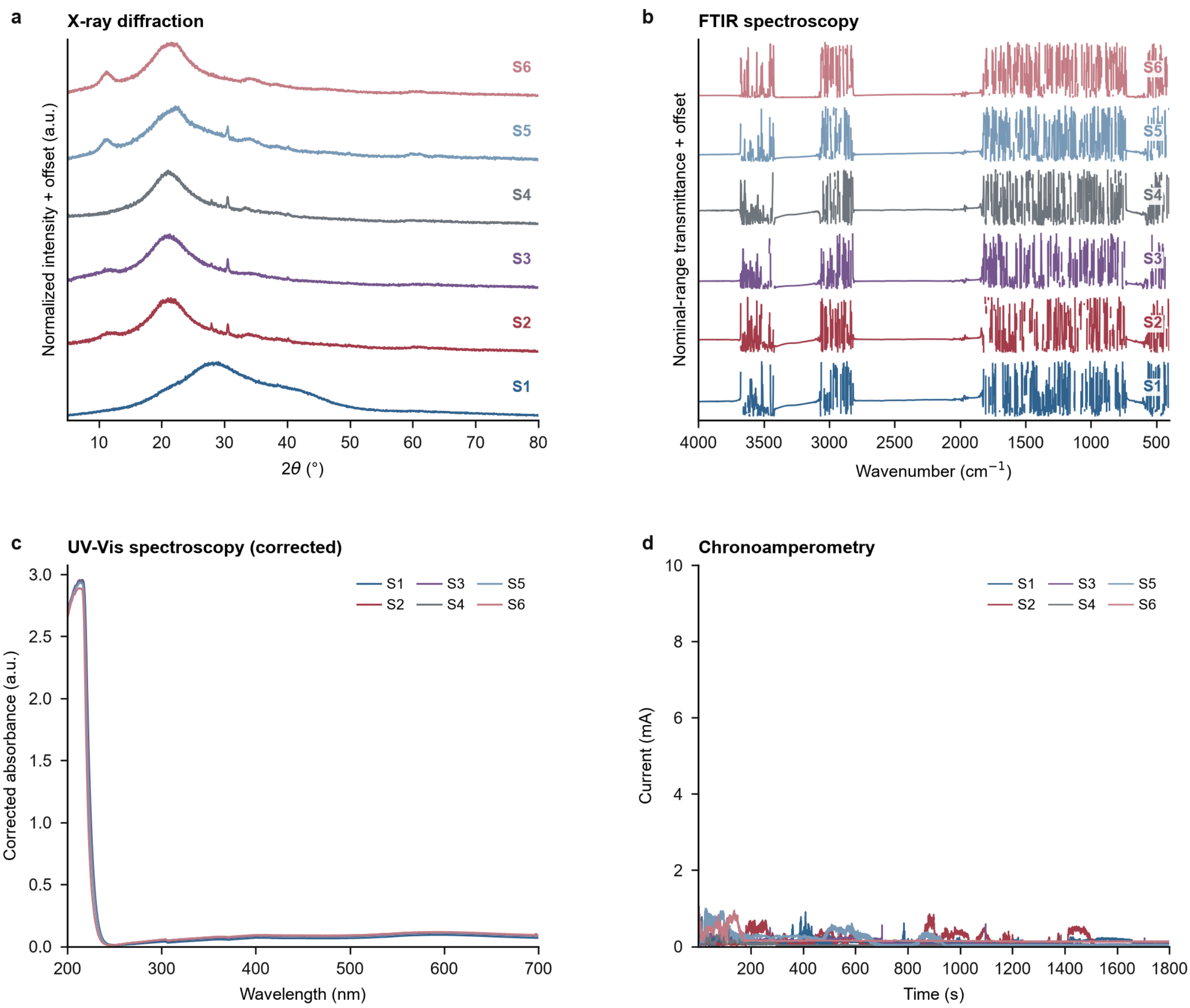

**Fig. S53. Round 3 - Low-carbon composition refinement. (a) X-ray diffraction patterns, (b) FTIR spectra, (c) corrected UV-Vis absorbance spectra and (d) chronoamperometric current for samples S1-S6. The UV-Vis display applies the documented offset correction around 400 nm; this display correction does not establish product-specific observability. Electrochemical source values are multiplied by 1000 from A to mA and shown on the fixed 0-10 mA range.**

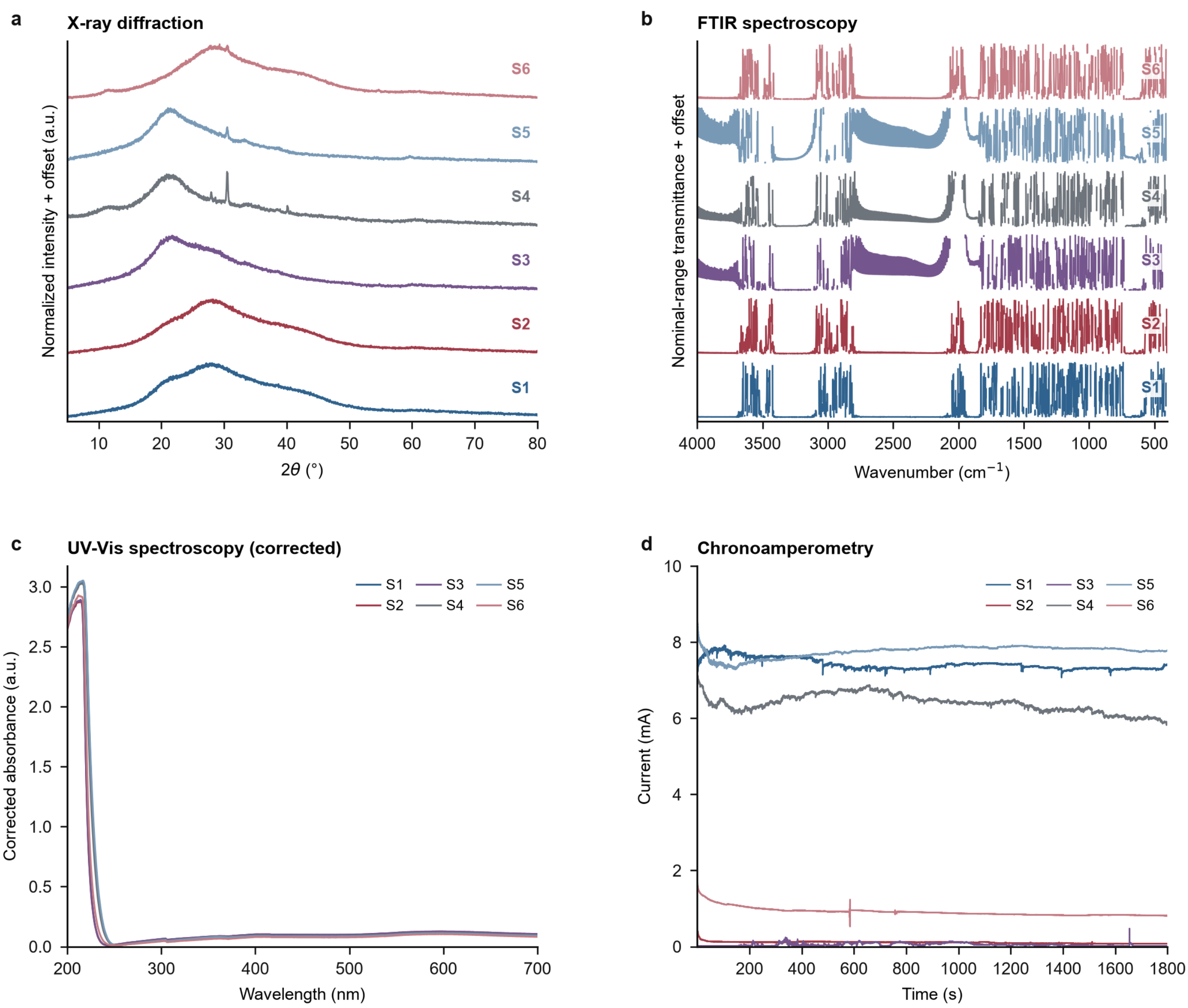

**Fig. S54. Round 4 - Follow-up characterization and control measurements. (a) X-ray diffraction patterns, (b) FTIR spectra, (c) corrected UV-Vis absorbance spectra and (d) chronoamperometric current for samples S1-S6. The UV-Vis display applies the documented offset correction around 400 nm; this display correction does not establish product-specific observability. Electrochemical source values are multiplied by 1000 from A to mA and shown on the fixed 0-10 mA range.**

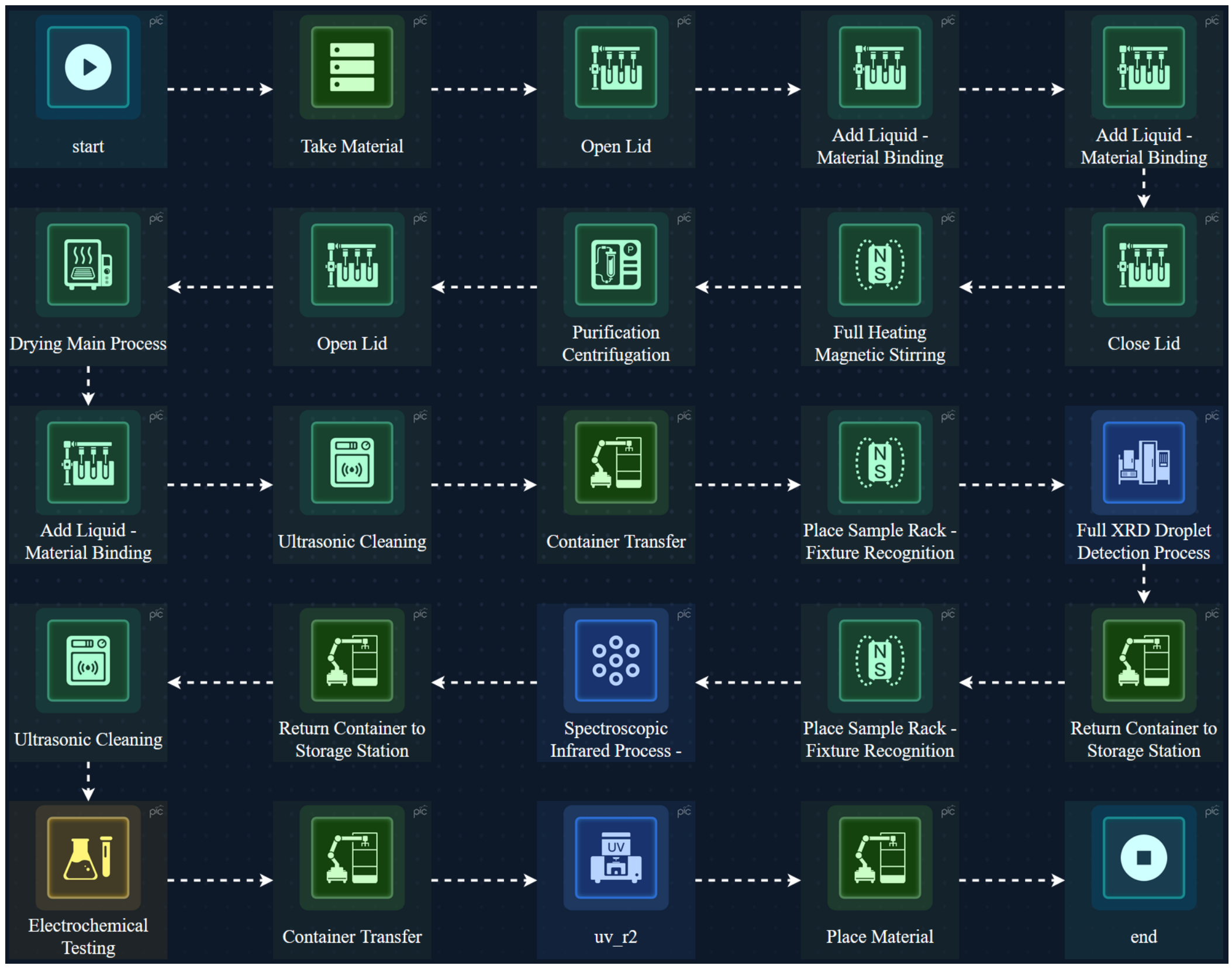

**Fig. S55. Round 2 - Expanded workflow after the documented recovery intervention. The schematic contains 23 experimental operations between Start and End and retains six samples. The sequence covers material handling and liquid addition, closed-vessel heating and magnetic stirring, centrifugation-based purification, drying, ultrasonication and container transfer, followed by XRD, infrared, electrochemical and UV-Vis characterization.**

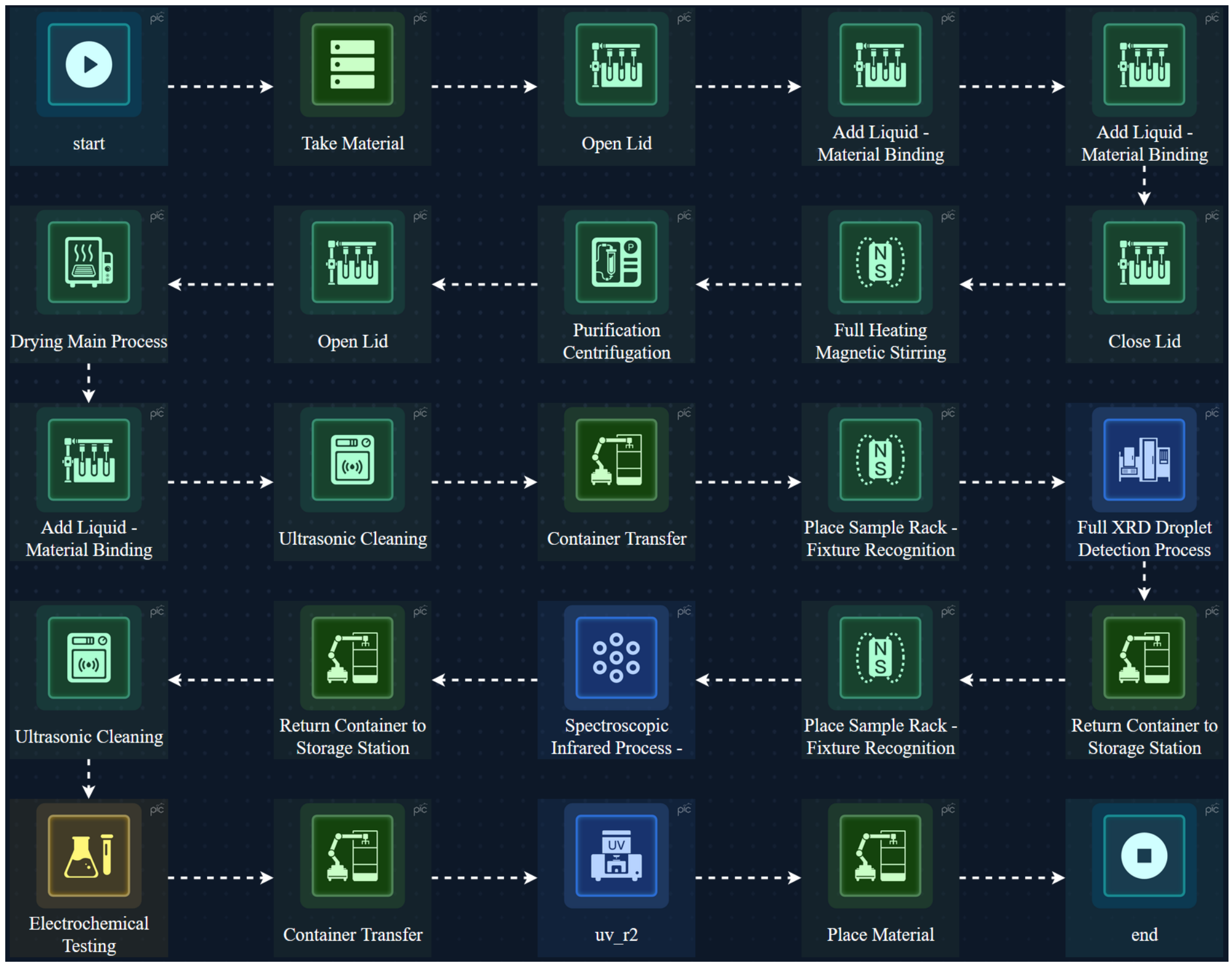

**Fig. S56. Round 3 - Retained expanded workflow. The schematic again contains 23 experimental operations between Start and End for six samples and preserves the overall synthesis, recovery and multimodal characterization structure established in Round 2. No separate documented human scientific judgement was recorded for the Round 2-to-Round 3 transition.**

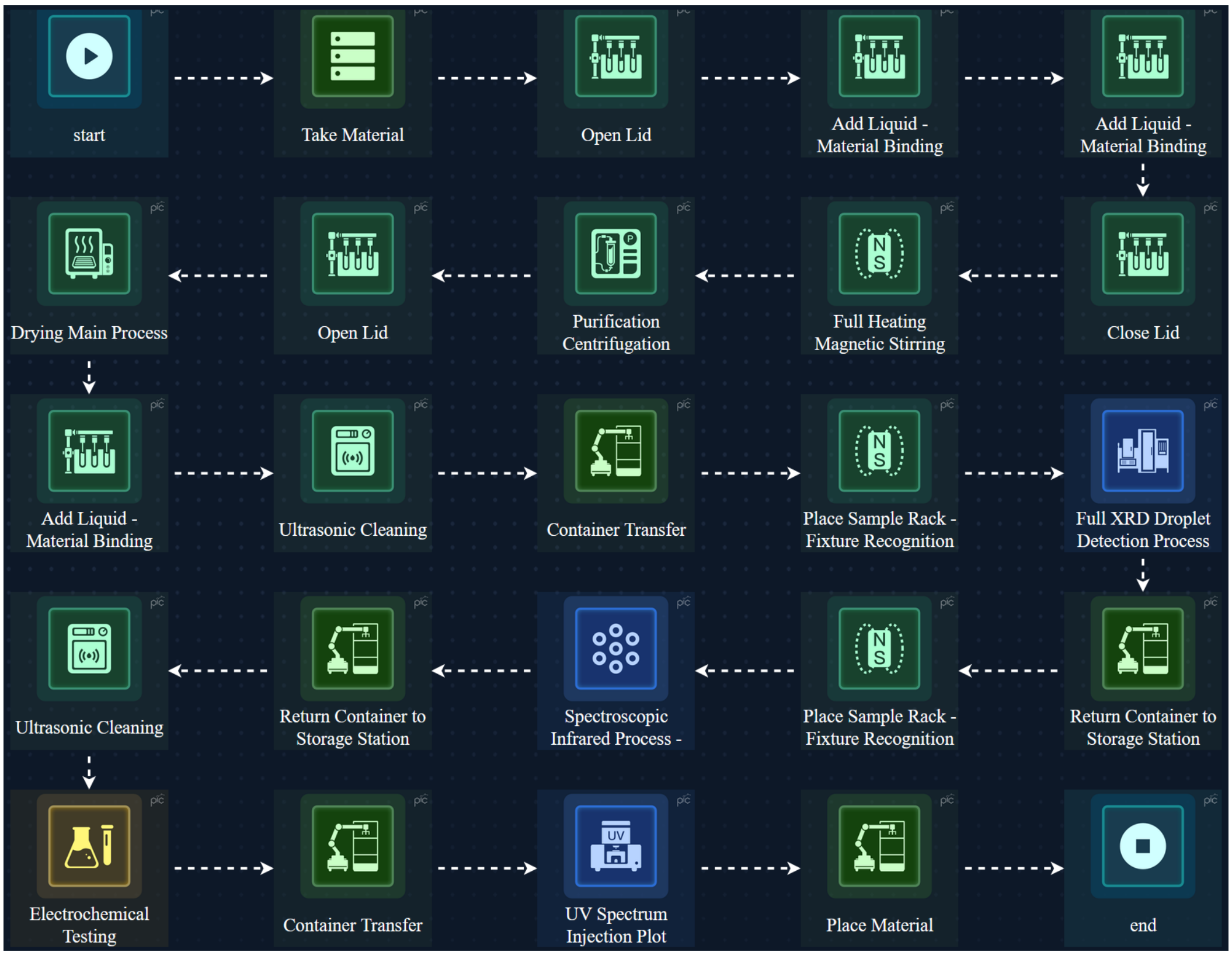

**Fig. S57. Round 4 - Continued expanded workflow. The schematic contains 23 experimental operations between Start and End for six samples and maintains the same overall material-handling, synthesis, purification, recovery and multimodal characterization sequence. No separate documented human scientific judgement was recorded for the Round 3-to-Round 4 transition.**

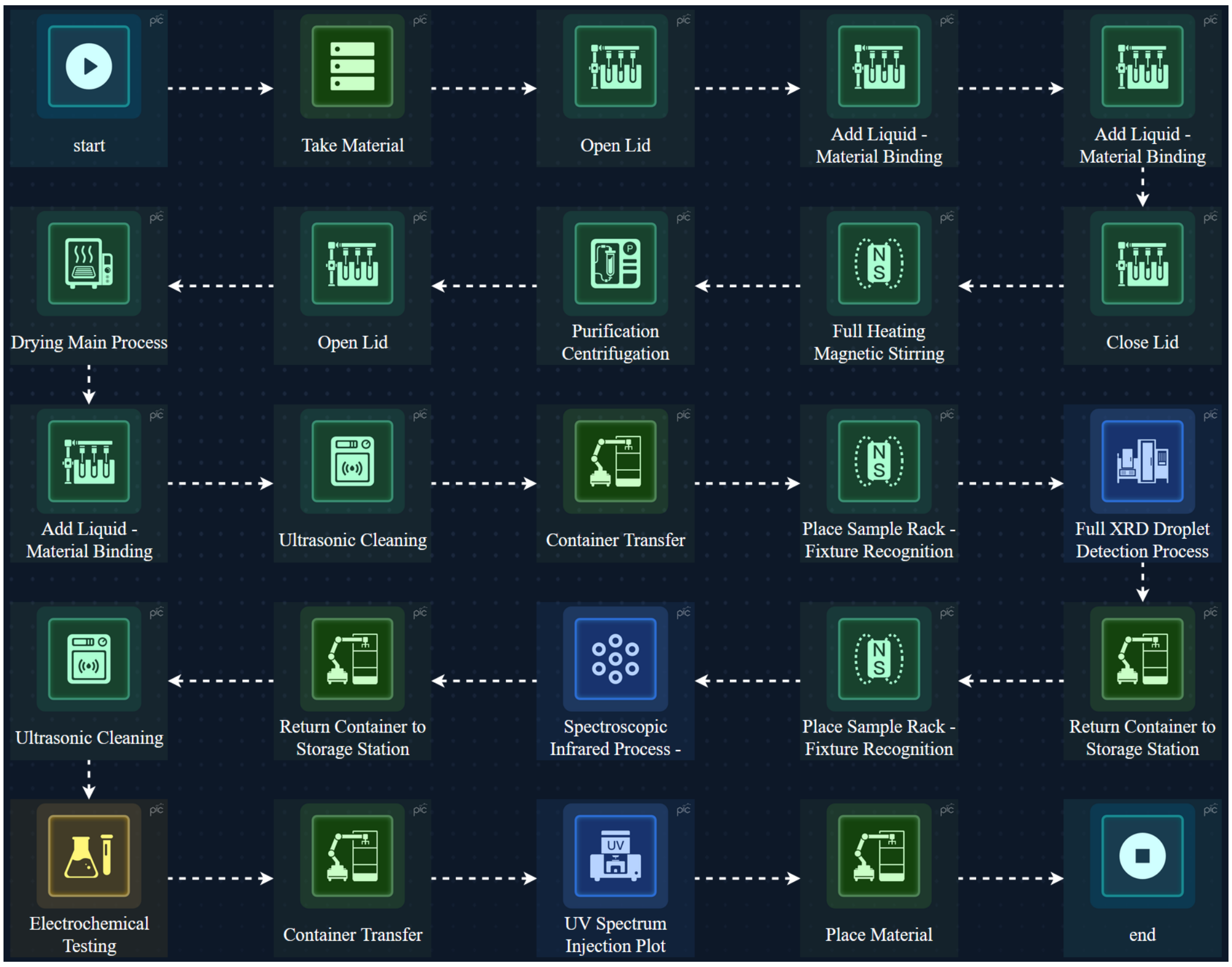

**Fig. S58. Round 5 - Final documented workflow. The schematic contains 23 experimental operations between Start and End for six samples and maintains the expanded synthesis, purification, recovery and multimodal characterization sequence. No separate documented human scientific judgement was recorded for the Round 4-to-Round 5 transition.**

# Data and movie legends

**Data S1.** Benchmark task prompts and task codes (Data_S1.xlsx), including the README, data dictionary and locked task-prompt table.

**Data S2.** Agent-visible workstation catalogue (Data_S2.xlsx), including the README, data dictionary and the 45-workstation catalogue.

**Data S3.** Benchmark and open-ended-task source data (Data_S3.xlsx), including configuration-level matrices, judge prompts and anchors, R1 constraint references, planning-round records, returned-data summaries and provenance.

**Data S4.** Node-level assessment data (Data_S4.xlsx), including pass-rate and pass-count matrices, the assessment matrix, long-form node decisions, Top-8 node rates and decision breakdowns.

**Movie S1.** Overview of the operational robotic laboratory (Supplementary Video_1.mp4; 60 s).

**Movie S2.** Demonstration of a high-throughput synthesis and characterization area (Supplementary Video_2.mp4; 60 s).

**Movie S3.** Demonstration of a catalysis reaction and characterization platform (Supplementary Video_3.mp4; 60 s).

**Movie S4.** Robotic electrochemical testing workflow (Supplementary video_Electrochemical.mp4; 60 s). This clip demonstrates the experimental procedure and is not a real-time recording of a task reported in this Article.

**Movie S5.** Robotic FT-IR measurement workflow (Supplementary Video_FT-IR.mp4; 27.2 s). This clip demonstrates the experimental procedure and is not a real-time recording of a task reported in this Article.

**Movie S6.** Robotic XRD measurement workflow (Supplementary Video_XRD.mp4; 22.2 s). This clip demonstrates the experimental procedure and is not a real-time recording of a task reported in this Article.

**References and Notes**

1. C. Lu, C. Lu, R. T. Lange, Y. Yamada, S. Hu, J. Foerster, D. Ha, J. Clune, Towards end-to-end automation of AI research. *Nature* **651**, 914–919 (2026).
2. J. Gottweis, W.-H. Weng, A. Daryin, T. Tu, P. Sirkovic, A. Myaskovsky, G. Glowaty, F. Weissenberger, A. Orlandi, D. Popovici, A. Palepu, K. Rong, R. Tanno, K. Saab, F. Zhang, J. Blum, A. Carroll, K. Kulkarni, N. Tomašev, D. Zverinski, I. Rendulic, E. Vedadi, F. Hasler, L. Rimanic, M. Boia, I. Budiselic, B. Feinstein, M. Bellaiche, T. Sheffer, J. Freyberg, J. Ratcliff, O. Bertolli, K. Chou, A. Hassidim, B. Gokturk, A. Vahdat, Y. Guan, V. Dhillon, E. D. Vaishnav, B. Lee, T. R. D. Costa, J. R. Penadés, G. Peltz, Y. Matias, J. Manyika, D. Hassabis, Y. Xu, P. Kohli, A. Pawlosky, A. Karthikesalingam, V. Natarajan, Accelerating scientific discovery with Co-Scientist. *Nature* **655**, 487–496 (2026).
3. A. E. Ghareeb, B. Chang, L. Mitchener, A. Yiu, C. J. Szostkiewicz, D. Shved, G. J. Gyimesi, J. M. Laurent, S. M. Wright, M. T. Razzak, A. D. White, S. C. Finnemann, M. M. Hinks, S. G. Rodriques, A multi-agent system for automating scientific discovery. *Nature* **655**, 497–505 (2026).
4. F. Trost, B. Zhang, I. Aring, M. Bauer, L. Glamann, M. Wessolly, K. Johnson, H. Göbel, T. Lerbs, T. Sangenne, P. Herrmann, F. Mairinger, C. Kopp, S. Michels, A. Rasokat, M. Heldwein, S. Wagner, B. Schömig-Markiefka, J. Wolf, S. Hartmann, C. Wickenhauser, A. Bychkov, J. P. Klussmann, A. Quaas, R. Buettner, Y. Tolkach, An agentic framework for autonomous scientific discovery in cancer pathology. *Nat. Med.* **32**, 2254–2266 (2026).
5. J. R. Penadés, J. Gottweis, L. He, J. B. Patkowski, A. Daryin, W.-H. Weng, T. Tu, A. Palepu, A. Myaskovsky, A. Pawlosky, V. Natarajan, A. Karthikesalingam, T. R. D. Costa, AI mirrors experimental science to uncover a mechanism of gene transfer crucial to bacterial evolution. *Cell* **188**, 6654–6665.e2 (2025).
6. B. Burger, P. M. Maffettone, V. V. Gusev, C. M. Aitchison, Y. Bai, X. Wang, X. Li, B. M. Alston, B. Li, R. Clowes, N. Rankin, B. Harris, R. S. Sprick, A. I. Cooper, A mobile robotic chemist. *Nature* **583**, 237–241 (2020).
7. B. A. Koscher, R. B. Canty, M. A. McDonald, K. P. Greenman, C. J. McGill, C. L. Bilodeau, W. Jin, H. Wu, F. H. Vermeire, B. Jin, T. Hart, T. Kulesza, S.-C. Li, T. S. Jaakkola, R. Barzilay, R. Gómez-Bombarelli, W. H. Green, K. F. Jensen, Autonomous, multiproperty-driven molecular discovery: From predictions to measurements and back. *Science* **382**, eadi1407 (2023).
8. F. Strieth-Kalthoff, H. Hao, V. Rathore, J. Derasp, T. Gaudin, N. H. Angello, M. Seifrid, E. Trushina, M. Guy, J. Liu, X. Tang, M. Mamada, W. Wang, T. Tsagaantsooj, C. Lavigne, R. Pollice, T. C. Wu, K. Hotta, L. Bodo, S. Li, M. Haddadnia, A. Wołos, R. Roszak, C. T. Ser, C. Bozal-Ginesta, R. J. Hickman, J. Vestfrid, A. Aguilar-Granda, E. L. Klimareva, R. C. Sigerson, W. Hou, D. Gahler, S. Lach, A. Warzybok, O. Borodin, S. Rohrbach, B. Sanchez-Lengeling, C. Adachi, B. A. Grzybowski, L. Cronin, J. E. Hein, M. D. Burke, A. Aspuru-Guzik, Delocalized, asynchronous, closed-loop discovery of organic laser emitters. *Science* **384**, eadk9227 (2024).
9. N. J. Szymanski, B. Rendy, Y. Fei, R. E. Kumar, T. He, D. Milsted, M. J. McDermott, M. Gallant, E. D. Cubuk, A. Merchant, H. Kim, A. Jain, C. J. Bartel, K. Persson, Y. Zeng, G. Ceder, An autonomous laboratory for the accelerated synthesis of inorganic materials. *Nature* **624**, 86–91 (2023).
10. J. Bai, S. Mosbach, C. J. Taylor, D. Karan, K. F. Lee, S. D. Rihm, J. Akroyd, A. A. Lapkin, M. Kraft, A dynamic knowledge graph approach to distributed self-driving laboratories. *Nat. Commun.* **15**, 462 (2024).
11. R. D. King, J. Rowland, S. G. Oliver, M. Young, W. Aubrey, E. Byrne, M. Liakata, M. Markham, P. Pir, L. N. Soldatova, A. Sparkes, K. E. Whelan, A. Clare, The automation of science. *Science* **324**, 85–89 (2009).
12. A. Slattery, Z. Wen, P. Tenblad, J. Sanjosé-Orduna, D. Pintossi, T. den Hartog, T. Noël, Automated self-optimization, intensification, and scale-up of photocatalysis in flow. *Science* **383**, eadj1817 (2024).
13. S. Steiner, J. Wolf, S. Glatzel, A. Andreou, J. M. Granda, G. Keenan, T. Hinkley, G. Aragon-Camarasa, P. J. Kitson, D. Angelone, L. Cronin, Organic synthesis in a modular robotic system driven by a chemical programming language. *Science* **363**, eaav2211 (2019).
14. K. Huang, S. Zhang, H. Wang, Y. Qu, Y. Lu, R. Li, Y. Roohani, L. Qiu, S. Cao, G. Li, J. Zhang, D. Yin, R. Wierenga, D. Kavi, S. Liu, T. She, S. Marwaha, J. N. Carter, X. Zhou, M. T. Wheeler, J. A. Bernstein, M. Wang, P. He, J. Zhou, M. P. Snyder, L. Cong, A. Regev, J. Leskovec, Autonomous biomedical research with an artificial intelligence agent. *Science* **393**, eadz4351 (2026).
15. D. A. Boiko, R. MacKnight, B. Kline, G. Gomes, Autonomous chemical research with large language models. *Nature* **624**, 570–578 (2023).
16. T. Song, M. Luo, X. Zhang, L. Chen, Y. Huang, J. Cao, Q. Zhu, D. Liu, B. Zhang, G. Zou, G. Zhang, F. Zhang, W. Shang, Y. Fu, J. Jiang, Y. Luo, A multiagent-driven robotic AI chemist enabling autonomous chemical research on demand. *J. Am. Chem. Soc.* **147**, 12534–12545 (2025).
17. L. Huang, C. Zhang, Y. Fu, Y. Jiang, E. He, M.-Q. Qi, M.-H. Du, X.-J. Kong, J. Cheng, L. Cronin, C. Wang, Natural-language-interfaced robotic synthesis for AI-copilot-assisted exploration of inorganic materials. *J. Am. Chem. Soc.* **147**, 23014–23025 (2025).
18. Z. Chen, A. N. Petsch, A. J. Israelski, R. Plumley, L. Shen, C. Wang, C. Peng, Y. Ni, A. Bansil, S. Chowdhury, M. Li, J. B. Thayer, V. Thampy, J. J. Turner, An agentic artificially intelligent X-ray scientist. *Nat. Mach. Intell.* **8**, 1075–1086 (2026).
19. Z. Chen, S. Chen, Y. Ning, Q. Zhang, B. Wang, B. Yu, Y. Li, Z. Liao, C. Wei, Z. Lu, V. Dey, M. Xue, F. N. Baker, B. Burns, D. Adu-Ampratwum, X. Huang, X. Ning, S. Gao, Y. Su, H. Sun, "ScienceAgentBench: Toward Rigorous Assessment of Language Agents for Data-Driven Scientific Discovery" in *The Thirteenth International Conference on Learning Representations* (ICLR, Singapore, 2025).

20. B. P. Majumder, H. Surana, D. Agarwal, B. Dalvi Mishra, A. Meena, A. Prakhar, T. Vora, T. Khot, A. Sabharwal, P. Clark, “DiscoveryBench: Towards Data-Driven Discovery with Large Language Models” in *The Thirteenth International Conference on Learning Representations* (ICLR, Singapore, 2025).
21. P. Scardi, P. L. Antonucci, XRD characterization of highly dispersed metal catalysts on carbon support. *J. Mater. Res.* **8**, 1829–1835 (1993).
22. J. M. O’Reilly, R. A. Mosher, Functional groups in carbon black by FTIR spectroscopy. *Carbon* **21**, 47–51 (1983).
23. S. M. Alia, K. S. Reeves, J. S. Baxter, D. A. Cullen, The impact of ink and spray variables on catalyst layer properties, electrolyzer performance, and electrolyzer durability. *J. Electrochem. Soc.* **167**, 144512 (2020).
24. B. K. Boggs, R. L. King, G. G. Botte, Urea electrolysis: direct hydrogen production from urine. *Chem. Commun.* **2009**, 4859–4861 (2009).
25. A. Mather, D. Roland, The automated thiosemicarbazide-diacetyl monoxime method for plasma urea. *Clin. Chem.* **15**, 393–396 (1969).
26. C. E. Bower, T. Holm-Hansen, A salicylate–hypochlorite method for determining ammonia in seawater. *Can. J. Fish. Aquat. Sci.* **37**, 794–798 (1980).
27. K. M. Miranda, M. G. Espey, D. A. Wink, A rapid, simple spectrophotometric method for simultaneous detection of nitrate and nitrite. *Nitric Oxide* **5**, 62–71 (2001).
28. J. Pineau, P. Vincent-Lamarre, K. Sinha, V. Larivière, A. Beygelzimer, F. d’Alché-Buc, E. Fox, H. Larochelle, Improving reproducibility in machine learning research (a report from the NeurIPS 2019 reproducibility program). *J. Mach. Learn. Res.* **22**, 1–20 (2021).
29. J. Dodge, S. Gururangan, D. Card, R. Schwartz, N. A. Smith, “Show Your Work: Improved Reporting of Experimental Results” in *Proceedings of the 2019 Conference on Empirical Methods in Natural Language Processing and the 9th International Joint Conference on Natural Language Processing (EMNLP-IJCNLP)* (Association for Computational Linguistics, Hong Kong, China, 2019), pp. 2185–2194.
30. Y. Liu, D. Iter, Y. Xu, S. Wang, R. Xu, C. Zhu, “G-Eval: NLG Evaluation using GPT-4 with Better Human Alignment” in *Proceedings of the 2023 Conference on Empirical Methods in Natural Language Processing* (Association for Computational Linguistics, Singapore, 2023), pp. 2511–2522.
31. L. Zheng, W.-L. Chiang, Y. Sheng, S. Zhuang, Z. Wu, Y. Zhuang, Z. Lin, Z. Li, D. Li, E. P. Xing, H. Zhang, J. E. Gonzalez, I. Stoica, “Judging LLM-as-a-Judge with MT-Bench and Chatbot Arena” in *Advances in Neural Information Processing Systems 36* (Neural Information Processing Systems Foundation, 2023), pp. 46595–46623.
32. A. Kumar, N. Poungpeth, D. Yang, E. Farrell, B. L. Lambert, M. Groh, When large language models are reliable for judging empathic communication. *Nat. Mach. Intell.* **8**, 173–185 (2026).
33. P. Wang, L. Li, L. Chen, Z. Cai, D. Zhu, B. Lin, Y. Cao, L. Kong, Q. Liu, T. Liu, Z. Sui, “Large Language Models are not Fair Evaluators” in *Proceedings of the 62nd Annual Meeting of the Association for Computational Linguistics (Volume 1: Long Papers)* (Association for Computational Linguistics, Bangkok, Thailand, 2024), pp. 9440–9450.
34. J. M. Bland, D. G. Altman, Statistical methods for assessing agreement between two methods of clinical measurement. *Lancet* **327**, 307–310 (1986).
35. B. Efron, Bootstrap methods: another look at the jackknife. *Ann. Stat.* **7**, 1–26 (1979).
36. C. A. Field, A. H. Welsh, Bootstrapping clustered data. *J. R. Stat. Soc. B* **69**, 369–390 (2007).